\pdfoutput=1

\documentclass[10pt,twocolumn,letterpaper]{article}

\usepackage{cvpr}
\usepackage{times}
\usepackage{epsfig}
\usepackage{graphicx}
\usepackage{amsmath}
\usepackage{amssymb}
\usepackage{ulem}
\usepackage{mathrsfs}
\usepackage{color}
\usepackage{subcaption}
\usepackage{footmisc}
\usepackage[]{algorithm2e}
\usepackage{makecell}
\usepackage{multirow}
\usepackage{fancyhdr}
\usepackage{appendix}
\usepackage{textcomp}

\usepackage[breaklinks=true,bookmarks=false]{hyperref}

\cvprfinalcopy 

\def\cvprPaperID{4754} 

\ifcvprfinal\fi
\def\x{\boldsymbol{x}}
\def\r{\boldsymbol{r}}
\def\w{\boldsymbol{w}}

\DeclareMathOperator*{\argmax}{arg\,max}
\DeclareMathOperator*{\sign}{sign}

\begin{document}
\title{SparseFool: a few pixels make a big difference}

\author{Apostolos Modas,~Seyed-Mohsen Moosavi-Dezfooli,~Pascal Frossard\\
\'Ecole Polytechnique F\'ed\'erale de Lausanne\\
{\tt\small \{apostolos.modas,~seyed.moosavi,~pascal.frossard\}@epfl.ch}
}

\maketitle

\thispagestyle{fancy}
\fancyhead[L]{Accepted at CVPR 2019}
\renewcommand{\headrulewidth}{0.5pt}


\begin{abstract}
Deep Neural Networks have achieved extraordinary results on image classification tasks, but have been shown to be vulnerable to attacks with carefully crafted perturbations of the input data. Although most attacks usually change values of many image's pixels, it has been shown that deep networks are also vulnerable to sparse alterations of the input. However, no computationally efficient method has been proposed to compute sparse perturbations. In this paper, we exploit the low mean curvature of the decision boundary, and propose SparseFool, a geometry inspired sparse attack that controls the sparsity of the perturbations. Extensive evaluations show that our approach computes sparse perturbations very fast, and scales efficiently to high dimensional data. We further analyze the transferability and the visual effects of the perturbations, and show the existence of shared semantic information across the images and the networks. Finally, we show that adversarial training can only slightly improve the robustness against sparse additive perturbations computed with SparseFool.
\footnote{The code of SparseFool is available on \url{https://github.com/LTS4/SparseFool}, and Foolbox~\protect\cite{rauber2017foolbox} \url{https://github.com/bethgelab/foolbox}.}
\end{abstract}

\section{Introduction}
Deep neural networks (DNNs) are powerful learning models that achieve state-of-the-art performance in many different classification tasks~\cite{coco,pmlr-v48-zhangc16,youtube,openimages,imagenet_2009}, but have been shown to be vulnerable to very small, and often imperceptible, adversarial manipulations of their input data ~\cite{szegedy_Intr}. Interestingly, the existence of such adversarial perturbations is not only limited to additive perturbations~\cite{manifool, BMVC2015_106, xiao2018spatially} or classification tasks~\cite{cisse}, but can be found in many other applications~\cite{tabacof2016, carlini_wagner_2018, lin_hong_liao_shih_liu_sun_2017, metzen_kumar_brox_fischer_2017, papernot_mcdaniel_swami_harang_2016, rozsa_gunther_rudd_boult_2016, rozsa}.

For image classification tasks, the most common type of adversarial perturbations are the $\ell_p$ minimization ones, since they are easier to analyze and optimize. Formally, for a given classifier and an image $\x\in\mathbb{R}^n$, we define as adversarial perturbation the minimal perturbation $\r$ that changes the classifier's estimated label $k(\x)$:
\begin{equation}
\label{eq:pert}
\underset{\r}{\text{min}}\|\r\|_p\enskip\text{s.t.}\enskip k(\x+\r)\neq k(\x),
\end{equation}

\begin{figure}[t]
\centering
    \begin{subfigure}[b]{0.33\columnwidth}
        \includegraphics[width=\linewidth]{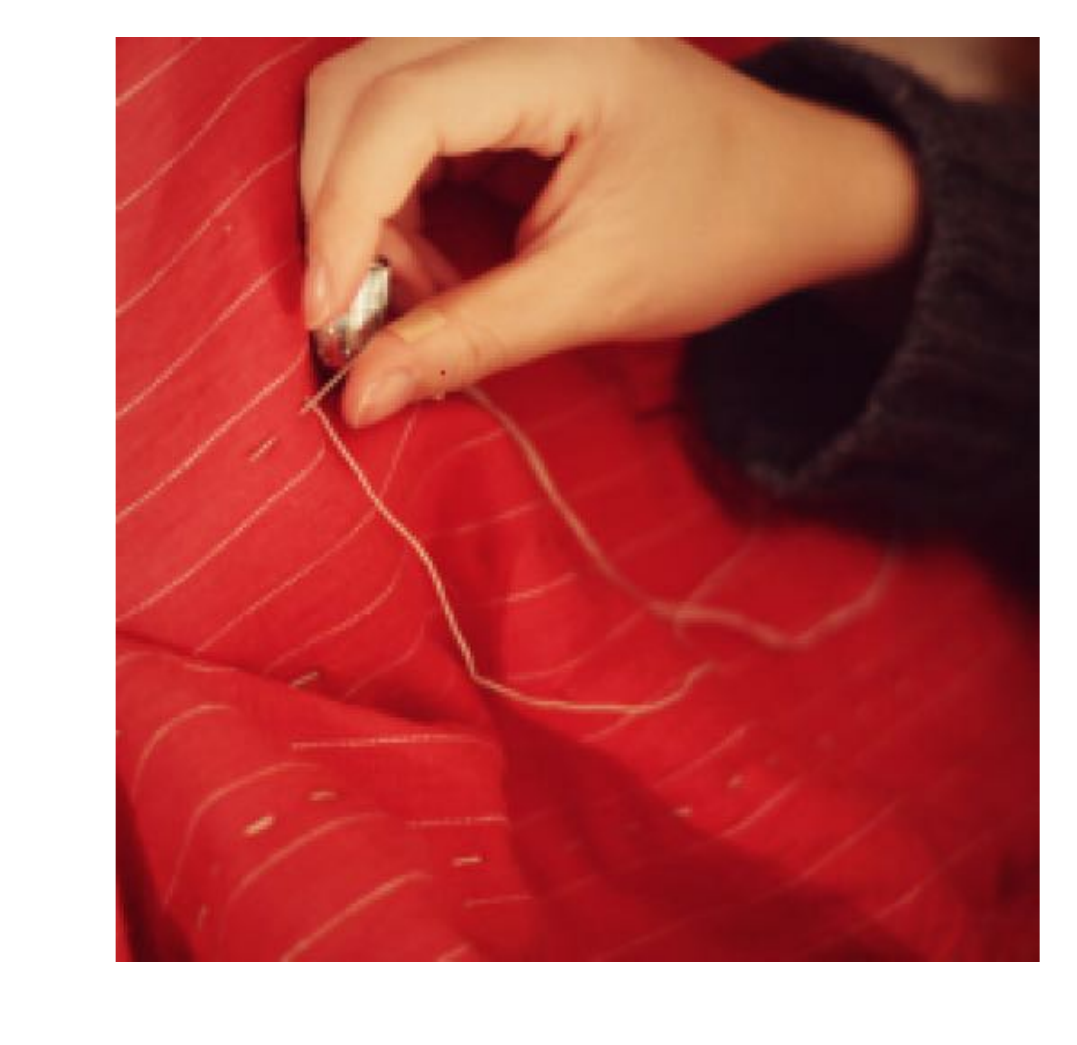}
        \vspace*{-7mm}
        \caption*{cockroach}
    \end{subfigure}\!
    \begin{subfigure}[b]{0.33\columnwidth}
        \includegraphics[width=\linewidth]{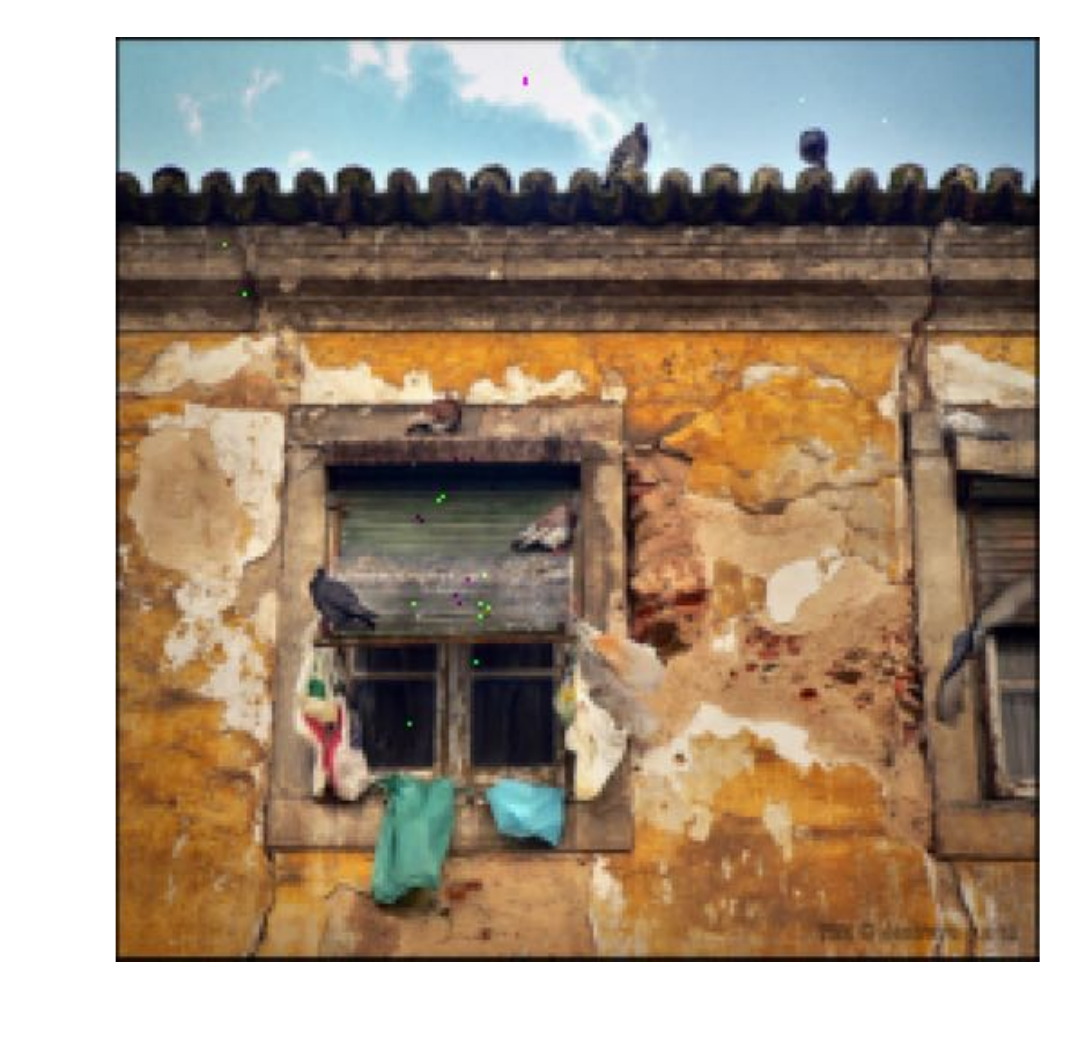}
        \vspace*{-7mm}
        \caption*{palace}
    \end{subfigure}\!
    \begin{subfigure}[b]{0.33\columnwidth}
        \includegraphics[width=\linewidth]{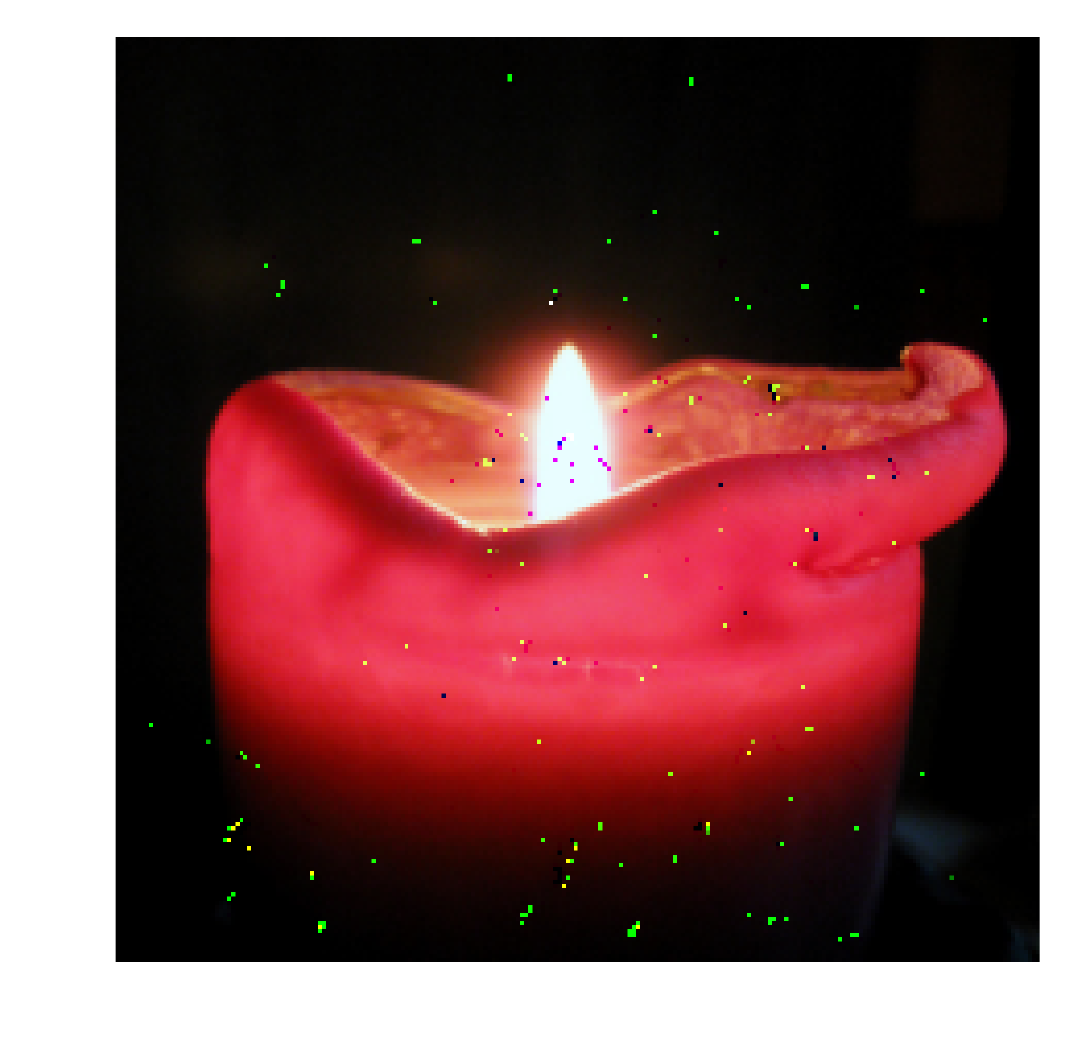}
        \vspace*{-7mm}
        \caption*{bathtub}
    \end{subfigure}
    
    \begin{subfigure}[b]{0.33\columnwidth}
        \includegraphics[width=\linewidth]{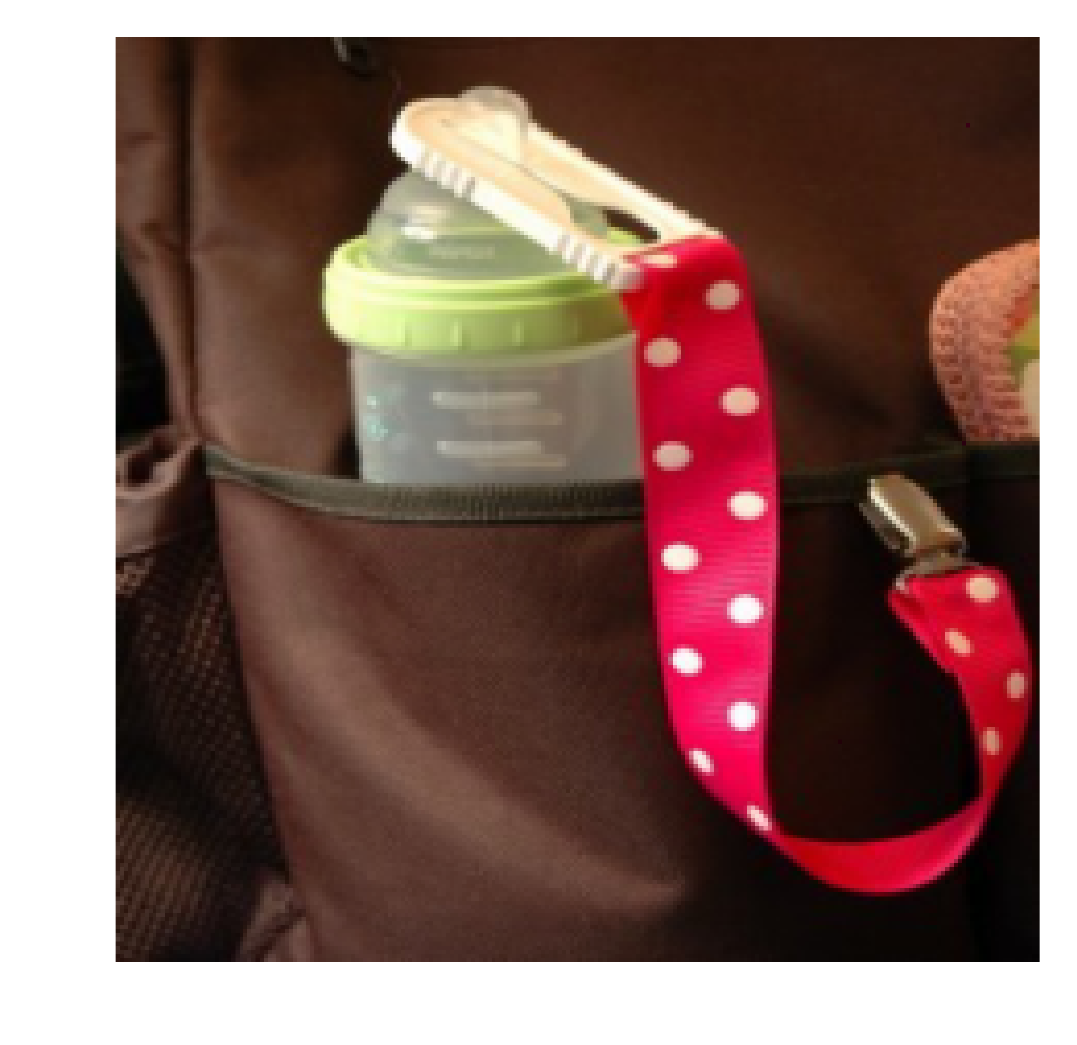}
        \vspace*{-7mm}
        \caption*{sandal}
    \end{subfigure}\!
    \begin{subfigure}[b]{0.33\columnwidth}
        \includegraphics[width=\linewidth]{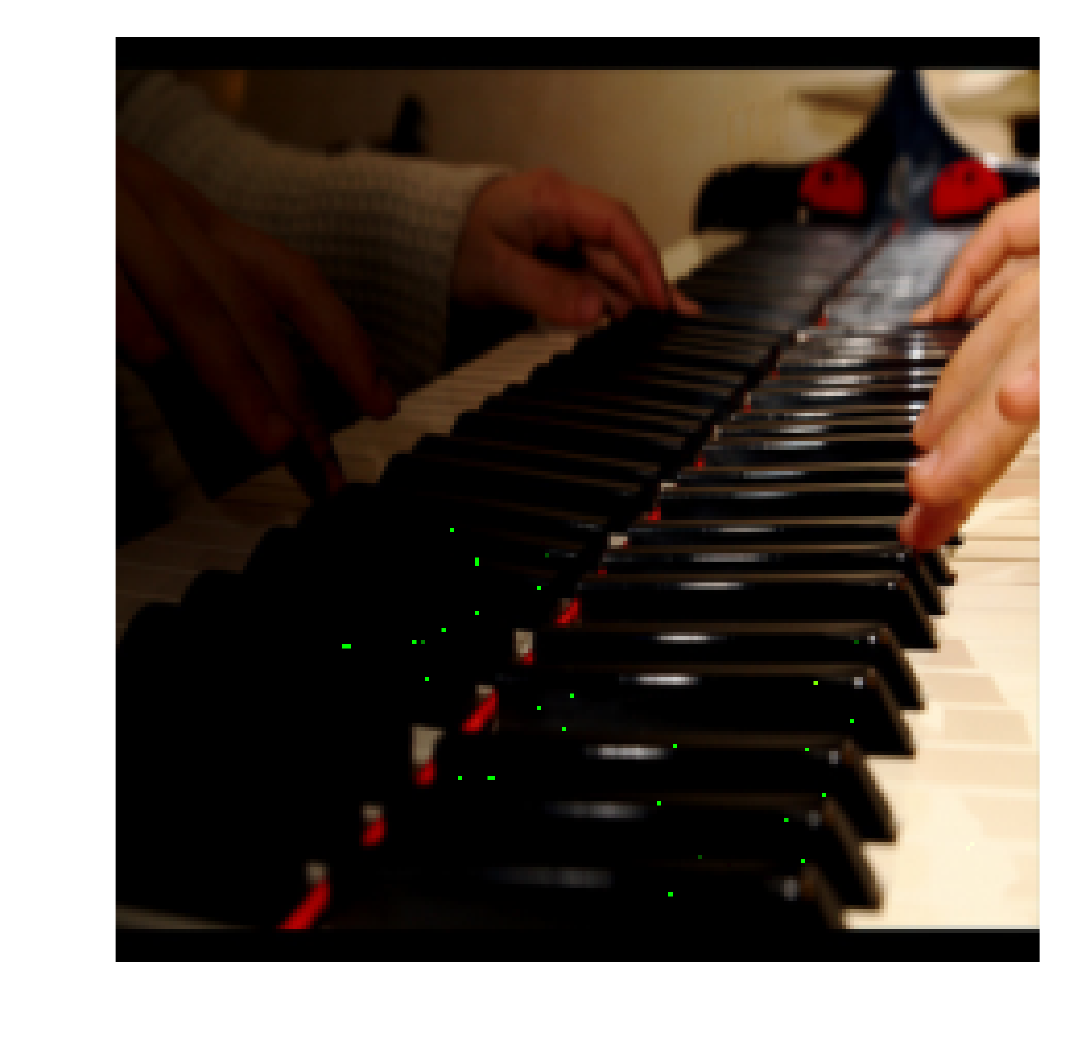}
        \vspace*{-7mm}
        \caption*{wine bottle}
    \end{subfigure}\!
    \begin{subfigure}[b]{0.33\columnwidth}
        \includegraphics[width=\linewidth]{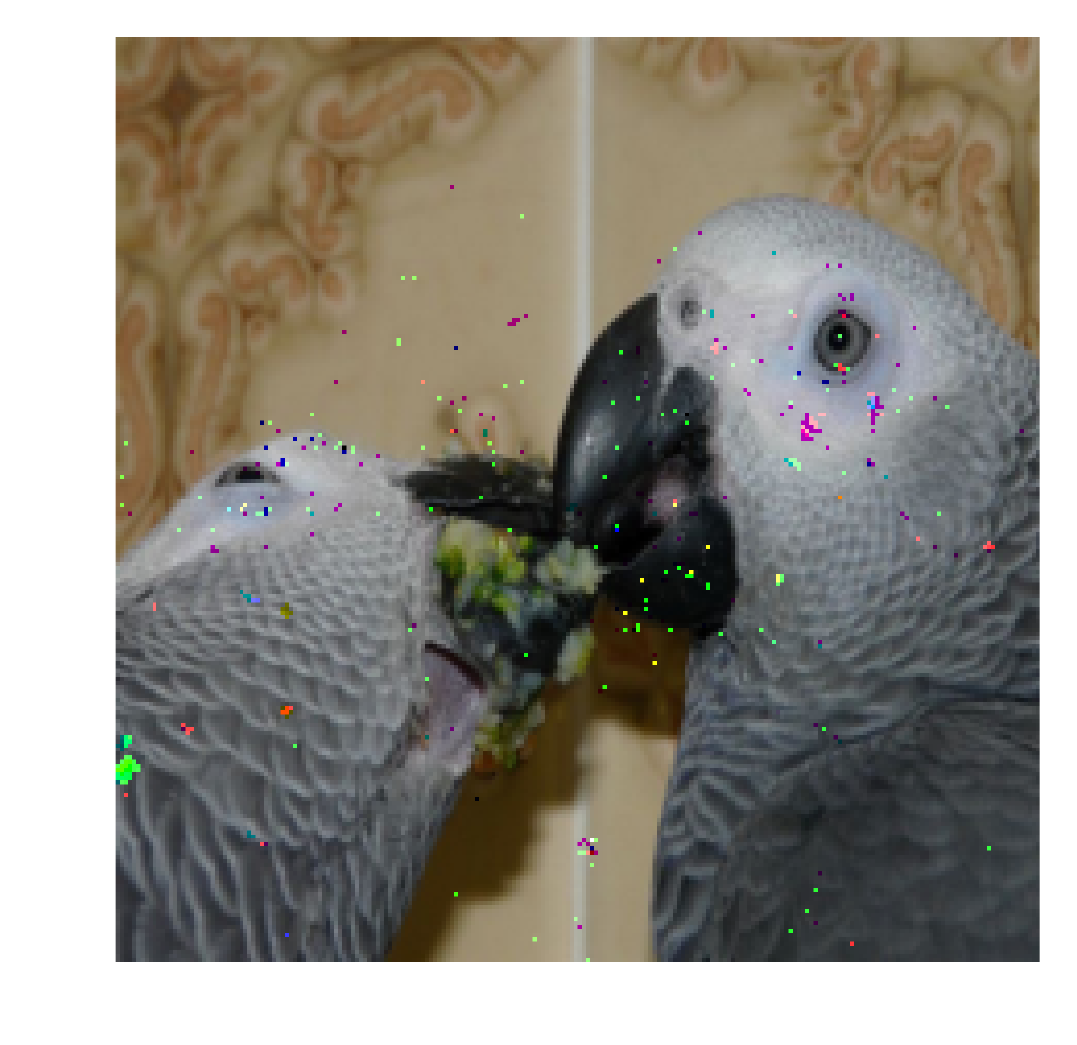}
        \vspace*{-7mm}
        \caption*{bubble}
    \end{subfigure}
    \caption{Adversarial examples for ImageNet, as computed by SparseFool on a ResNet-101 architecture. Each column corresponds to different level of perturbed pixels. The fooling labels are shown below the images.}
    \label{fig:pert_imagenet}
\end{figure}

Several methods (attacks) have been proposed for computing $\ell_2$ and $\ell_\infty$ adversarial perturbations~\cite{szegedy_Intr, goodfellow_shlens_szegedy_2016, bim, moosavi-dezfooli_fawzi_frossard_2016, carlini_wagner_2017, universal}. However, understanding the vulnerabilities of deep neural networks in other perturbation regimes remains important. In particular, it has been shown~\cite{papernot_mcdaniel_jha_fredrikson_celik_swami_2016, su_vargas_kouichi_2018, narodytska_kasiviswanathan_2017, bibi2018analytic, hein2017formal} that DNNs can misclassify an image, when only a small fraction of the input is altered (sparse perturbations). In practice, sparse perturbations could correspond to some raindrops that reflect the sun on a ``STOP" sign, but that are sufficient to fool an autonomous vehicle; or a crop-field with some sparse colorful flowers that force a UAV to spray pesticide on non affected areas. Understanding the vulnerability of deep networks to such a simple perturbation regime can further help to design methods to improve the robustness of deep classifiers.

Some prior work on sparse perturbations have been developed recently. The authors in~\cite{papernot_mcdaniel_jha_fredrikson_celik_swami_2016} proposed the JSMA method, which perturbs the pixels based on their saliency score. Furthermore, the authors in~\cite{su_vargas_kouichi_2018} exploit Evolutionary Algorithms (EAs) to achieve extremely sparse perturbations, while the authors in~\cite{narodytska_kasiviswanathan_2017} finally proposed a black-box attack that computes sparse adversarial perturbations using a greedy local search algorithm.

However, solving the optimization problem in Eq.~\eqref{eq:pert} in an $\ell_0$ sense is NP-hard, and current algorithms are all characterized by high complexity. The resulting perturbations usually consist of high magnitude noise, concentrated over a small number of pixels. This makes them quite perceptible and in many cases, the perturbed pixels might even exceed the dynamic range of the image.

Therefore, we propose in this paper an efficient and principled way to compute sparse perturbations, while at the same time ensuring the validity of the perturbed pixels.\smallskip

\noindent Our main contributions are the following:
\begin{itemize}
    \item We propose SparseFool, a geometry inspired sparse attack that exploits the boundaries' low mean curvature to compute adversarial perturbations efficiently.
    \item We show through extensive evaluations that (a) our method computes sparse perturbations much faster than the existing methods, and (b) it can scale efficiently to high dimensional data.
    \item We further propose a method to control the perceptibility of the resulted perturbations, while retaining the levels of sparsity and complexity.
    \item We analyze the visual features affected by our attack, and show the existence of some shared semantic information across different images and networks.
    \item We finally show that adversarial training using slightly lowers the vulnerability against sparse perturbations, but not enough to lead to more robust classifiers yet.
\end{itemize}
The rest of the paper is organized as follows: in Section~\ref{sec:problem}, we describe the challenges and the problems for computing sparse adversarial perturbations. In Section~\ref{sec:sparse}, we provide an efficient method for computing sparse adversarial perturbations, by linearizing and solving the initial optimization problem. Finally, the evaluation and analysis of the computed sparse perturbations is provided in Section~\ref{sec:results}.

\section{Problem description}
\label{sec:problem}
\subsection{Finding sparse perturbations}
Most of the existing adversarial attack algorithms solve the optimization problem of Eq.~\eqref{eq:pert} for $p=2$ or $\infty$, resulting in dense but imperceptible perturbations. For the case of sparse perturbations, the goal is to minimize the number of perturbed pixels required to fool the network, which corresponds to minimizing $\|\r\|_0$ in Eq.~\eqref{eq:pert}. Unfortunately, this leads to NP-hard problems, for which reaching a global minimum cannot be guaranteed in general~\cite{blumensath_davies_2008, nikolova_2013, patrascu_necoara_2015}. There exist different methods~\cite{nagahara_quevedo_ostergaard_2014, patrascu_necoara_2015} to avoid the computational burden of this problem, with the $\ell_1$ relaxation being the most common; the minimization of $\|\r\|_0$ under linear constraints can be approximated by solving the corresponding convex $\ell_1$ problem~\cite{candes_tao_2005, donoho_2006, natarajan_1995}\footnote{Under some conditions, the solution of such approximation is indeed optimal~\cite{candes_rudelson_tao_vershynin_2005, donoho_elad_2003, gribonval_nielsen_2003}.}. Thus, we are looking for an efficient way to exploit such a relaxation to solve the optimization problem in Eq.~\eqref{eq:pert}.

DeepFool~\cite{moosavi-dezfooli_fawzi_frossard_2016} is an algorithm that exploits such a relaxation, by adopting an iterative procedure that includes a linearization of the classifier at each iteration, in order to estimate the minimal adversarial perturbation $\r$. Specifically, assuming $f$ is a classifier, at each iteration $i$, $f$ is linearized around the current point $\x^{(i)}$, the minimal perturbation $\r^{(i)}$ (in an $\ell_2$ sense) is computed as the projection of $\x^{(i)}$ onto the linearized hyperplane, and the next iterate $\x^{(i+1)}$ is updated. One can use such a linearization procedure to solve Eq.~(\ref{eq:pert})
for $p=1$, so as to obtain an approximation to the $\ell_0$ solution. Thus, by generalizing the projection to $\ell_p$ norms $(p\in[1, \infty))$ and setting $p=1$, $\ell_1$-DeepFool provides an efficient way for computing sparse adversarial perturbations using the $\ell_1$ projection.

\subsection{Validity of perturbations}
\label{subsec:validity}
Although the $\ell_1$-DeepFool efficiently computes sparse perturbations, it does not explicitly respect the constraint on the validity of the adversarial image values. When computing adversarial perturbations, it is very important to ensure that the pixel values of the adversarial image $\x+\r$ lie within the valid range of color images (\eg, $[0,255]$). For $\ell_2$ and $\ell_\infty$ perturbations, almost every pixel of the image is distorted with noise of small magnitude, so that most common algorithms usually ignore such constraints~\cite{moosavi-dezfooli_fawzi_frossard_2016, goodfellow_shlens_szegedy_2016}. In such cases, it is unlikely that many pixels will be out of their valid range; and even then, clipping the invalid values after the computation of such adversarial images have a minor impact.

This is unfortunately not the case for sparse perturbations however; solving the $\ell_1$ optimization problem results in a few distorted pixels of high magnitude noise, and clipping the values after computing the adversarial image can have a significant impact on the success of the attack. In other words, as the perturbation becomes sparser, the contribution of each pixel is typically much stronger compared to $\ell_2$ or $\ell_\infty$ perturbations.

\begin{figure}[t]
\begin{center}
\includegraphics[width=0.7\linewidth, page=1]{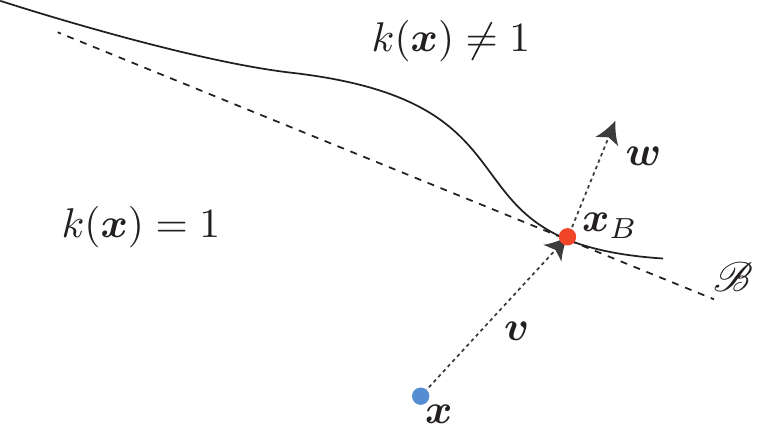}
\end{center}
   \caption{The approximated decision boundary $\mathscr{B}$ in the vicinity of the datapoint $\x$ that belongs to class $1$. $\mathscr{B}$ can be seen as a one-vs-all linear classifier for class $1$.}
\label{fig:flat}
\end{figure}

We demonstrate the effect of such clipping operation on the quality of adversarial perturbations generated by $\ell_1$-DeepFool. For example, with perturbations computed for a VGG-16~\cite{simonyan_zisserman_2015} network trained on ImageNet~\cite{imagenet_2009}, we observed that $\ell_1$-DeepFool achieves almost $100\%$ of fooling rate by perturbing only $0.037\%$ of the pixels on average. However, clipping the pixel values of adversarial images to $[0,255]$ results in a fooling rate of merely $13\%$. Furthermore, incorporating the clipping operator inside the iterative procedure of the algorithm does not improve the results. In other words, $\ell_1$-DeepFool fails to properly compute sparse perturbations. This underlies the need for an improved attack algorithm that natively takes into account the validity of generated adversarial images, as proposed in the next sections.

\subsection{Problem formulation}
\label{subsec:2.3}
Based on the above discussion, sparse adversarial perturbations are obtained by solving the following general form optimization problem:
\begin{equation}
\label{eq:l1_optim}
\begin{aligned}
& \underset{\r}{\text{minimize}}
& & \|\r\|_1 \\
& \text{subject to}
& & k(\x+\r)\neq k(\x)\\
& & & \boldsymbol{l}\preccurlyeq \x+\r\preccurlyeq \boldsymbol{u},
\end{aligned}
\end{equation}
where $\boldsymbol{l}, \boldsymbol{u}\in\mathbb{R}^n$ denote the lower and upper bounds of the values of $\x+\r$, such that $l_i \leq x_i + r_i \leq u_i, \enskip i=1\dots n$.

To find an efficient relaxation to problem~\eqref{eq:l1_optim}, we focus on the geometric characteristics of the decision boundary, and specifically on its curvature. It has been shown~\cite{fawzi_moosavi-dezfooli_frossard_2017, fawzi_moosavi-dezfooli_frossard_soatto_2018, jetley_lord_torr_2018} that the decision boundaries of state-of-the-art deep networks have a quite low mean curvature in the neighborhood of data samples. In other words, for a datapoint $\x$ and its corresponding minimal $\ell_2$ adversarial perturbation $\boldsymbol{v}$, the decision boundary at the vicinity of $\x$ can be locally well approximated by a hyperplane passing through the datapoint $\x_B=\x+\boldsymbol{v}$, and a normal vector $\w$~(see Fig.~\ref{fig:flat}).

Hence, we exploit this property and linearize the optimization problem~\eqref{eq:l1_optim}, so that sparse adversarial perturbations can be computed by solving the following box-constrained optimization problem:
\begin{equation}
\label{eq:l1_linear}
\begin{aligned}
& \underset{\r}{\text{minimize}}
& & \|\r\|_1 \\
& \text{subject to}
& & \w^T(\x + \r)- \w^T\x_{B}=0\\
& & & \boldsymbol{l}\preccurlyeq \x+\r\preccurlyeq \boldsymbol{u}.
\end{aligned}
\end{equation}

In the following section, we provide a method for solving the optimization problem~\eqref{eq:l1_linear}, and introduce SparseFool, a fast yet efficient algorithm for computing sparse adversarial perturbations, which linearizes the constraints by approximating the decision boundary as an affine hyperplane.

\AlgoDontDisplayBlockMarkers
\RestyleAlgo{ruled}
\SetAlgoNoLine
\LinesNumbered
\begin{algorithm}[t]
 	\KwIn{image $\x$, normal $\w$, boundary point $\x_B$, projection operator $Q$.}
 	\KwOut{perturbed point $\x^{(i)}$}
 	\BlankLine
	Initialize: $\x^{(0)}\leftarrow\x$,\enskip$i\leftarrow 0$,\enskip$S=\{\}$\\
	
	\smallskip
	\While{$\w^T(\x^{(i)}-\x_B)\neq0$}
	{
	    $\r\leftarrow\boldsymbol{0}$
	    
	    \smallskip
	    $d\leftarrow\underset{j\notin S}{\argmax}|w_j|$
	    
	    \smallskip
	    $r_d\leftarrow\dfrac{|\w^T(\x^{(i)}-\x_B)|}{|w_d|}\cdot\sign (w_d)$
	    
	    \smallskip
	    $\x^{(i+1)}\leftarrow Q(\x^{(i)}+\r)$
	    
		\smallskip
	    $S \leftarrow S\cup \{d\}$
	    
		\smallskip
		$i\leftarrow i+1$
	}	
 	\KwRet{$\x^{(i)}$}
\caption{LinearSolver}
\label{alg:linear_solver}
\end{algorithm}

\section{Sparse adversarial perturbations}
\label{sec:sparse}
\subsection{Linearized problem solution}
In solving the optimization problem~\eqref{eq:l1_linear}, the computation of the $\ell_1$ projection of $\x$ onto the approximated hyperplane does not guarantee a solution. For a perturbed image, consider the case where some of its values exceed the bounds defined by $\boldsymbol{l}$ and $\boldsymbol{u}$. Thus, by readjusting the invalid values to match the constraints, the resulted adversarial image may eventually not lie onto the approximated hyperplane.

For this reason, we propose an iterative procedure, where at each iteration we project only towards one single coordinate of the normal vector $\w$ at a time. If projecting $\x$ towards a specific direction does not provide a solution, then the perturbed image at this coordinate has reached its extrema value. Therefore, at the next iteration, this direction should be ignored, since it cannot contribute any further to finding a better solution.

Formally, let $S$ be a set containing all the directions of $\w$ that cannot contribute to the minimal perturbation. Then, the minimal perturbation $\r$ is updated through the $\ell_1$ projection of the current datapoint $\x^{(i)}$ onto the estimated hyperplane as:
\begin{equation}
\label{eq:r_d}
    r_d\leftarrow\dfrac{|\w^T(\x^{(i)}-\x_B)|}{|w_d|}\cdot\sign (w_d),
\end{equation}
where $d$ is the index of the maximum absolute value of $\w$ that has not already been used
\begin{equation}
\label{eq:argmax}
    d\leftarrow\underset{j\notin S}{\argmax|w_j|}.
\end{equation}

\begin{figure}[t]
\begin{center}
\includegraphics[width=0.7\linewidth, page=2]{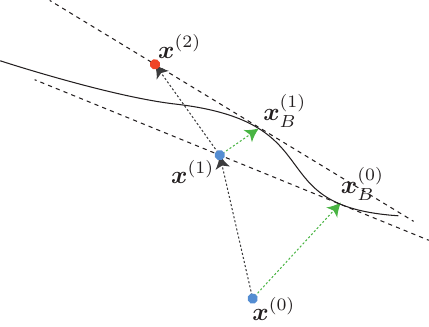}
\end{center}
   \caption{Illustration of SparseFool algorithm. With green we denote the $\ell_2$-DeepFool adversarial perturbations computed at each iteration. In this example, the algorithm converges after $2$ iterations, and the total perturbation is $\r=\x^{(2)}-\x^{(0)}$.}
\label{fig:curv_convg}
\end{figure}

Before proceeding to the next iteration, we must ensure the validity of the values of the next iterate $\x^{(i+1)}$. For this reason, we use a projection operator $Q(\cdot)$ that readjusts the values of the updated point that are out of bounds, by projecting $\x^{(i)} + \r$ onto the box-constraints defined by $\boldsymbol{l}$ and $\boldsymbol{u}$. Hence, the new iterate $\x^{(i+1)}$ is updated as
$\x^{(i+1)}\leftarrow Q(\x^{(i)} + \r)$.
Note here that the bounds $\boldsymbol{l}$, $\boldsymbol{u}$ are not limited to only represent the dynamic range of an image, but can be generalized to satisfy any similar restriction. For example, as we will describe later in Section~\ref{subsec:results}, they can be used to control the perceptibility of the computed adversarial images.

The next step is to check if the new iterate $\x^{(i+1)}$ has reached the approximated hyperplane. Otherwise, it means that the perturbed image at the coordinate $d$ has reached its extrema value, and thus we cannot change it any further; perturbing towards the corresponding direction will have no effect. Thus, we reduce the search space, by adding to the forbidden set $S$ the direction $d$, and repeat the procedure until we reach the approximated hyperplane. The algorithm for solving the linearized problem is summarized in Algorithm~\ref{alg:linear_solver}.

\AlgoDontDisplayBlockMarkers
\RestyleAlgo{ruled}
\SetAlgoNoLine
\LinesNumbered
\begin{algorithm}[t]
	\SetKwFunction{Union}{Union}\SetKwFunction{deepfool}{DeepFool}\SetKwFunction{linear}{LinearSolver}
 	\KwIn{image $\x$, projection operator $Q$, classifier $f$.}
 	\KwOut{perturbation $\r$}
 	\BlankLine
	Initialize: $\x^{(0)}\leftarrow\x,\enskip i\leftarrow 0$\\
	
	\While{$k(\x^{(i)})=k(\x^{(0)})$}
	{
	    \smallskip
		$\r_\text{adv}=$ \deepfool{$\x^{(i)}$}
		
		\smallskip
		$\x^{(i)}_B=\x^{(i)}+\r_\text{adv}$
		
		\smallskip
		$\w^{(i)}=\nabla f_{k(\x^{(i)}_B)}(\x^{(i)}_B) - \nabla f_{k(\x^{(i)})}(\x^{(i)}_B)$
		
		\smallskip
		$\x^{(i+1)}=$ \linear{$\x^{(i)}$, $\w^{(i)}$, $\x^{(i)}_B$, $Q$}
		
		\smallskip
		$i\leftarrow i+1$
	}	
 	\KwRet{$\boldsymbol{r}=\boldsymbol{x}^{(i)}-\boldsymbol{x}^{(0)}$}
\caption{SparseFool}
\label{alg:SparseFool}
\end{algorithm}

\begin{figure*}[t]
    \centering
    \begin{subfigure}[b]{0.6\columnwidth}
        \includegraphics[width=\linewidth]{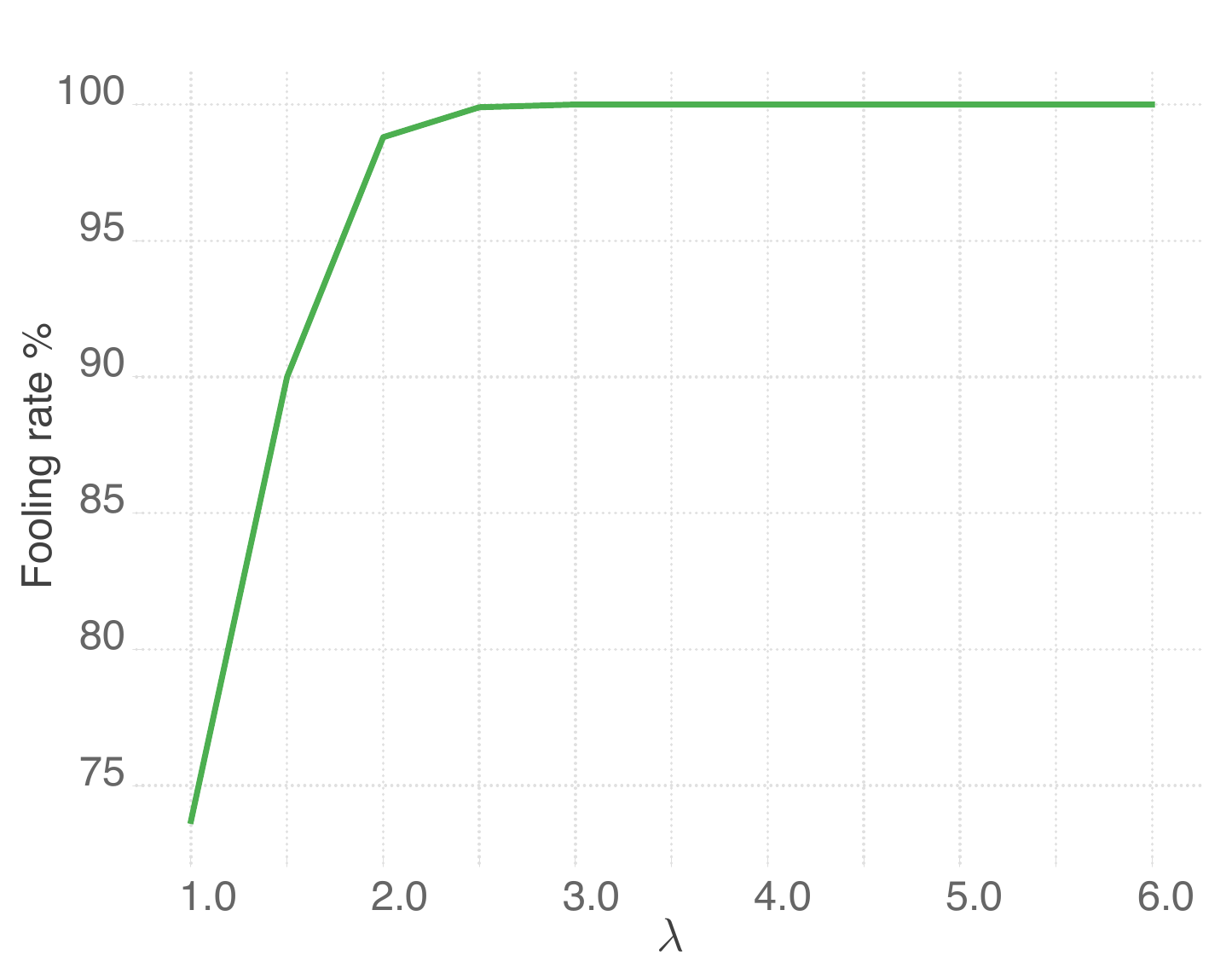}
    \end{subfigure}
    \begin{subfigure}[b]{0.6\columnwidth}
        \includegraphics[width=\linewidth]{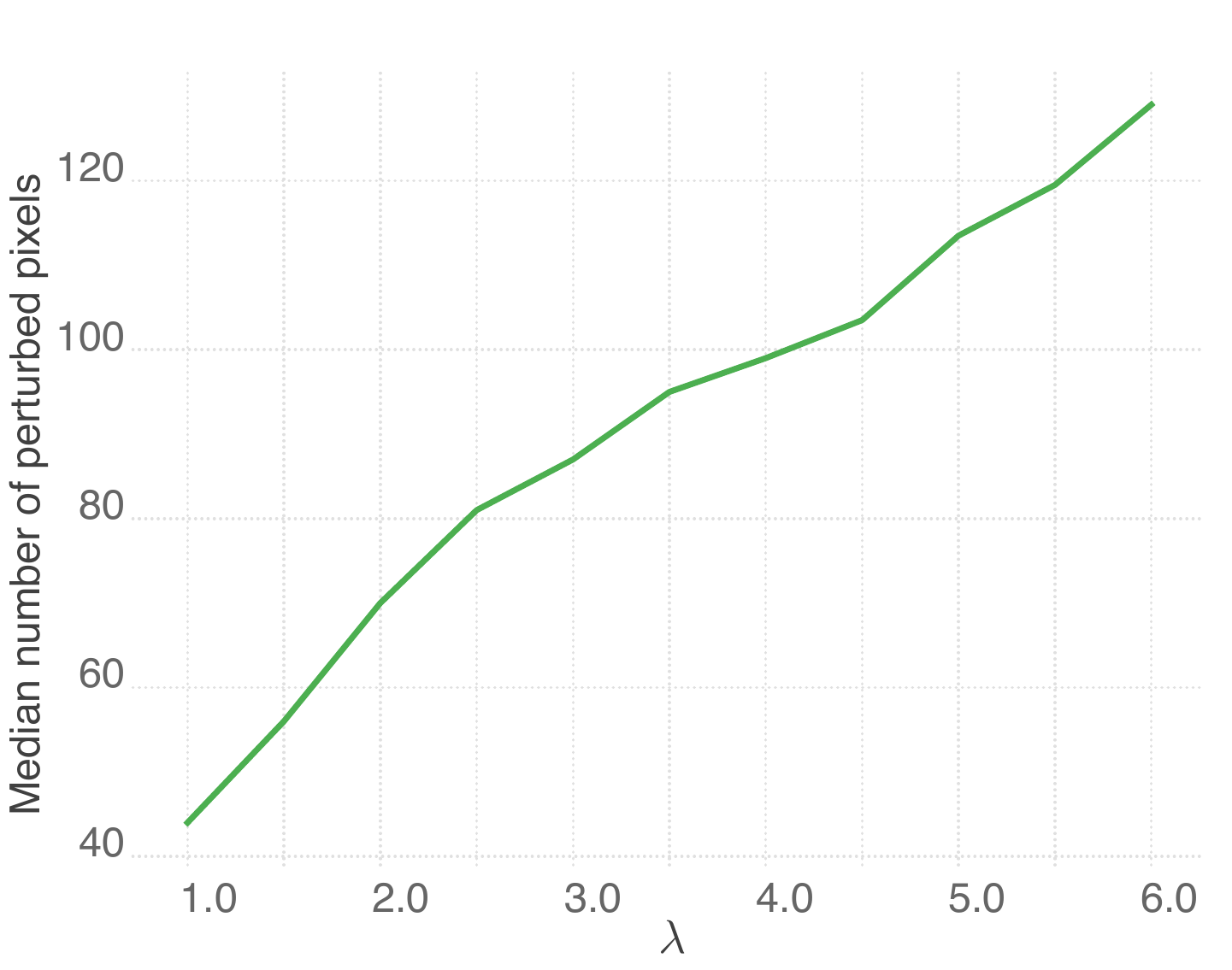}
    \end{subfigure}
    \begin{subfigure}[b]{0.6\columnwidth}
        \includegraphics[width=\linewidth]{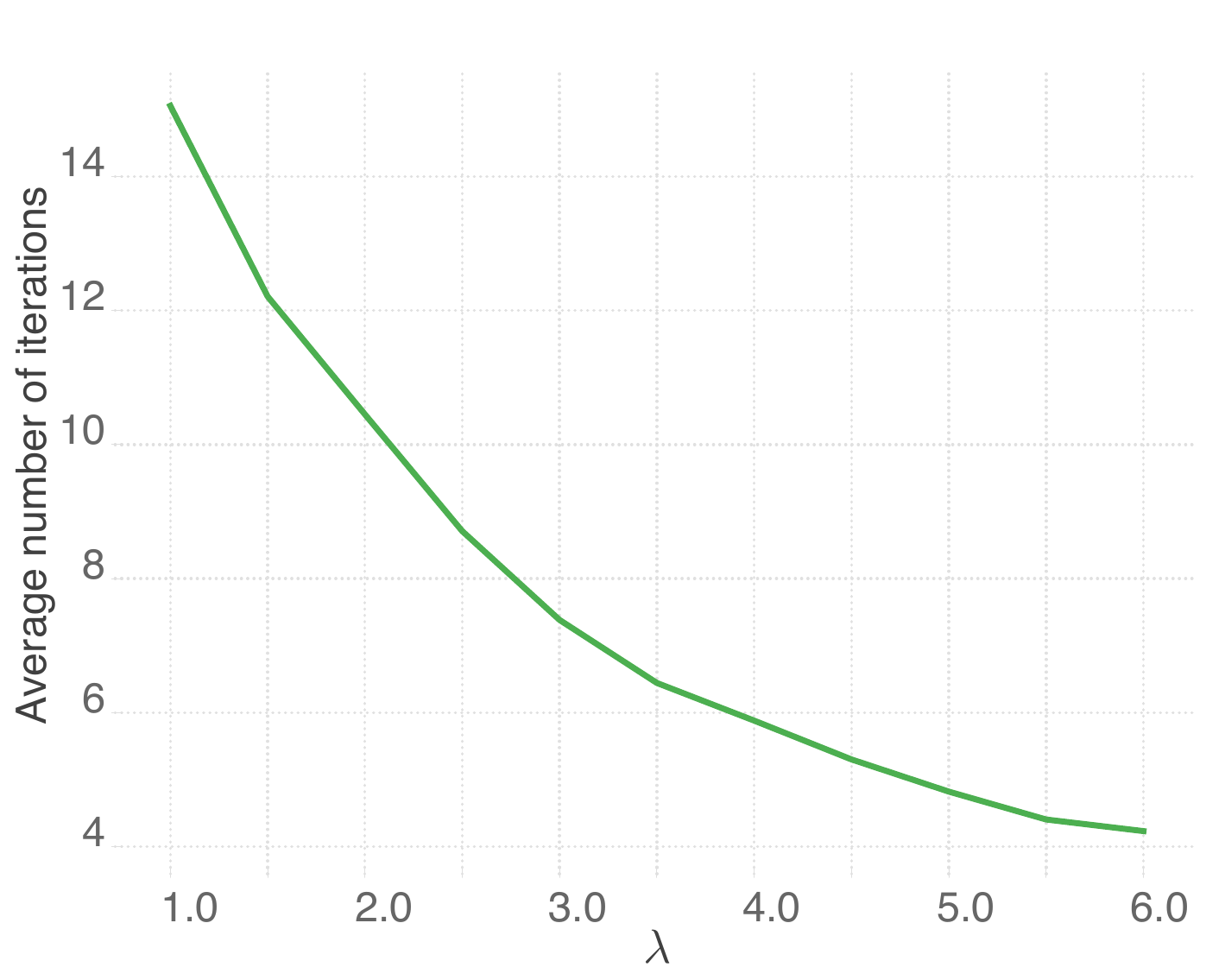}
    \end{subfigure}
    \caption{The fooling rate, the sparsity of the perturbations, and the average iterations of SparseFool for different values of $\lambda$, on $4000$ images from ImageNet dataset using an Inception-v3~\protect\cite{inception} model.}
    \label{fig:lambda_effect}
\end{figure*}

\subsection{Finding the point $\x_B$ and the normal $\w$}
\label{subsec:finding_x_w}
In order to complete our solution to the optimization problem~\eqref{eq:l1_linear}, we now focus on the linear approximation of the decision boundary. Recall from Section~\ref{subsec:2.3} that we need to find the boundary point $\x_B$, along with the corresponding normal vector $\w$.

Finding $\x_{B}$ is analogous to computing (in a $\ell_2$ sense) an adversarial example of $\x$, so it can be approximated by applying one of the existing $\ell_2$ attack algorithms. However, not all of these attacks are proper for our task; we need a fast method that finds adversarial examples that are as close to the original image $\x$ as possible. Recall that DeepFool~\cite{moosavi-dezfooli_fawzi_frossard_2016} iteratively moves $\x$ towards the decision boundary, and stops as soon as the perturbed data point reaches the other side of the boundary. Therefore, the resulting perturbed sample usually lies very close to the decision boundary, and thus, $\x_{B}$ can be very well approximated by $\x+\r_{\text{adv}}$, with $\r_{\text{adv}}$ being the corresponding $\ell_2$-DeepFool perturbation of $\x$. Then, if we denote the network's classification function as $f$, one can estimate the normal vector to the \textit{decision boundary} at the datapoint $\x_B$ as:
\begin{equation}
\label{eq:normal_vec}
    \boldsymbol{w}:=\nabla f_{k(\x_B)}(\x_B) - \nabla f_{k(\x)}(\x_B).
\end{equation}

\noindent Hence, the decision boundary can now be approximated by the affine hyperplane $\mathscr{B}\triangleq\big\{\x:\w^T(\x- \x_B)=0\big\}$, and sparse adversarial perturbations are computed by applying Algorithm~\ref{alg:linear_solver}.

\subsection{SparseFool}
\label{subsec:sparsefool}
However, although we expected to have a one-step solution, in many cases the algorithm does not converge. The reason behind this behavior lies on the fact that the decision boundaries of the networks are only \textit{locally flat}~\cite{fawzi_moosavi-dezfooli_frossard_2017, fawzi_moosavi-dezfooli_frossard_soatto_2018, jetley_lord_torr_2018}. Thus, if the $\ell_1$ perturbation moves the datapoint $\x$ away from its neighborhood, then the ``local flatness'' property might be lost, and eventually the perturbed point will not reach the other side of the decision boundary (projected onto a ``curved'' area).

We mitigate the convergence issue with an iterative method, namely SparseFool, where each iteration includes the linear approximation of the decision boundary. Specifically, at iteration $i$, the boundary point $\x_B^{(i)}$ and the normal vector $\w^{(i)}$ are estimated using $\ell_2$-DeepFool based on the current iterate $\x^{(i)}$. Then, the next iterate $\x^{(i+1)}$ is updated through the solution of Algorithm~\ref{alg:linear_solver}, having though $\x^{(i)}$ as the initial point, and the algorithm terminates when $\x^{(i)}$ changes the label of the network. An illustration of SparseFool is given in Fig.~\ref{fig:curv_convg}, and the algorithm is summarized in Algorithm~\ref{alg:SparseFool}.

However, we observed that instead of using the boundary point $\x^{(i)}_B$ at the step $6$ of SparseFool, better convergence (misclassification) can be achieved by going further into the other side of the boundary, and find a solution for the hyperplane passing through the datapoint $\x^{(i)}+\lambda (\x^{(i)}_B-\x^{(i)})$, where $\lambda\geq1$. Specifically, as shown in Fig.~\ref{fig:lambda_effect}, this parameter is used to control the trade-off between the fooling rate, the sparsity, and the complexity. Values close to $1$, lead to sparser perturbations, but also to lower fooling rate and increased complexity. On the contrary, higher values of $\lambda$ lead to fast convergence -- even one step solutions --, but the resulted perturbations are less sparse. Since $\lambda$ is the \textit{only parameter} of the algorithm, it can be easily adjusted to meet the corresponding needs in terms of fooling rate, sparsity, and complexity.

Finally, note that $\mathscr{B}$ corresponds to the boundary between the adversarial and the estimated true class, and thus it can be seen as an affine binary classifier. Since at each iteration the adversarial class is computed as the closest (in an $\ell_2$ sense) to the true one, we can say that SparseFool operates as an untargeted attack. Even though, it can be easily transformed to a targeted one, by simply computing at each iteration the adversarial example -- and thus approximating the decision boundary -- of a specific category.

\section{Experimental results}
\label{sec:results}
\subsection{Setup}
We test SparseFool on deep convolutional neural network architectures with the $10000$ images of the MNIST~\cite{lecun-mnisthandwrittendigit-2010} test set, $10000$ images of the CIFAR-10~\cite{Krizhevsky09learningmultiple} test set, and $4000$ randomly selected images from the ILSVRC2012 validation set. In order to evaluate our algorithm and compare with related works, we compute the fooling rate, the median perturbation percentage, and the average execution time. Given a dataset $\mathscr{D}$, the fooling rate measures the efficiency of the algorithm, using the formula: $\big| \x\in\mathscr{D}:k(\x+\r_{\x})\neq k(\x)\big|/\big| \mathscr{D} \big|$, where $\r_{\x}$ is the perturbation that corresponds to the image $\x$. The median perturbation percentage corresponds to the median percentage of the pixels that are perturbed per fooling sample, while the average execution time measures the average execution time of the algorithm per sample\footnote{All the experiments were conducted on a GTX TITAN X GPU. \label{gpu}}.

We compare SparseFool with the implementation of JSMA attack~\cite{papernot_mcdaniel_jha_fredrikson_celik_swami_2016}. Since JSMA is a targeted attack, we evaluate it on its ``untargeted'' version, where the target class is chosen at random. We also make one more modification at the  success condition; instead of checking if the predicted class is equal to the target one, we simply check if it is different from the source one. Let us note that JSMA is not evaluated on ImageNet dataset, due to its huge computational cost for searching over all pairs of candidates~\cite{carlini_wagner_2017}. We also compare SparseFool with the ``one-pixel attack" proposed in~\cite{su_vargas_kouichi_2018}. Since ``one-pixel attack" perturbs exactly $\kappa$ pixels, for every image we start with $\kappa=1$ and increase it till ``one-pixel attack" finds an adversarial example. Again, we do not evaluate the ``one-pixel attack" on the ImageNet dataset, due to its high computational cost for high dimensional images.

\subsection{Results}
\label{subsec:results}
\paragraph{Overall performance.}
We first evaluate the performance of SparseFool, JSMA, and ``one-pixel attack" on different datasets and architectures. The control parameter $\lambda$ in SparseFool was set to $1$ and $3$ for the MNIST and CIFAR-10 datasets respectively. We observe in Table~~\ref{tab:mnist_cifar} that SparseFool computes $2.9$x sparser perturbations, and is $4.7$x faster compared to JSMA for the MNIST dataset. This behavior remains similar for the CIFAR-10 dataset, where SparseFool computes on average, perturbations of $2.4$x higher sparsity, and is $15.5$x faster. Notice here the difference in the execution time: JSMA becomes much slower as the dimension of the input data increases, while SparseFool's time complexity remains at very low levels.

Then, in comparison to ``one-pixel attack", we observe that for the MNIST dataset, our method computes $5.5$x sparser perturbations, and is more than $3$ orders of magnitude faster. For the CIFAR-10 dataset, SparseFool still manages to find very sparse perturbations, but less so than the ``one-pixel attack" in this case. The reason is that our method does not solve the original $\ell_0$ optimization problem, but it rather computes sparse perturbations through the $\ell_1$ solution of the linearized one. The resulting solution is often suboptimal, and may be optimal when the datapoint is very close to the boundary, where the linear approximation is more accurate. However, solving our linearized problem is fast, and enables our method to efficiently scale to high dimensional data, which is not the case for the ``one-pixel attack". Considering the tradeoff between the sparsity of the solutions and the required complexity, we choose to sacrifice the former, rather than following a complex exhaustive approach like~\cite{su_vargas_kouichi_2018}. In fact, our method is able to compute sparse perturbations $270$x faster, and by requiring $2$ orders of magnitude less queries to the network, than the ``one-pixel attack" algorithm.

Finally, due to the huge computational cost of both JSMA and ``one-pixel attack", we do not use it for the large ImageNet dataset. In this case, we instead compare SparseFool with an algorithm that randomly selects a subset of elements from each color channel (RGB), and replaces their intensity with a random value from the set $V=\{0,255\}$. The cardinality of each channel subset is constrained to match SparseFool's per-channel median number of perturbed elements; for each channel, we select as many elements as the median, across all images, of SparseFool's perturbed elements for this channel. The performance of SparseFool for the ImageNet dataset is reported in Table~\ref{tab:imagenet}, while the corresponding fooling rates for the random algorithm were $18.2\%$, $13.2\%$, $14.5\%$, and $9.6\%$ respectively. Observe that the fooling rates obtained for the random algorithm are far from comparable with SparseFool's, indicating that the proposed algorithm cleverly finds very sparse solutions. Indeed, our method is consistent among different architectures, perturbing on average $0.21\%$ of the pixels, with an average execution time of $7$ seconds per sample.

To the best of our knowledge, we are the first to provide an adequate sparse attack that efficiently achieves such fooling rates and sparsity, and at the same time scales to high dimensional data. ``One-pixel attack" does not necessarily find good solutions for all the studied datasets, however, SparseFool -- as it relies on the high dimensional geometry of the classifiers --  successfully computes sparse enough perturbations for all three datasets. 

\begin{table*}[tb]
\begin{center}
{
	\begin{tabular}{| c | c | c | c | c | c| c | c | c | c | c | c |}
	\hline
	\multirow{2}{*}{Dataset}
	& \multirow{2}{*}{Network}
	& \multirow{2}{*}{Acc. (\%)}
	& \multicolumn{3}{c|}{Fooling rate (\%)}
	& \multicolumn{3}{c|}{Perturbation (\%)}
	& \multicolumn{3}{c|}{Time (sec)} \\
	\cline{4-12}
	\multicolumn{1}{|c|}{}
	& \multicolumn{1}{c|}{}
	& \multicolumn{1}{c|}{}
	& \multicolumn{1}{c|}{SF}
	& \multicolumn{1}{c|}{JSMA}
	& \multicolumn{1}{c|}{$1$-PA}
	& \multicolumn{1}{c|}{SF}
	& \multicolumn{1}{c|}{JSMA}
	& \multicolumn{1}{c|}{$1$-PA}
	& \multicolumn{1}{c|}{SF}
	& \multicolumn{1}{c|}{JSMA}
	& \multicolumn{1}{c|}{$1$-PA} \\
	\hline
	\multicolumn{1}{|c|}{MNIST}
	& \multicolumn{1}{c|}{LeNet~\protect\cite{lenet}}
	& \multicolumn{1}{c|}{$99.14$}
	& \multicolumn{1}{c|}{$99.93$}
	& \multicolumn{1}{c|}{$95.73$} 
	& \multicolumn{1}{c|}{$100$} 
	& \multicolumn{1}{c|}{$1.66$}
	& \multicolumn{1}{c|}{$4.85$} 
	& \multicolumn{1}{c|}{$9.43$} 
	& \multicolumn{1}{c|}{$0.14$}
	& \multicolumn{1}{c|}{$0.66$}
	& \multicolumn{1}{c|}{$310.2$}  \\
	\hline
	\hline
	\multirow{2}{*}{CIFAR-10}
	& \multicolumn{1}{c|}{VGG-19}
	& \multicolumn{1}{c|}{$92.71$}
	& \multicolumn{1}{c|}{$100$}
	& \multicolumn{1}{c|}{$98.12$} 
	& \multicolumn{1}{c|}{$100$} 
	& \multicolumn{1}{c|}{$1.07$}
	& \multicolumn{1}{c|}{$2.25$} 
	& \multicolumn{1}{c|}{$0.15$} 
	& \multicolumn{1}{c|}{$0.34$}
	& \multicolumn{1}{c|}{$6.28$}
	& \multicolumn{1}{c|}{$102.7$}  \\
	\cline{2-12}
	& \multicolumn{1}{c|}{ResNet18~\protect\cite{resnet}}
	& \multicolumn{1}{c|}{$92.74$}
	& \multicolumn{1}{c|}{$100$}
	& \multicolumn{1}{c|}{$100$}
	& \multicolumn{1}{c|}{$100$}  
	& \multicolumn{1}{c|}{$1.27$}
	& \multicolumn{1}{c|}{$3.91$}
	& \multicolumn{1}{c|}{$0.2$}  
	& \multicolumn{1}{c|}{$0.69$}
	& \multicolumn{1}{c|}{$8.73$}
	& \multicolumn{1}{c|}{$167.4$}  \\
	\hline
\end{tabular}
}
\end{center}
\caption{The performance of SparseFool (SF), JSMA~\cite{papernot_mcdaniel_jha_fredrikson_celik_swami_2016}, and ``one-pixel attack" ($1$-PA)~\cite{su_vargas_kouichi_2018} on the MNIST and the CIFAR-10 datasets.\footref{gpu} Note that due to its high complexity, ``one-pixel attack" was evaluated on only $100$ samples.}
\label{tab:mnist_cifar}
\end{table*}

\begin{table}[tb]
\begin{center}
{
	\begin{tabular}{| c | c | c | c | c |}
	\hline
	\multicolumn{1}{|c|}{Network}
	& \multicolumn{1}{c|}{\makecell{Acc. (\%)}}
	& \multicolumn{1}{c|}{\makecell{Fooling\\rate (\%)}}
	& \multicolumn{1}{c|}{\makecell{Pert.\\(\%)}}
	& \multicolumn{1}{c|}{\makecell{Time \\ (sec)}}\\
	\hline
	\multicolumn{1}{|c|}{VGG-16}
	& \multicolumn{1}{c|}{$71.59$}
	& \multicolumn{1}{c|}{$100$} 
	& \multicolumn{1}{c|}{$0.18$}
	& \multicolumn{1}{c|}{$5.09$} \\
	\hline
	\multicolumn{1}{|c|}{ResNet-101}
	& \multicolumn{1}{c|}{$77.37$}
	& \multicolumn{1}{c|}{$100$} 
	& \multicolumn{1}{c|}{$0.23$}
	& \multicolumn{1}{c|}{$8.07$}\\
	\hline
	\multicolumn{1}{|c|}{DenseNet-161}
	& \multicolumn{1}{c|}{$77.65$}
	& \multicolumn{1}{c|}{$100$}
	& \multicolumn{1}{c|}{$0.29$} 
	& \multicolumn{1}{c|}{$10.07$} \\
	\hline
	\multicolumn{1}{|c|}{Inception-v3}
	& \multicolumn{1}{c|}{$77.45$}
	& \multicolumn{1}{c|}{$100$}
	& \multicolumn{1}{c|}{$0.14$} 
	& \multicolumn{1}{c|}{$4.94$} \\
	\hline
\end{tabular}
}
\end{center}
\caption{The performance of SparseFool on the ImageNet dataset, using the pre-trained models provided by PyTorch.\footref{gpu}}
\label{tab:imagenet}
\end{table}

\begin{figure}[t]
    \centering
    \begin{subfigure}[b]{0.5\columnwidth}
        \centering
        \captionsetup{font=scriptsize}
        \includegraphics[width=0.9\linewidth, page=1]{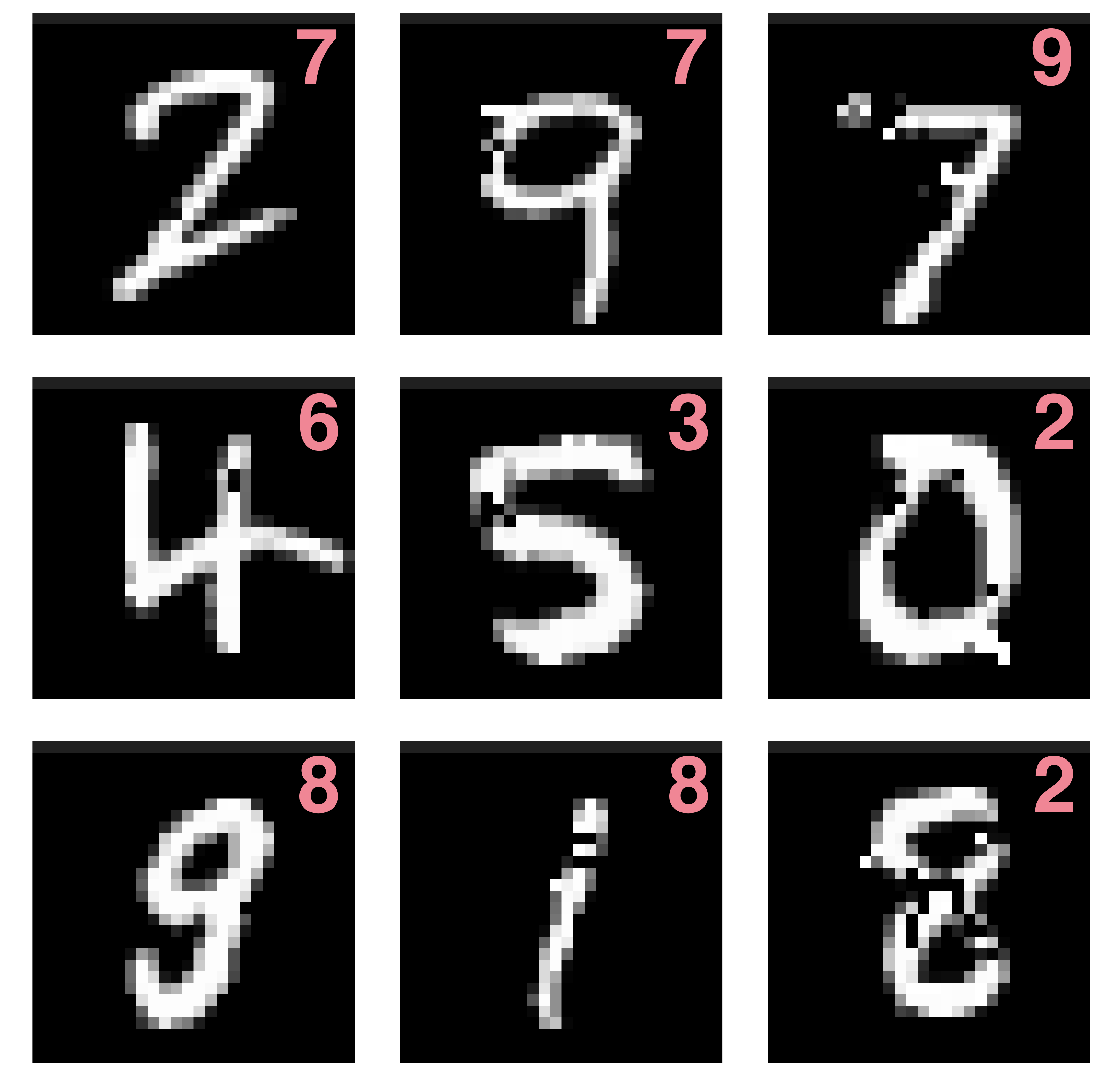}
    \captionsetup{font=small}
    \caption{MNIST}
    \label{fig:pert_mnist}
    \end{subfigure}\hspace{-0.15em}%
    \begin{subfigure}[b]{0.5\columnwidth}
        \centering
        \captionsetup{font=scriptsize}
        \includegraphics[width=0.9\linewidth, page=2]{./mnist_cifar.pdf}
    \captionsetup{font=small}
    \caption{CIFAR-10}
    \label{fig:pert_cifar}
    \end{subfigure}
    \caption{Adversarial examples for (a) MNIST and (b) CIFAR-10 datasets, as computed by SparseFool on LeNet and ResNet-18 architectures respectively. Each column corresponds to different level of perturbed pixels.}
    \label{fig:pert_mnist_cifar}
\end{figure}

\textbf{Perceptibility.}\quad In this section, we illustrate some adversarial examples generated by SparseFool, for three different levels of sparsity; highly sparse perturbations, sparse perturbations, and somewhere in the middle. For the MNIST and CIFAR-10 datasets (Figures~\ref{fig:pert_mnist} and~\ref{fig:pert_cifar} respectively), we observe that for highly sparse cases, the perturbation is either imperceptible or so sparse (i.e., $1$ pixel) that it can be easily ignored. However, as the number of perturbed pixels increases, the distortion becomes even more perceptible, and in some cases the noise is detectable and far from imperceptible. The same behavior is observed for the ImageNet dataset (Fig.~\ref{fig:pert_imagenet}). For highly sparse perturbations, the noise is again either imperceptible or negligible, but as the density increases, it becomes more and more visible.

\begin{figure}[t]
\centering
\includegraphics[width=0.7\columnwidth]{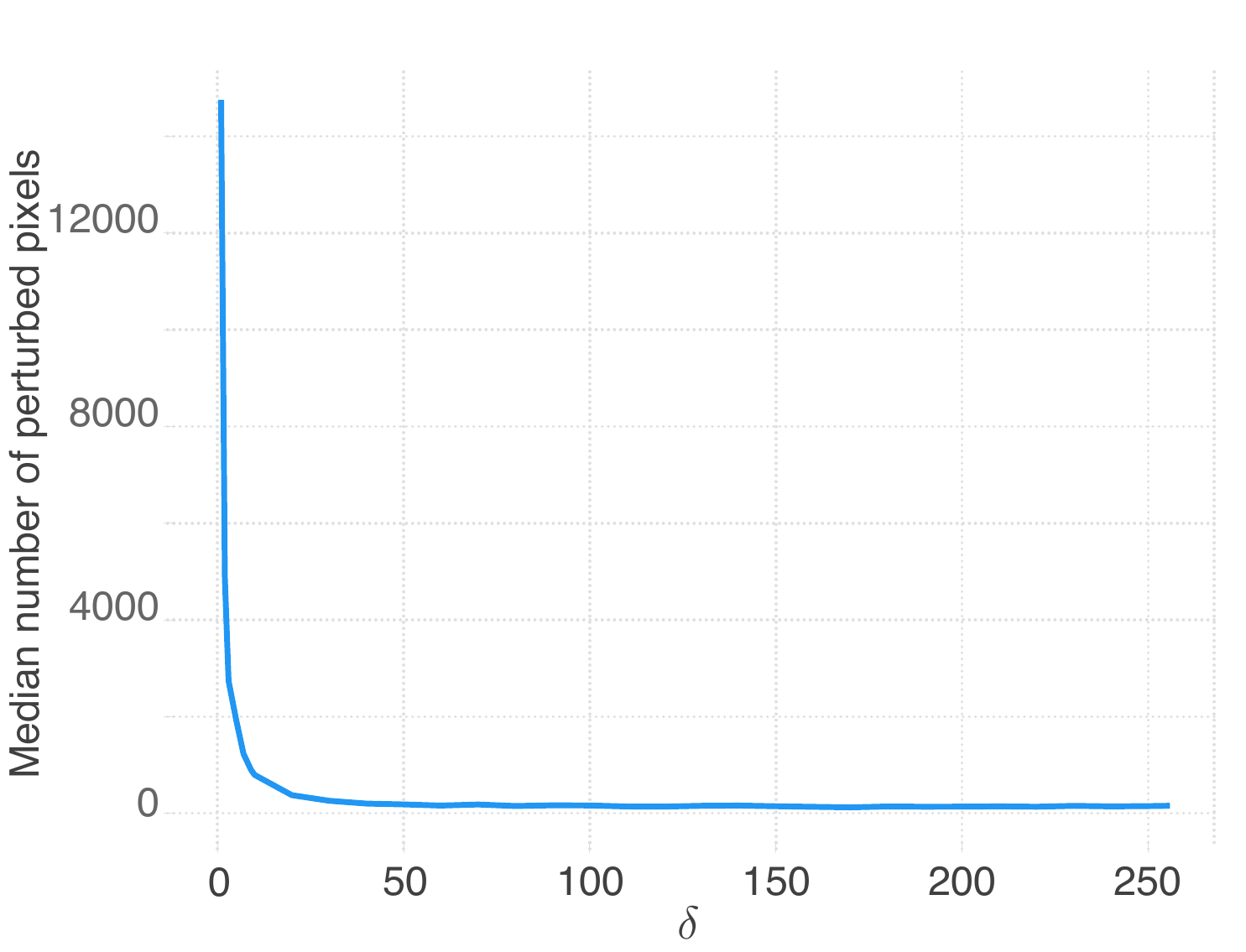}
\caption{The resulted sparsity of SparseFool perturbations for $\pm \delta$ around the values of $\x$, for $100$ samples from ImageNet on a ResNet-101 architecture.}
\label{fig:perc_sparse_const}
\end{figure}
\begin{figure}[ht]
    \captionsetup[subfigure]{justification=centering}
    \centering
    \begin{subfigure}[b]{0.33\columnwidth}
        \caption*{$x_i \pm 255$}
        \includegraphics[width=0.9\linewidth]{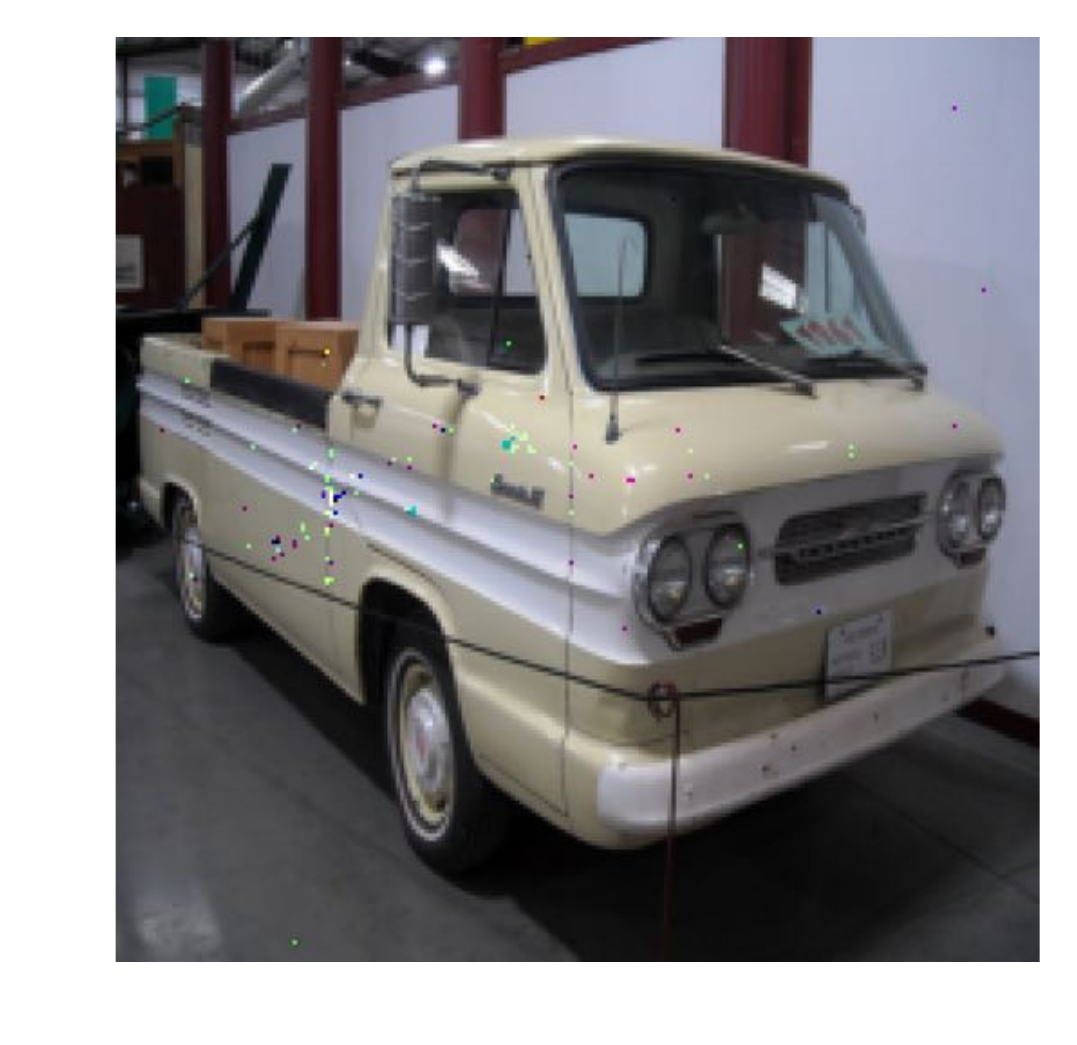}
        \caption*{amphibian\\($0.227\%$)}
    \end{subfigure}
    \begin{subfigure}[b]{0.33\columnwidth}
        \caption*{$x_i \pm 30$}
        \includegraphics[width=0.9\linewidth]{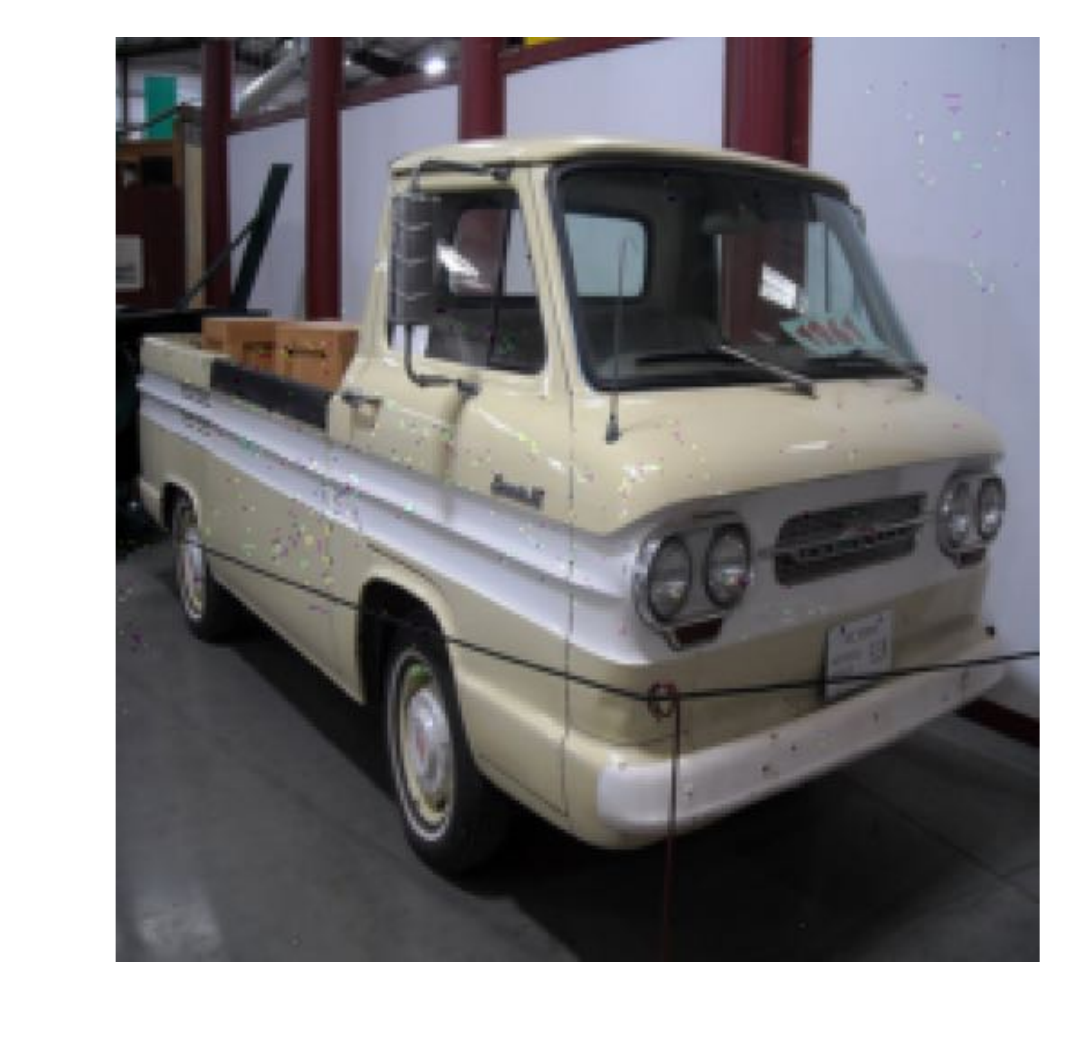}
        \caption*{amphibian\\($1.058\%$)}
    \end{subfigure}
    \begin{subfigure}[b]{0.33\columnwidth}
        \caption*{$x_i \pm 10$}
        \includegraphics[width=0.9\linewidth]{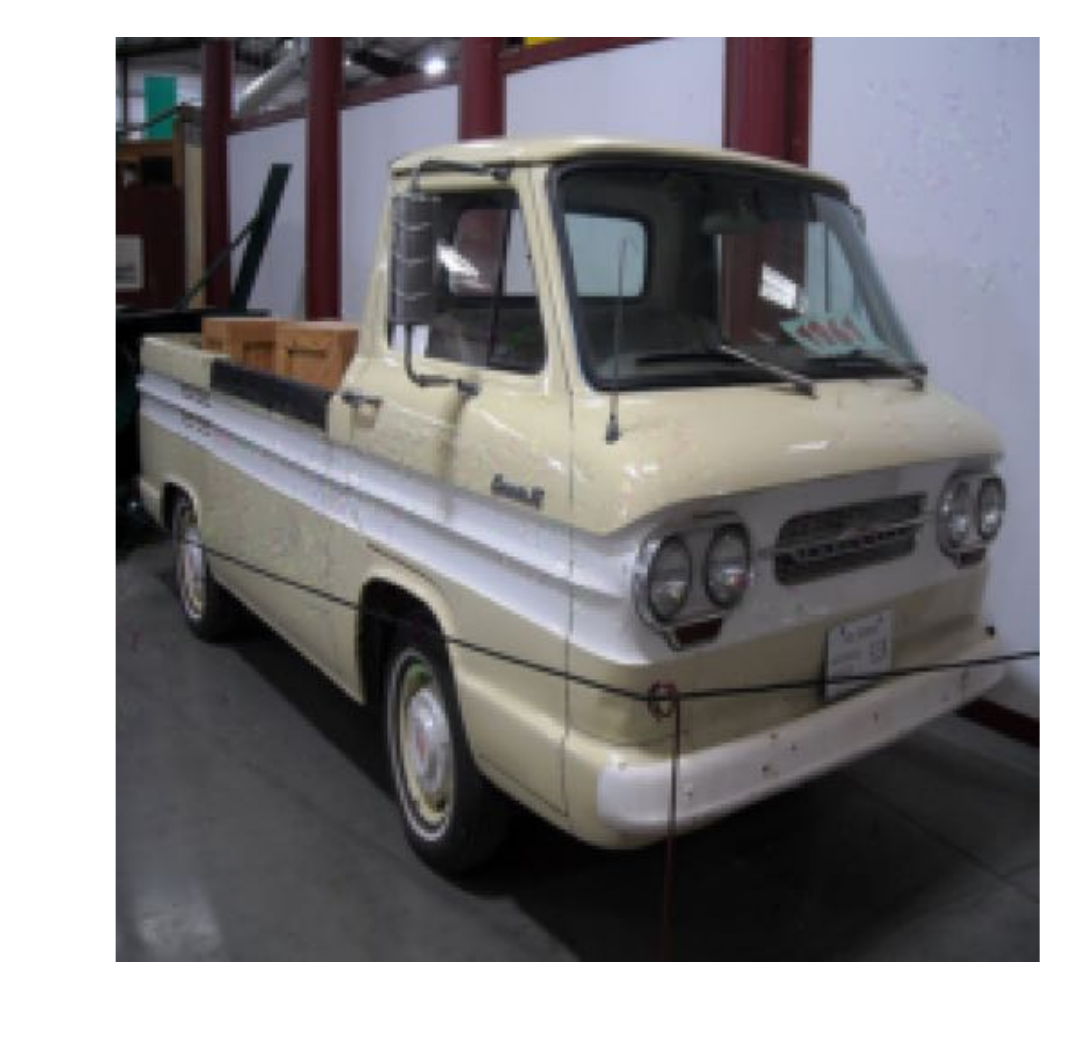}
        \caption*{amphibian\\($4.296\%$)}
    \end{subfigure}
    
    \begin{subfigure}[b]{0.33\columnwidth}
        \includegraphics[width=0.9\linewidth]{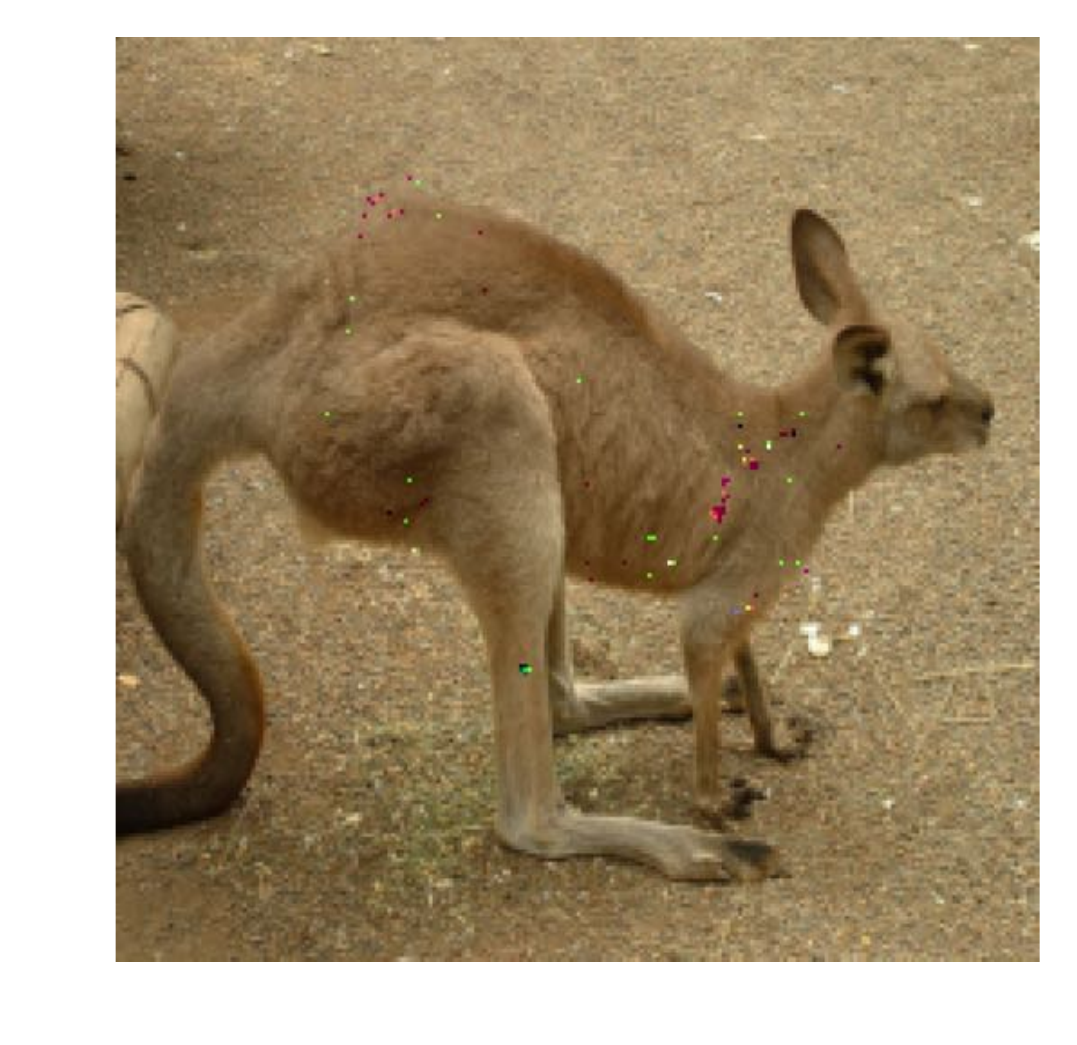}
        \caption*{Arabian camel\\($0.169\%$)}
    \end{subfigure}
    \begin{subfigure}[b]{0.33\columnwidth}
        \includegraphics[width=0.9\linewidth]{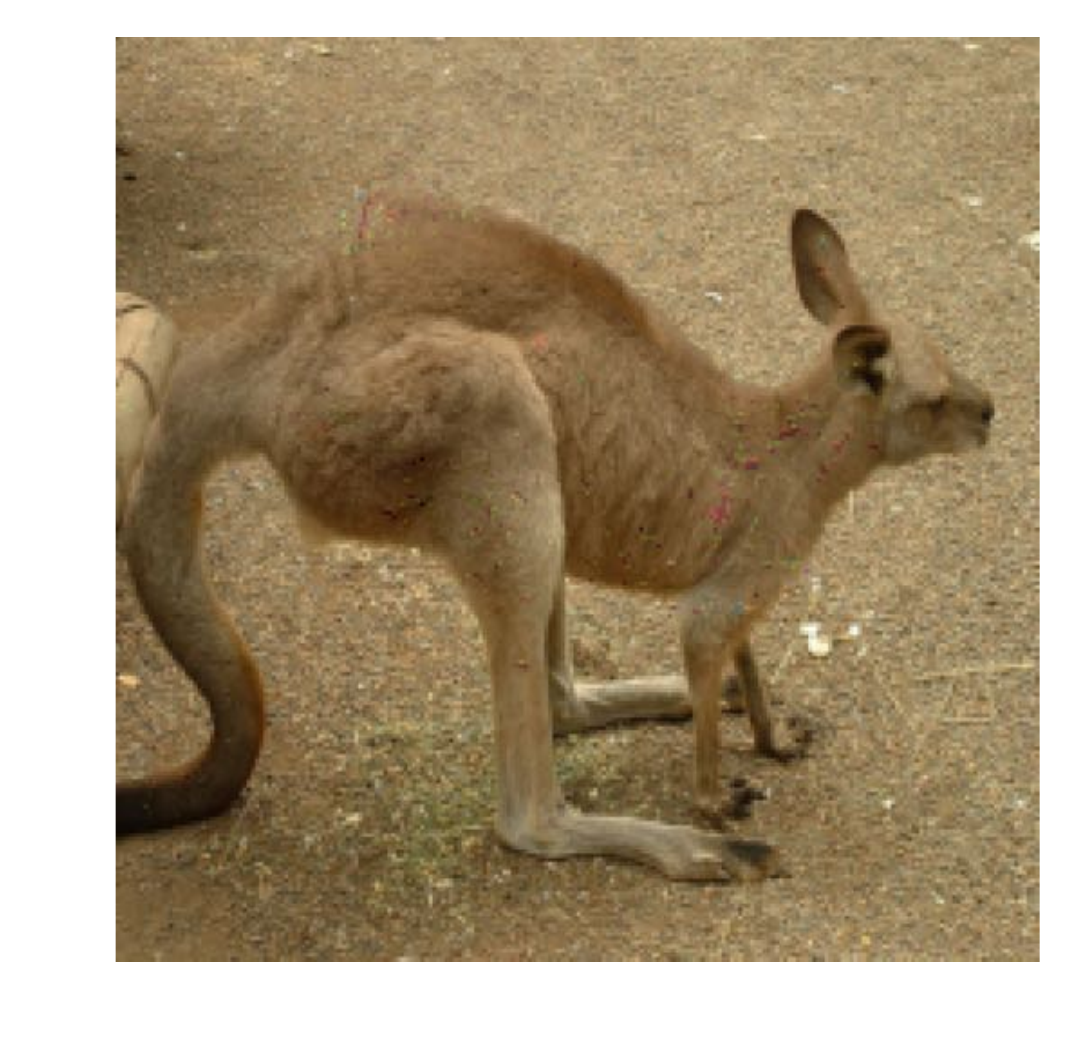}
        \caption*{Arabian camel\\($0.839\%$)}
    \end{subfigure}
    \begin{subfigure}[b]{0.33\columnwidth}
        \includegraphics[width=0.9\linewidth]{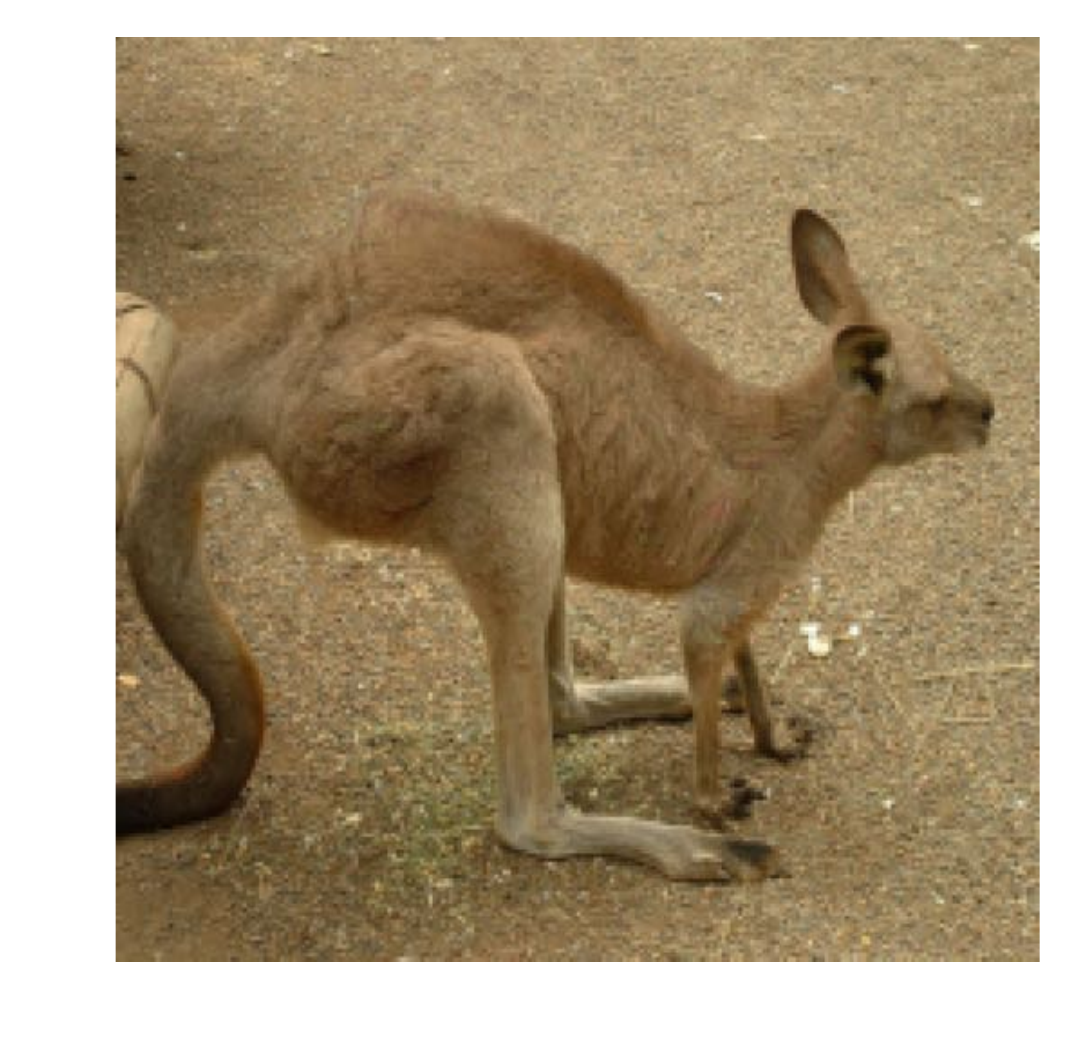}
        \caption*{Arabian camel\\($3.202\%$)}
    \end{subfigure}
    \caption{The effect of $\delta$ on the perceptibility and the sparsity of SparseFool perturbations. The values of $\delta$ are shown on top of each column, while the fooling label and the percentage of perturbed pixels are written below each image.}
    \label{fig:percept_delta}
\end{figure}

To eliminate this perceptibility effect, we focus on the lower and upper bounds of the values of an adversarial image $\hat{\x}$. Recall from Section~\ref{subsec:2.3} that the bounds $\boldsymbol{l}$, $\boldsymbol{u}$ are defined in way such that $l_i \leq \hat{x}_i \leq u_i, \enskip i=1\dots n$. If these bounds represent the dynamic range of the image, then $\hat{x}_i$ can take every possible value from this range, and the magnitude of the noise at the element $i$ can reach some visible levels. However, if the perturbed values of each element lie close to the original values $x_i$, then we might prevent the magnitude from reaching very high levels. For this reason, assuming a dynamic range of $[0,255]$, we explicitly constrain the values of $\hat{x}_i$ to lie in a small interval $\pm\delta$ around $x_i$, such that $0\leq x_i-\delta\leq\hat{x}_i\leq x_i+\delta \leq 255$.

The resulted sparsity for different values of $\delta$ is depicted in Fig.~\ref{fig:perc_sparse_const}. The higher the value, the more freedom we give to the perturbations, and for $\delta=255$ we exploit the whole dynamic range. Observe though that after $\delta\approx25$, the sparsity levels seem to remain almost constant, which indicates that we do not need to use the whole dynamic range. Furthermore, we observed that the average execution time per sample of SparseFool from this value onward seems to remain constant as well, while the fooling rate is always $100\%$ regardless $\delta$. Thus, by selecting appropriate values for $\delta$, we can control the perceptibility of the resulted perturbations, retain the sparsity around a sufficient level. The influence of $\delta$ on the perceptibility and the sparsity of the perturbations is demonstrated in Fig.~\ref{fig:percept_delta}.

\textbf{Transferability and semantic information.}\quad We now study if SparseFool adversarial perturbations can generalize across different architectures. For the VGG-16, ResNet-101, and DenseNet-161~\cite{densenet} architectures, we report in Table~\ref{tab:transfer} the fooling rate of each model when fed with adversarial examples generated by another one. We observe that sparse adversarial perturbations can generalize only to some extent, and also, as expected~\cite{Liu2016DelvingIT}, they are more transferable from larger to smaller architectures. This indicates that there should be some shared semantic information between different architectures that SparseFool exploits, but the perturbations are mostly network dependent.
\begin{table}[t]
\begin{center}
{
	\begin{tabular}{| m{1.8cm} | m{1.7cm} | m{1.7cm} | m{1.6cm} |}
	\cline{2-4}
	\multicolumn{1}{c|}{}
	& \multicolumn{1}{c|}{\makecell{VGG16}}
	& \multicolumn{1}{c|}{\makecell{ResNet101}}
	& \multicolumn{1}{c|}{\makecell{DenseNet161}}\\
	\hline
	VGG16
	& \multicolumn{1}{c|}{$100\%$}
	& \multicolumn{1}{c|}{$10.8\%$} 
	& \multicolumn{1}{c|}{$8.2\%$} \\
	\hline
	ResNet101
	& \multicolumn{1}{c|}{$25.3\%$}
	& \multicolumn{1}{c|}{$100\%$} 
	& \multicolumn{1}{c|}{$12.1\%$} \\
	\hline
	\multicolumn{1}{|c|}{DenseNet161}
	& \multicolumn{1}{c|}{$28.2\%$}
	& \multicolumn{1}{c|}{$17.5\%$} 
	& \multicolumn{1}{c|}{$100\%$} \\
	\hline
\end{tabular}
}
\end{center}
\caption{Fooling rates of SparseFool perturbations between pairs of models for $4000$ samples from ImageNet. The row and column denote the source and target model respectively.}
\label{tab:transfer}
\end{table}
 
\begin{figure}[t]
\centering
    \begin{subfigure}[b]{0.33\columnwidth}
        \caption{VGG-16}
        \includegraphics[width=0.9\linewidth]{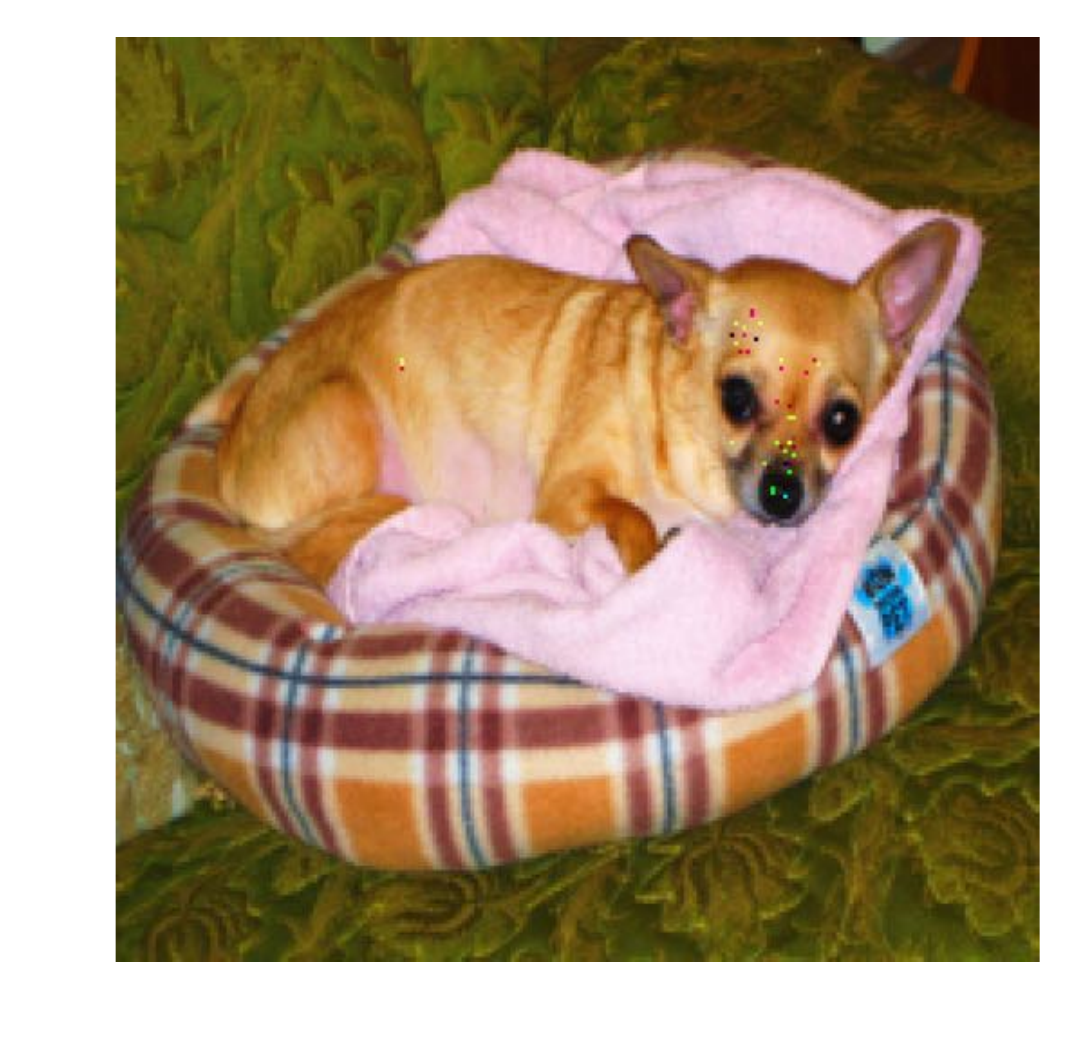}
    \end{subfigure}\!
    \begin{subfigure}[b]{0.33\columnwidth}
        \caption{ResNet-101}
        \includegraphics[width=0.9\linewidth]{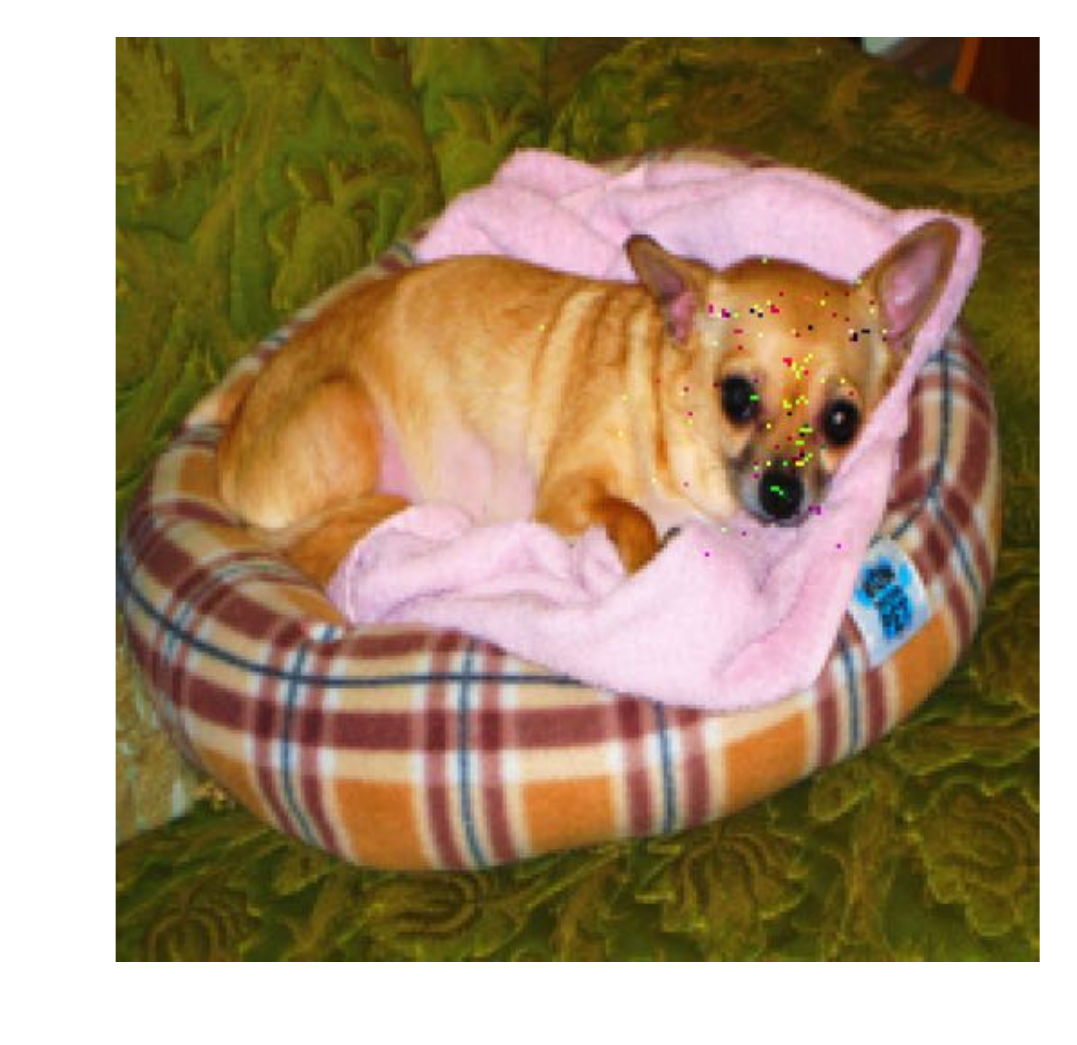}
    \end{subfigure}\!
    \begin{subfigure}[b]{0.33\columnwidth}
        \caption{DenseNet-161}
        \includegraphics[width=0.9\linewidth]{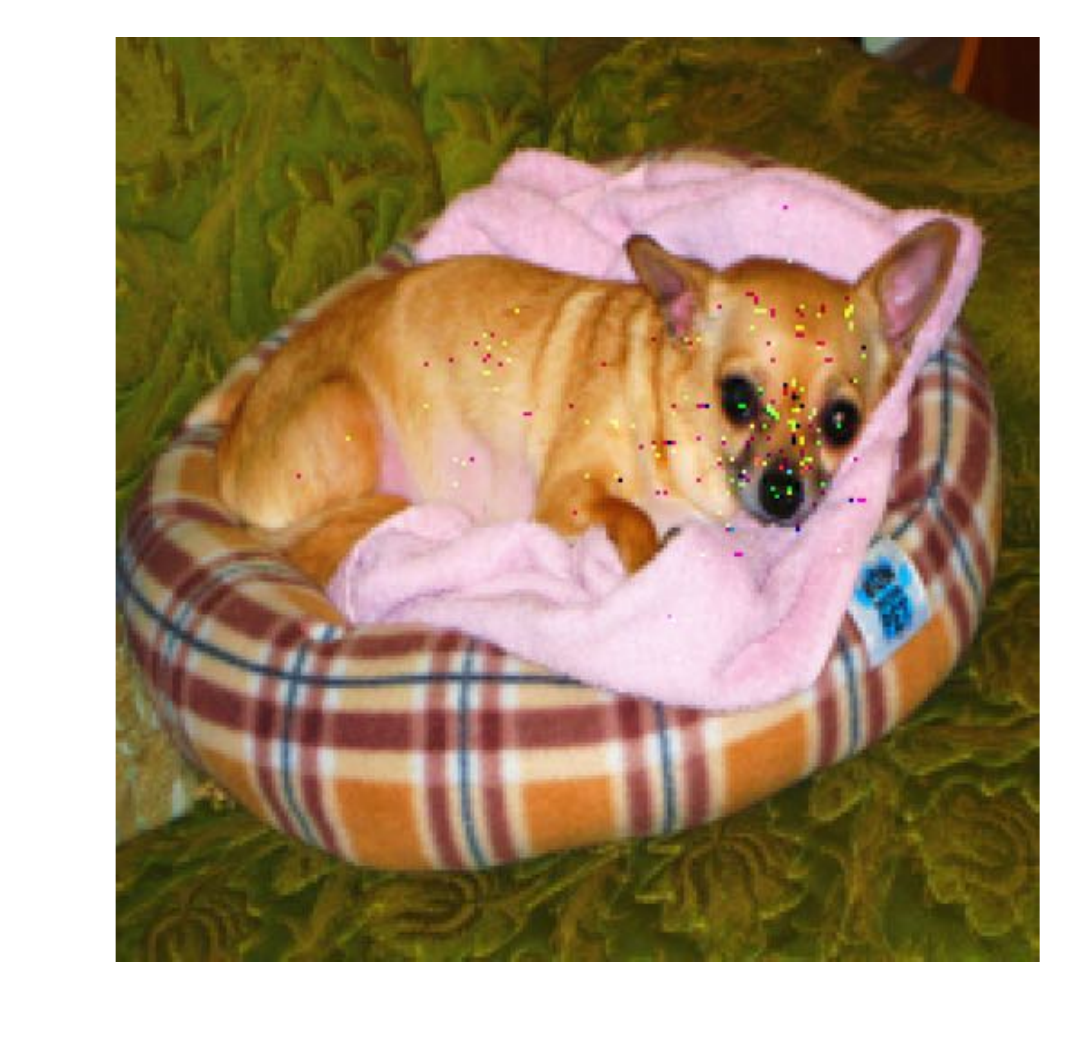}
    \end{subfigure}
    
    \begin{subfigure}[b]{0.33\columnwidth}
        \includegraphics[width=0.9\linewidth]{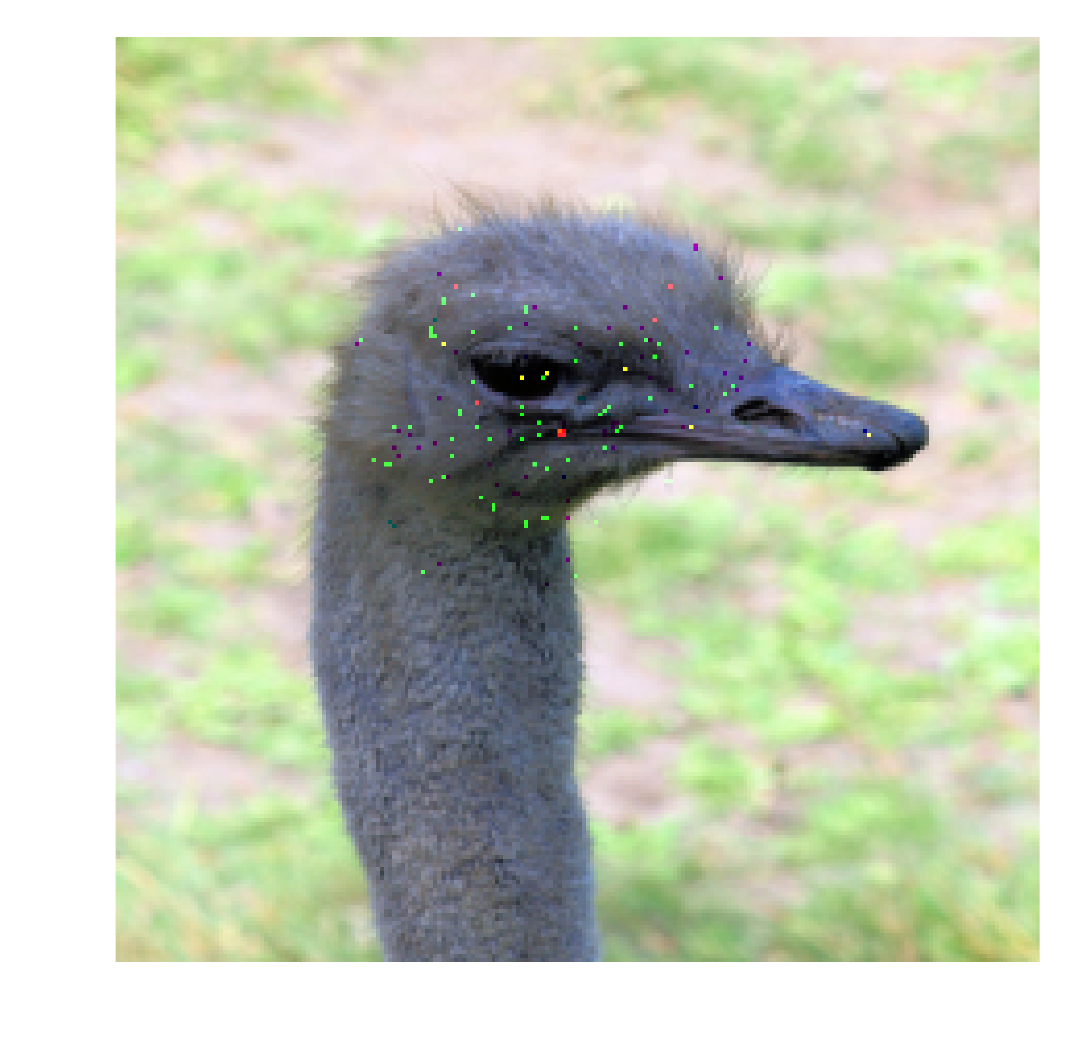}
    \end{subfigure}\!
    \begin{subfigure}[b]{0.33\columnwidth}
        \includegraphics[width=0.9\linewidth]{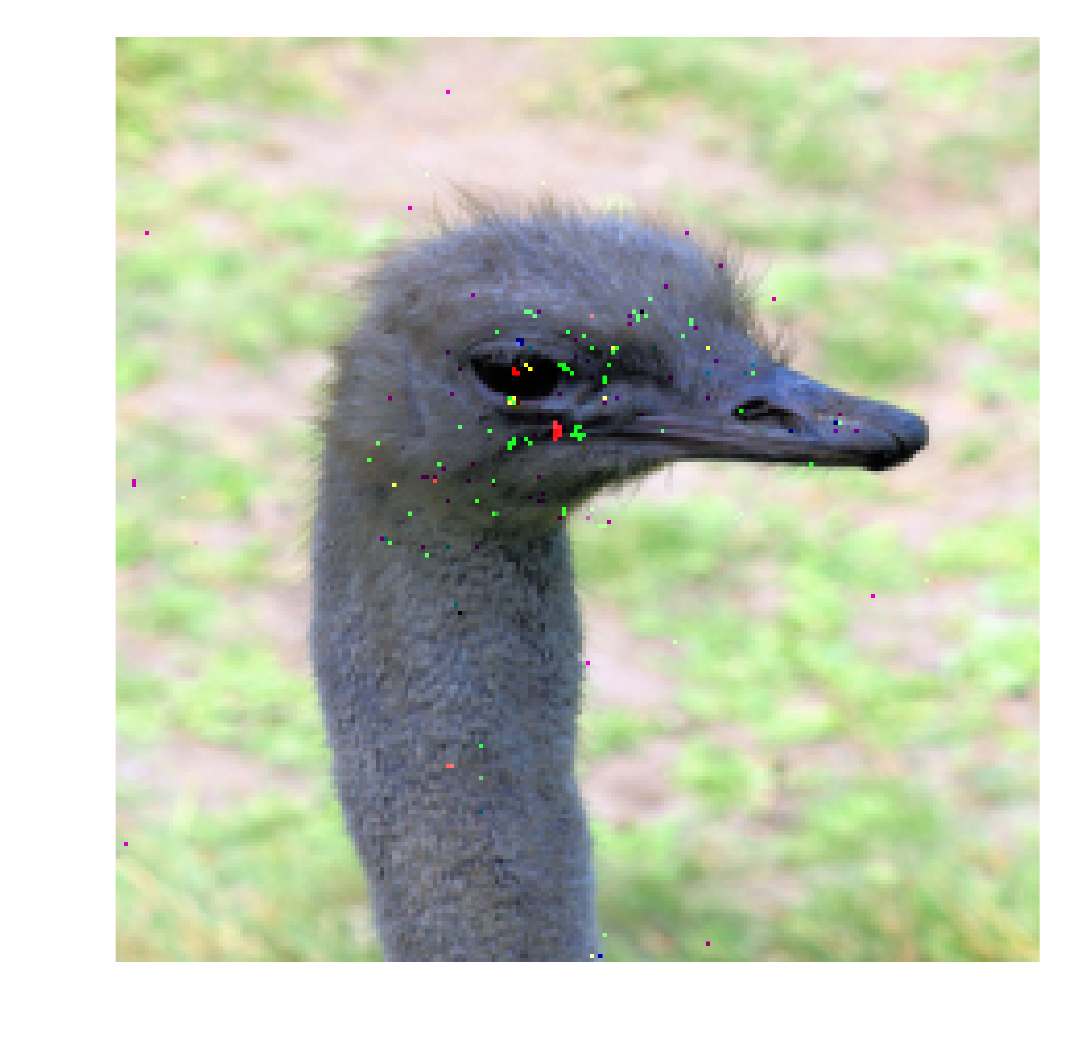}
    \end{subfigure}\!
    \begin{subfigure}[b]{0.33\columnwidth}
        \includegraphics[width=0.9\linewidth]{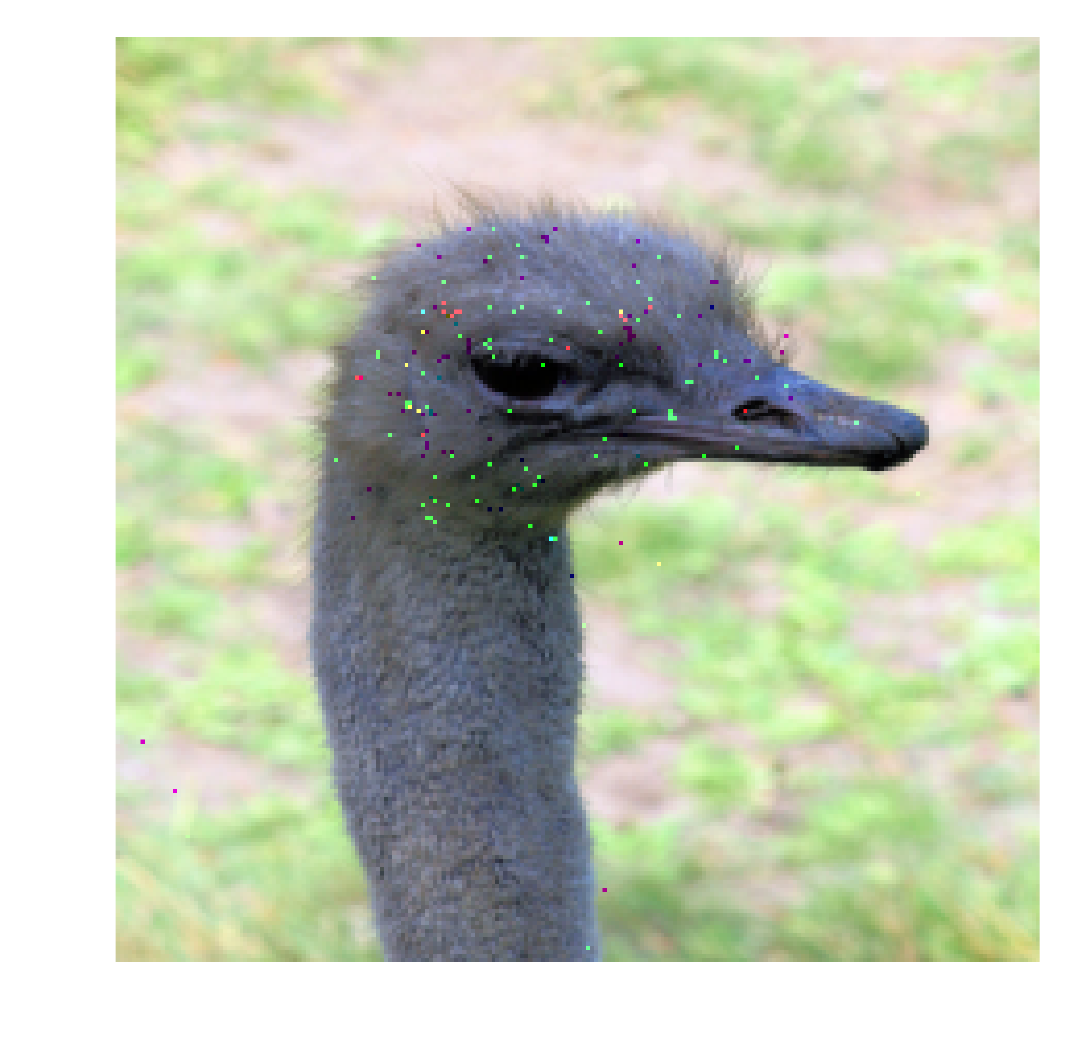}
    \end{subfigure}
    \caption{Shared information of the SparseFool perturbations for ImageNet, as computed on three different architectures. For all networks, the first row image was classified as ``Chihuahua" and misclassified as ``French Bulldog", and second row image as ``Ostrich" and ``Crane" respectively.}
    \label{fig:univ_imagenet}
    \vspace{-5mm}
\end{figure}

Focusing on some animal categories of the ImageNet dataset, we observe that indeed the perturbations are concentrated around ``important" areas (i.e., head), but there is not a consistent pattern to indicate specific features that are the most important for the network (i.e., eyes, ears, nose etc.); in many cases the noise also spreads around different parts of the image. Examining now if semantic information is shared across different architectures (Fig.~\ref{fig:univ_imagenet}), we observe that in all the networks, the noise consistently lies around the important areas of the image, but the way it concentrates or spreads is different for each network.

\begin{figure}[t]
    \centering
    \begin{subfigure}[b]{0.5\columnwidth}
        \centering
        \includegraphics[width=0.9\linewidth, page=1]{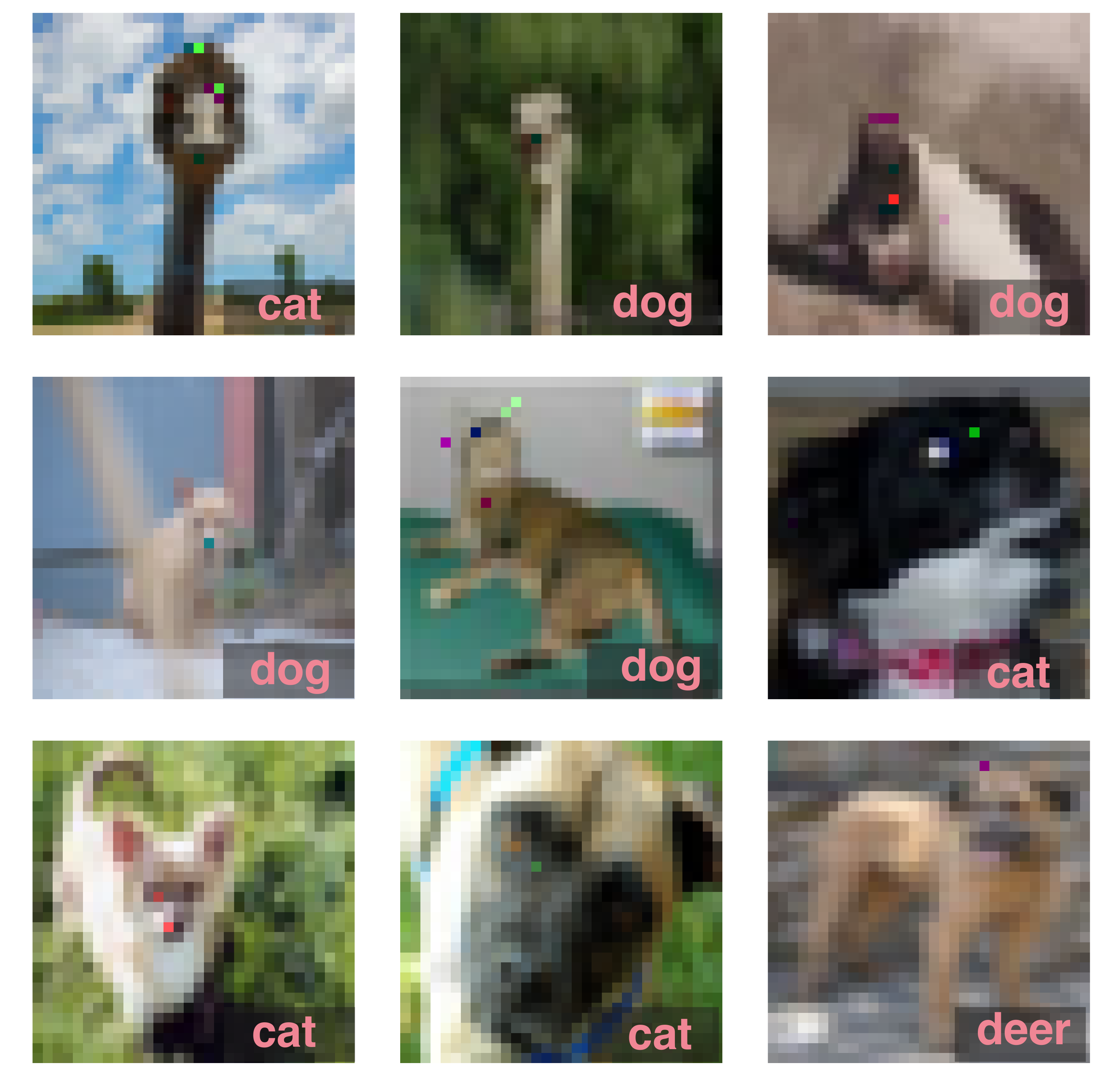}
    \caption{Universal features}
    \label{fig:univ_features}
    \end{subfigure}\hspace{-0.15em}%
    \begin{subfigure}[b]{0.5\columnwidth}
        \centering
        \includegraphics[width=0.9\linewidth, page=2]{./univ_feat.pdf}
    \caption{Fooling class features}
    \label{fig:univ_target}
    \end{subfigure}
    \caption{Semantic information of SparseFool perturbations for the CIFAR-10 dataset, on a ResNet-18 architecture. Observe that the perturbation is concentrated on (a) some features around the area of the face, and (b) on areas that are important for the fooling class.}
    \label{fig:univ_cifar}
\end{figure}

For the CIFAR-10 dataset, we observe that in many cases of animal classes, SparseFool tends to perturb some universal features around the area of the head (i.e., eyes, ears, nose, mouth etc.), as shown in Fig.~\ref{fig:univ_features}. Furthermore, we tried to understand if there is a correlation between the perturbed pixels and the fooling class. Interestingly, we observed that in many cases, the algorithm was perturbing those regions of the image that correspond to important features of the fooling class, i.e., when changing a ``bird" label to a ``plane", where the perturbation seems to represent some parts of the plane (i.e., wings, tail, nose, turbine). This behavior becomes even more obvious when the fooling label is a ``deer", where the noise lies mostly around the area of the head in a way that resembles the antlers.

\textbf{Robustness of adversarially trained network.}\quad Finally, we study the robustness to sparse perturbations of an adversarially trained ResNet-18 network on the CIFAR-10 dataset, using $\ell_\infty$ perturbations. The accuracy of this more robust network is $82.17\%$, while the training procedure and its overall performance are similar to the ones provided in~\cite{madry_makelov_schmidt_tsipras_vladu_2017}. Compared to the results of Table~\ref{tab:mnist_cifar}, the average time dropped to $0.3$ sec, but the perturbation percentage increased to $2.44\%$. In other words, the adversarially trained network leads to just a slight change in the sparsity, and thus training it on $\ell_\infty$ perturbations does not significantly improve its robustness against sparse perturbations.

\section{Conclusion}
Computing adversarial perturbations beyond simple $\ell_p$ norms is a challenging problem from different aspects. For sparse perturbations, apart from the NP-hardness of the $\ell_0$ minimization, one needs to ensure the validity of the adversarial example values. In this work, we provide a novel geometry inspired sparse attack algorithm that is fast and can scale to high dimensional data, where one can also have leverage over the sparsity of the resulted perturbations. Furthermore, we provide a simple method to improve the perceptibility of the perturbations, while retaining the levels of sparsity and complexity.
We also note that for some datasets, the proposed sparse attack alters features that are shared among different images. Finally, we show that adversarial training does not significantly improve the robustness against sparse perturbations computed with SparseFool. We believe that the proposed approach can be used to further understand the behavior and the geometry of deep image classifiers, and provide insights for building more robust networks.

\section*{Acknowledgements}
We thank Mireille El Gheche and Stefano D\textquotesingle Aronco for the fruitful discussions. This work has been supported by a Google Faculty Research Award, and the Hasler Foundation, Switzerland, in the framework of the ROBERT project. We also gratefully acknowledge the support of NVIDIA Corporation with the donation of the GTX Titan X GPU used for this research.
\newpage
{\small
\bibliographystyle{ieee}
\bibliography{SpaseFool}
}

\newpage
\onecolumn

\begin{appendices}

\section{SparseFool adversarial examples}
\begin{figure}[ht]
\centering
\begin{subfigure}[b]{0.33\linewidth}
\captionsetup[subfigure]{skip=1pt, font=scriptsize}
\centering
    \begin{subfigure}[b]{0.49\linewidth}
        \includegraphics[width=\linewidth]{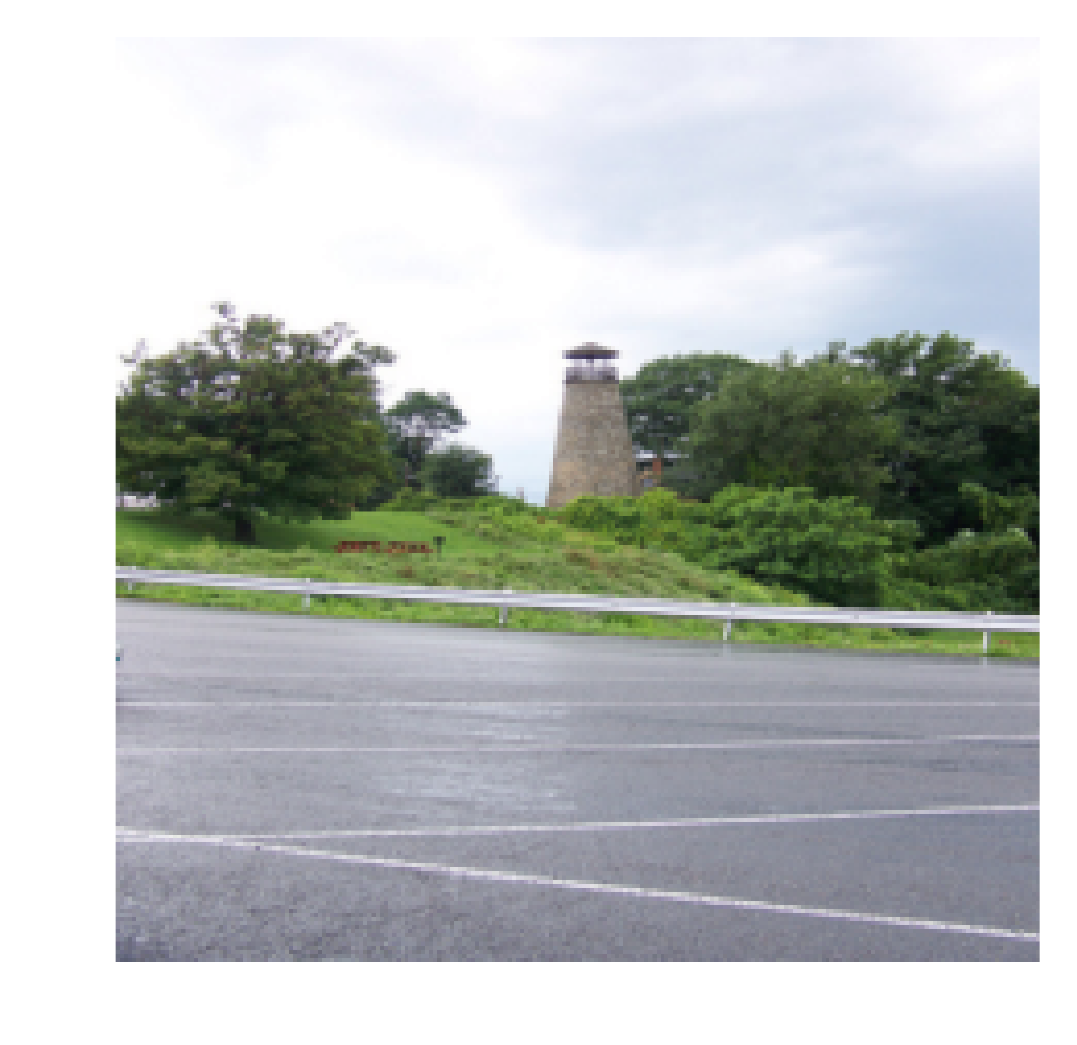}\!
        \caption*{beacon / go-kart (1)}
    \end{subfigure}
    \begin{subfigure}[b]{0.49\linewidth}
        \includegraphics[width=\linewidth]{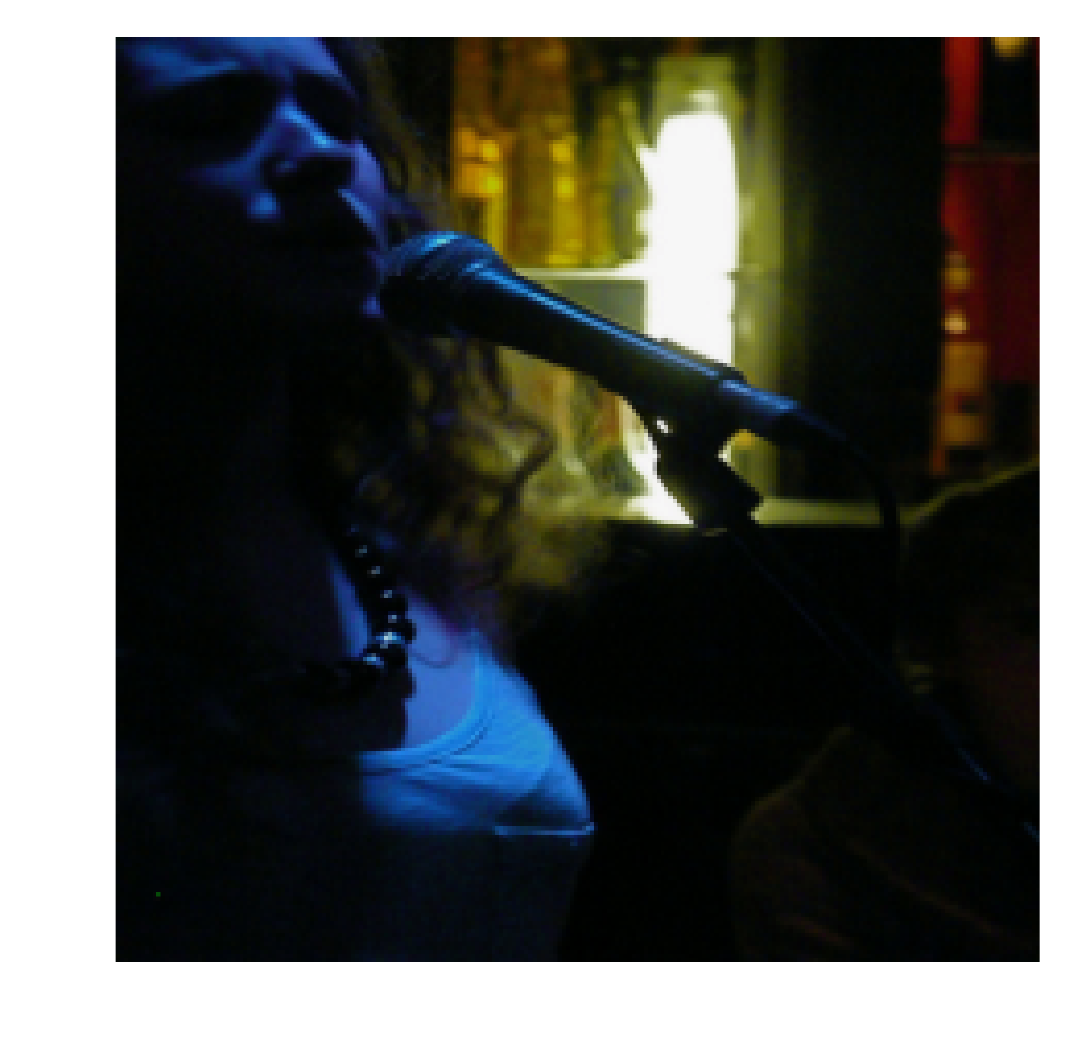}\!
        \caption*{microphone / scuba diver (1)}
    \end{subfigure}
    
    \begin{subfigure}[b]{0.49\linewidth}
        \includegraphics[width=\linewidth]{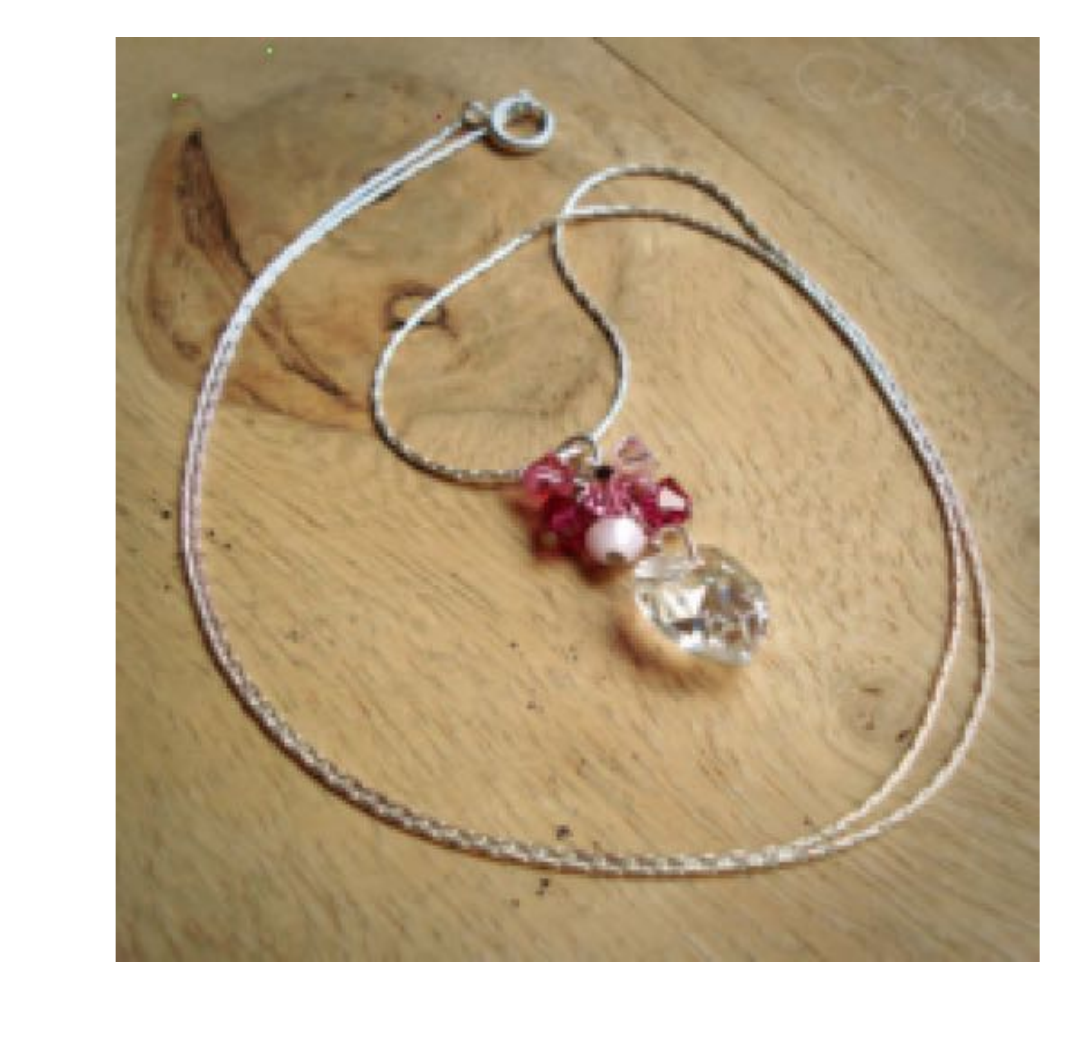}\!
        \caption*{necklace / safety pin (4)}
    \end{subfigure}
    \begin{subfigure}[b]{0.49\linewidth}
        \includegraphics[width=\linewidth]{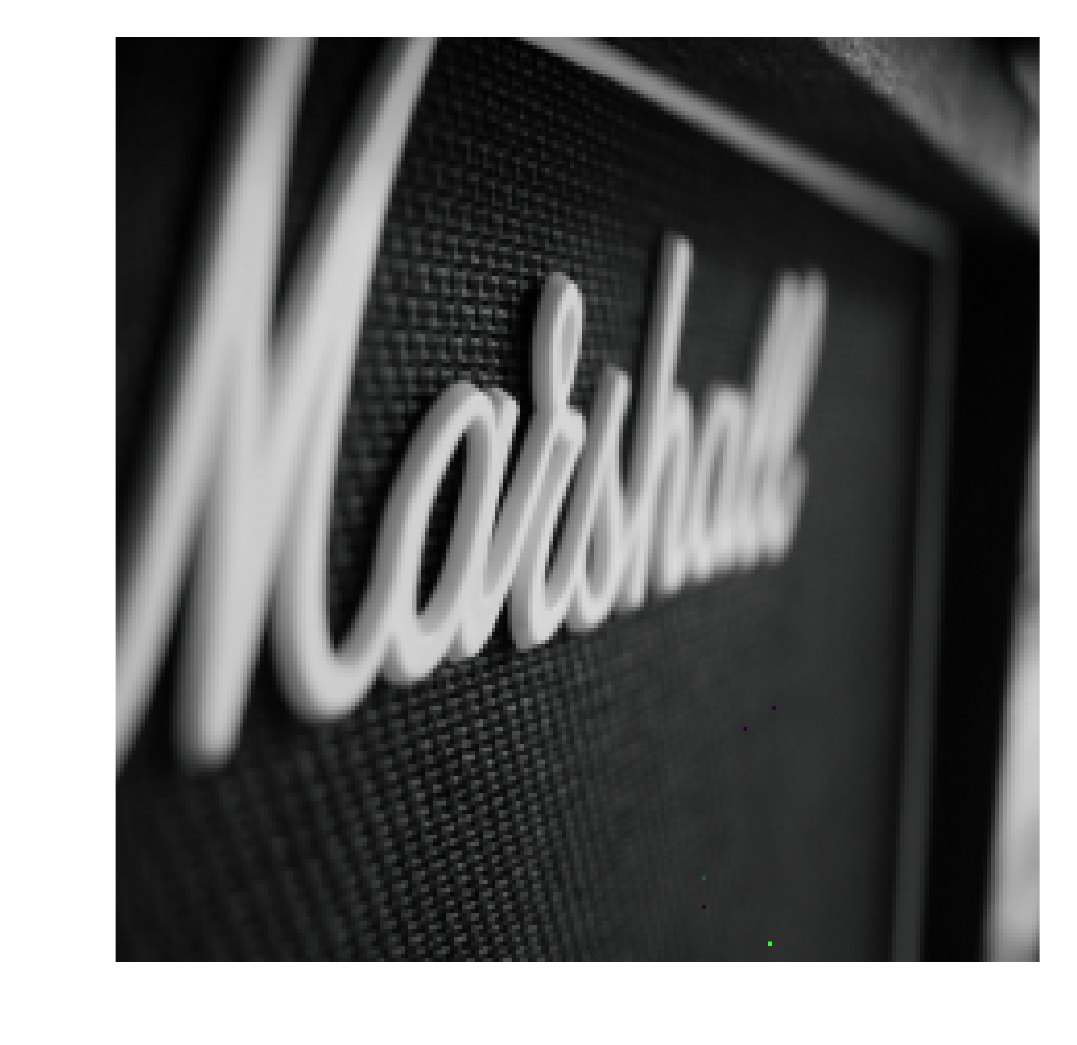}\!
        \caption*{loudspeaker / lens cap (5)}
    \end{subfigure}
    
    \begin{subfigure}[b]{0.49\linewidth}
        \includegraphics[width=\linewidth]{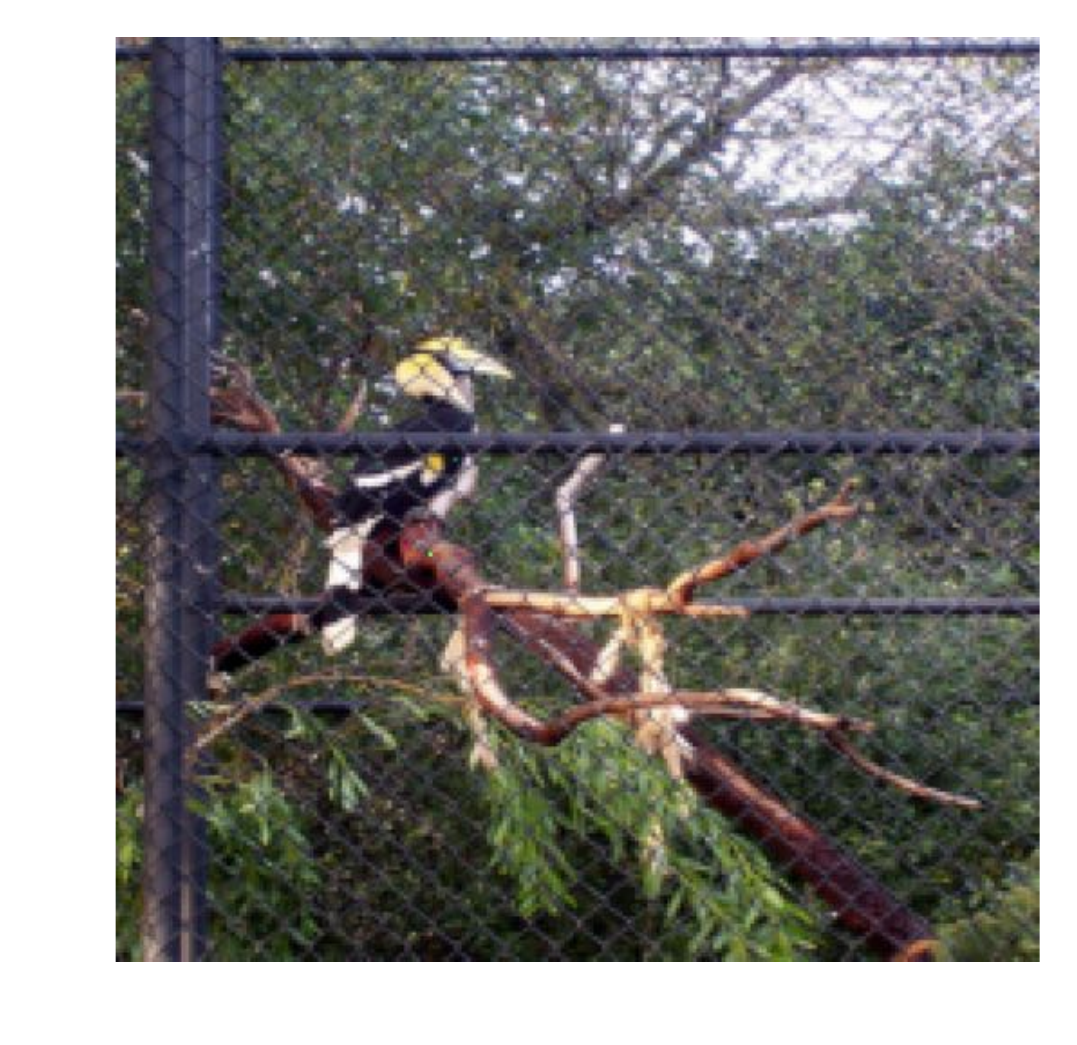}\!
        \caption*{toucan / spider (5)}
    \end{subfigure}
    \begin{subfigure}[b]{0.49\linewidth}
        \includegraphics[width=\linewidth]{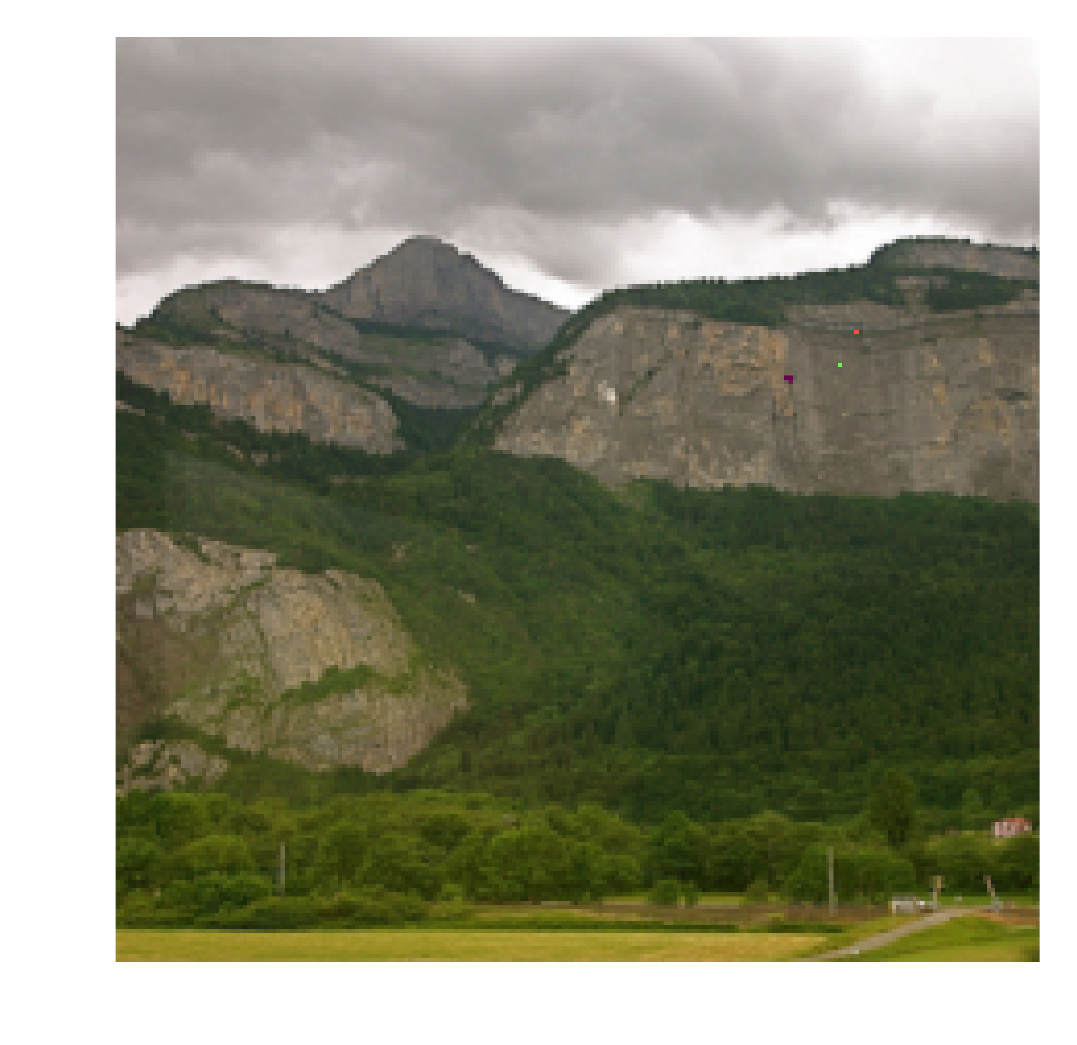}\!
        \caption*{valley / castle (5)}
    \end{subfigure}
\caption{Highly sparse perturbations.}
\end{subfigure}\hfill
\begin{subfigure}[b]{0.33\linewidth}
\captionsetup[subfigure]{skip=1pt, font=scriptsize}
\centering
    \begin{subfigure}[b]{0.49\linewidth}
        \includegraphics[width=\linewidth]{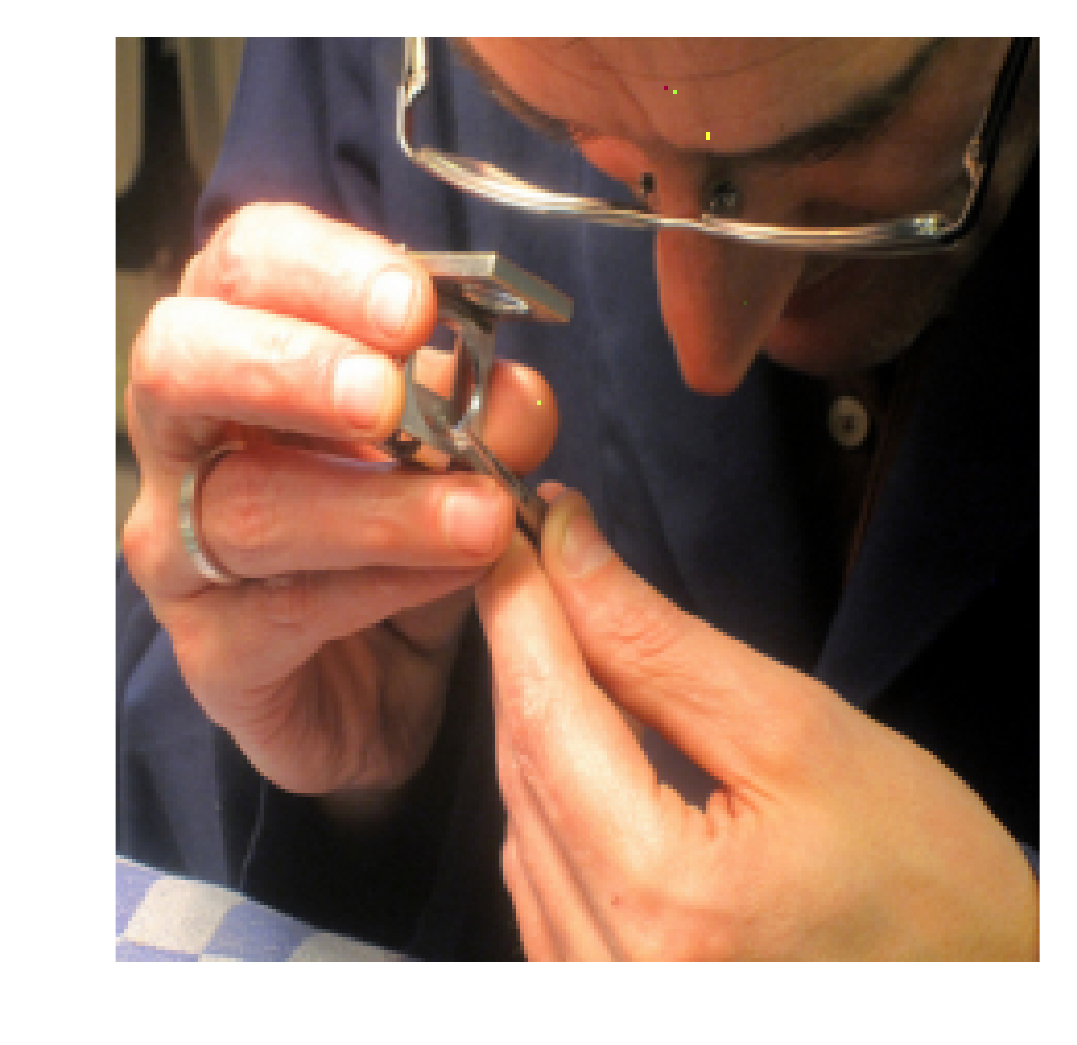}\!
        \caption*{loupe / violin (6)}
    \end{subfigure}
    \begin{subfigure}[b]{0.49\linewidth}
        \includegraphics[width=\linewidth]{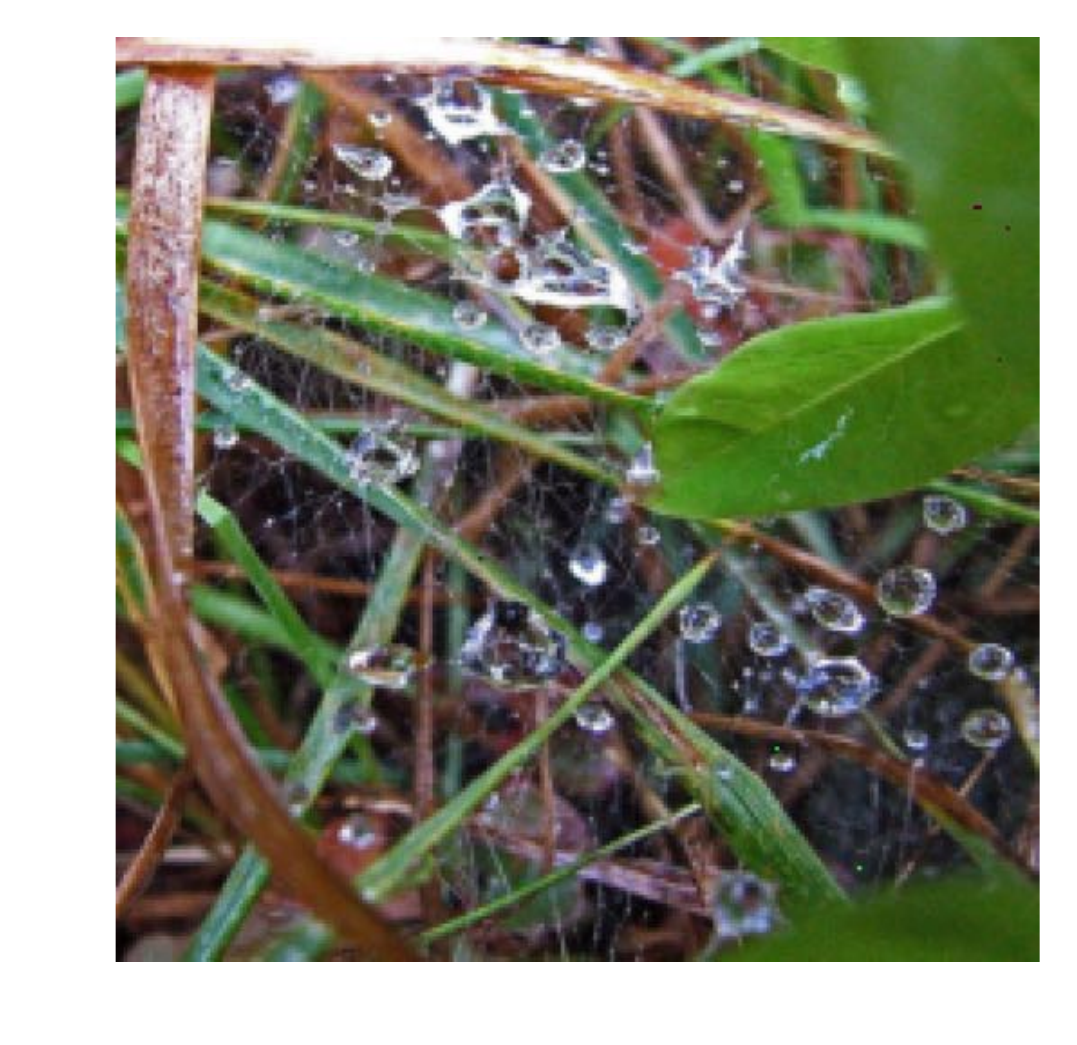}\!
        \caption*{web / walking-stick (7)}
    \end{subfigure}
    
    \begin{subfigure}[b]{0.49\linewidth}
        \includegraphics[width=\linewidth]{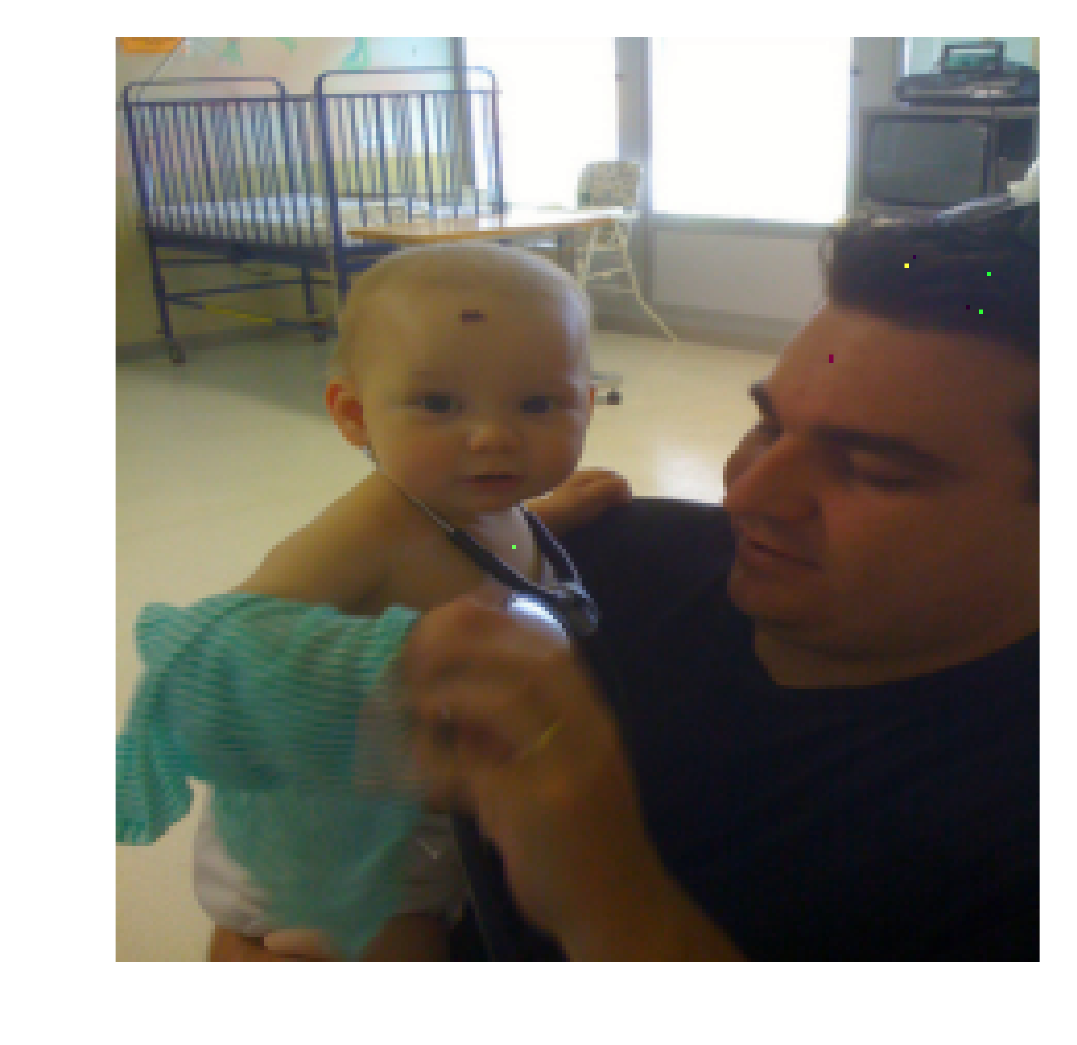}\!
        \caption*{stethoscope / piggy-bank (8)}
    \end{subfigure}
    \begin{subfigure}[b]{0.49\linewidth}
        \includegraphics[width=\linewidth]{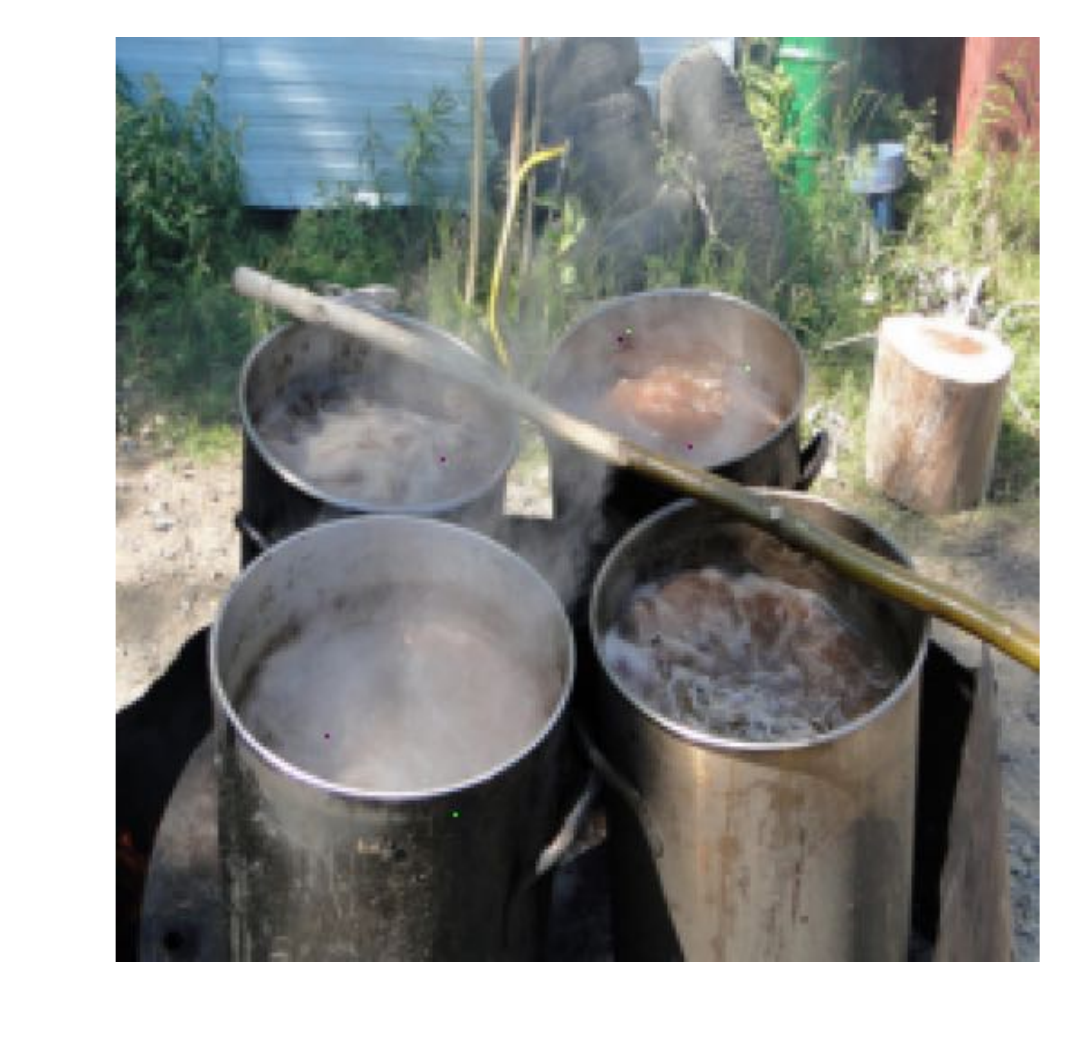}\!
        \caption*{caldron / steel drum (9)}
    \end{subfigure}
    
    \begin{subfigure}[b]{0.49\linewidth}
        \includegraphics[width=\linewidth]{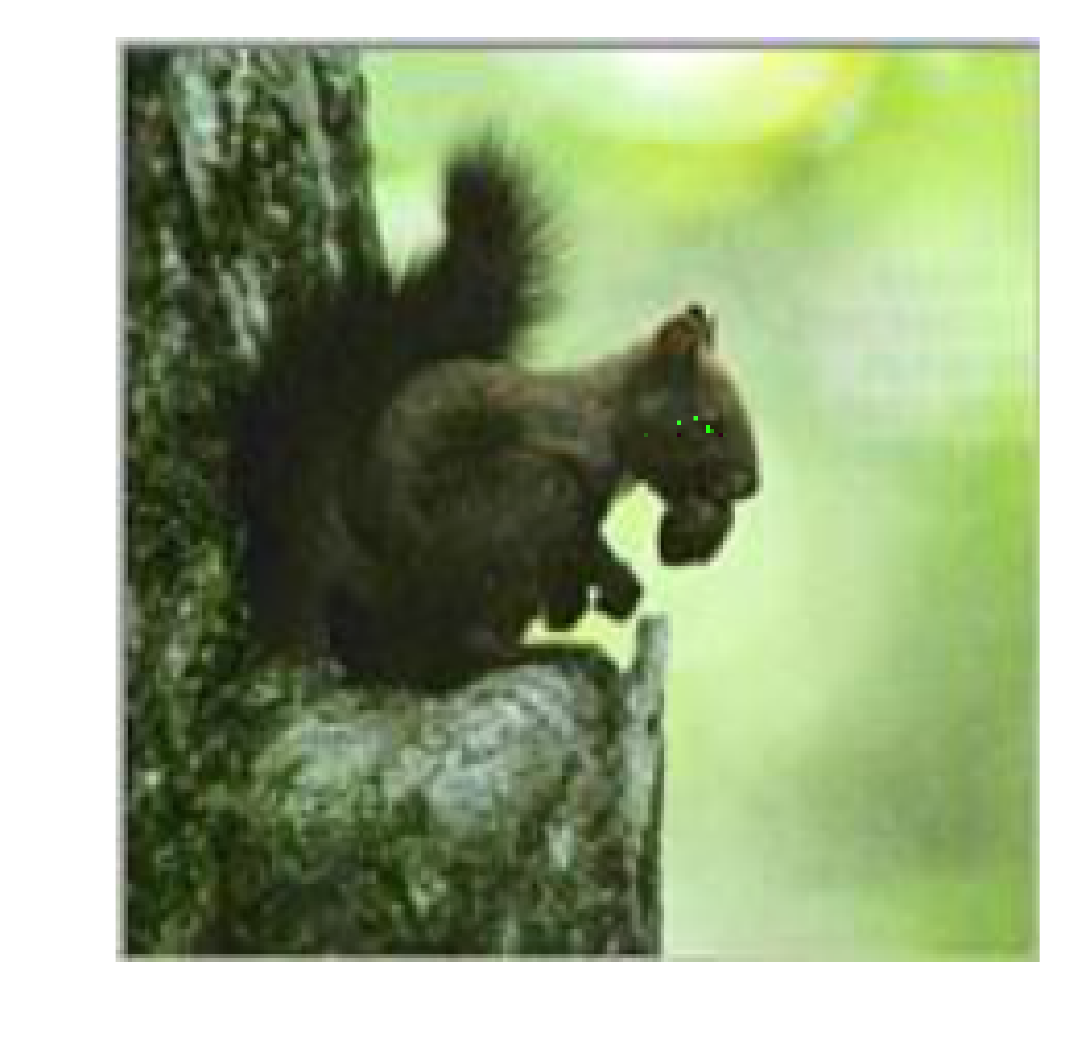}\!
        \caption*{fox-squirrel / indri (9)}
    \end{subfigure}
    \begin{subfigure}[b]{0.49\linewidth}
        \includegraphics[width=\linewidth]{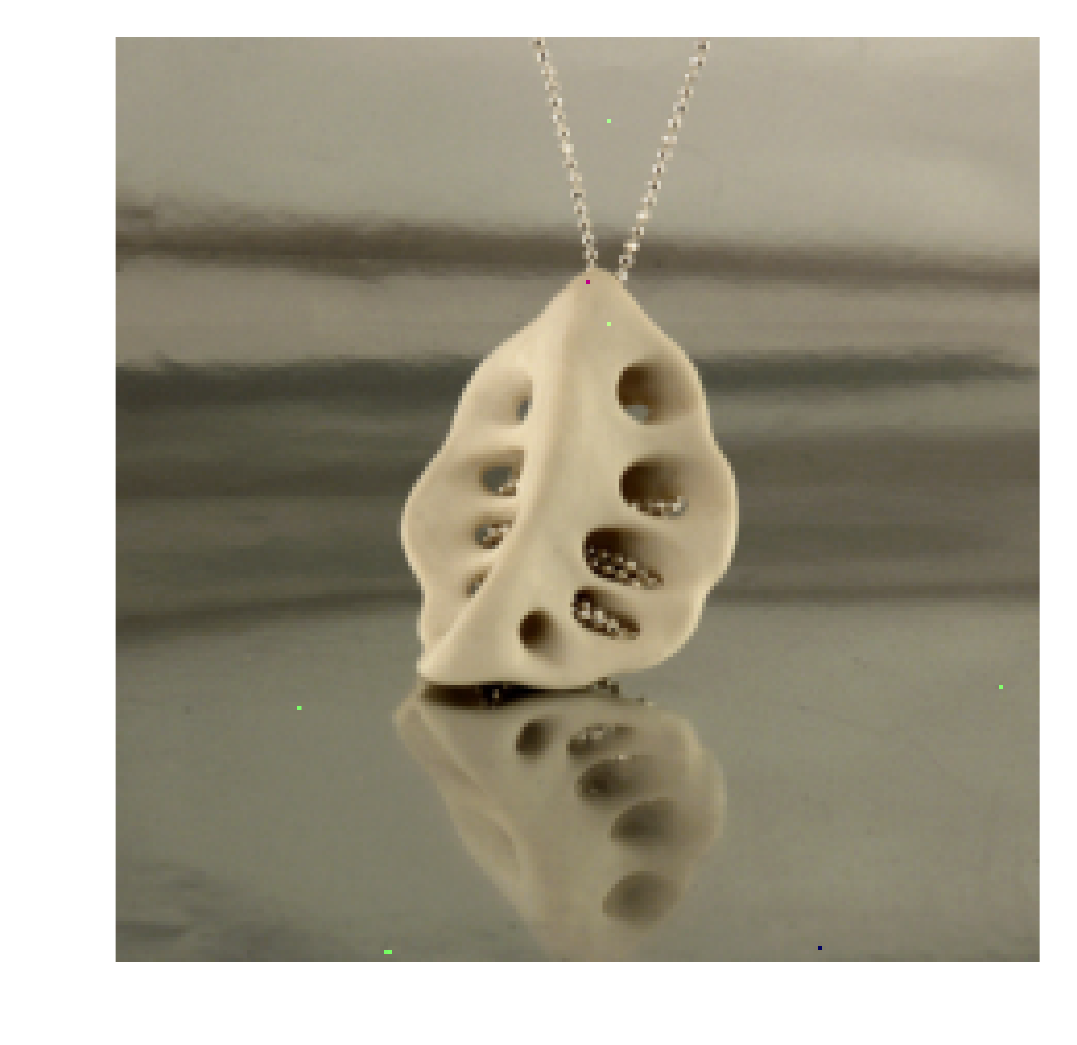}\!
        \caption*{necklace / chime (9)}
    \end{subfigure}
\caption{Highly sparse perturbations.}
\end{subfigure}\hfill
\begin{subfigure}[b]{0.33\linewidth}
\captionsetup[subfigure]{skip=1pt, font=scriptsize}
\centering
    \begin{subfigure}[b]{0.49\linewidth}
        \includegraphics[width=\linewidth]{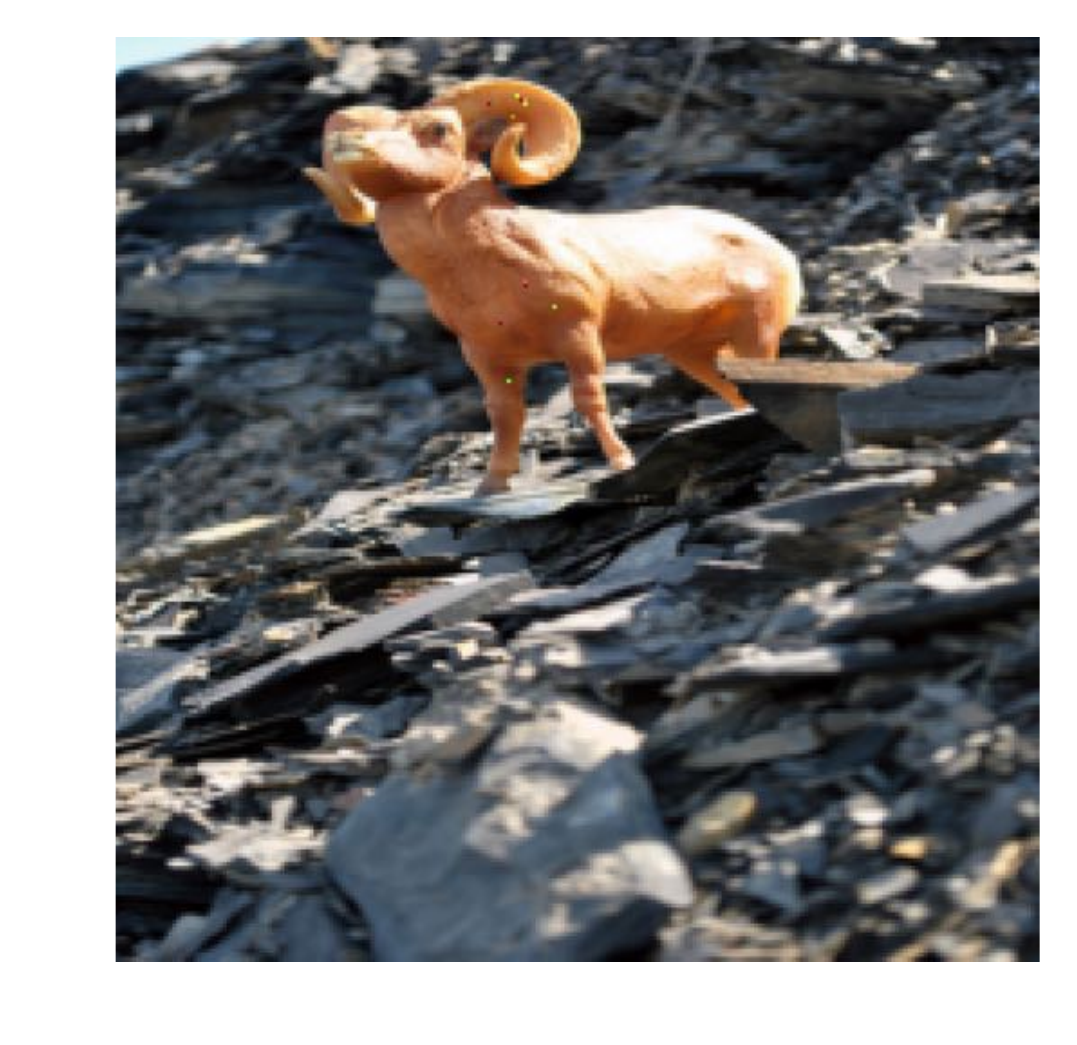}\!
        \caption*{ram / conch (11)}
    \end{subfigure}
    \begin{subfigure}[b]{0.49\linewidth}
        \includegraphics[width=\linewidth]{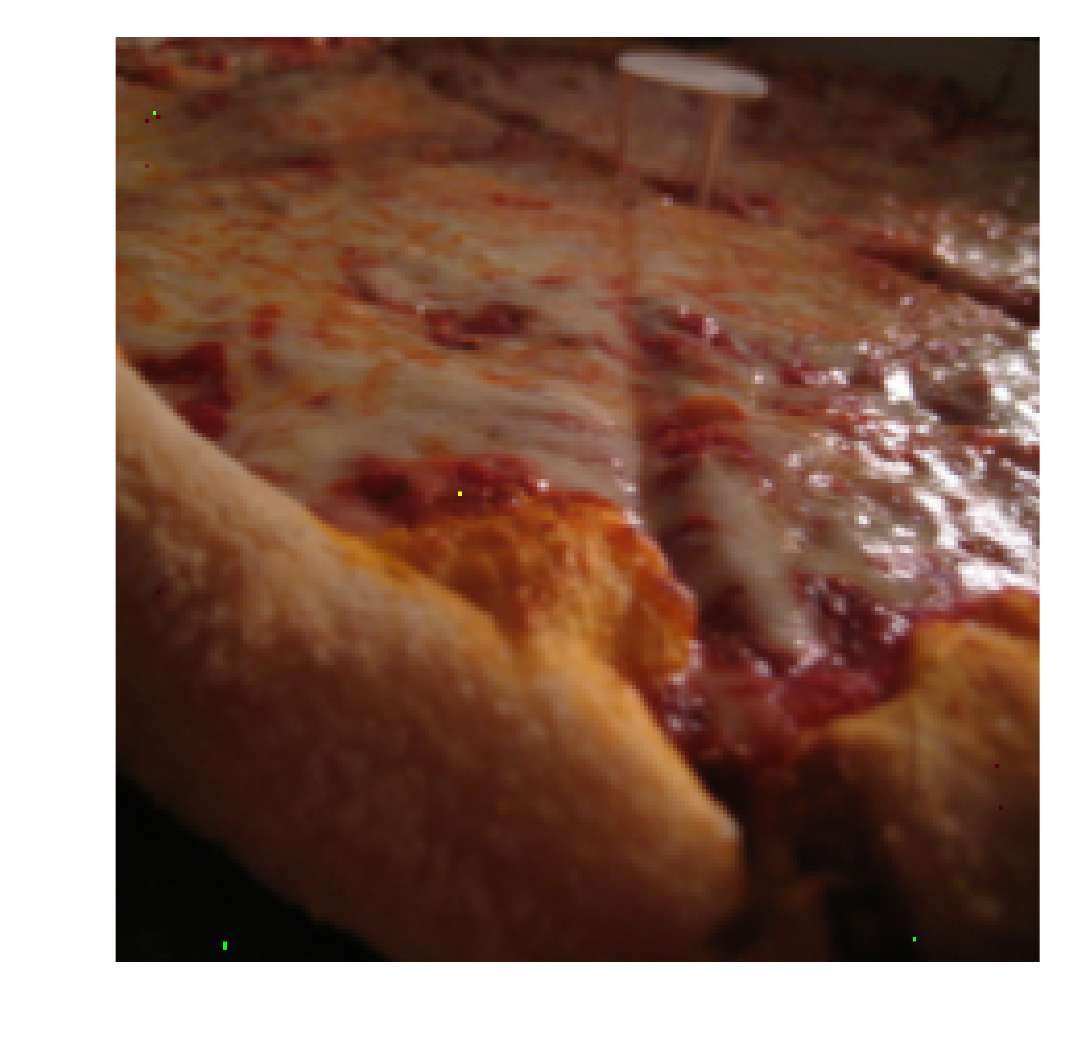}\!
        \caption*{pizza / hot-dog (13)}
    \end{subfigure}
    
    \begin{subfigure}[b]{0.49\linewidth}
        \includegraphics[width=\linewidth]{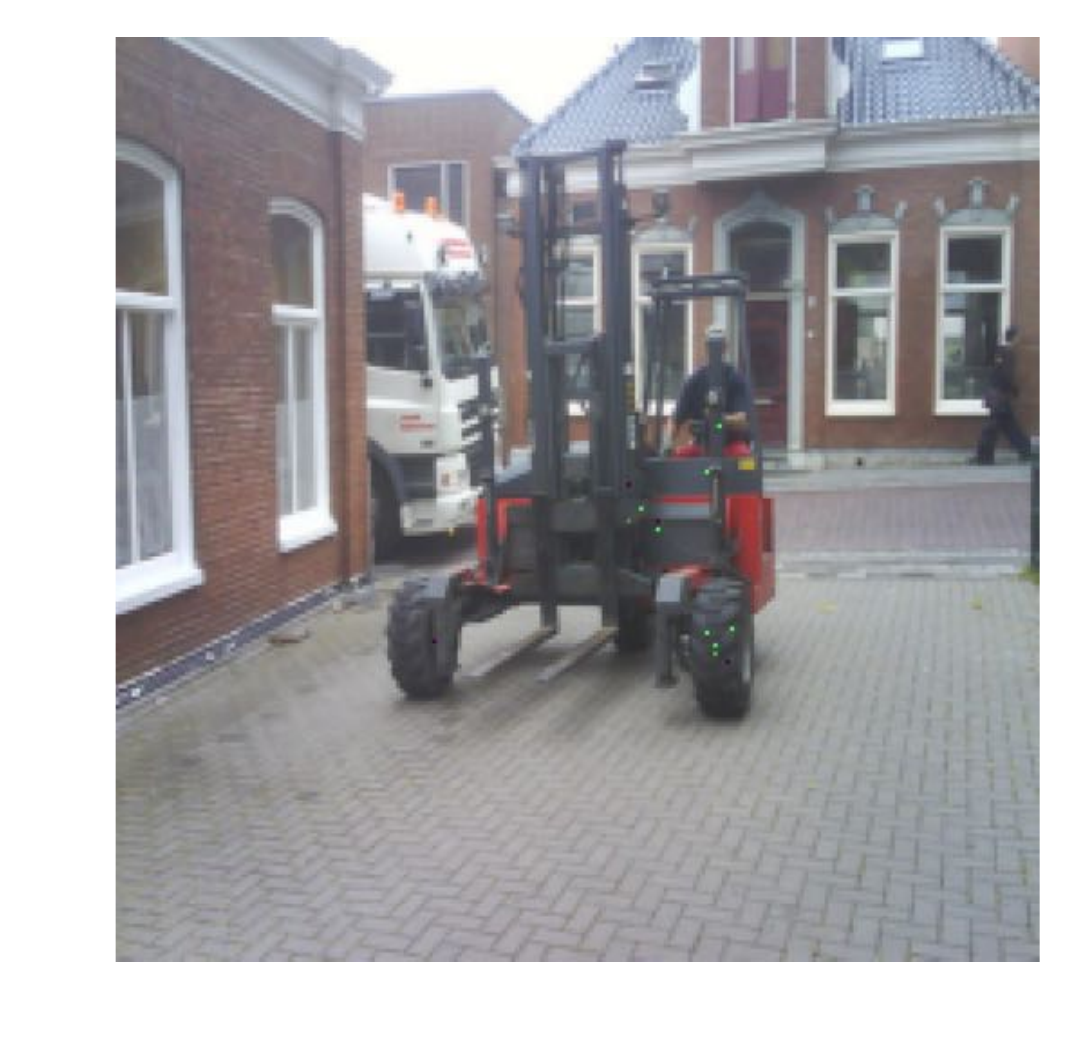}\!
        \caption*{forklift / scooter (14)}
    \end{subfigure}
    \begin{subfigure}[b]{0.49\linewidth}
        \includegraphics[width=\linewidth]{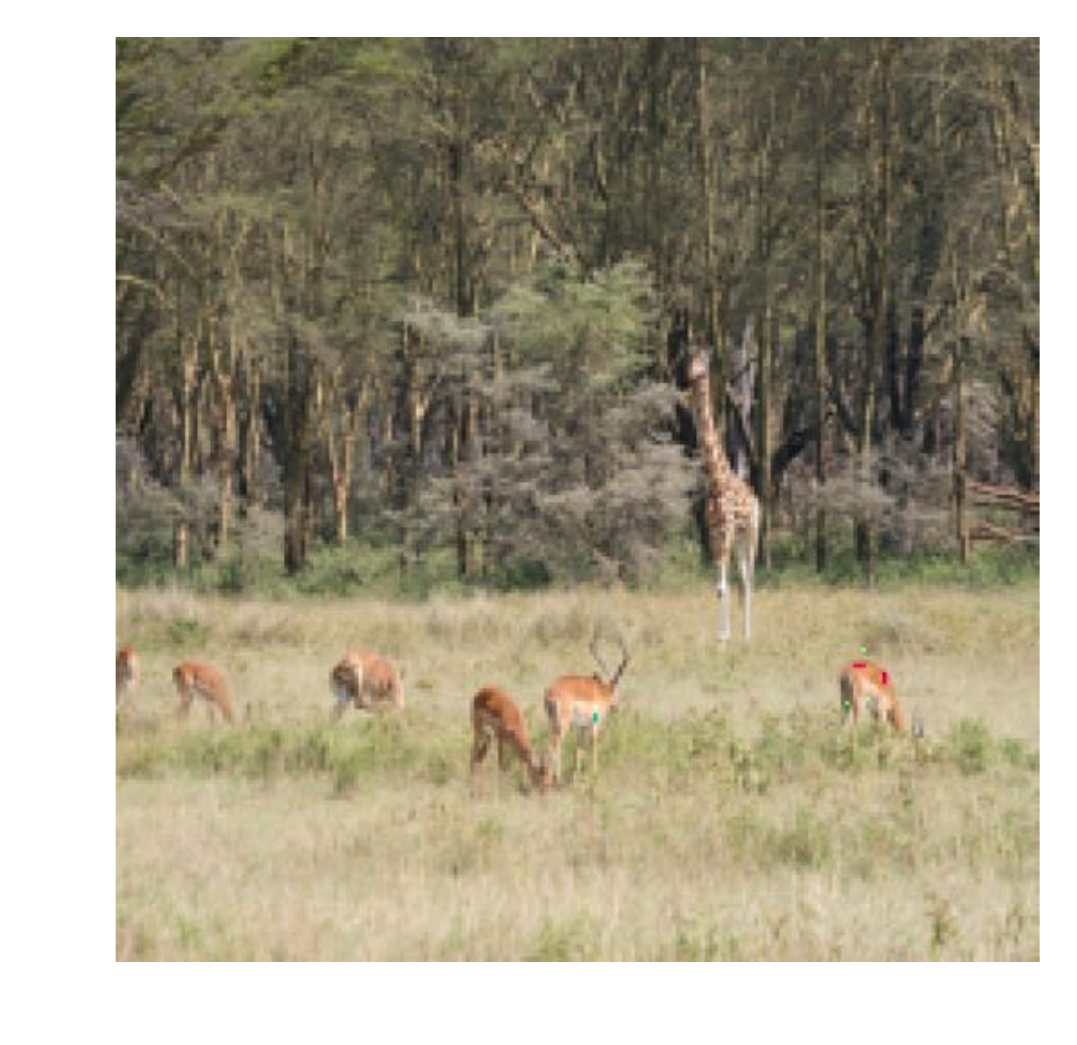}\!
        \caption*{impala / crane (15)}
    \end{subfigure}
    
    \begin{subfigure}[b]{0.49\linewidth}
        \includegraphics[width=\linewidth]{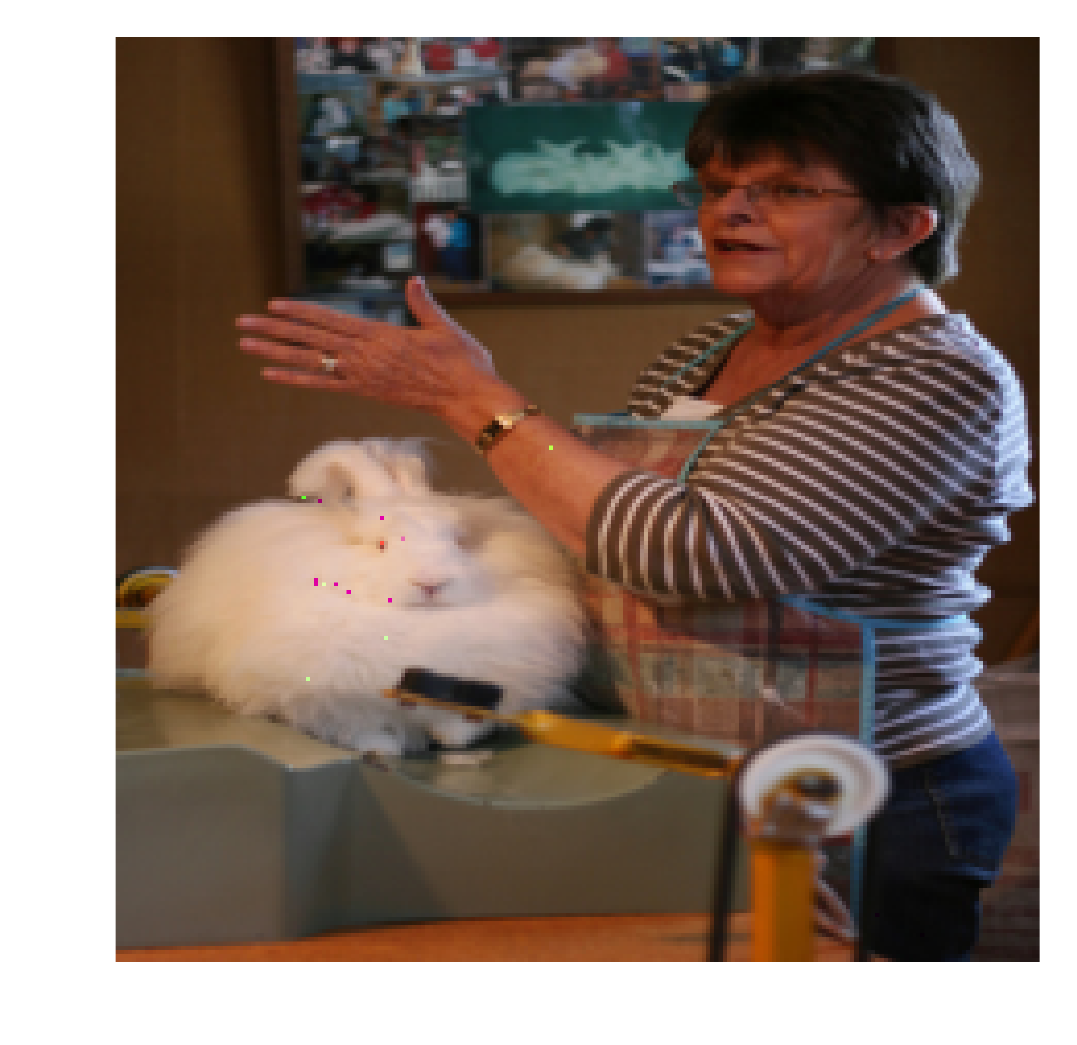}\!
        \caption*{angora / bassoon (16)}
    \end{subfigure}
    \begin{subfigure}[b]{0.49\linewidth}
        \includegraphics[width=\linewidth]{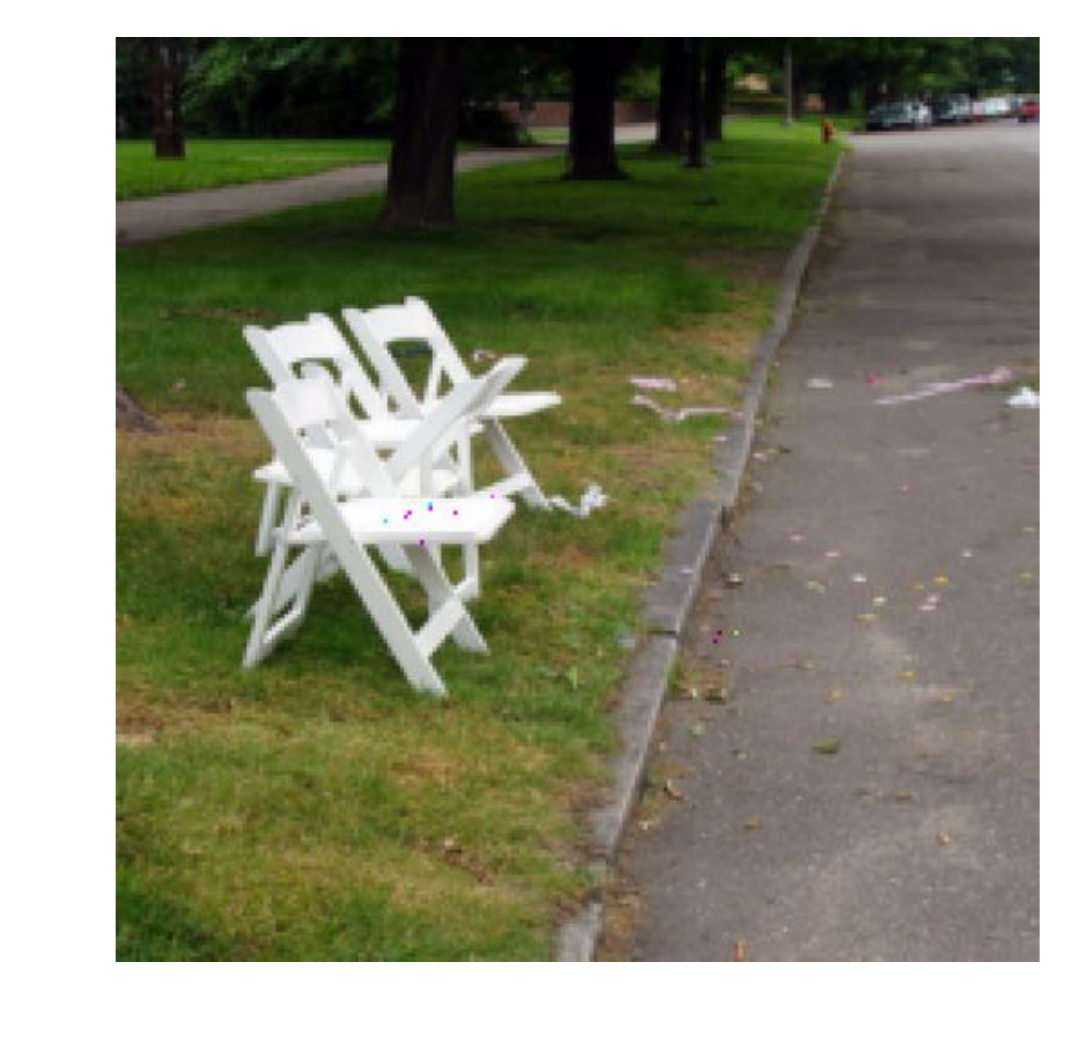}\!
        \caption*{chair / warm fence (18)}
    \end{subfigure}
\caption{Highly sparse perturbations.}
\end{subfigure}\hfill
\caption{SparseFool adversarial examples for the ImageNet dataset for different levels of sparsity. The predicted label is shown above the image, the fooling one below, and the number of perturbed pixels is written inside the parentheses.}
\label{fig:imnet_visual}
\end{figure}

\begin{figure}[ht]
\centering
\begin{subfigure}[b]{0.3\linewidth}
\captionsetup[subfigure]{skip=1pt}
    \begin{subfigure}[b]{0.20\linewidth}
        \includegraphics[width=\linewidth]{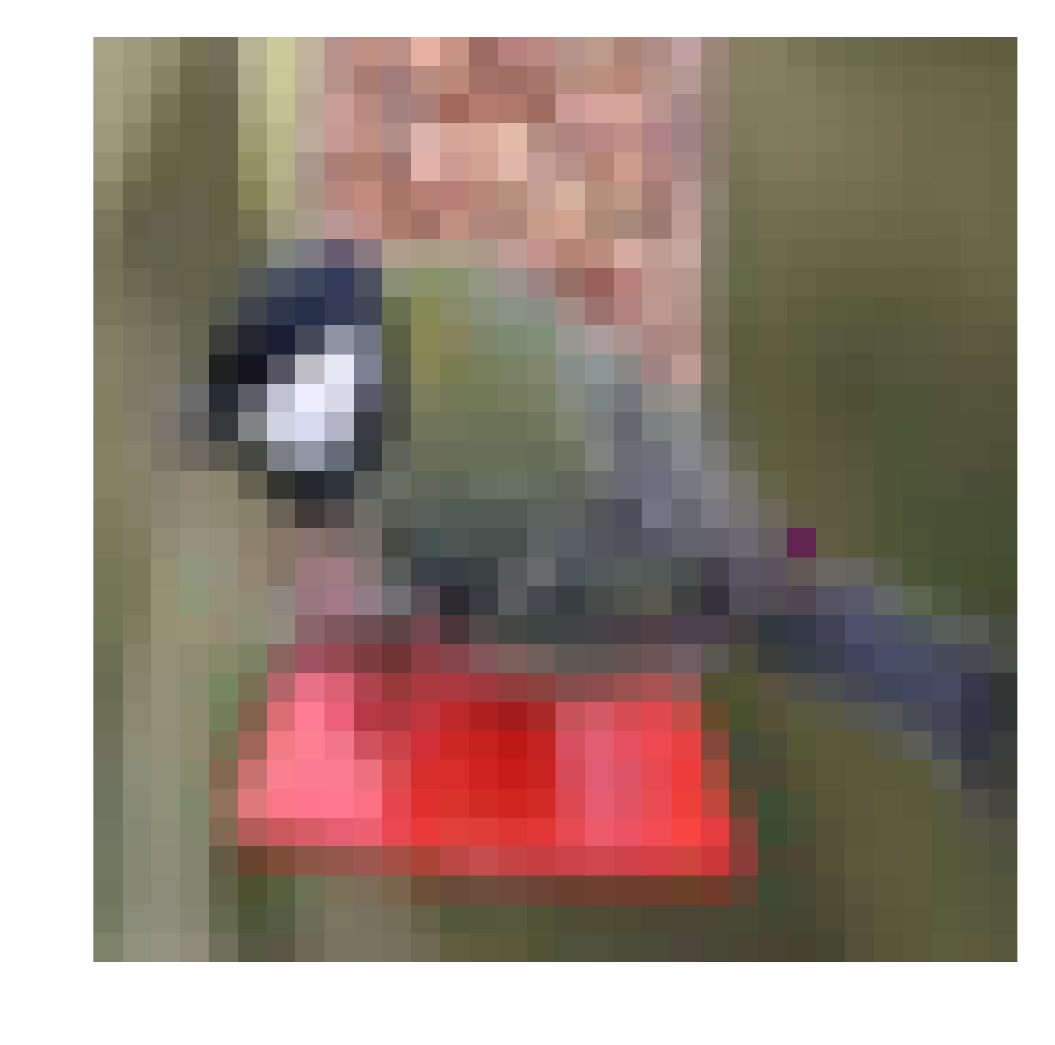}\!
        \captionsetup{font=scriptsize}
        \caption*{car (1)}
    \end{subfigure}\!
    \begin{subfigure}[b]{0.20\linewidth}
        \includegraphics[width=\linewidth]{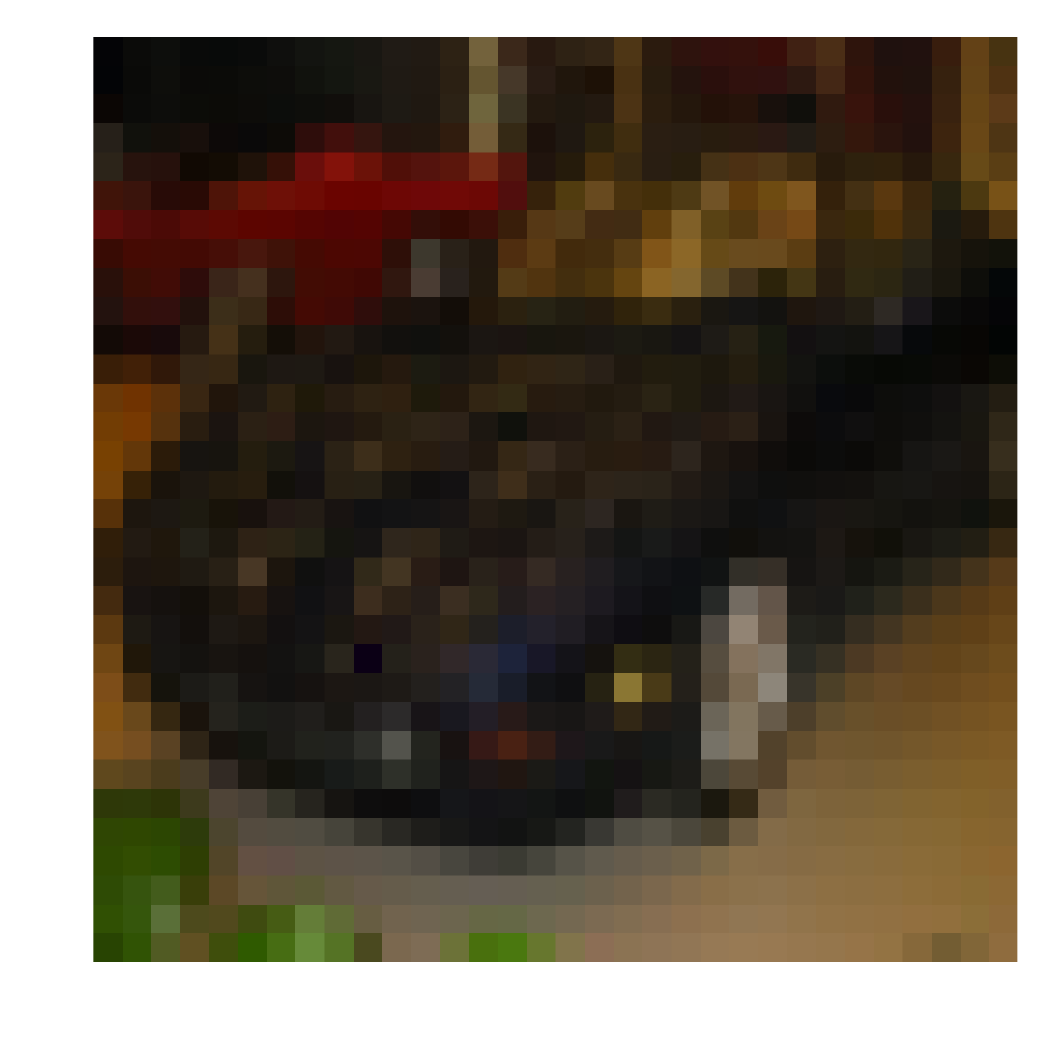}\!
        \captionsetup{font=scriptsize}
        \caption*{frog (1)}
    \end{subfigure}\!
    \begin{subfigure}[b]{0.20\linewidth}
        \includegraphics[width=\linewidth]{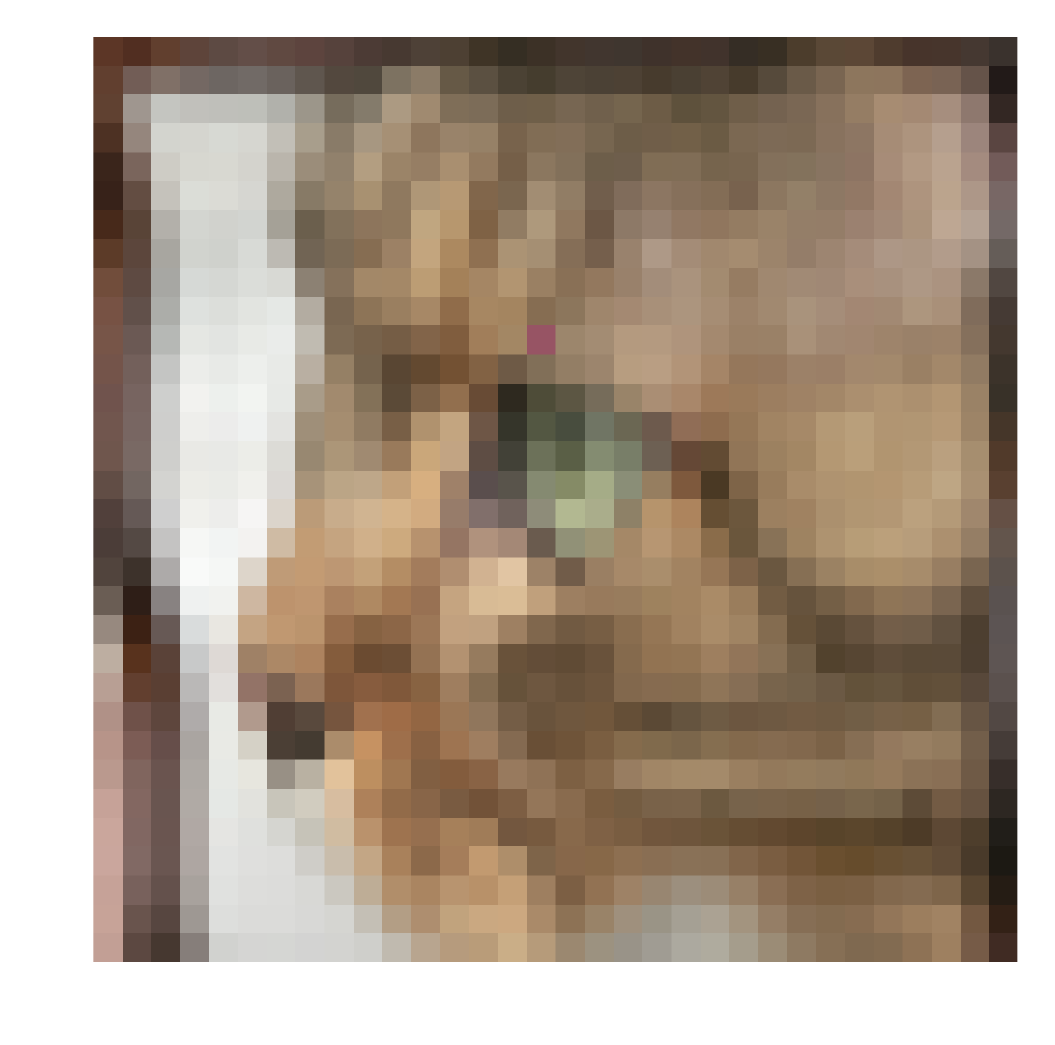}\!
        \captionsetup{font=scriptsize}
        \caption*{dog (1)}
    \end{subfigure}\!
    \begin{subfigure}[b]{0.20\linewidth}
        \includegraphics[width=\linewidth]{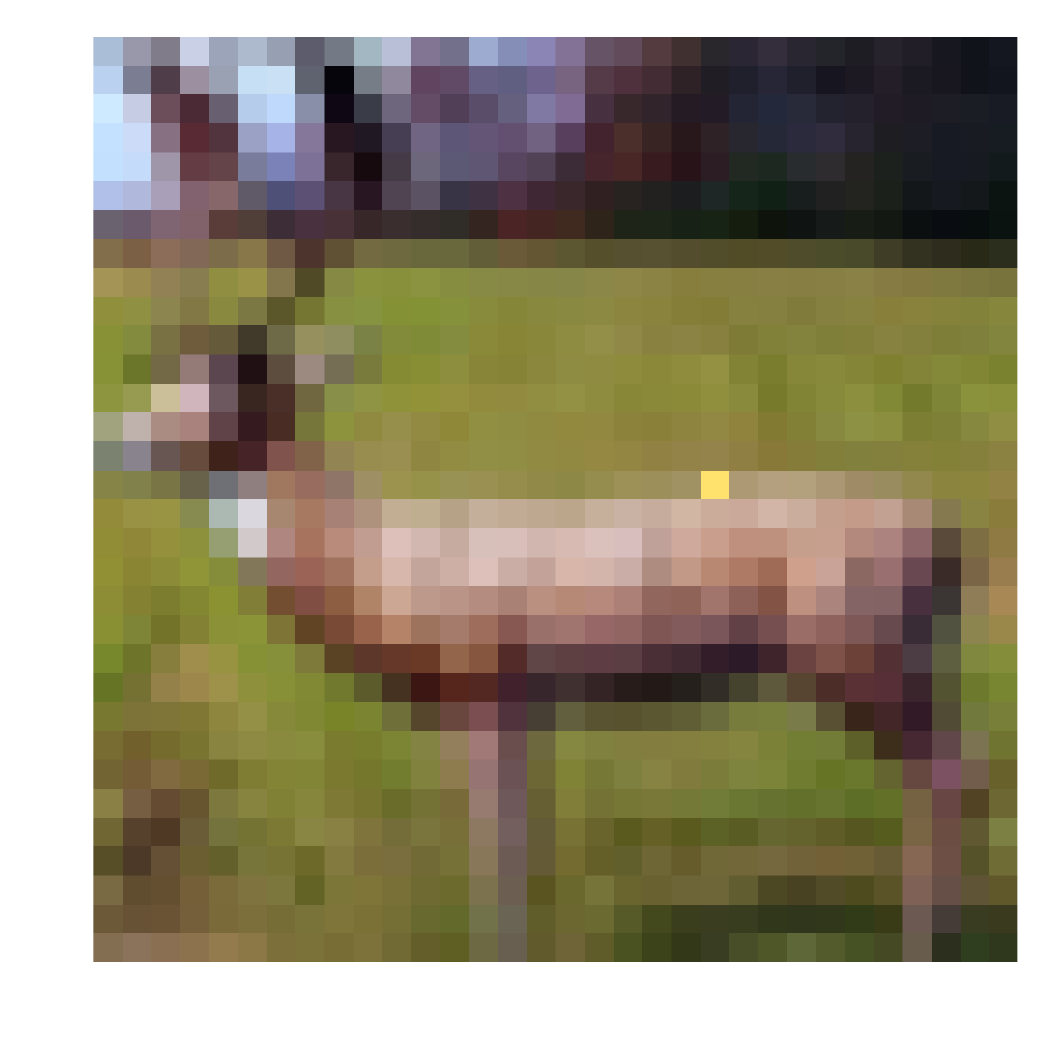}\!
        \captionsetup{font=scriptsize}
        \caption*{bird (1)}
    \end{subfigure}\!
    \begin{subfigure}[b]{0.20\linewidth}
        \includegraphics[width=\linewidth]{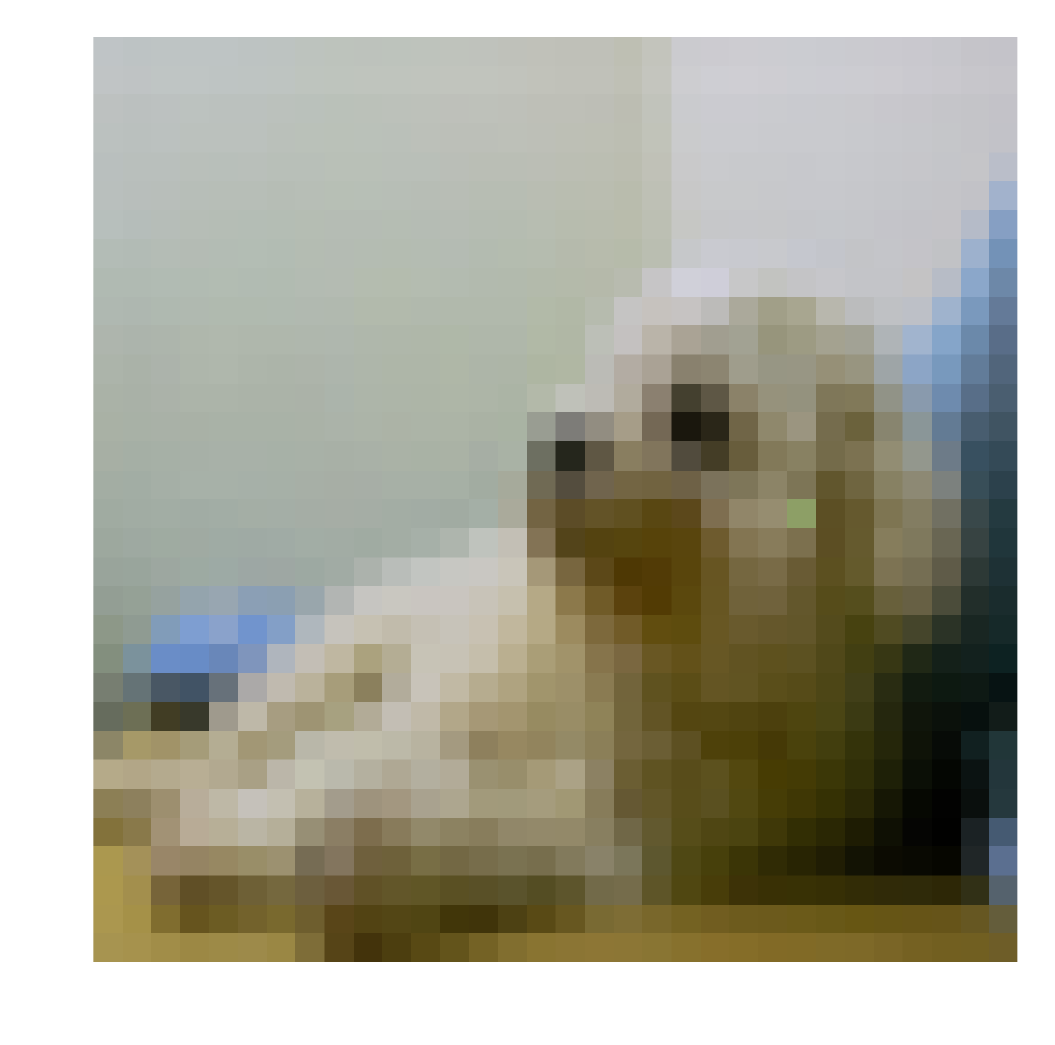}\!
        \captionsetup{font=scriptsize}
        \caption*{bird (1)}
    \end{subfigure}\!
    
    \begin{subfigure}[b]{0.20\linewidth}
        \includegraphics[width=\linewidth]{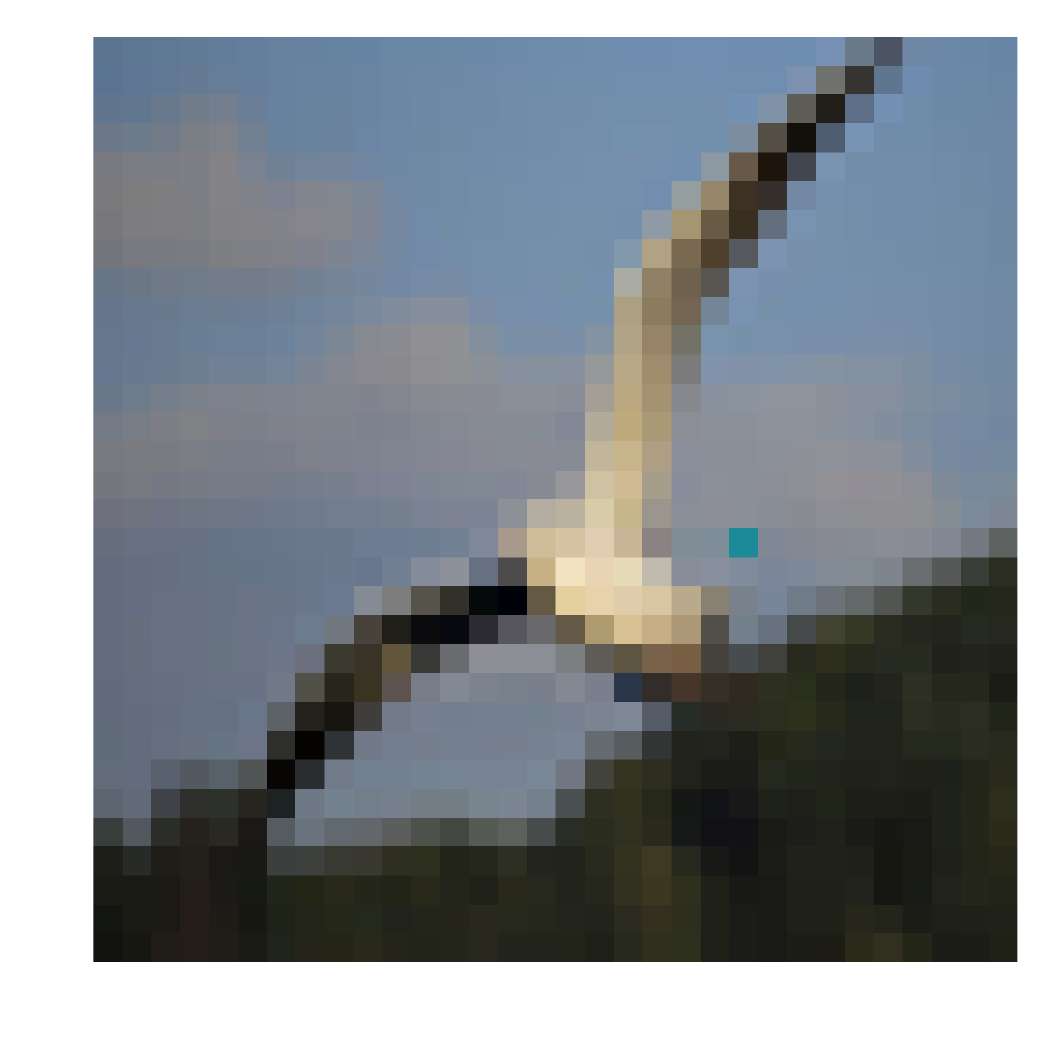}\!
        \captionsetup{font=scriptsize}
        \caption*{plane (1)}
    \end{subfigure}\!
    \begin{subfigure}[b]{0.20\linewidth}
        \includegraphics[width=\linewidth]{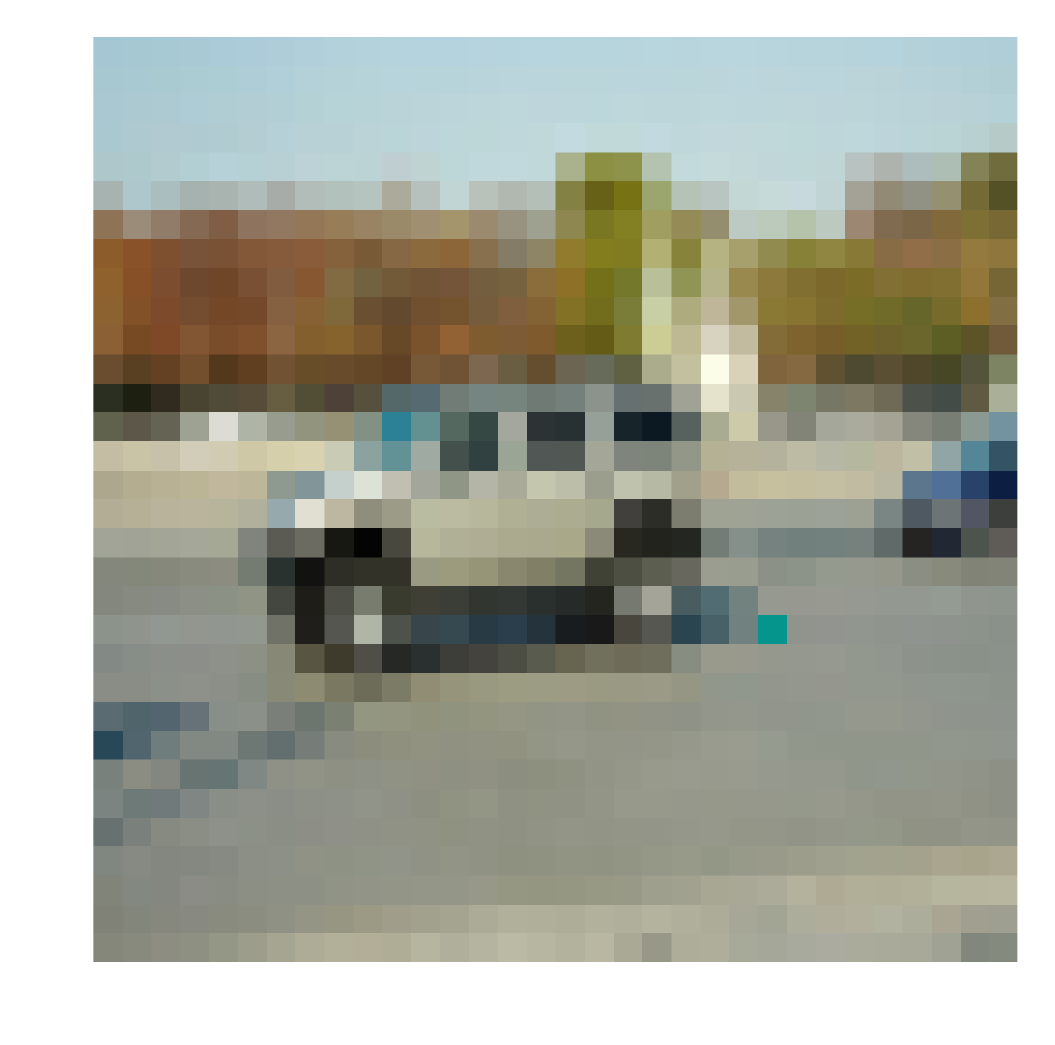}\!
        \captionsetup{font=scriptsize}
        \caption*{plane (1)}
    \end{subfigure}\!
    \begin{subfigure}[b]{0.20\linewidth}
        \includegraphics[width=\linewidth]{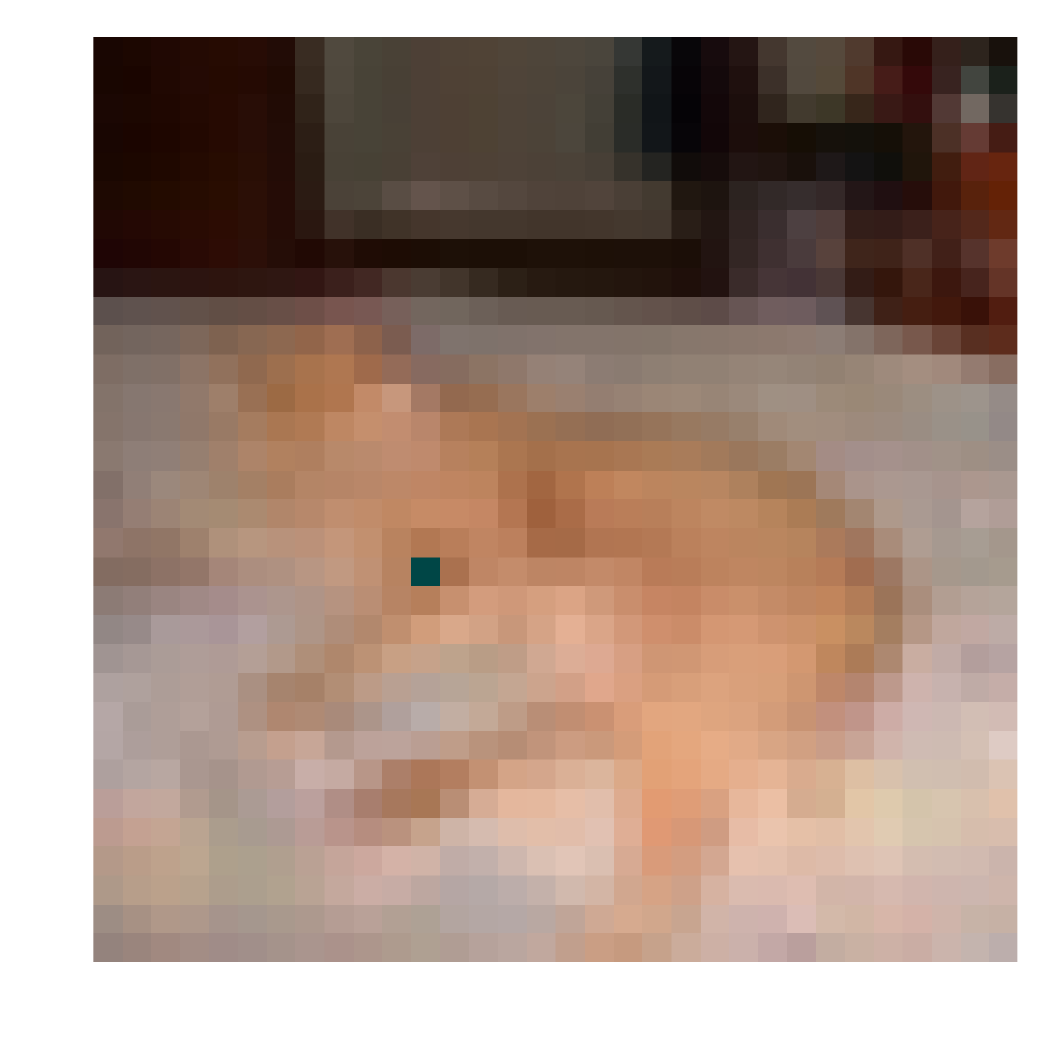}\!
        \captionsetup{font=scriptsize}
        \caption*{frog (1)}
    \end{subfigure}\!
    \begin{subfigure}[b]{0.20\linewidth}
        \includegraphics[width=\linewidth]{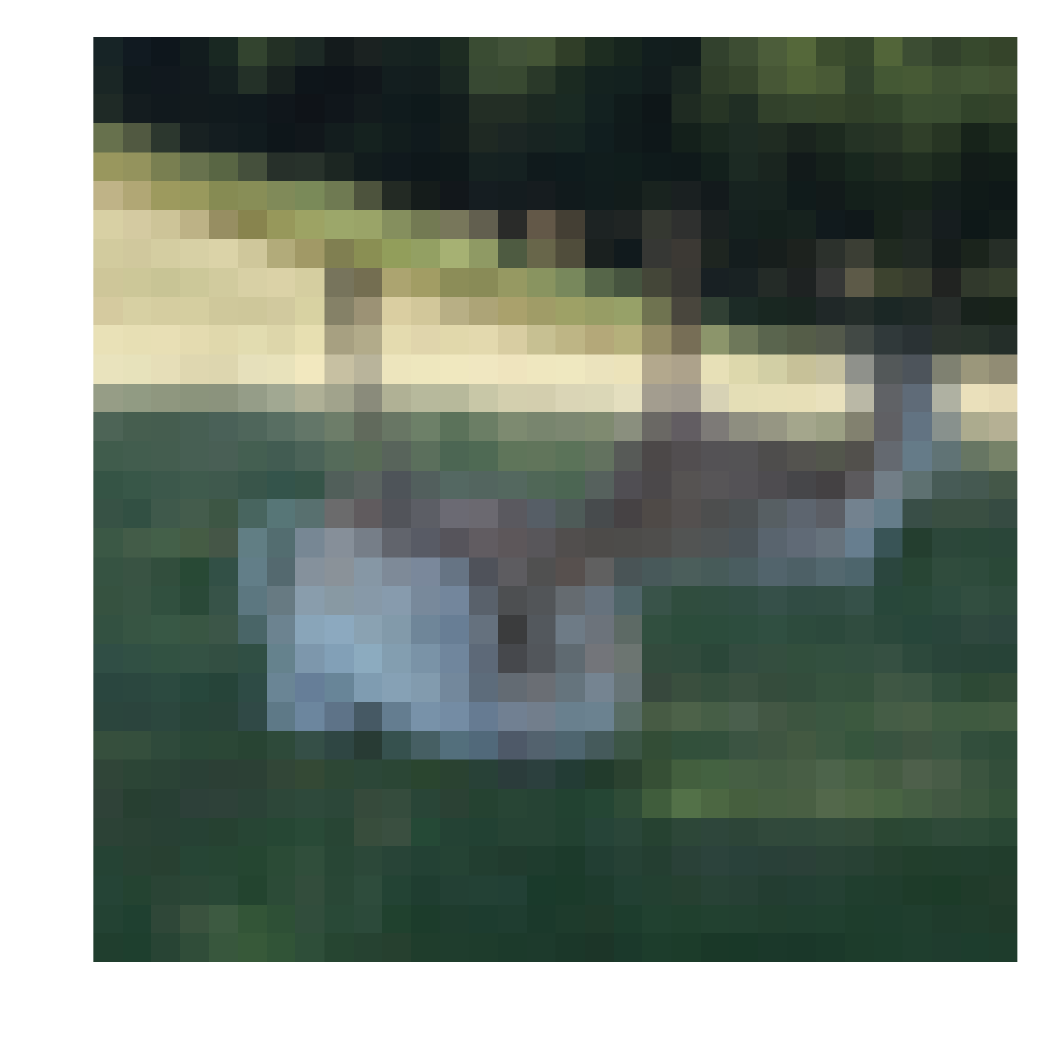}\!
        \captionsetup{font=scriptsize}
        \caption*{ship (1)}
    \end{subfigure}\!
    \begin{subfigure}[b]{0.20\linewidth}
        \includegraphics[width=\linewidth]{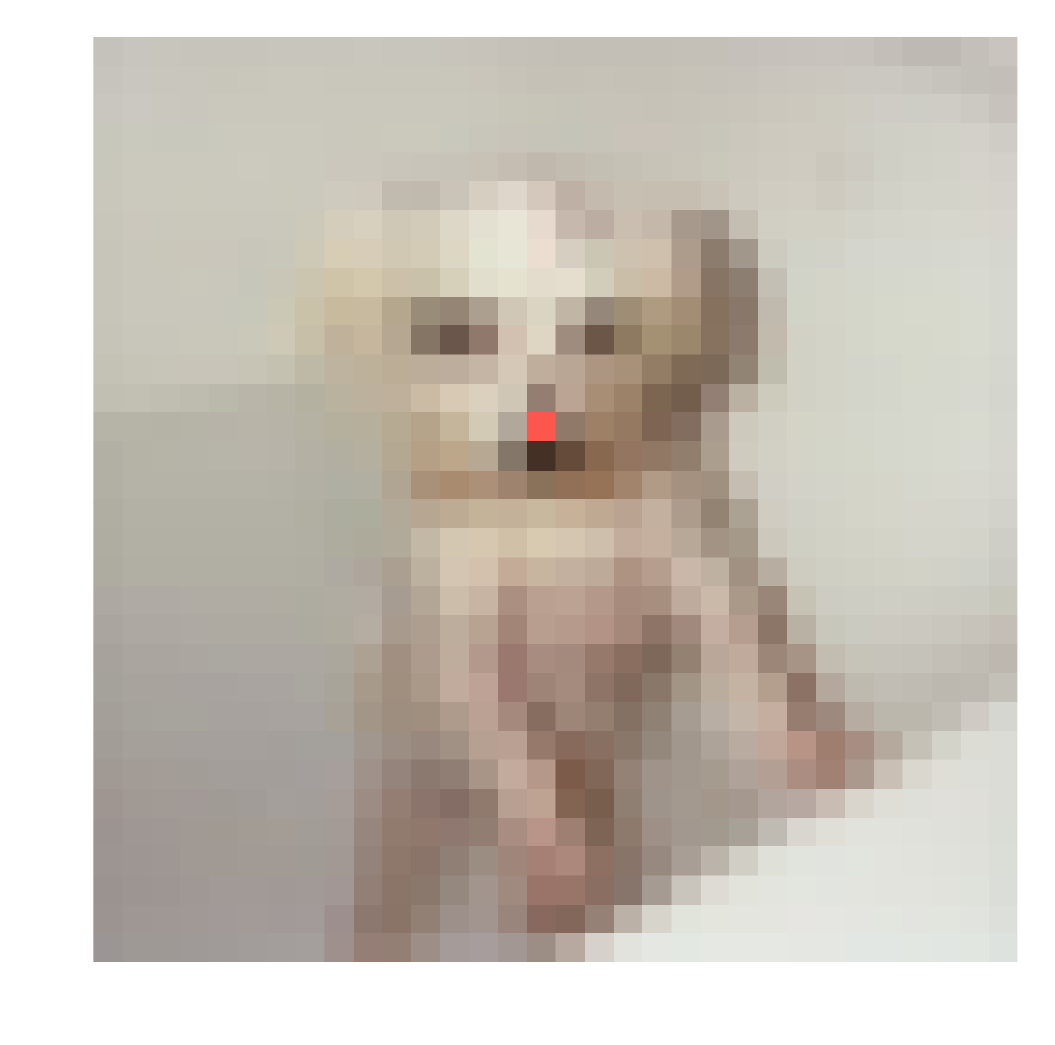}\!
        \captionsetup{font=scriptsize}
        \caption*{cat (1)}
    \end{subfigure}\!
    
    \begin{subfigure}[b]{0.20\linewidth}
        \includegraphics[width=\linewidth]{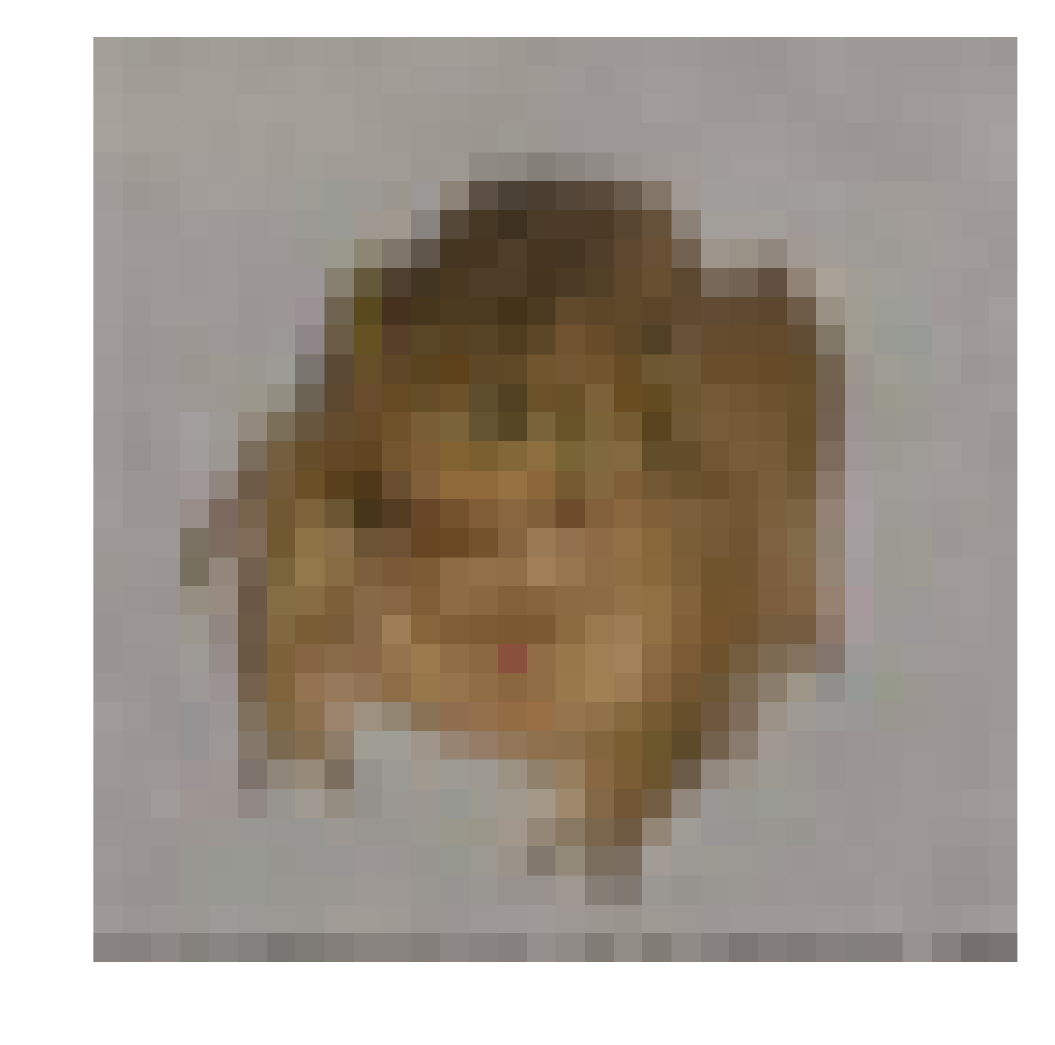}\!
        \captionsetup{font=scriptsize}
        \caption*{bird (1)}
    \end{subfigure}\!
    \begin{subfigure}[b]{0.20\linewidth}
        \includegraphics[width=\linewidth]{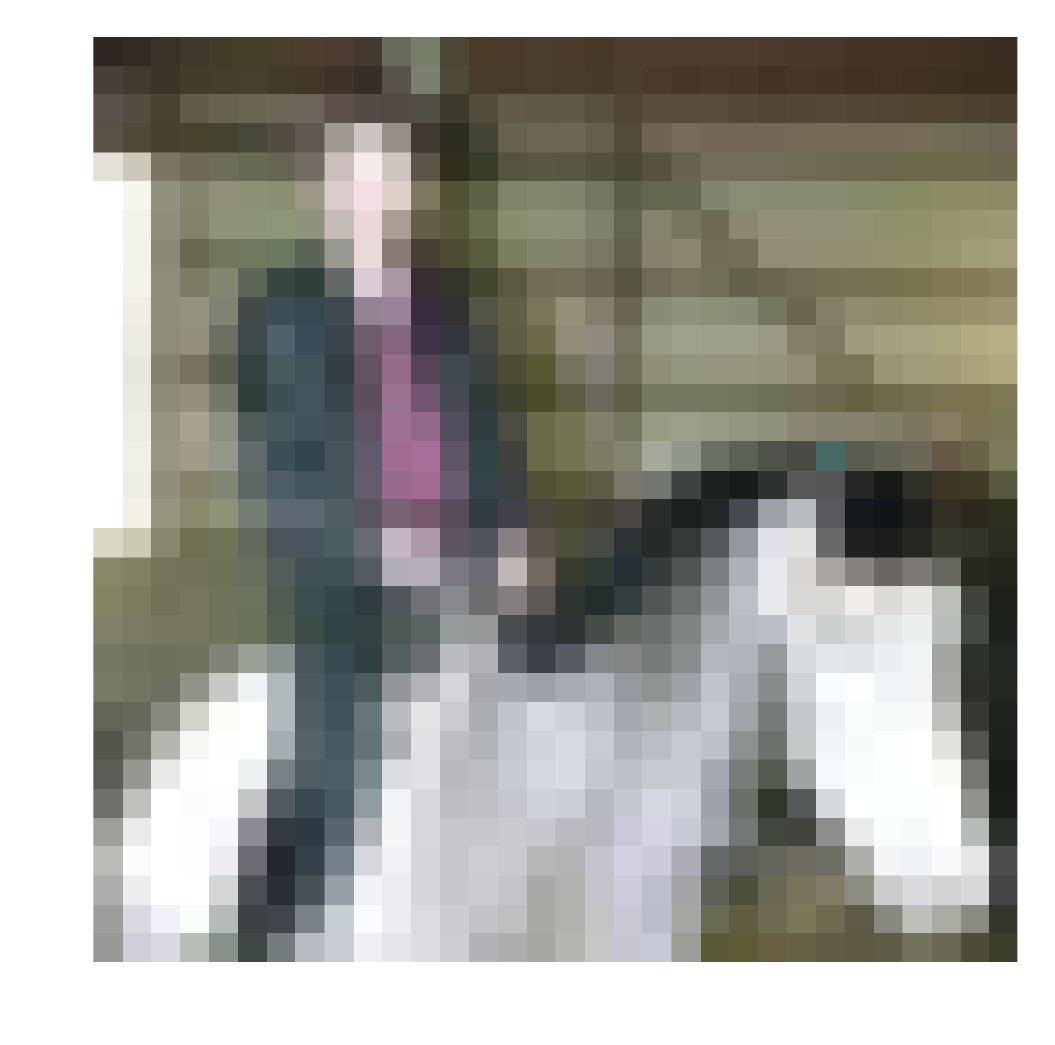}\!
        \captionsetup{font=scriptsize}
        \caption*{bird (1)}
    \end{subfigure}\!
    \begin{subfigure}[b]{0.20\linewidth}
        \includegraphics[width=\linewidth]{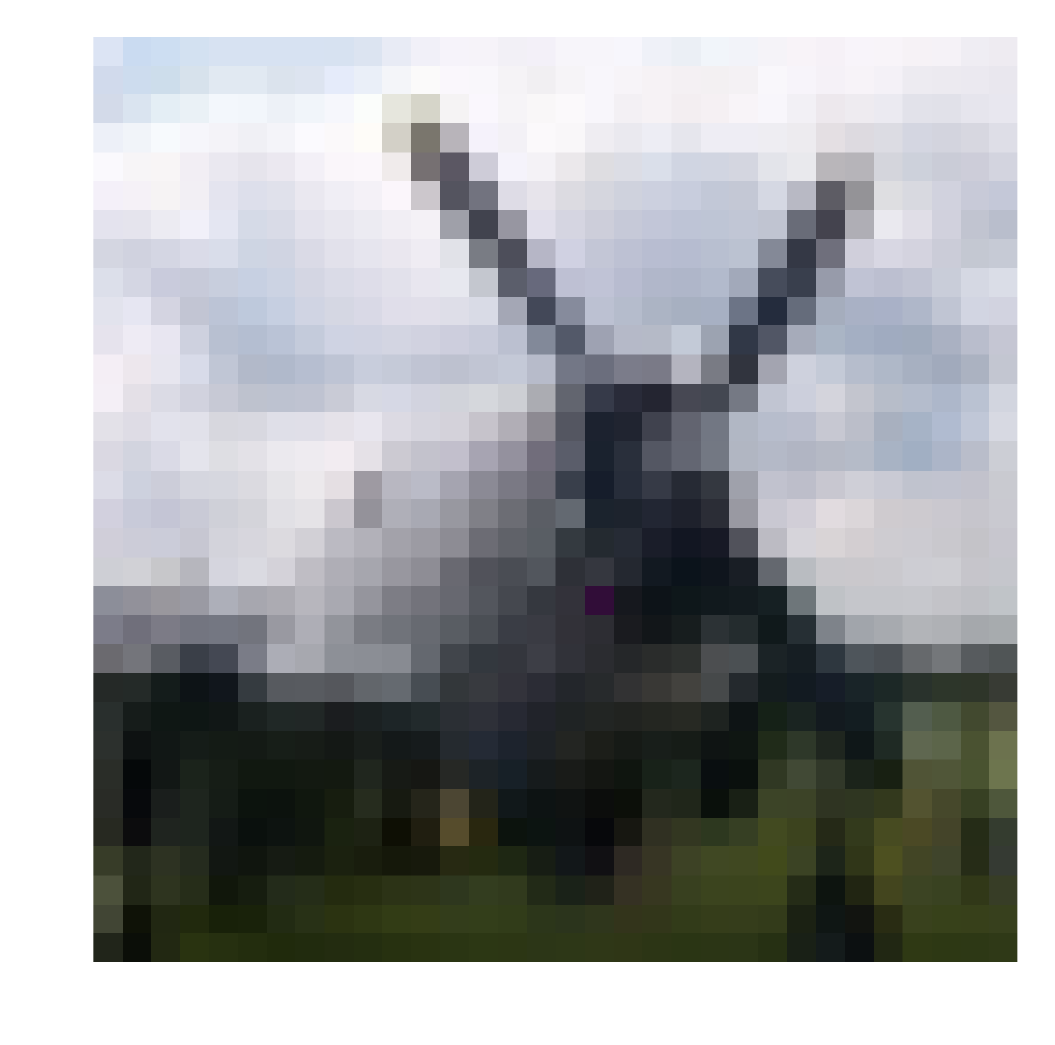}\!
        \captionsetup{font=scriptsize}
        \caption*{deer (1)}
    \end{subfigure}\!
    \begin{subfigure}[b]{0.20\linewidth}
        \includegraphics[width=\linewidth]{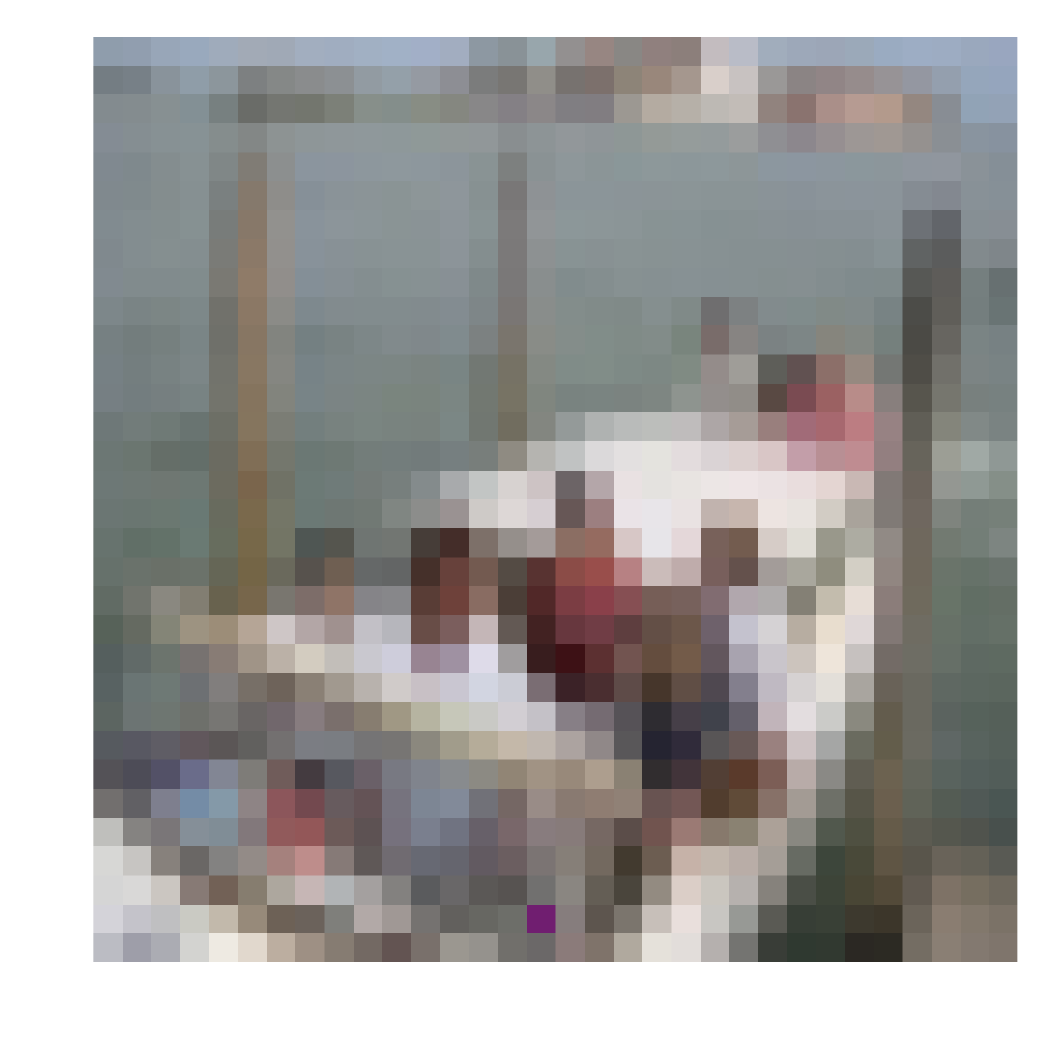}\!
        \captionsetup{font=scriptsize}
        \caption*{car (1)}
    \end{subfigure}\!
    \begin{subfigure}[b]{0.20\linewidth}
        \includegraphics[width=\linewidth]{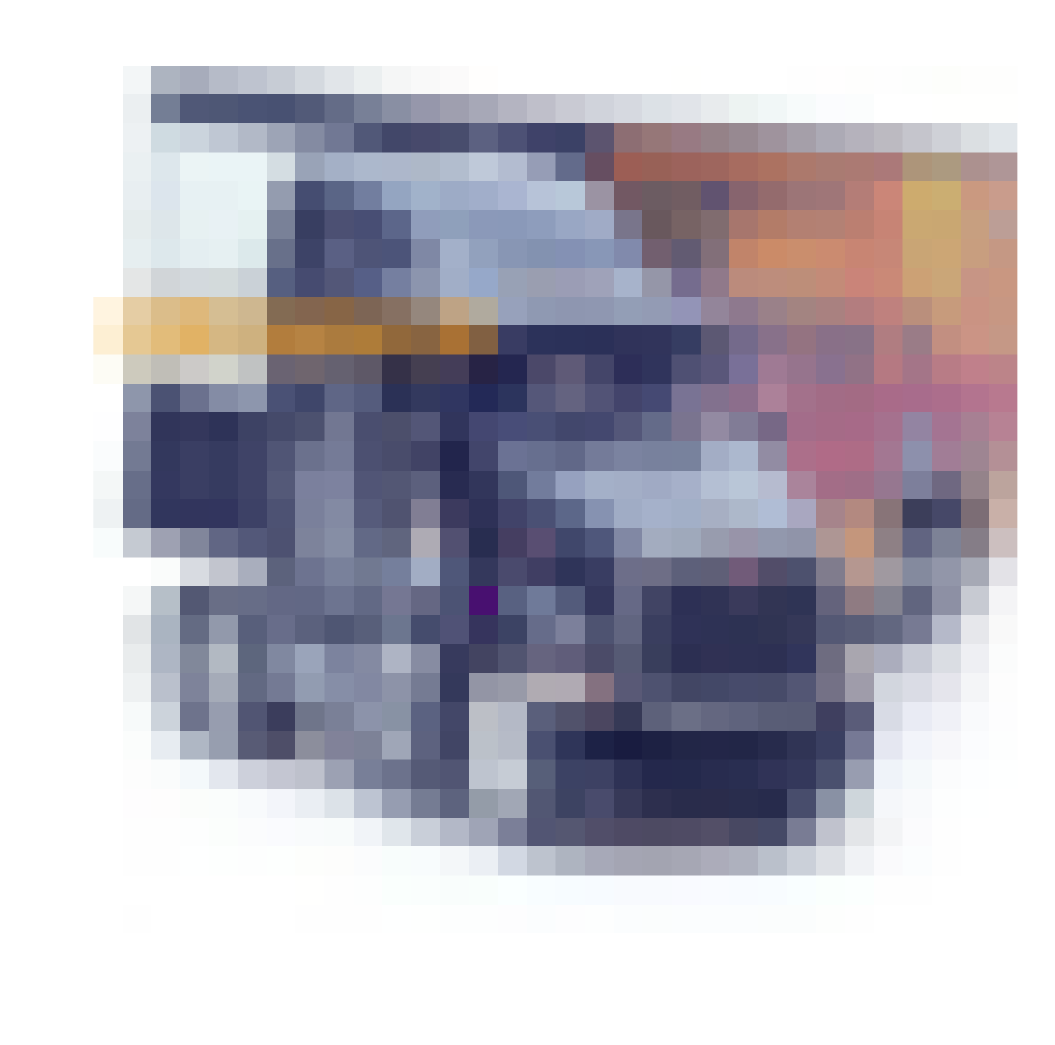}\!
        \captionsetup{font=scriptsize}
        \caption*{car (2)}
    \end{subfigure}\!
    
    \begin{subfigure}[b]{0.20\linewidth}
        \includegraphics[width=\linewidth]{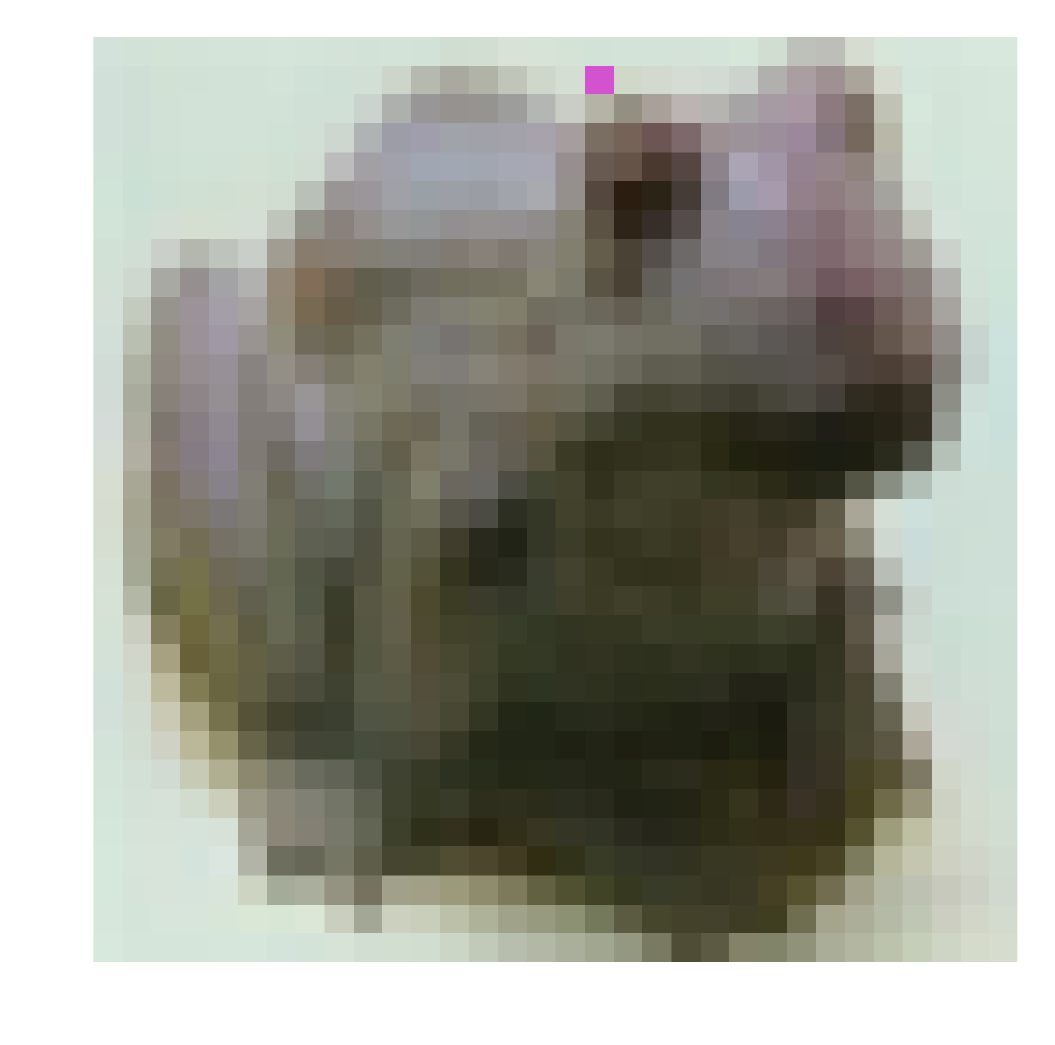}\!
        \captionsetup{font=scriptsize}
        \caption*{cat (1)}
    \end{subfigure}\!
    \begin{subfigure}[b]{0.20\linewidth}
        \includegraphics[width=\linewidth]{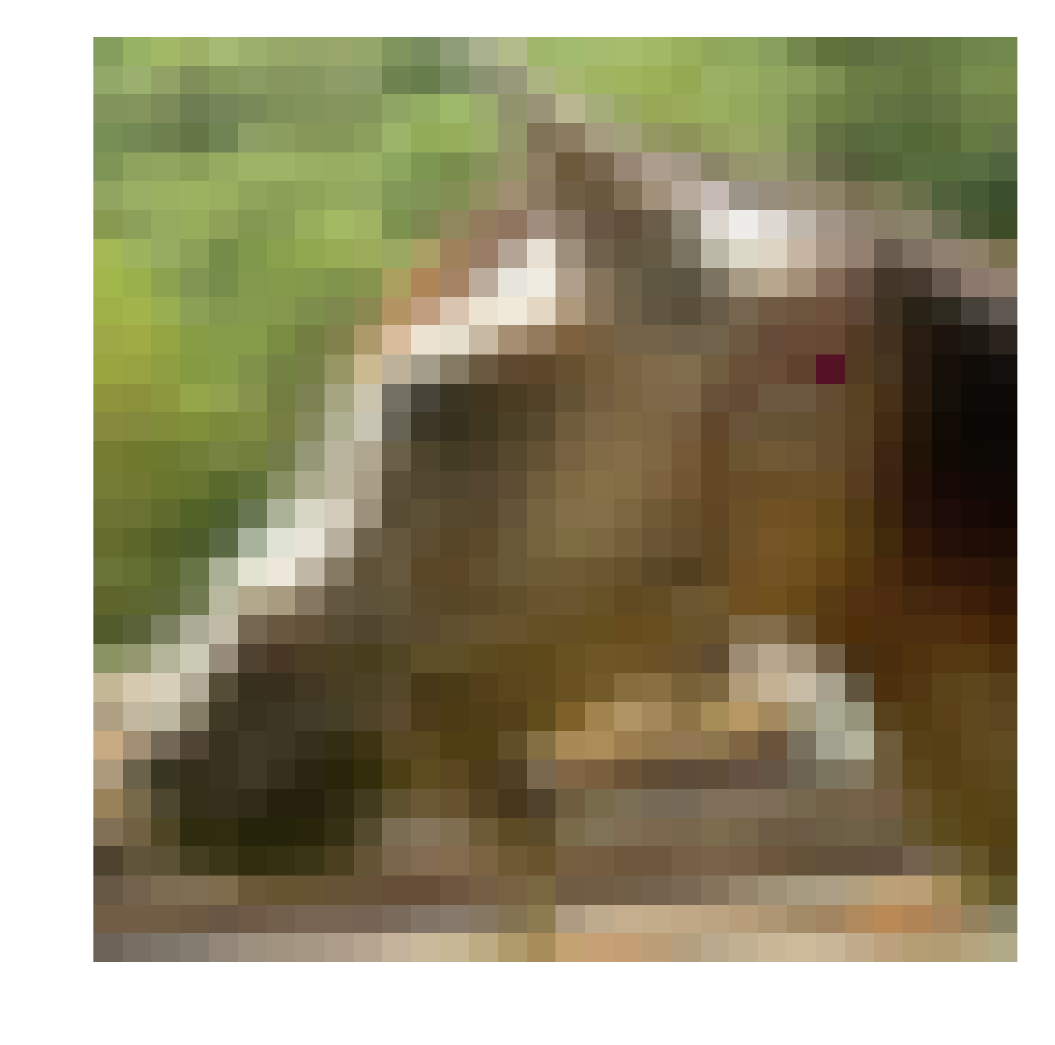}\!
        \captionsetup{font=scriptsize}
        \caption*{deer (1)}
    \end{subfigure}\!
    \begin{subfigure}[b]{0.20\linewidth}
        \includegraphics[width=\linewidth]{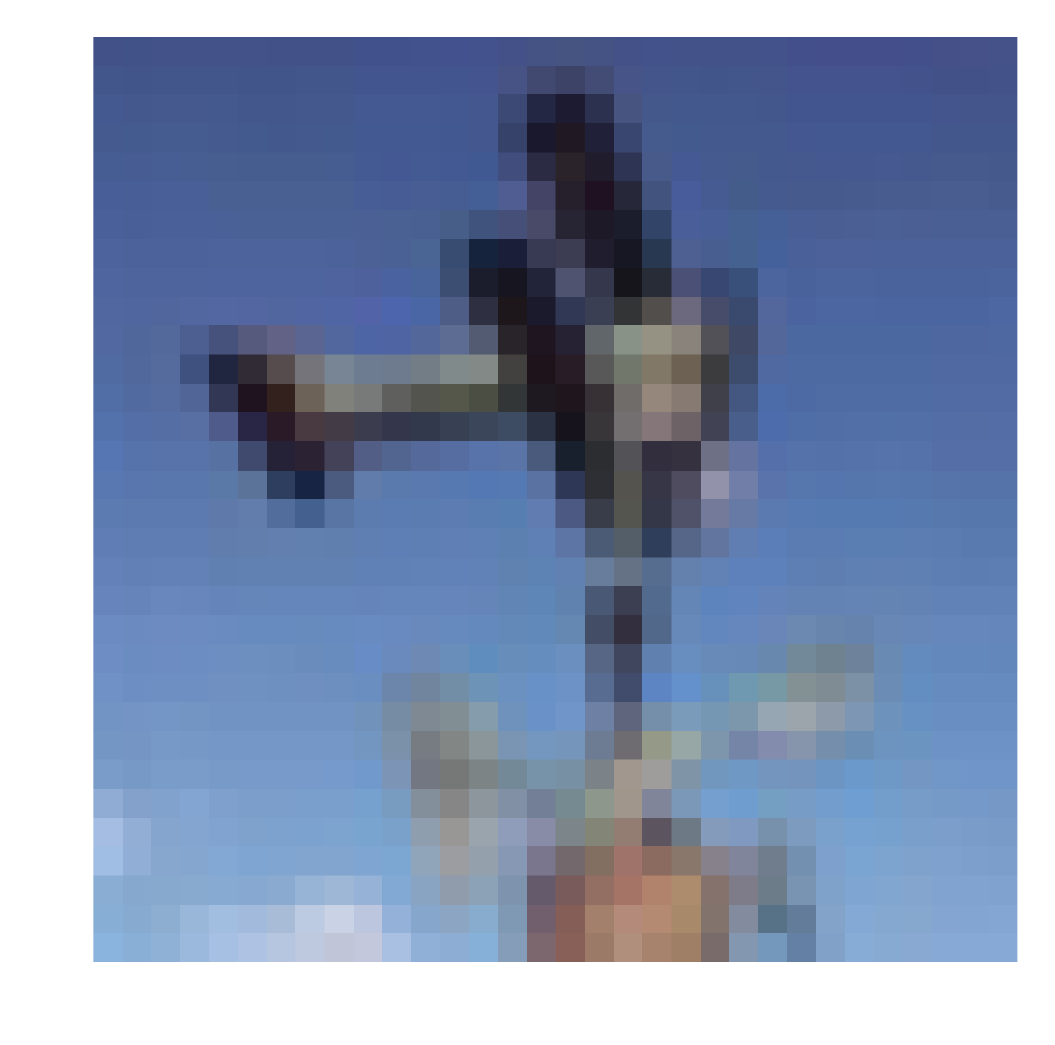}\!
        \captionsetup{font=scriptsize}
        \caption*{bird (1)}
    \end{subfigure}\!
    \begin{subfigure}[b]{0.20\linewidth}
        \includegraphics[width=\linewidth]{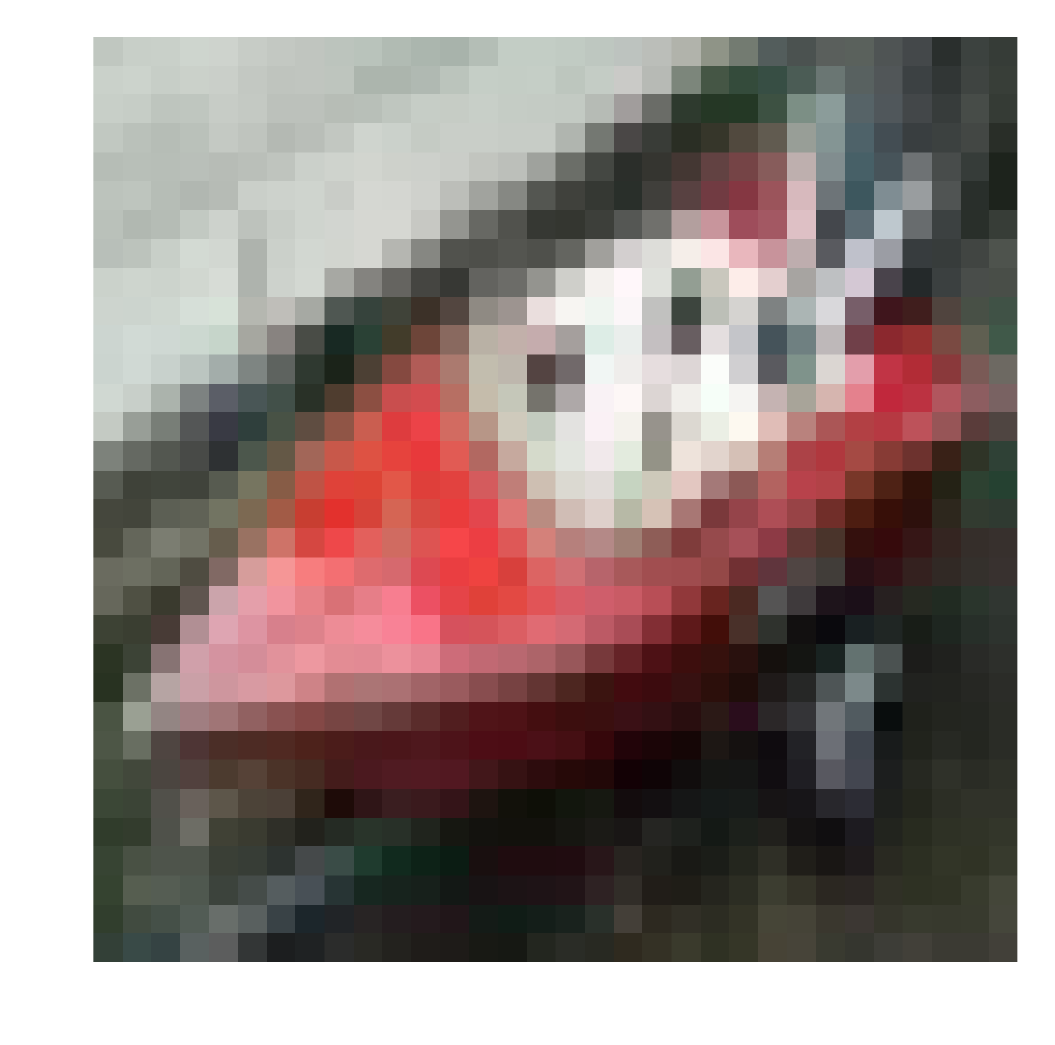}\!
        \captionsetup{font=scriptsize}
        \caption*{plane (1)}
    \end{subfigure}\!
    \begin{subfigure}[b]{0.20\linewidth}
        \includegraphics[width=\linewidth]{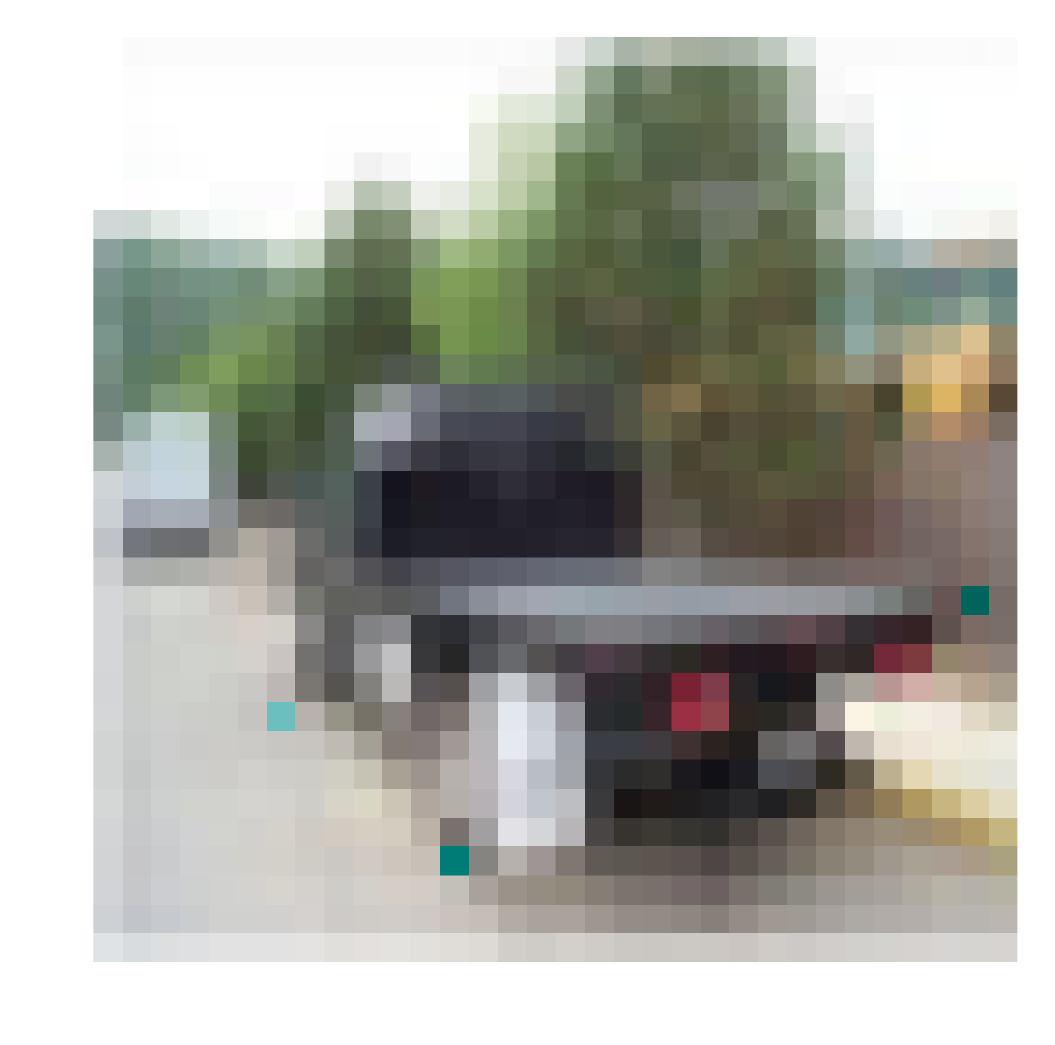}\!
        \captionsetup{font=scriptsize}
        \caption*{plane (3)}
    \end{subfigure}\!
\captionsetup{font=small, skip=8pt}
\caption{Highly sparse perturbations.}
\label{fig:cifar_high}
\end{subfigure}\hfill
\begin{subfigure}[b]{0.3\linewidth}
\captionsetup[subfigure]{skip=1pt}
    \begin{subfigure}[b]{0.20\linewidth}
        \includegraphics[width=\linewidth]{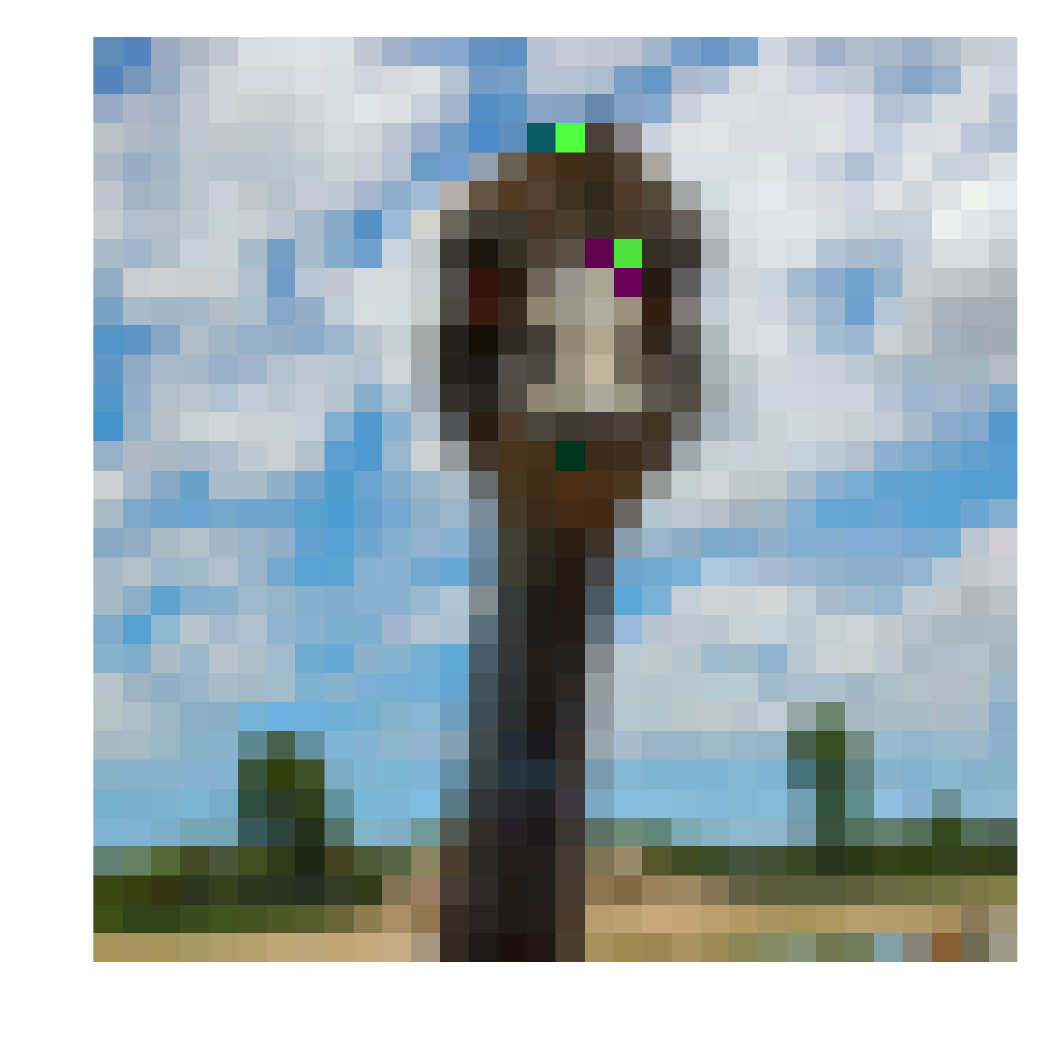}\!
        \captionsetup{font=scriptsize}
        \caption*{cat (6)}
    \end{subfigure}\!
    \begin{subfigure}[b]{0.20\linewidth}
        \includegraphics[width=\linewidth]{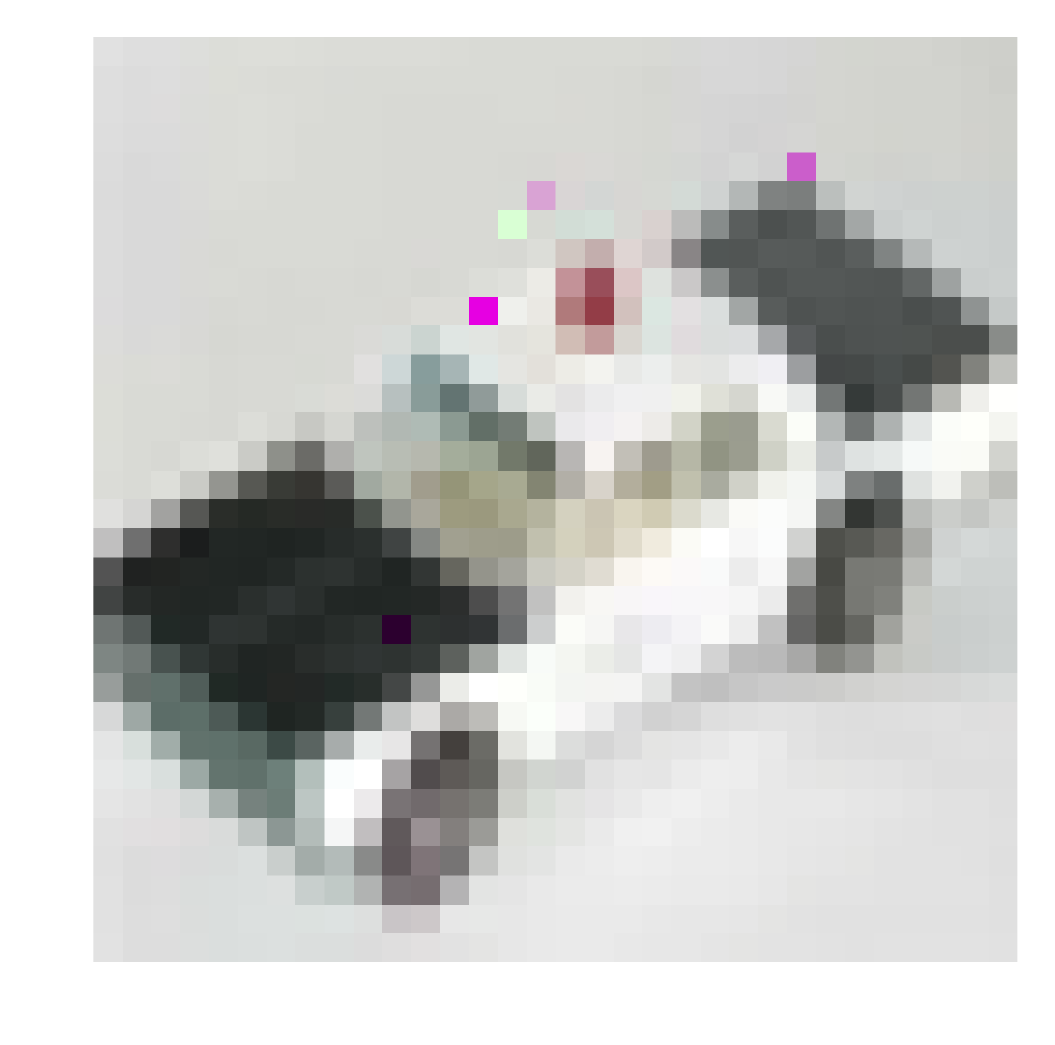}\!
        \captionsetup{font=scriptsize}
        \caption*{cat (6)}
    \end{subfigure}\!
    \begin{subfigure}[b]{0.20\linewidth}
        \includegraphics[width=\linewidth]{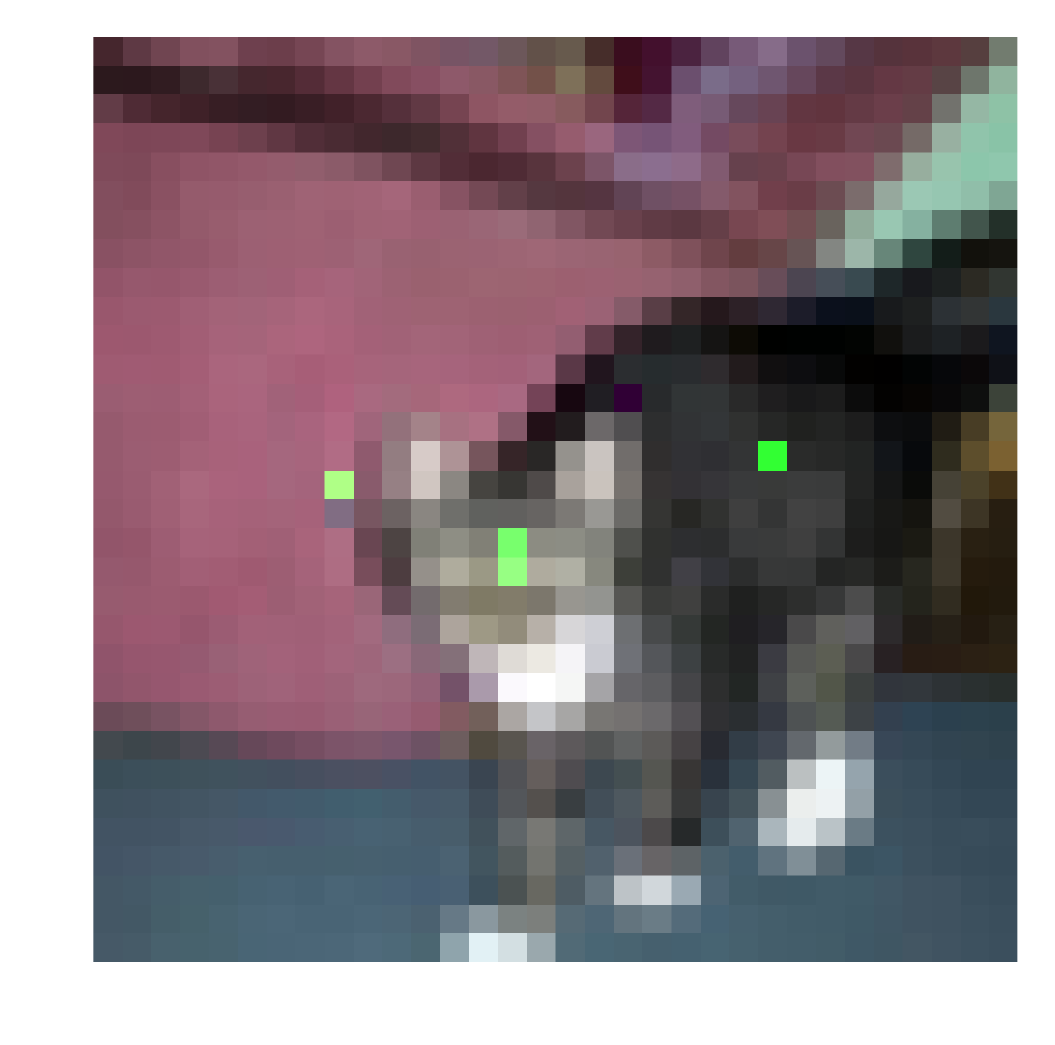}\!
        \captionsetup{font=scriptsize}
        \caption*{deer (6)}
    \end{subfigure}\!
    \begin{subfigure}[b]{0.20\linewidth}
        \includegraphics[width=\linewidth]{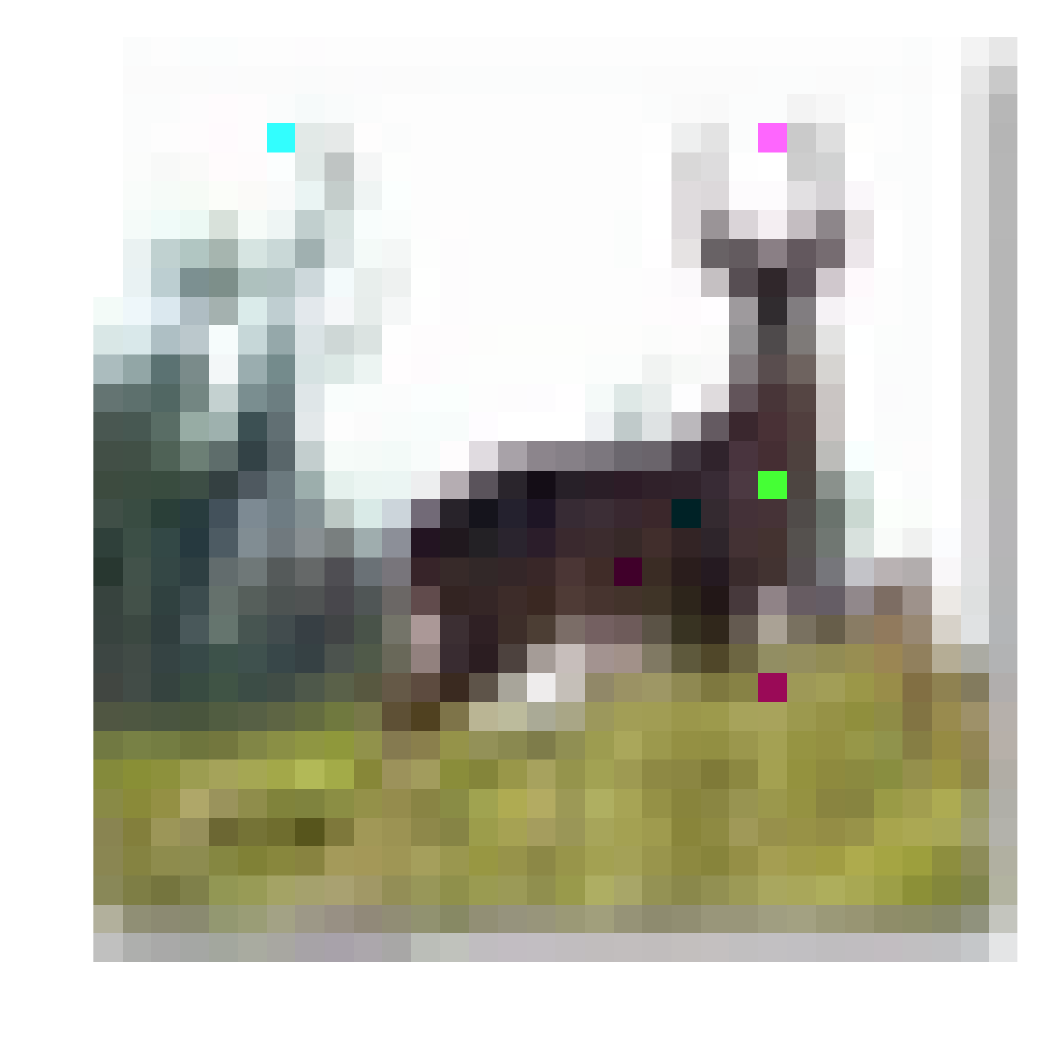}\!
        \captionsetup{font=scriptsize}
        \caption*{cat (6)}
    \end{subfigure}\!
    \begin{subfigure}[b]{0.20\linewidth}
        \includegraphics[width=\linewidth]{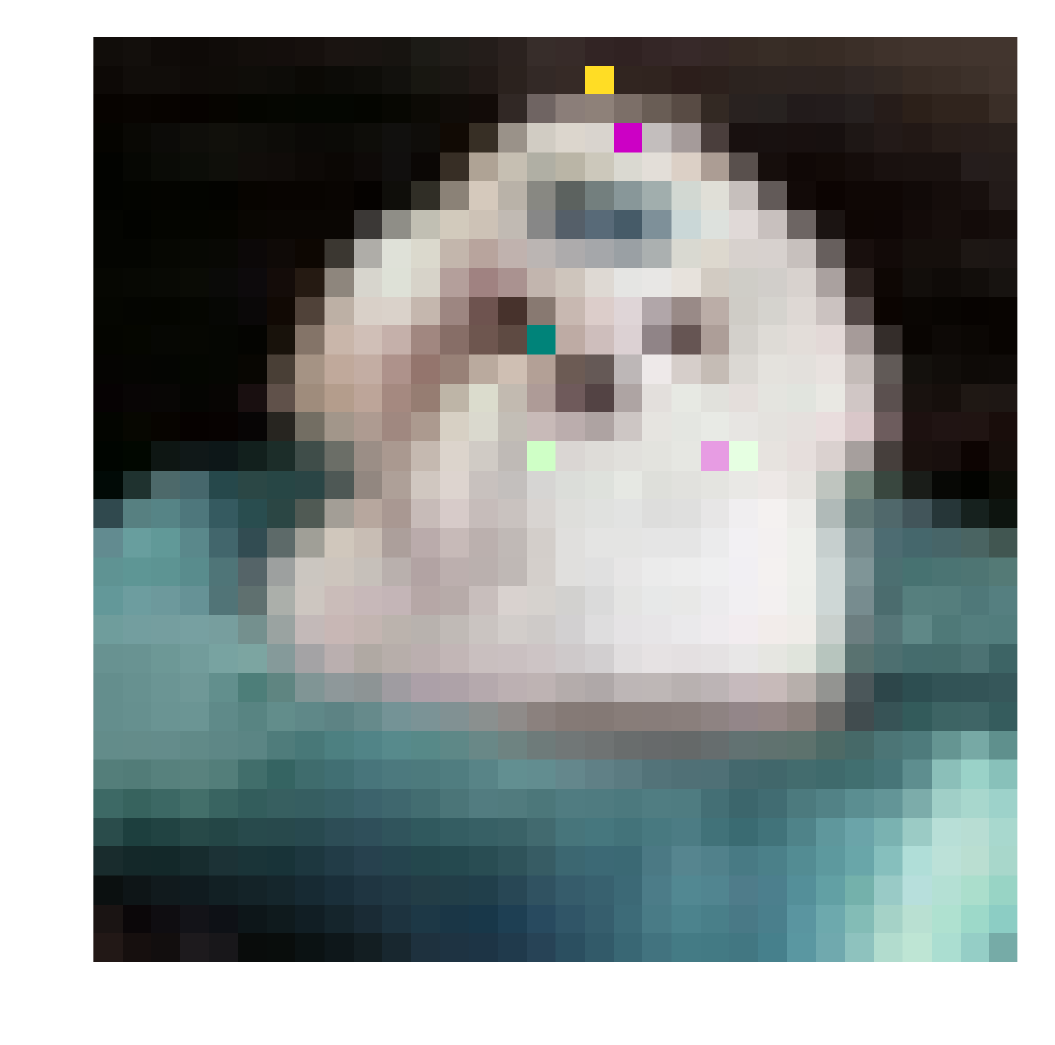}\!
        \captionsetup{font=scriptsize}
        \caption*{horse (6)}
    \end{subfigure}\!
    
    \begin{subfigure}[b]{0.20\linewidth}
        \includegraphics[width=\linewidth]{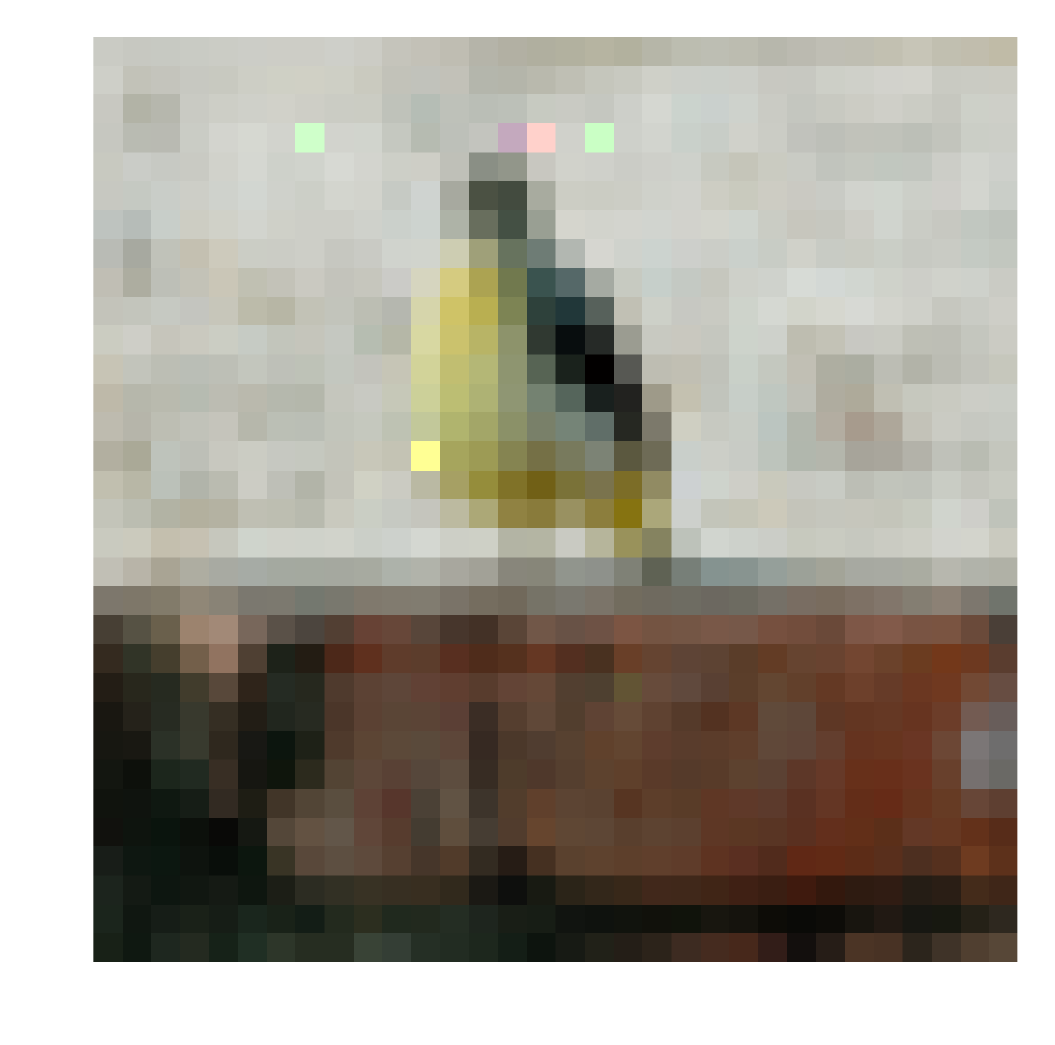}\!
        \captionsetup{font=scriptsize}
        \caption*{ship (6)}
    \end{subfigure}\!
    \begin{subfigure}[b]{0.20\linewidth}
        \includegraphics[width=\linewidth]{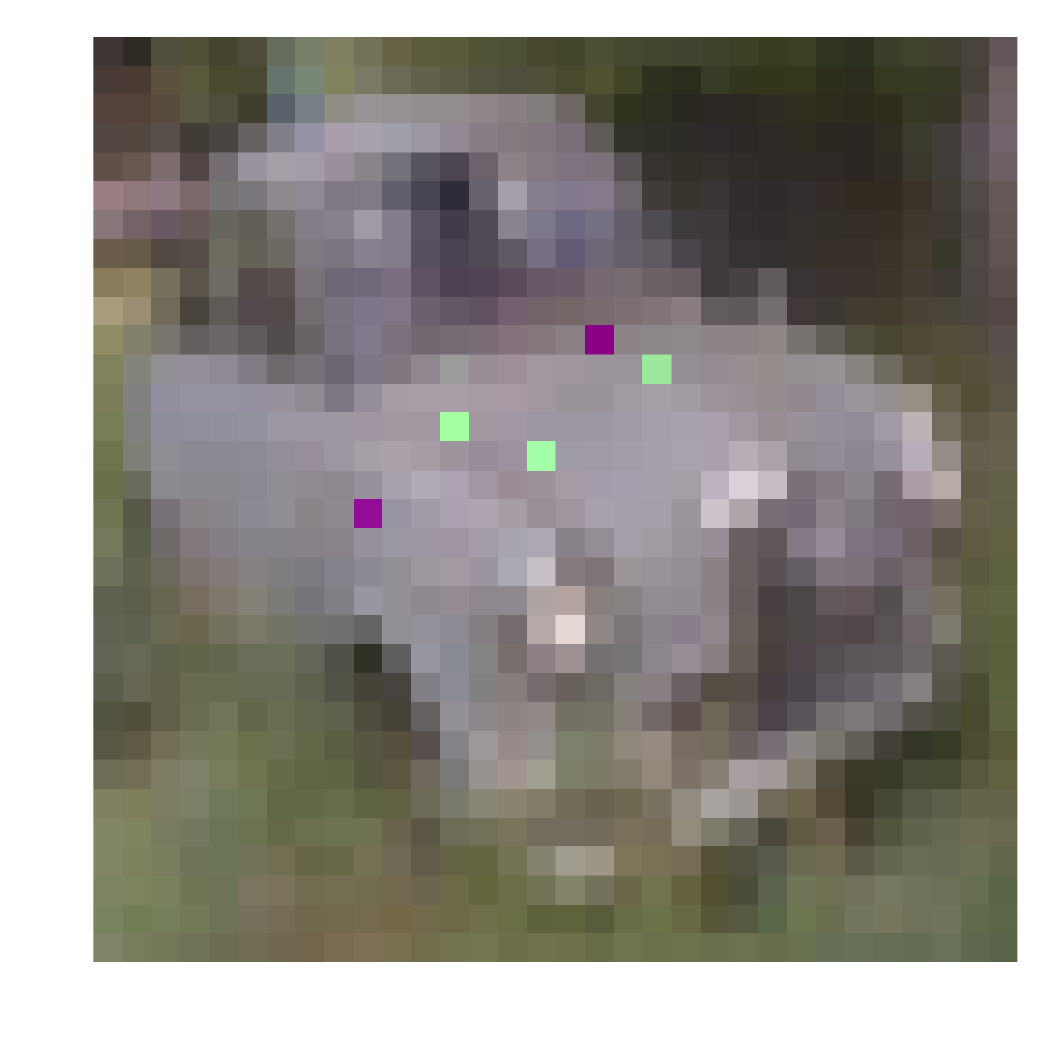}\!
        \captionsetup{font=scriptsize}
        \caption*{frog (6)}
    \end{subfigure}\!
    \begin{subfigure}[b]{0.20\linewidth}
        \includegraphics[width=\linewidth]{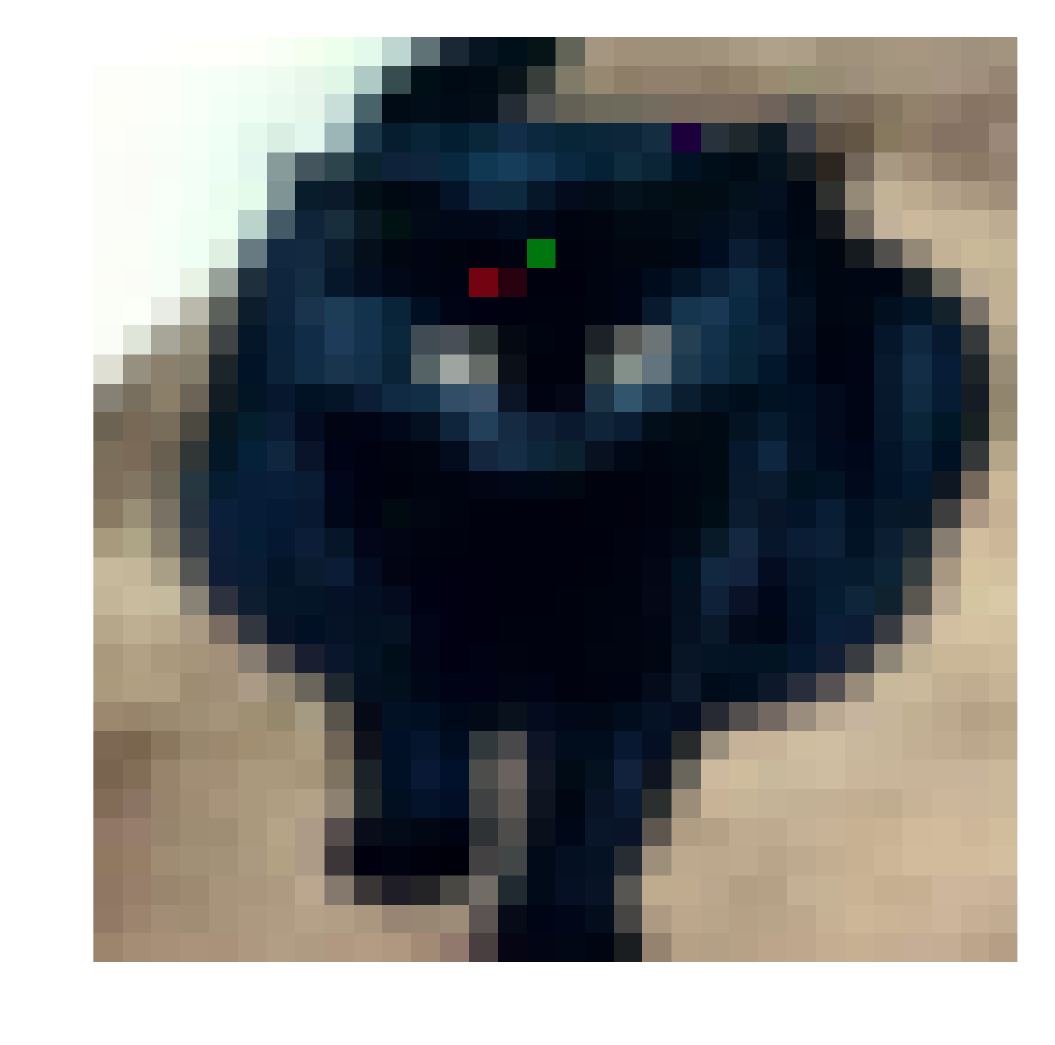}\!
        \captionsetup{font=scriptsize}
        \caption*{car (6)}
    \end{subfigure}\!
    \begin{subfigure}[b]{0.20\linewidth}
        \includegraphics[width=\linewidth]{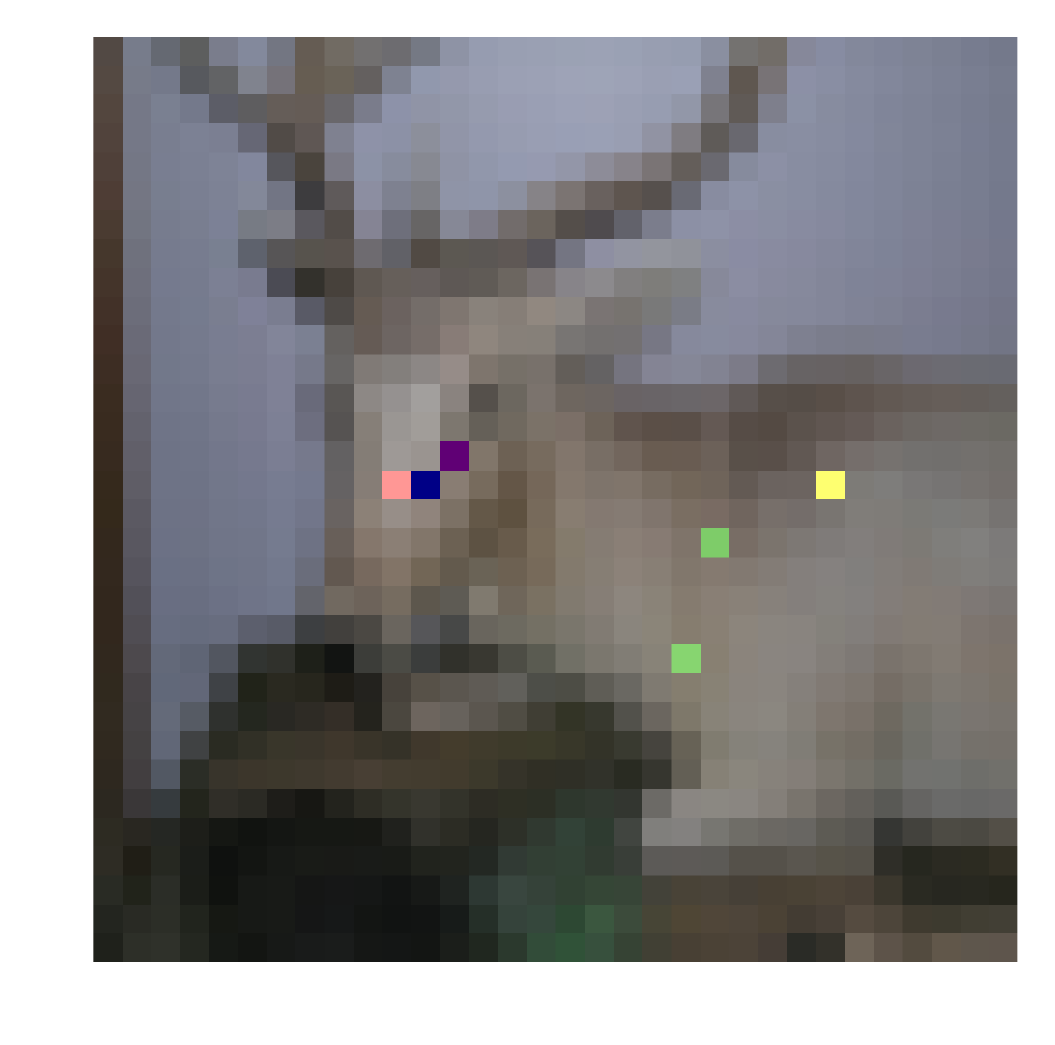}\!
        \captionsetup{font=scriptsize}
        \caption*{plane (6)}
    \end{subfigure}\!
    \begin{subfigure}[b]{0.20\linewidth}
        \includegraphics[width=\linewidth]{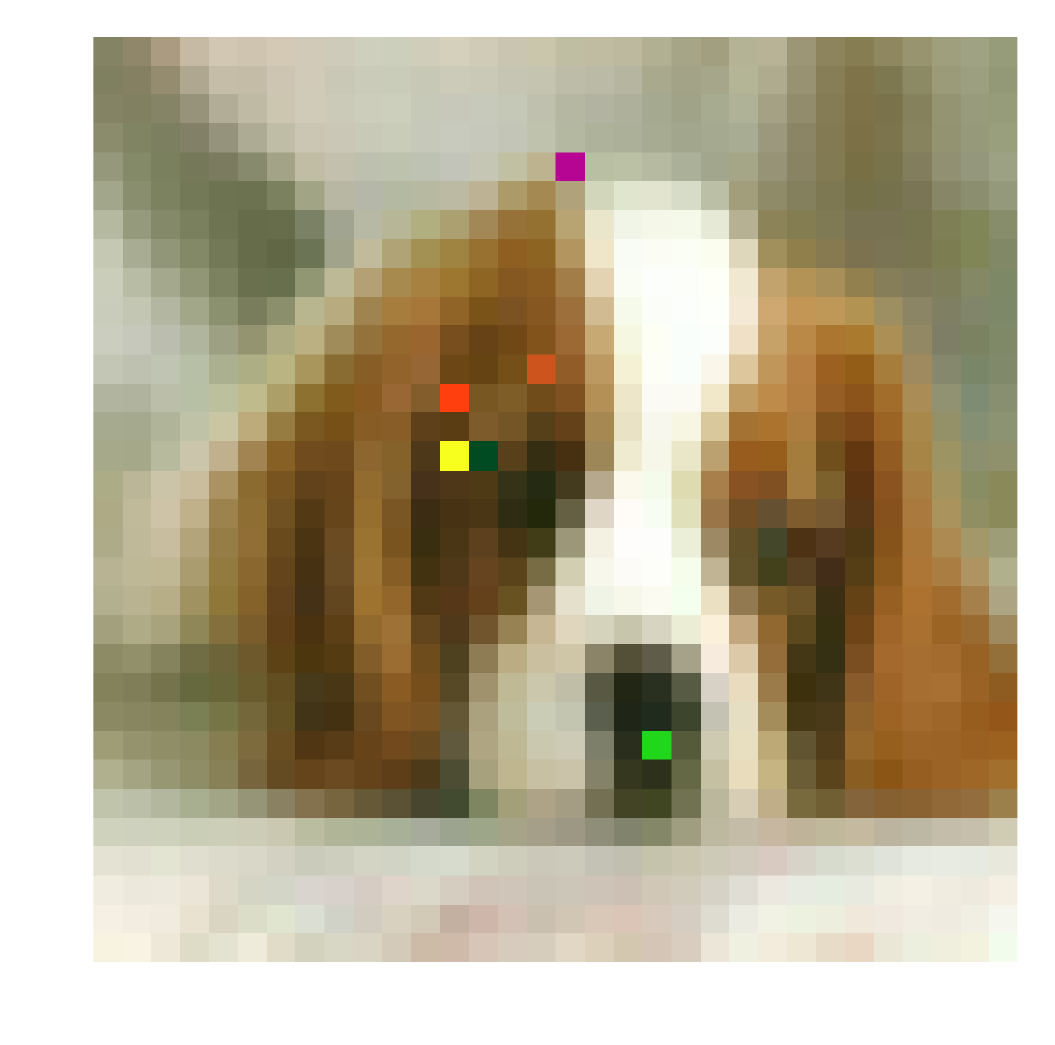}\!
        \captionsetup{font=scriptsize}
        \caption*{cat (6)}
    \end{subfigure}\!
    
    \begin{subfigure}[b]{0.20\linewidth}
        \includegraphics[width=\linewidth]{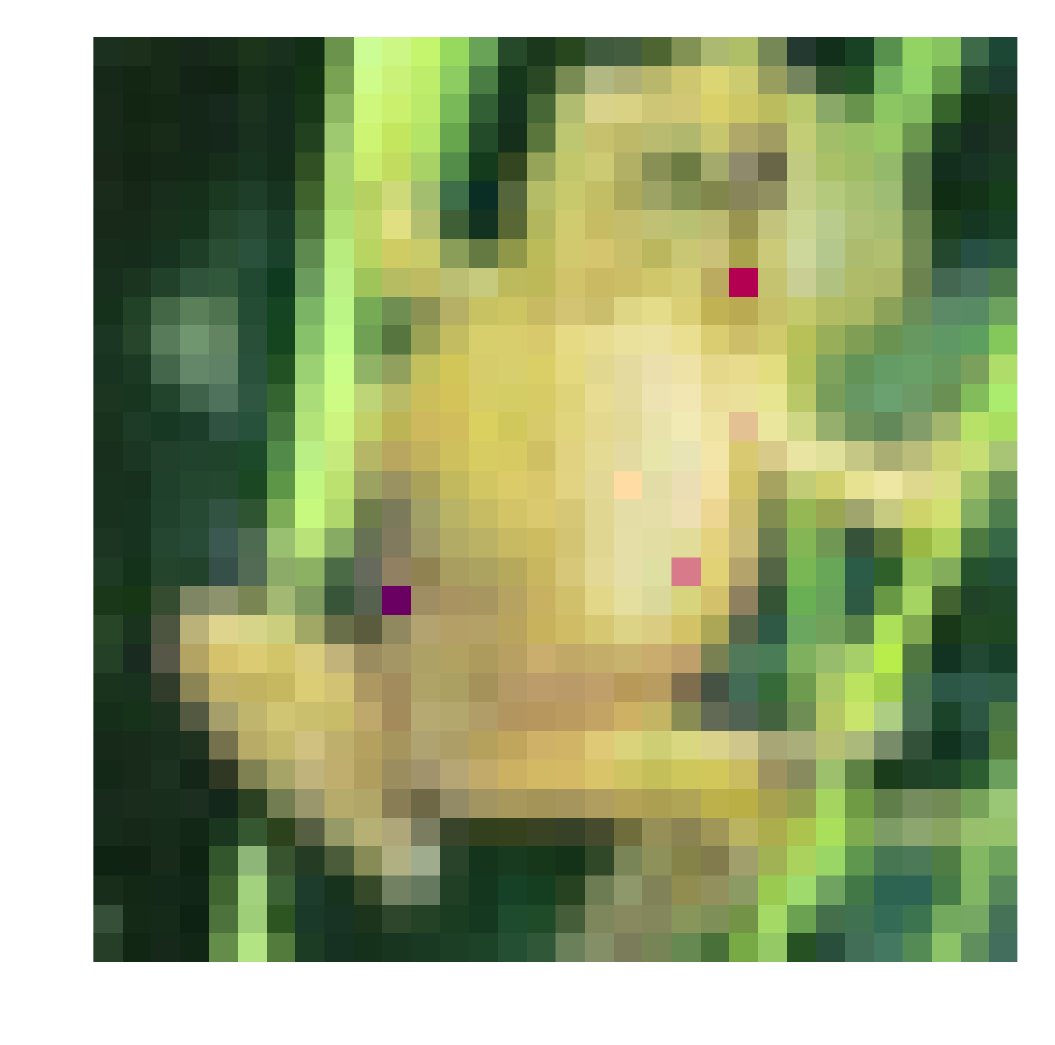}\!
        \captionsetup{font=scriptsize}
        \caption*{bird (6)}
    \end{subfigure}\!
    \begin{subfigure}[b]{0.20\linewidth}
        \includegraphics[width=\linewidth]{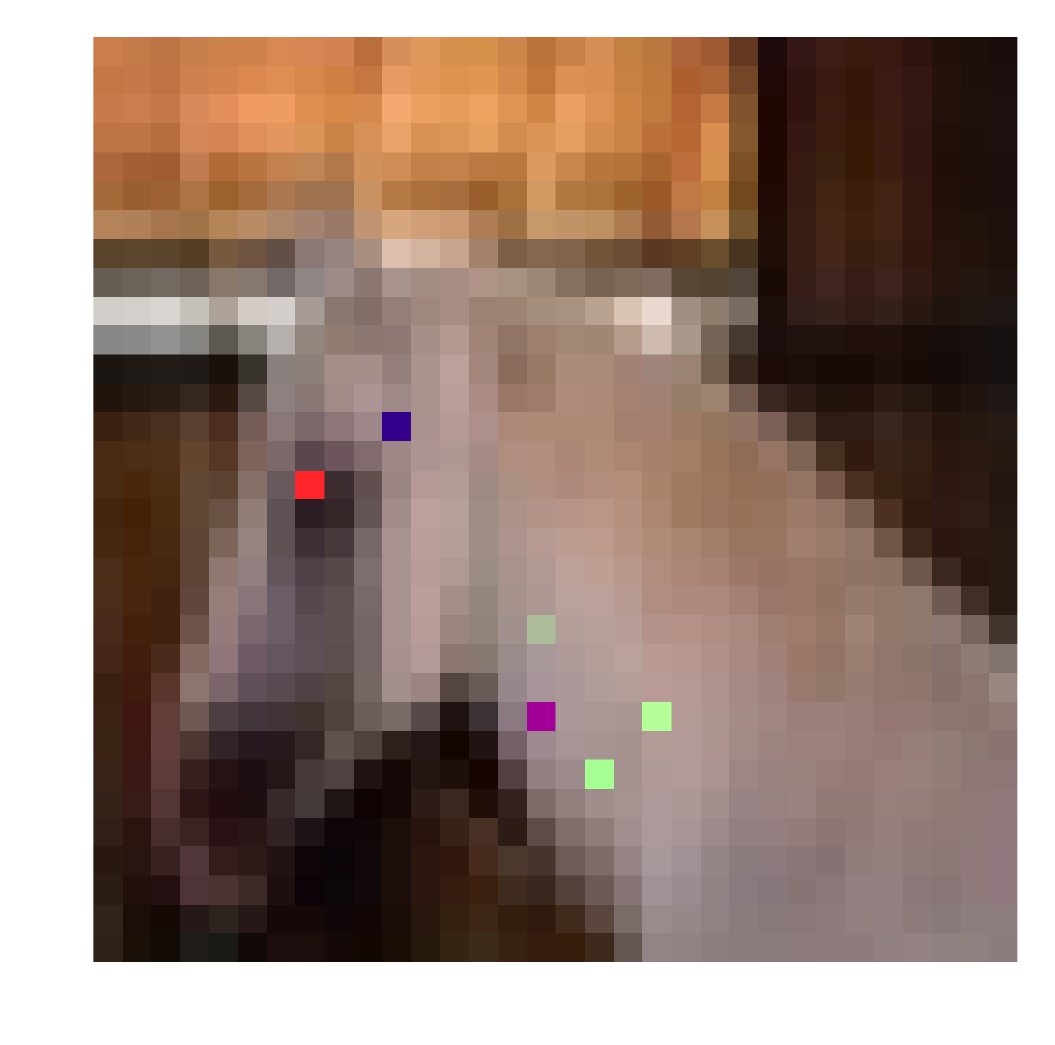}\!
        \captionsetup{font=scriptsize}
        \caption*{cat (6)}
    \end{subfigure}\!
    \begin{subfigure}[b]{0.20\linewidth}
        \includegraphics[width=\linewidth]{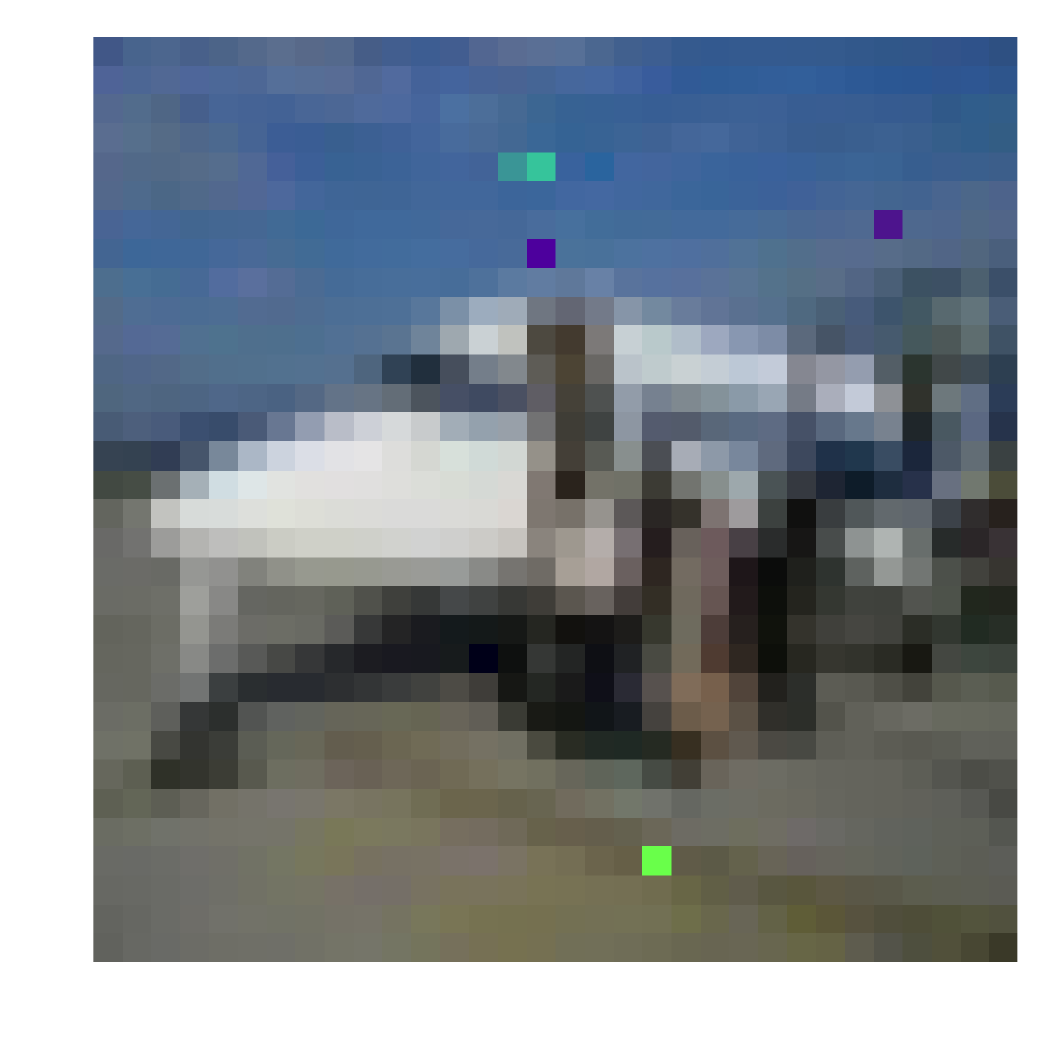}\!
        \captionsetup{font=scriptsize}
        \caption*{horse (7)}
    \end{subfigure}\!
    \begin{subfigure}[b]{0.20\linewidth}
        \includegraphics[width=\linewidth]{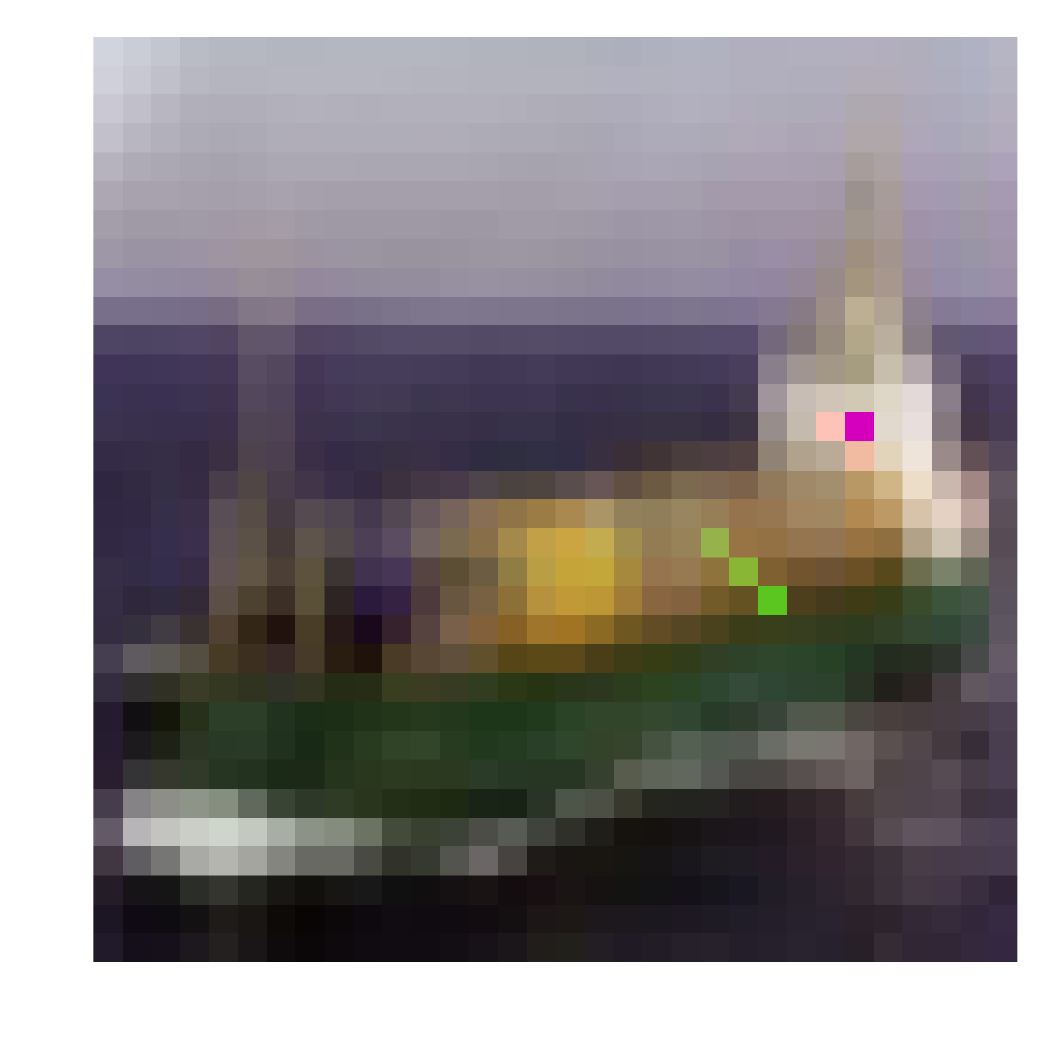}\!
        \captionsetup{font=scriptsize}
        \caption*{cat (6)}
    \end{subfigure}\!
    \begin{subfigure}[b]{0.20\linewidth}
        \includegraphics[width=\linewidth]{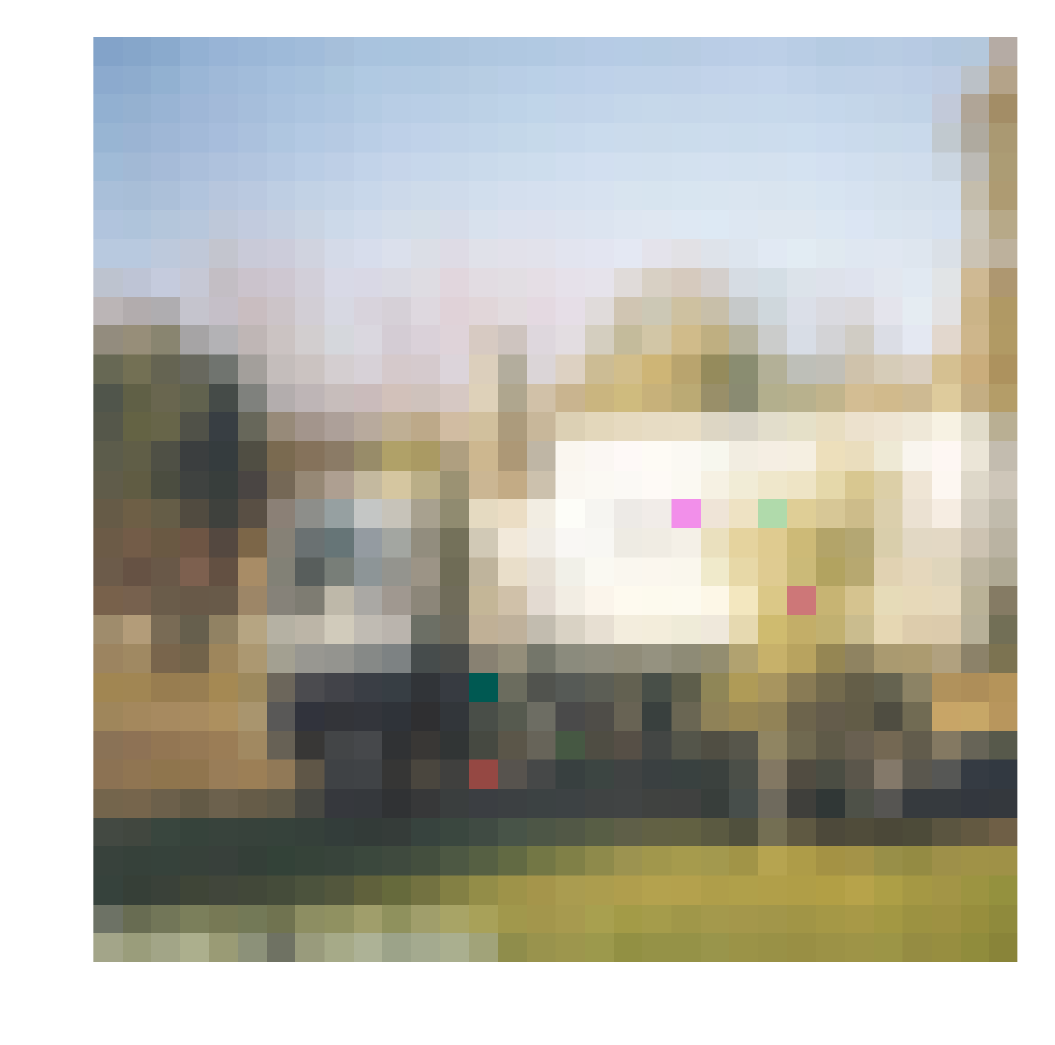}\!
        \captionsetup{font=scriptsize}
        \caption*{plane (6)}
    \end{subfigure}\!
    
    \begin{subfigure}[b]{0.20\linewidth}
        \includegraphics[width=\linewidth]{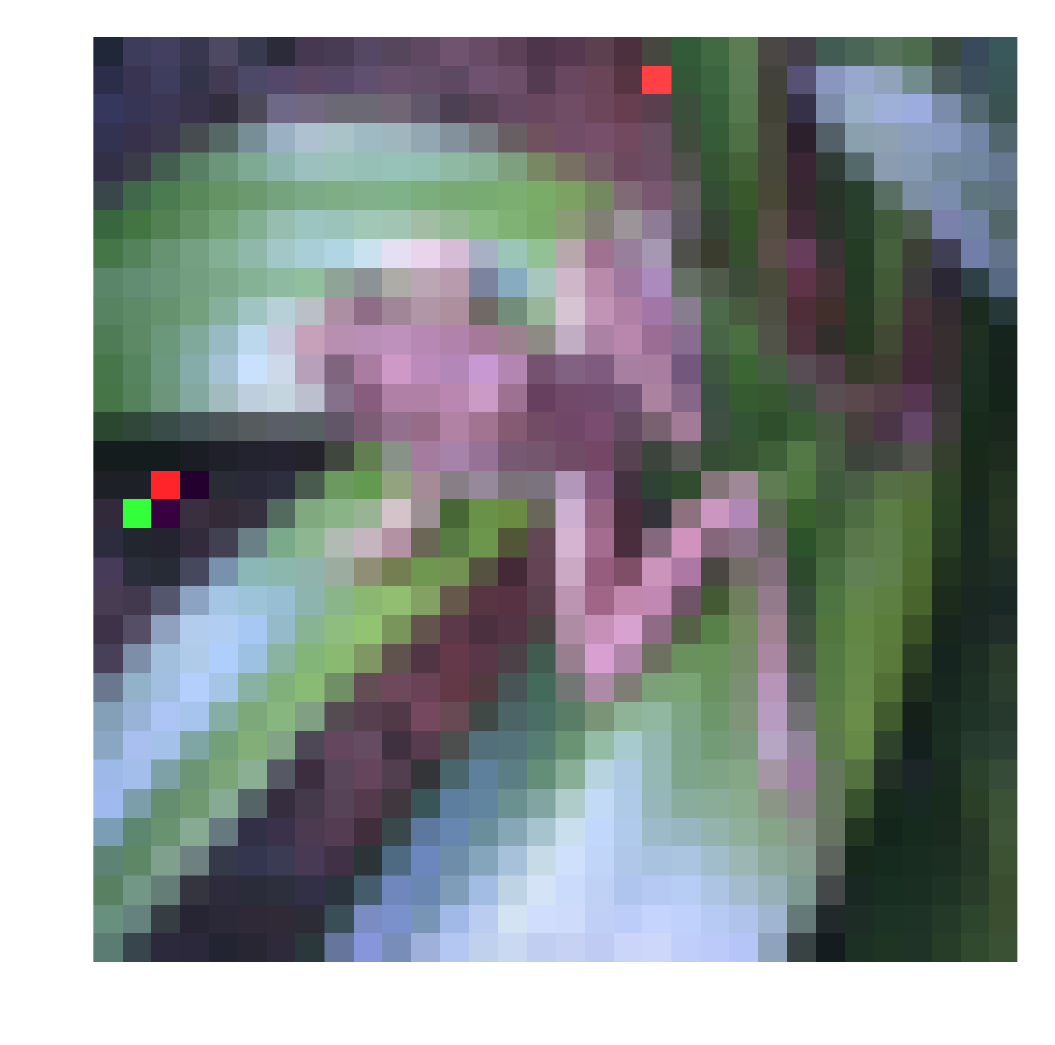}\!
        \captionsetup{font=scriptsize}
        \caption*{car (6)}
    \end{subfigure}\!
    \begin{subfigure}[b]{0.20\linewidth}
        \includegraphics[width=\linewidth]{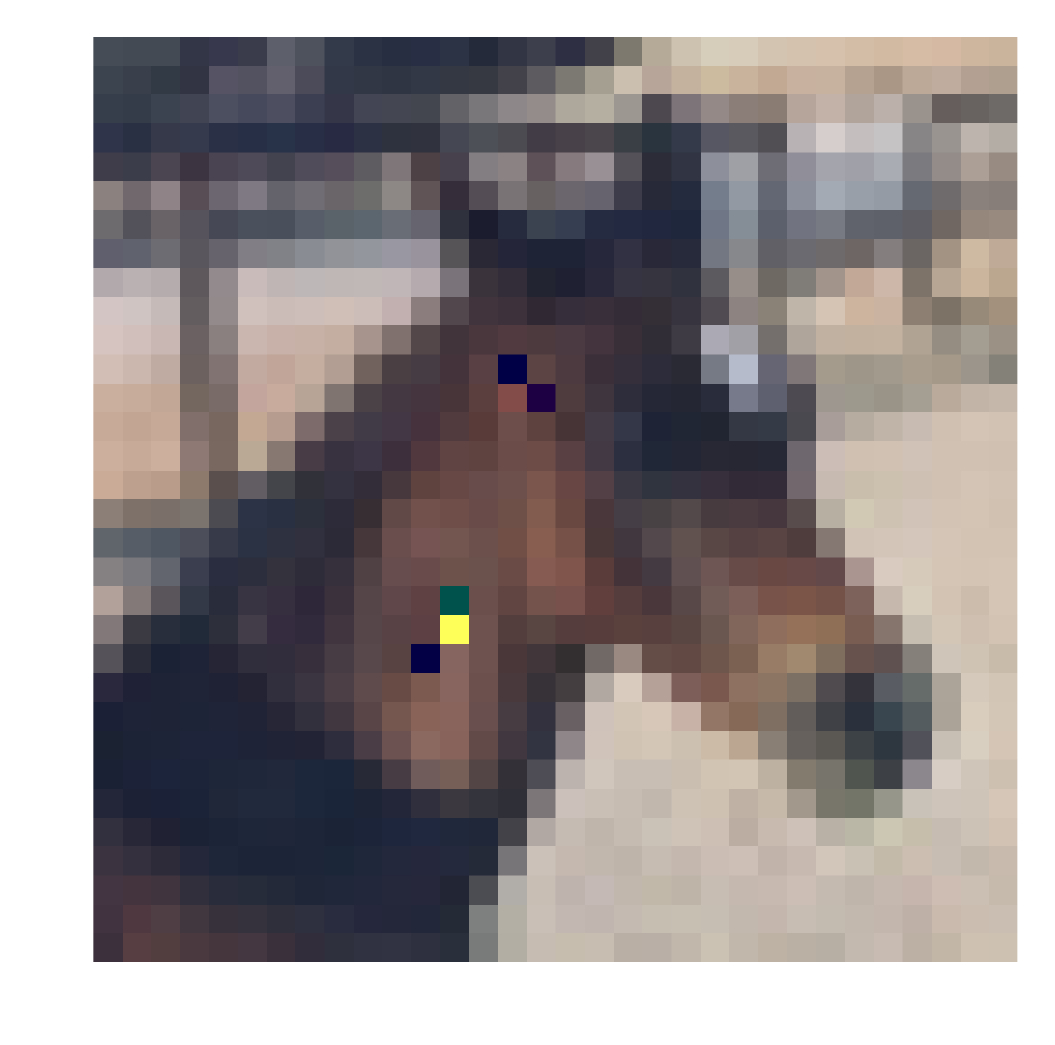}\!
        \captionsetup{font=scriptsize}
        \caption*{plane (6)}
    \end{subfigure}\!
    \begin{subfigure}[b]{0.20\linewidth}
        \includegraphics[width=\linewidth]{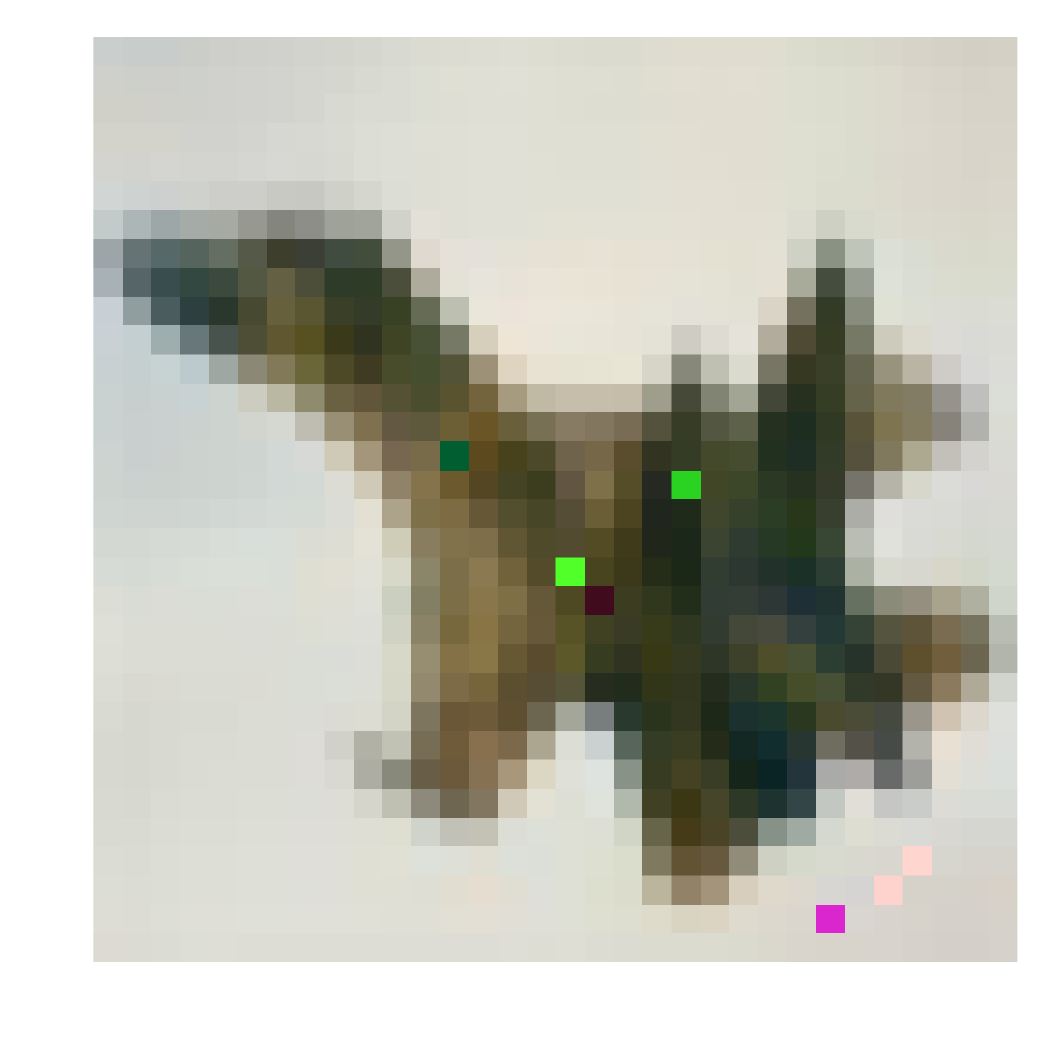}\!
        \captionsetup{font=scriptsize}
        \caption*{frog (8)}
    \end{subfigure}\!
    \begin{subfigure}[b]{0.20\linewidth}
        \includegraphics[width=\linewidth]{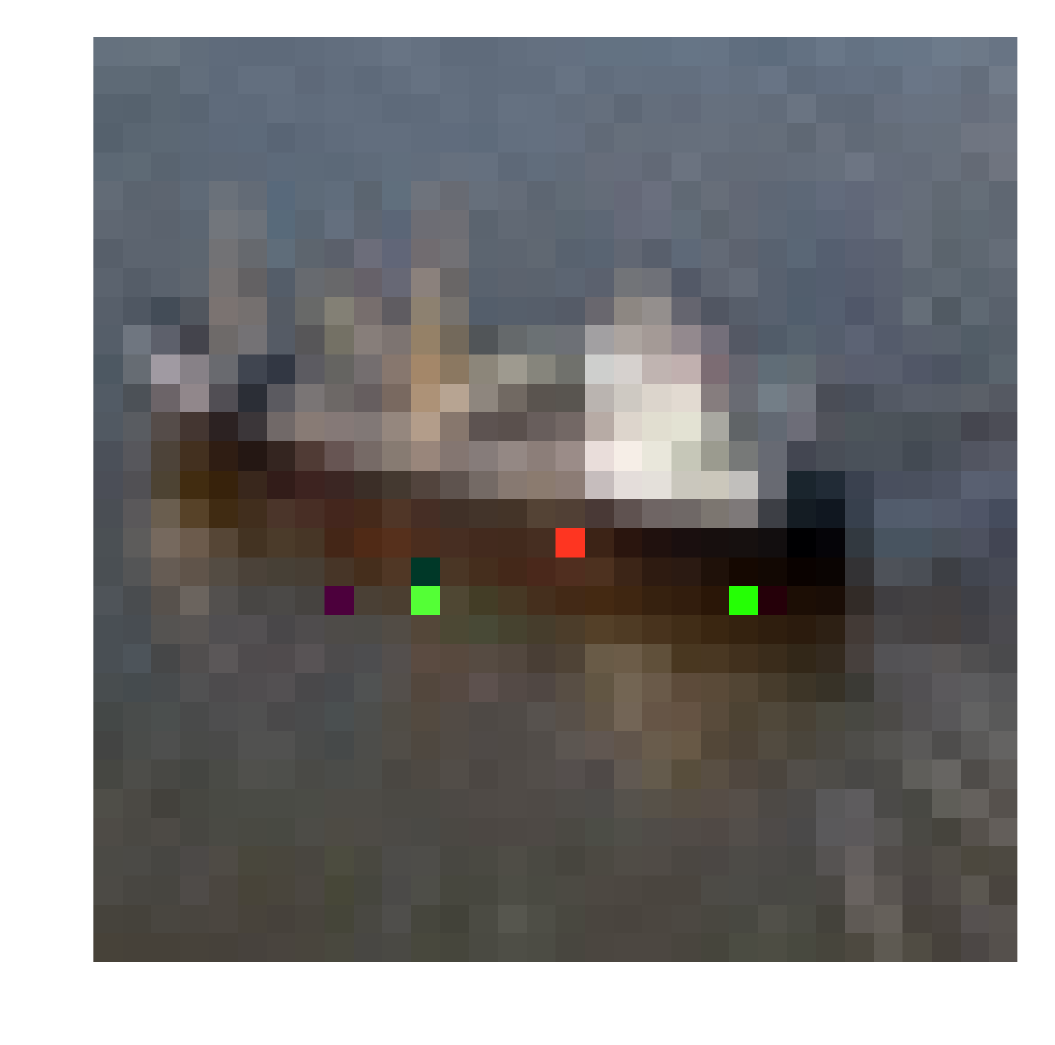}\!
        \captionsetup{font=scriptsize}
        \caption*{plane (6)}
    \end{subfigure}\!
    \begin{subfigure}[b]{0.20\linewidth}
        \includegraphics[width=\linewidth]{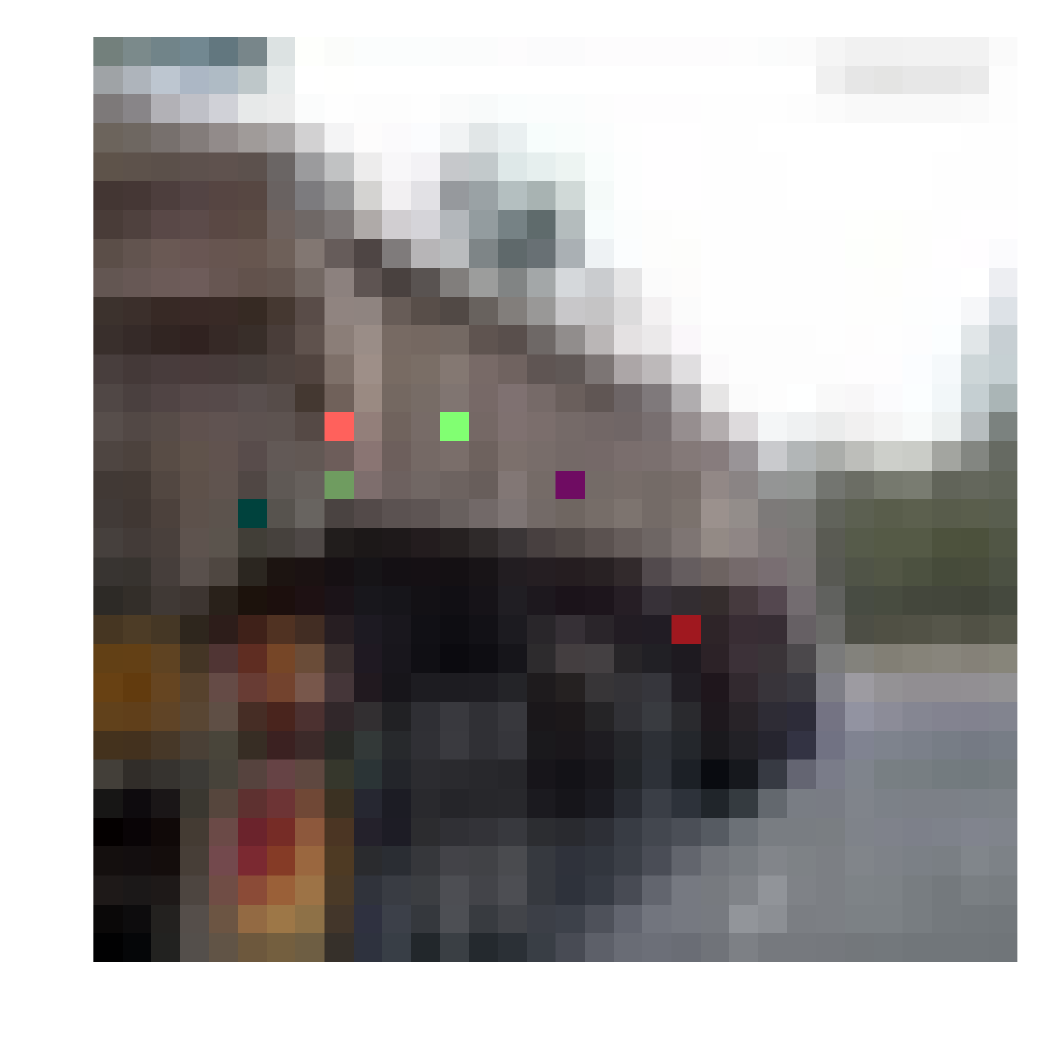}\!
        \captionsetup{font=scriptsize}
        \caption*{ship (6)}
    \end{subfigure}\!
\captionsetup{font=small, skip=8pt}
\caption{Very sparse perturbations.}
\label{fig:cifar_very}
\end{subfigure}\hfill
\begin{subfigure}[b]{0.3\linewidth}
\captionsetup[subfigure]{skip=1pt}
    \begin{subfigure}[b]{0.20\linewidth}
        \includegraphics[width=\linewidth]{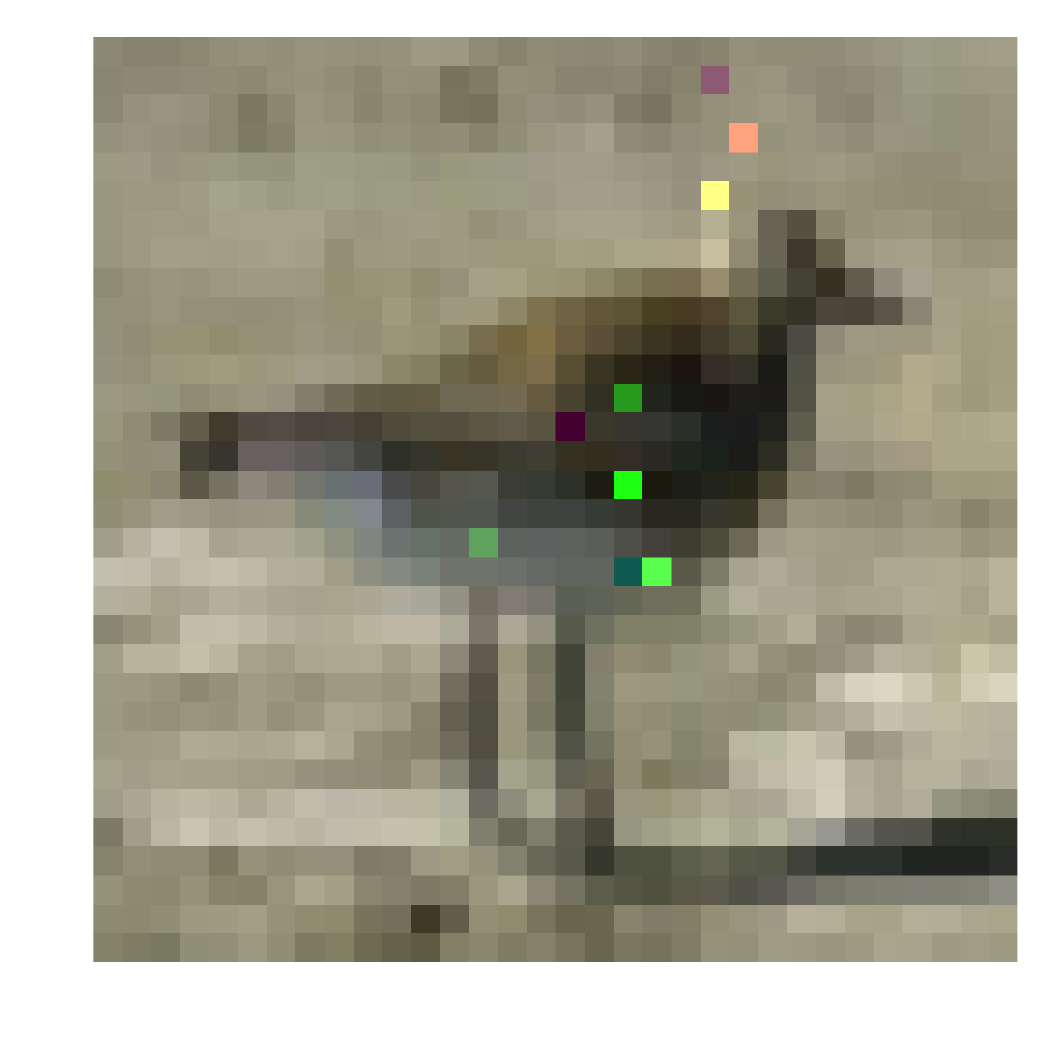}\!
        \captionsetup{font=scriptsize}
        \caption*{deer (9)}
    \end{subfigure}\!
    \begin{subfigure}[b]{0.20\linewidth}
        \includegraphics[width=\linewidth]{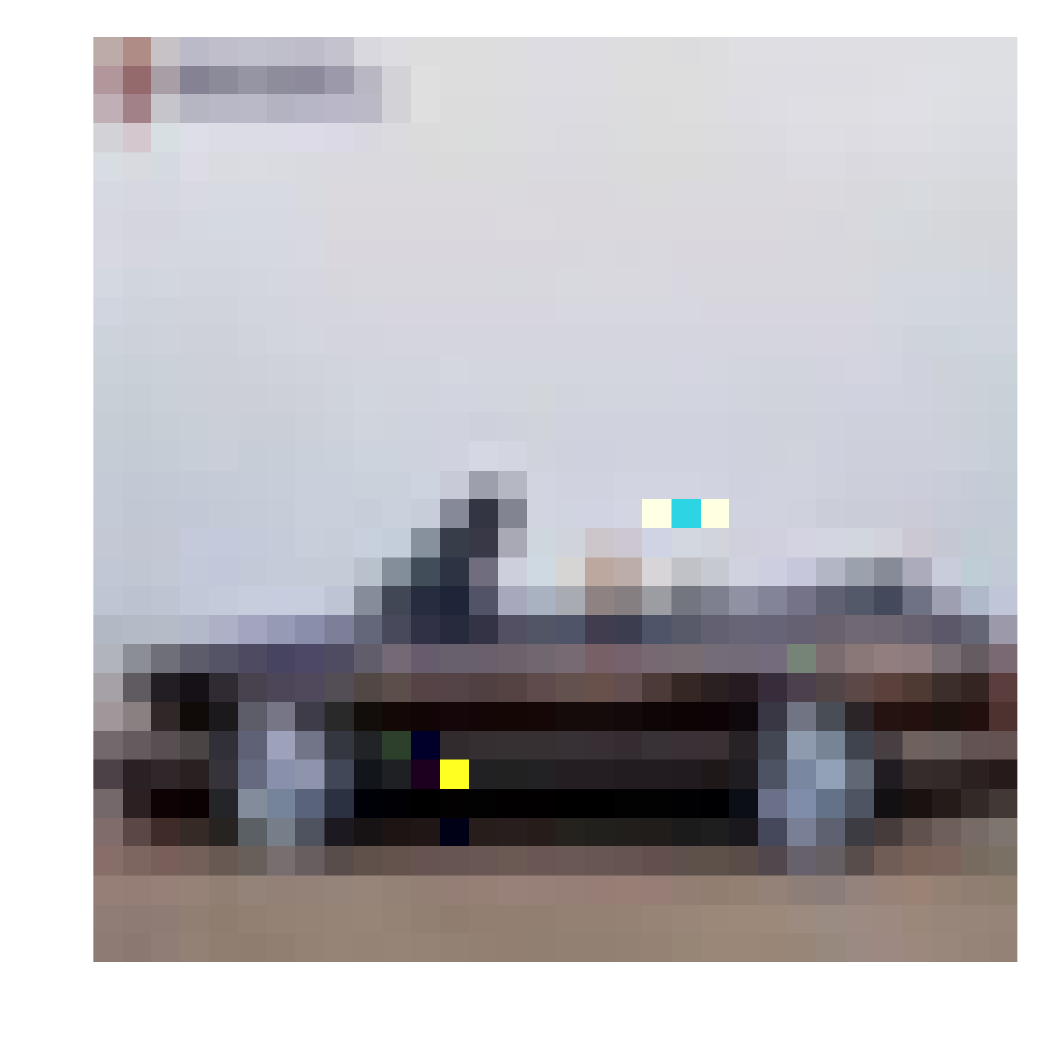}\!
        \captionsetup{font=scriptsize}
        \caption*{plane (9)}
    \end{subfigure}\!
    \begin{subfigure}[b]{0.20\linewidth}
        \includegraphics[width=\linewidth]{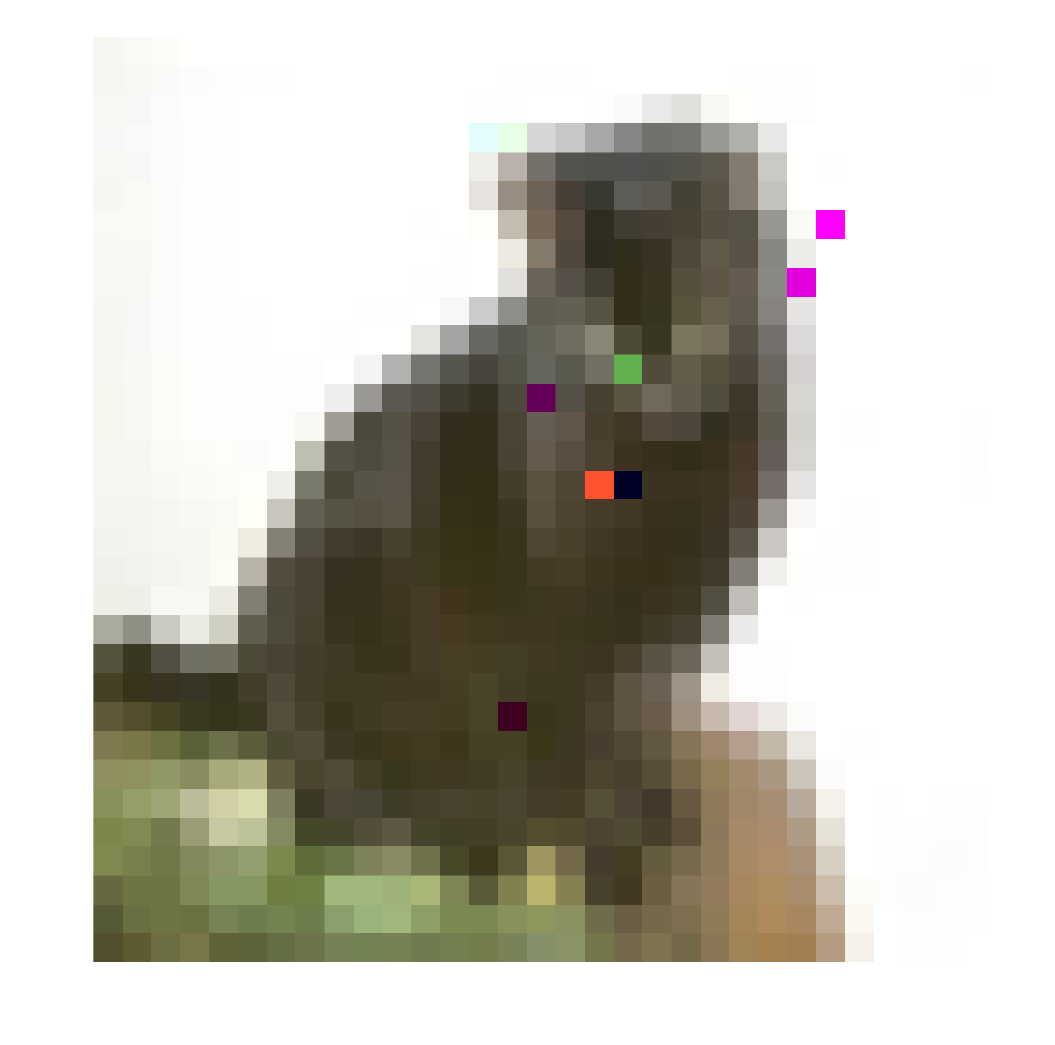}\!
        \captionsetup{font=scriptsize}
        \caption*{bird (9)}
    \end{subfigure}\!
    \begin{subfigure}[b]{0.20\linewidth}
        \includegraphics[width=\linewidth]{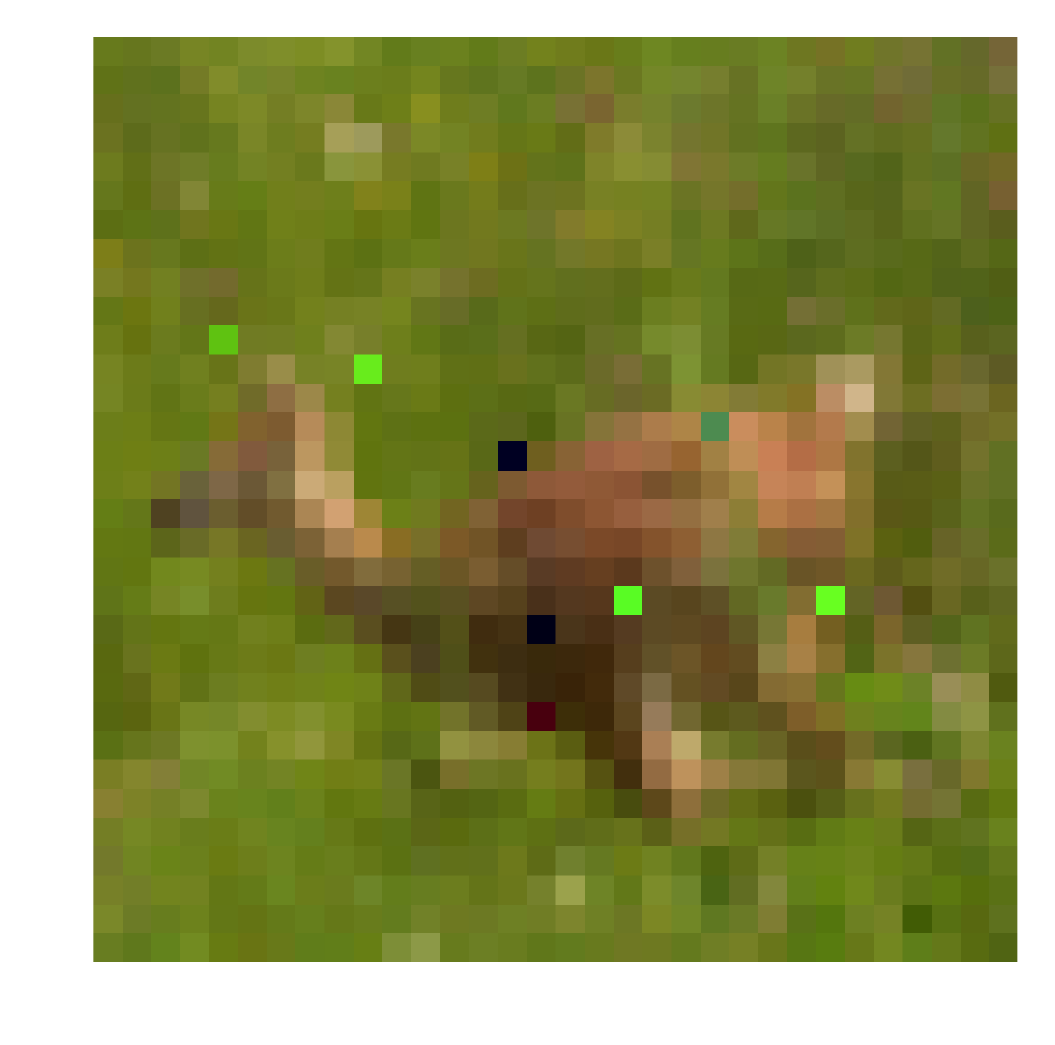}\!
        \captionsetup{font=scriptsize}
        \caption*{frog (9)}
    \end{subfigure}\!
    \begin{subfigure}[b]{0.20\linewidth}
        \includegraphics[width=\linewidth]{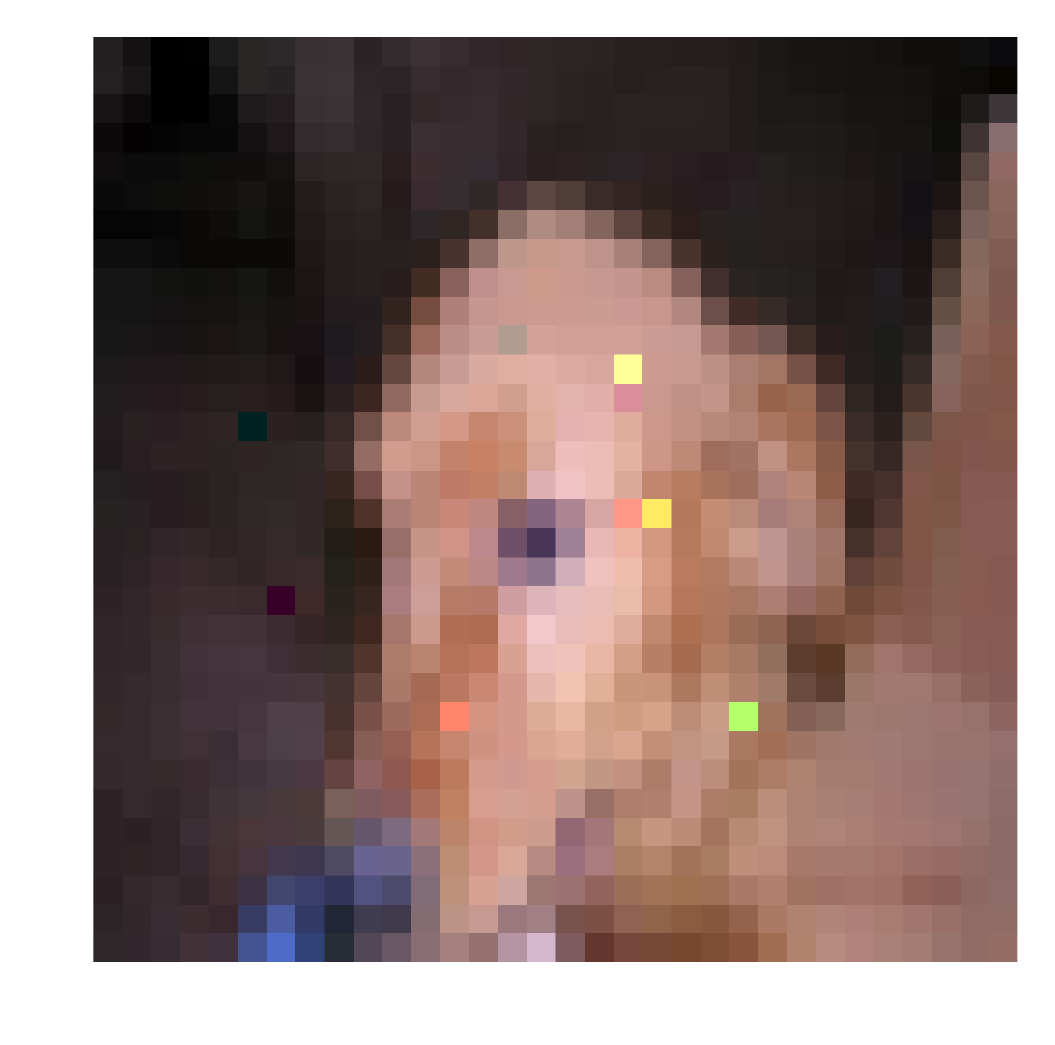}\!
        \captionsetup{font=scriptsize}
        \caption*{bird (9)}
    \end{subfigure}\!
    
    \begin{subfigure}[b]{0.20\linewidth}
        \includegraphics[width=\linewidth]{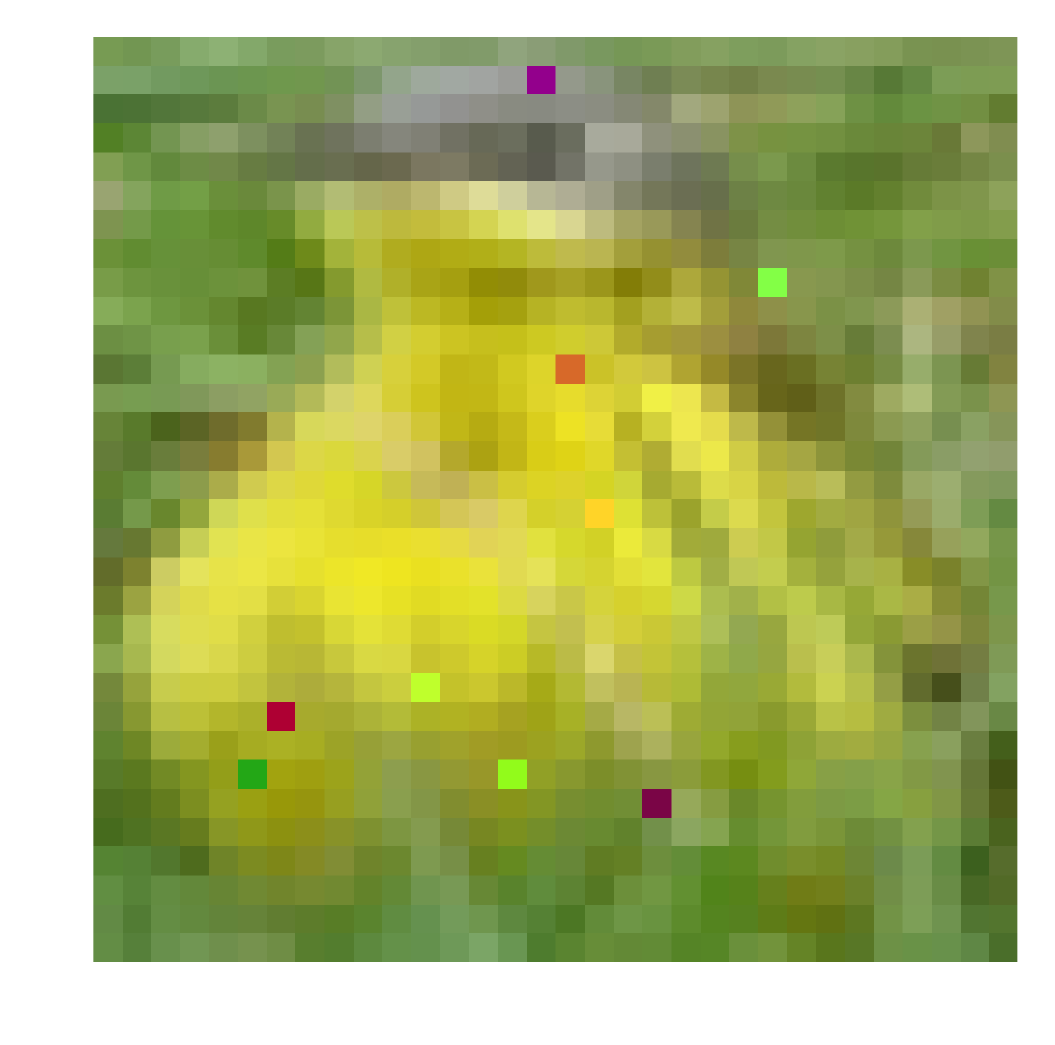}\!
        \captionsetup{font=scriptsize}
        \caption*{frog (9)}
    \end{subfigure}\!
    \begin{subfigure}[b]{0.20\linewidth}
        \includegraphics[width=\linewidth]{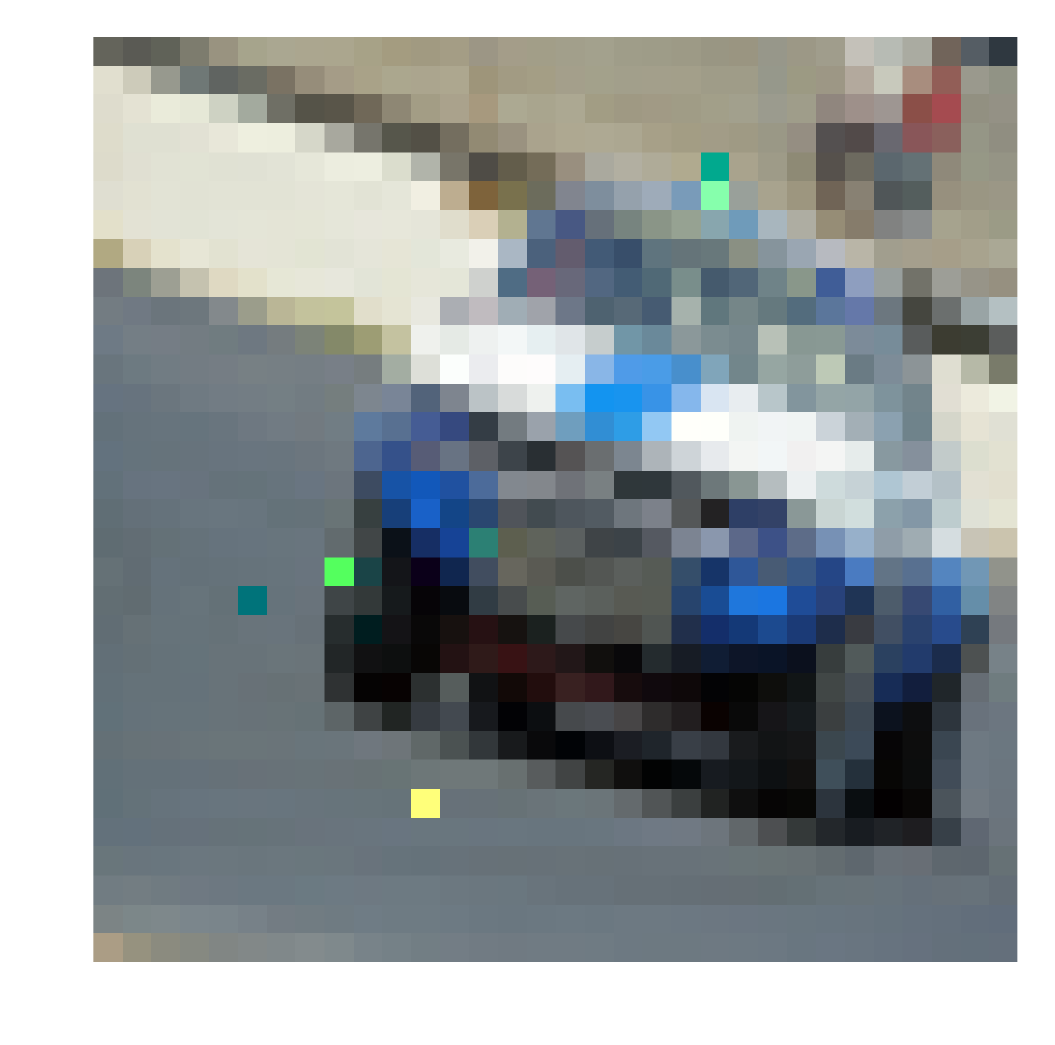}\!
        \captionsetup{font=scriptsize}
        \caption*{ship (9)}
    \end{subfigure}\!
    \begin{subfigure}[b]{0.20\linewidth}
        \includegraphics[width=\linewidth]{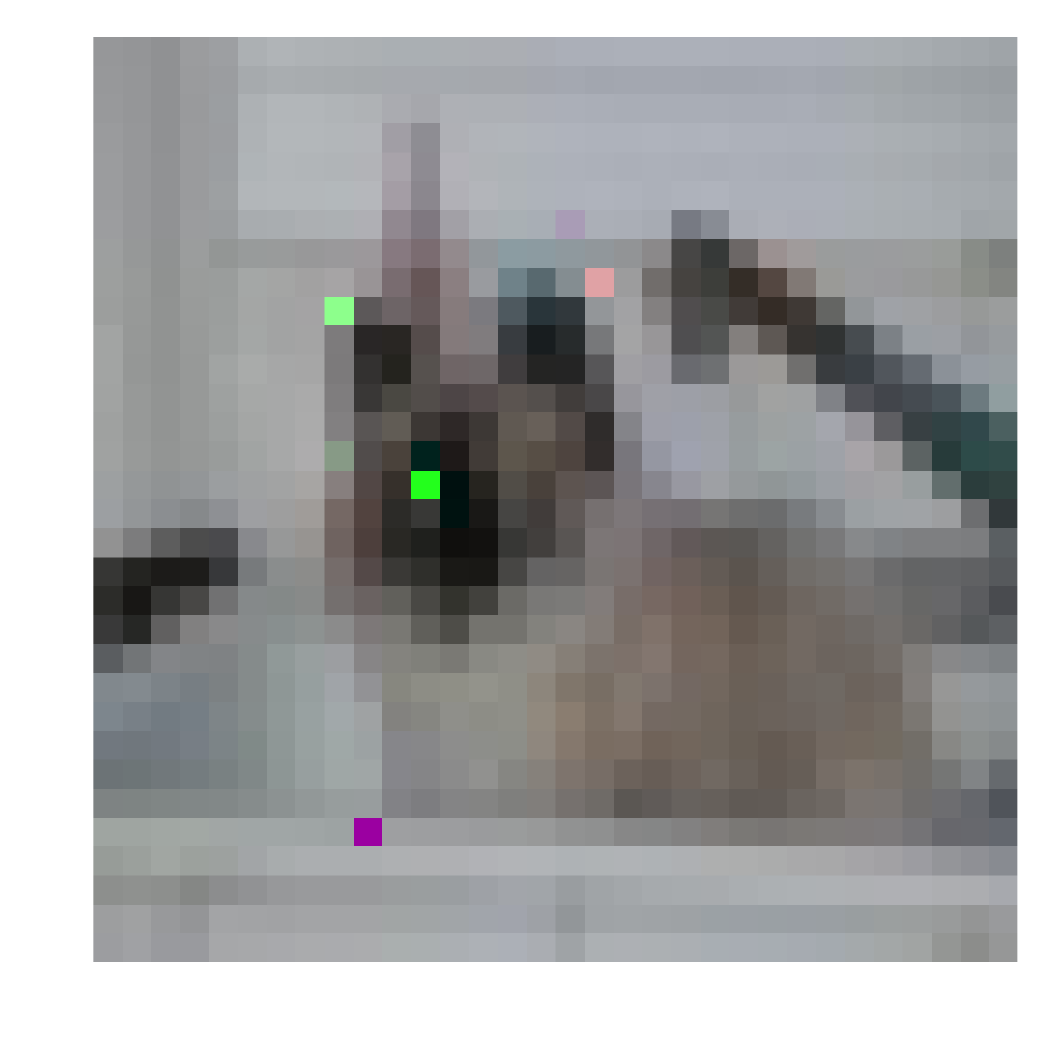}\!
        \captionsetup{font=scriptsize}
        \caption*{dog (9)}
    \end{subfigure}\!
    \begin{subfigure}[b]{0.20\linewidth}
        \includegraphics[width=\linewidth]{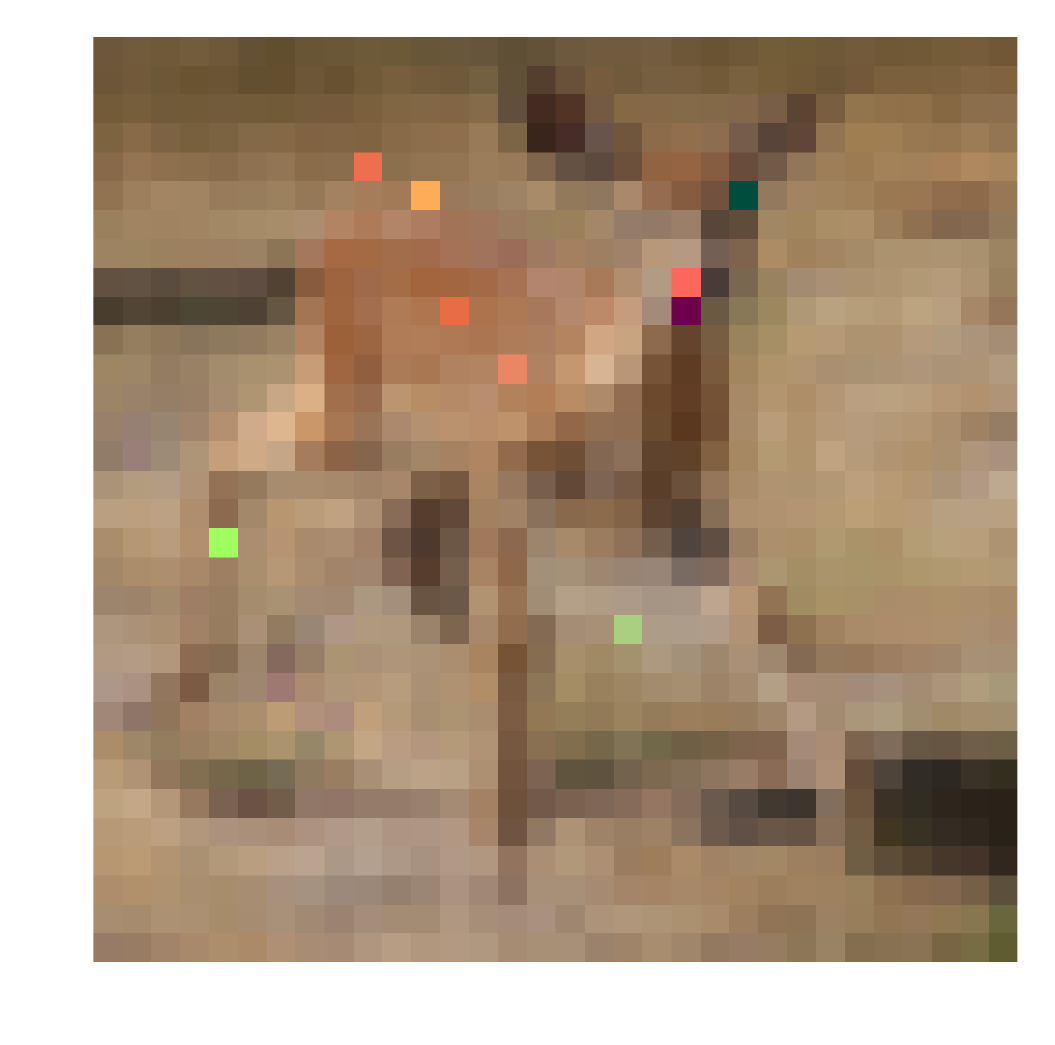}\!
        \captionsetup{font=scriptsize}
        \caption*{bird (9)}
    \end{subfigure}\!
    \begin{subfigure}[b]{0.20\linewidth}
        \includegraphics[width=\linewidth]{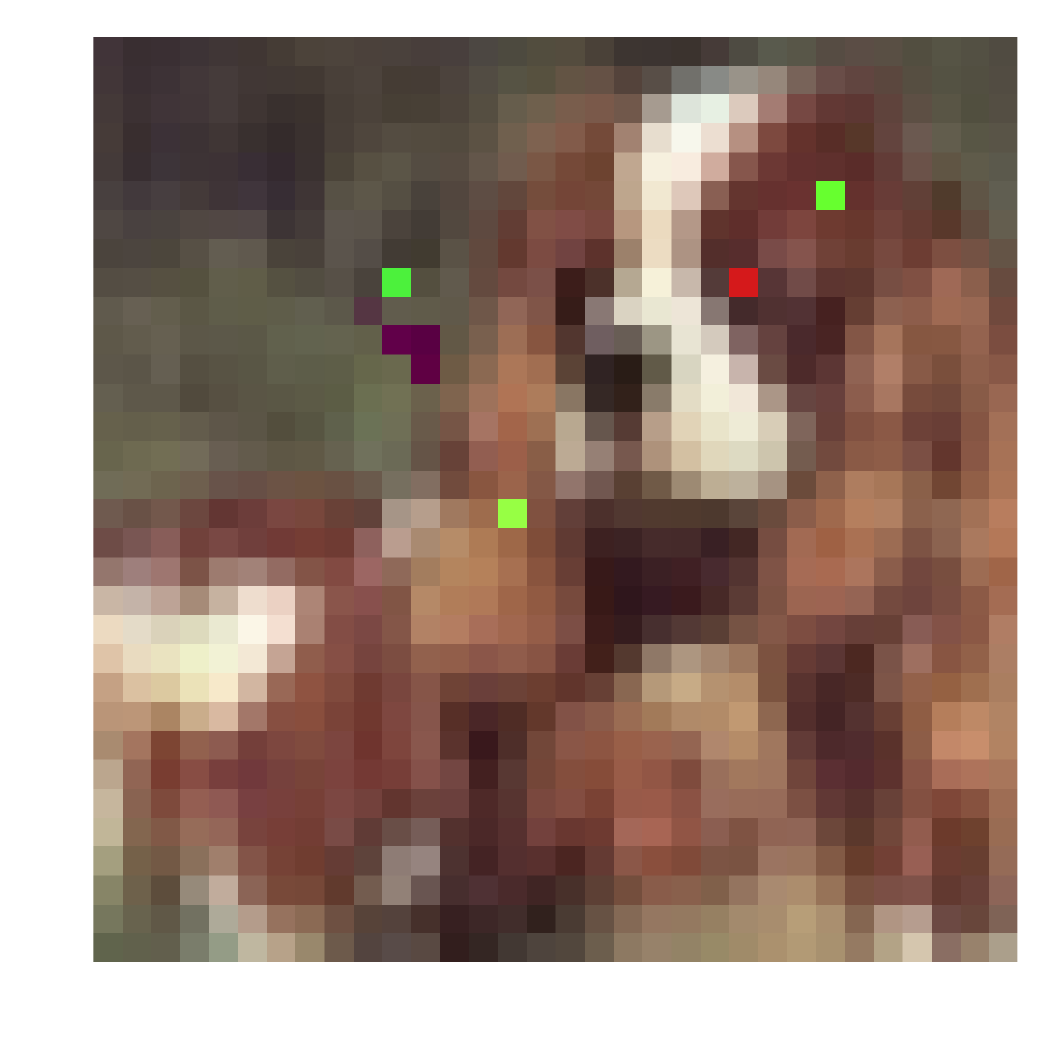}\!
        \captionsetup{font=scriptsize}
        \caption*{deer (9)}
    \end{subfigure}\!
    
    \begin{subfigure}[b]{0.20\linewidth}
        \includegraphics[width=\linewidth]{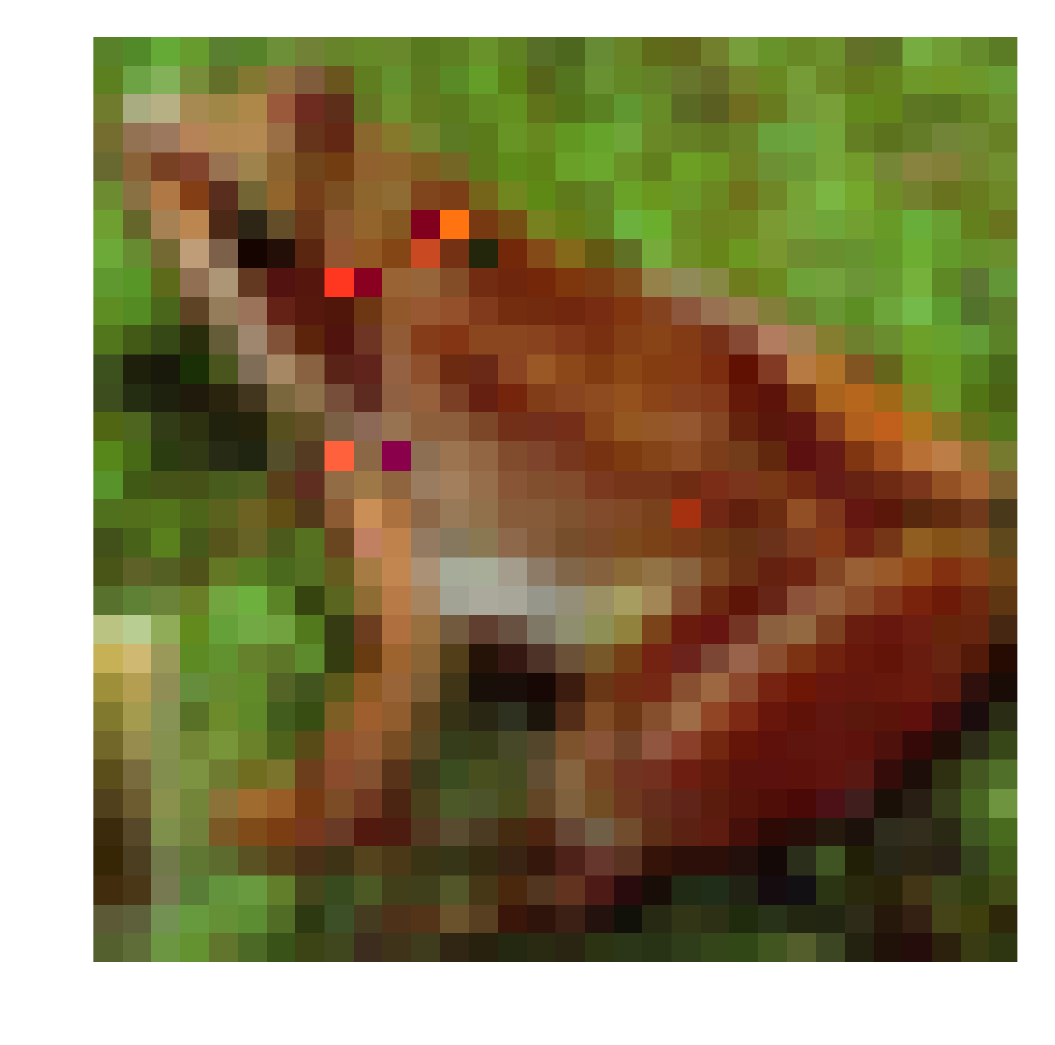}\!
        \captionsetup{font=scriptsize}
        \caption*{deer (9)}
    \end{subfigure}\!
    \begin{subfigure}[b]{0.20\linewidth}
        \includegraphics[width=\linewidth]{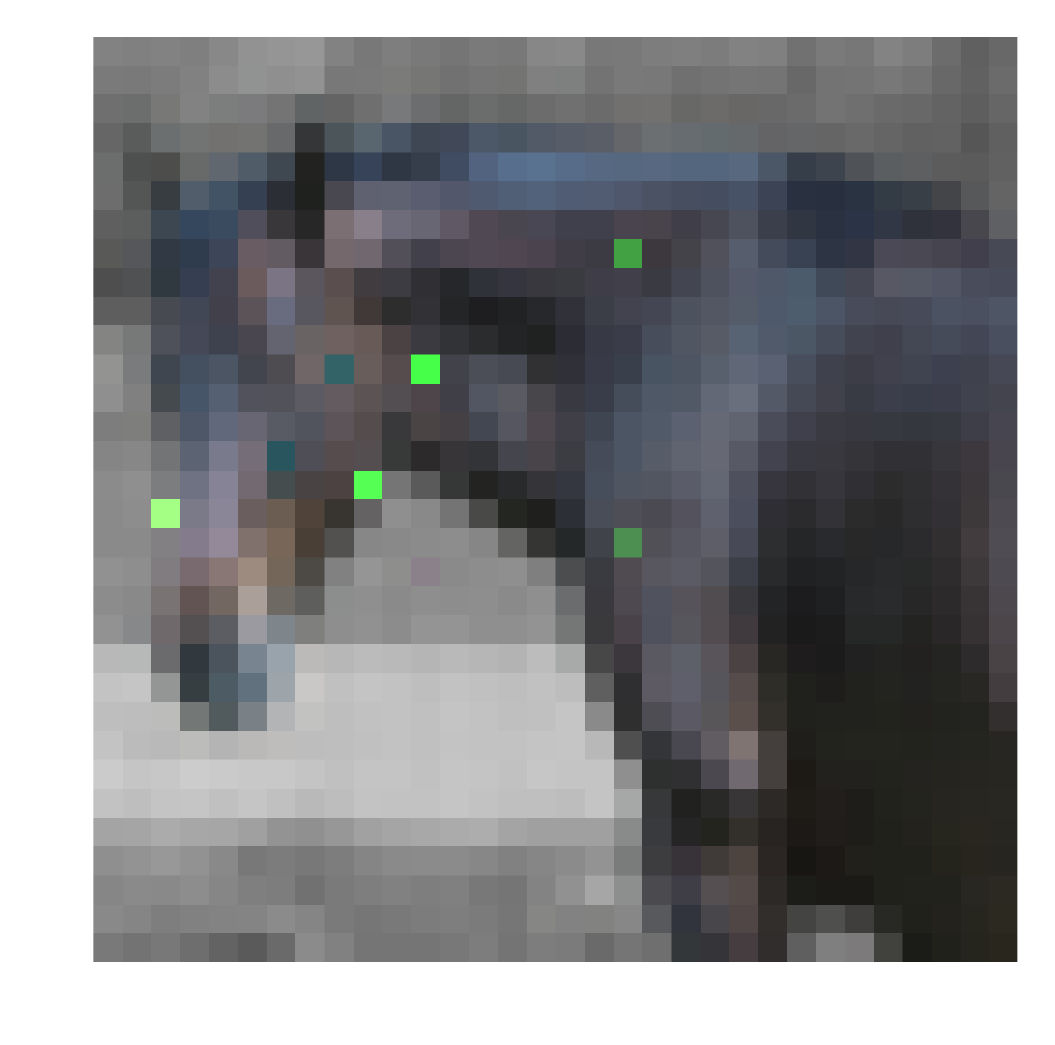}\!
        \captionsetup{font=scriptsize}
        \caption*{cat (9)}
    \end{subfigure}\!
    \begin{subfigure}[b]{0.20\linewidth}
        \includegraphics[width=\linewidth]{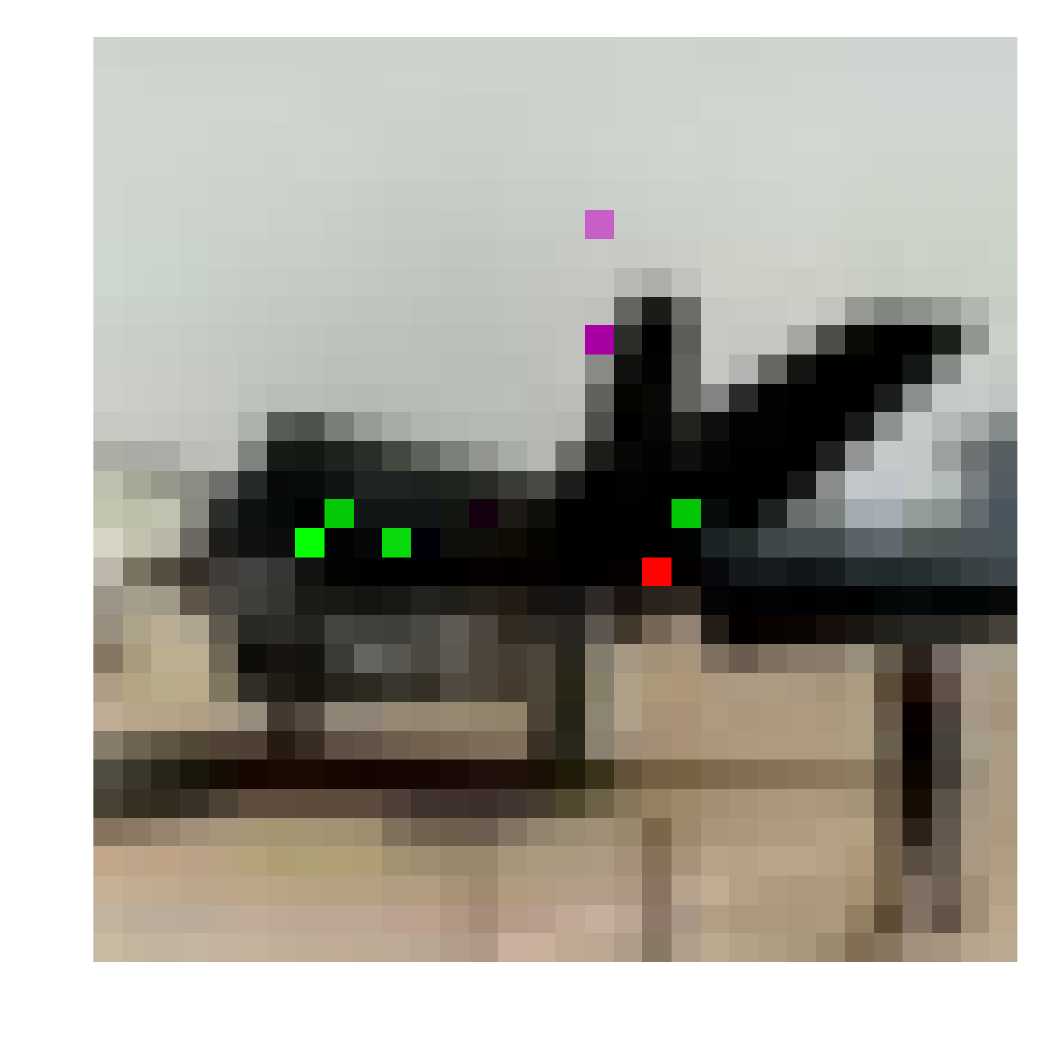}\!
        \captionsetup{font=scriptsize}
        \caption*{horse (9)}
    \end{subfigure}\!
    \begin{subfigure}[b]{0.20\linewidth}
        \includegraphics[width=\linewidth]{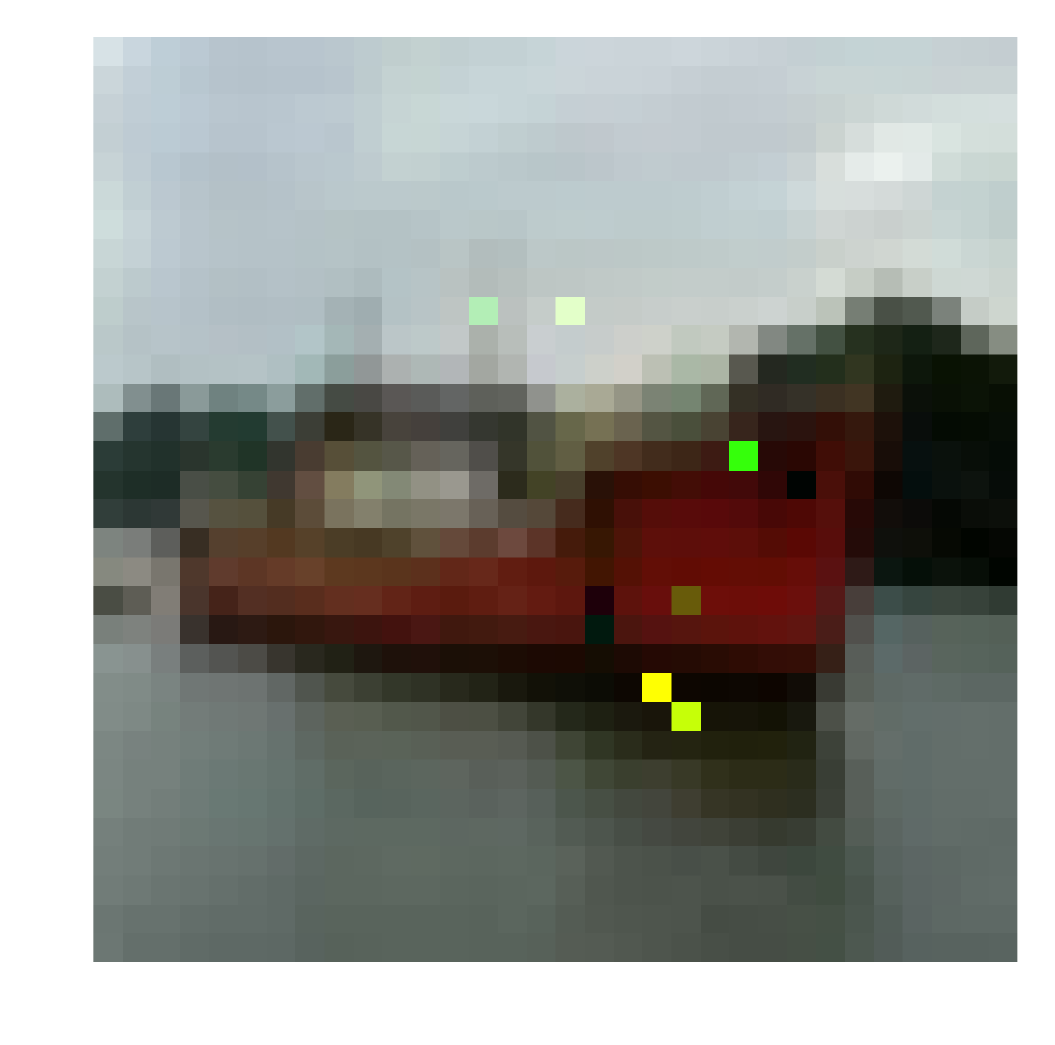}\!
        \captionsetup{font=scriptsize}
        \caption*{car (9)}
    \end{subfigure}\!
    \begin{subfigure}[b]{0.20\linewidth}
        \includegraphics[width=\linewidth]{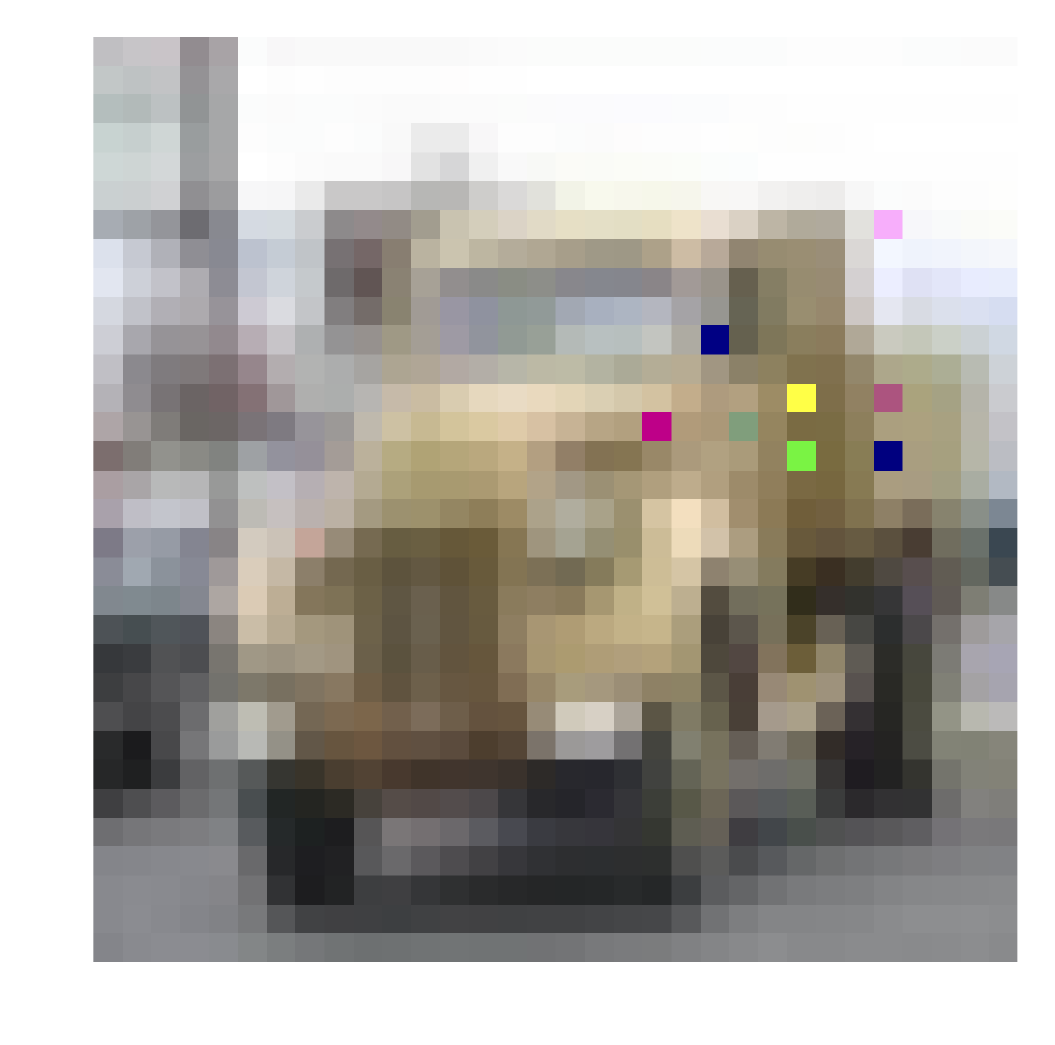}\!
        \captionsetup{font=scriptsize}
        \caption*{car (9)}
    \end{subfigure}\!
    
    \begin{subfigure}[b]{0.20\linewidth}
        \includegraphics[width=\linewidth]{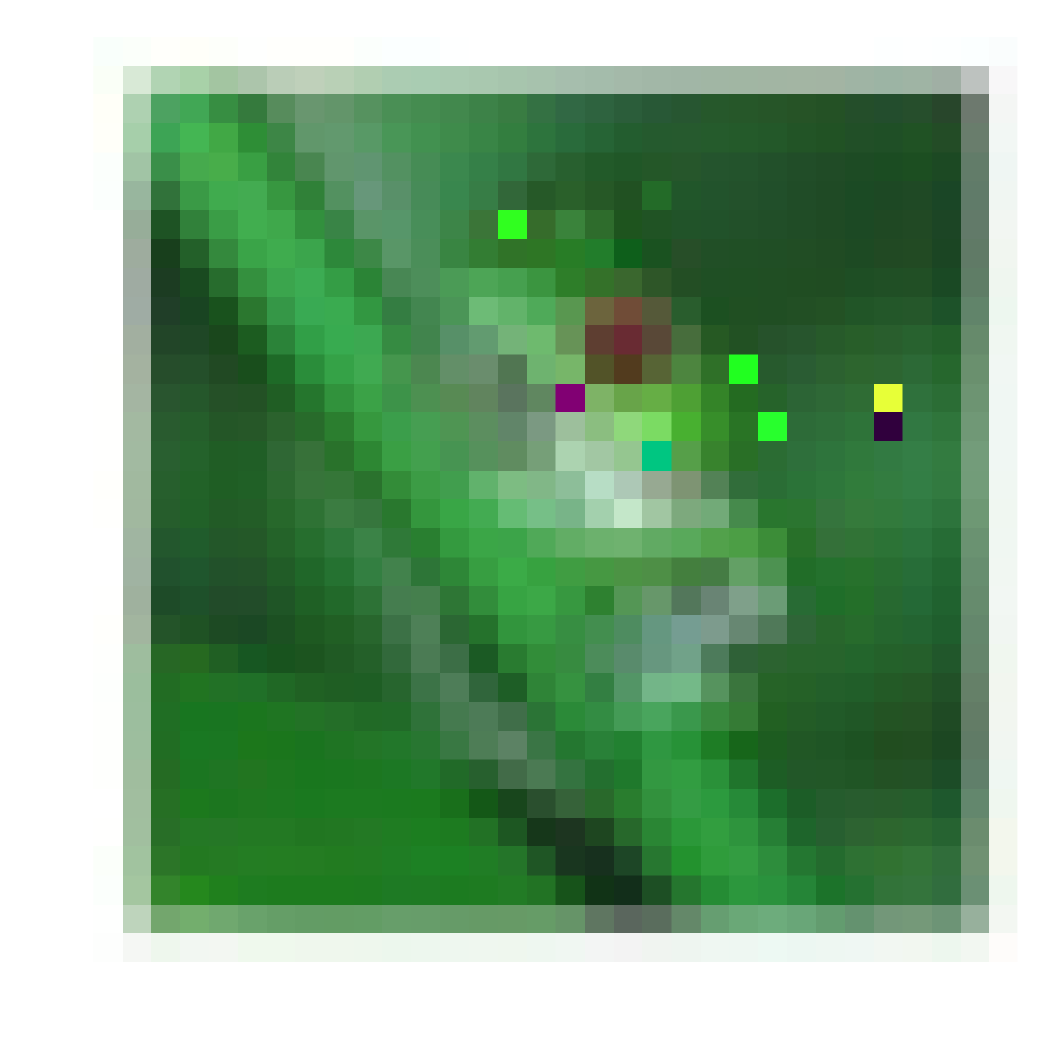}\!
        \captionsetup{font=scriptsize}
        \caption*{plane (9)}
    \end{subfigure}\!
    \begin{subfigure}[b]{0.20\linewidth}
        \includegraphics[width=\linewidth]{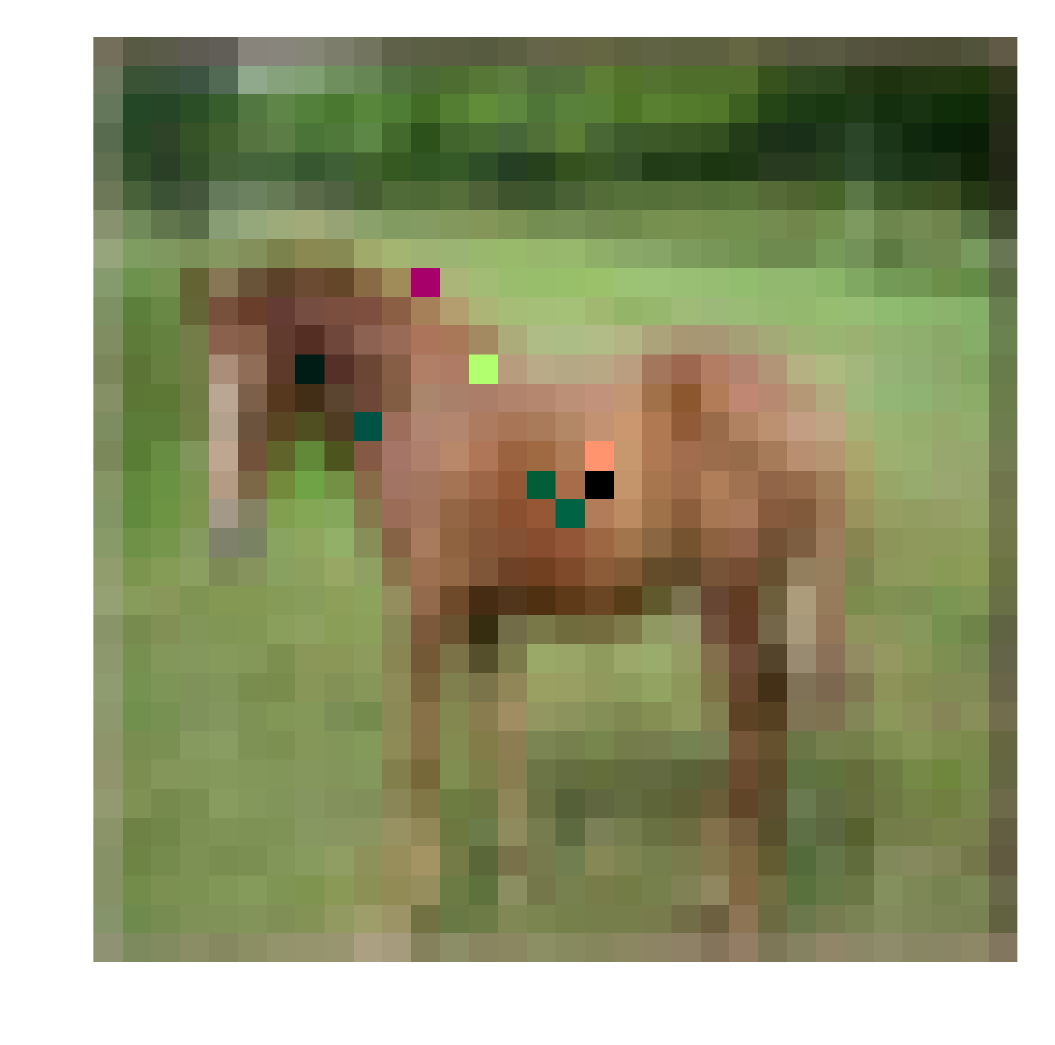}\!
        \captionsetup{font=scriptsize}
        \caption*{dog (9)}
    \end{subfigure}\!
    \begin{subfigure}[b]{0.20\linewidth}
        \includegraphics[width=\linewidth]{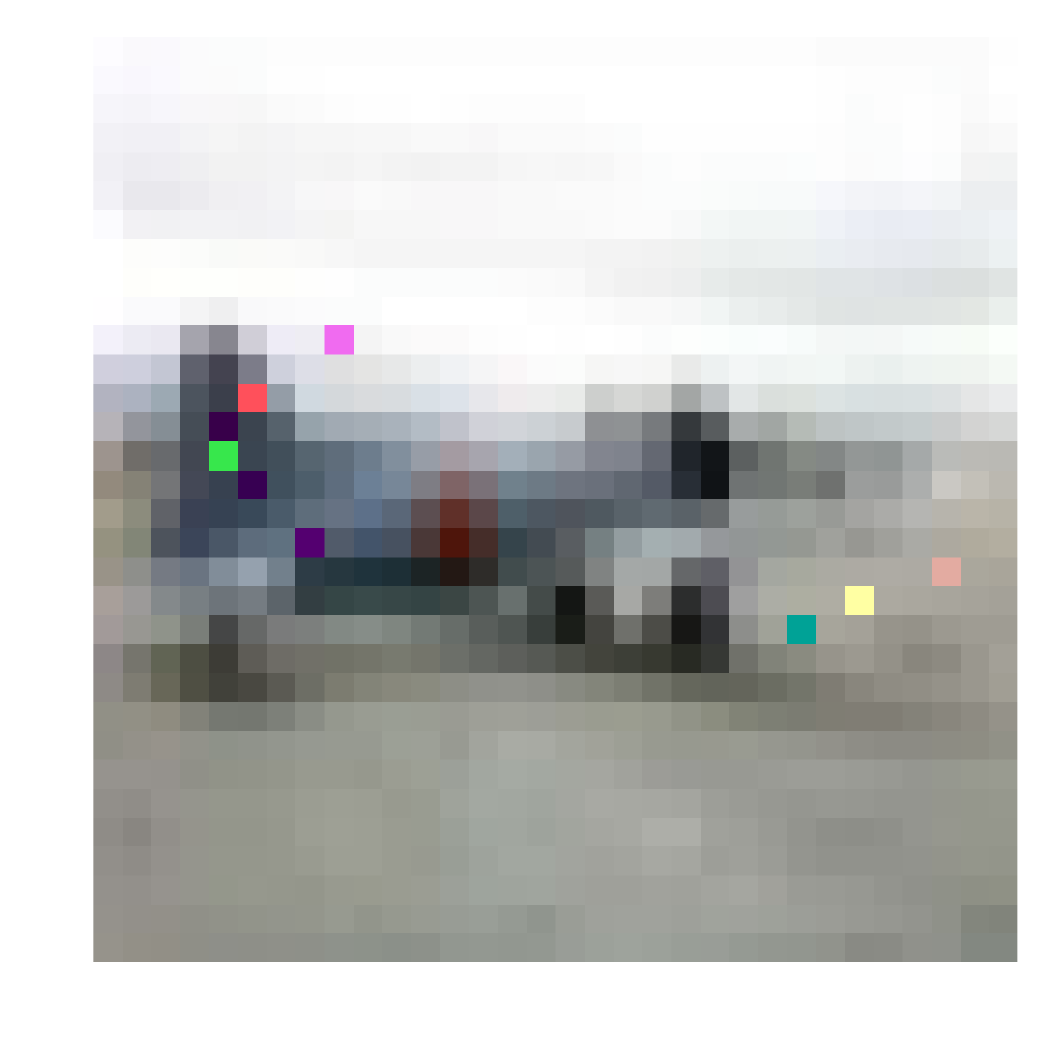}\!
        \captionsetup{font=scriptsize}
        \caption*{truck (9)}
    \end{subfigure}\!
    \begin{subfigure}[b]{0.20\linewidth}
        \includegraphics[width=\linewidth]{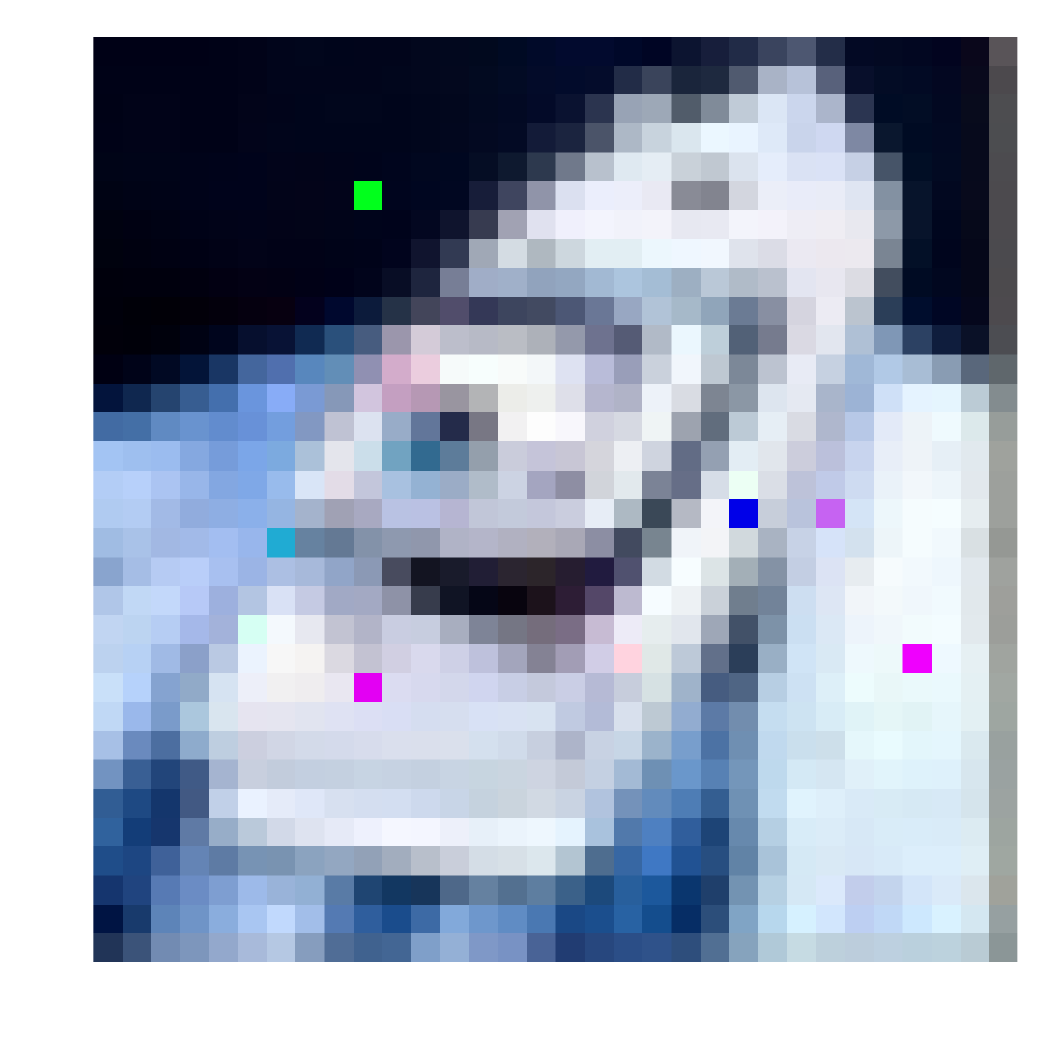}\!
        \captionsetup{font=scriptsize}
        \caption*{horse (9)}
    \end{subfigure}\!
    \begin{subfigure}[b]{0.20\linewidth}
        \includegraphics[width=\linewidth]{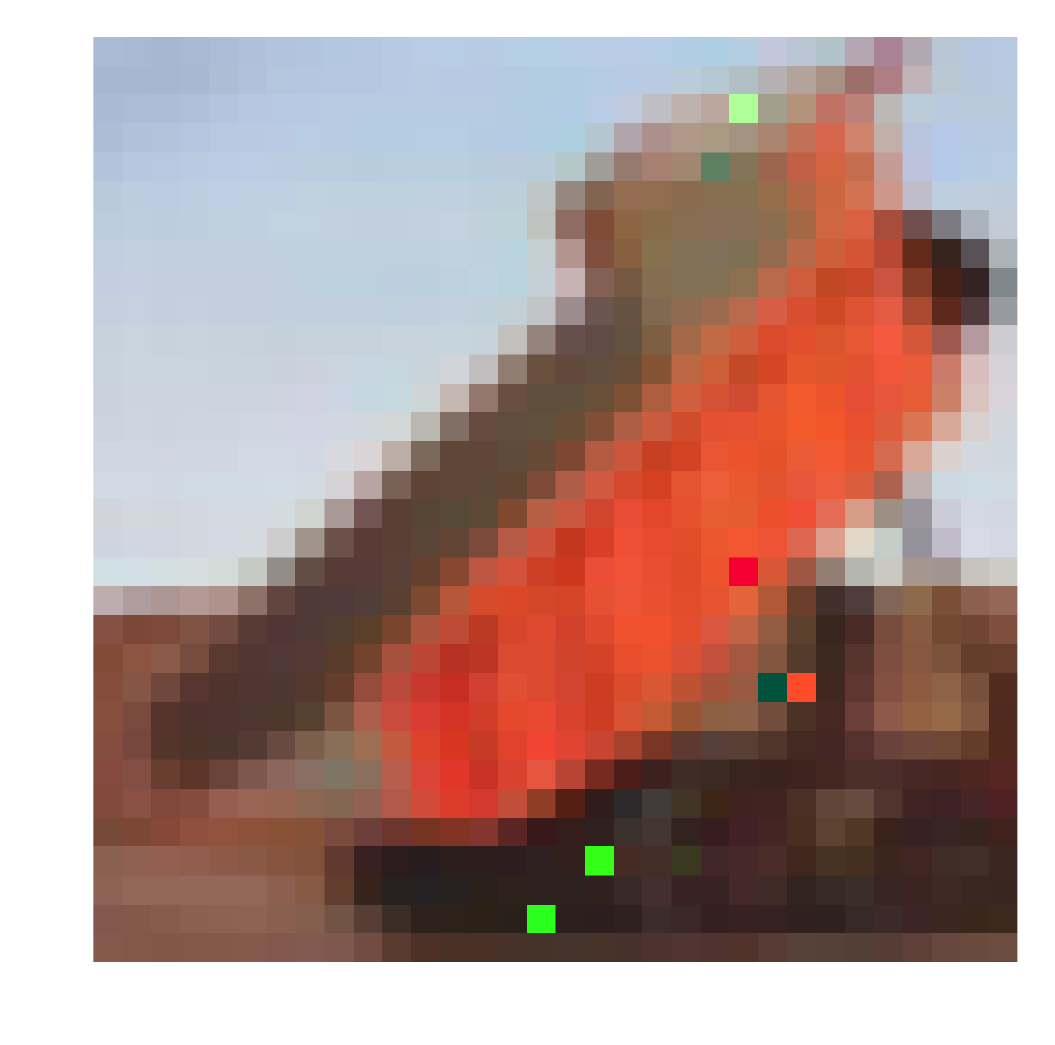}\!
        \captionsetup{font=scriptsize}
        \caption*{plane (9)}
    \end{subfigure}\!
\captionsetup{font=small, skip=8pt}
\caption{Sparse perturbations.}
\label{fig:cifar_sparse}
\end{subfigure}\hfill
\caption{SparseFool adversarial examples for the CIFAR-10 dataset for different levels of sparsity. The fooling label is shown below the image, and the number of perturbed pixels is written inside the parentheses.}
\label{fig:cifar_visual}
\end{figure}

\begin{figure}[th]
\centering
\begin{subfigure}[b]{0.3\linewidth}
\captionsetup[subfigure]{skip=1pt}
    \begin{subfigure}[b]{0.2\linewidth}
        \includegraphics[width=\linewidth]{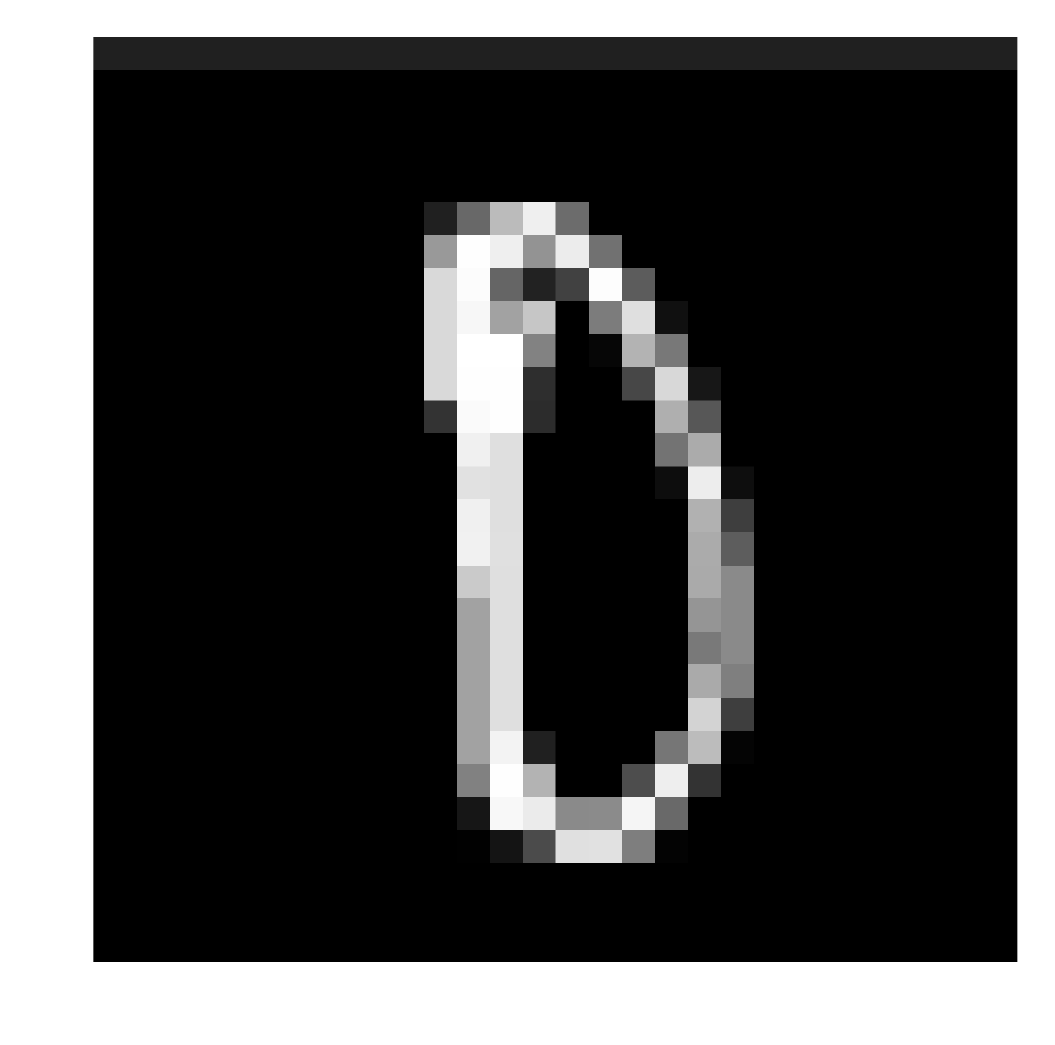}\!
        \captionsetup{font=scriptsize}
        \caption*{$8$ (1)}
    \end{subfigure}\!
    \begin{subfigure}[b]{0.2\linewidth}
        \includegraphics[width=\linewidth]{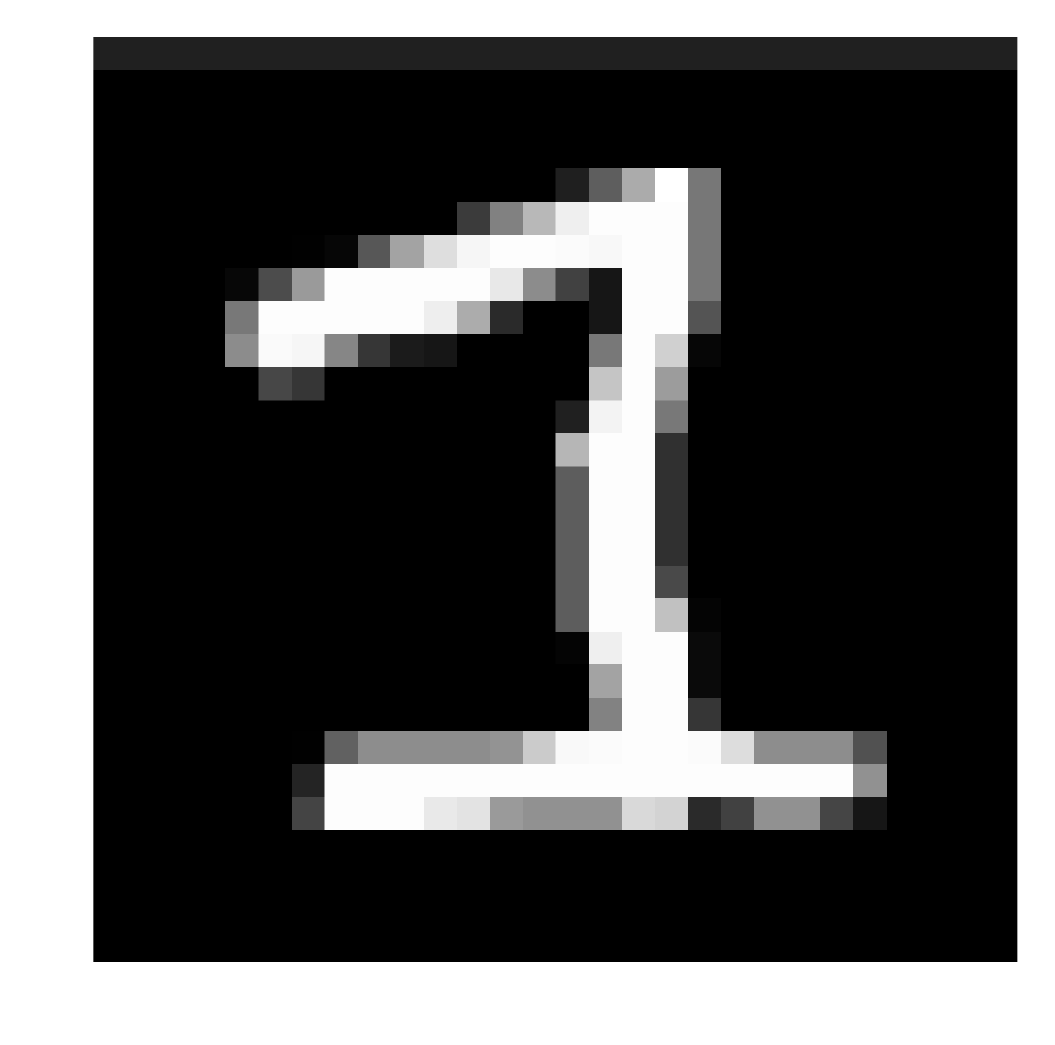}\!
        \captionsetup{font=scriptsize}
        \caption*{$3$ (1)}
    \end{subfigure}\!
    \begin{subfigure}[b]{0.2\linewidth}
        \includegraphics[width=\linewidth]{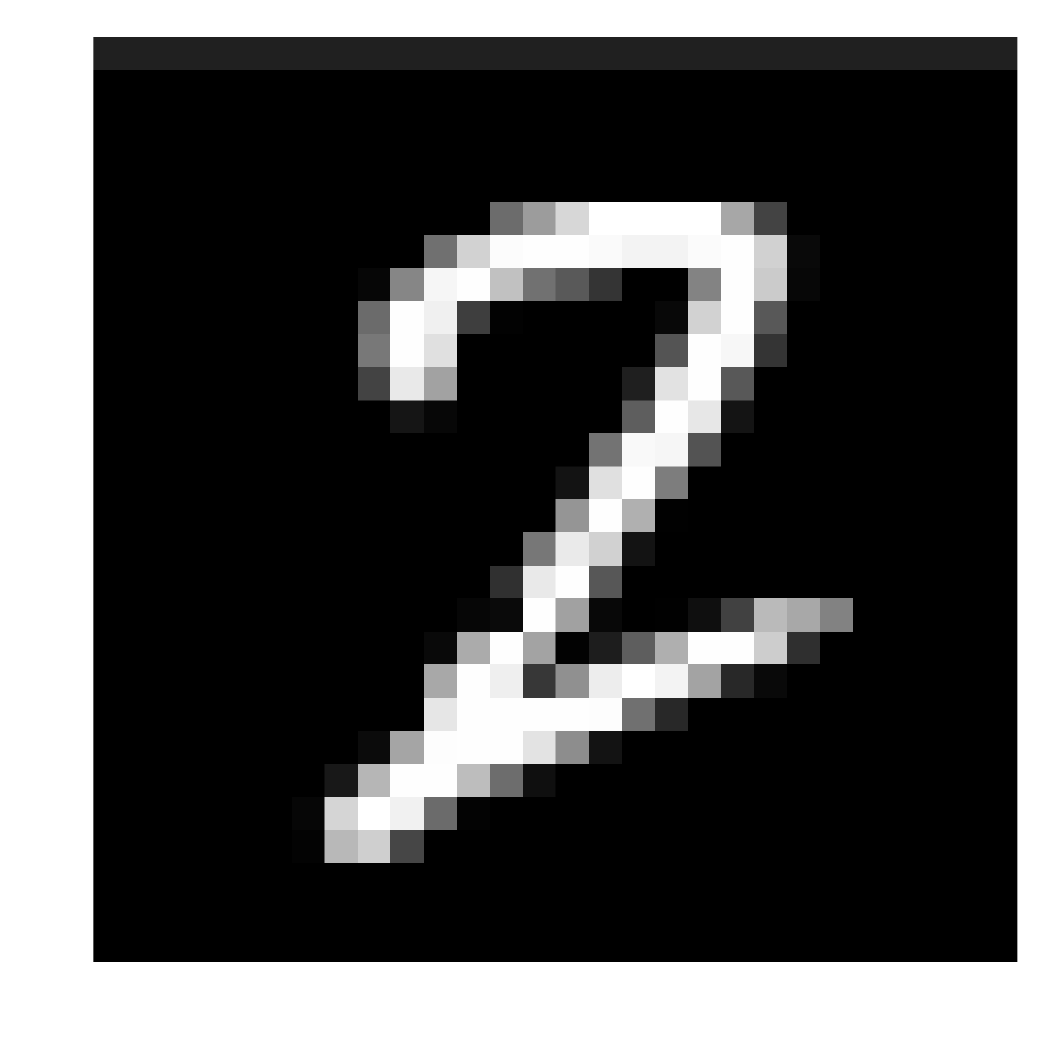}\!
        \captionsetup{font=scriptsize}
        \caption*{$7$ (1)}
    \end{subfigure}\!
    \begin{subfigure}[b]{0.2\linewidth}
        \includegraphics[width=\linewidth]{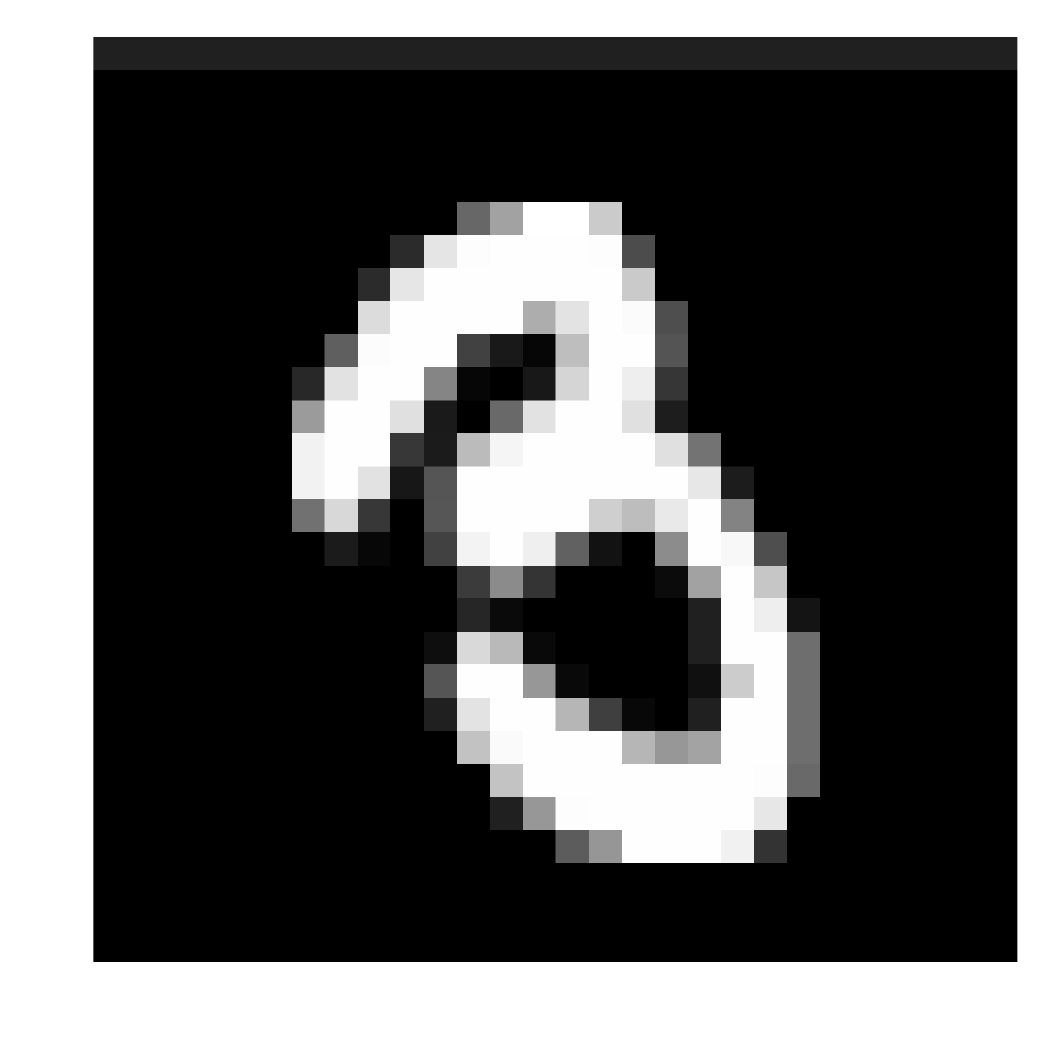}\!
        \captionsetup{font=scriptsize}
        \caption*{$8$ (1)}
    \end{subfigure}\!
    \begin{subfigure}[b]{0.2\linewidth}
        \includegraphics[width=\linewidth]{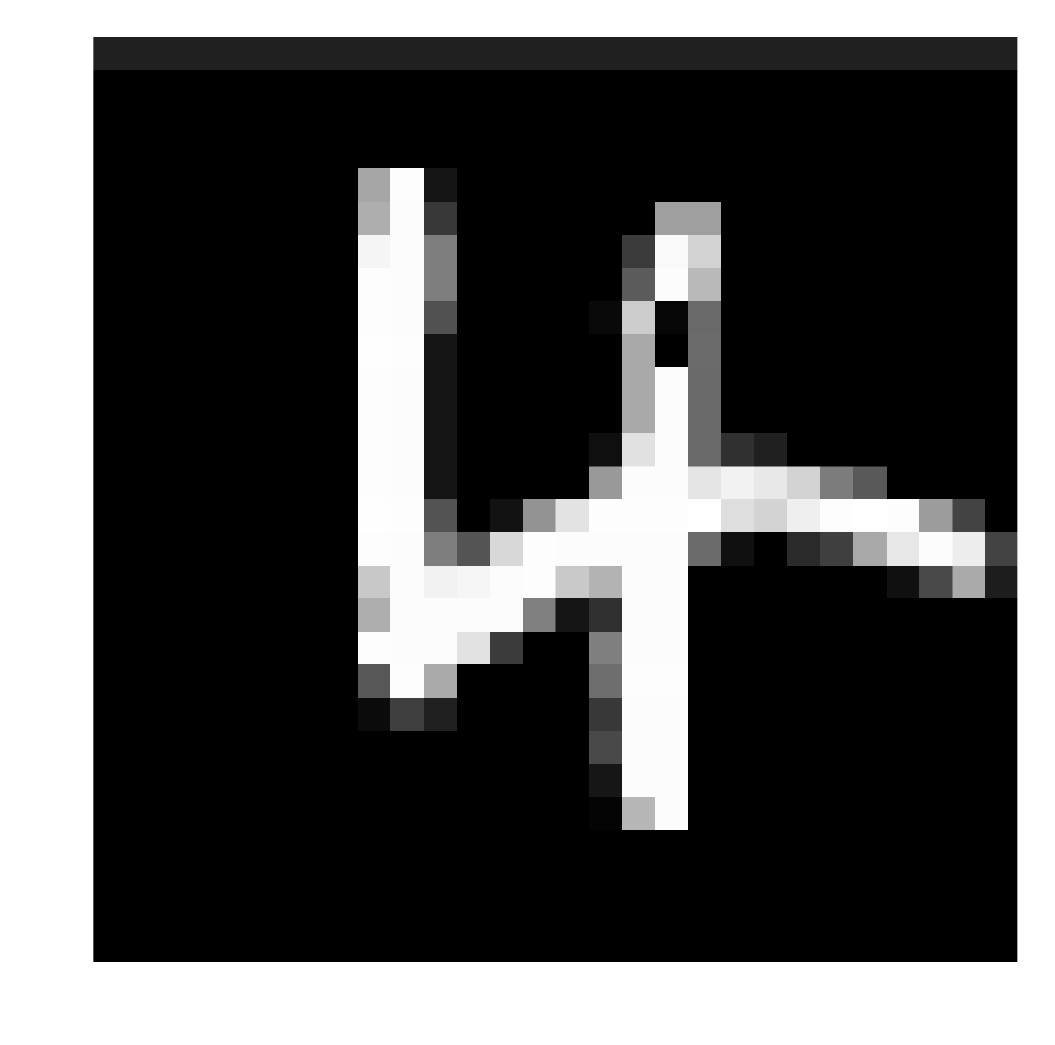}\!
        \captionsetup{font=scriptsize}
        \caption*{$6$ (2)}
    \end{subfigure}\!
    
    \begin{subfigure}[b]{0.2\linewidth}
        \includegraphics[width=\linewidth]{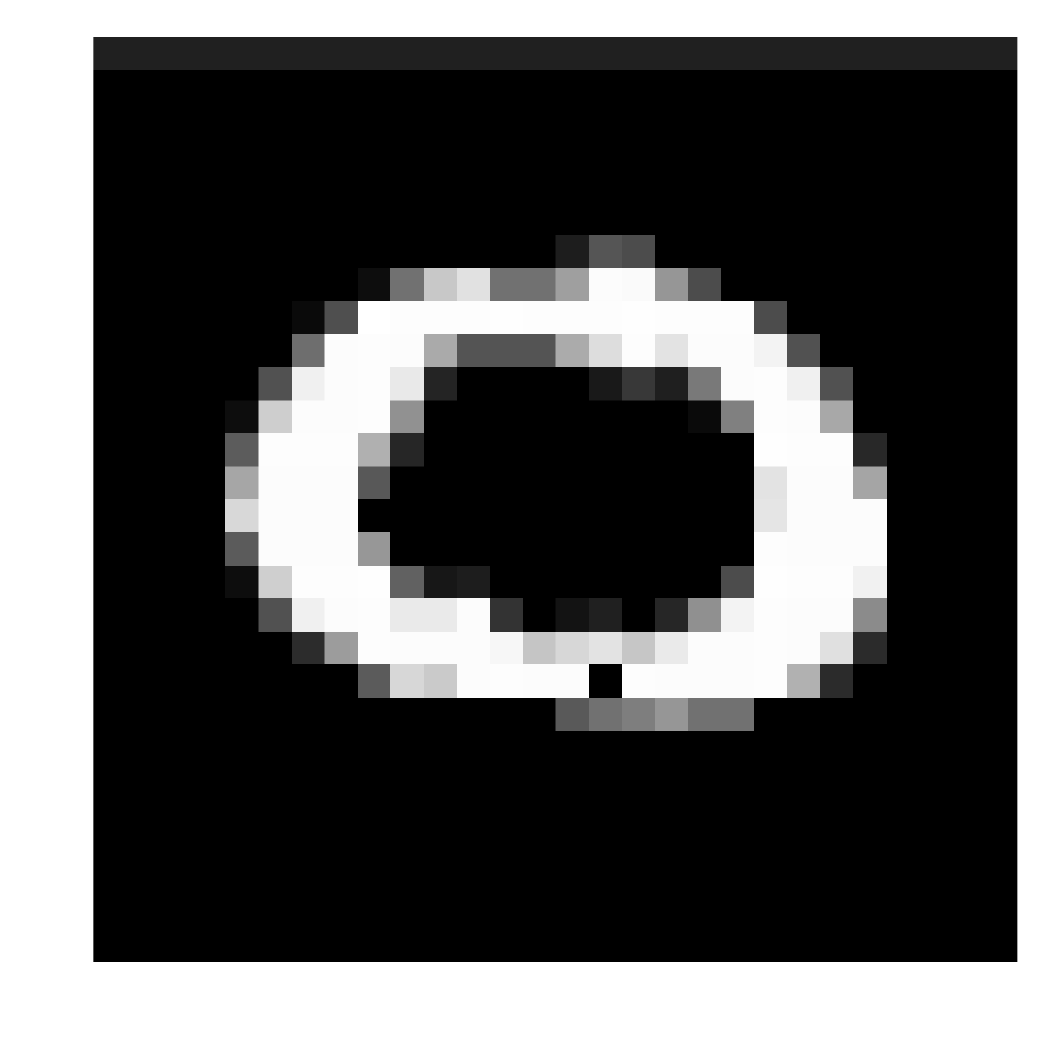}\!
        \captionsetup{font=scriptsize}
        \caption*{$9$ (1)}
    \end{subfigure}\!
    \begin{subfigure}[b]{0.2\linewidth}
        \includegraphics[width=\linewidth]{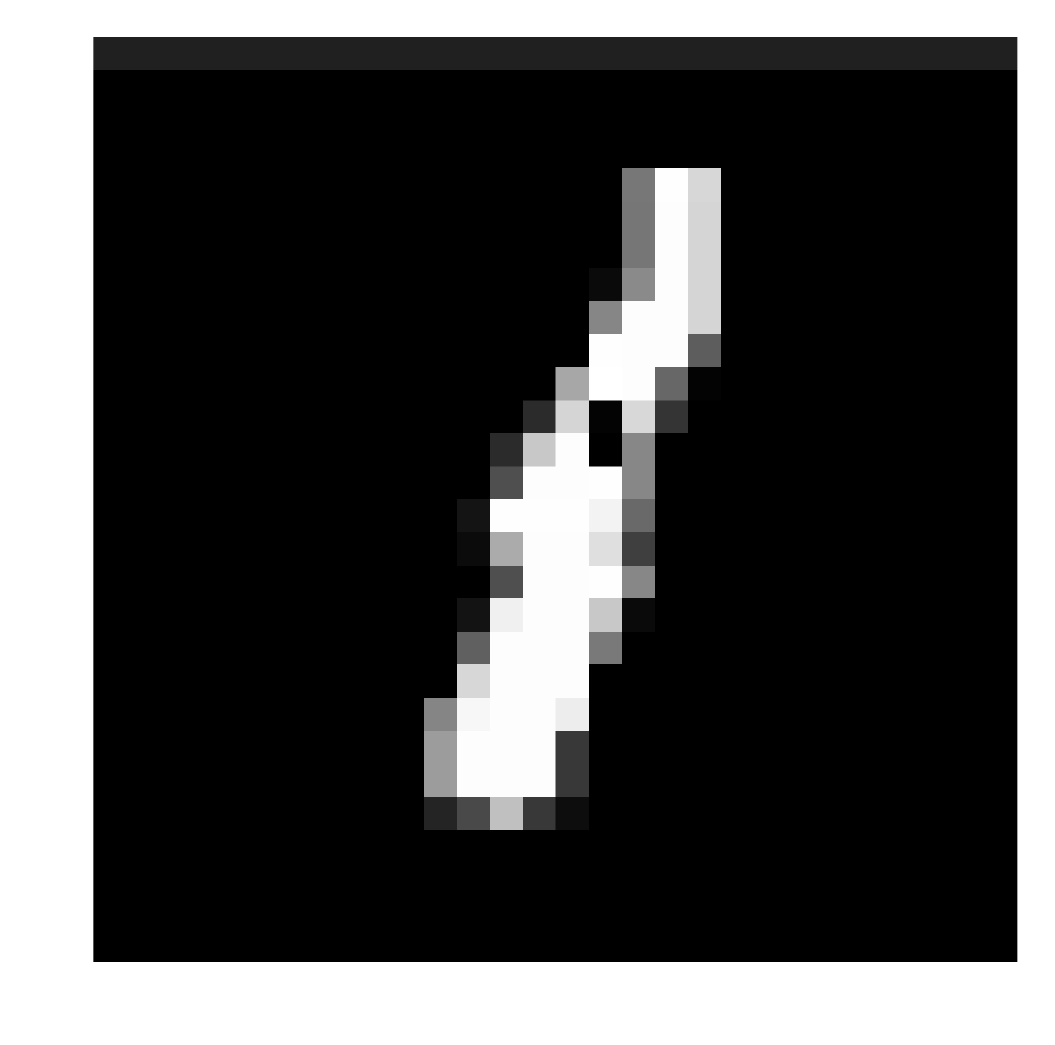}\!
        \captionsetup{font=scriptsize}
        \caption*{$8$ (2)}
    \end{subfigure}\!
    \begin{subfigure}[b]{0.2\linewidth}
        \includegraphics[width=\linewidth]{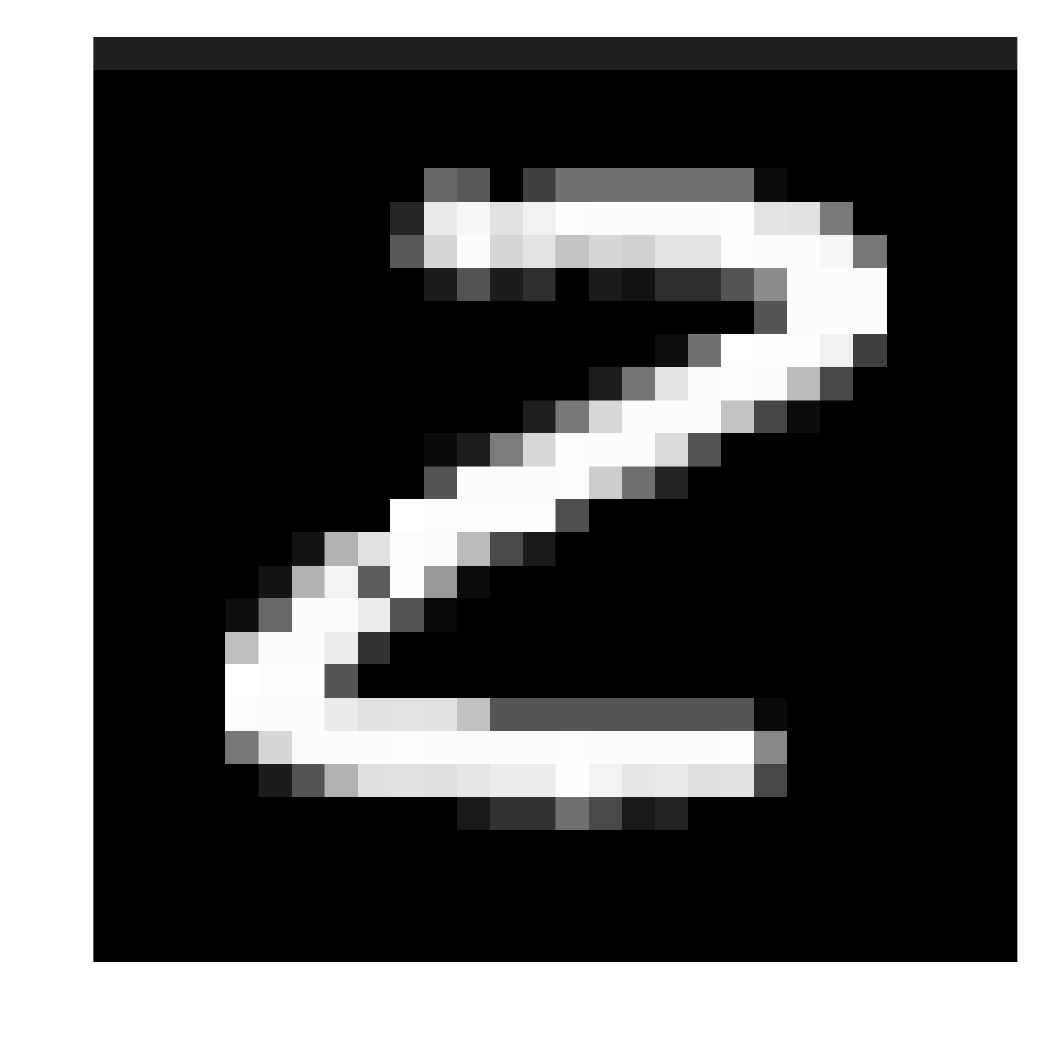}\!
        \captionsetup{font=scriptsize}
        \caption*{$3$ (2)}
    \end{subfigure}\!
    \begin{subfigure}[b]{0.2\linewidth}
        \includegraphics[width=\linewidth]{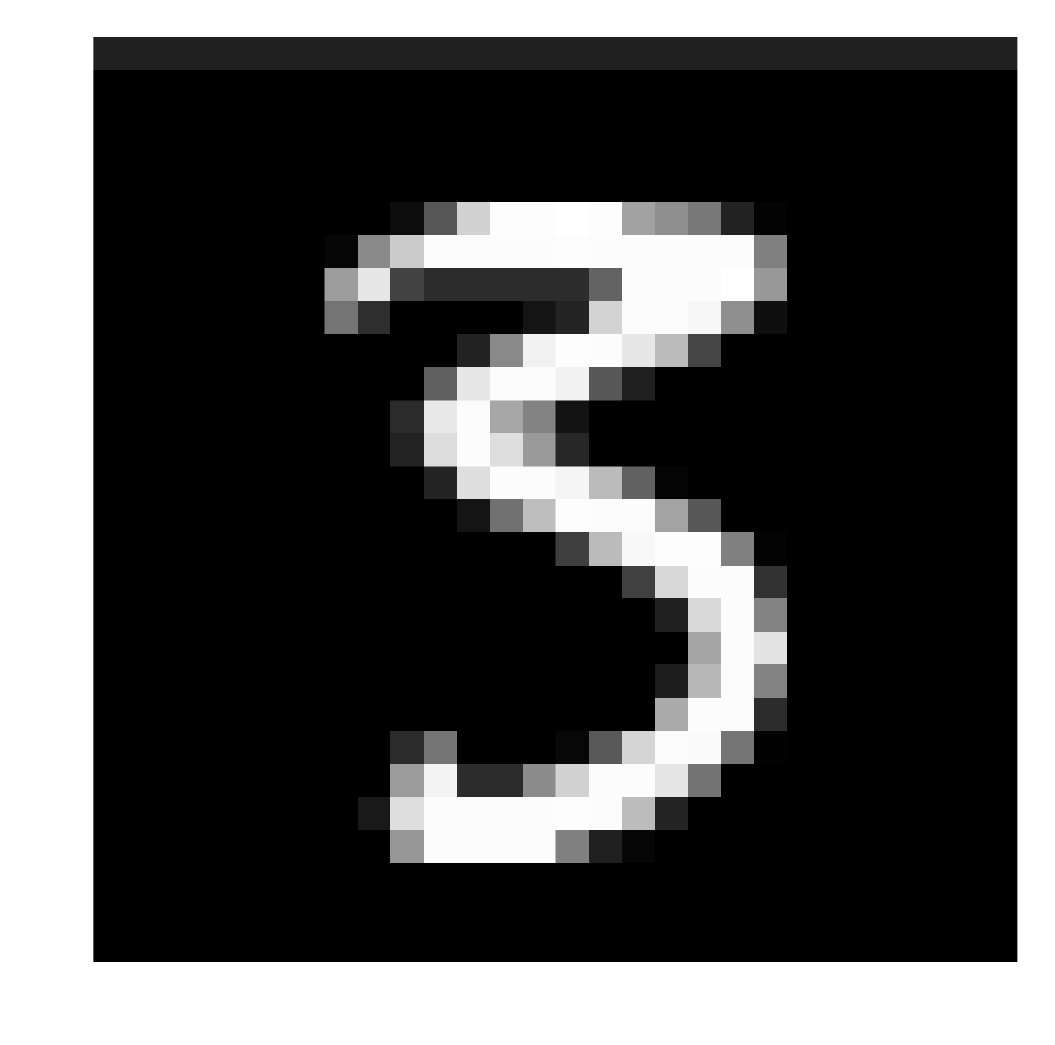}\!
        \captionsetup{font=scriptsize}
        \caption*{$5$ (2)}
    \end{subfigure}\!
    \begin{subfigure}[b]{0.2\linewidth}
        \includegraphics[width=\linewidth]{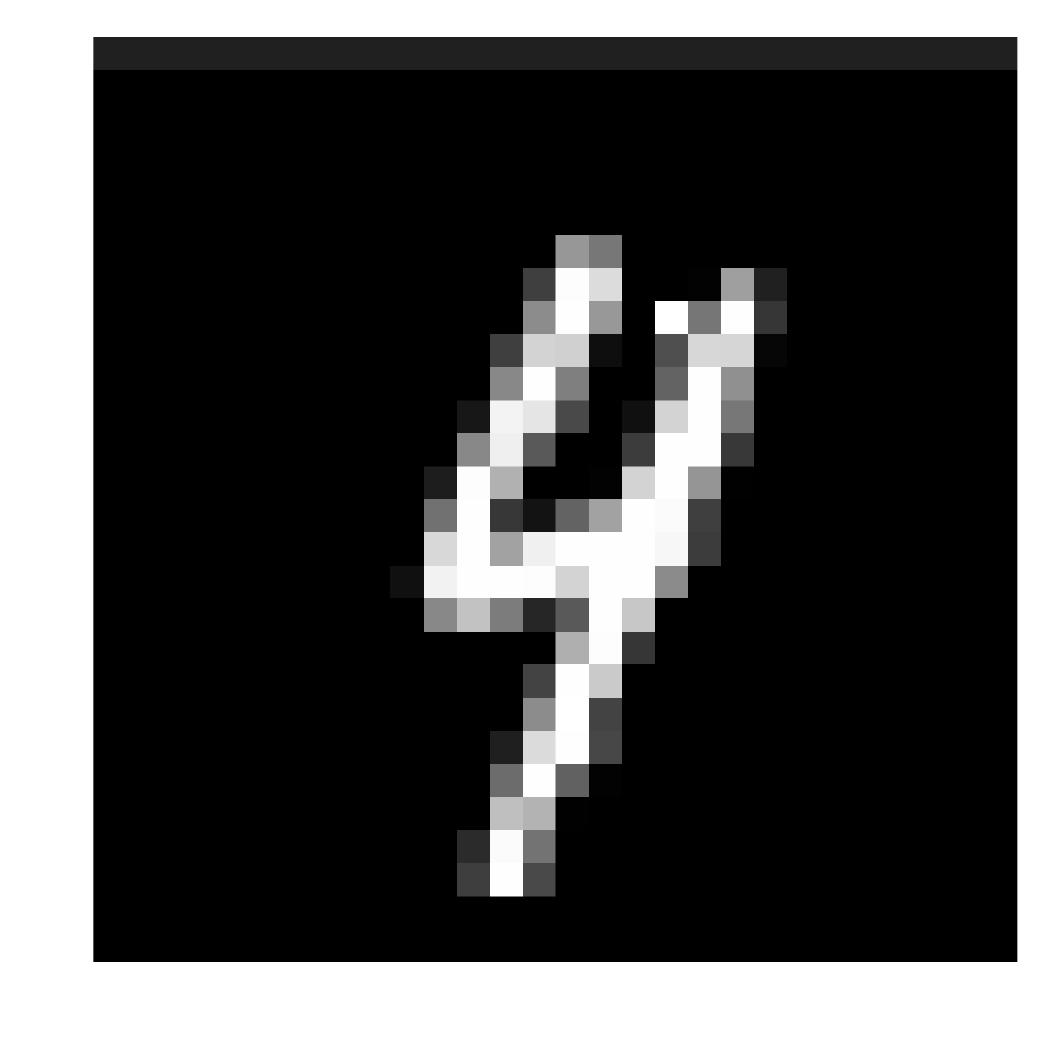}\!
        \captionsetup{font=scriptsize}
        \caption*{$9$ (2)}
    \end{subfigure}\!
    
    \begin{subfigure}[b]{0.2\linewidth}
        \includegraphics[width=\linewidth]{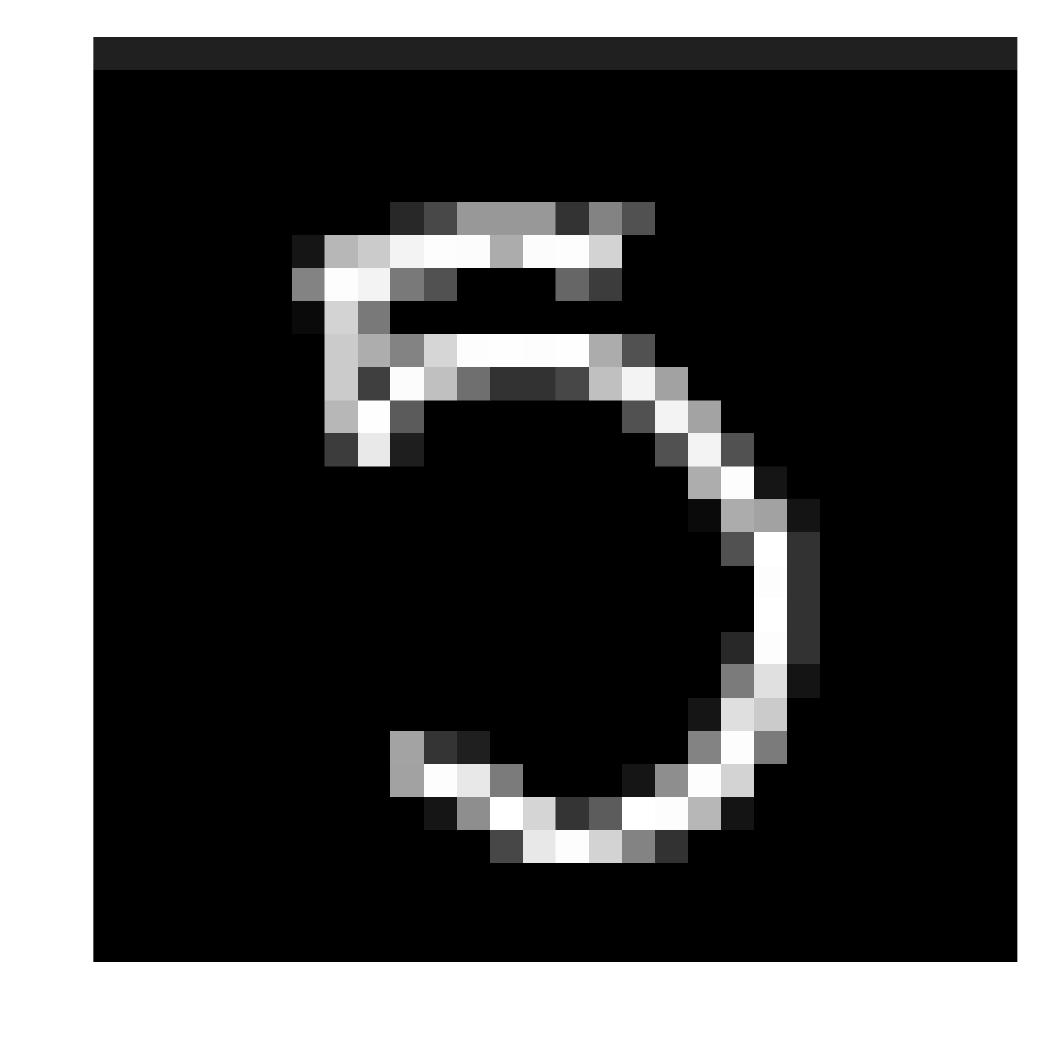}\!
        \captionsetup{font=scriptsize}
        \caption*{$3$ (1)}
    \end{subfigure}\!
    \begin{subfigure}[b]{0.2\linewidth}
        \includegraphics[width=\linewidth]{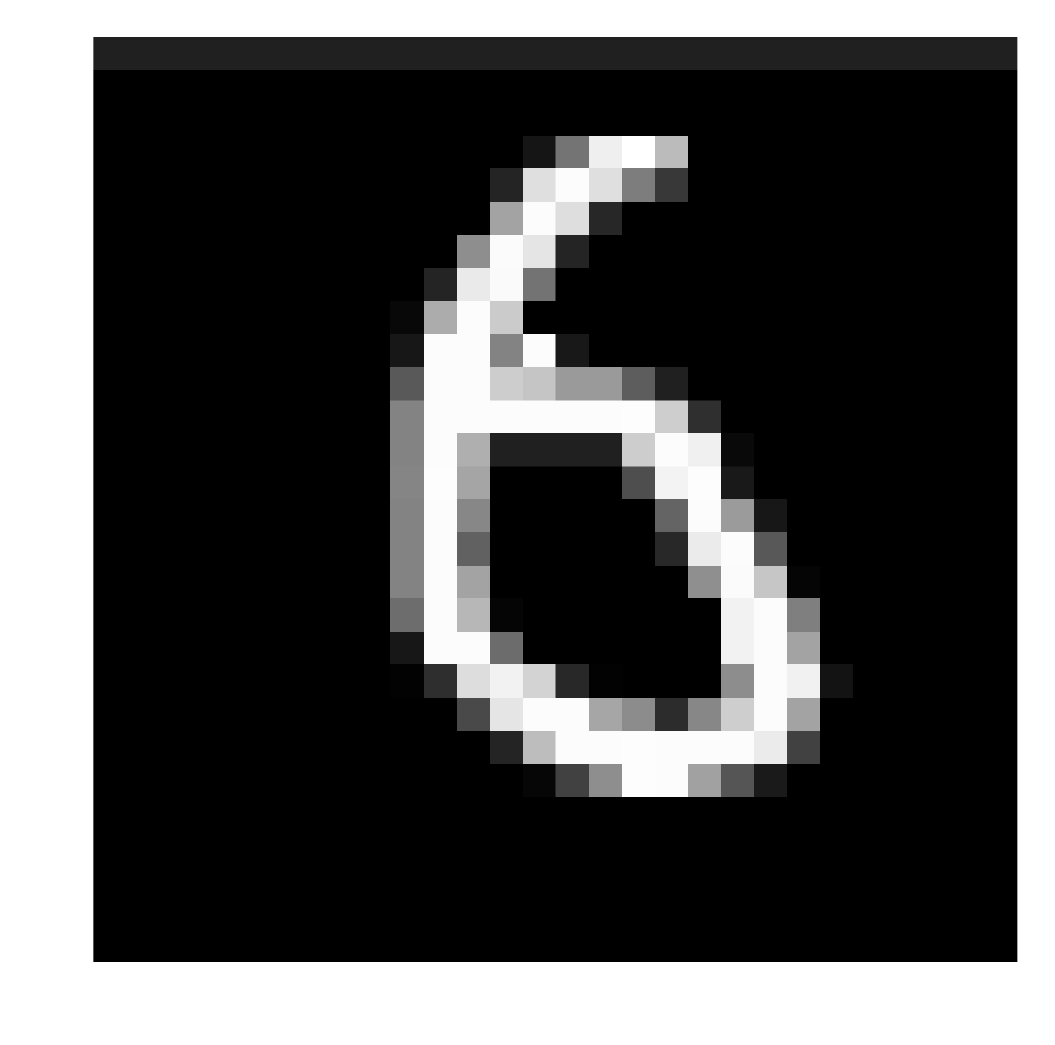}\!
        \captionsetup{font=scriptsize}
        \caption*{$0$ (2)}
    \end{subfigure}\!
    \begin{subfigure}[b]{0.2\linewidth}
        \includegraphics[width=\linewidth]{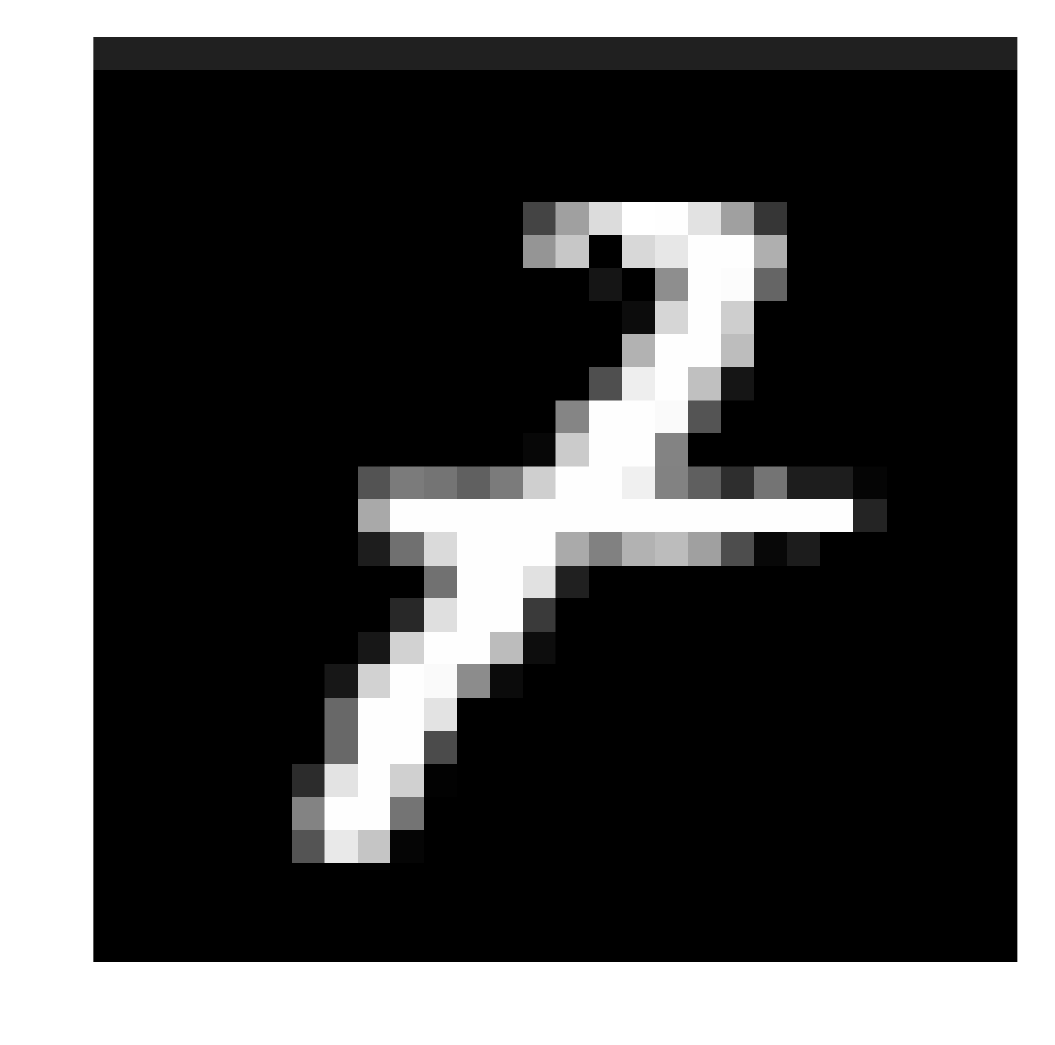}\!
        \captionsetup{font=scriptsize}
        \caption*{$2$ (2)}
    \end{subfigure}\!
    \begin{subfigure}[b]{0.2\linewidth}
        \includegraphics[width=\linewidth]{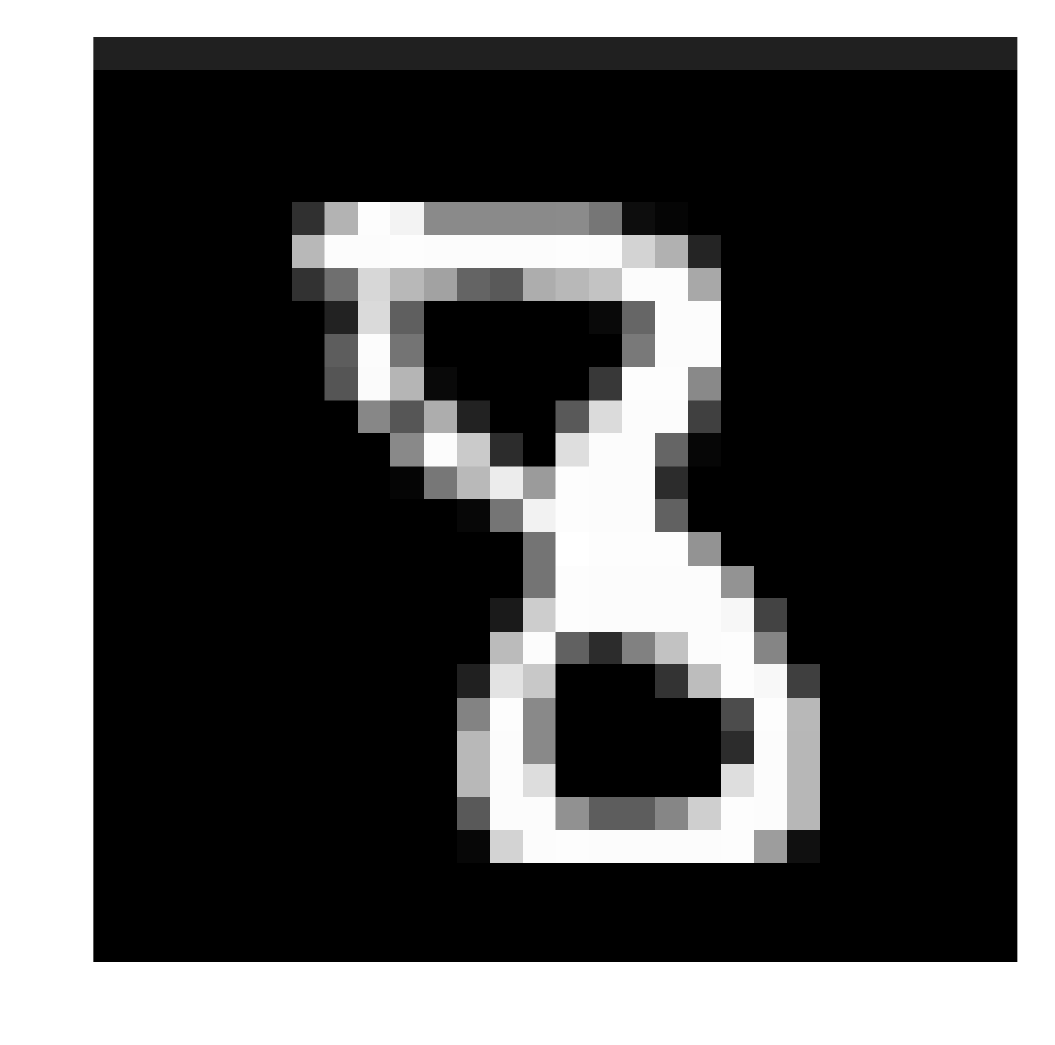}\!
        \captionsetup{font=scriptsize}
        \caption*{$3$ (1)}
    \end{subfigure}\!
    \begin{subfigure}[b]{0.2\linewidth}
        \includegraphics[width=\linewidth]{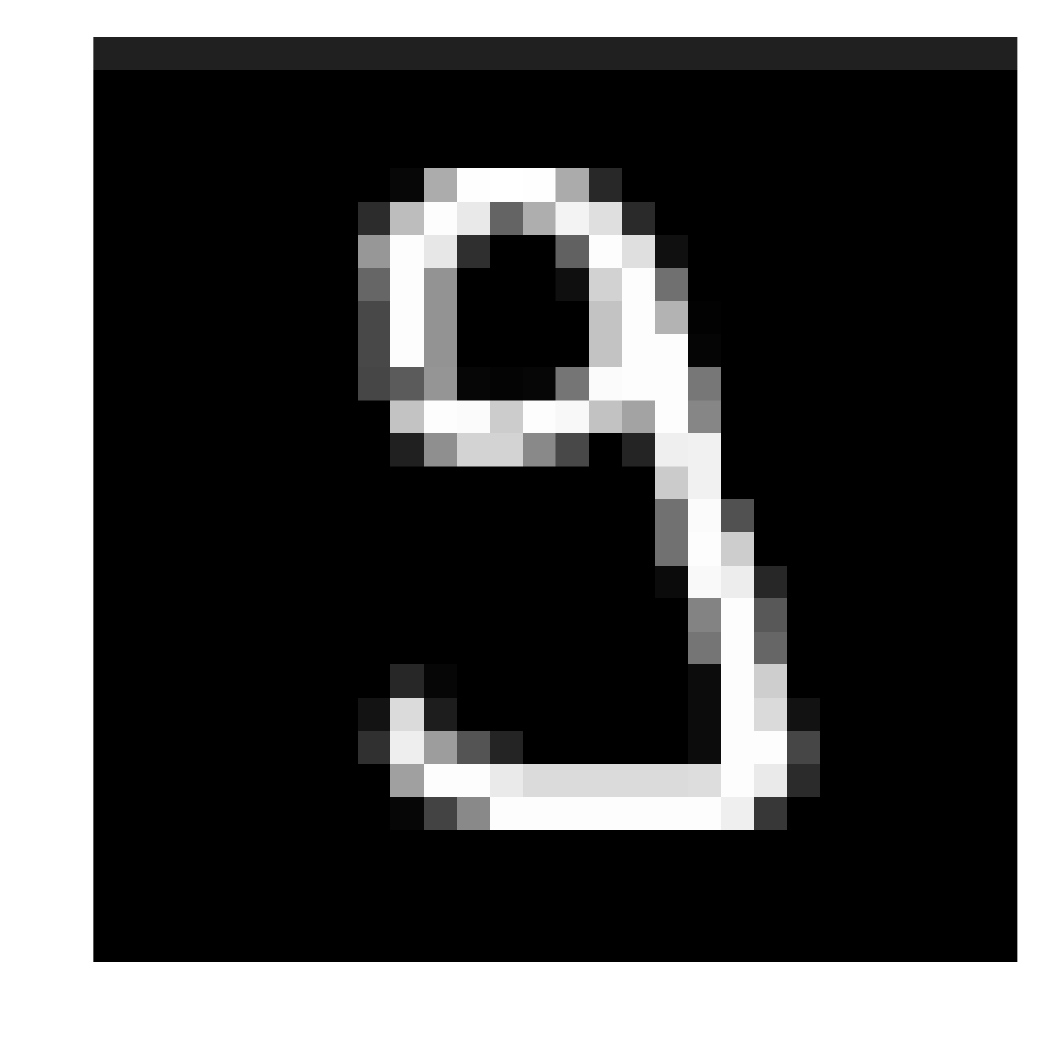}\!
        \captionsetup{font=scriptsize}
        \caption*{$3$ (1)}
    \end{subfigure}\!
    
    \begin{subfigure}[b]{0.2\linewidth}
        \includegraphics[width=\linewidth]{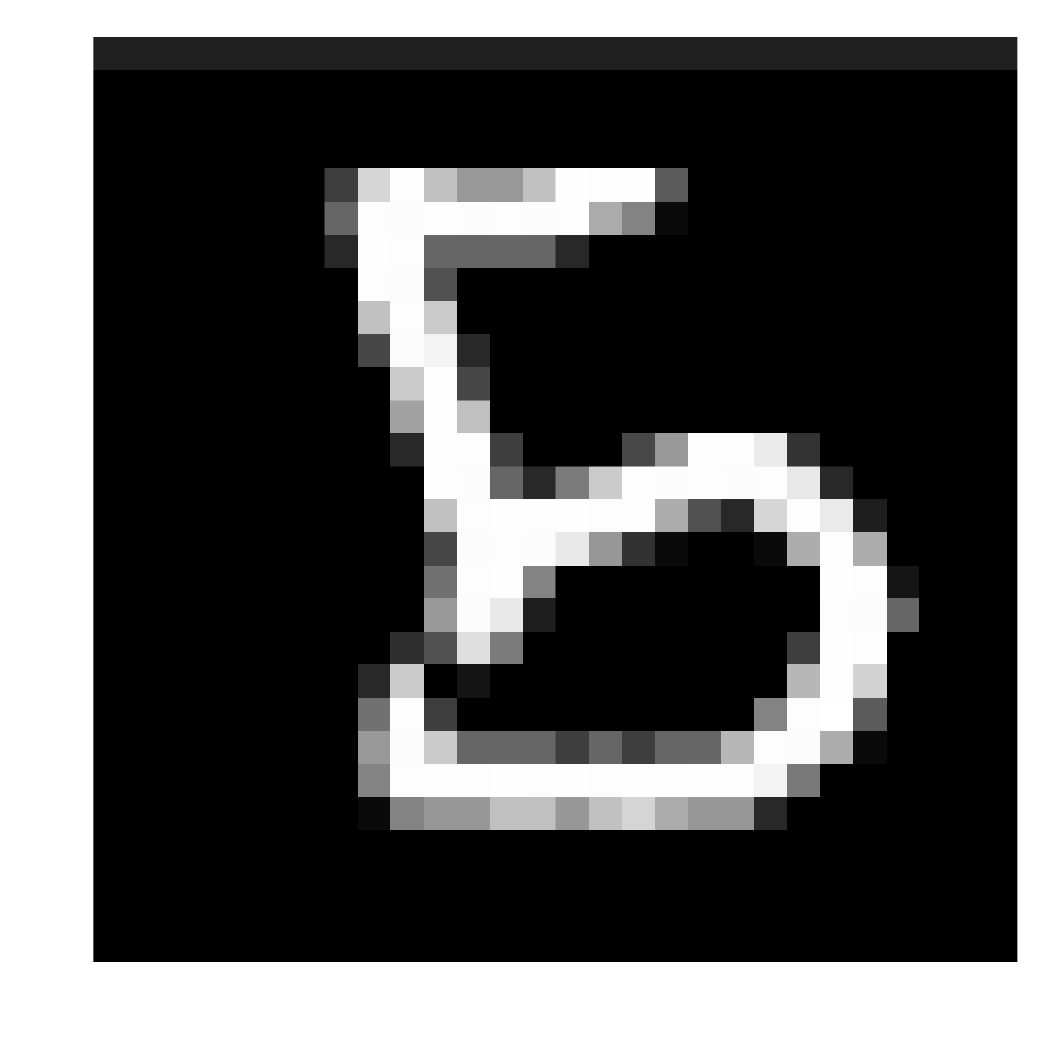}\!
        \captionsetup{font=scriptsize}
        \caption*{$8$ (1)}
    \end{subfigure}\!
    \begin{subfigure}[b]{0.2\linewidth}
        \includegraphics[width=\linewidth]{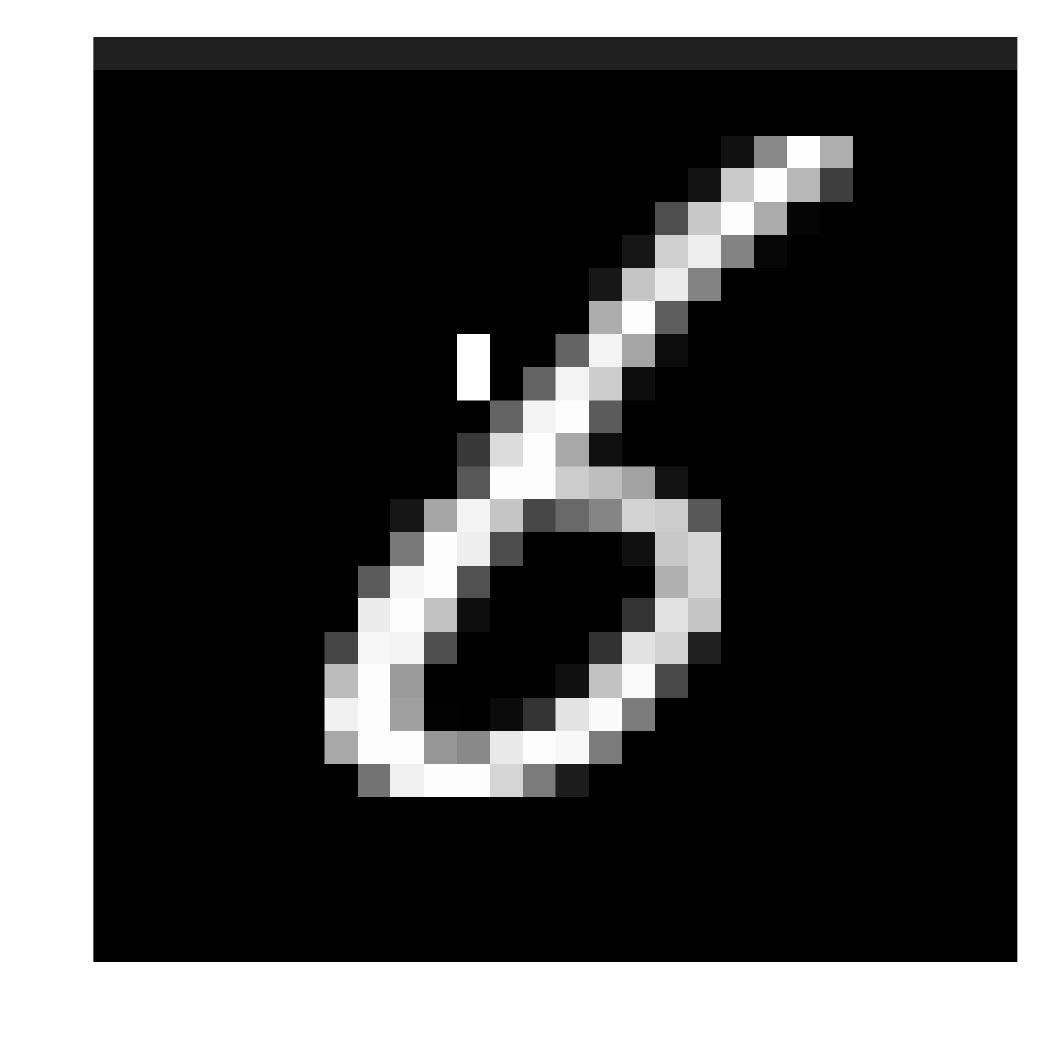}\!
        \captionsetup{font=scriptsize}
        \caption*{$8$ (3)}
    \end{subfigure}\!
    \begin{subfigure}[b]{0.2\linewidth}
        \includegraphics[width=\linewidth]{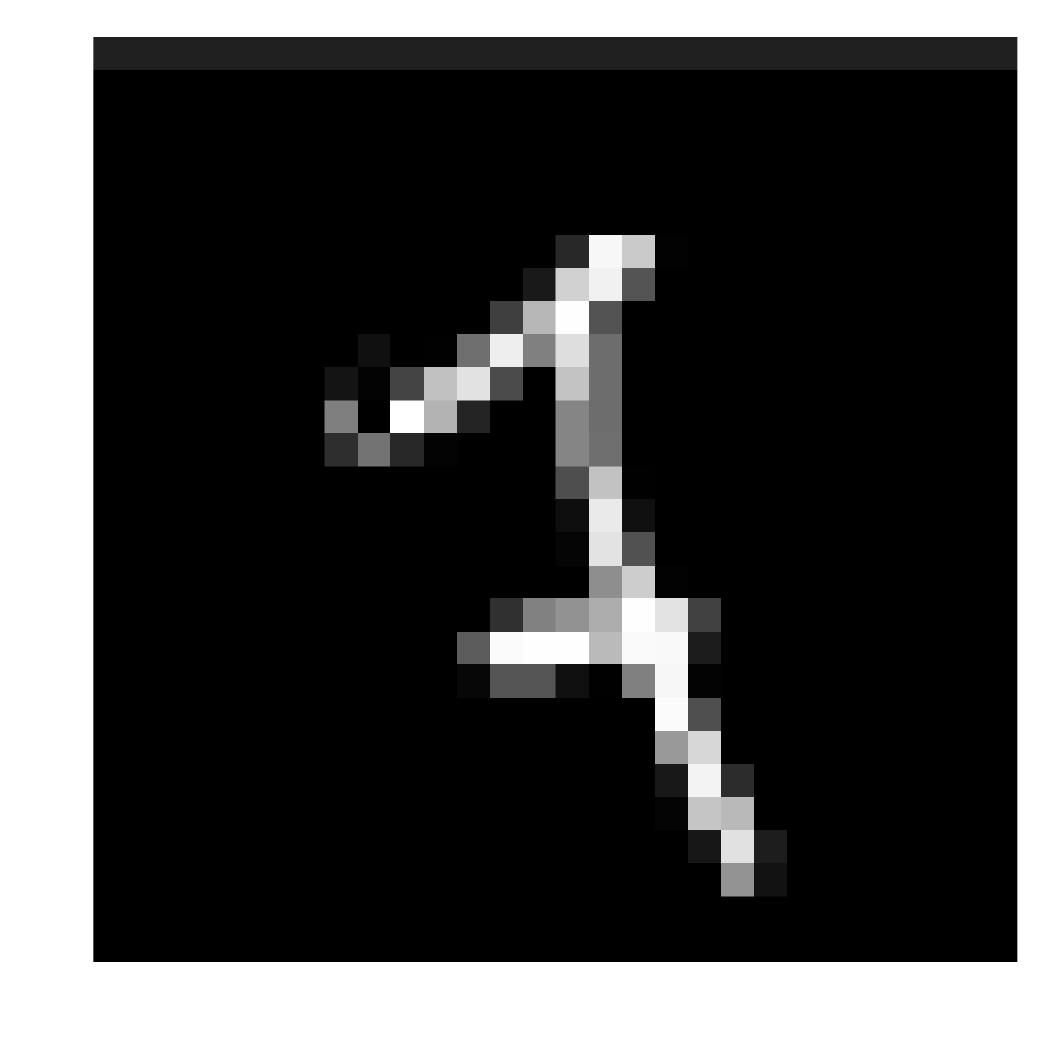}\!
        \captionsetup{font=scriptsize}
        \caption*{$1$ (3)}
    \end{subfigure}\!
    \begin{subfigure}[b]{0.2\linewidth}
        \includegraphics[width=\linewidth]{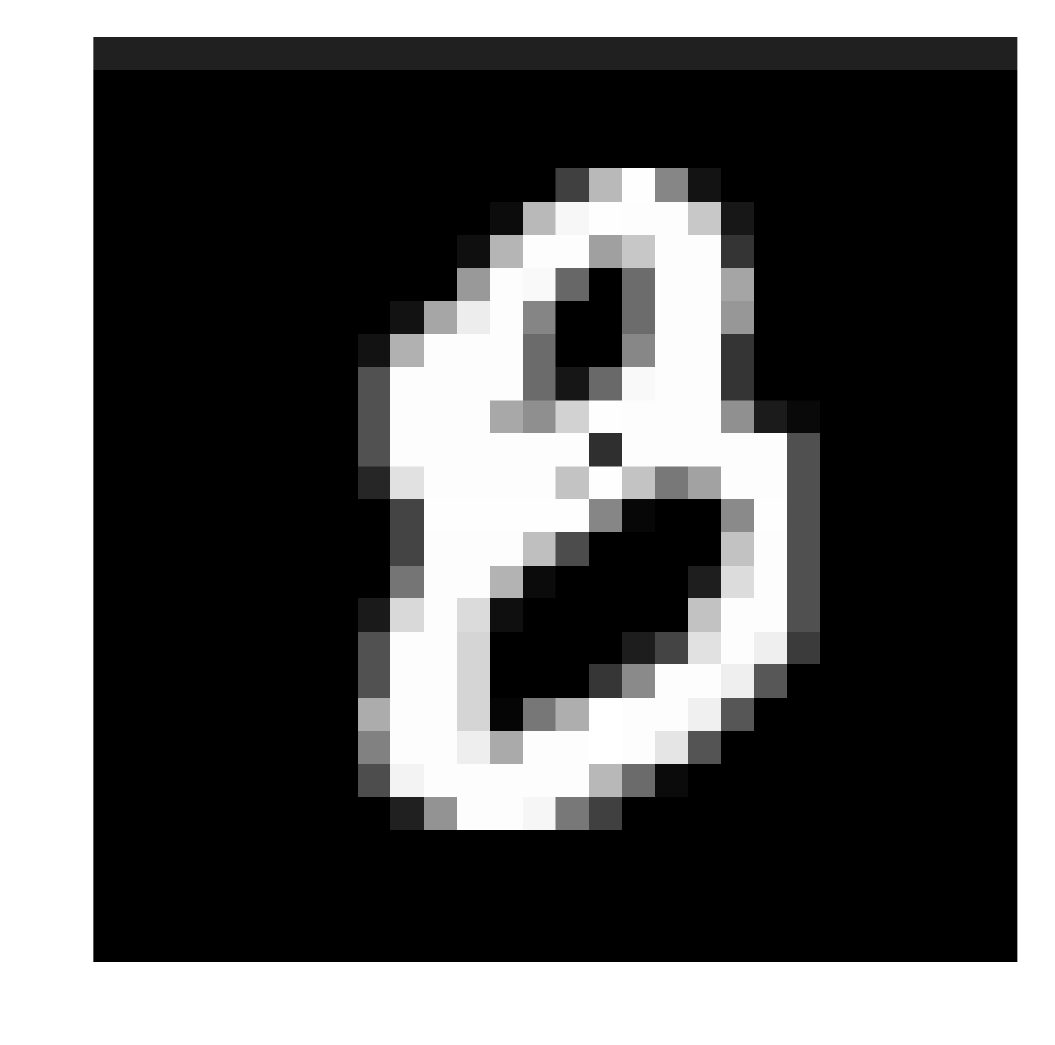}\!
        \captionsetup{font=scriptsize}
        \caption*{$0$ (2)}
    \end{subfigure}\!
    \begin{subfigure}[b]{0.2\linewidth}
        \includegraphics[width=\linewidth]{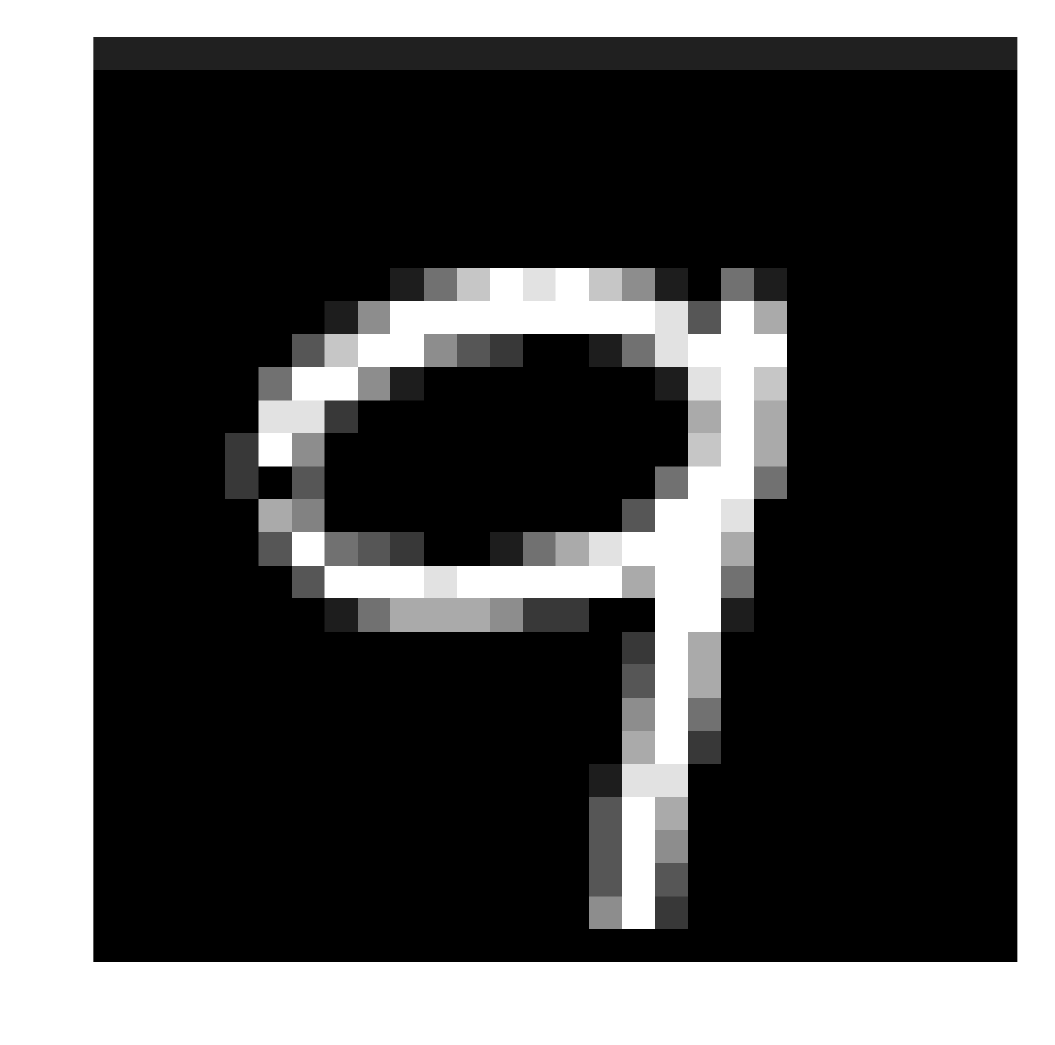}\!
        \captionsetup{font=scriptsize}
        \caption*{$7$ (2)}
    \end{subfigure}\!
\captionsetup{font=small, skip=8pt}
\caption{Highly sparse perturbations.}
\label{fig:mnist_high}
\end{subfigure}\hfill
\begin{subfigure}[b]{0.3\linewidth}
\captionsetup[subfigure]{skip=1pt}
    \begin{subfigure}[b]{0.2\linewidth}
        \includegraphics[width=\linewidth]{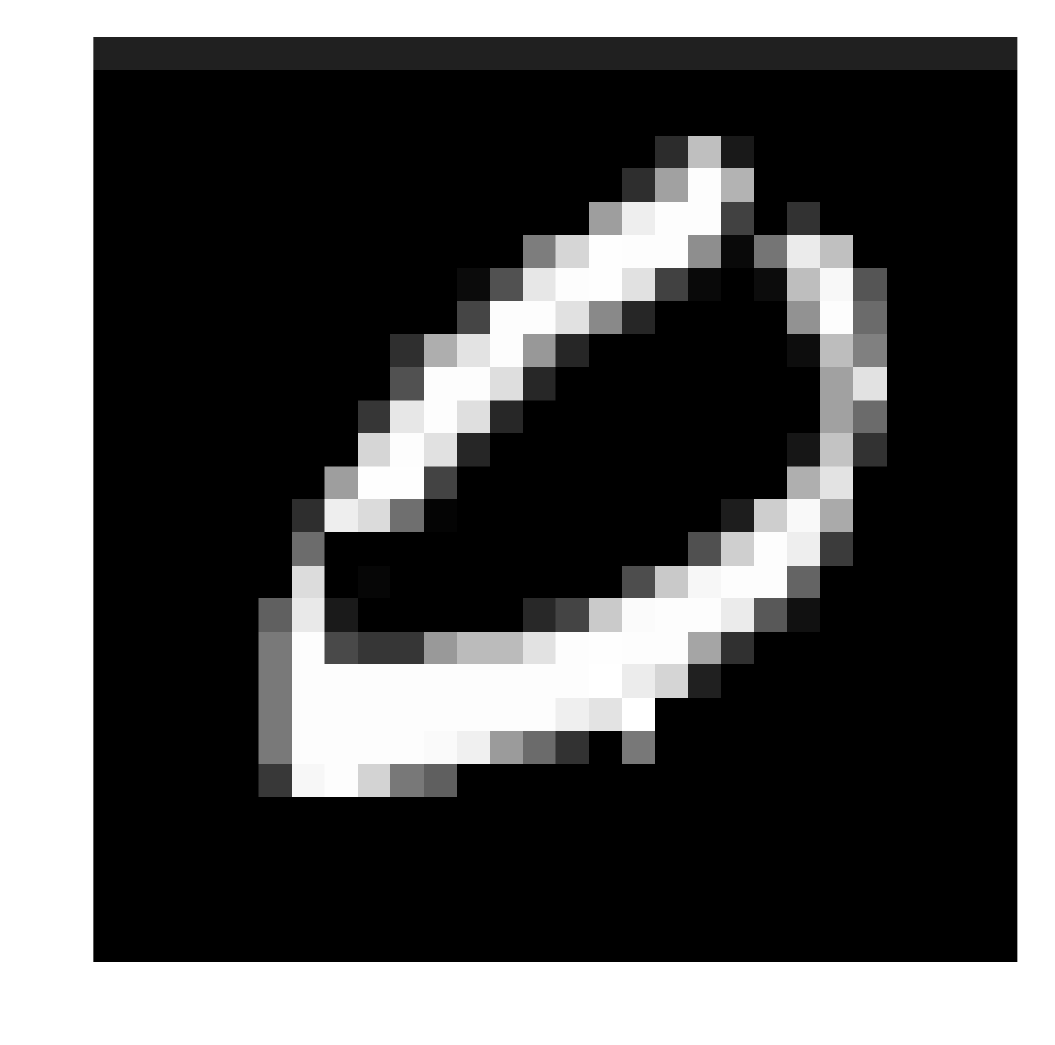}\!
        \captionsetup{font=scriptsize}
        \caption*{$2$ (6)}
    \end{subfigure}\!
    \begin{subfigure}[b]{0.2\linewidth}
        \includegraphics[width=\linewidth]{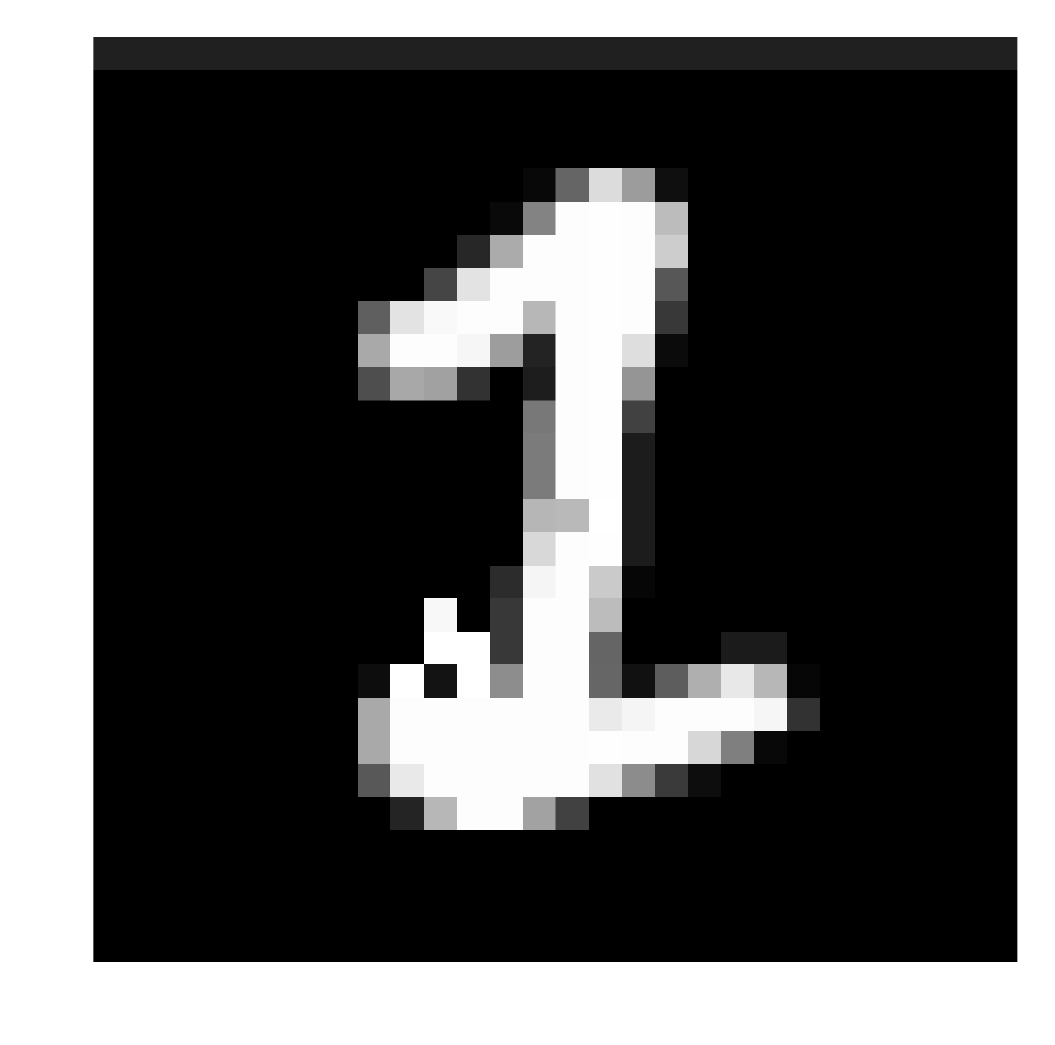}\!
        \captionsetup{font=scriptsize}
        \caption*{$2$ (6)}
    \end{subfigure}\!
    \begin{subfigure}[b]{0.2\linewidth}
        \includegraphics[width=\linewidth]{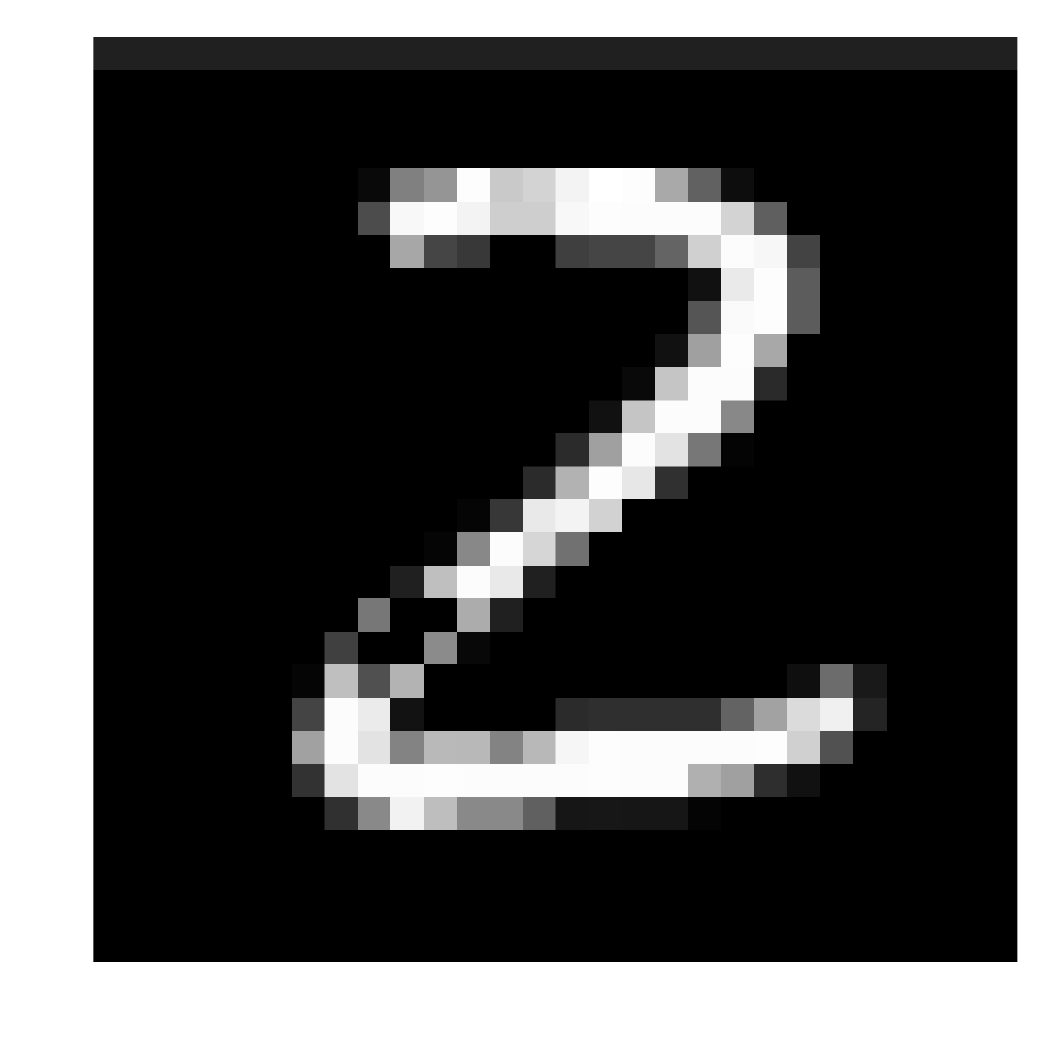}\!
        \captionsetup{font=scriptsize}
        \caption*{$3$ (6)}
    \end{subfigure}\!
    \begin{subfigure}[b]{0.2\linewidth}
        \includegraphics[width=\linewidth]{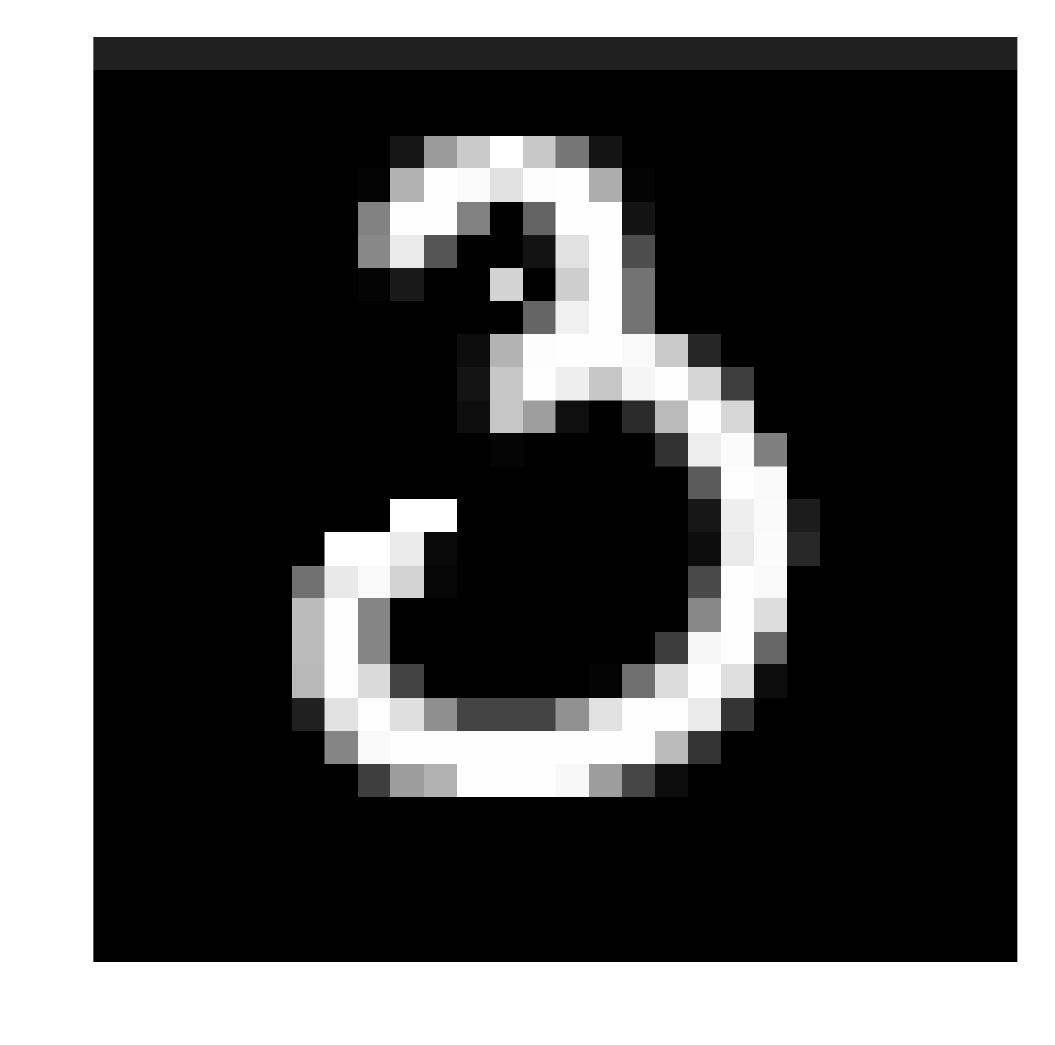}\!
        \captionsetup{font=scriptsize}
        \caption*{$0$ (6)}
    \end{subfigure}\!
    \begin{subfigure}[b]{0.2\linewidth}
        \includegraphics[width=\linewidth]{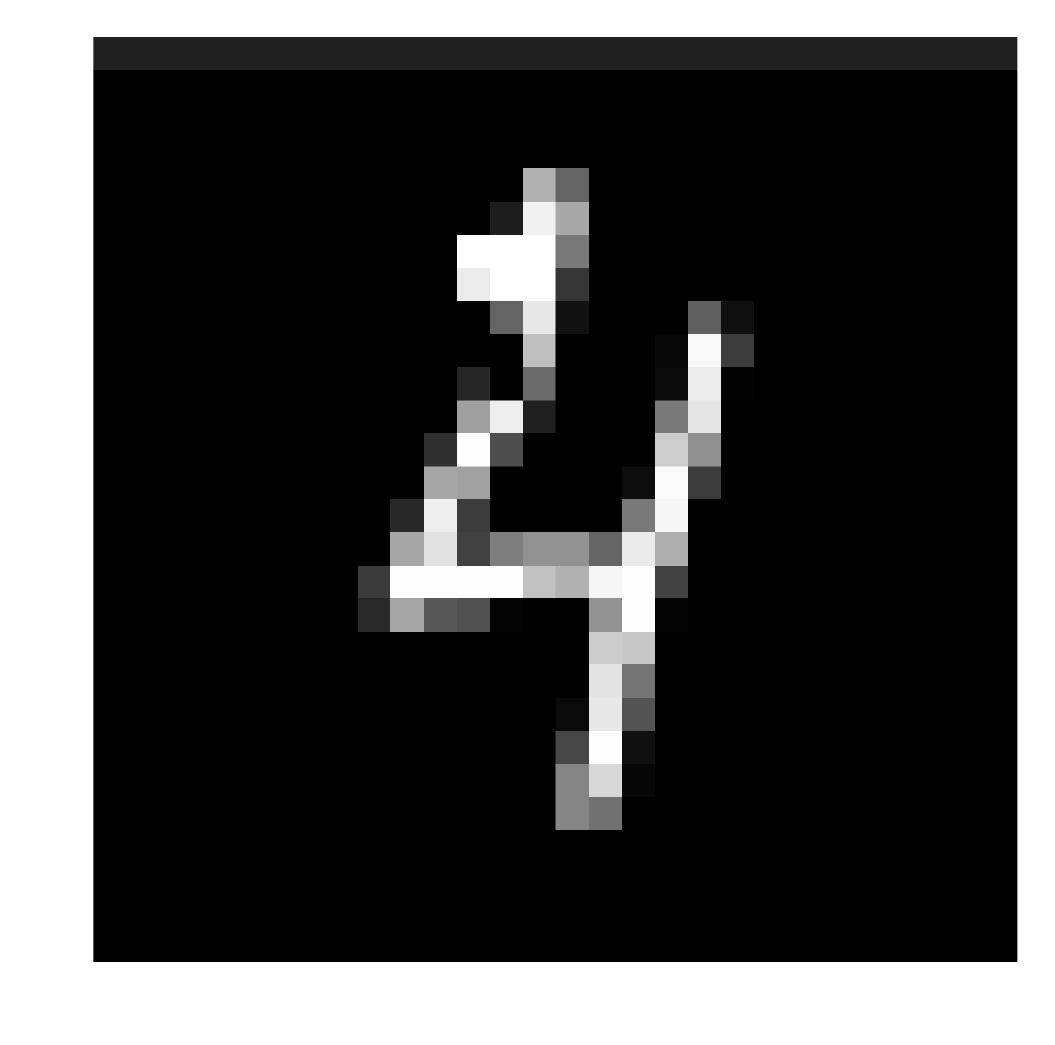}\!
        \captionsetup{font=scriptsize}
        \caption*{$7$ (6)}
    \end{subfigure}\!
    
    \begin{subfigure}[b]{0.2\linewidth}
        \includegraphics[width=\linewidth]{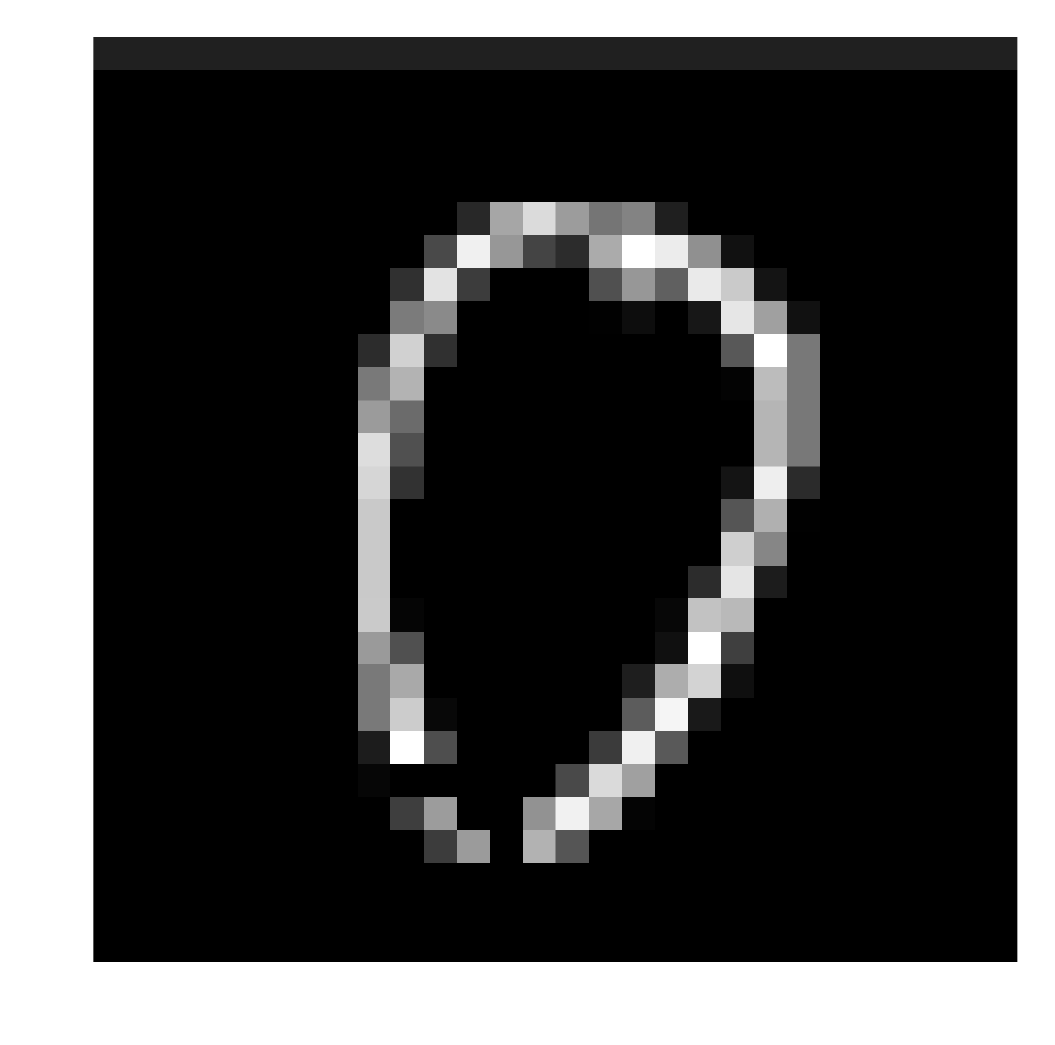}\!
        \captionsetup{font=scriptsize}
        \caption*{$9$ (6)}
    \end{subfigure}\!
    \begin{subfigure}[b]{0.2\linewidth}
        \includegraphics[width=\linewidth]{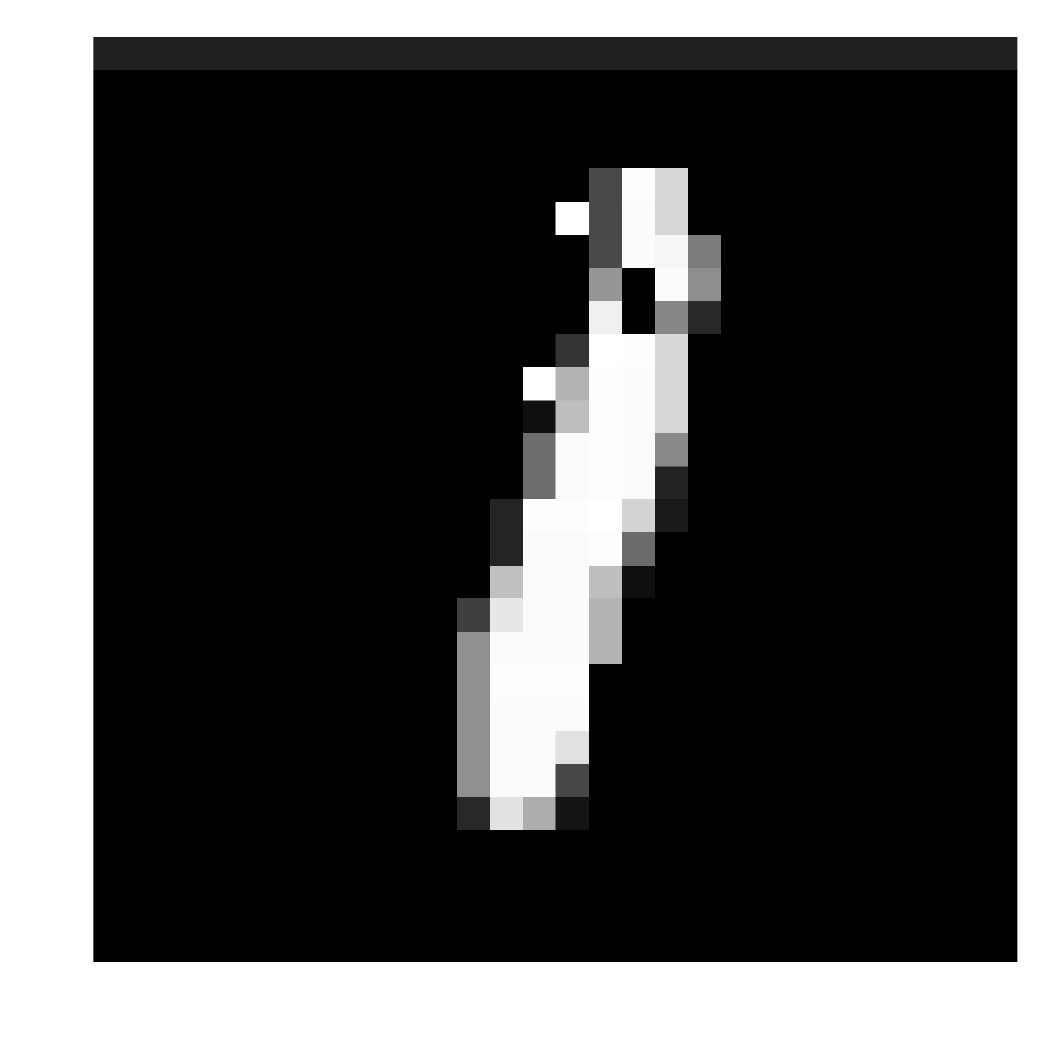}\!
        \captionsetup{font=scriptsize}
        \caption*{$8$ (6)}
    \end{subfigure}\!
    \begin{subfigure}[b]{0.2\linewidth}
        \includegraphics[width=\linewidth]{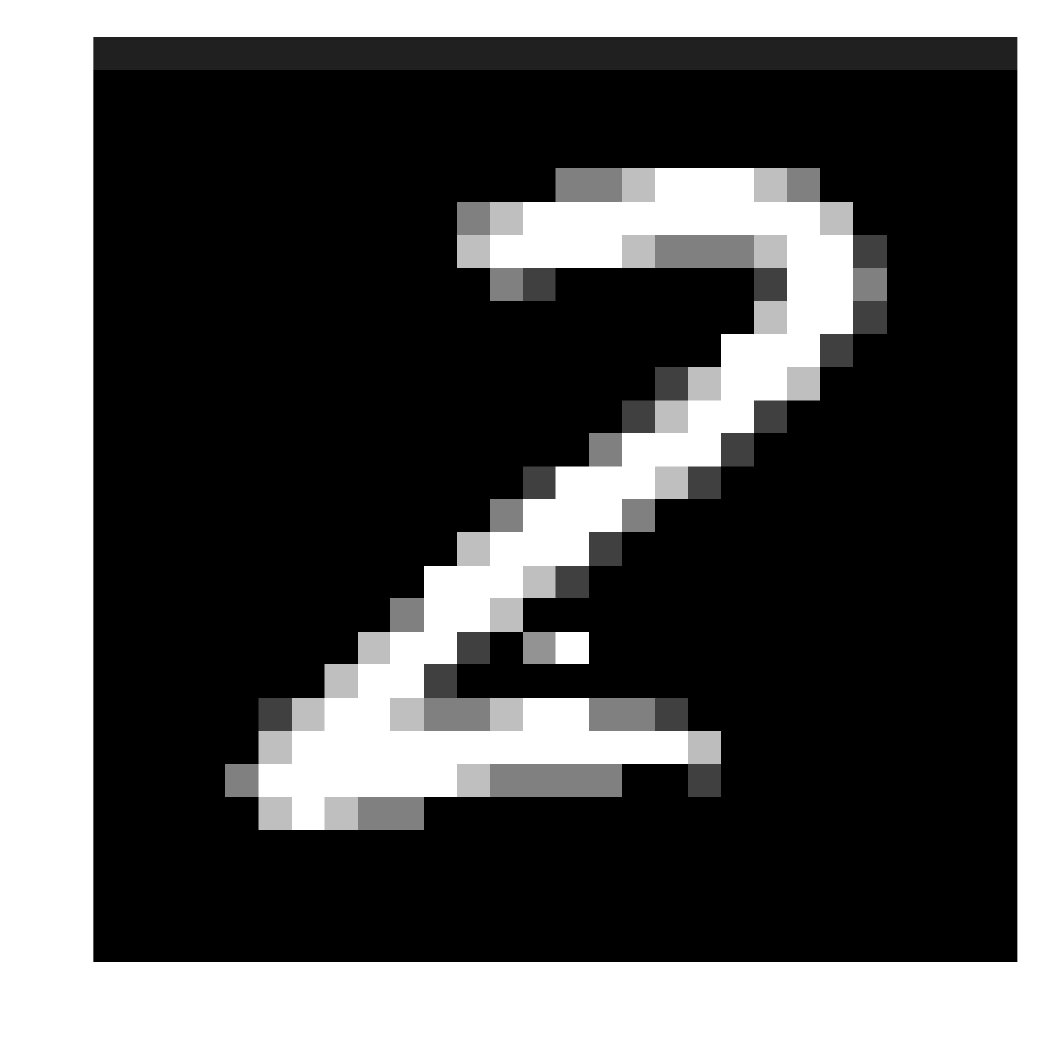}\!
        \captionsetup{font=scriptsize}
        \caption*{$8$ (6)}
    \end{subfigure}\!
    \begin{subfigure}[b]{0.2\linewidth}
        \includegraphics[width=\linewidth]{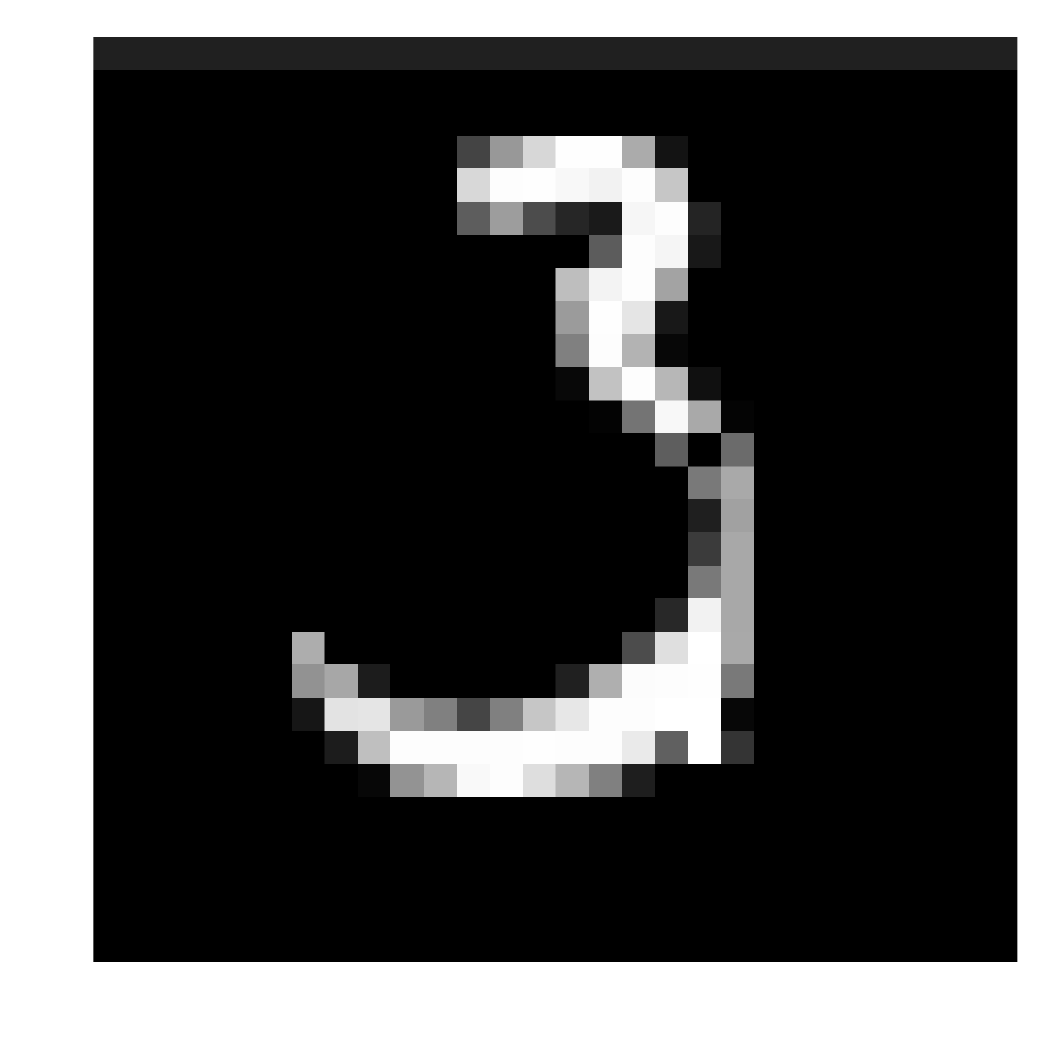}\!
        \captionsetup{font=scriptsize}
        \caption*{$1$ (6)}
    \end{subfigure}\!
    \begin{subfigure}[b]{0.2\linewidth}
        \includegraphics[width=\linewidth]{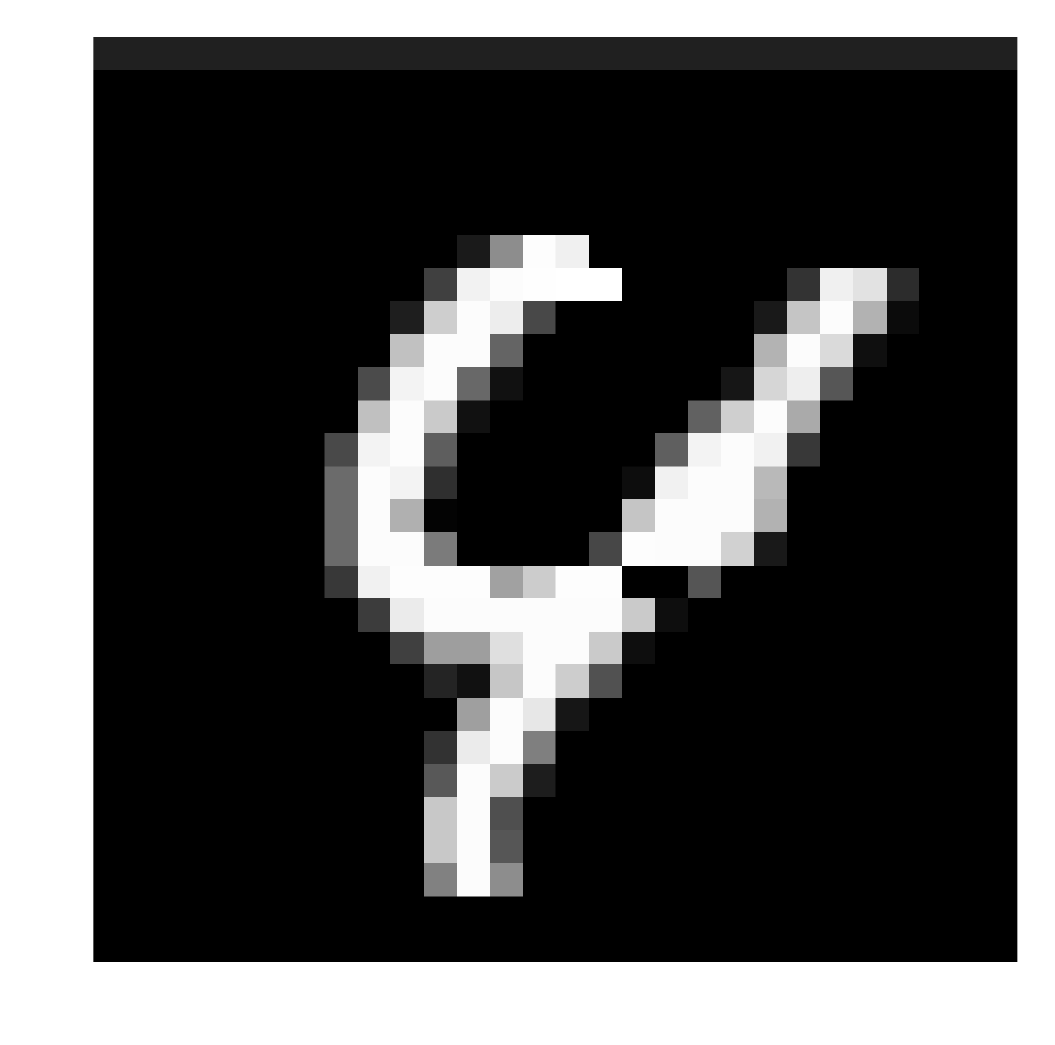}\!
        \captionsetup{font=scriptsize}
        \caption*{$8$ (6)}
    \end{subfigure}\!
    
    \begin{subfigure}[b]{0.2\linewidth}
        \includegraphics[width=\linewidth]{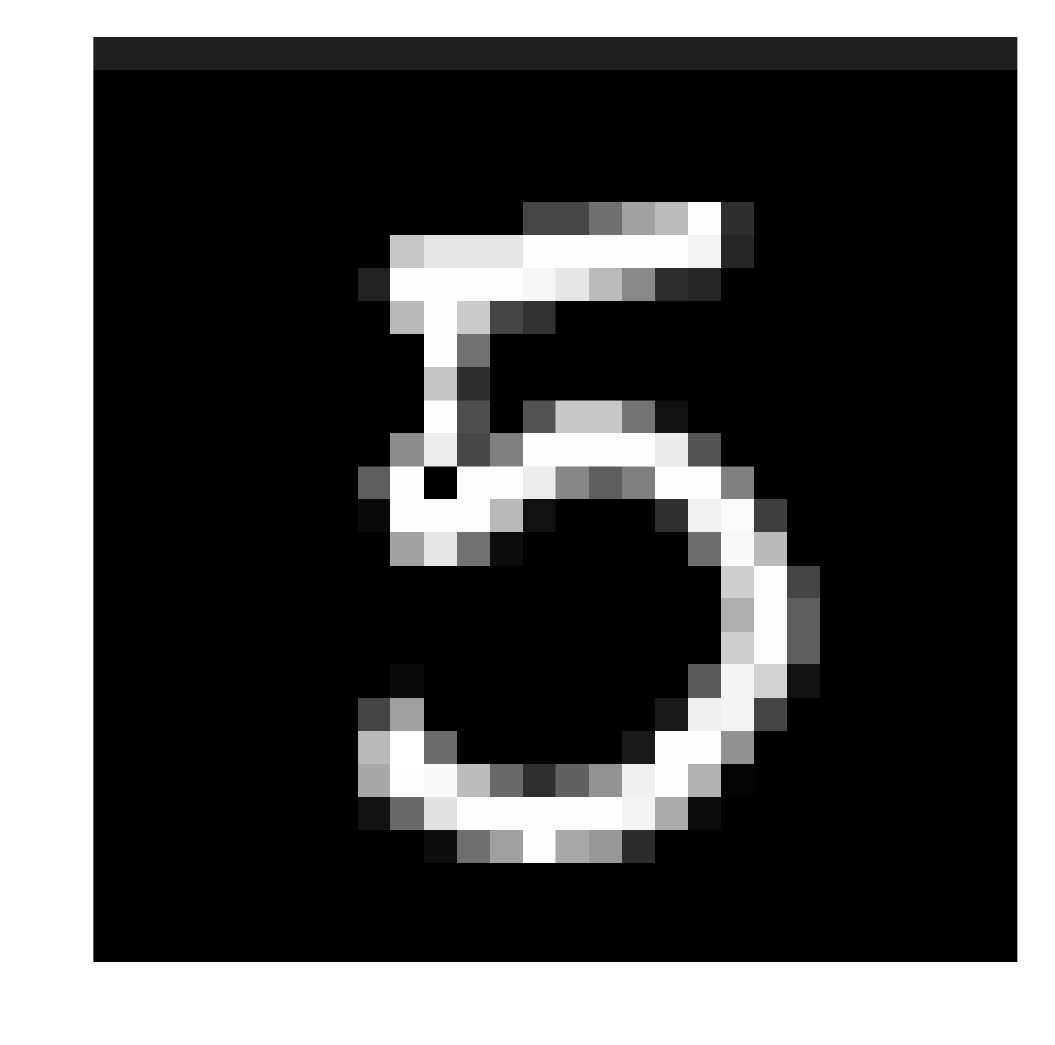}\!
        \captionsetup{font=scriptsize}
        \caption*{$3$ (6)}
    \end{subfigure}\!
    \begin{subfigure}[b]{0.2\linewidth}
        \includegraphics[width=\linewidth]{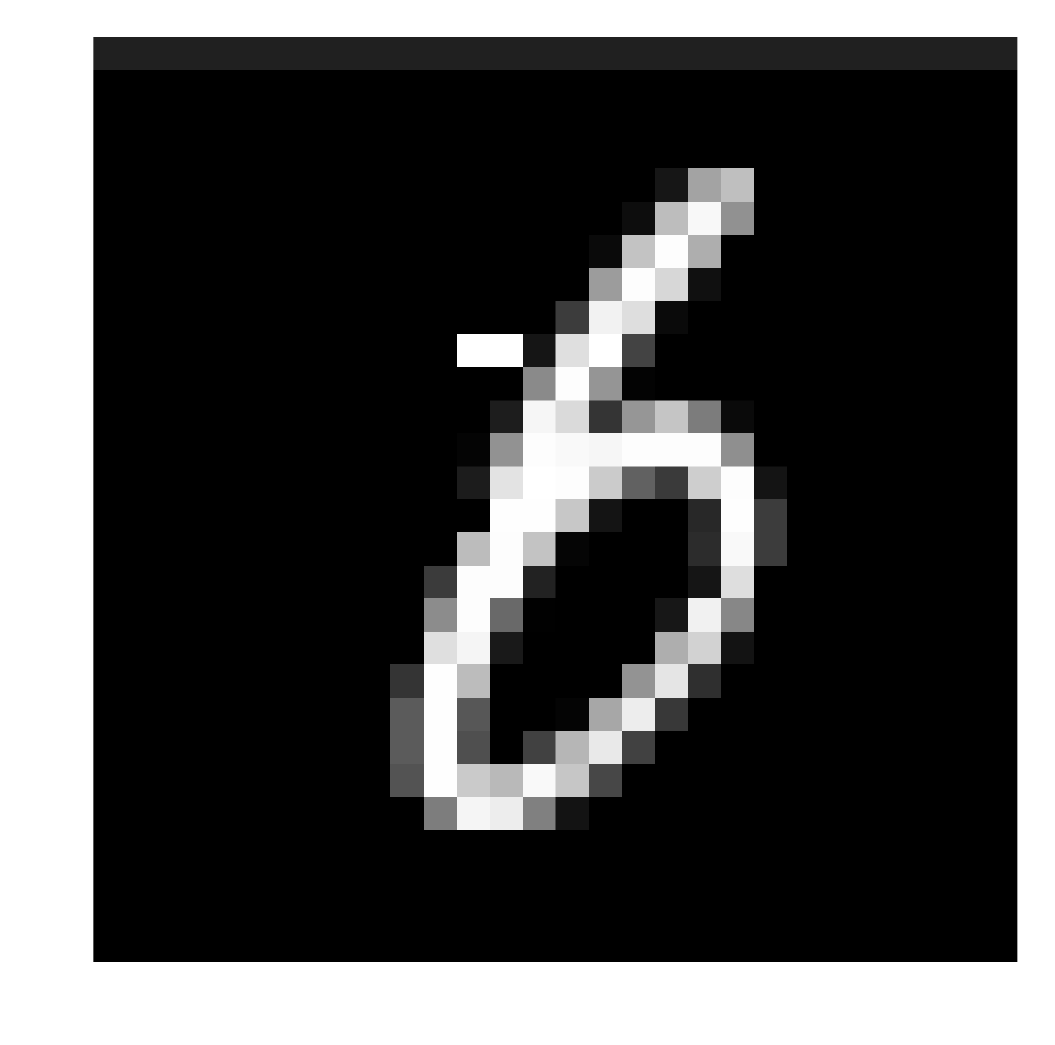}\!
        \captionsetup{font=scriptsize}
        \caption*{$8$ (6)}
    \end{subfigure}\!
    \begin{subfigure}[b]{0.2\linewidth}
        \includegraphics[width=\linewidth]{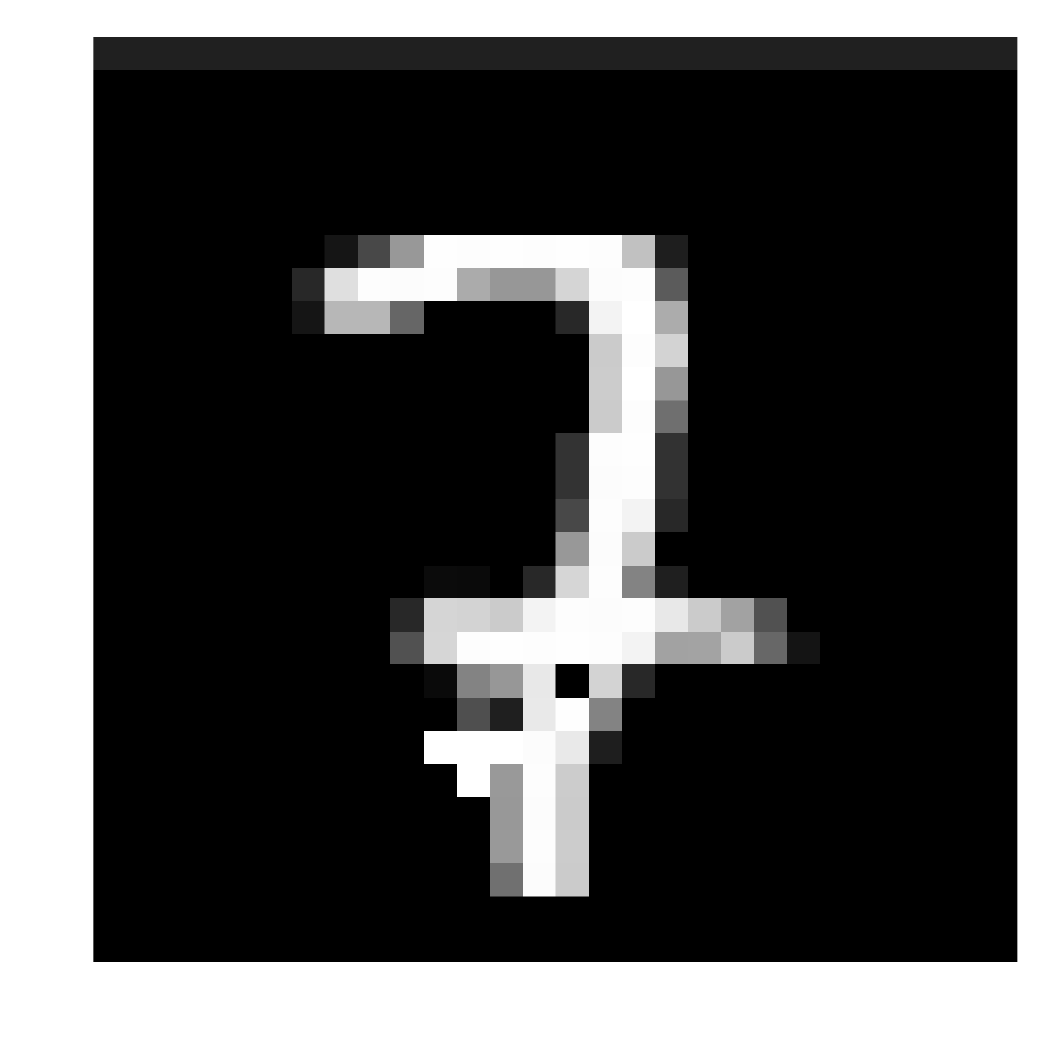}\!
        \captionsetup{font=scriptsize}
        \caption*{$2$ (6)}
    \end{subfigure}\!
    \begin{subfigure}[b]{0.2\linewidth}
        \includegraphics[width=\linewidth]{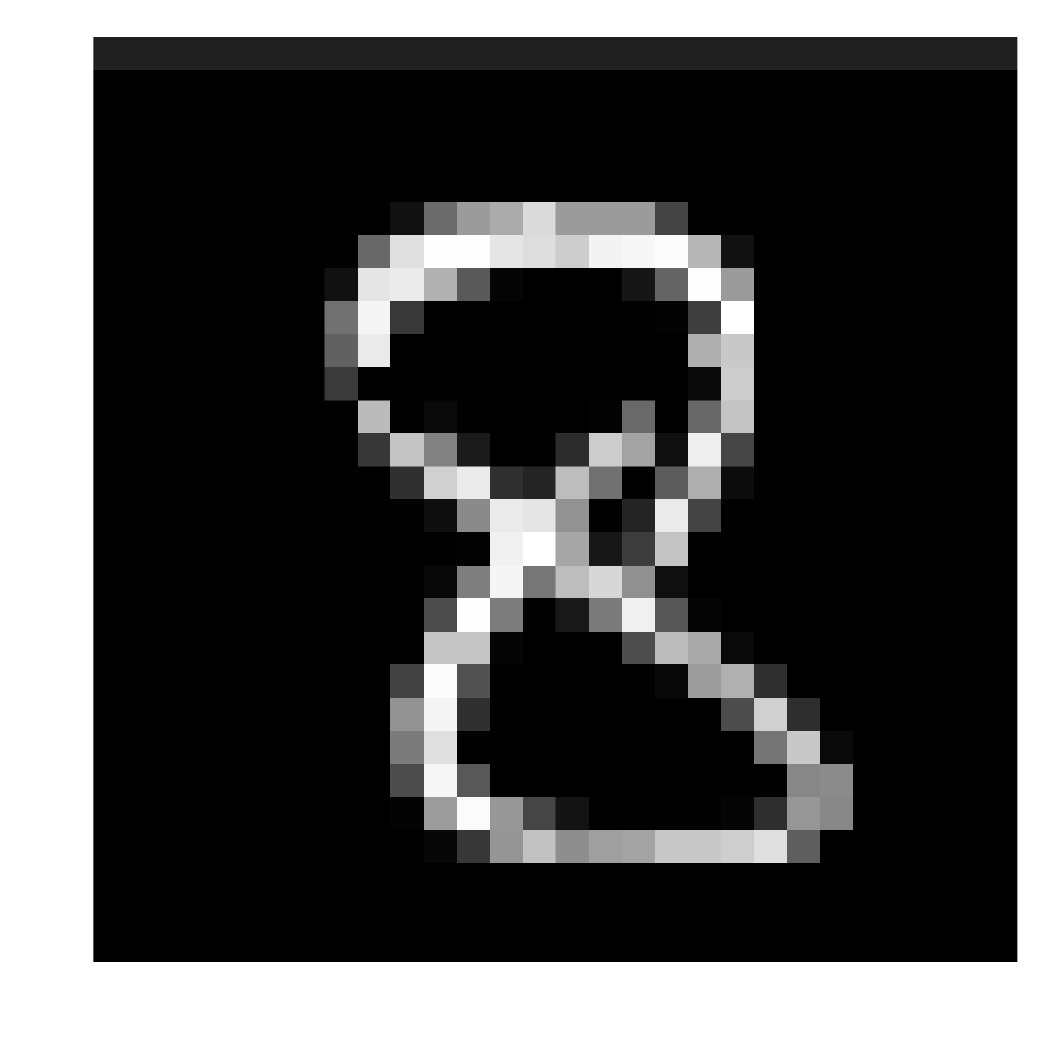}\!
        \captionsetup{font=scriptsize}
        \caption*{$2$ (6)}
    \end{subfigure}\!
    \begin{subfigure}[b]{0.2\linewidth}
        \includegraphics[width=\linewidth]{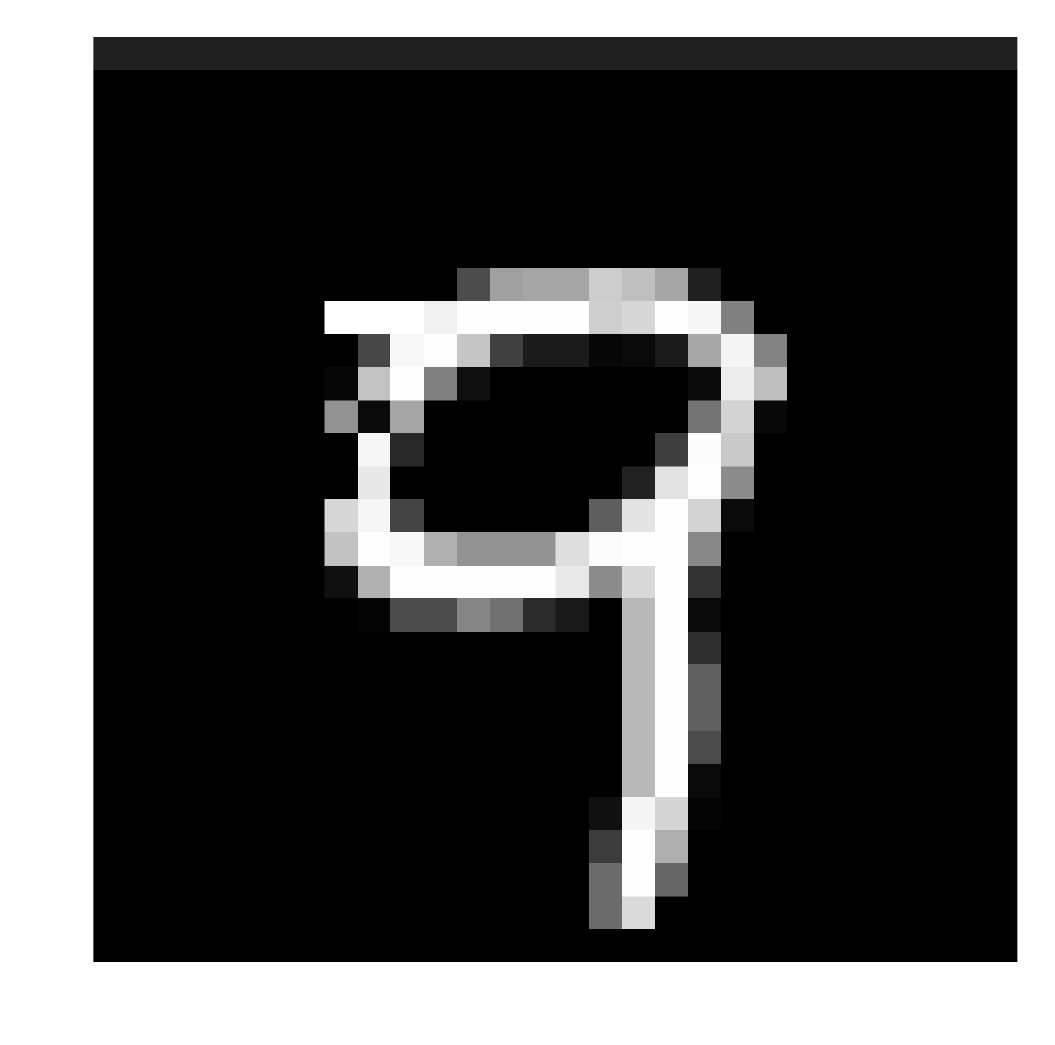}\!
        \captionsetup{font=scriptsize}
        \caption*{$7$ (6)}
    \end{subfigure}\!
    
    \begin{subfigure}[b]{0.2\linewidth}
        \includegraphics[width=\linewidth]{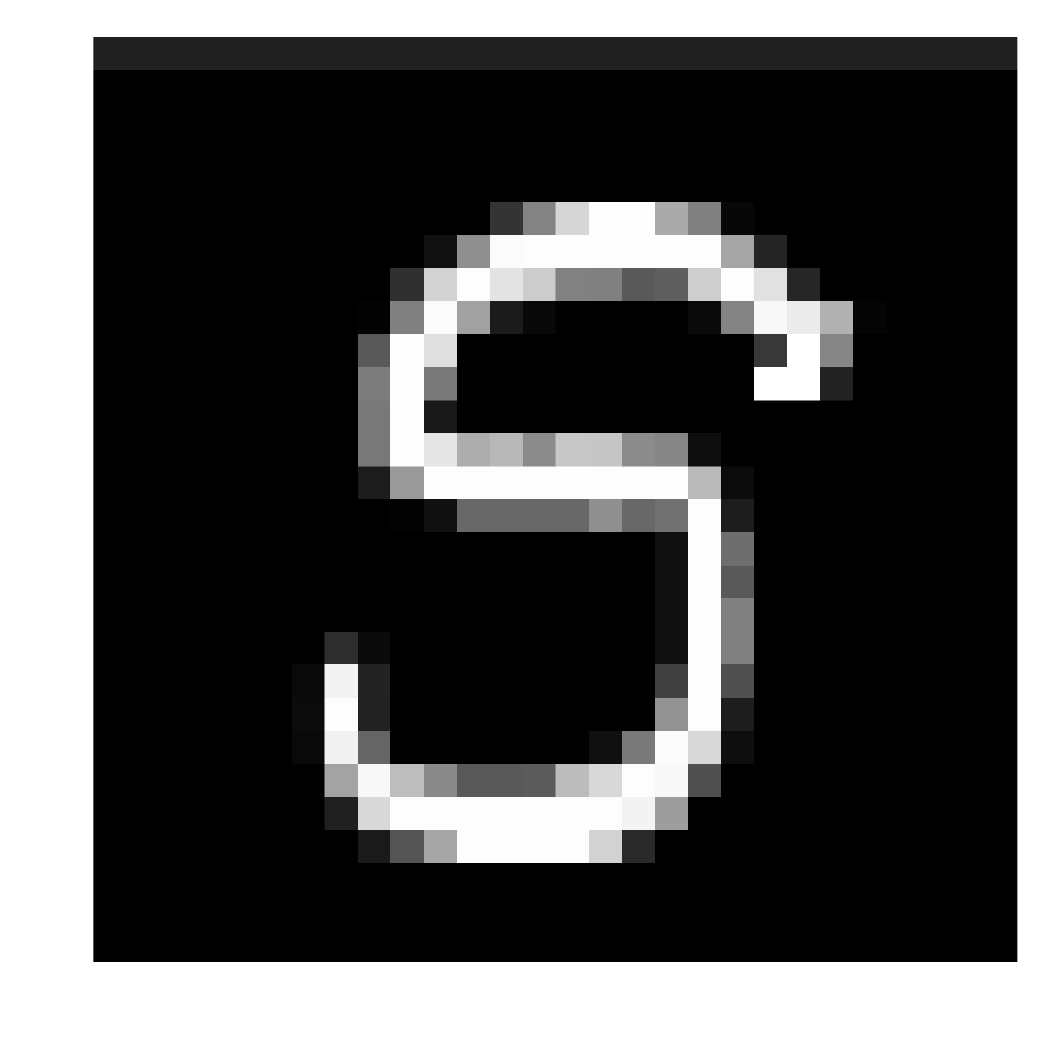}\!
        \captionsetup{font=scriptsize}
        \caption*{$9$ (6)}
    \end{subfigure}\!
    \begin{subfigure}[b]{0.2\linewidth}
        \includegraphics[width=\linewidth]{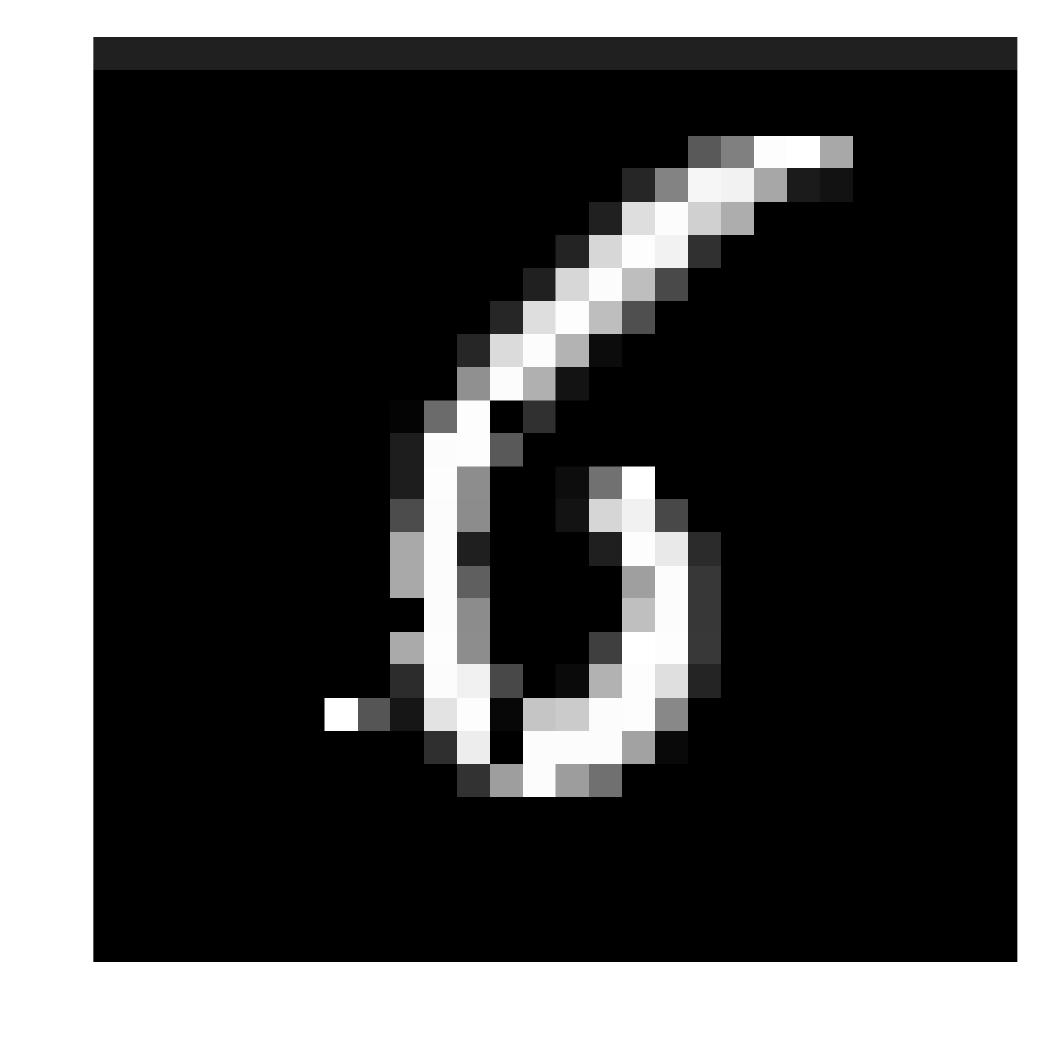}\!
        \captionsetup{font=scriptsize}
        \caption*{$5$ (7)}
    \end{subfigure}\!
    \begin{subfigure}[b]{0.2\linewidth}
        \includegraphics[width=\linewidth]{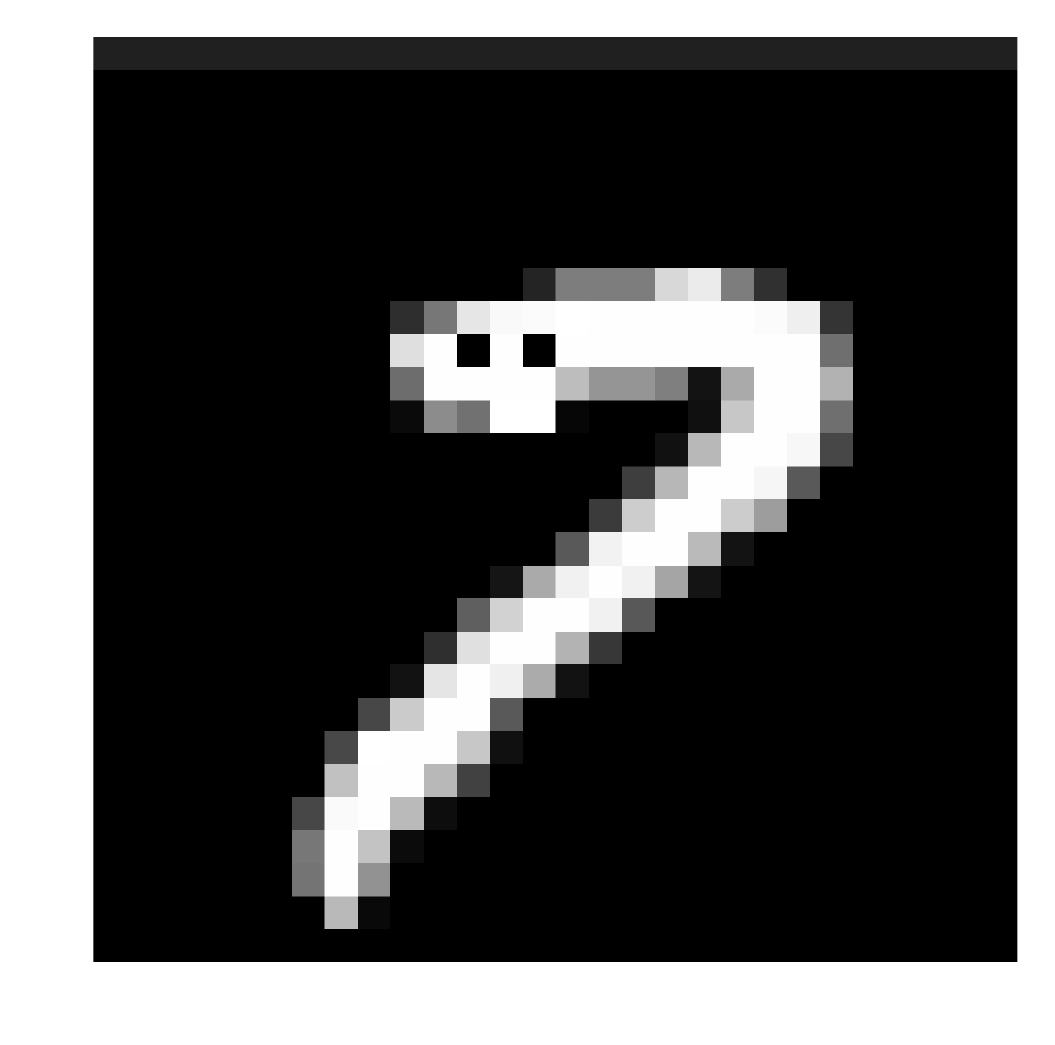}\!
        \captionsetup{font=scriptsize}
        \caption*{$9$ (6)}
    \end{subfigure}\!
    \begin{subfigure}[b]{0.2\linewidth}
        \includegraphics[width=\linewidth]{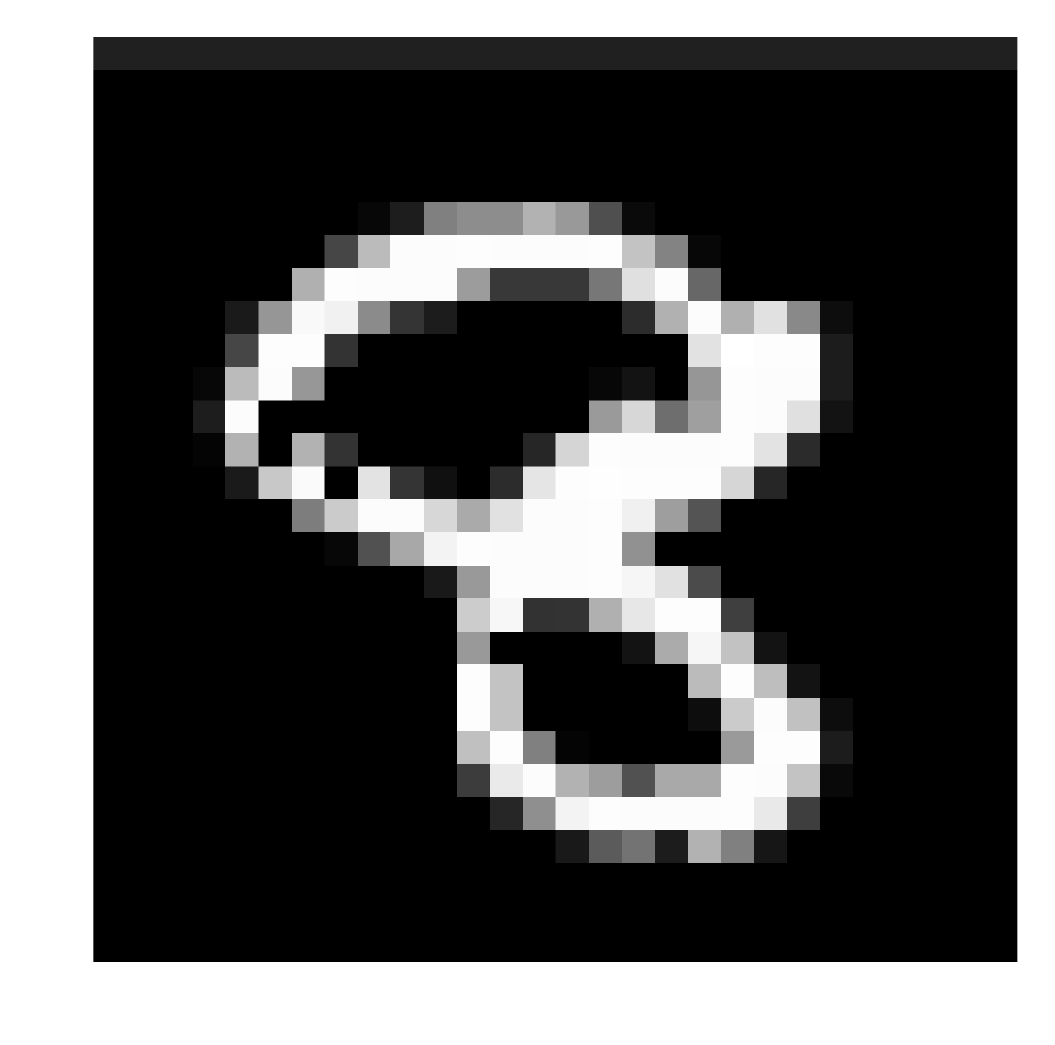}\!
        \captionsetup{font=scriptsize}
        \caption*{$3$ (6)}
    \end{subfigure}\!
    \begin{subfigure}[b]{0.2\linewidth}
        \includegraphics[width=\linewidth]{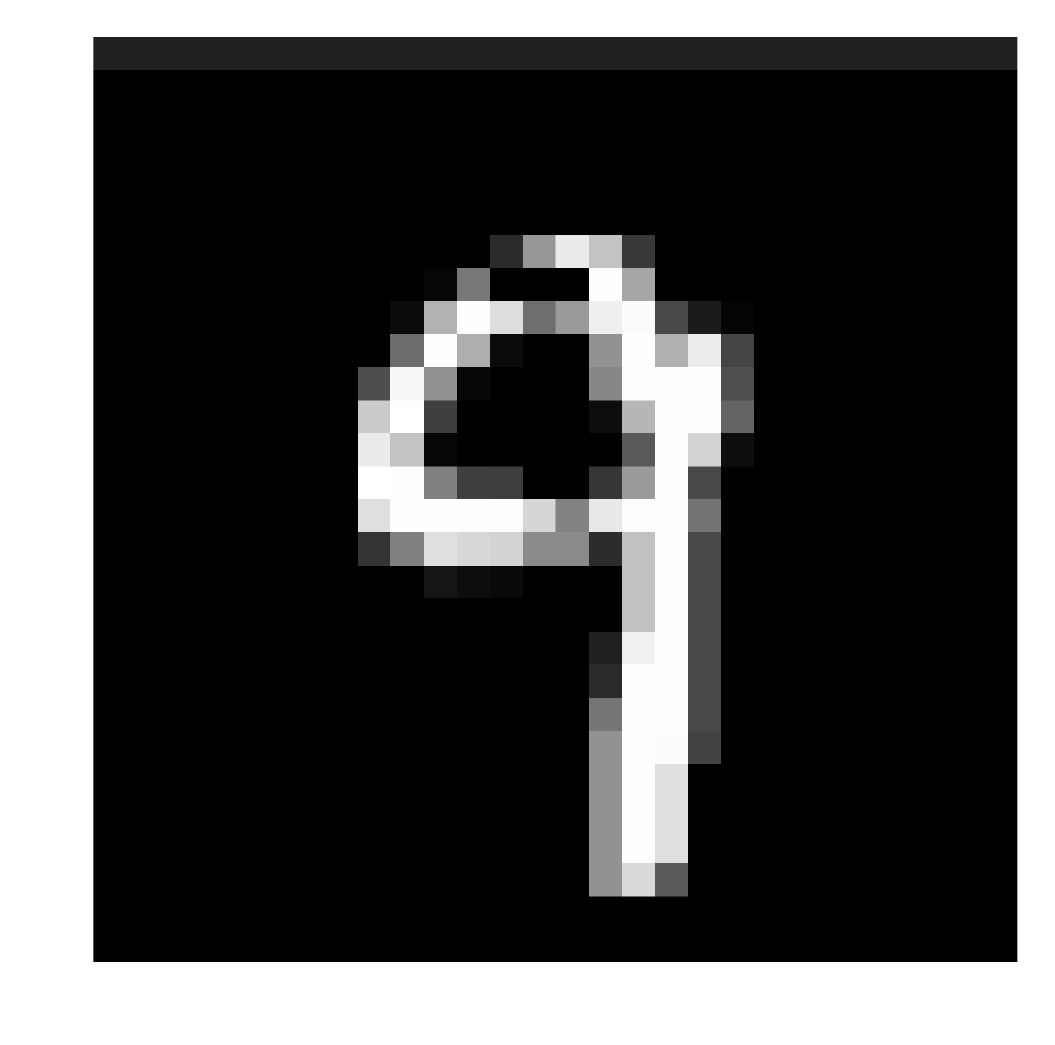}\!
        \captionsetup{font=scriptsize}
        \caption*{$4$ (6)}
    \end{subfigure}\!
\captionsetup{font=small, skip=8pt}
\caption{Very sparse perturbations.}
\label{fig:mnist_very}
\end{subfigure}\hfill
\begin{subfigure}[b]{0.3\linewidth}
\captionsetup[subfigure]{skip=1pt}
    \begin{subfigure}[b]{0.2\linewidth}
        \includegraphics[width=\linewidth]{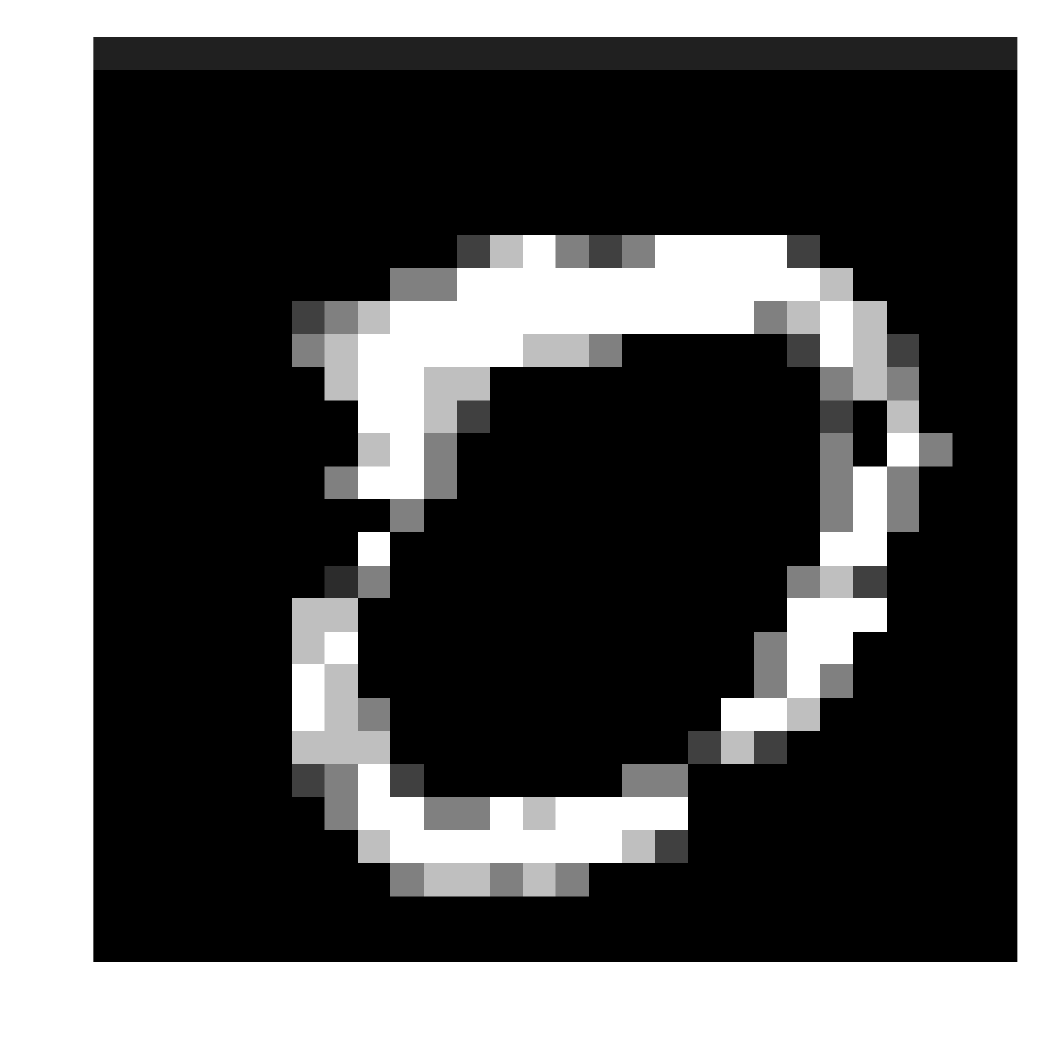}\!
        \captionsetup{font=scriptsize}
        \caption*{$5$ (9)}
    \end{subfigure}\!
    \begin{subfigure}[b]{0.2\linewidth}
        \includegraphics[width=\linewidth]{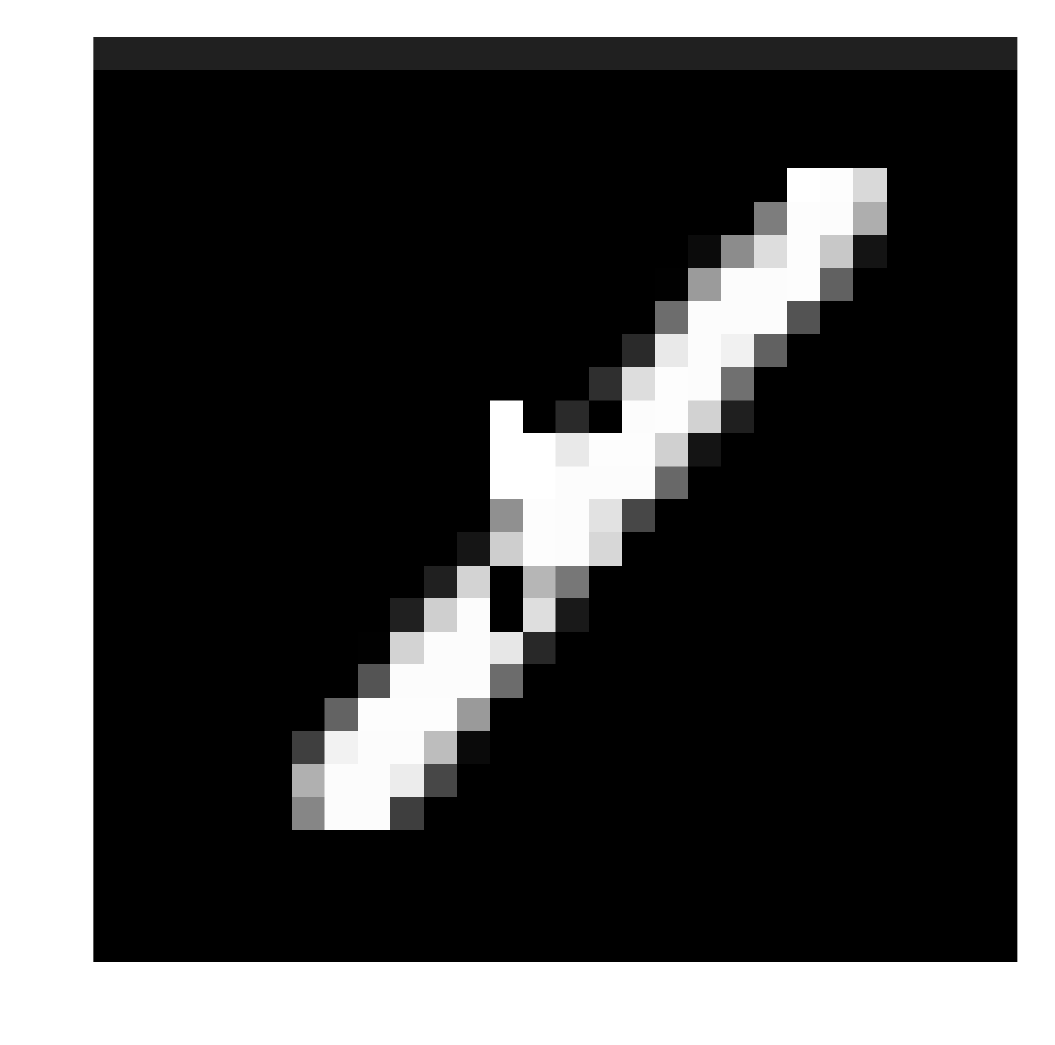}\!
        \captionsetup{font=scriptsize}
        \caption*{$4$ (9)}
    \end{subfigure}\!
    \begin{subfigure}[b]{0.2\linewidth}
        \includegraphics[width=\linewidth]{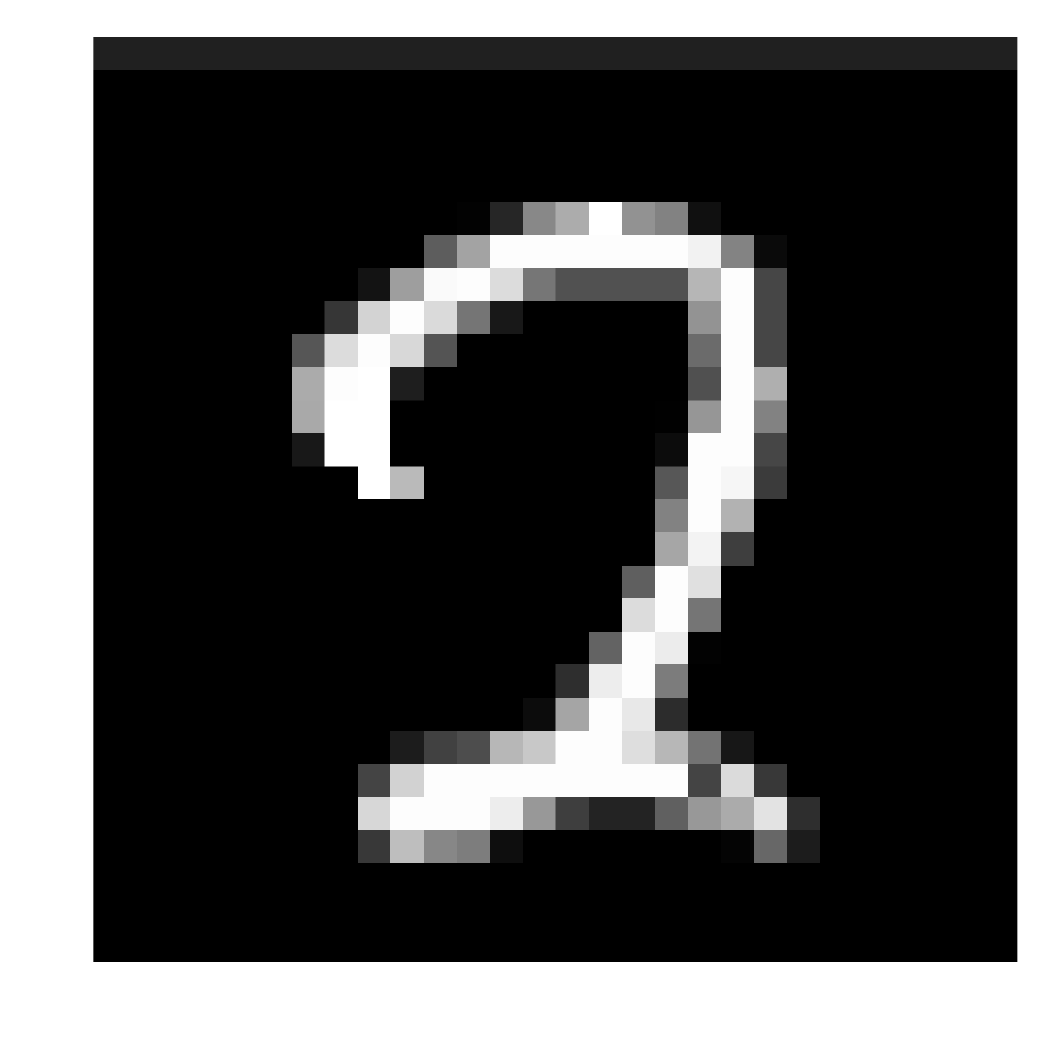}\!
        \captionsetup{font=scriptsize}
        \caption*{$9$ (9)}
    \end{subfigure}\!
    \begin{subfigure}[b]{0.2\linewidth}
        \includegraphics[width=\linewidth]{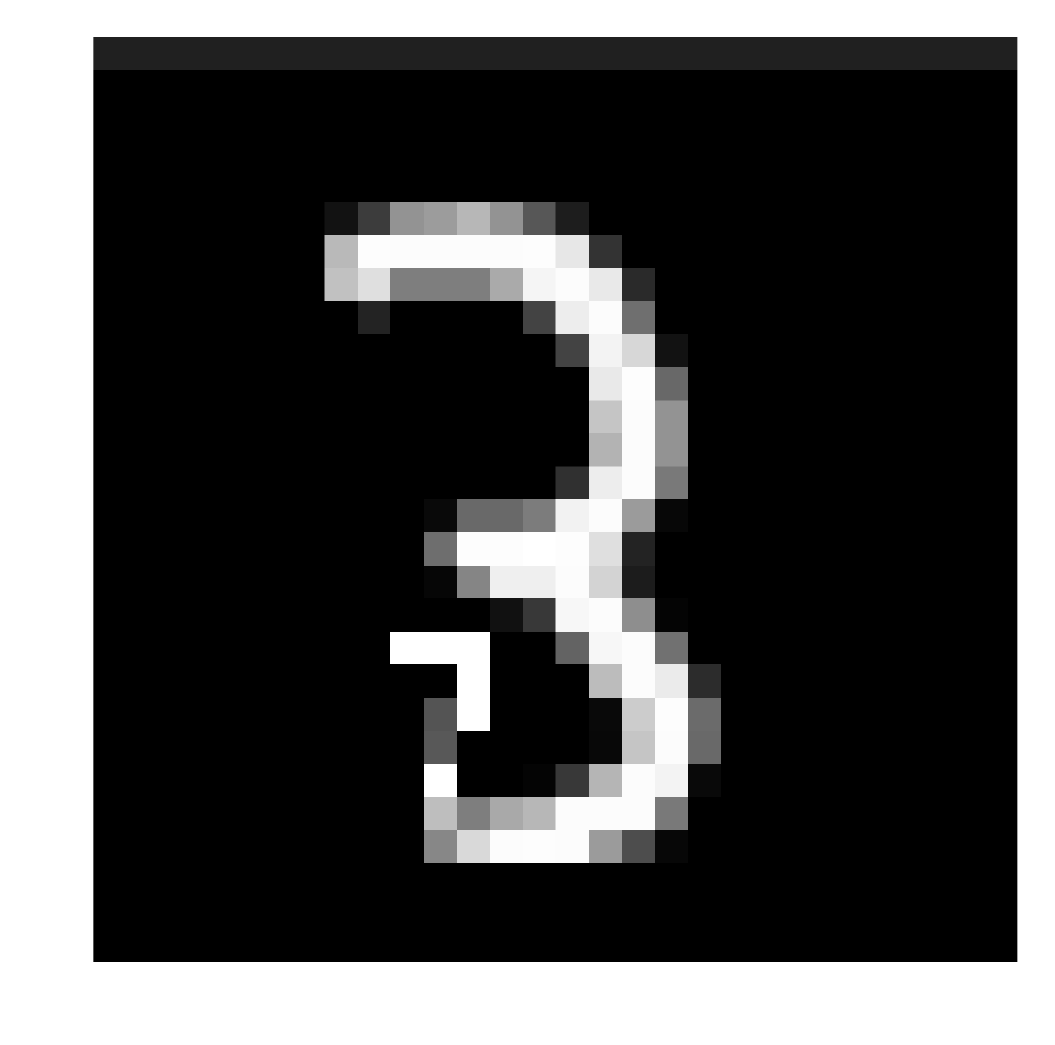}\!
        \captionsetup{font=scriptsize}
        \caption*{$2$ (9)}
    \end{subfigure}\!
    \begin{subfigure}[b]{0.2\linewidth}
        \includegraphics[width=\linewidth]{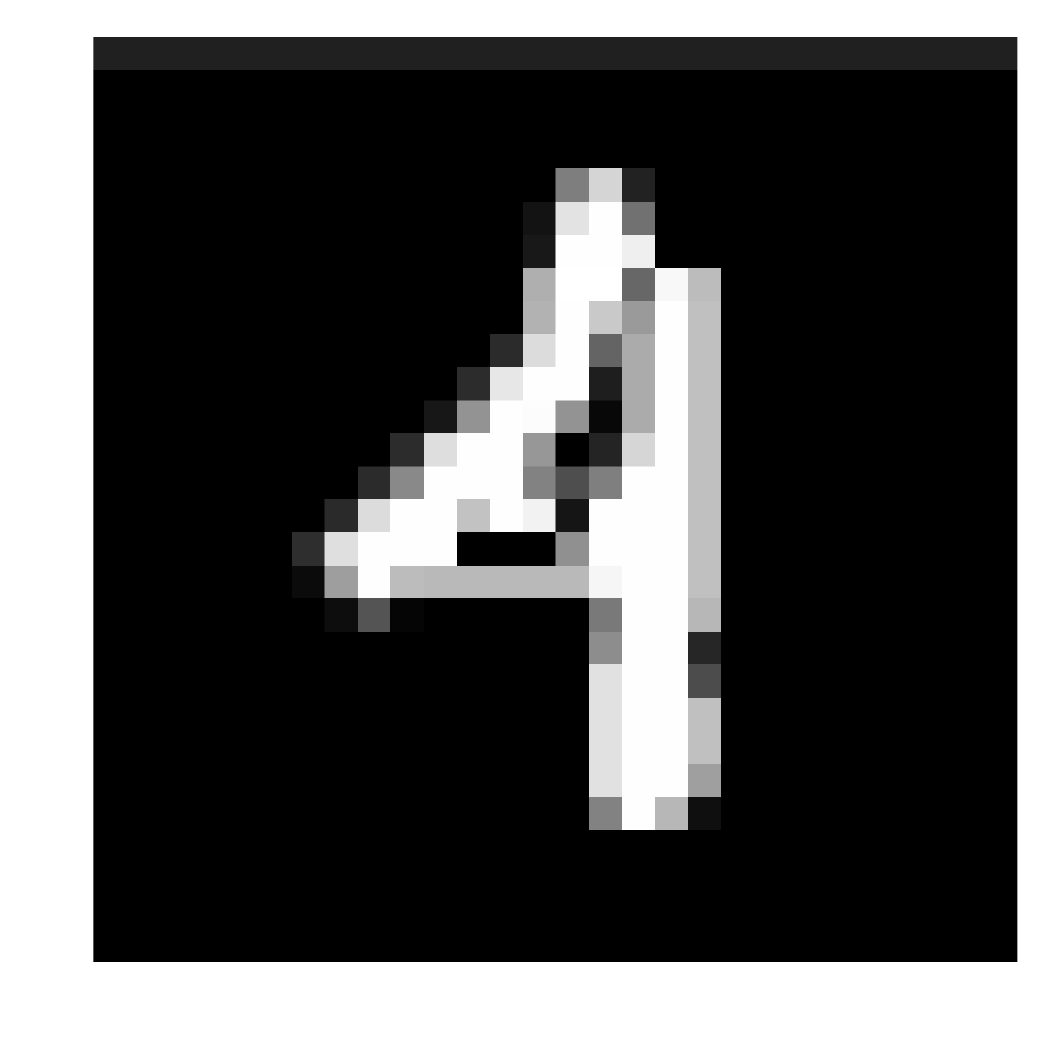}\!
        \captionsetup{font=scriptsize}
        \caption*{$1$ (9)}
    \end{subfigure}\!
    
    \begin{subfigure}[b]{0.2\linewidth}
        \includegraphics[width=\linewidth]{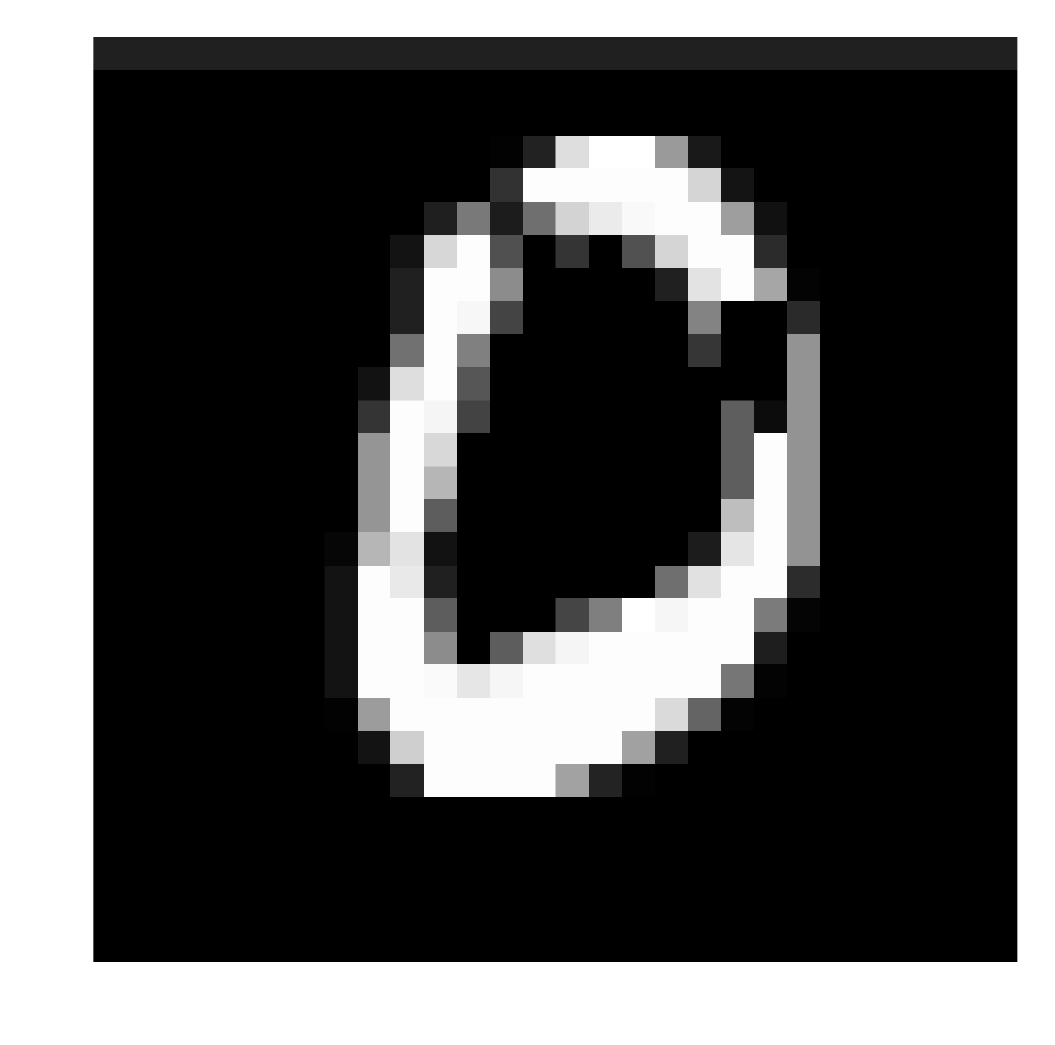}\!
        \captionsetup{font=scriptsize}
        \caption*{$6$ (9)}
    \end{subfigure}\!
    \begin{subfigure}[b]{0.2\linewidth}
        \includegraphics[width=\linewidth]{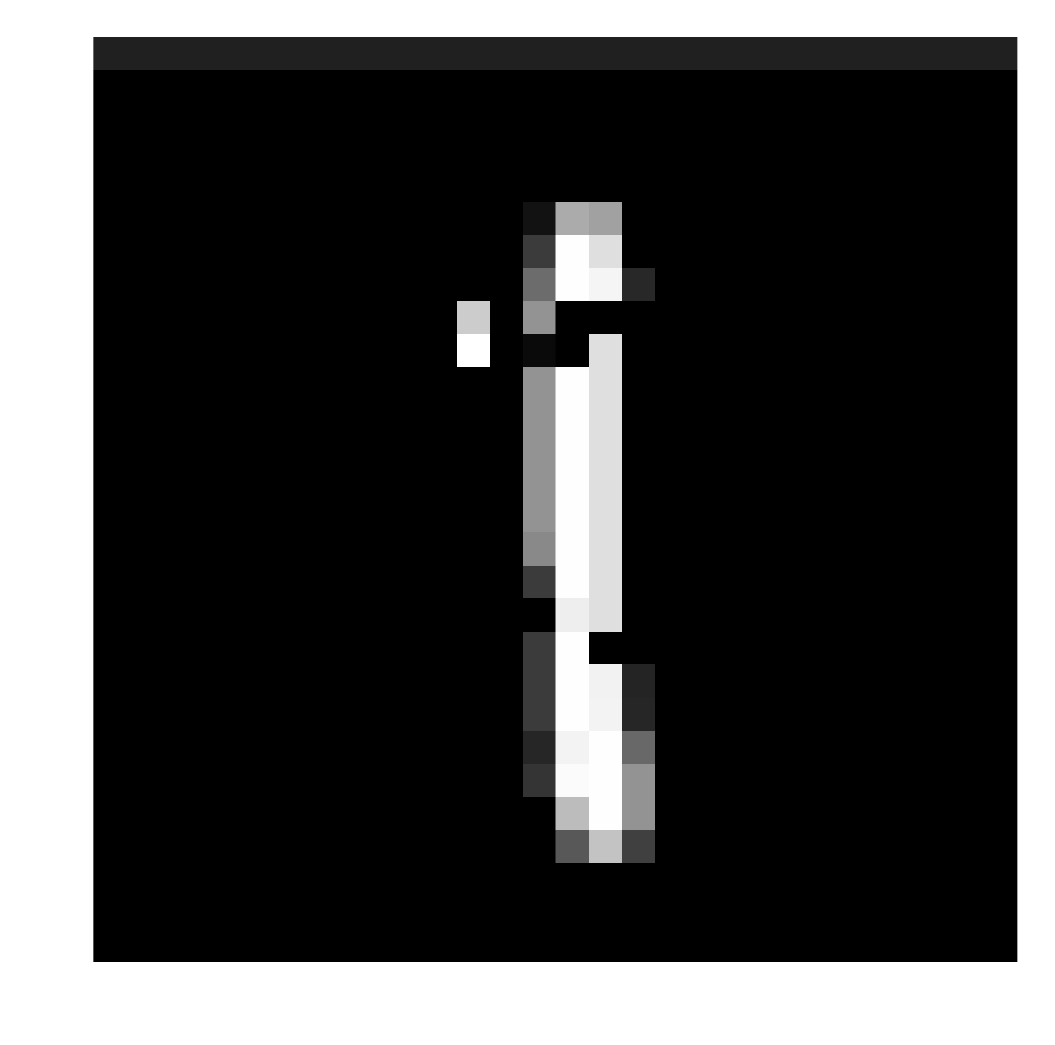}\!
        \captionsetup{font=scriptsize}
        \caption*{$9$ (9)}
    \end{subfigure}\!
    \begin{subfigure}[b]{0.2\linewidth}
        \includegraphics[width=\linewidth]{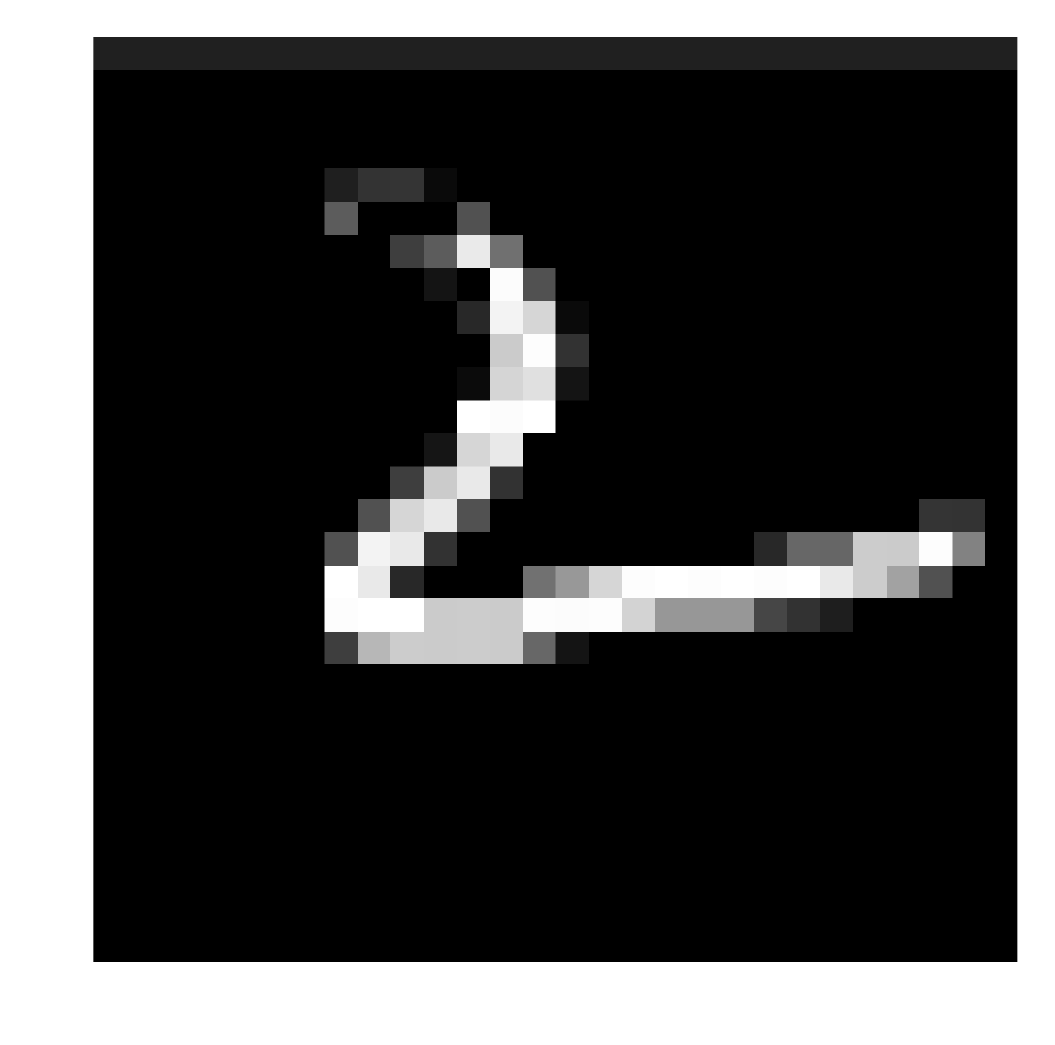}\!
        \captionsetup{font=scriptsize}
        \caption*{$4$ (9)}
    \end{subfigure}\!
    \begin{subfigure}[b]{0.2\linewidth}
        \includegraphics[width=\linewidth]{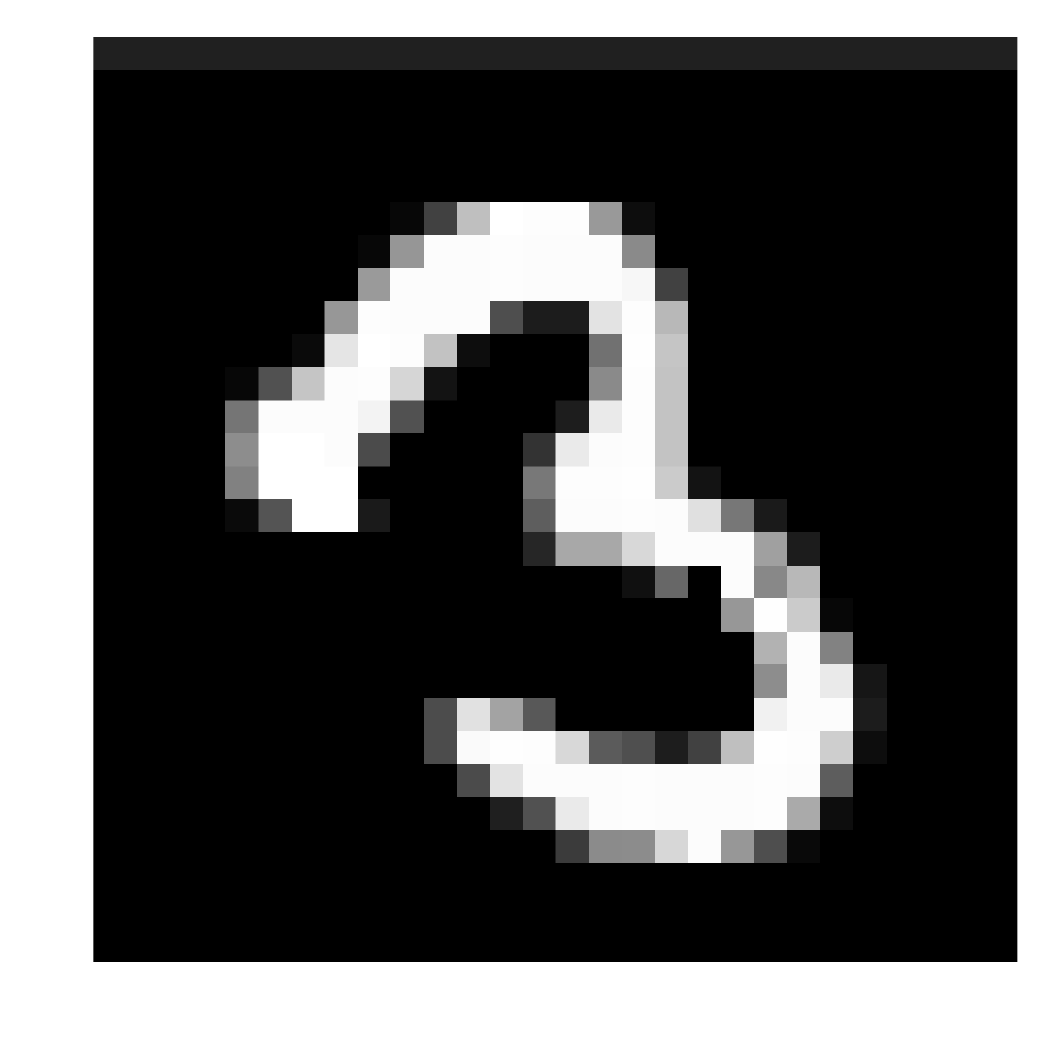}\!
        \captionsetup{font=scriptsize}
        \caption*{$9$ (10)}
    \end{subfigure}\!
    \begin{subfigure}[b]{0.2\linewidth}
        \includegraphics[width=\linewidth]{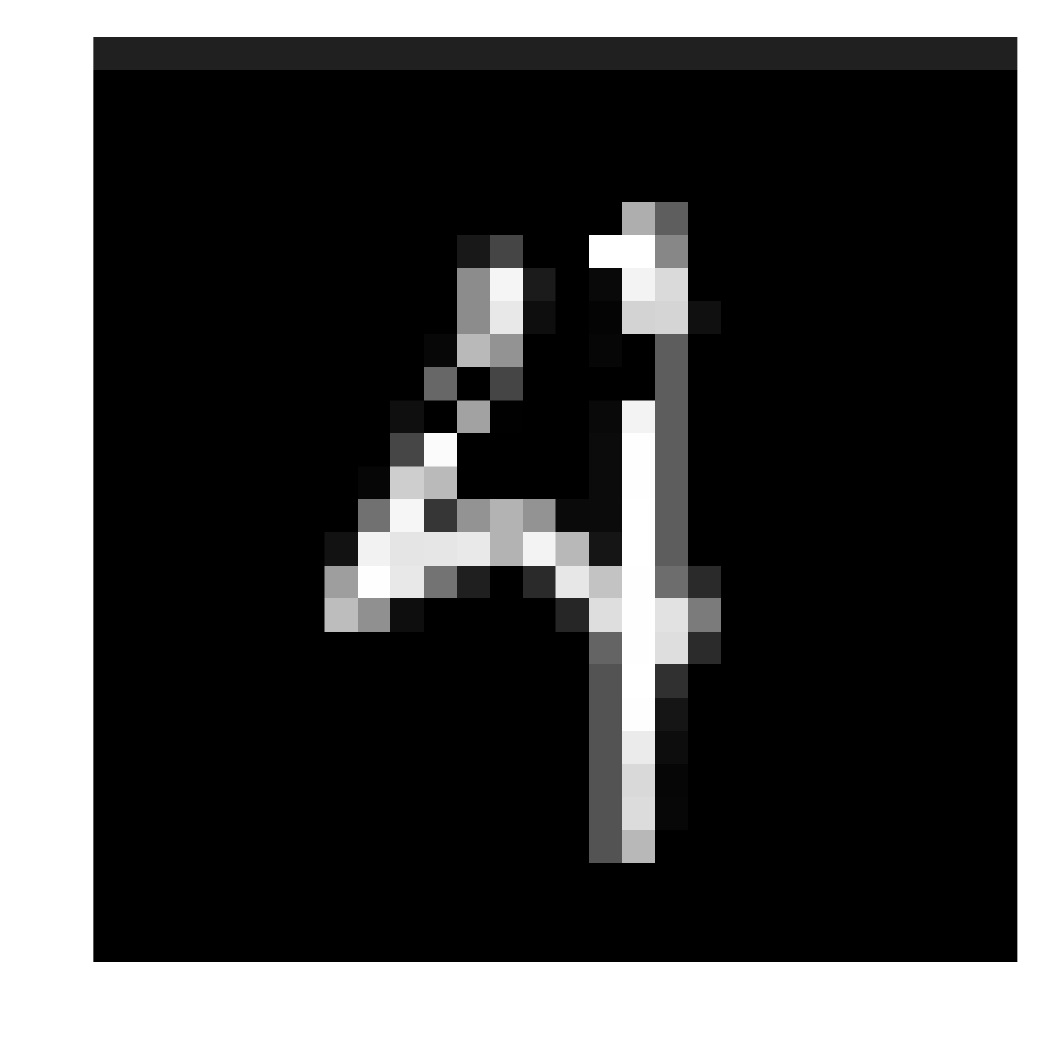}\!
        \captionsetup{font=scriptsize}
        \caption*{$2$ (9)}
    \end{subfigure}\!
    
    \begin{subfigure}[b]{0.2\linewidth}
        \includegraphics[width=\linewidth]{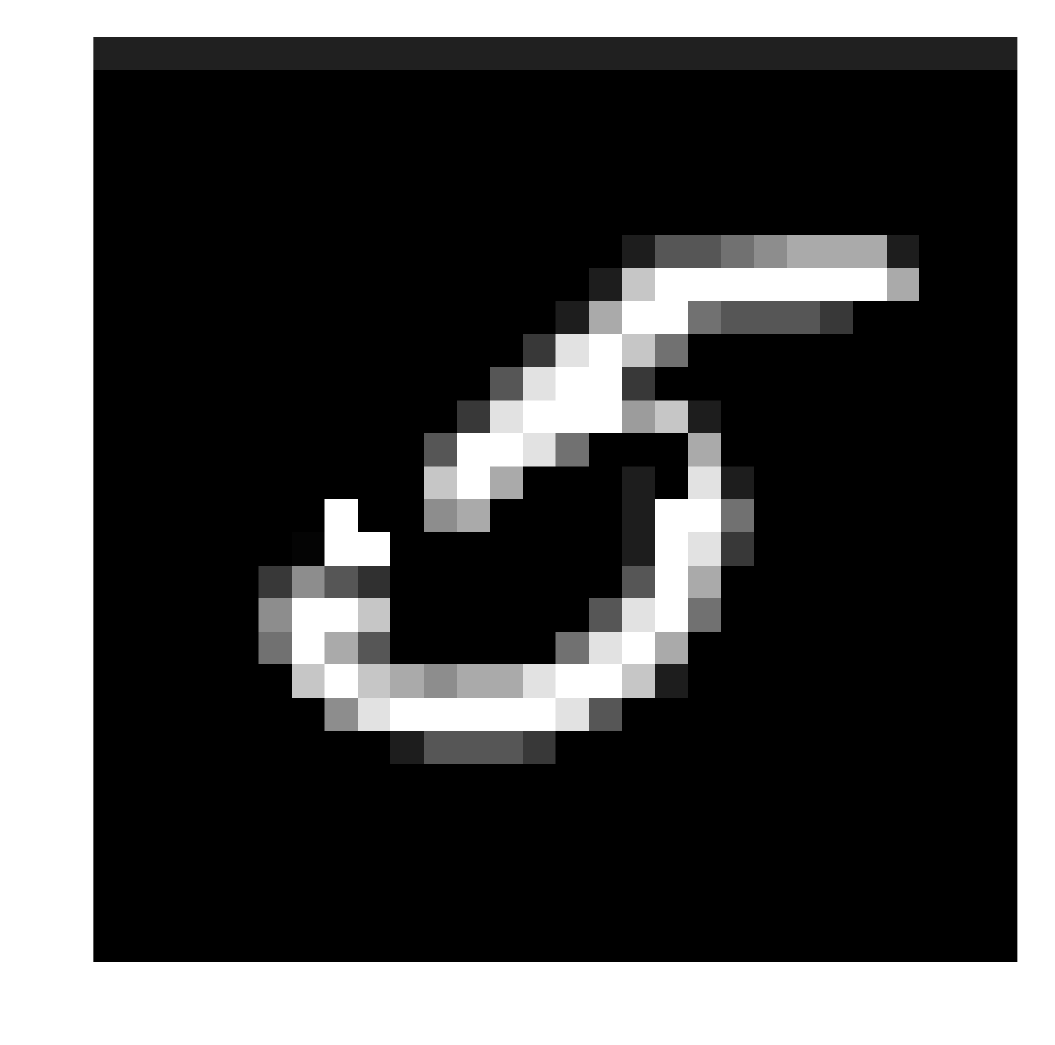}\!
        \captionsetup{font=scriptsize}
        \caption*{$0$ (10)}
    \end{subfigure}\!
    \begin{subfigure}[b]{0.2\linewidth}
        \includegraphics[width=\linewidth]{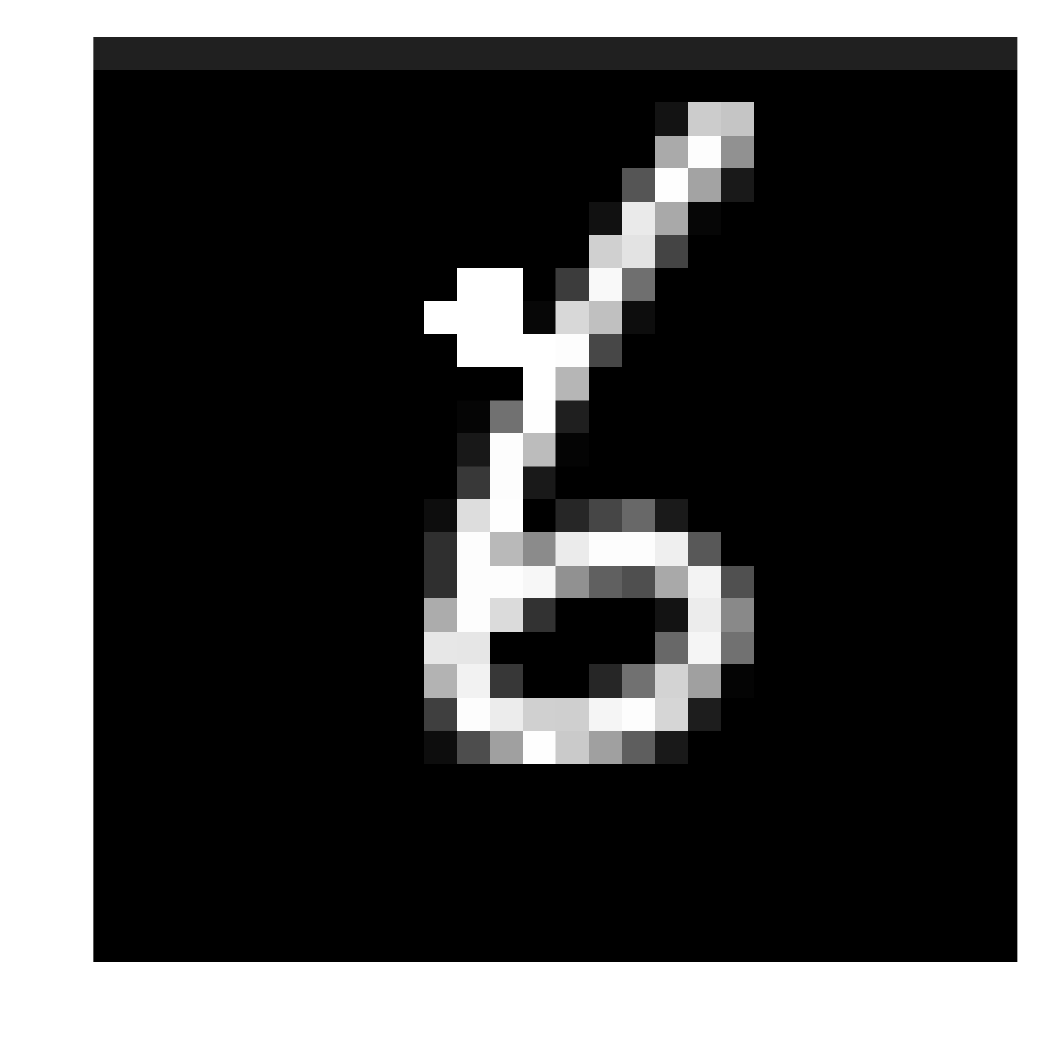}\!
        \captionsetup{font=scriptsize}
        \caption*{$2$ (9)}
    \end{subfigure}\!
    \begin{subfigure}[b]{0.2\linewidth}
        \includegraphics[width=\linewidth]{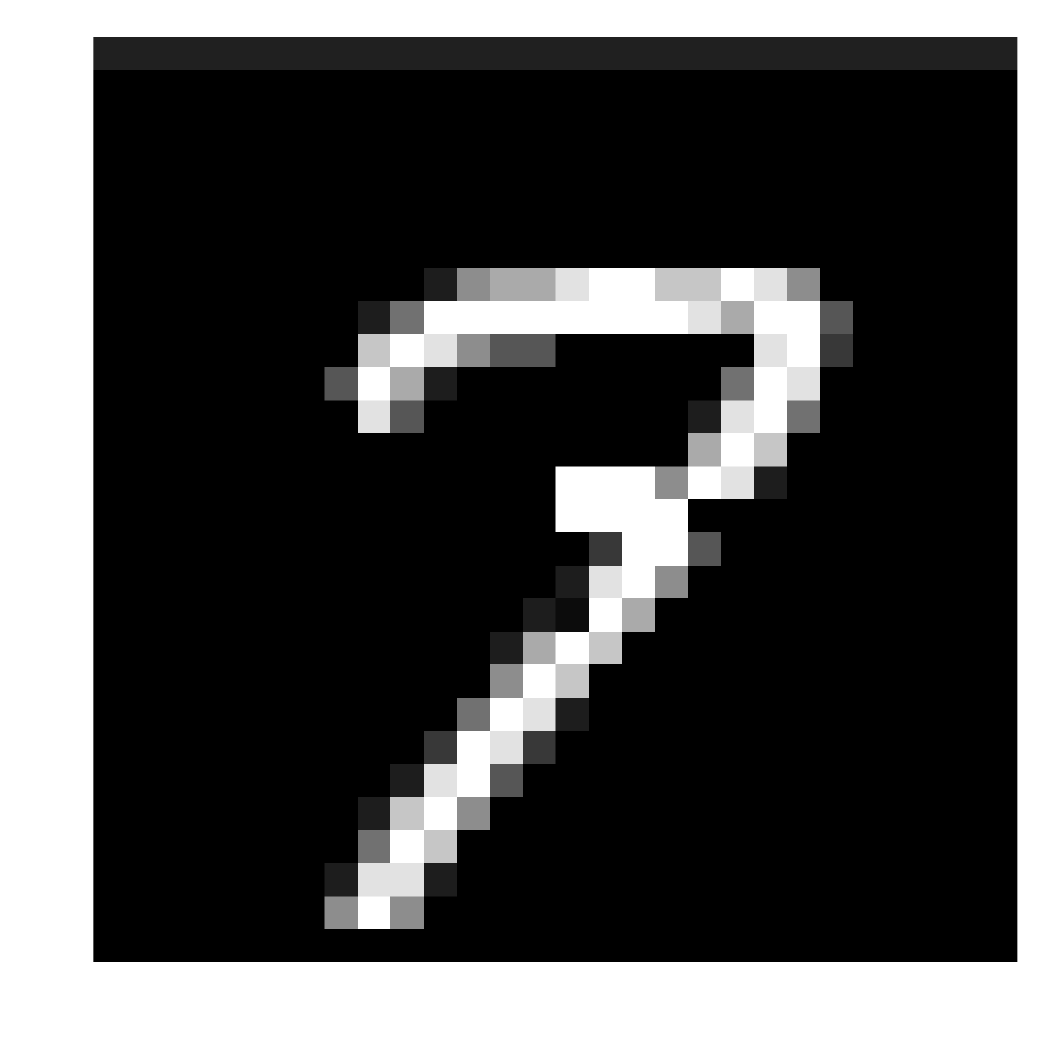}\!
        \captionsetup{font=scriptsize}
        \caption*{$3$ (9)}
    \end{subfigure}\!
    \begin{subfigure}[b]{0.2\linewidth}
        \includegraphics[width=\linewidth]{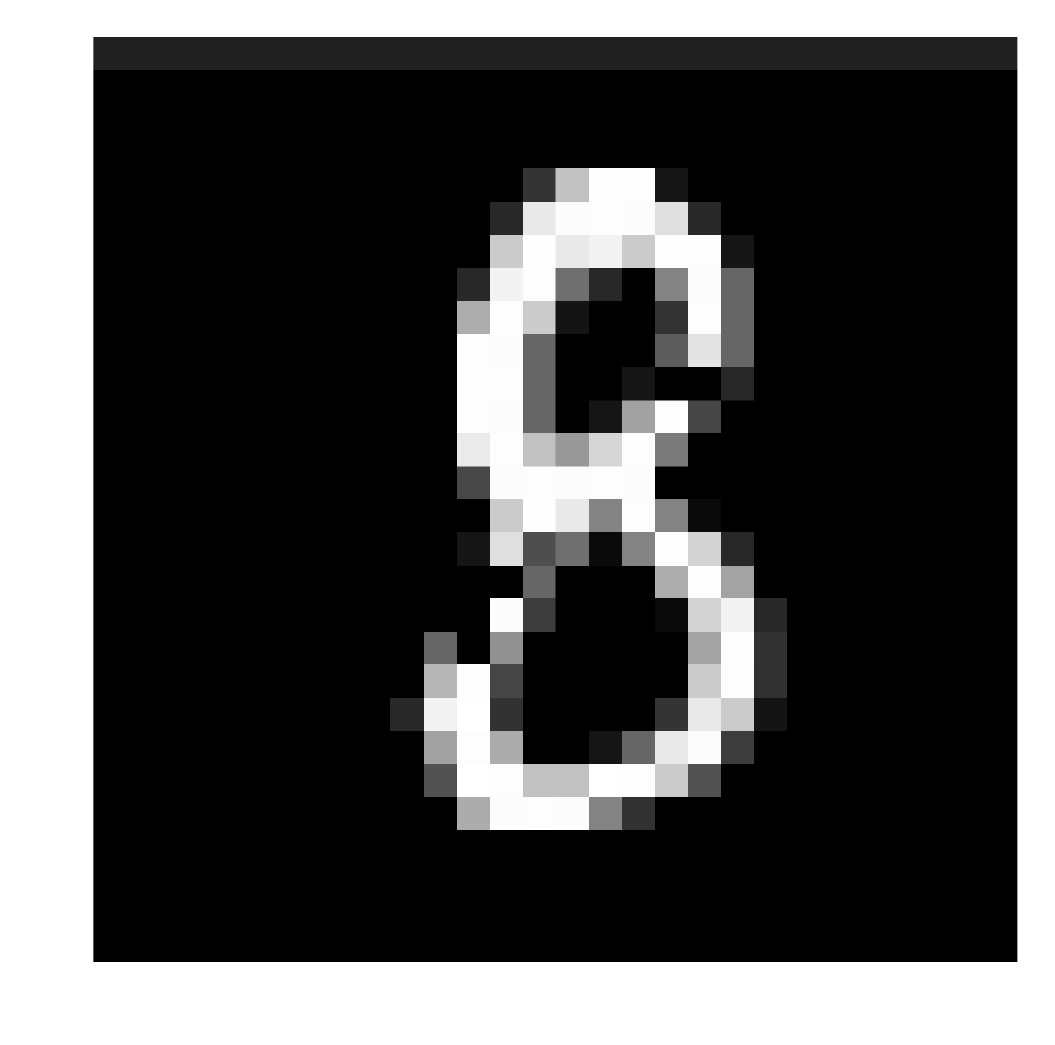}\!
        \captionsetup{font=scriptsize}
        \caption*{$5$ (9)}
    \end{subfigure}\!
    \begin{subfigure}[b]{0.2\linewidth}
        \includegraphics[width=\linewidth]{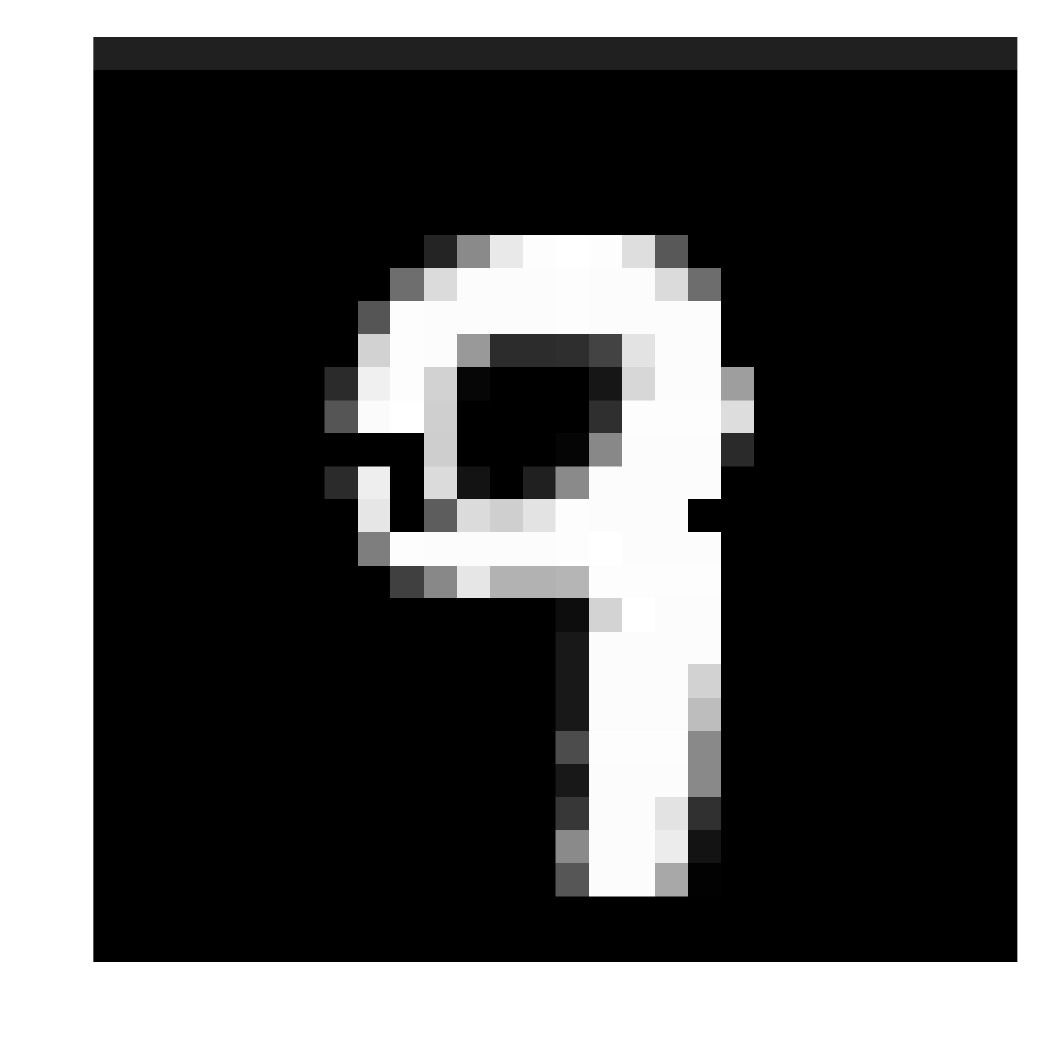}\!
        \captionsetup{font=scriptsize}
        \caption*{$3$ (9)}
    \end{subfigure}\!
    
    \begin{subfigure}[b]{0.2\linewidth}
        \includegraphics[width=\linewidth]{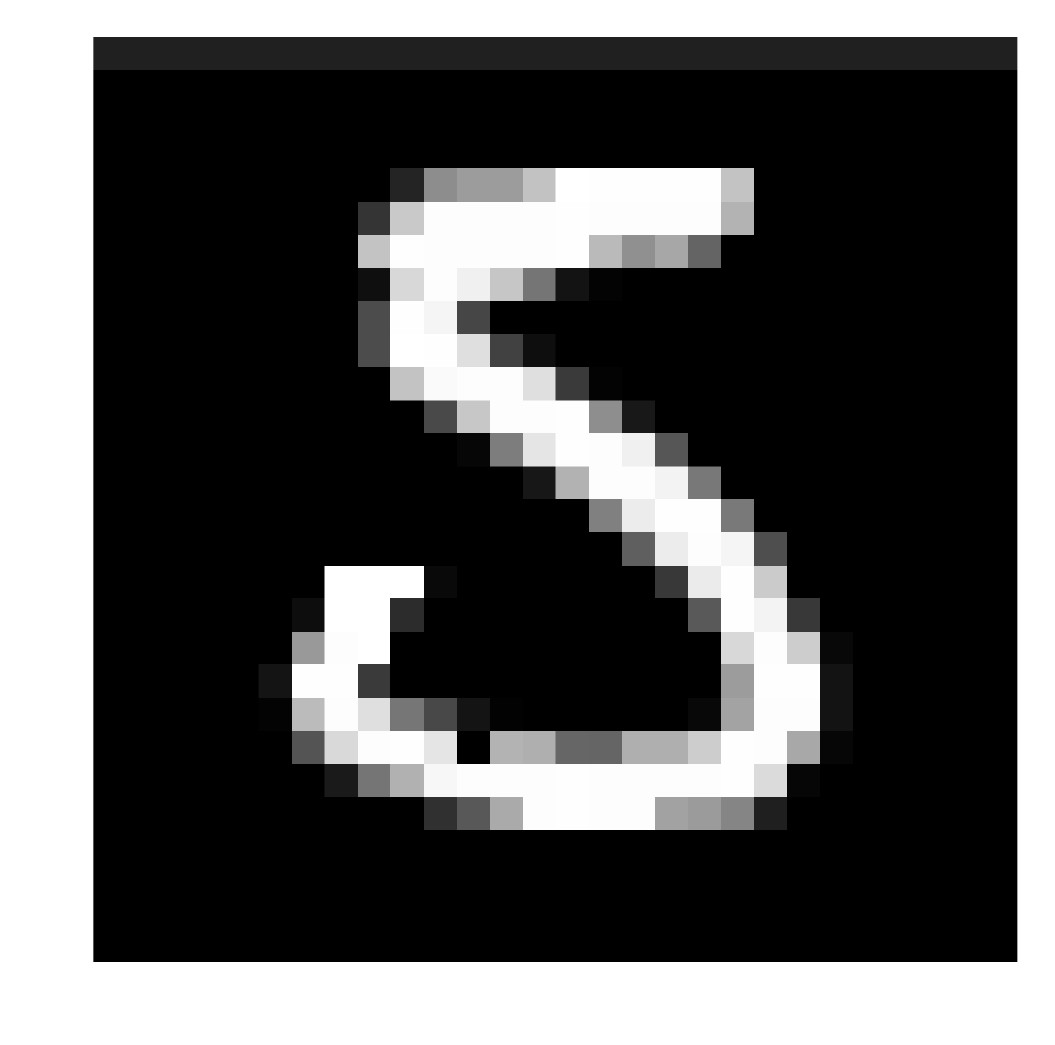}\!
        \captionsetup{font=scriptsize}
        \caption*{$8$ (10)}
    \end{subfigure}\!
    \begin{subfigure}[b]{0.2\linewidth}
        \includegraphics[width=\linewidth]{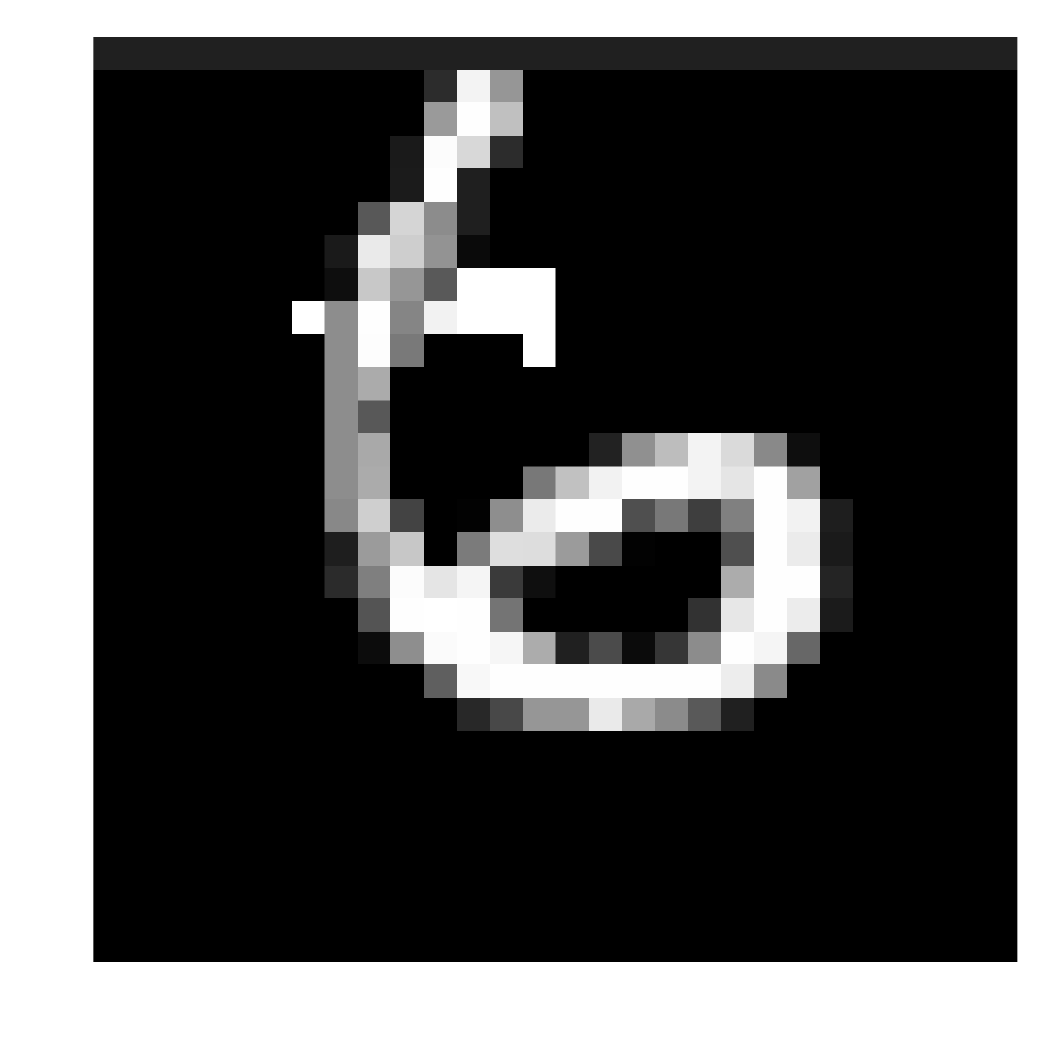}\!
        \captionsetup{font=scriptsize}
        \caption*{$3$ (9)}
    \end{subfigure}\!
    \begin{subfigure}[b]{0.2\linewidth}
        \includegraphics[width=\linewidth]{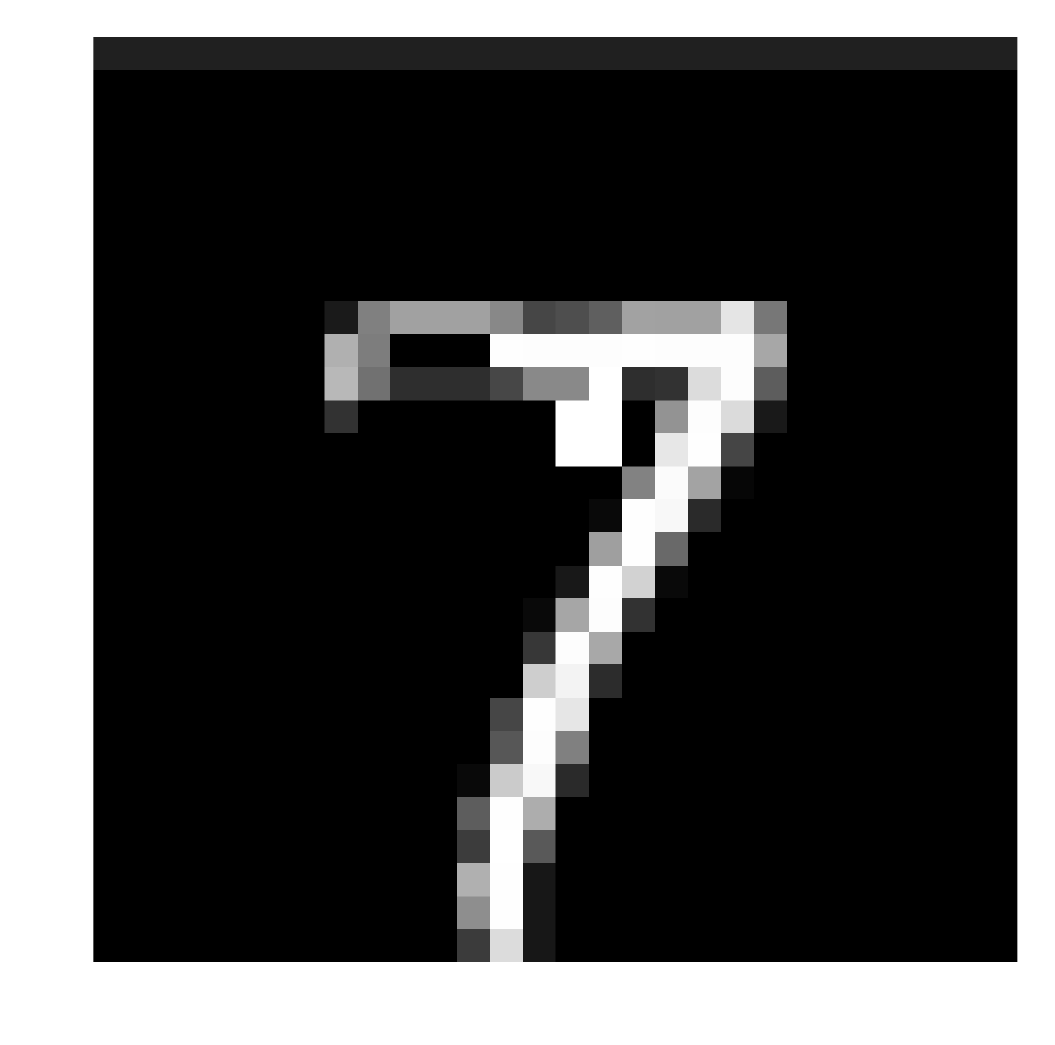}\!
        \captionsetup{font=scriptsize}
        \caption*{$9$ (9)}
    \end{subfigure}\!
    \begin{subfigure}[b]{0.2\linewidth}
        \includegraphics[width=\linewidth]{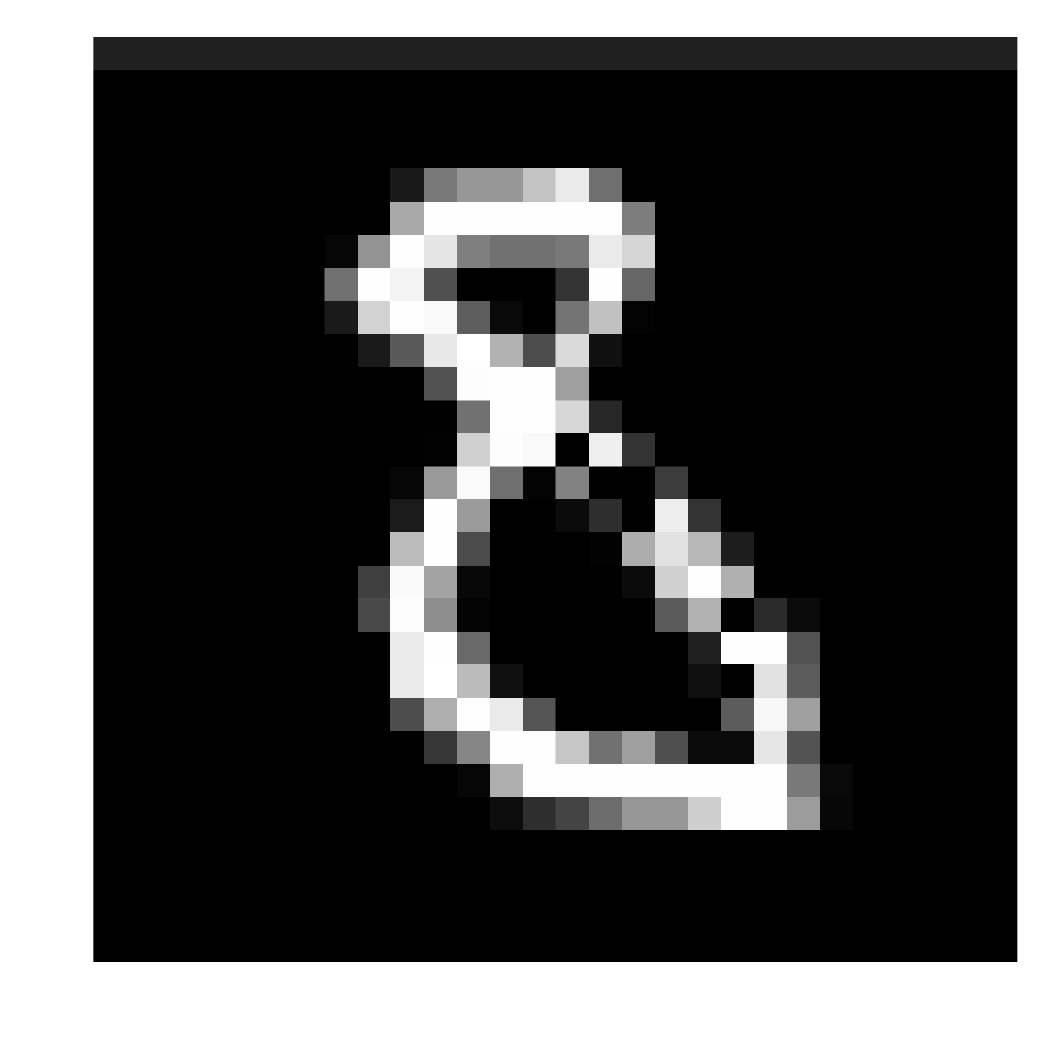}\!
        \captionsetup{font=scriptsize}
        \caption*{$2$ (9)}
    \end{subfigure}\!
    \begin{subfigure}[b]{0.2\linewidth}
        \includegraphics[width=\linewidth]{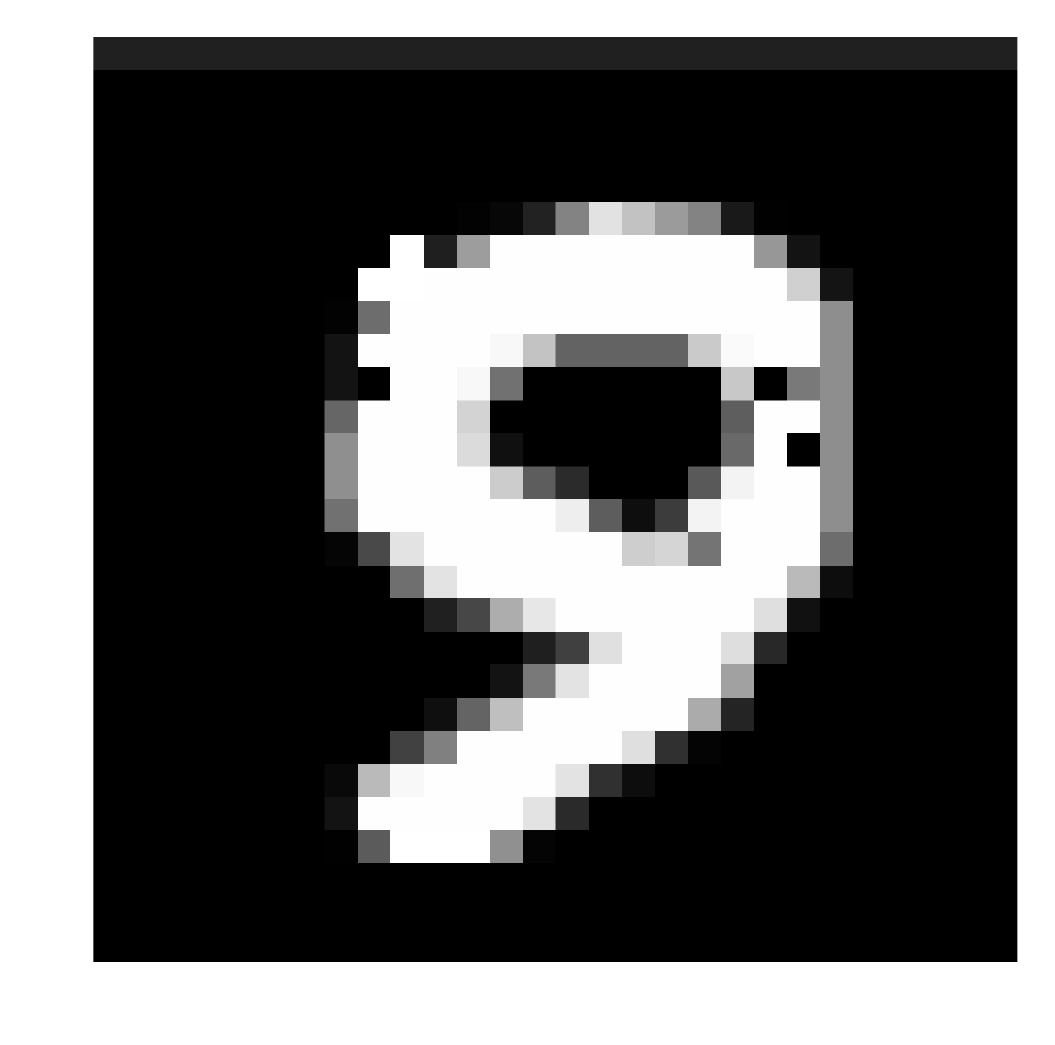}\!
        \captionsetup{font=scriptsize}
        \caption*{$5$ (9)}
    \end{subfigure}\!
\captionsetup{font=small, skip=8pt}
\caption{Sparse perturbations.}
\label{fig:mnist_sparse}
\end{subfigure}\hfill
\caption{SparseFool adversarial examples for the MNIST dataset for different levels of sparsity. The fooling label is shown below the image, and the number of perturbed pixels is written inside the parentheses.}
\label{fig:mnist_visual}
\end{figure}

In this section, we provide some supplementary adversarial examples generated by SparseFool for different datasets. These examples correspond to three different levels of sparsity: (a) highly sparse, (b) very sparse, and (c) sparse perturbations. The corresponding results for the ImageNet, CIFAR-10, and MNIST datasets are shown in Figures~[\ref{fig:imnet_visual} -- \ref{fig:mnist_visual}] respectively. Observer that for highly and very sparse perturbations, the noise is either imperceptible, or sparse enough to be negligible. However, as the number of perturbed pixels is increased, the noise may become quite perceptible.

\section{Controlling the perceptibility of the perturbations}
\begin{figure}[hb]
\centering
\begin{subfigure}[b]{\linewidth}
\captionsetup[subfigure]{skip=1pt, font=scriptsize}
\centering
    \begin{subfigure}[b]{0.16\linewidth}
        \includegraphics[width=\linewidth]{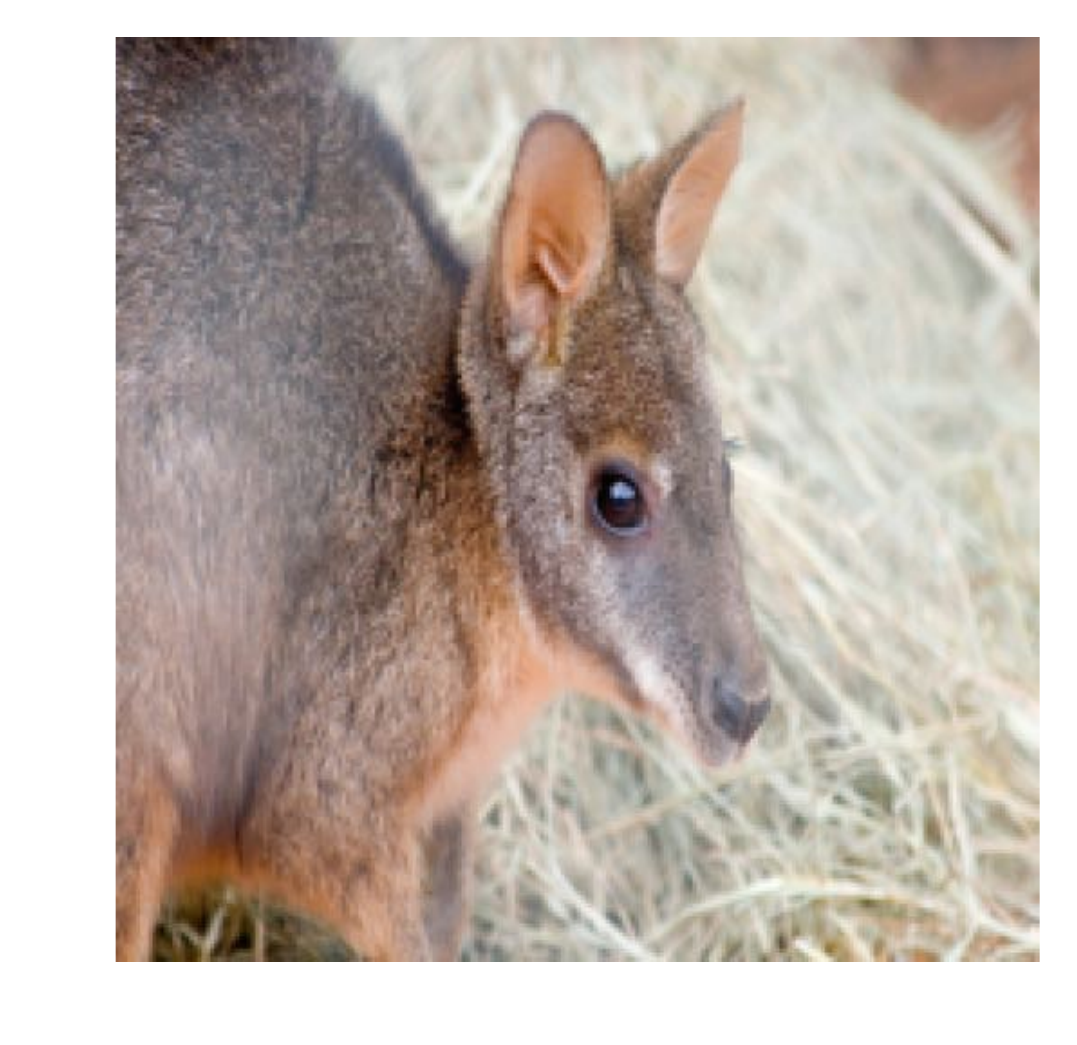}\!
        \caption*{wallaby}
    \end{subfigure}
    \begin{subfigure}[b]{0.16\linewidth}
        \includegraphics[width=\linewidth]{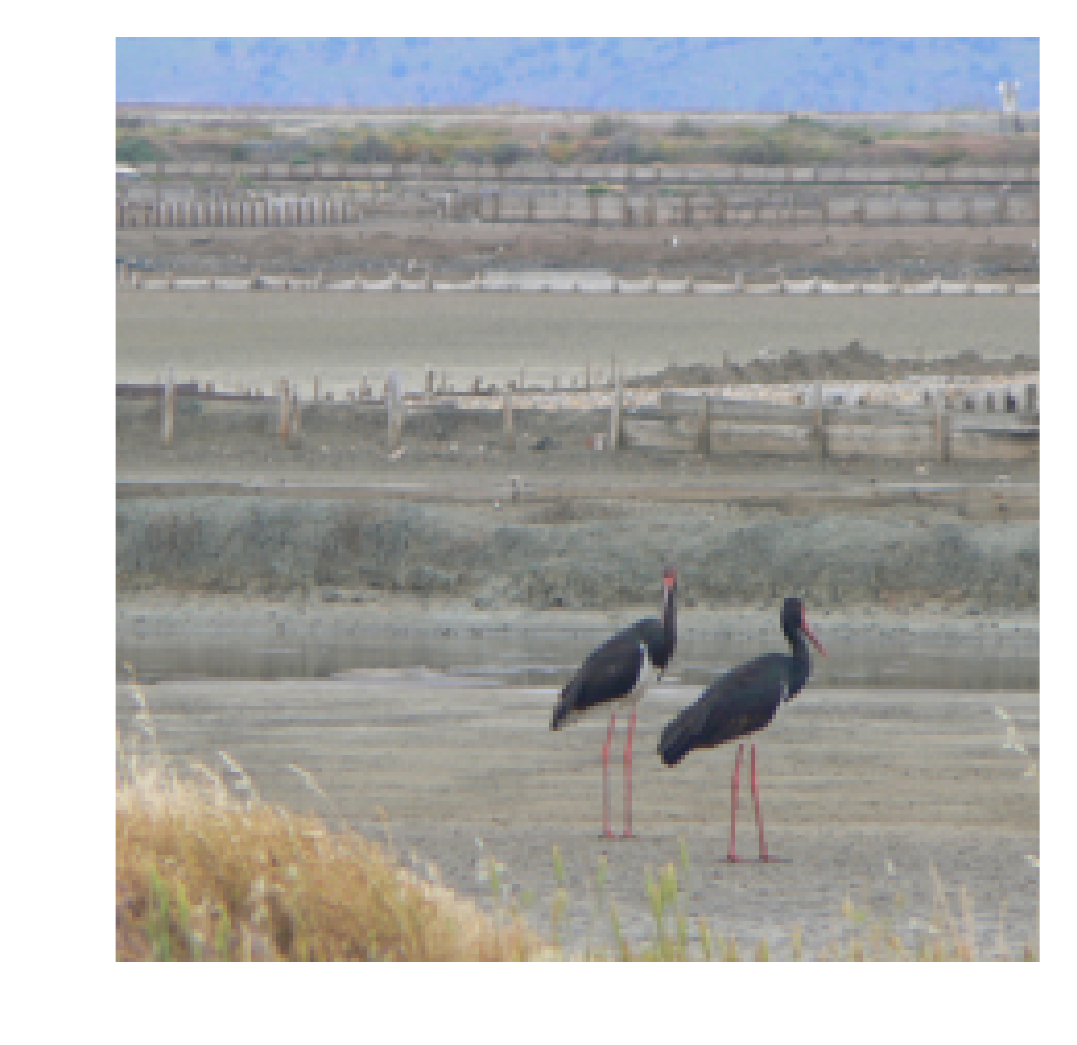}\!
        \caption*{black stork}
    \end{subfigure}
    \begin{subfigure}[b]{0.16\linewidth}
        \includegraphics[width=\linewidth]{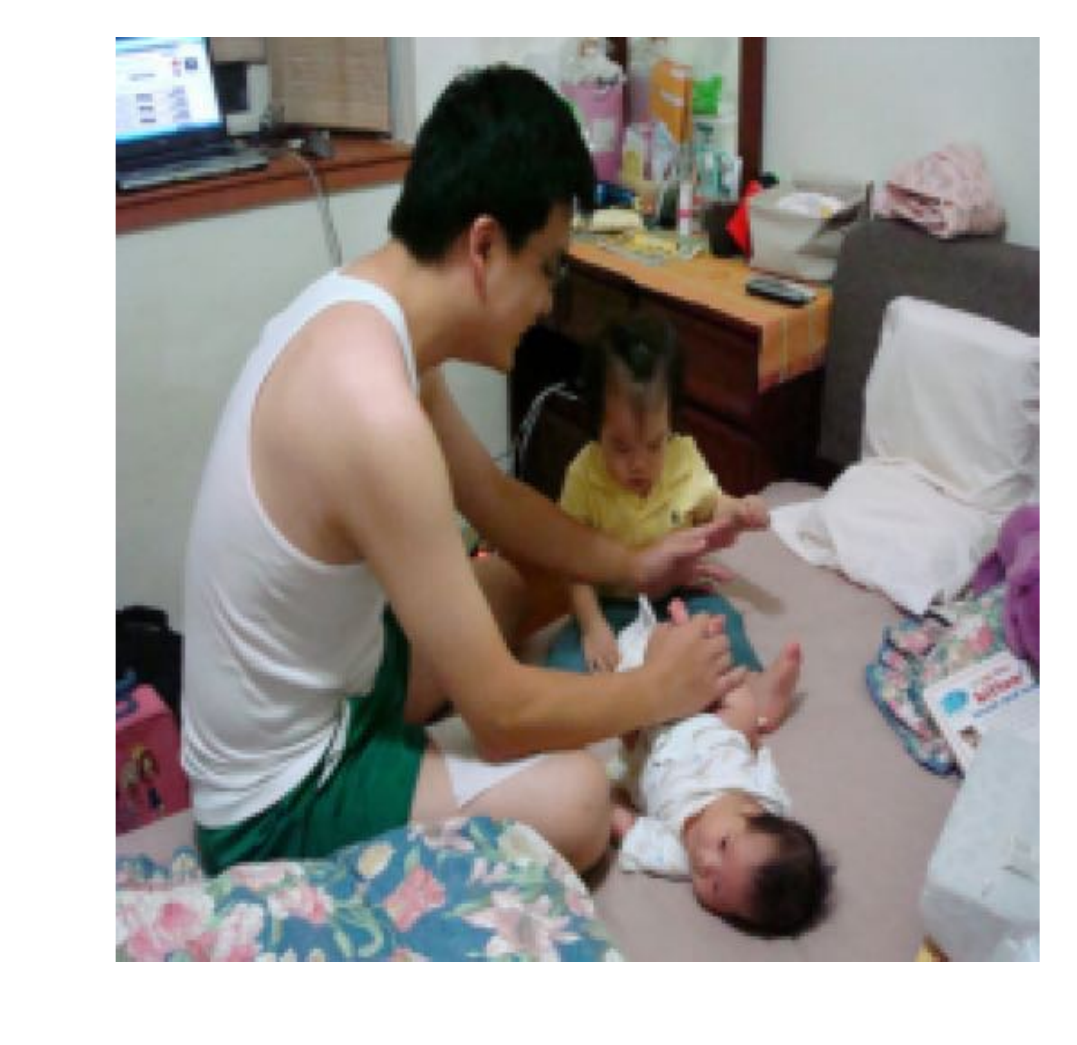}\!
        \caption*{diaper}
    \end{subfigure}
    \begin{subfigure}[b]{0.16\linewidth}
        \includegraphics[width=\linewidth]{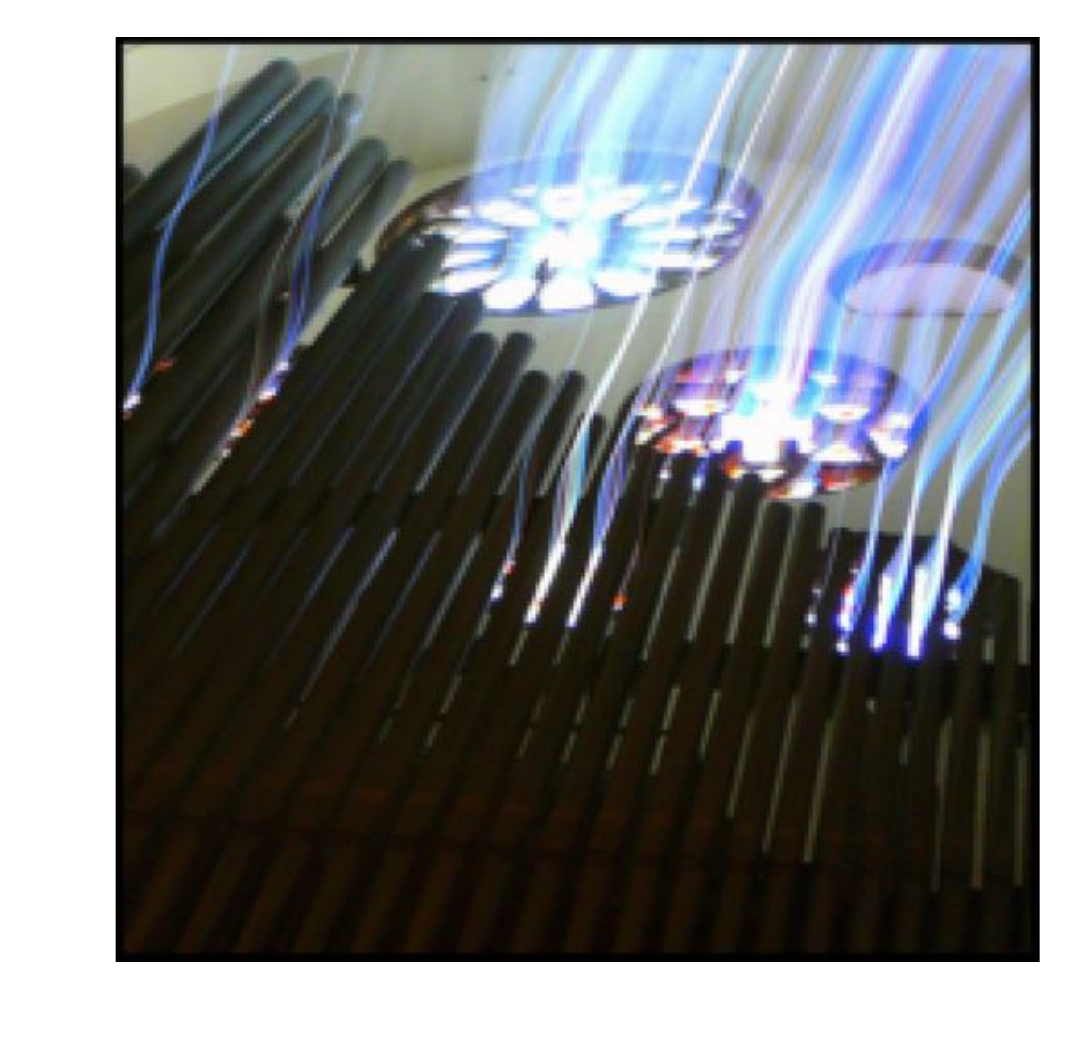}\!
        \caption*{organ}
    \end{subfigure}
    \begin{subfigure}[b]{0.16\linewidth}
        \includegraphics[width=\linewidth]{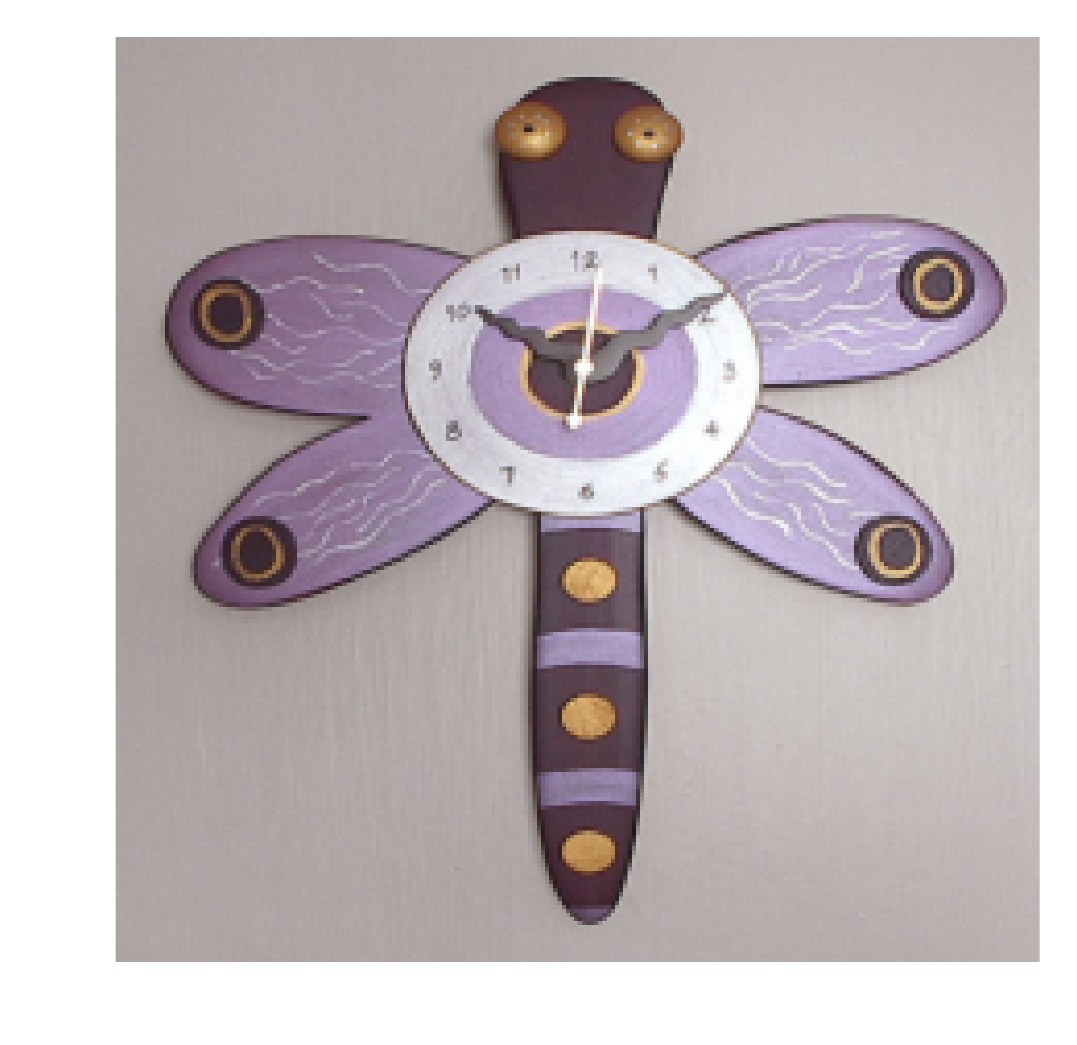}\!
        \caption*{wall clock}
    \end{subfigure}
    \begin{subfigure}[b]{0.16\linewidth}
        \includegraphics[width=\linewidth]{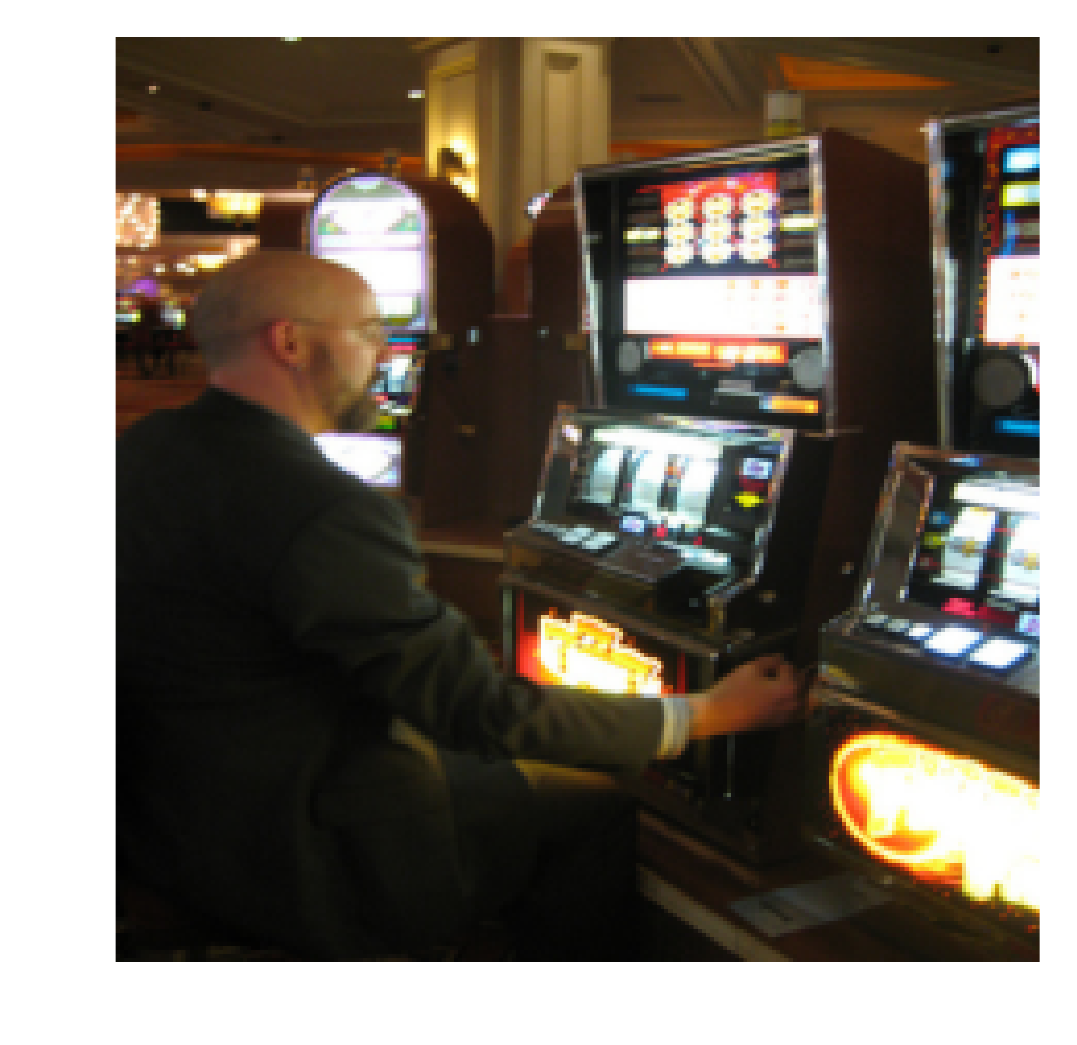}\!
        \caption*{slot}
    \end{subfigure}
    
    \begin{subfigure}[b]{0.16\linewidth}
        \includegraphics[width=\linewidth]{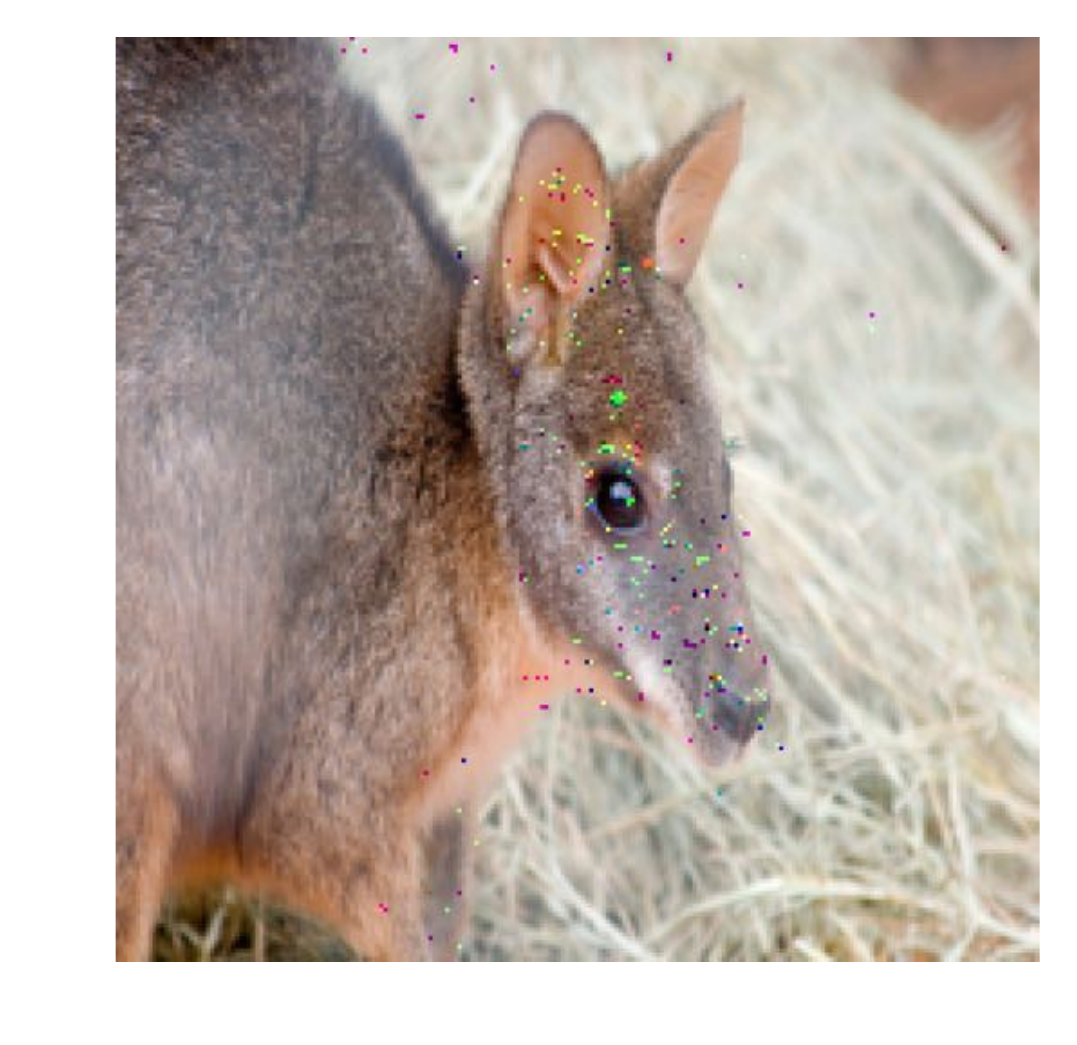}\!
        \caption*{wood rabbit ($0.6637\%$)}
    \end{subfigure}
    \begin{subfigure}[b]{0.16\linewidth}
        \includegraphics[width=\linewidth]{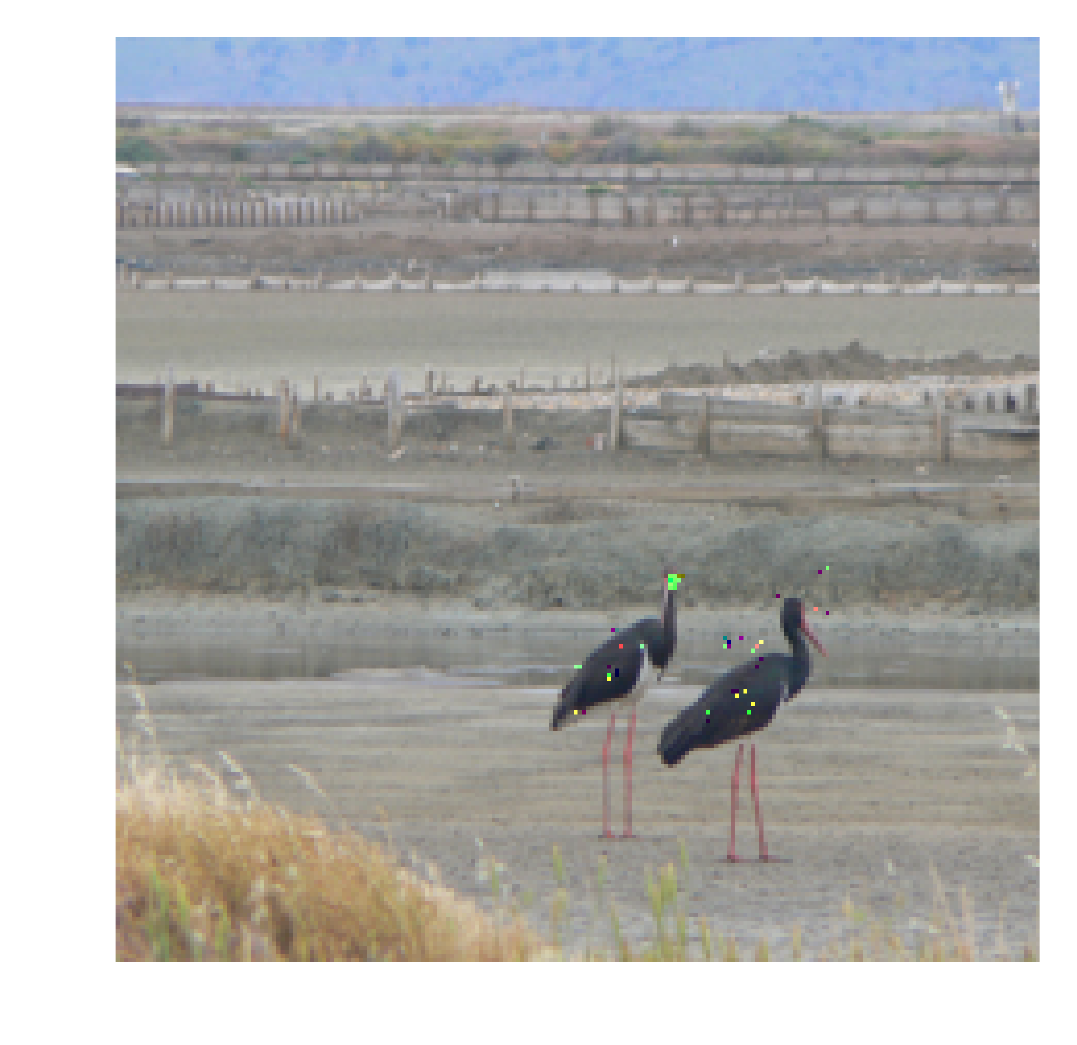}\!
        \caption*{white stork ($0.0957\%$)}
    \end{subfigure}
    \begin{subfigure}[b]{0.16\linewidth}
        \includegraphics[width=\linewidth]{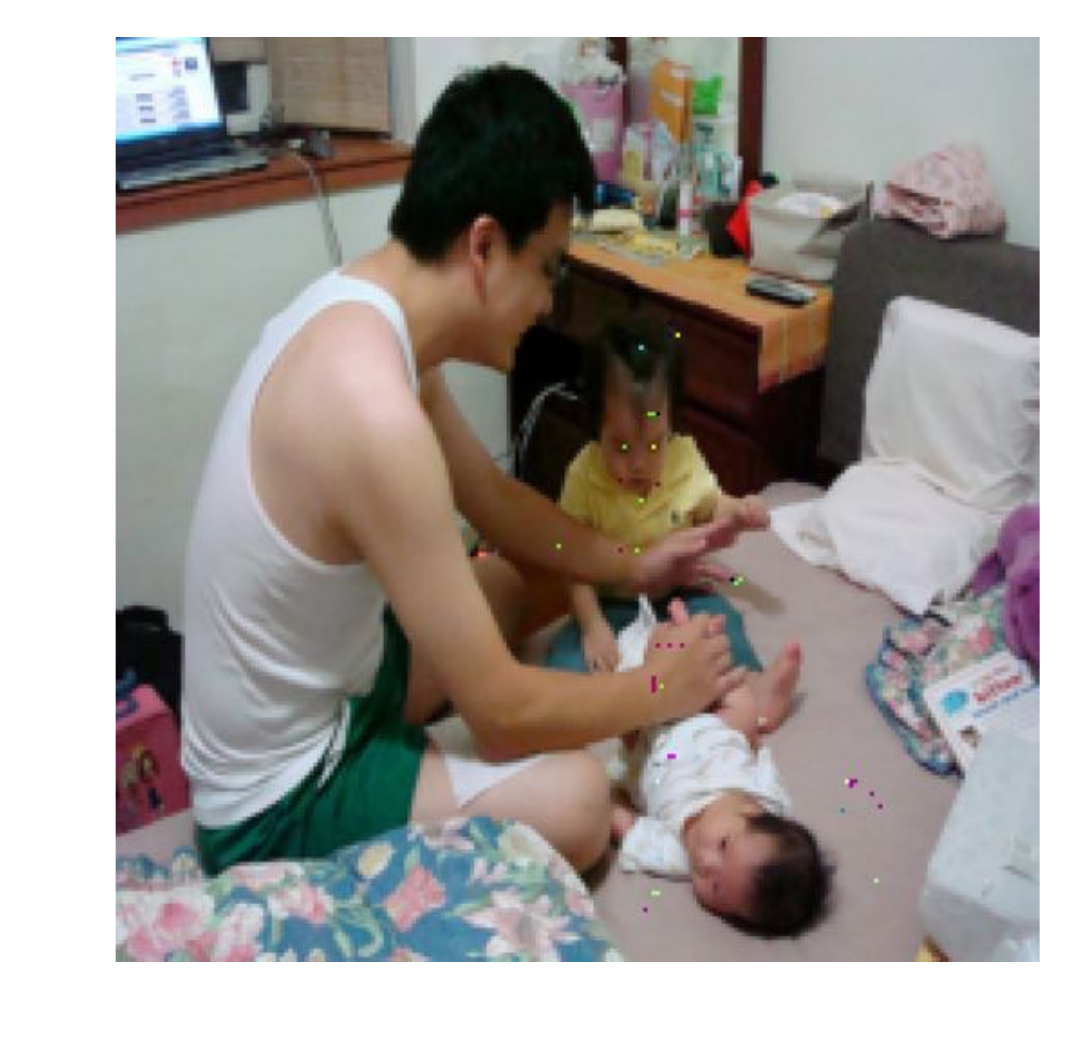}\!
        \caption*{potter's wheel ($0.0797\%$)}
    \end{subfigure}
    \begin{subfigure}[b]{0.16\linewidth}
        \includegraphics[width=\linewidth]{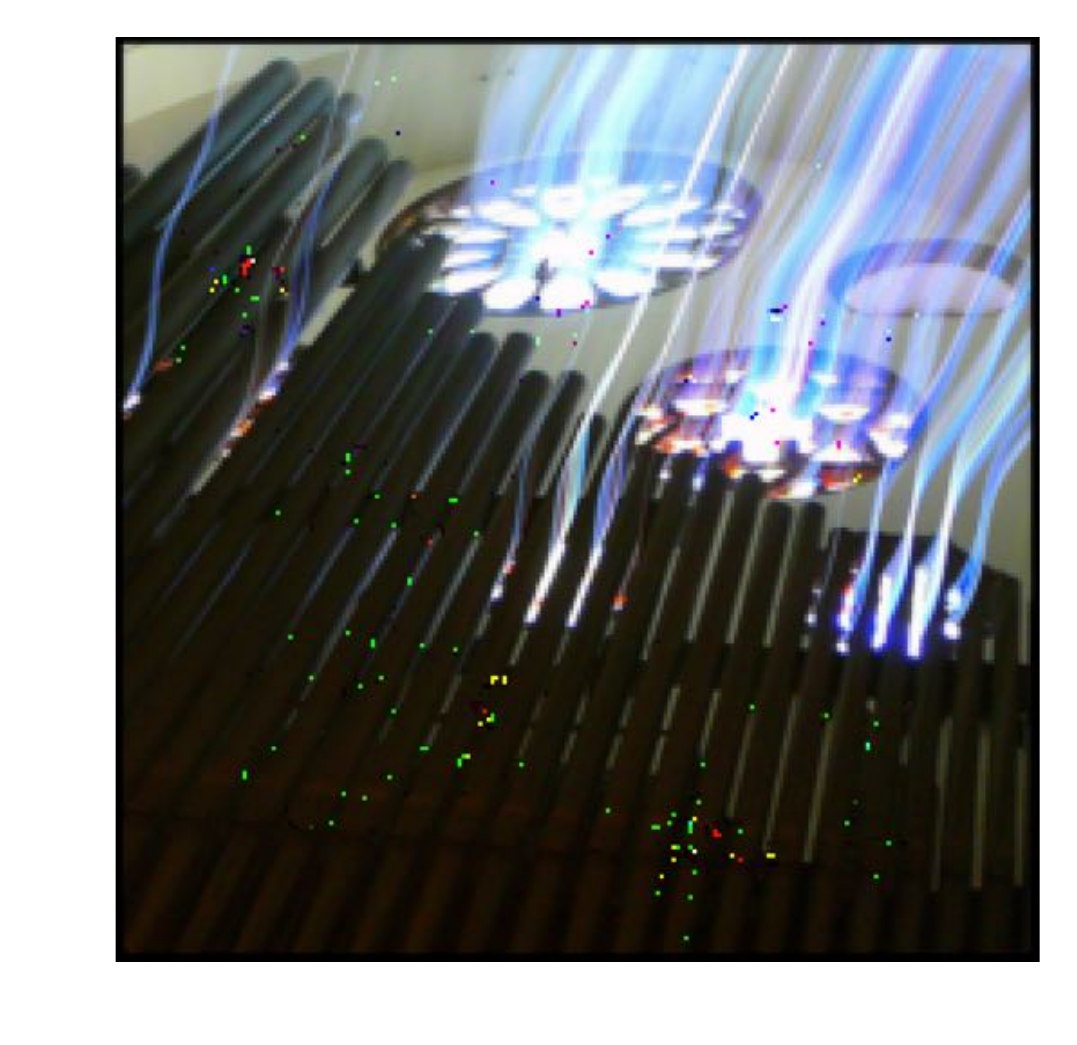}\!
        \caption*{whine bottle ($0.57\%$)}
    \end{subfigure}
    \begin{subfigure}[b]{0.16\linewidth}
        \includegraphics[width=\linewidth]{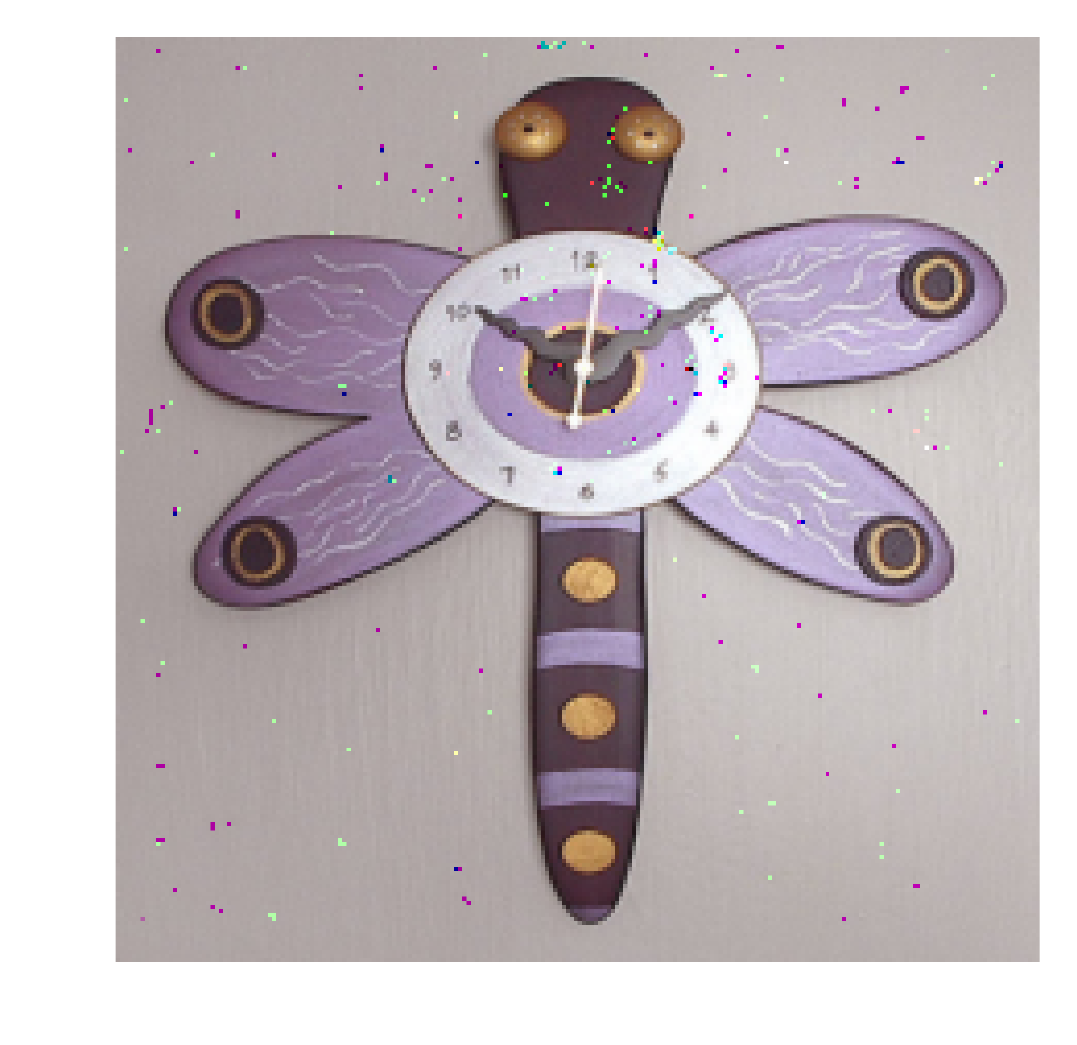}\!
        \caption*{buckle ($0.6019\%$)}
    \end{subfigure}
    \begin{subfigure}[b]{0.16\linewidth}
        \includegraphics[width=\linewidth]{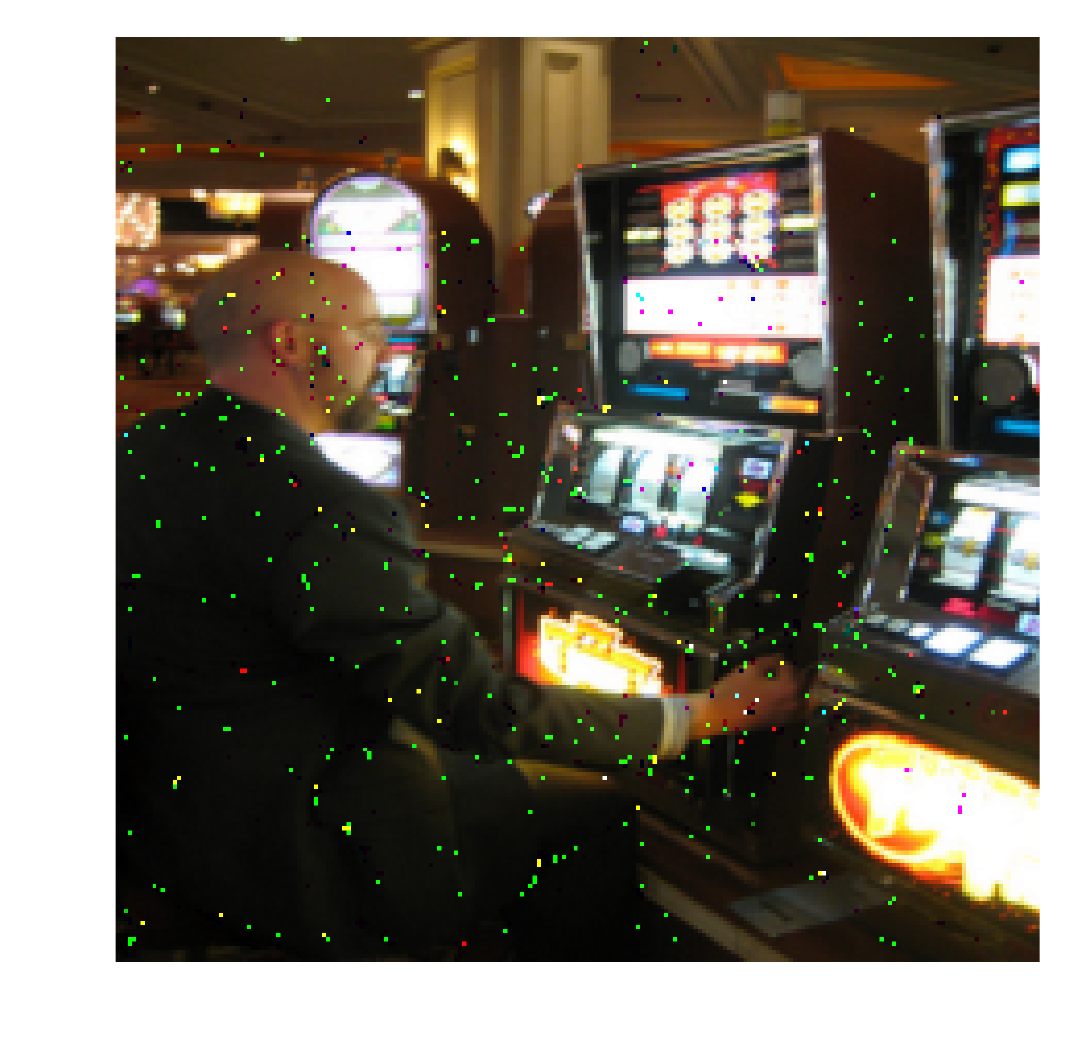}\!
        \caption*{television ($1.8056\%$)}
    \end{subfigure}
    
    \begin{subfigure}[b]{0.16\linewidth}
        \includegraphics[width=\linewidth]{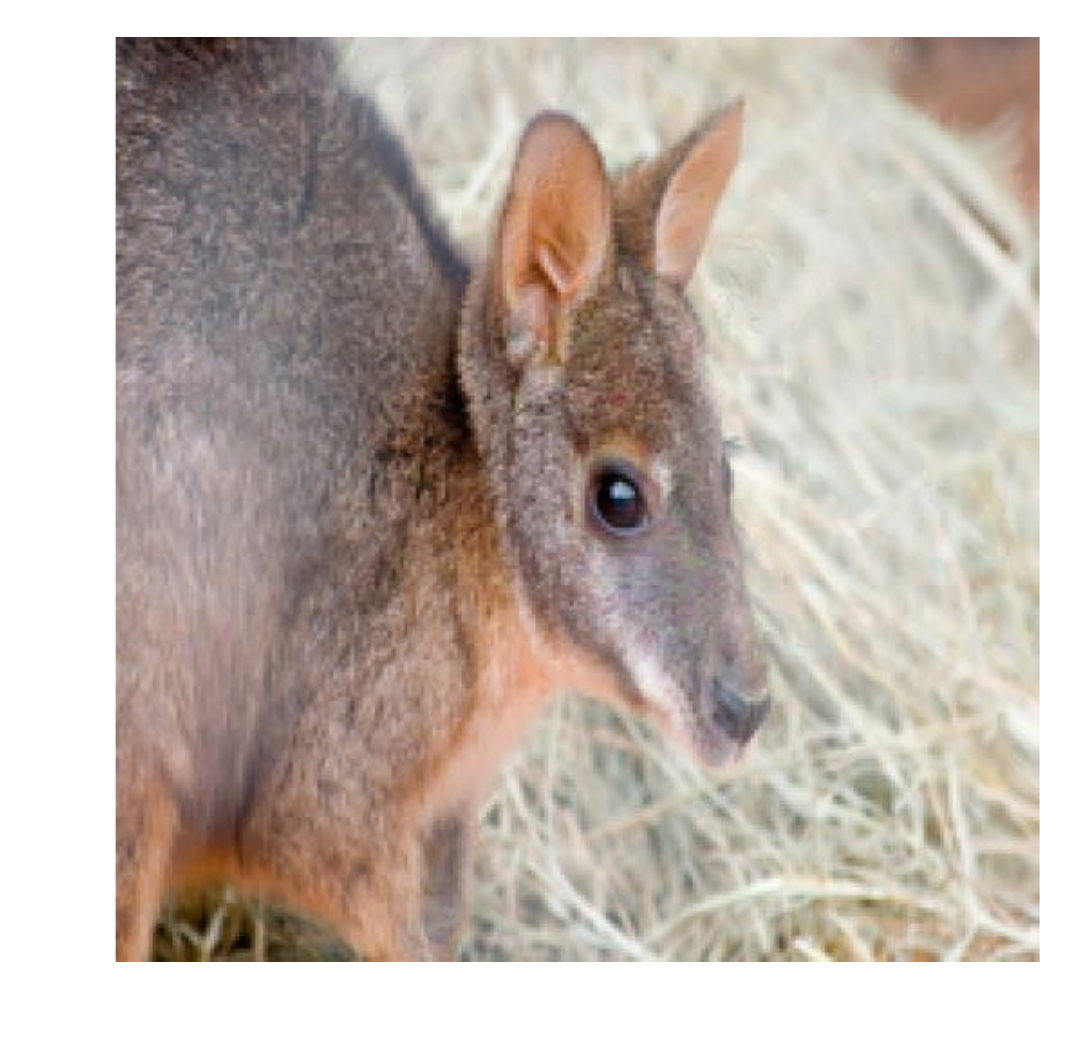}\!
        \caption*{wood rabbit ($6.8399\%$)}
    \end{subfigure}
    \begin{subfigure}[b]{0.16\linewidth}
        \includegraphics[width=\linewidth]{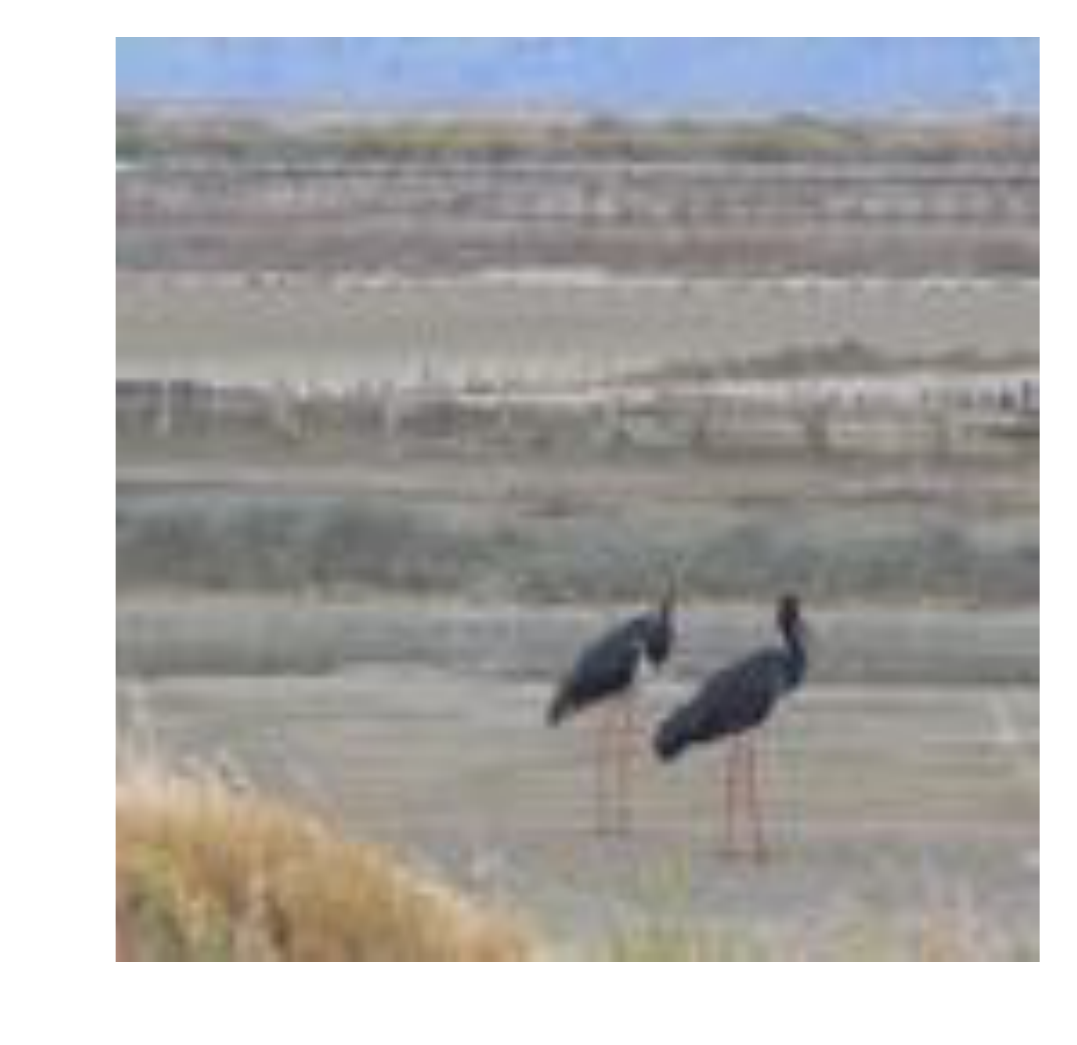}\!
        \caption*{white stork ($2.0548\%$)}
    \end{subfigure}
    \begin{subfigure}[b]{0.16\linewidth}
        \includegraphics[width=\linewidth]{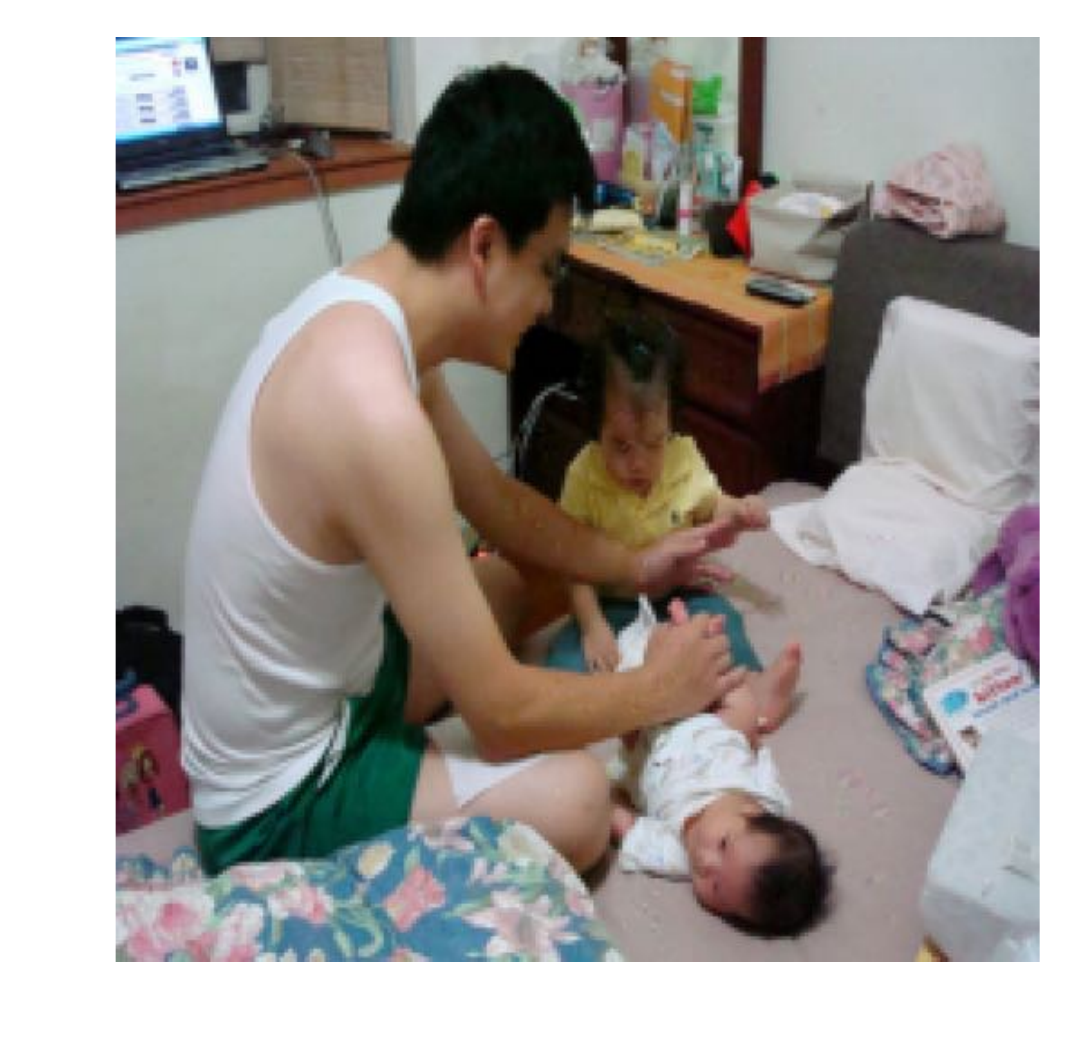}\!
        \caption*{barber shop ($1.5007\%$)}
    \end{subfigure}
    \begin{subfigure}[b]{0.16\linewidth}
        \includegraphics[width=\linewidth]{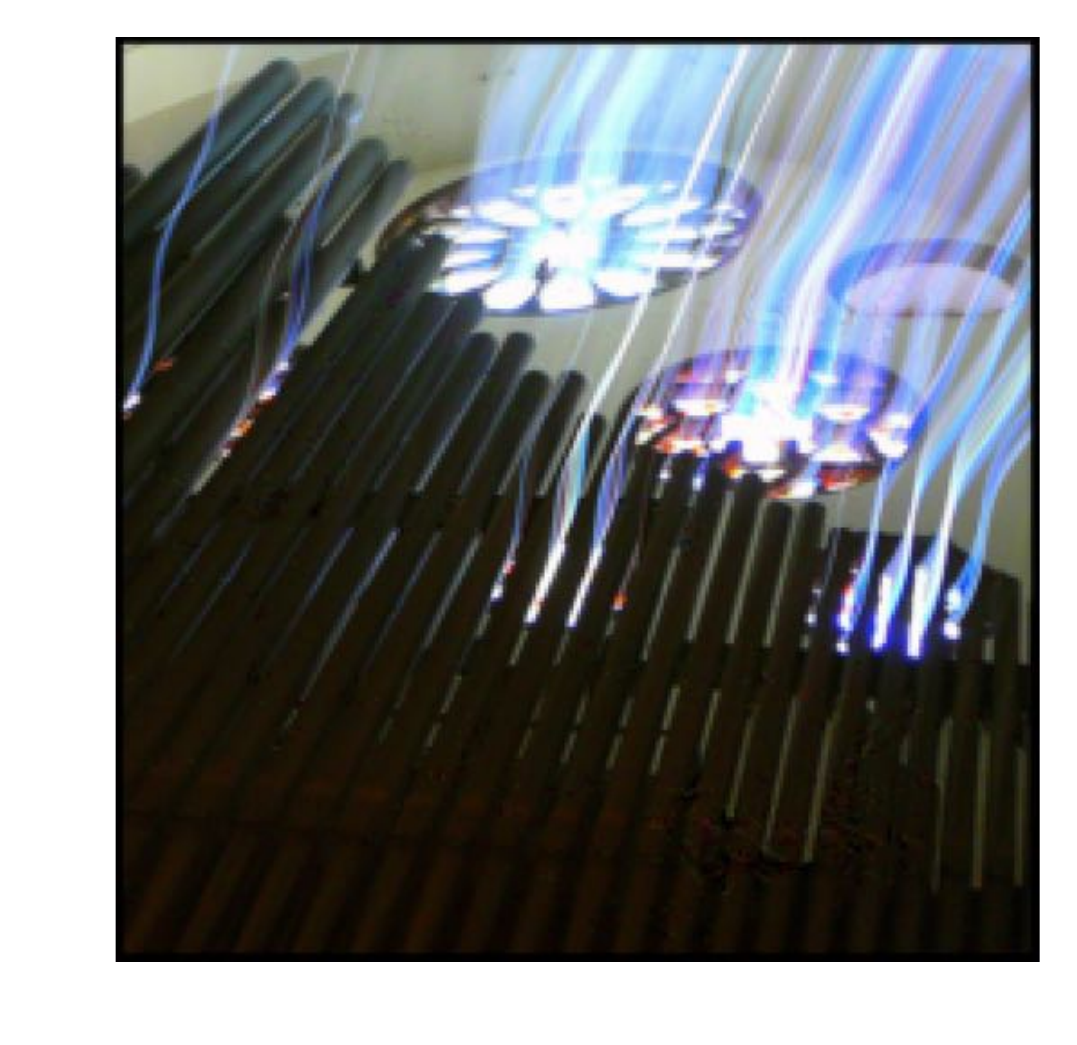}\!
        \caption*{matchstick ($3.6332\%$)}
    \end{subfigure}
    \begin{subfigure}[b]{0.16\linewidth}
        \includegraphics[width=\linewidth]{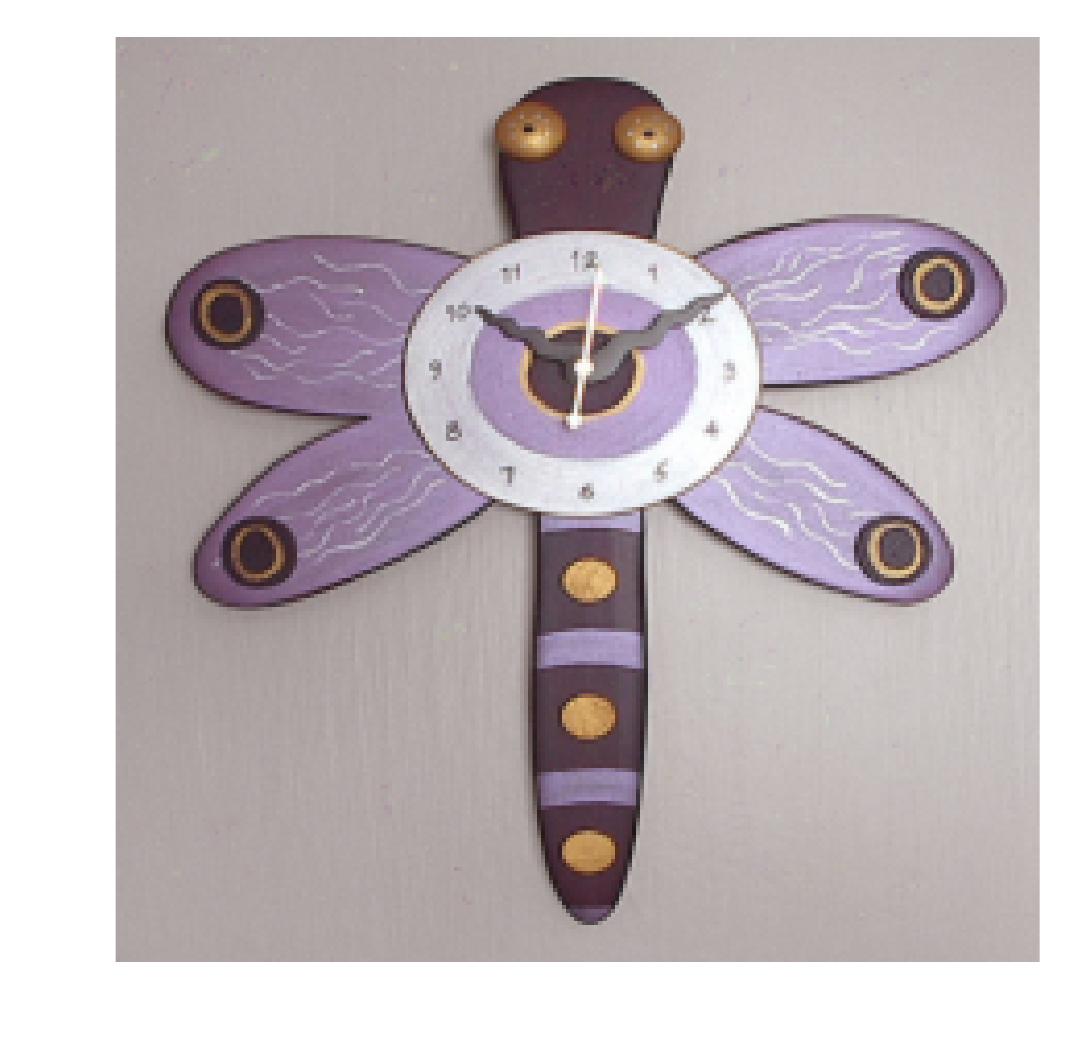}\!
        \caption*{buckle ($0.6537\%$)}
    \end{subfigure}
    \begin{subfigure}[b]{0.16\linewidth}
        \includegraphics[width=\linewidth]{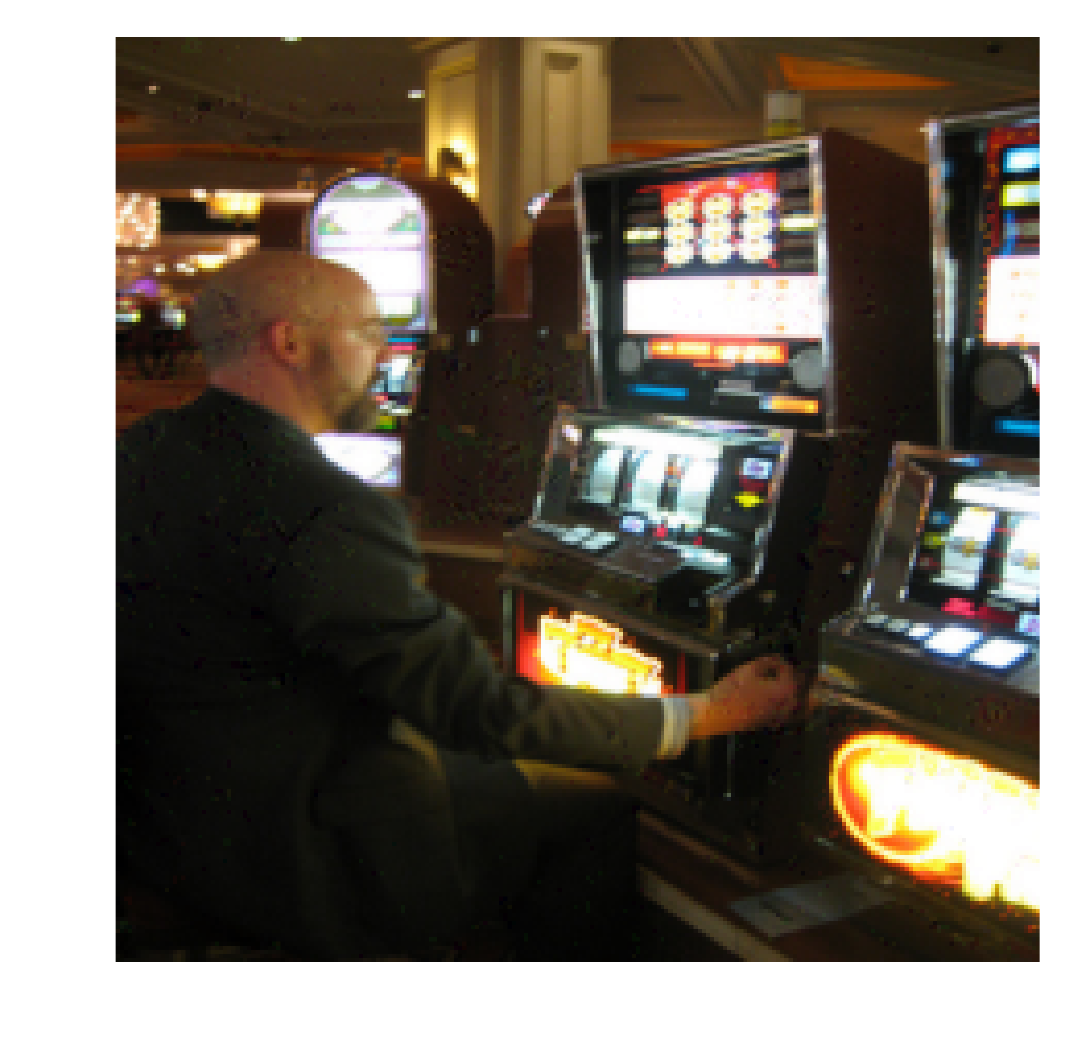}\!
        \caption*{limousine ($4.9645\%$)}
    \end{subfigure}
\end{subfigure}\hfill
\caption{Controlling the perceptibility of the perturbations computed by SparseFool. First row: the original images and the predicted labels below them. Second row: adversarial examples generated by SparseFool, using the whole dynamic range. Third row: adversarial examples generated by SparseFool, constraining the noise $\pm10$ around the image values. For the adversarial examples, the fooling label is shown below the images, and the percentage of perturbed pixels is written inside the parentheses. Note that the fooling label might change, since SparseFool is operating as an untargeted attack.}
\label{fig:delta}
\end{figure}

We now present in Figure~\ref{fig:delta} some adversarial examples on the ImageNet dataset, when we control the perceptibility of the perturbations computed by SparseFool. This can be done by properly constraining the values of the perturbed image to lie $\pm\delta$ around the values of the original image.

\section{The control parameter $\lambda$}
\begin{figure}[hb]
\centering
    \begin{subfigure}[b]{0.33\linewidth}
    \centering
    \includegraphics[width=\linewidth]{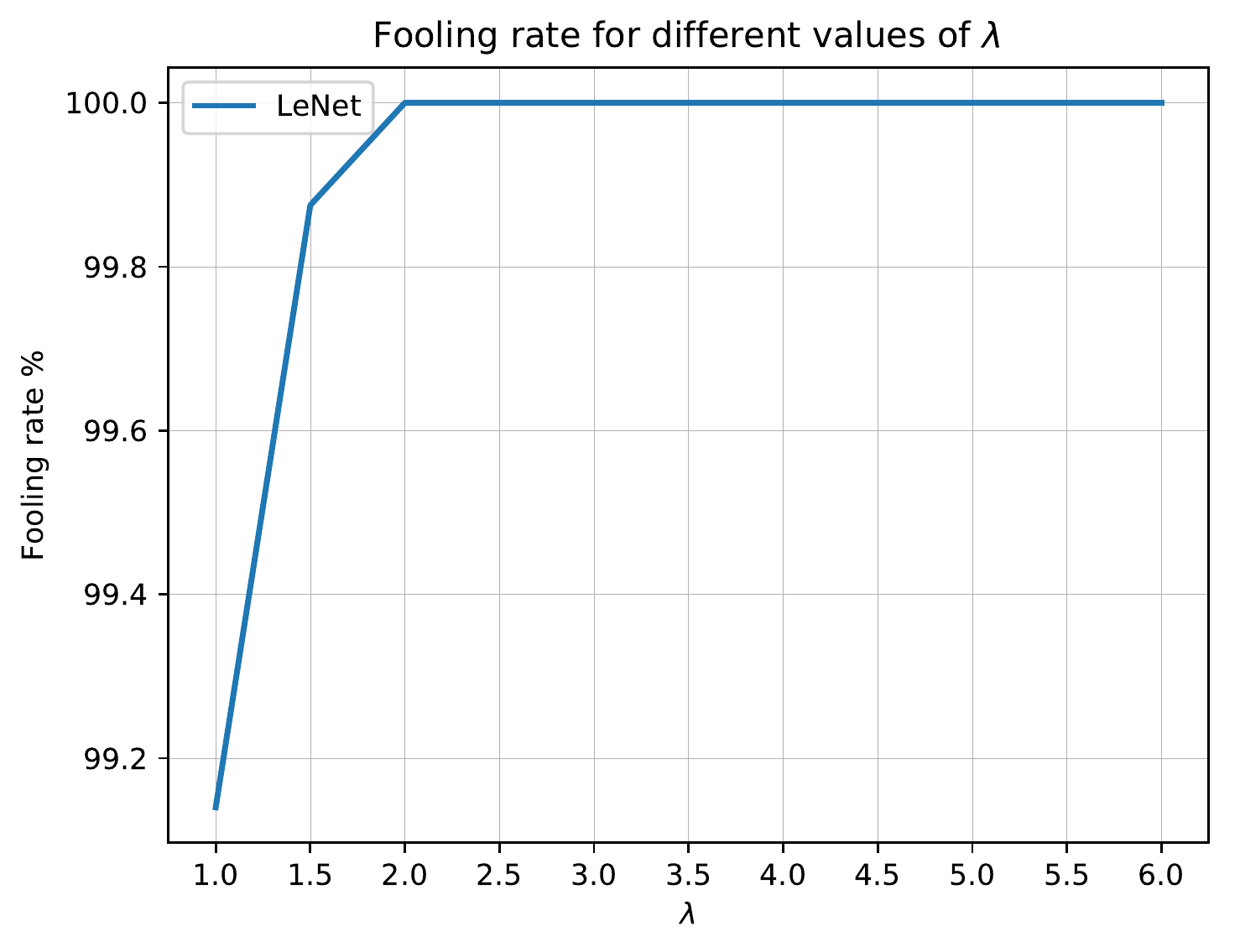}\!
    \end{subfigure}\hfill
    \begin{subfigure}[b]{0.33\linewidth}
    \centering
    \includegraphics[width=\linewidth]{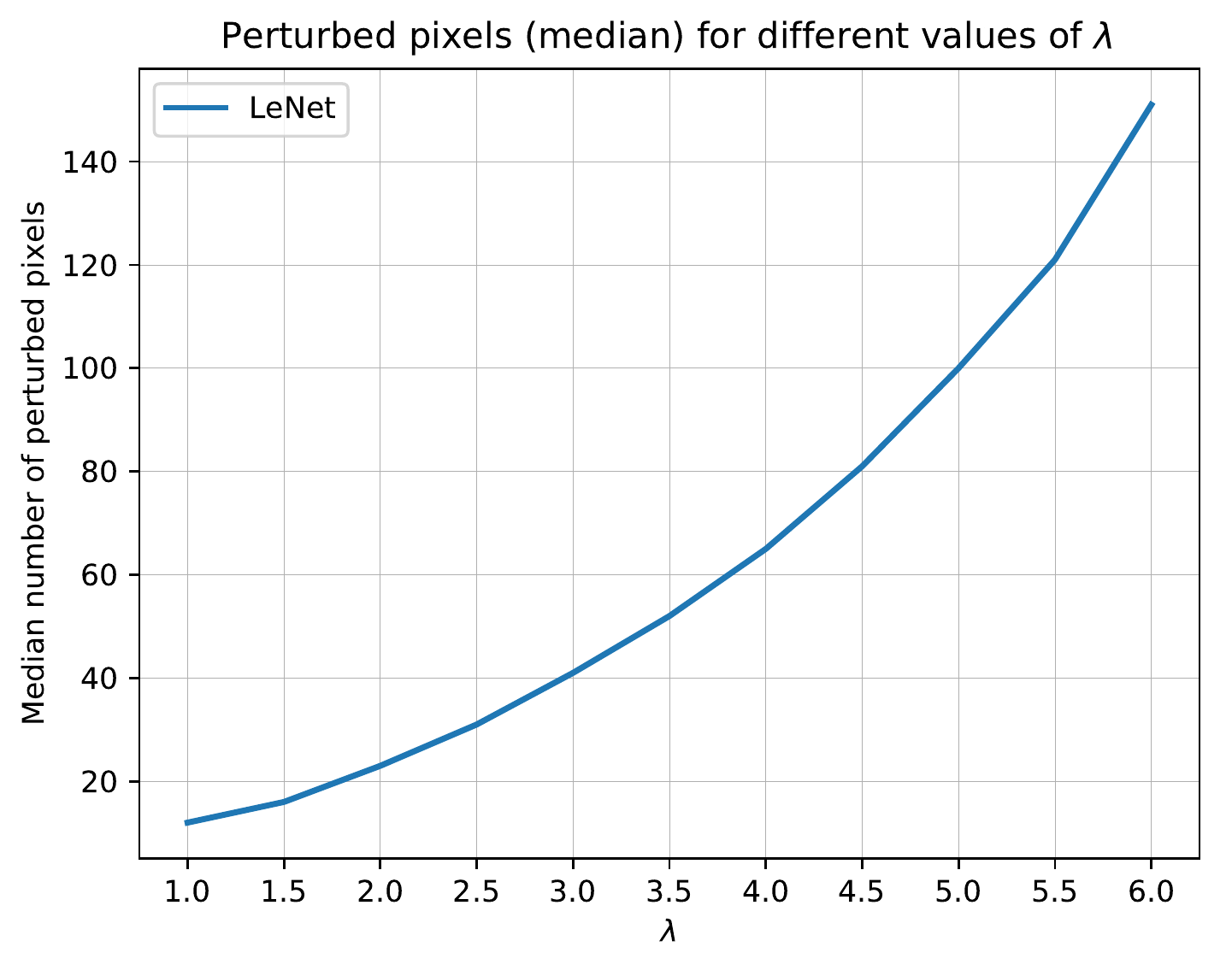}\!
    \end{subfigure}\hfill
    \begin{subfigure}[b]{0.33\linewidth}
    \centering
    \includegraphics[width=\linewidth]{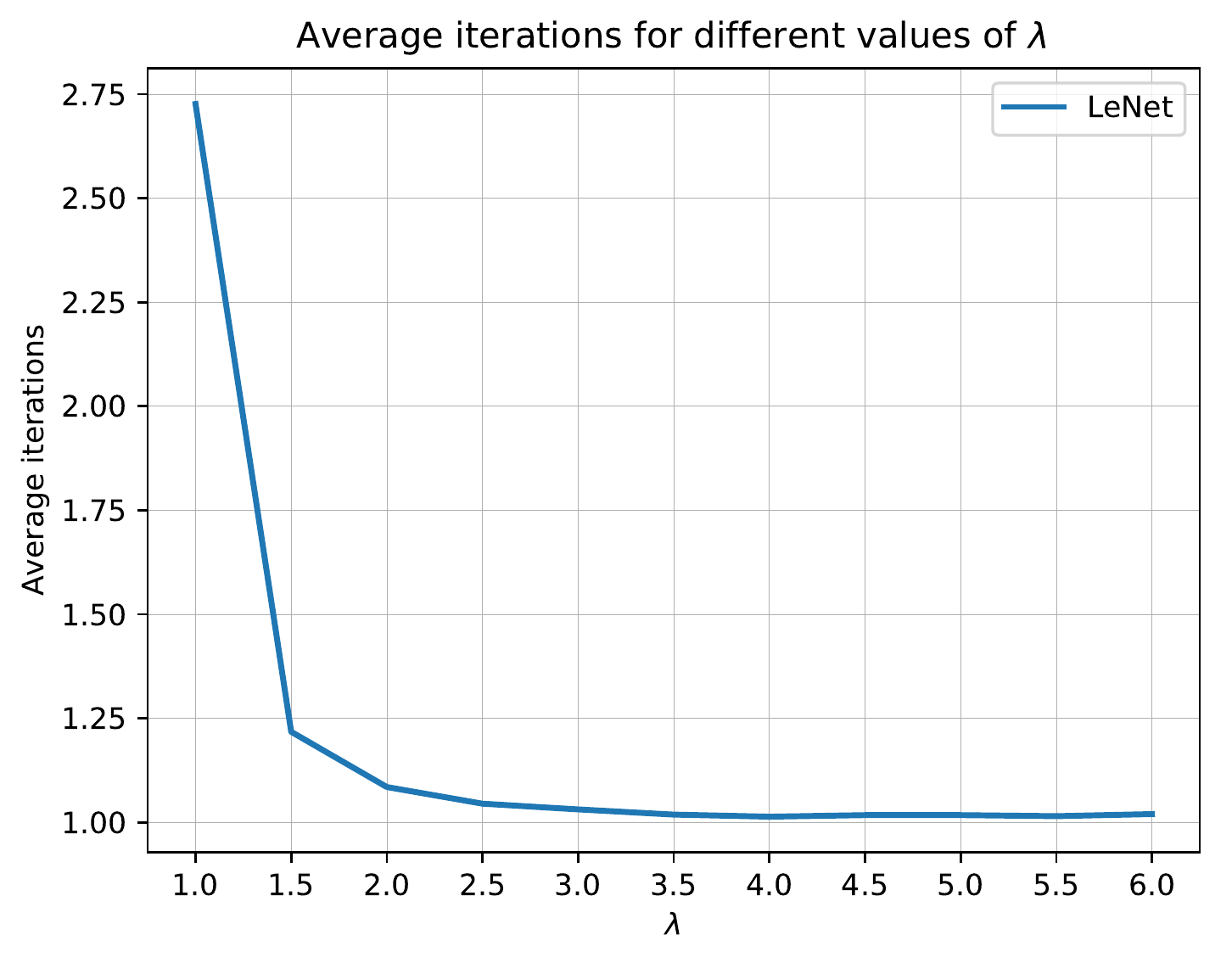}\!
    \end{subfigure}\hfill
\caption{The effect of $\lambda$ on SparseFool for a LeNet model trained on the MNIST dataset.}
\label{fig:lambda_mnist}
\end{figure}

\begin{figure}[hb]
\centering
    \begin{subfigure}[b]{0.33\linewidth}
    \centering
    \includegraphics[width=\linewidth]{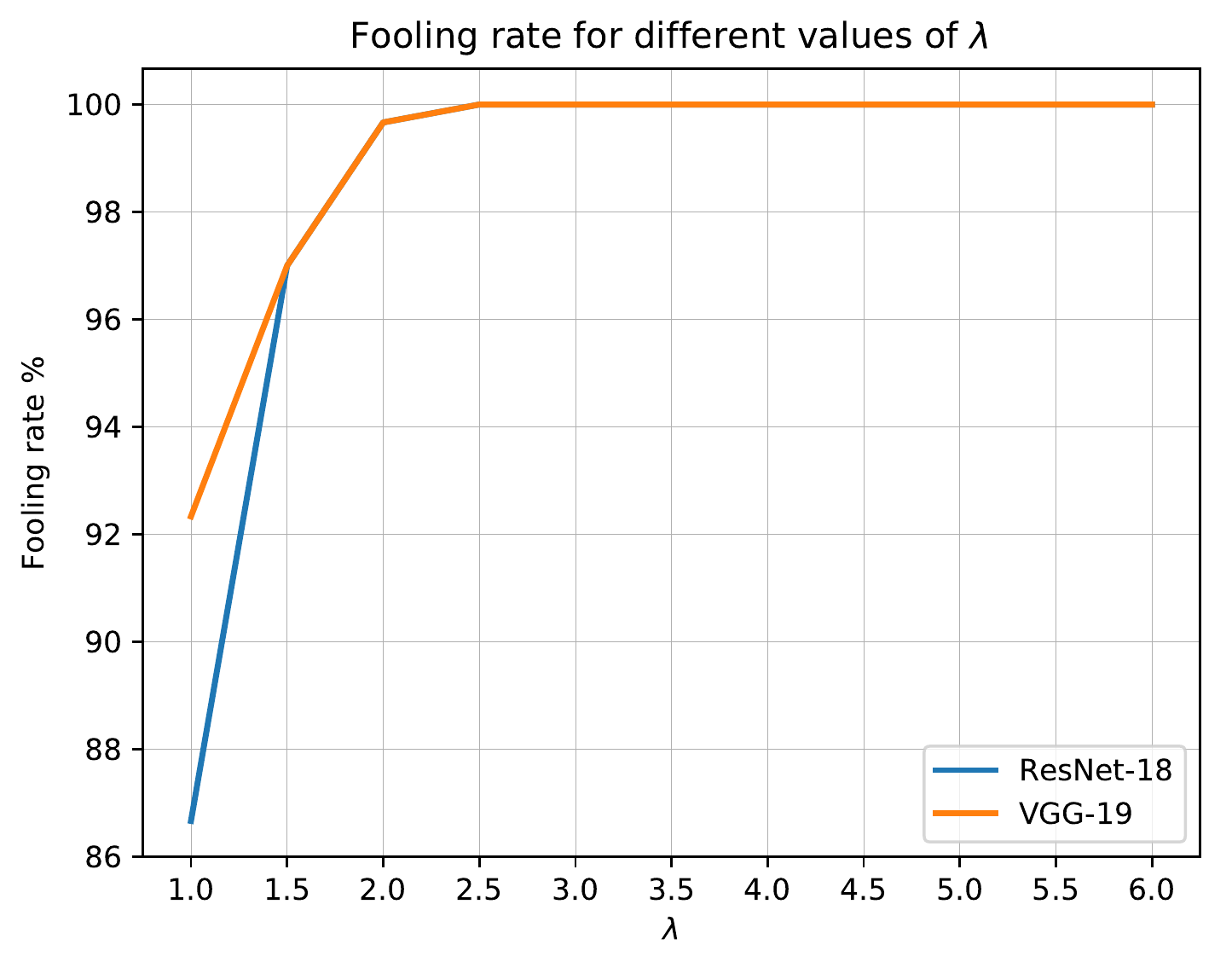}\!
    \end{subfigure}\hfill
    \begin{subfigure}[b]{0.33\linewidth}
    \centering
    \includegraphics[width=\linewidth]{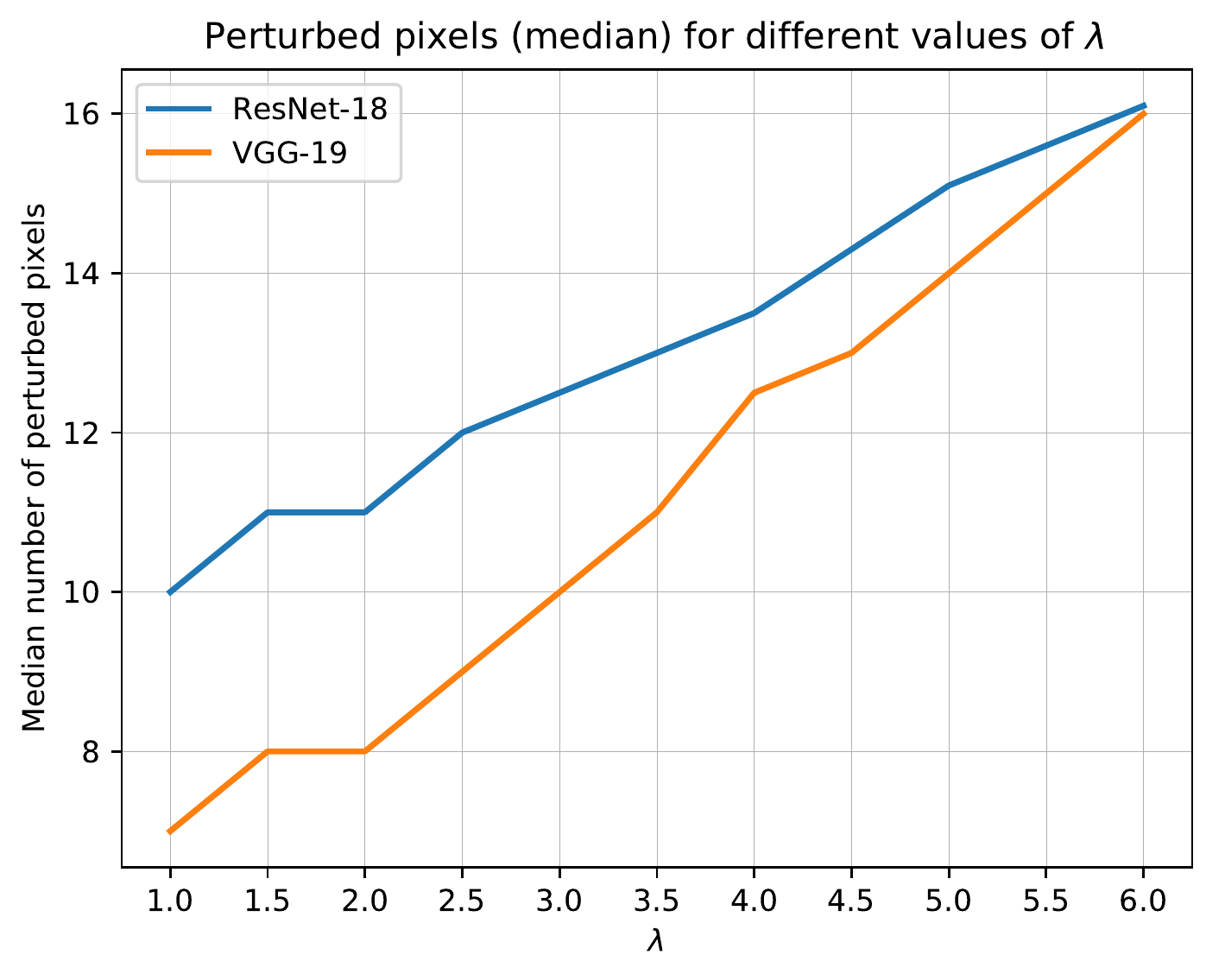}\!
    \end{subfigure}\hfill
    \begin{subfigure}[b]{0.33\linewidth}
    \centering
    \includegraphics[width=\linewidth]{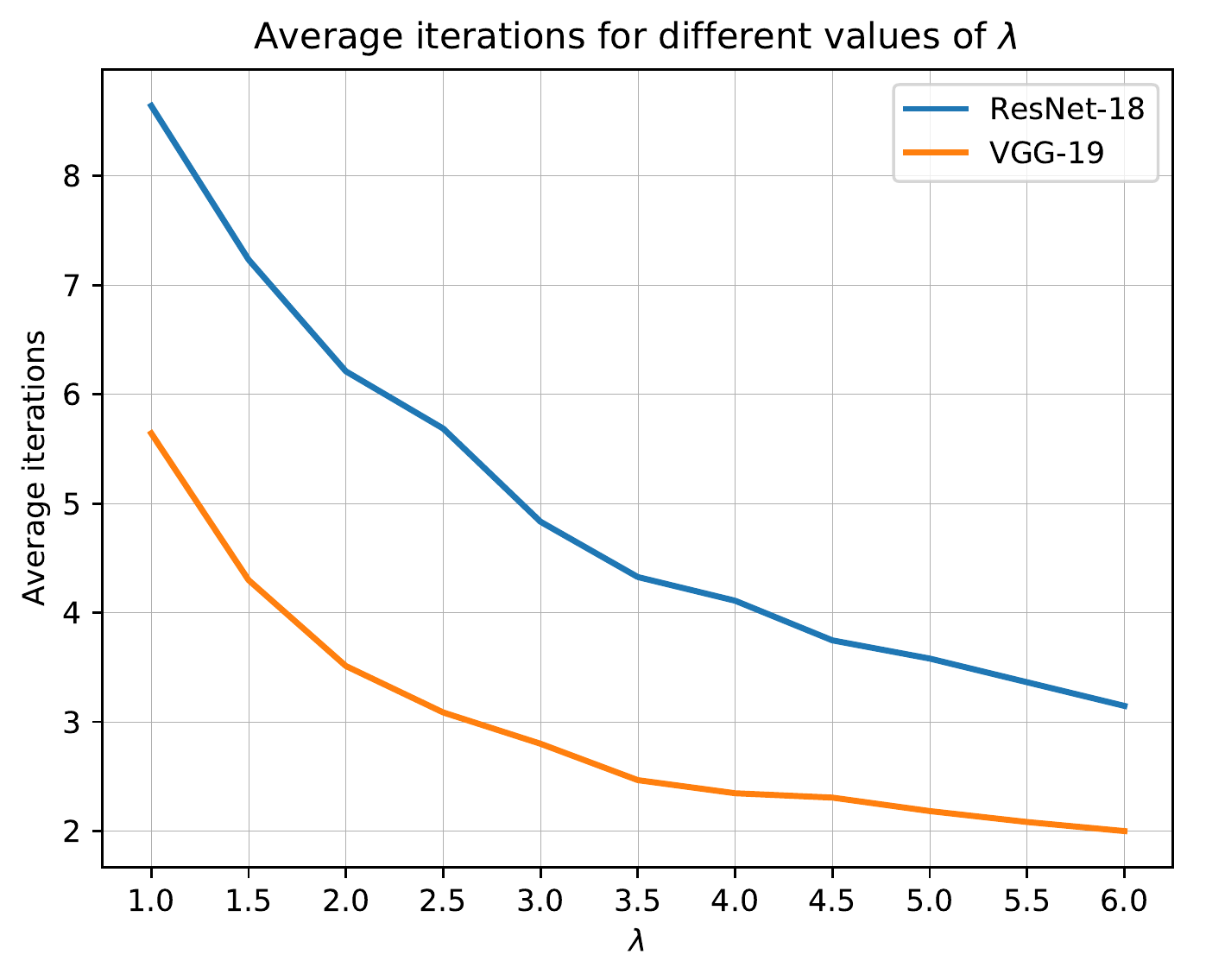}\!
    \end{subfigure}\hfill
\caption{The effect of $\lambda$ on SparseFool for different networks trained on the CIFAR-10 dataset.}
\label{fig:lambda_cifar}
\end{figure}

\begin{figure}[hb]
\centering
    \begin{subfigure}[b]{0.33\linewidth}
    \centering
    \includegraphics[width=\linewidth]{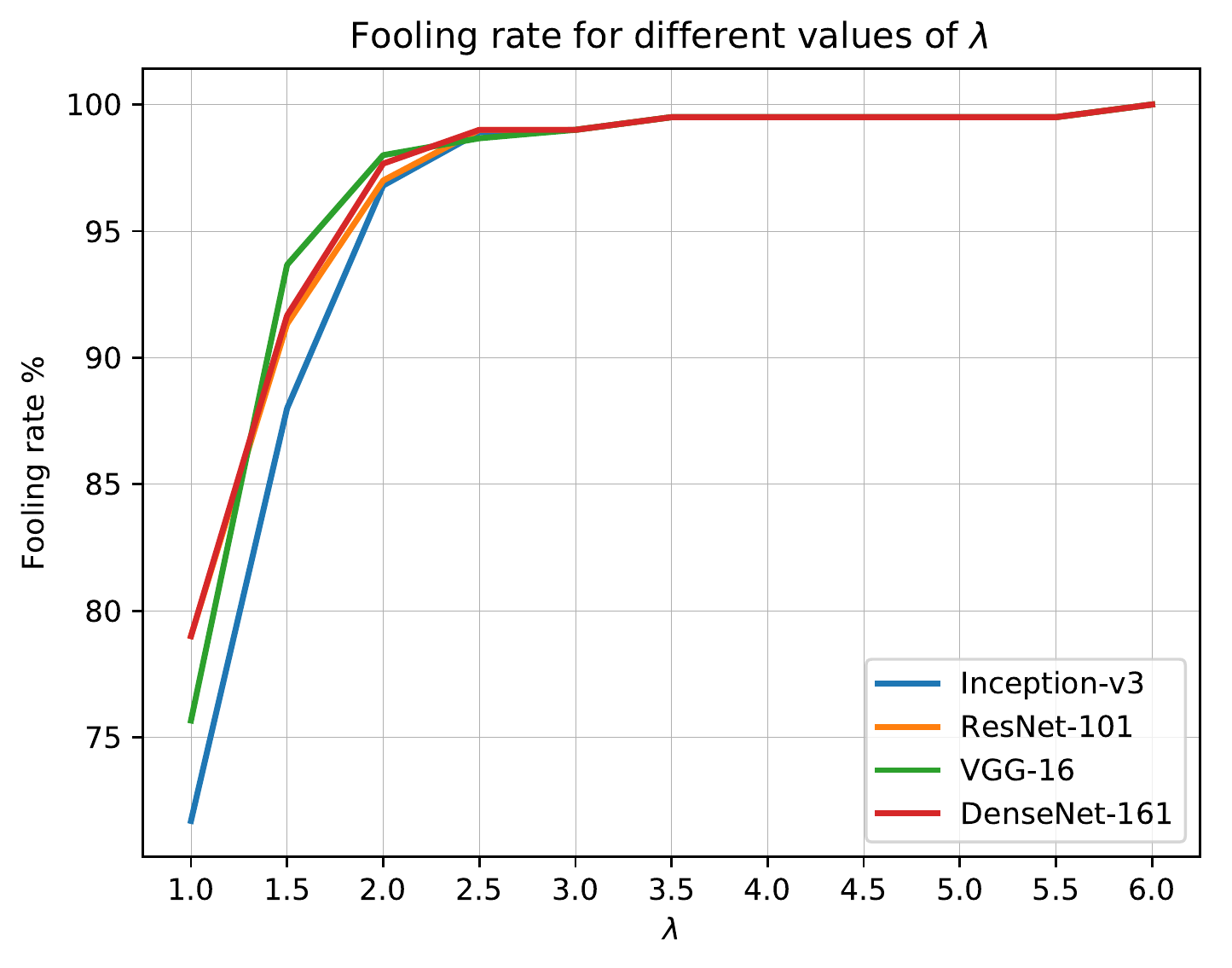}\!
    \end{subfigure}\hfill
    \begin{subfigure}[b]{0.33\linewidth}
    \centering
    \includegraphics[width=\linewidth]{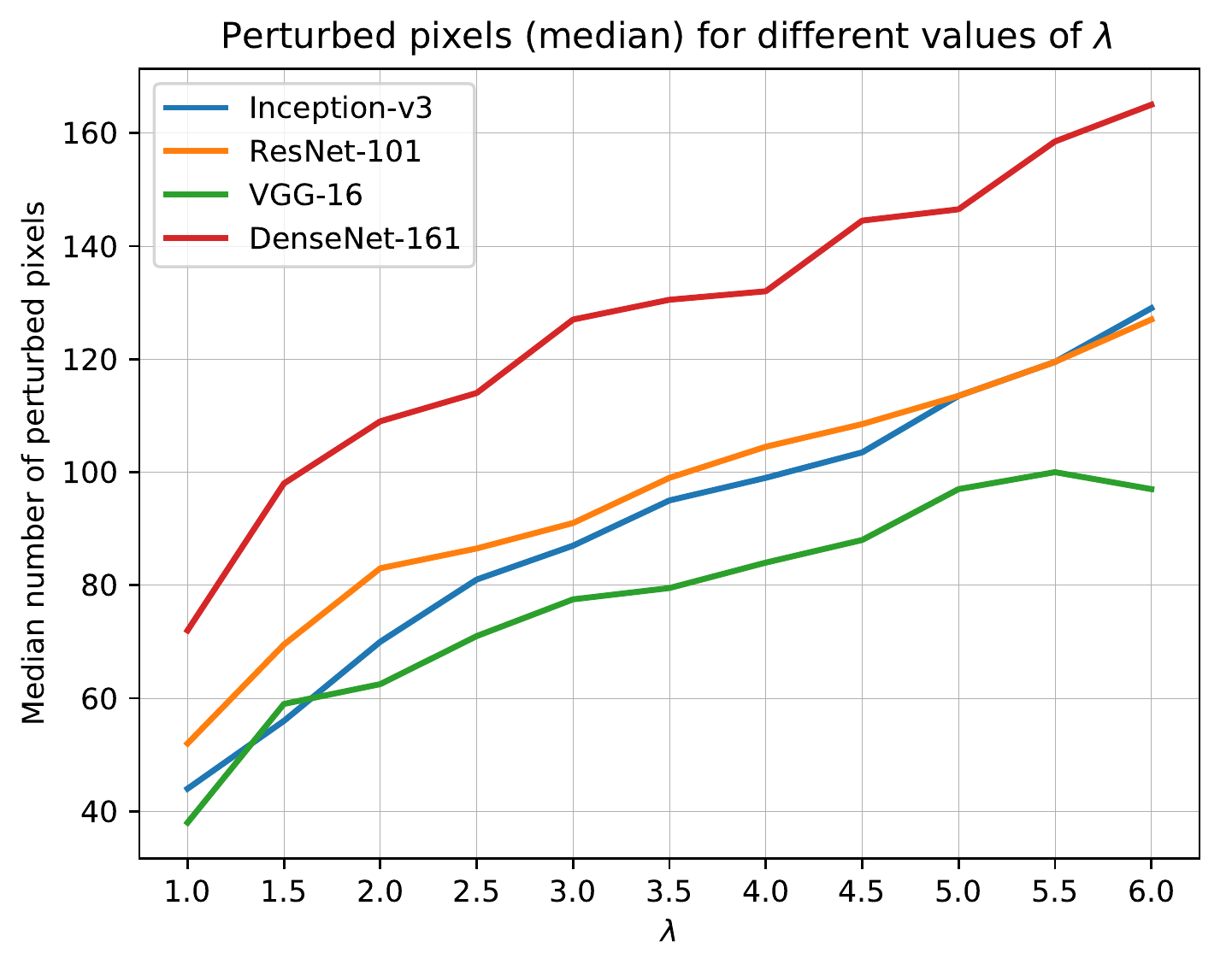}\!
    \end{subfigure}\hfill
    \begin{subfigure}[b]{0.33\linewidth}
    \centering
    \includegraphics[width=\linewidth]{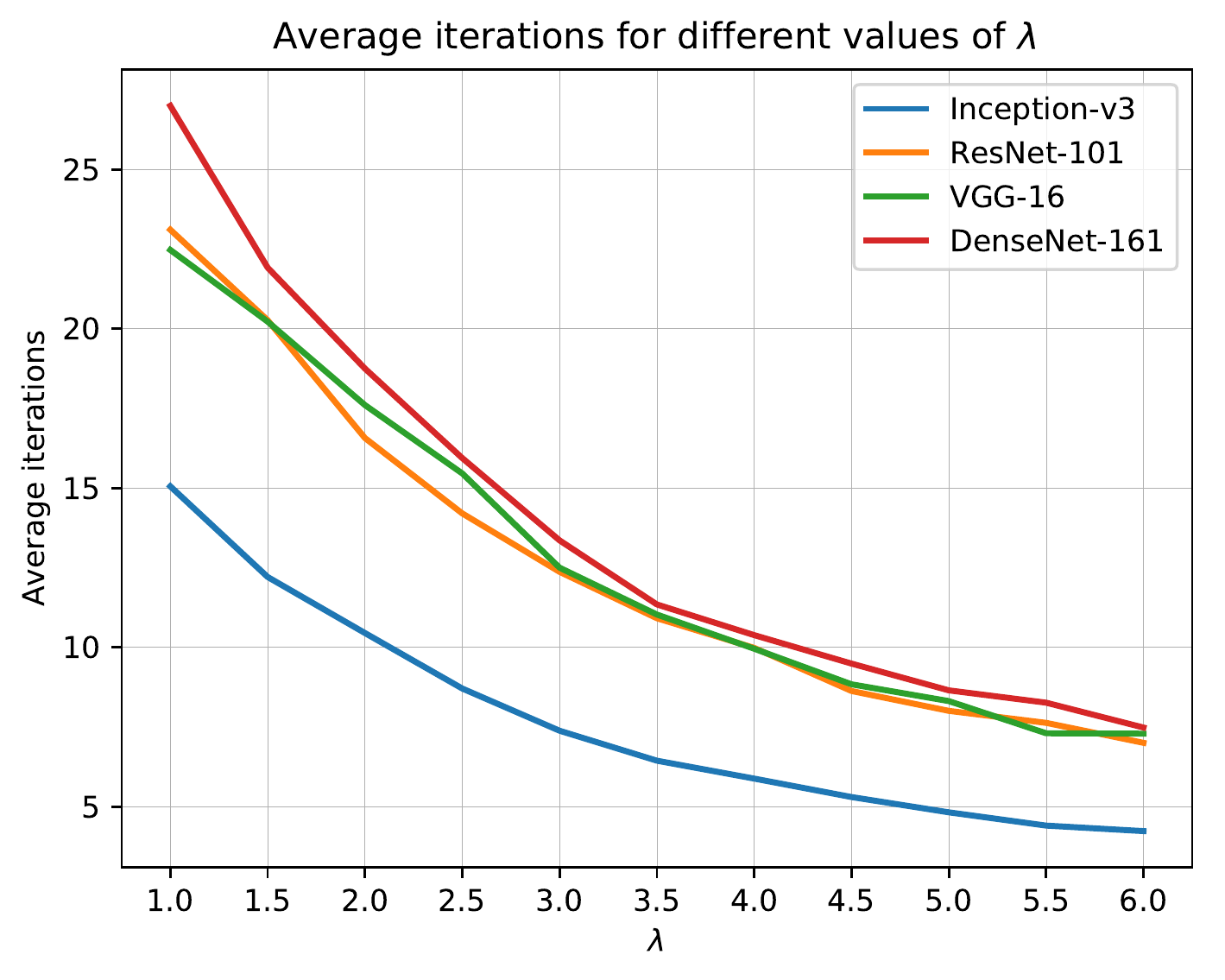}\!
    \end{subfigure}\hfill
\caption{The effect of $\lambda$ on SparseFool for different networks trained on the ImageNet dataset.}
\label{fig:lambda_imagenet}
\end{figure}

In this section we provide the effect of the control parameter $\lambda$ on (a) the fooling rate, (b) the sparsity of the perturbations, and (c) the convergence of SparseFool. The effect of different values of $\lambda$ for different networks on the MNIST, CIFAR-10, and ImageNet datasets is shown in Figures~[\ref{fig:lambda_mnist} -- \ref{fig:lambda_imagenet}] respectively.

\newpage
\section{Adversarial examples compared to related methods}
\begin{figure}[hb]
\centering
    \begin{subfigure}[b]{0.1\linewidth}
    \includegraphics[width=\linewidth]{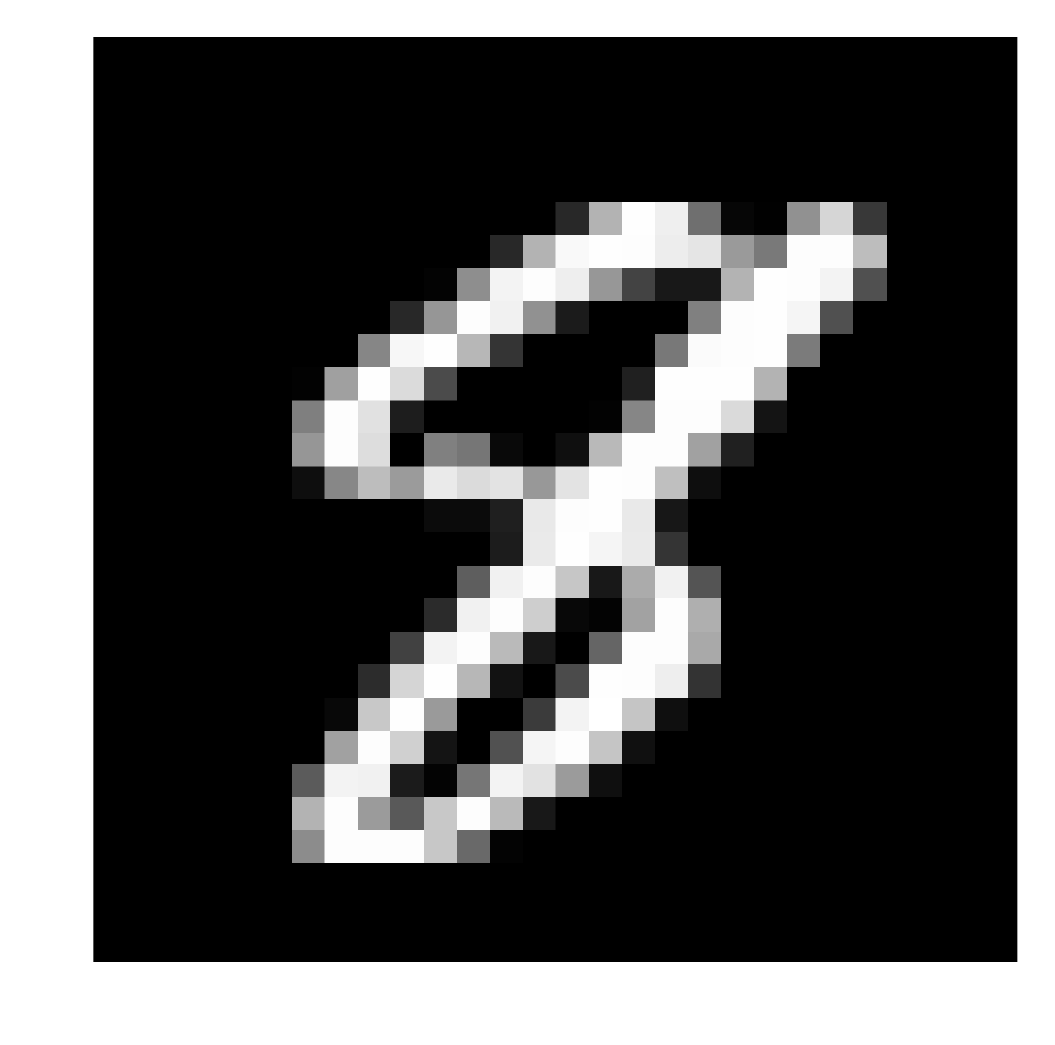}\!
    \caption*{$3$ (3)}
    \end{subfigure}
    \begin{subfigure}[b]{0.1\linewidth}
    \includegraphics[width=\linewidth]{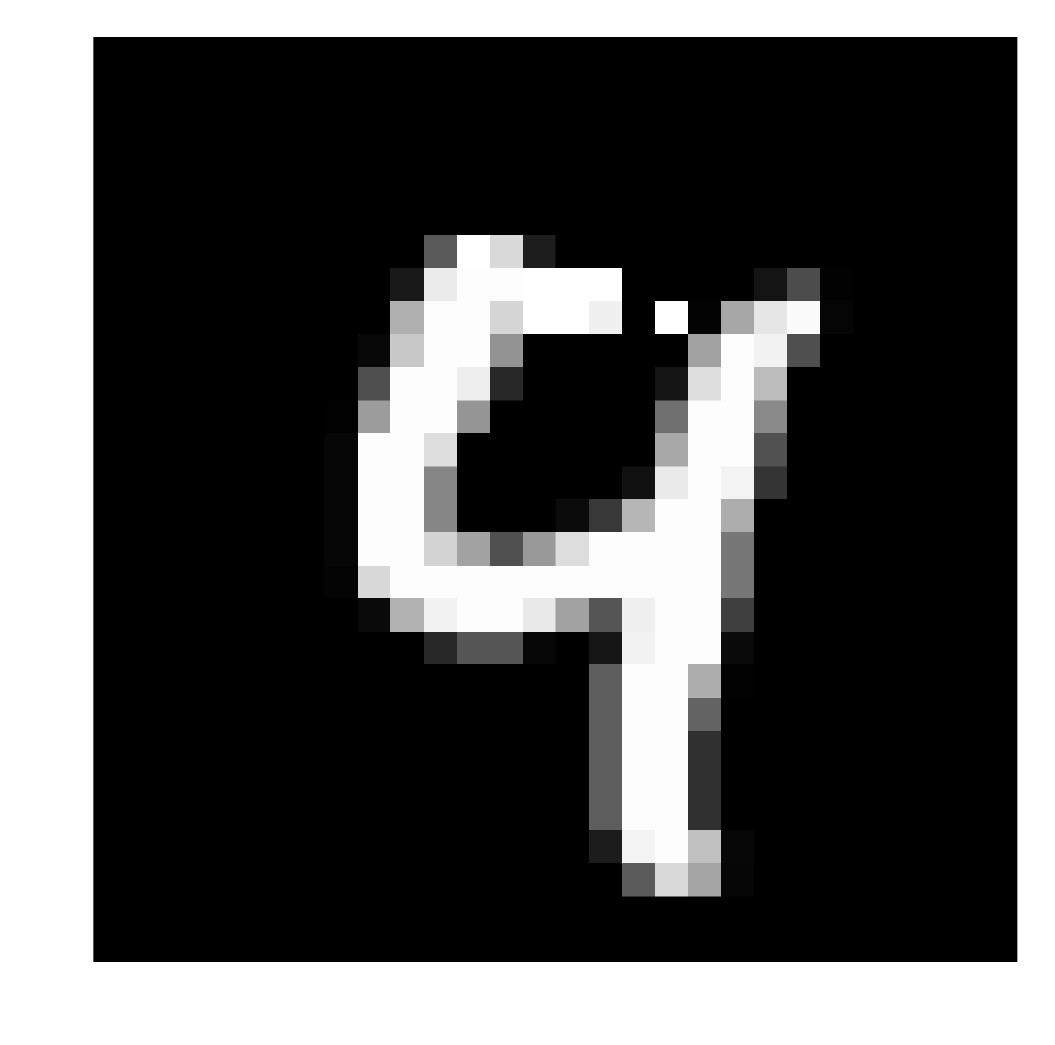}\!
    \caption*{$9$ (7)}
    \end{subfigure}
    \begin{subfigure}[b]{0.1\linewidth}
    \includegraphics[width=\linewidth]{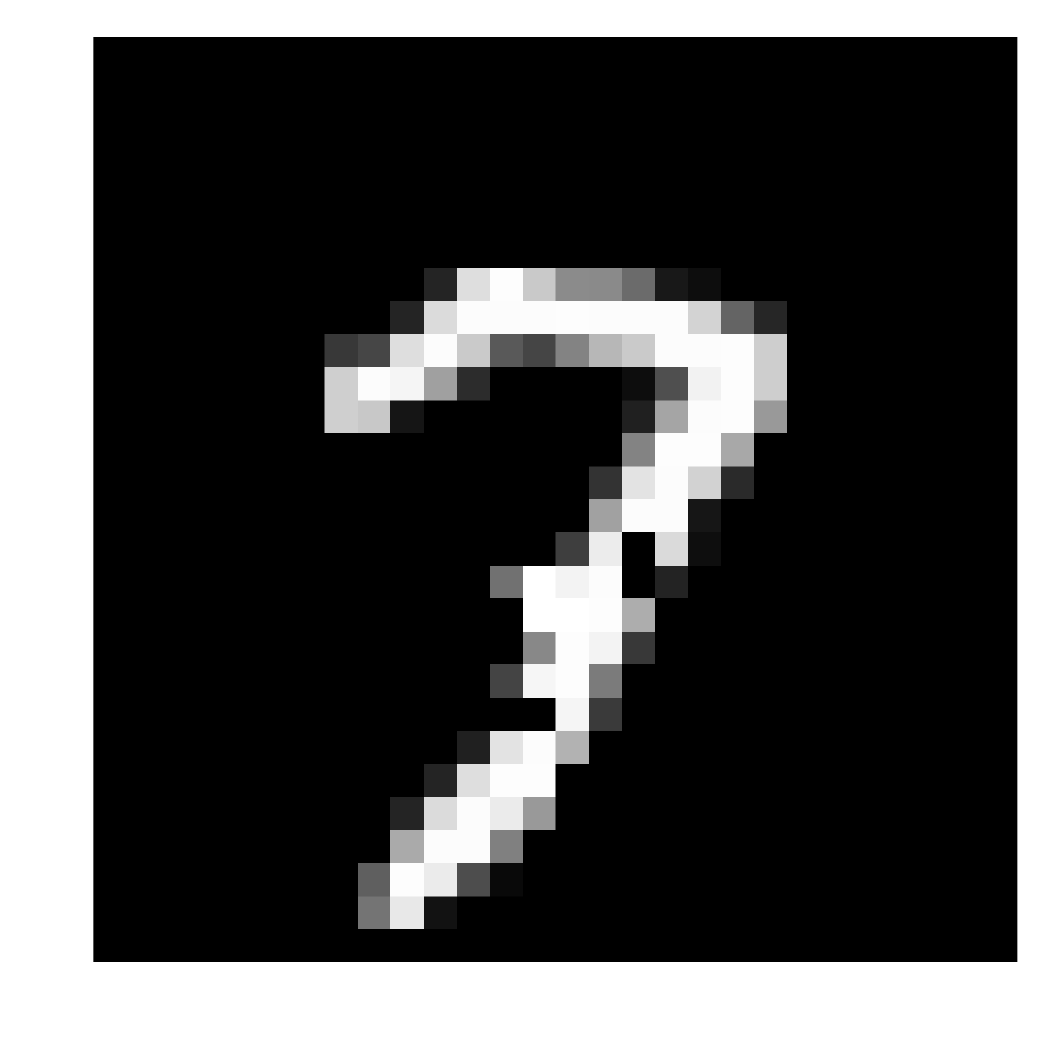}\!
    \caption*{$3$ (7)}
    \end{subfigure}
    \begin{subfigure}[b]{0.1\linewidth}
    \includegraphics[width=\linewidth]{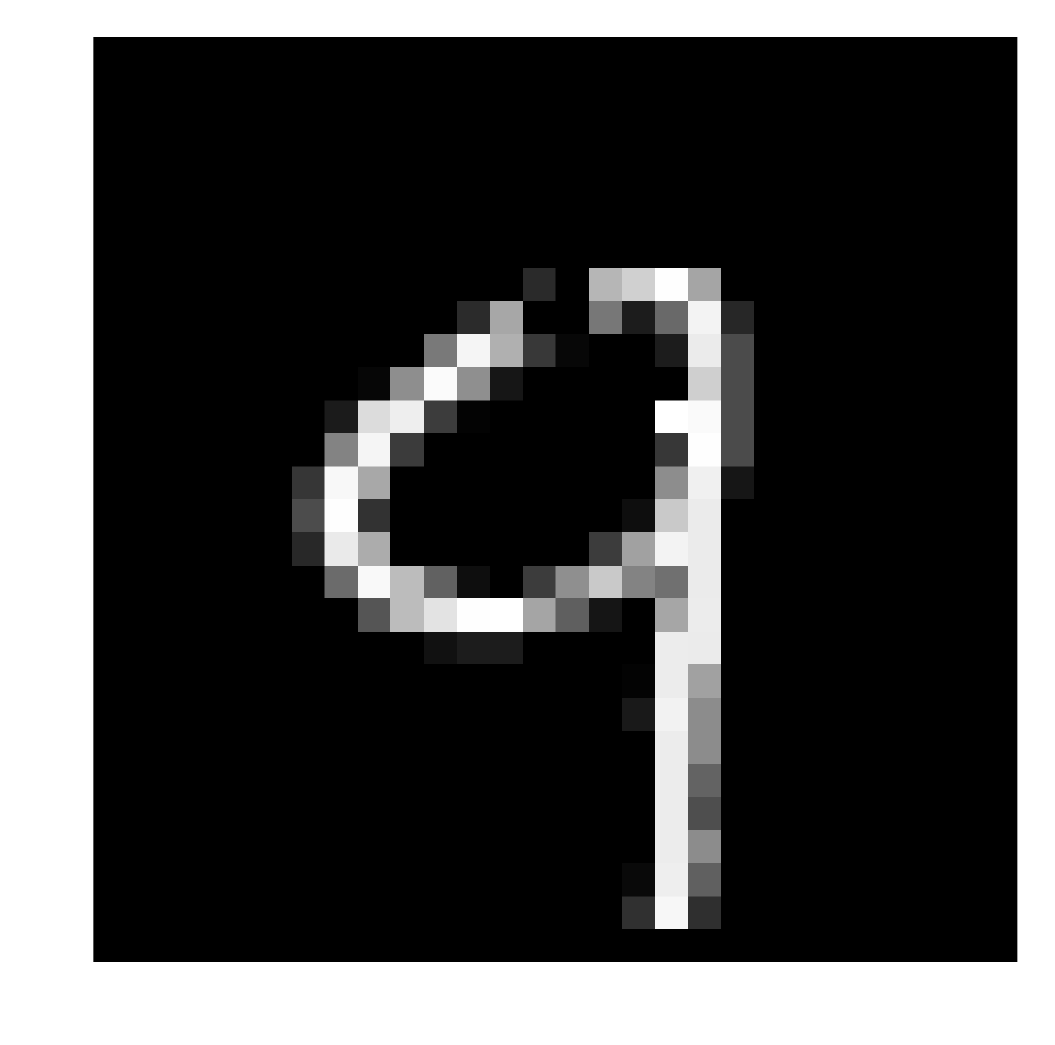}\!
    \caption*{$4$ (7)}
    \end{subfigure}
    \begin{subfigure}[b]{0.1\linewidth}
    \includegraphics[width=\linewidth]{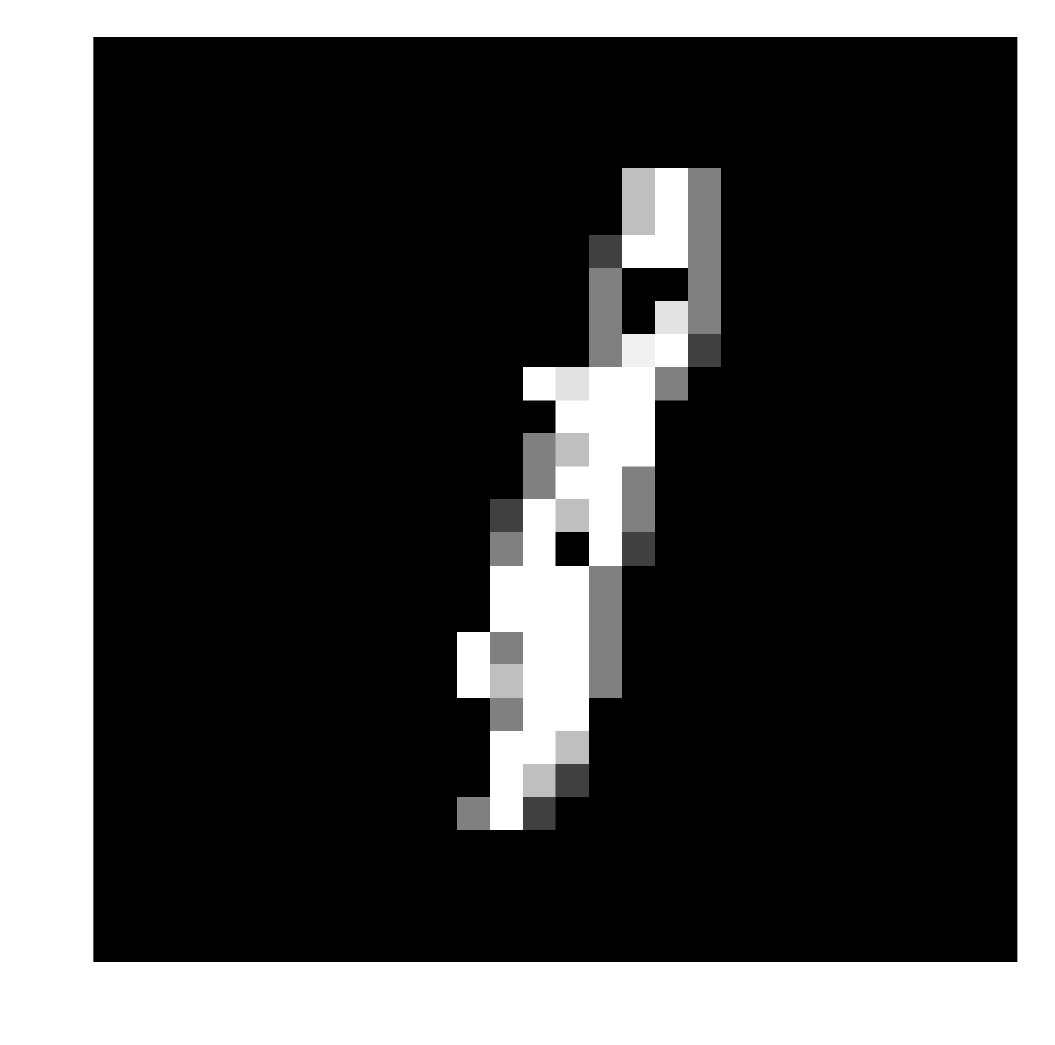}\!
    \caption*{$8$ (8)}
    \end{subfigure}
    \begin{subfigure}[b]{0.1\linewidth}
    \includegraphics[width=\linewidth]{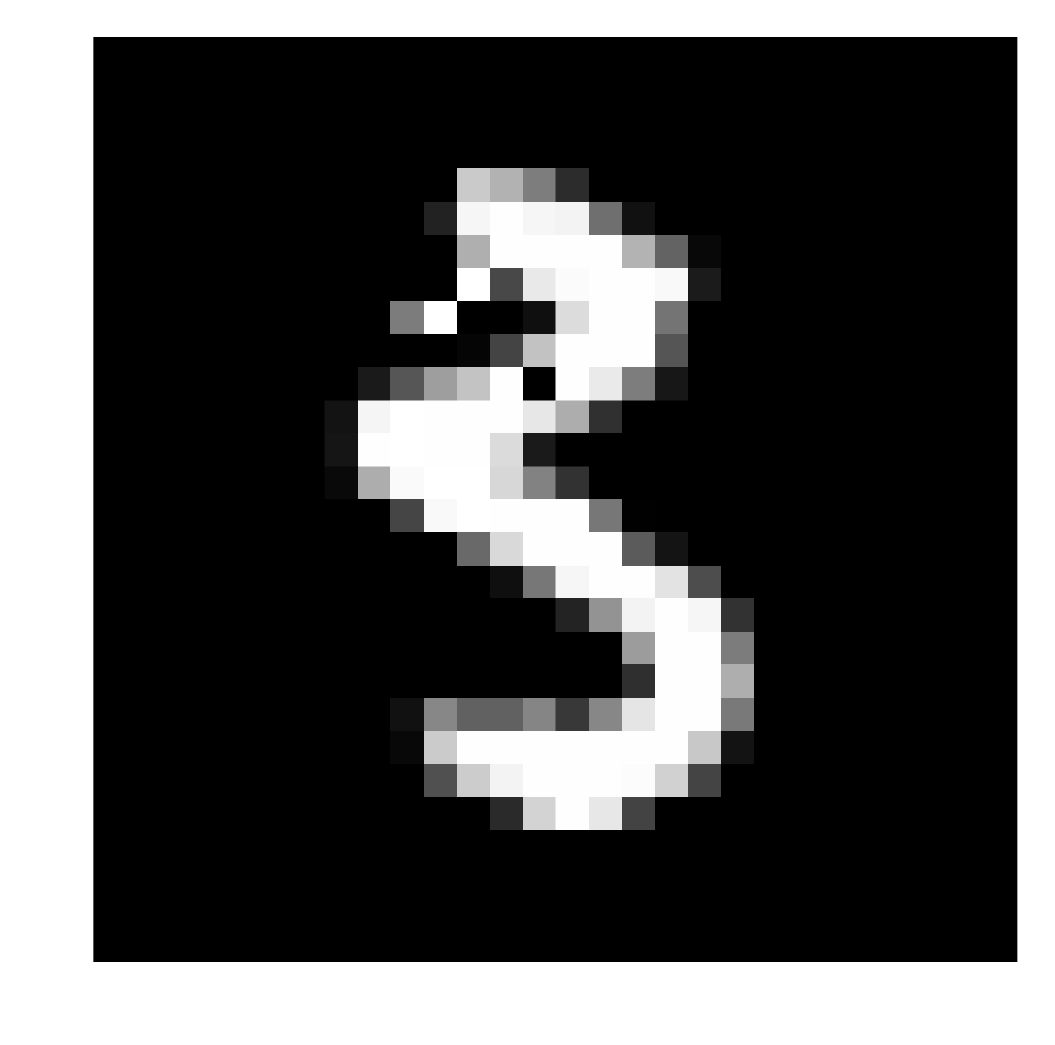}\!
    \caption*{$5$ (9)}
    \end{subfigure}
    \begin{subfigure}[b]{0.1\linewidth}
    \includegraphics[width=\linewidth]{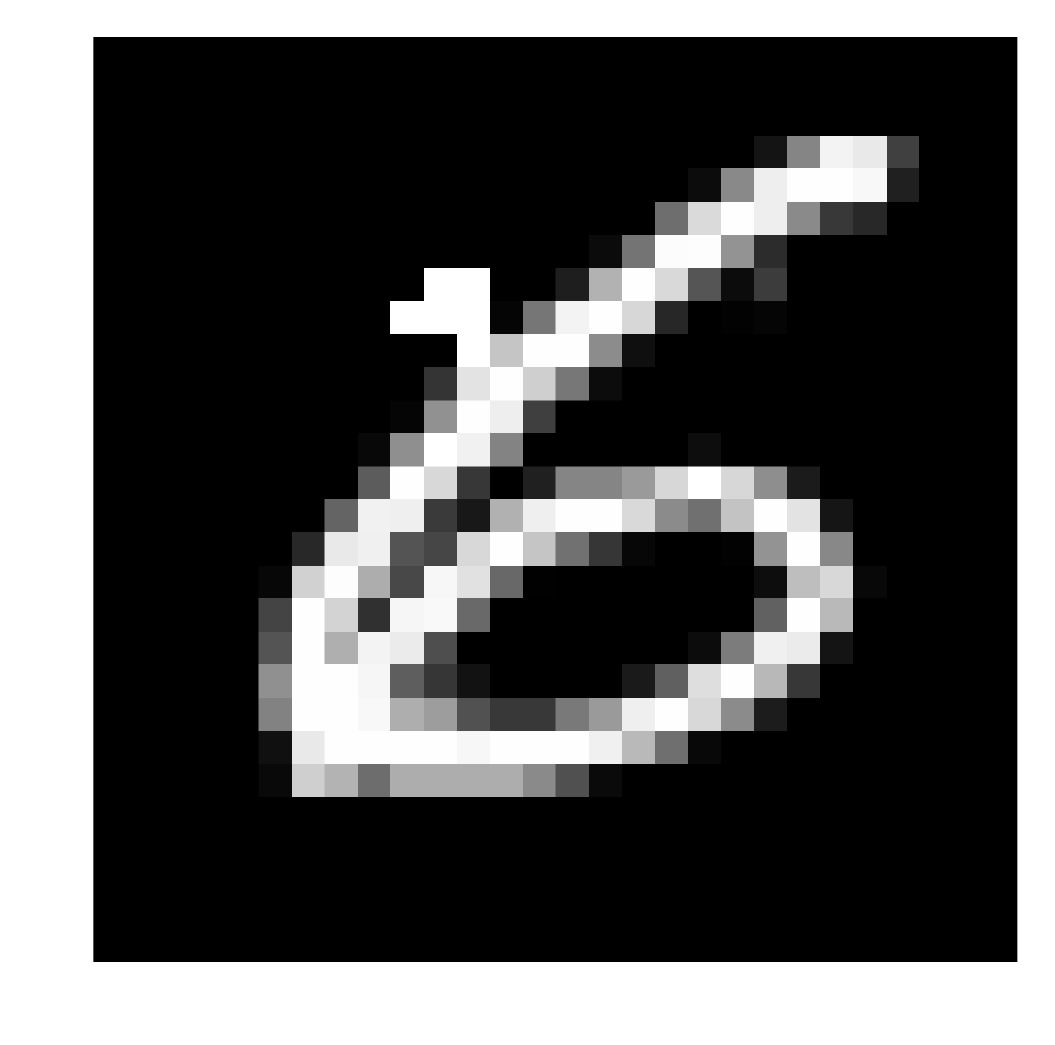}\!
    \caption*{$8$ (9)}
    \end{subfigure}
    
    \begin{subfigure}[b]{0.1\linewidth}
    \includegraphics[width=\linewidth]{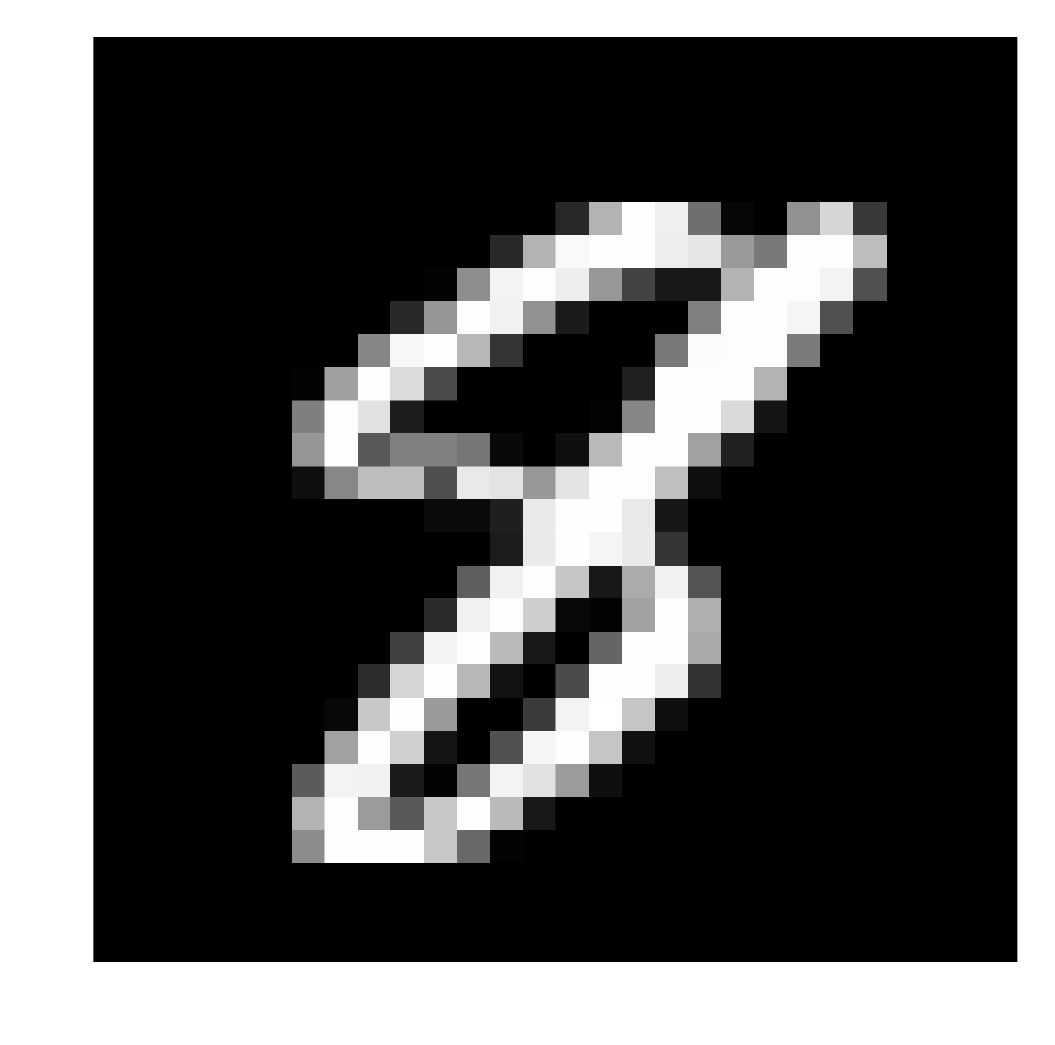}\!
    \caption*{$3$ (2)}
    \end{subfigure}
    \begin{subfigure}[b]{0.1\linewidth}
    \includegraphics[width=\linewidth]{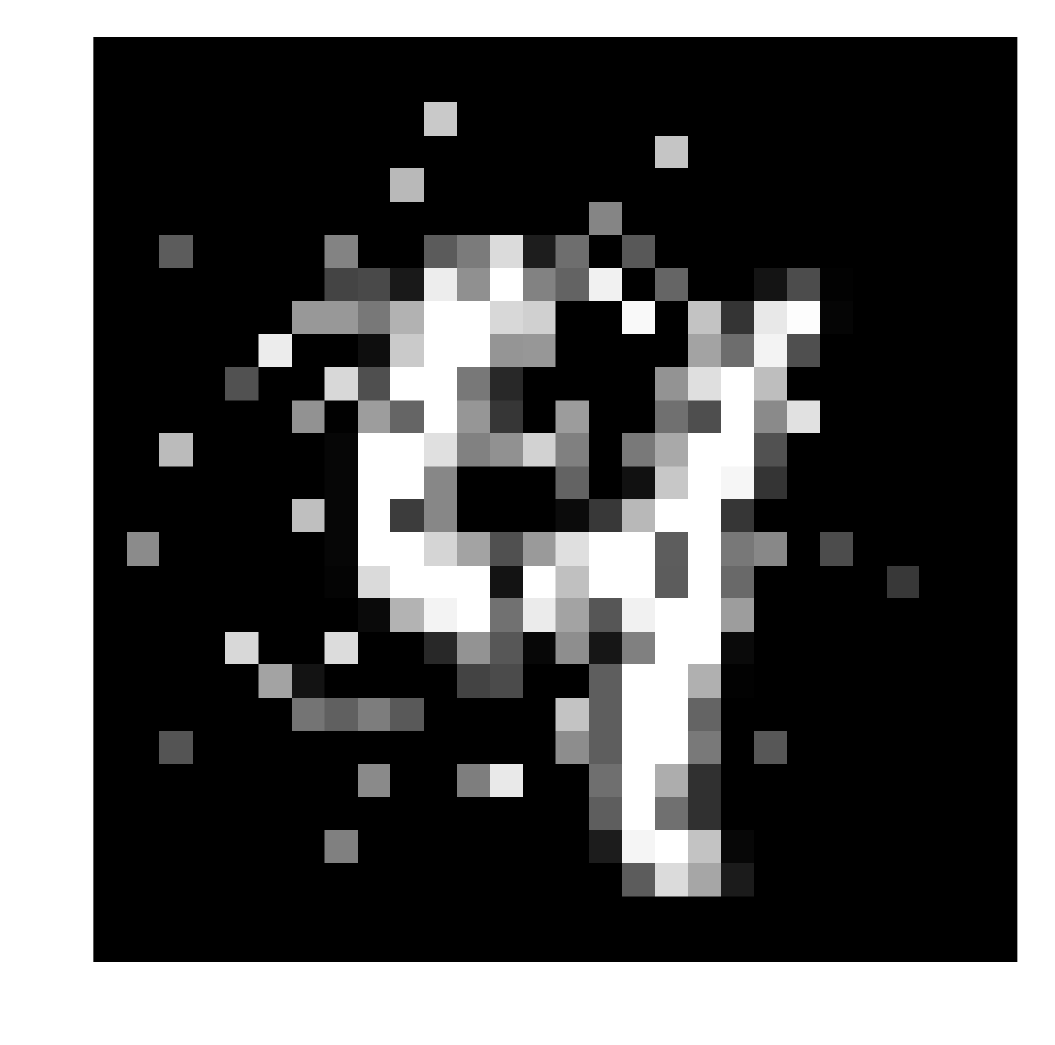}\!
    \caption*{$9$ (84)}
    \end{subfigure}
    \begin{subfigure}[b]{0.1\linewidth}
    \includegraphics[width=\linewidth]{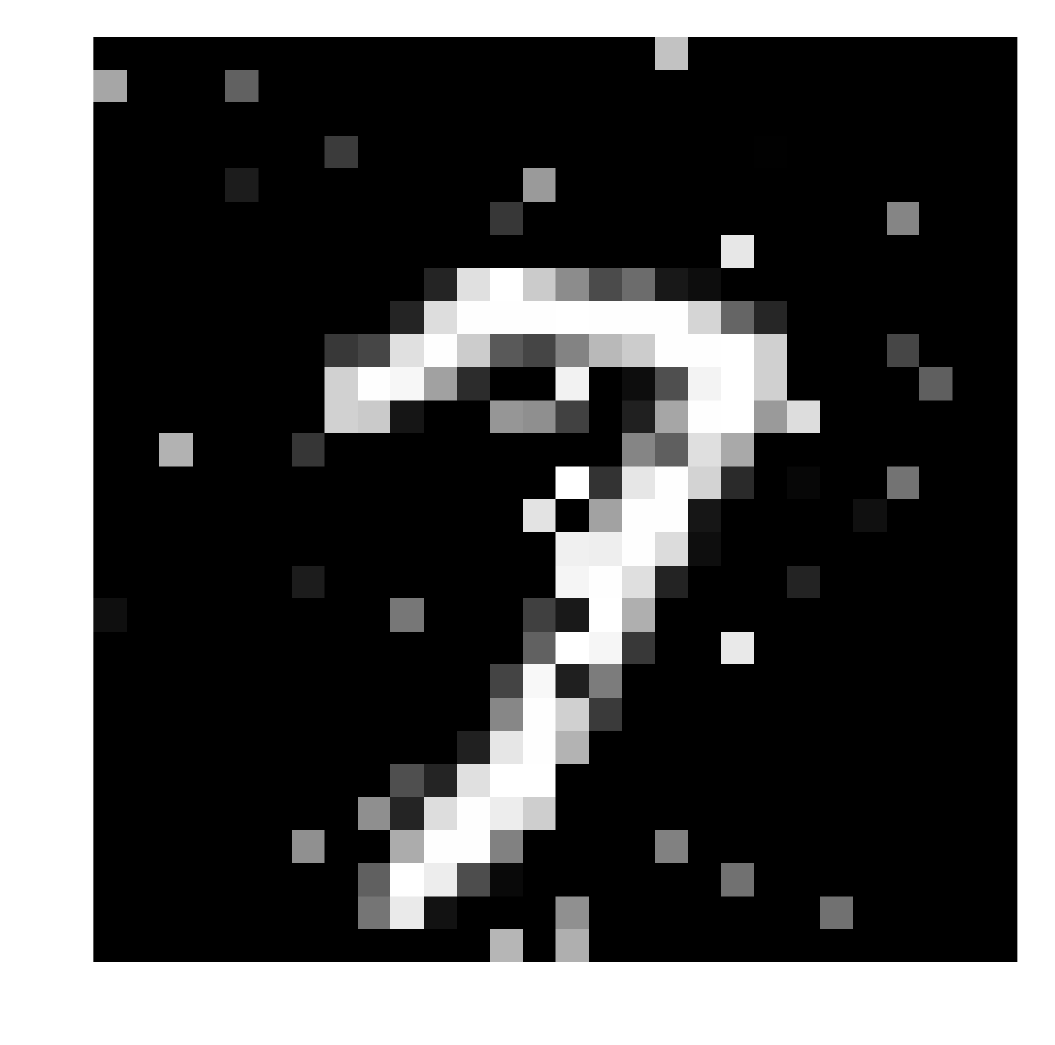}\!
    \caption*{$3$ (47)}
    \end{subfigure}
    \begin{subfigure}[b]{0.1\linewidth}
    \includegraphics[width=\linewidth]{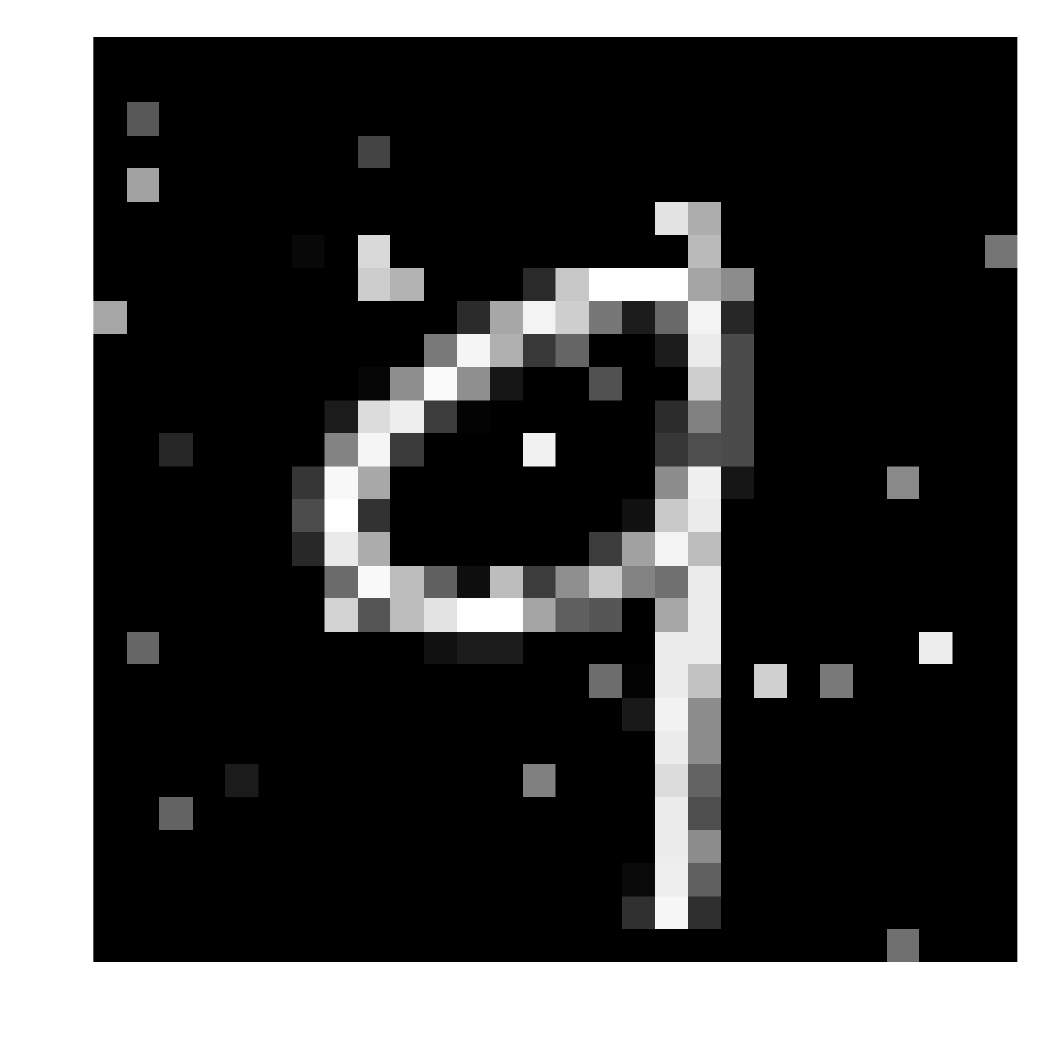}\!
    \caption*{$4$ (35)}
    \end{subfigure}
    \begin{subfigure}[b]{0.1\linewidth}
    \includegraphics[width=\linewidth]{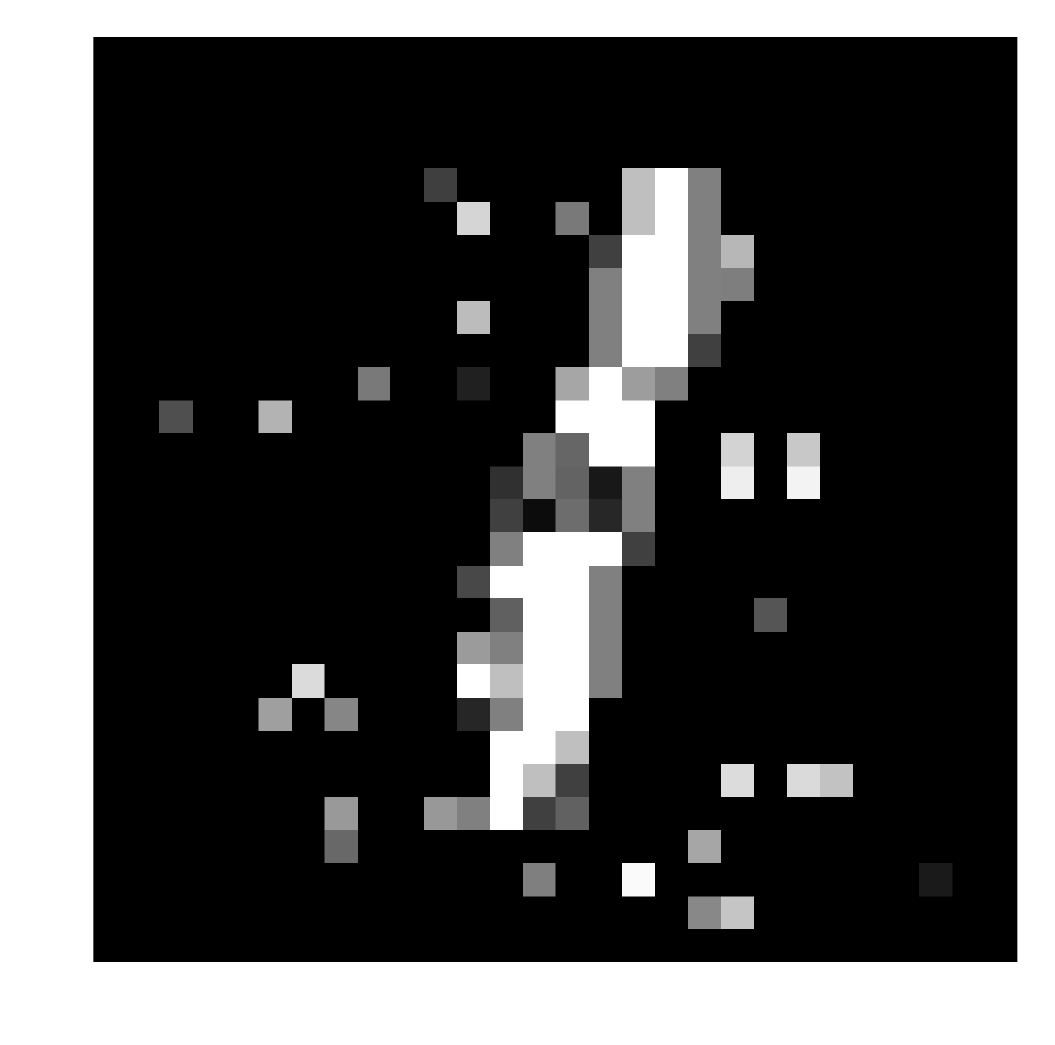}\!
    \caption*{$8$ (44)}
    \end{subfigure}
    \begin{subfigure}[b]{0.1\linewidth}
    \includegraphics[width=\linewidth]{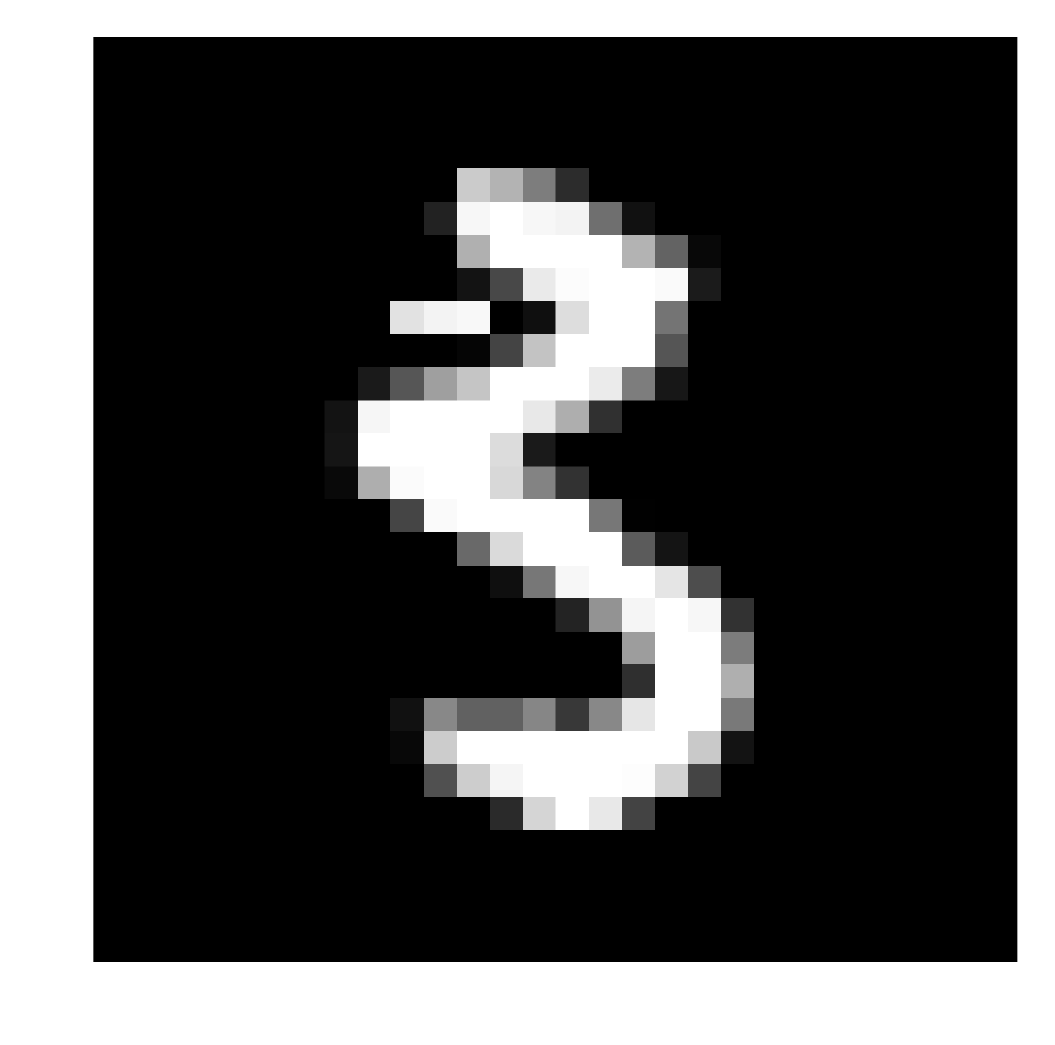}\!
    \caption*{$5$ (3)}
    \end{subfigure}
    \begin{subfigure}[b]{0.1\linewidth}
    \includegraphics[width=\linewidth]{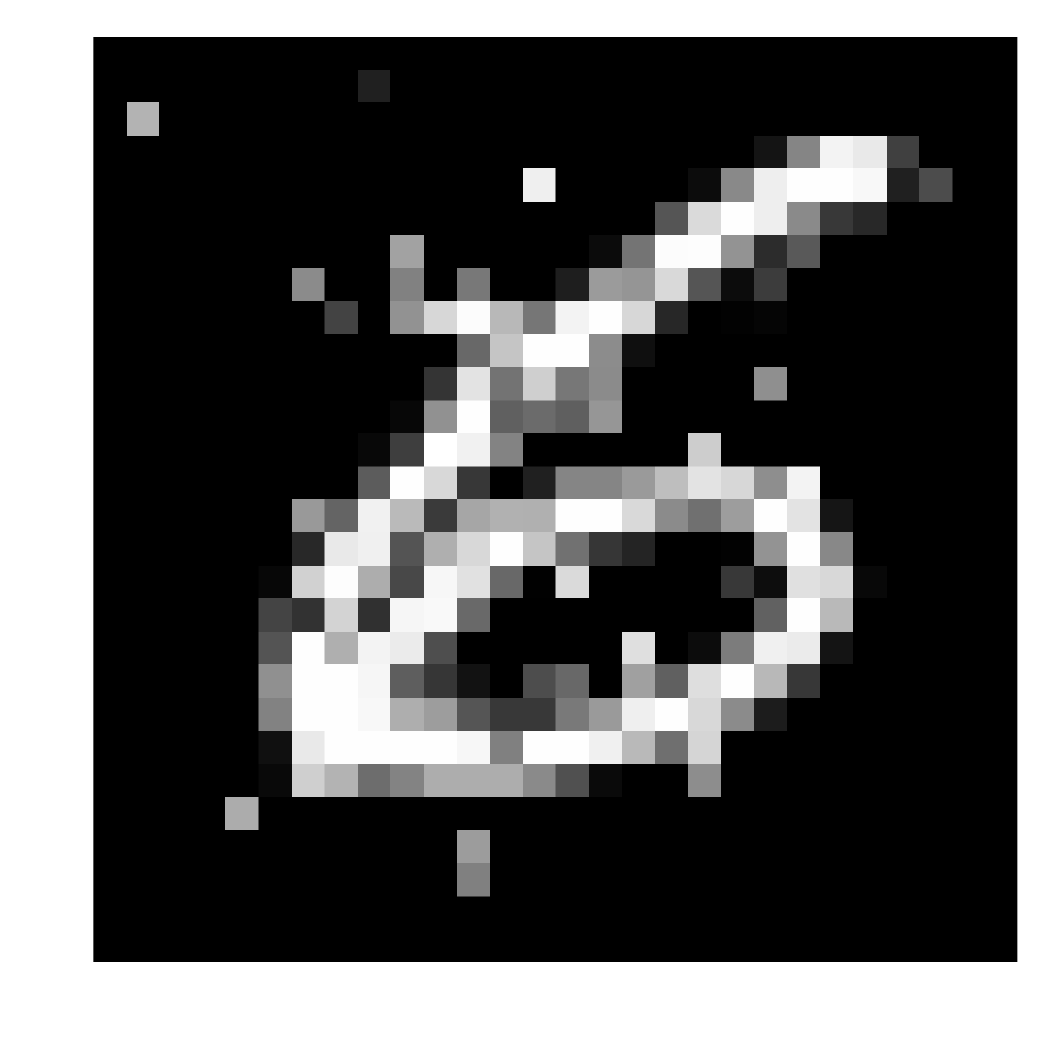}\!
    \caption*{$8$ (51)}
    \end{subfigure}
    
    \begin{subfigure}[b]{0.1\linewidth}
    \includegraphics[width=\linewidth]{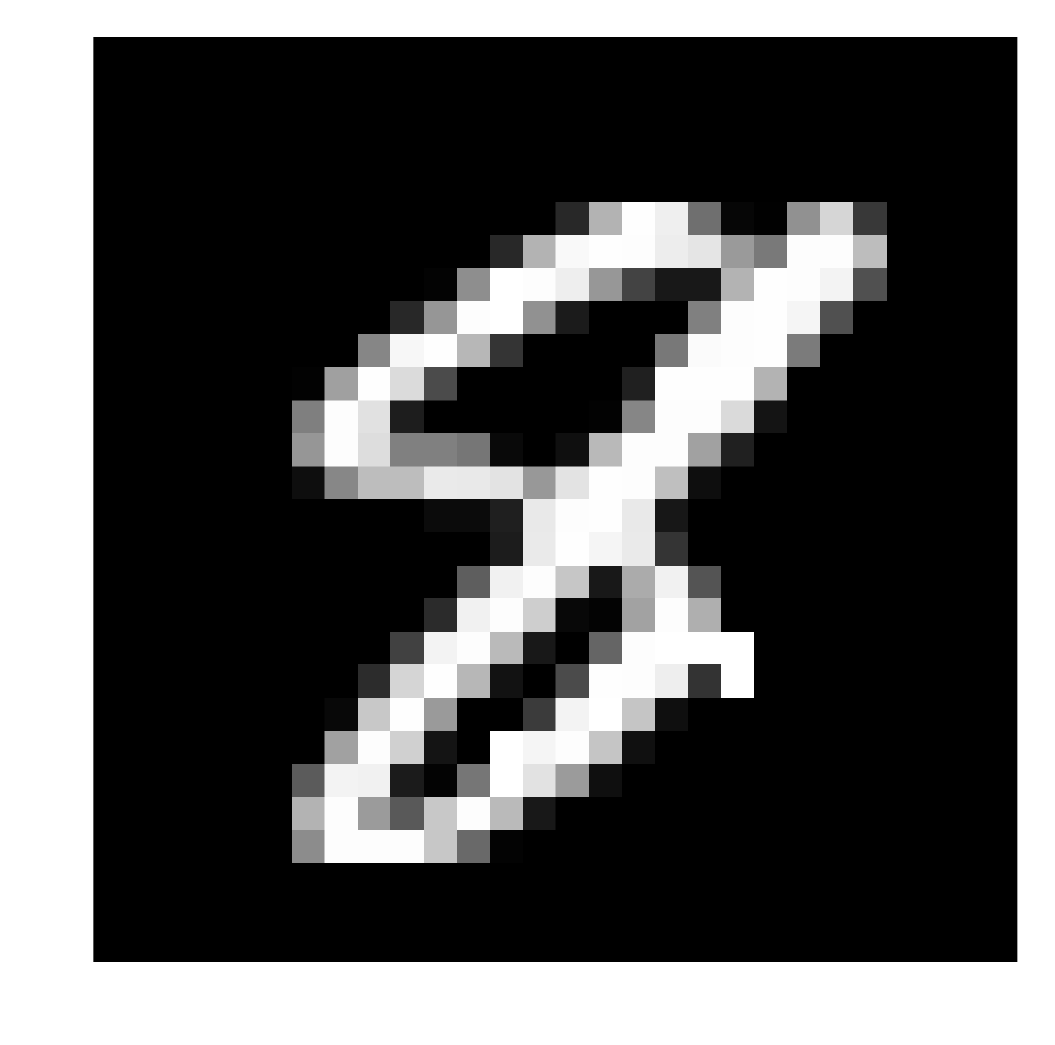}\!
    \caption*{$3$ (8)}
    \end{subfigure}
    \begin{subfigure}[b]{0.1\linewidth}
    \includegraphics[width=\linewidth]{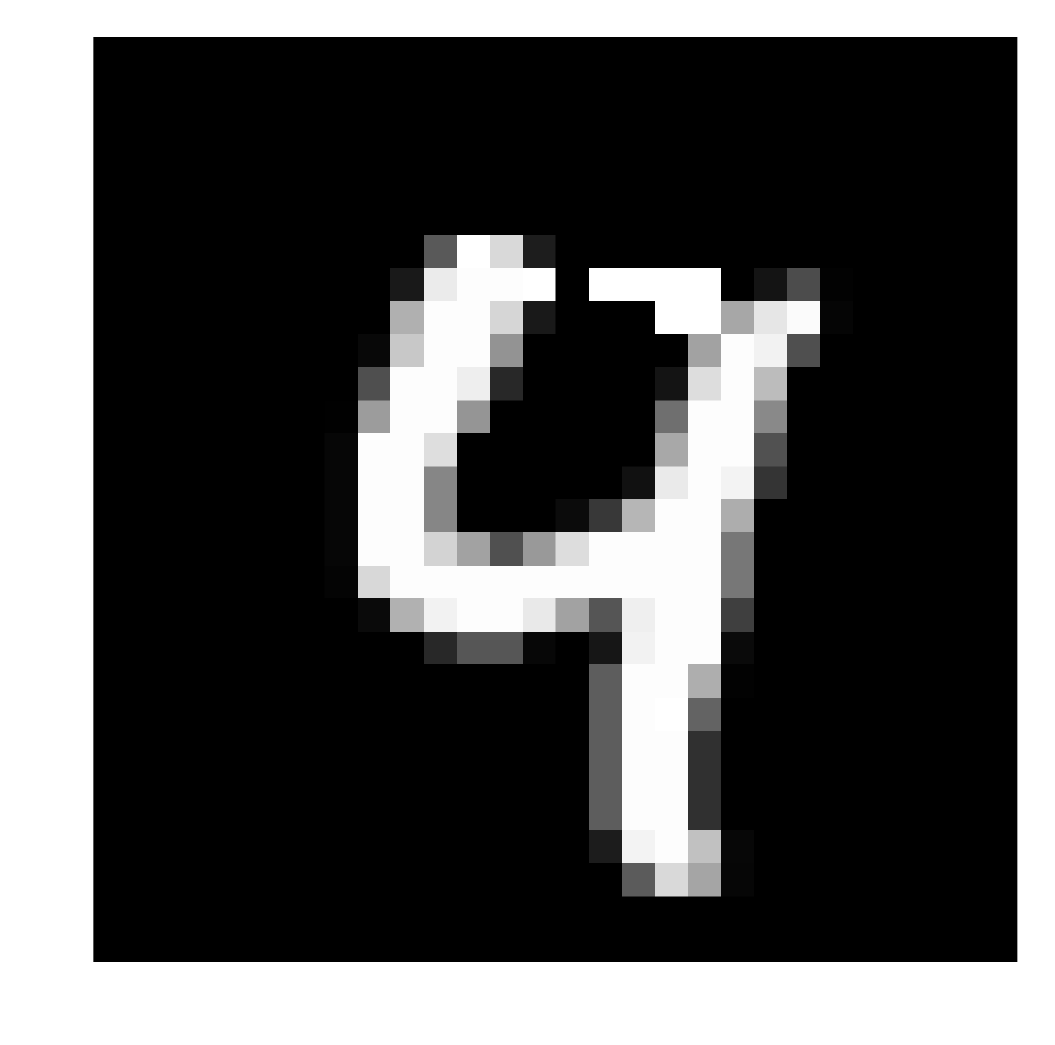}\!
    \caption*{$9$ (8)}
    \end{subfigure}
    \begin{subfigure}[b]{0.1\linewidth}
    \includegraphics[width=\linewidth]{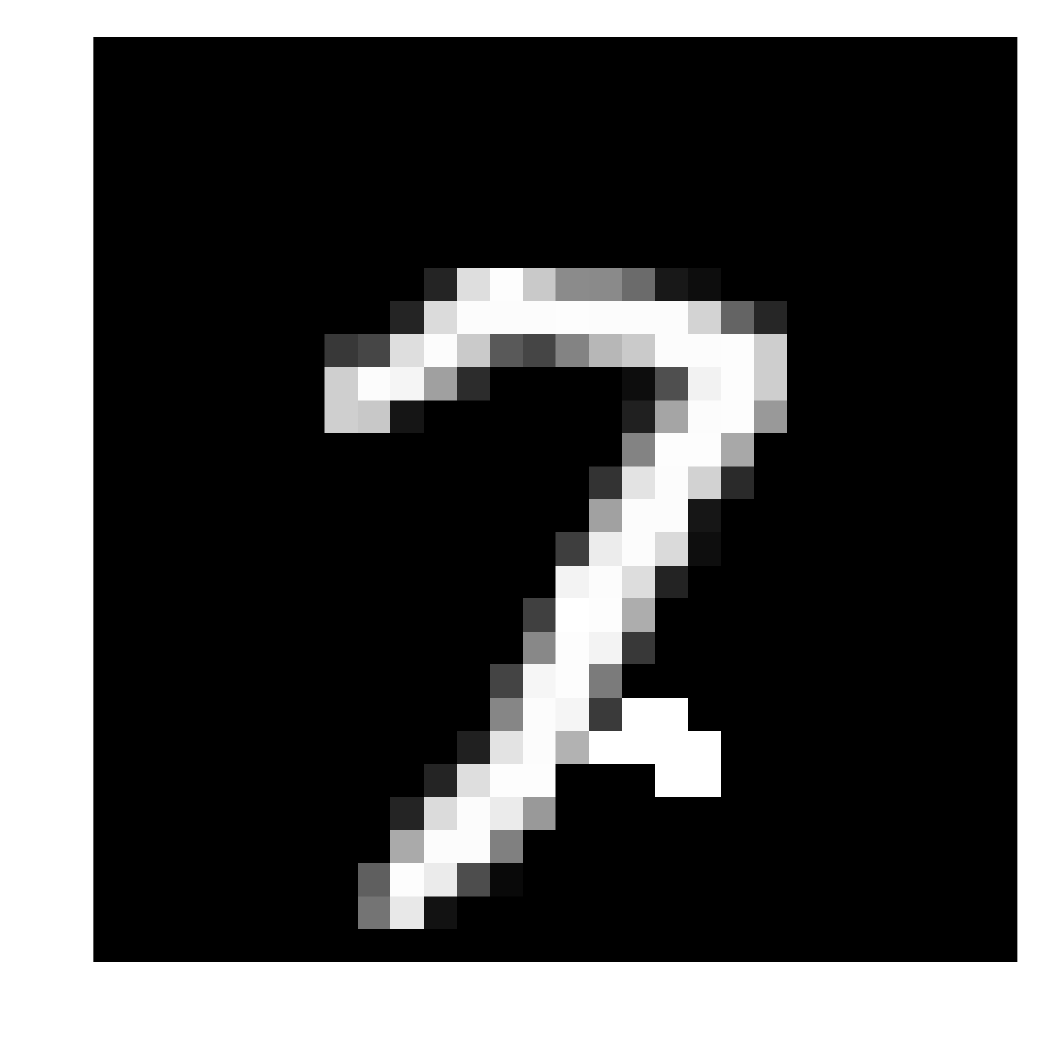}\!
    \caption*{$3$ (8)}
    \end{subfigure}
    \begin{subfigure}[b]{0.1\linewidth}
    \includegraphics[width=\linewidth]{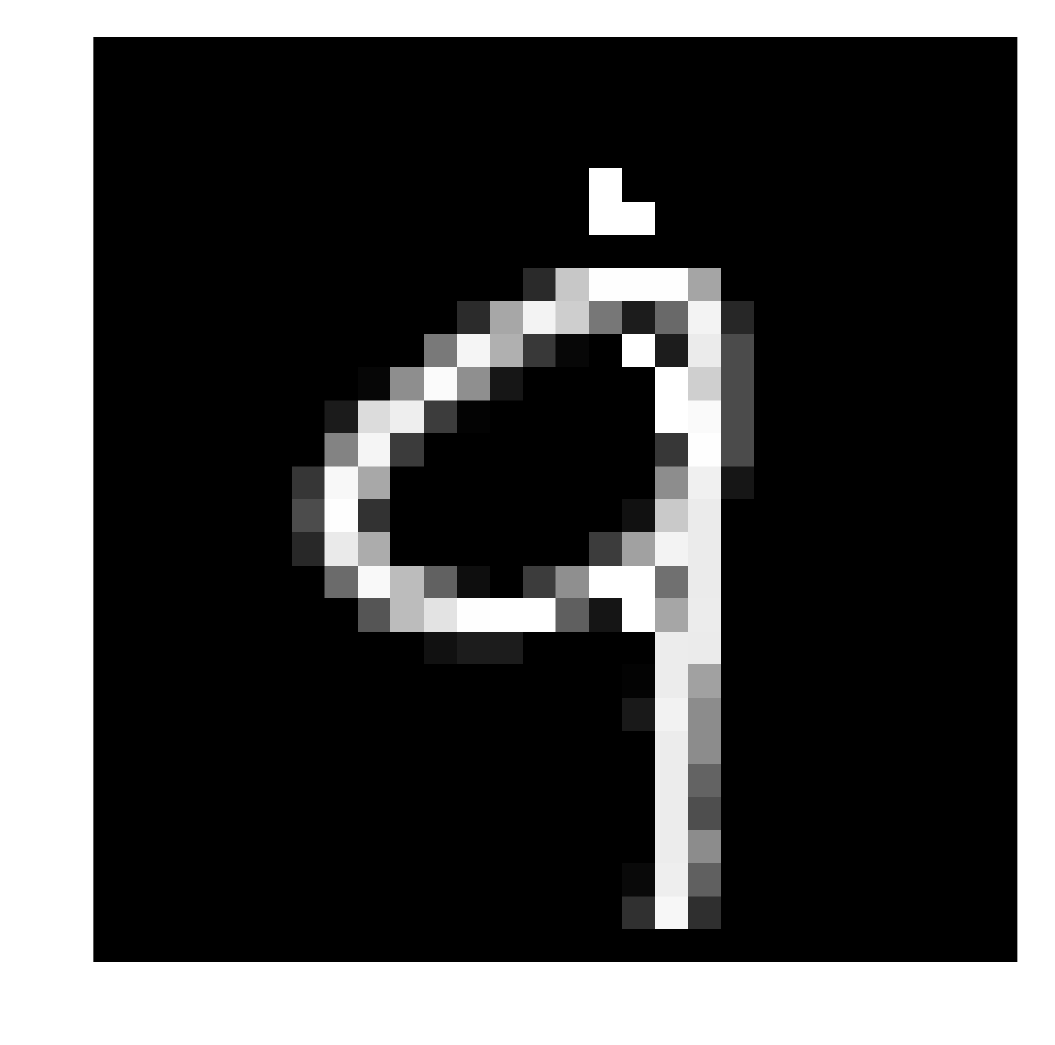}\!
    \caption*{$4$ (10)}
    \end{subfigure}
    \begin{subfigure}[b]{0.1\linewidth}
    \includegraphics[width=\linewidth]{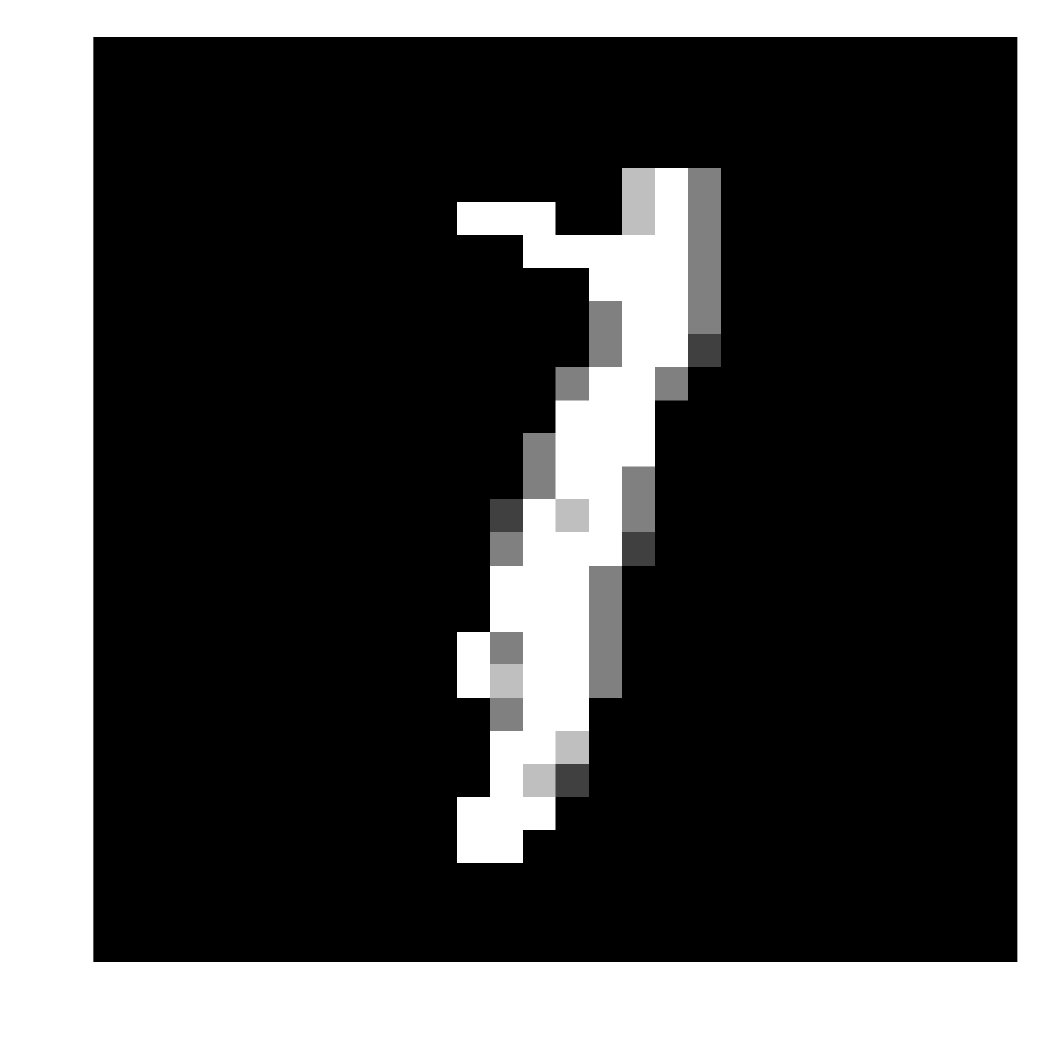}\!
    \caption*{$7$ (12)}
    \end{subfigure}
    \begin{subfigure}[b]{0.1\linewidth}
    \includegraphics[width=\linewidth]{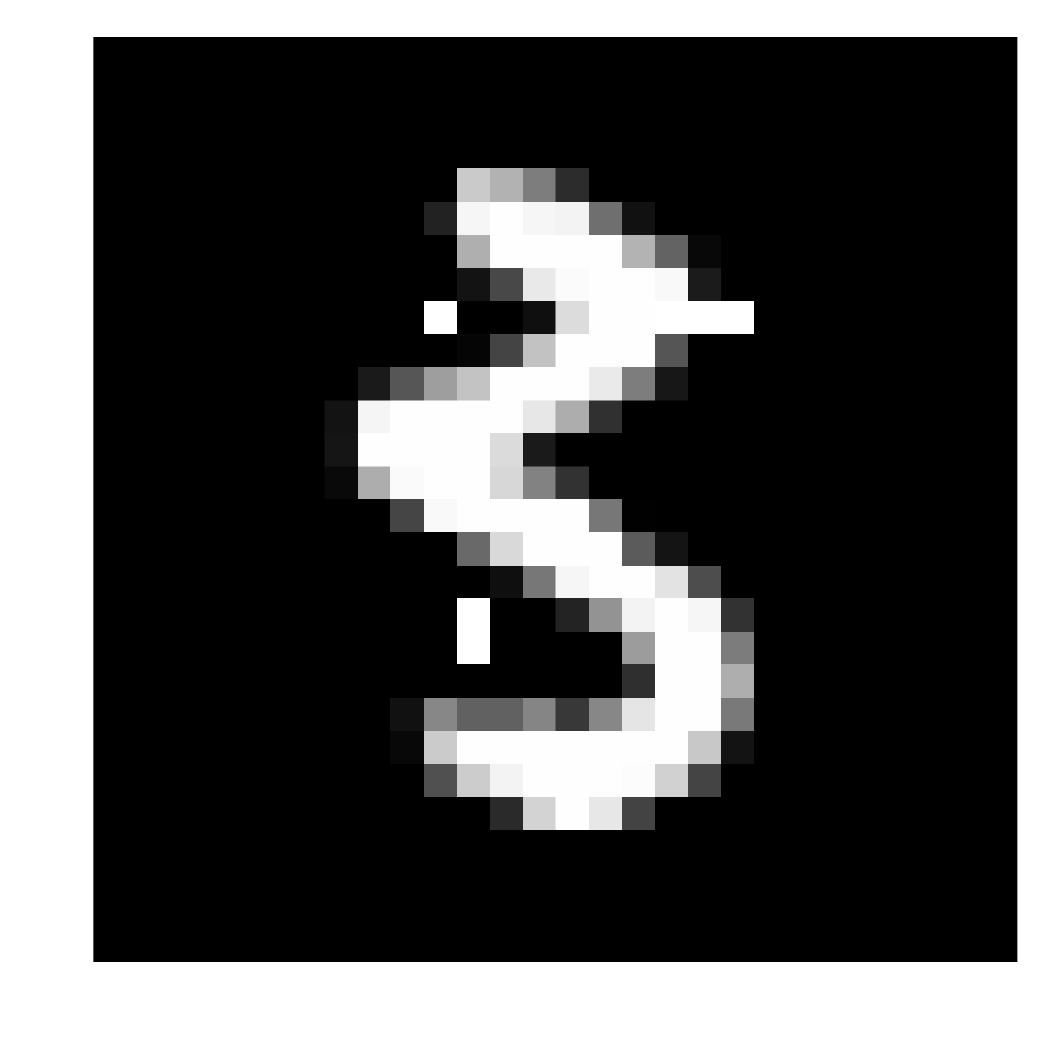}\!
    \caption*{$5$ (6)}
    \end{subfigure}
    \begin{subfigure}[b]{0.1\linewidth}
    \includegraphics[width=\linewidth]{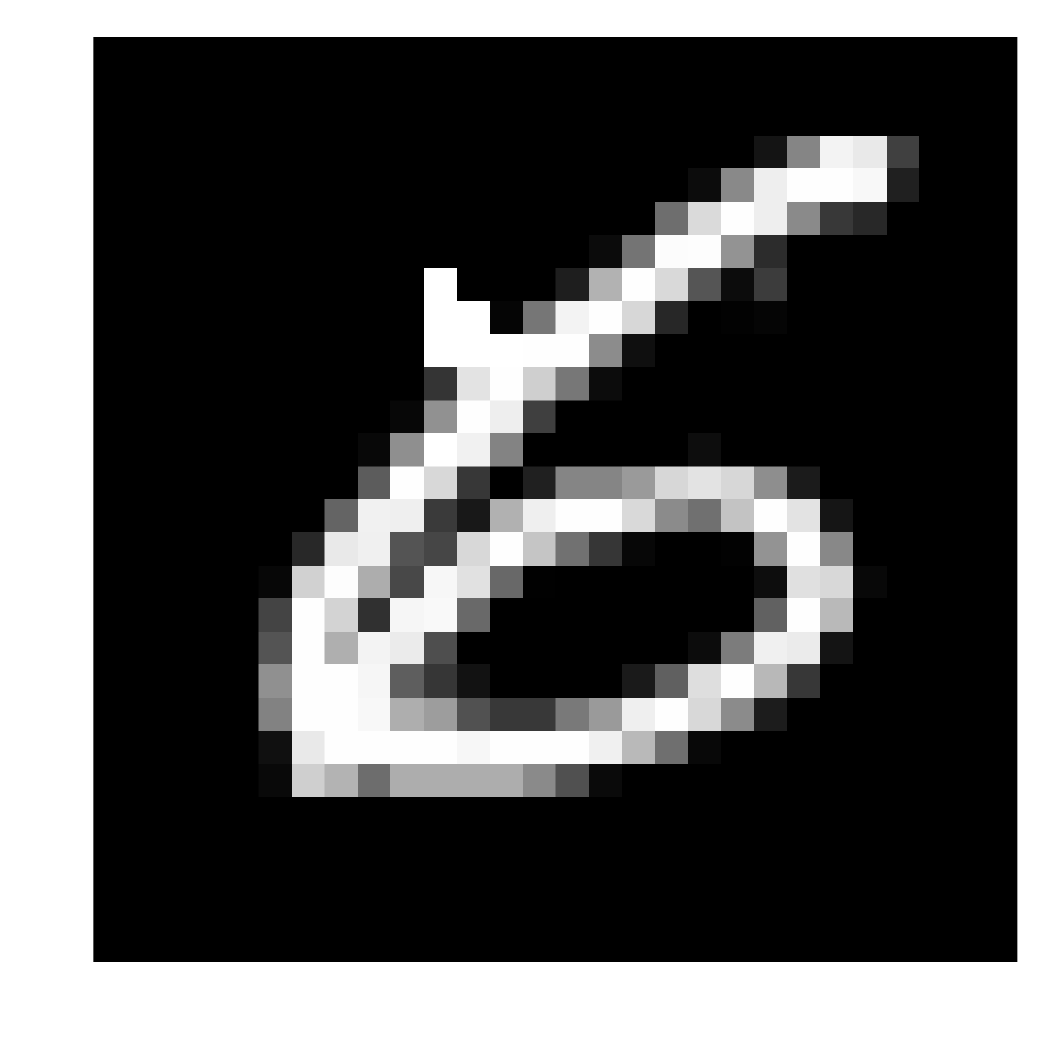}\!
    \caption*{$5$ (6)}
    \end{subfigure}
    
\caption{MNIST adversarial examples generated by (a) SparseFool (first row), (b) ``One pixel attack" (second row), and (c) JSMA (third row). The fooling label is shown below each image, and the number of perturbed pixels is written inside the parentheses.}
\label{fig:compare_mnist}
\end{figure}

\begin{figure}[hb]
\centering
\captionsetup[subfigure]{skip=1pt}
    \begin{subfigure}[b]{0.1\linewidth}
    \includegraphics[width=\linewidth]{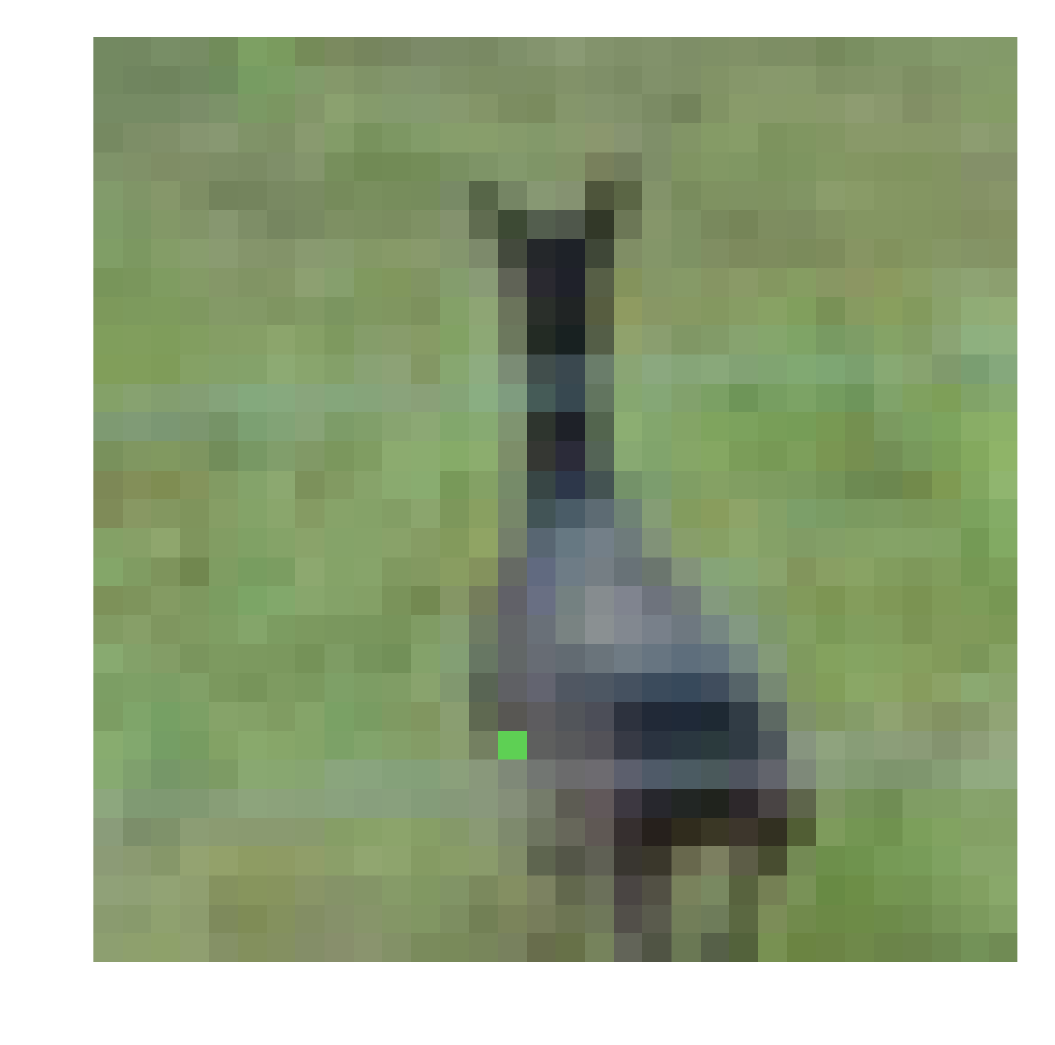}\!
    \caption*{deer (1)}
    \end{subfigure}
    \begin{subfigure}[b]{0.1\linewidth}
    \includegraphics[width=\linewidth]{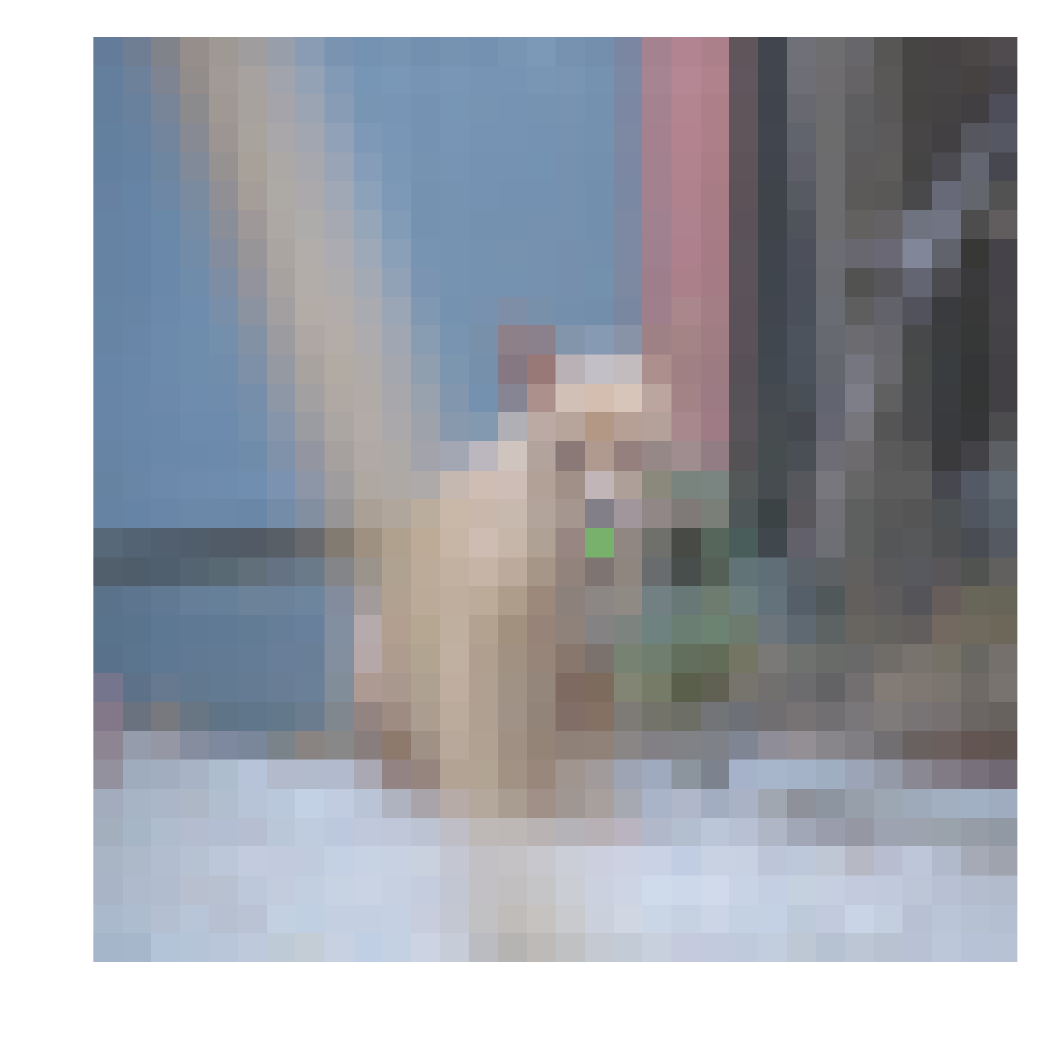}\!
    \caption*{cat (1)}
    \end{subfigure}
    \begin{subfigure}[b]{0.1\linewidth}
    \includegraphics[width=\linewidth]{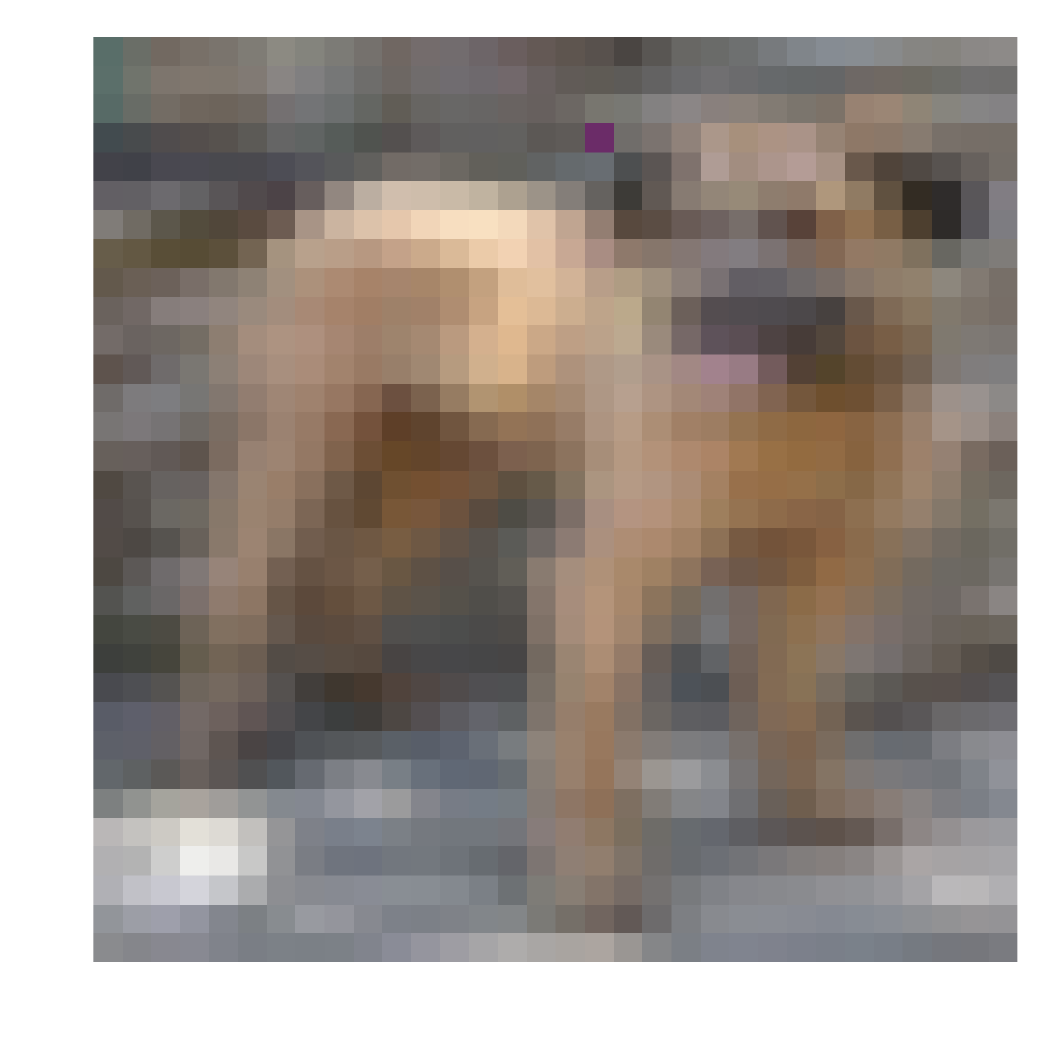}\!
    \caption*{deer (1)}
    \end{subfigure}
    \begin{subfigure}[b]{0.1\linewidth}
    \includegraphics[width=\linewidth]{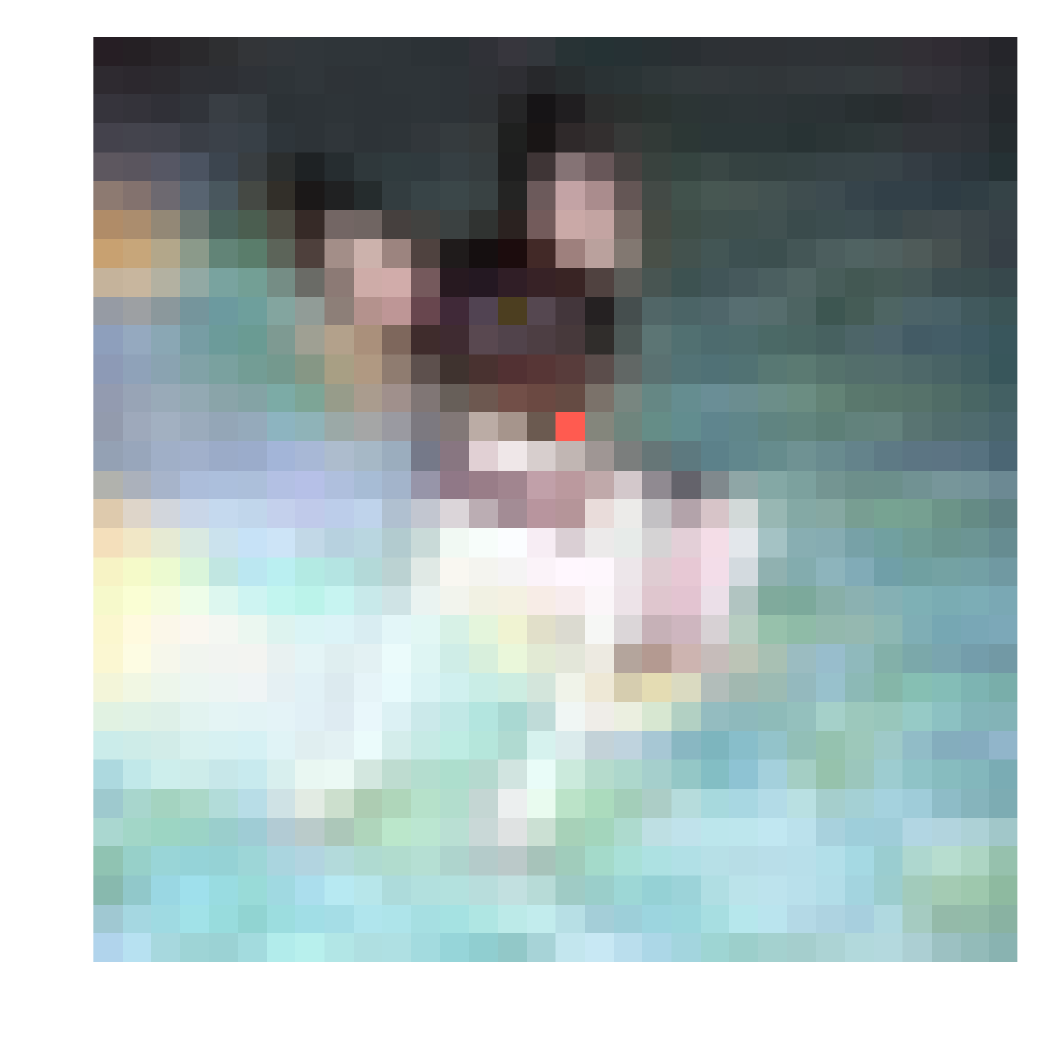}\!
    \caption*{cat (2)}
    \end{subfigure}
    \begin{subfigure}[b]{0.1\linewidth}
    \includegraphics[width=\linewidth]{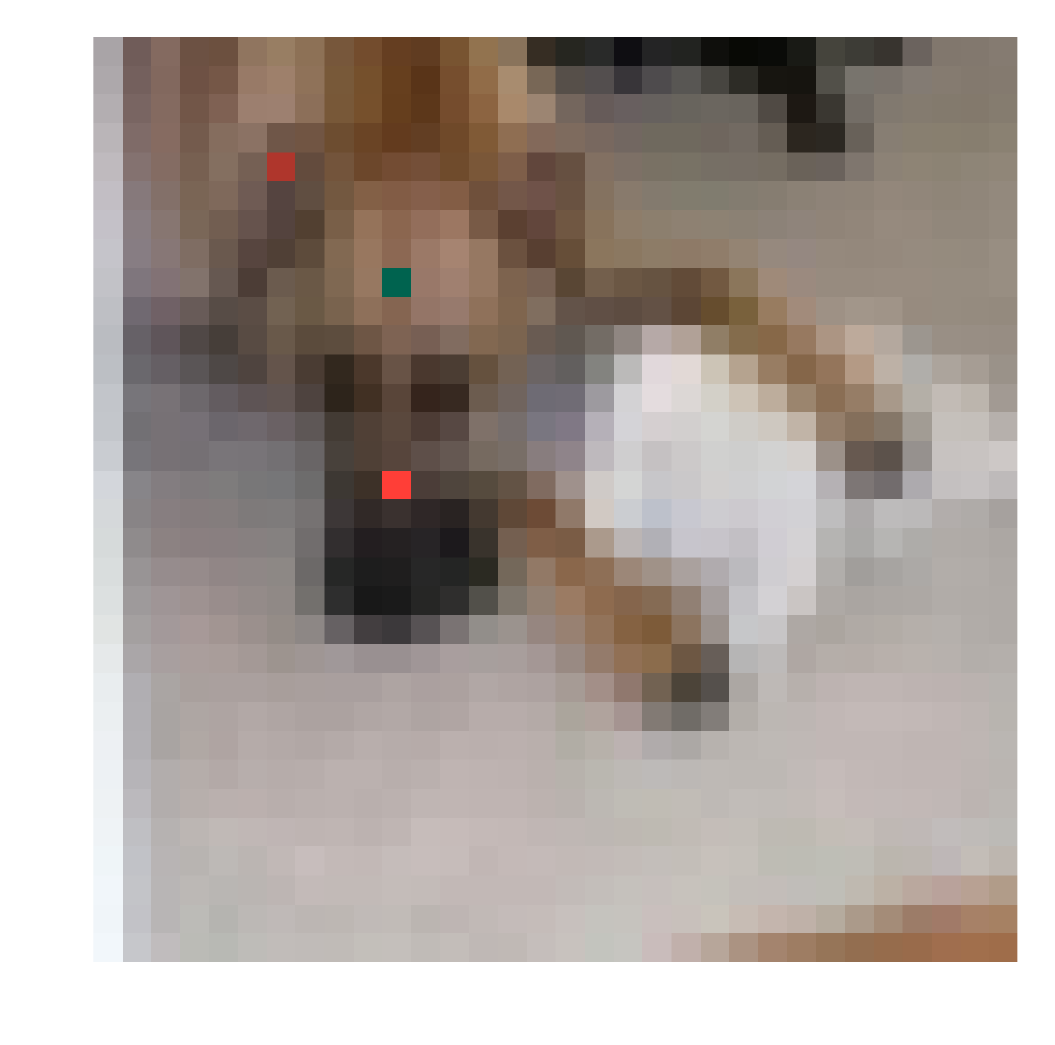}\!
    \caption*{cat (3)}
    \end{subfigure}
    \begin{subfigure}[b]{0.1\linewidth}
    \includegraphics[width=\linewidth]{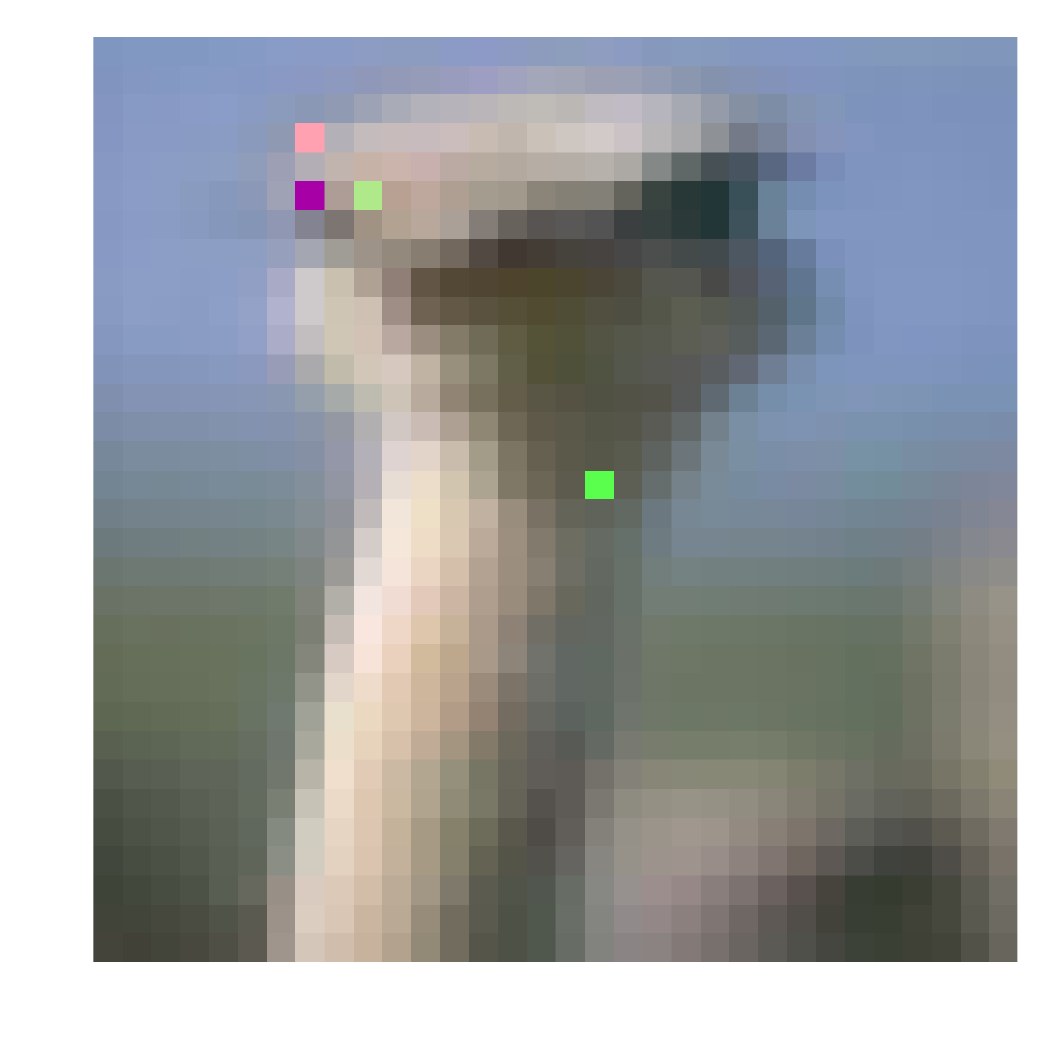}\!
    \caption*{plane (4)}
    \end{subfigure}
    \begin{subfigure}[b]{0.1\linewidth}
    \includegraphics[width=\linewidth]{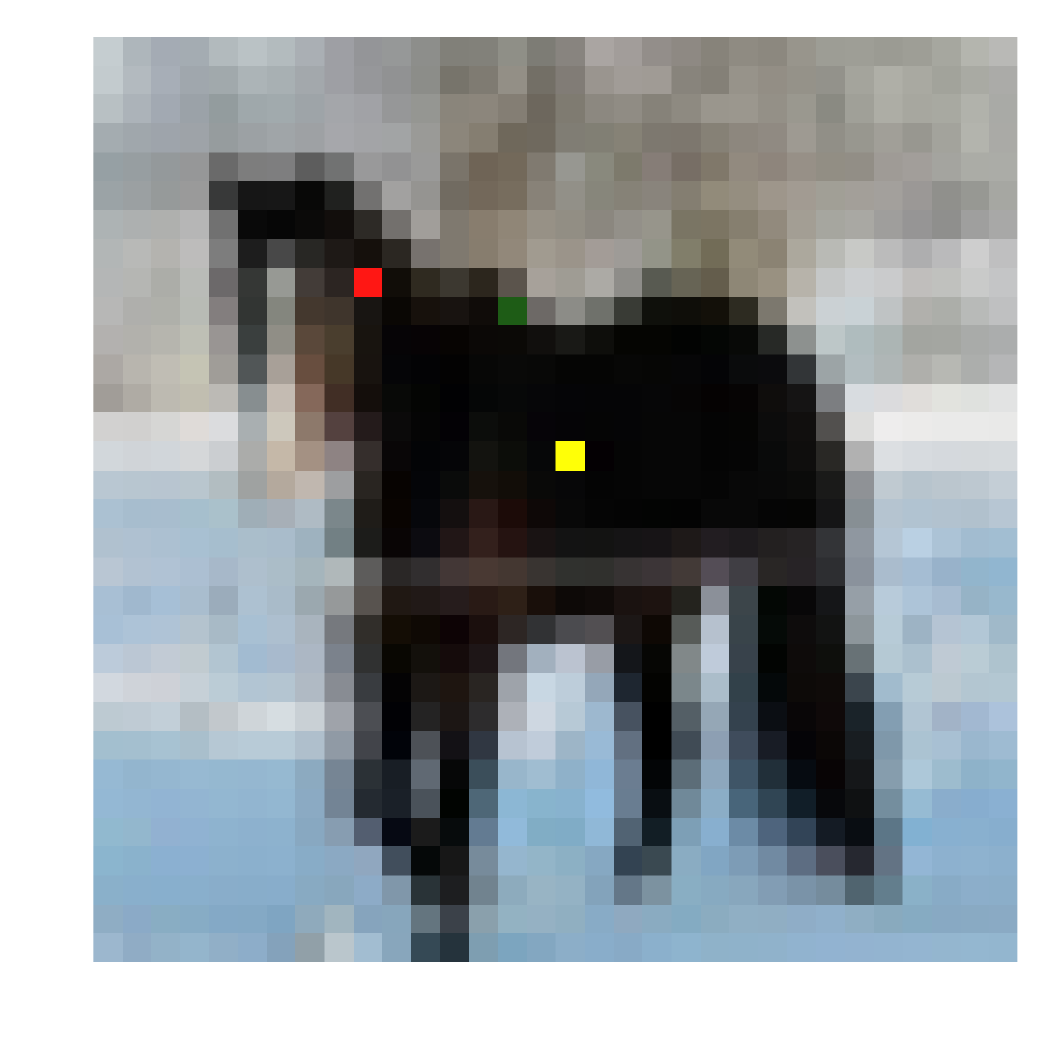}\!
    \caption*{cat (4)}
    \end{subfigure}
    
    \begin{subfigure}[b]{0.1\linewidth}
    \includegraphics[width=\linewidth]{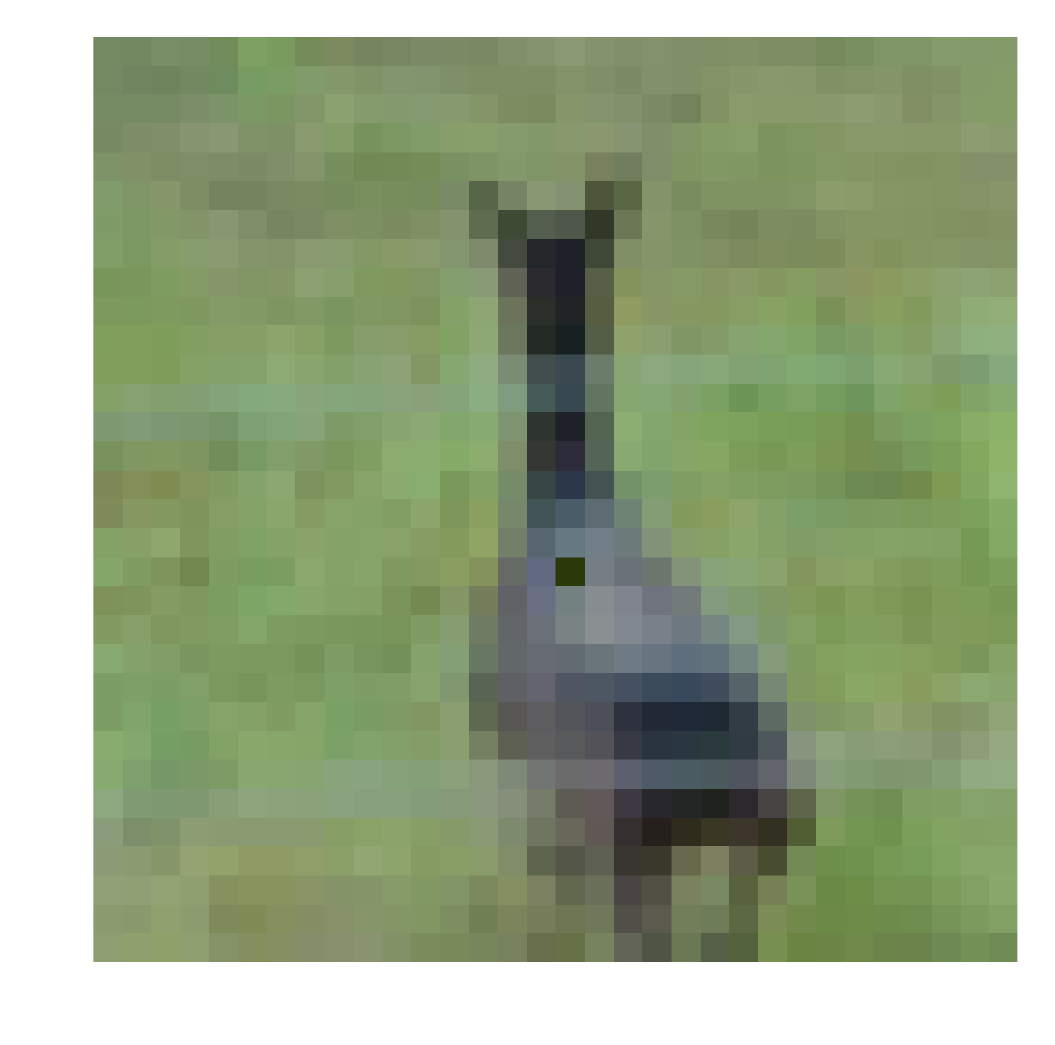}\!
    \caption*{deer (1)}
    \end{subfigure}
    \begin{subfigure}[b]{0.1\linewidth}
    \includegraphics[width=\linewidth]{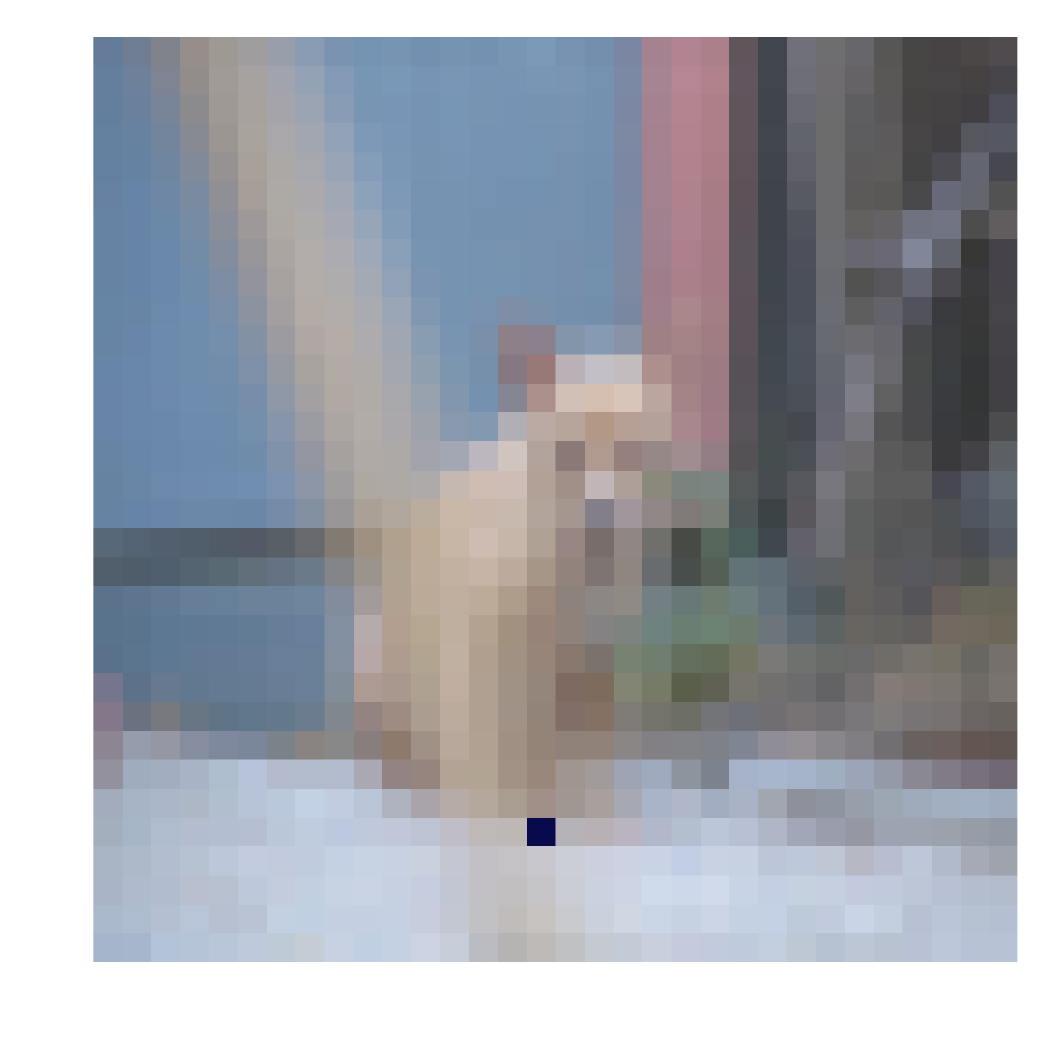}\!
    \caption*{cat (1)}
    \end{subfigure}
    \begin{subfigure}[b]{0.1\linewidth}
    \includegraphics[width=\linewidth]{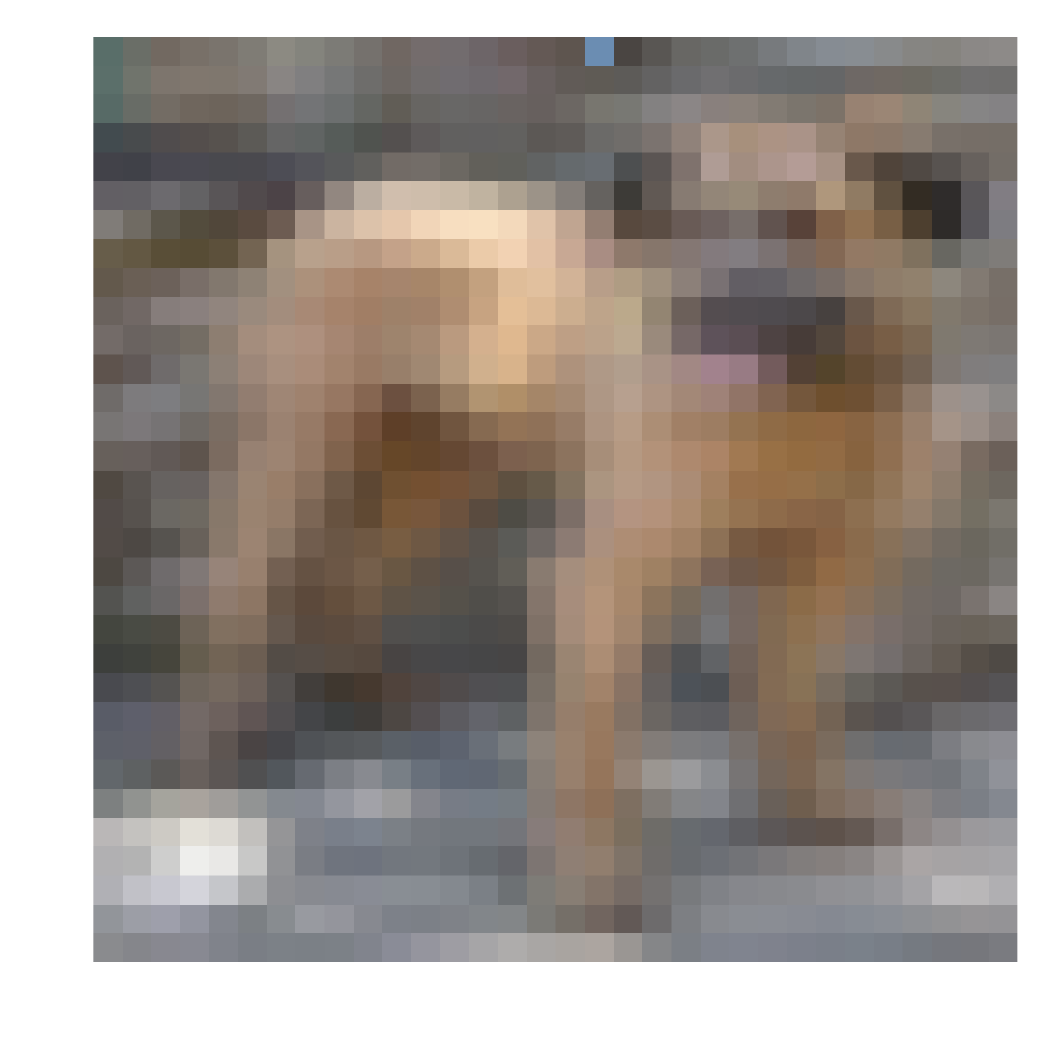}\!
    \caption*{deer (1)}
    \end{subfigure}
    \begin{subfigure}[b]{0.1\linewidth}
    \includegraphics[width=\linewidth]{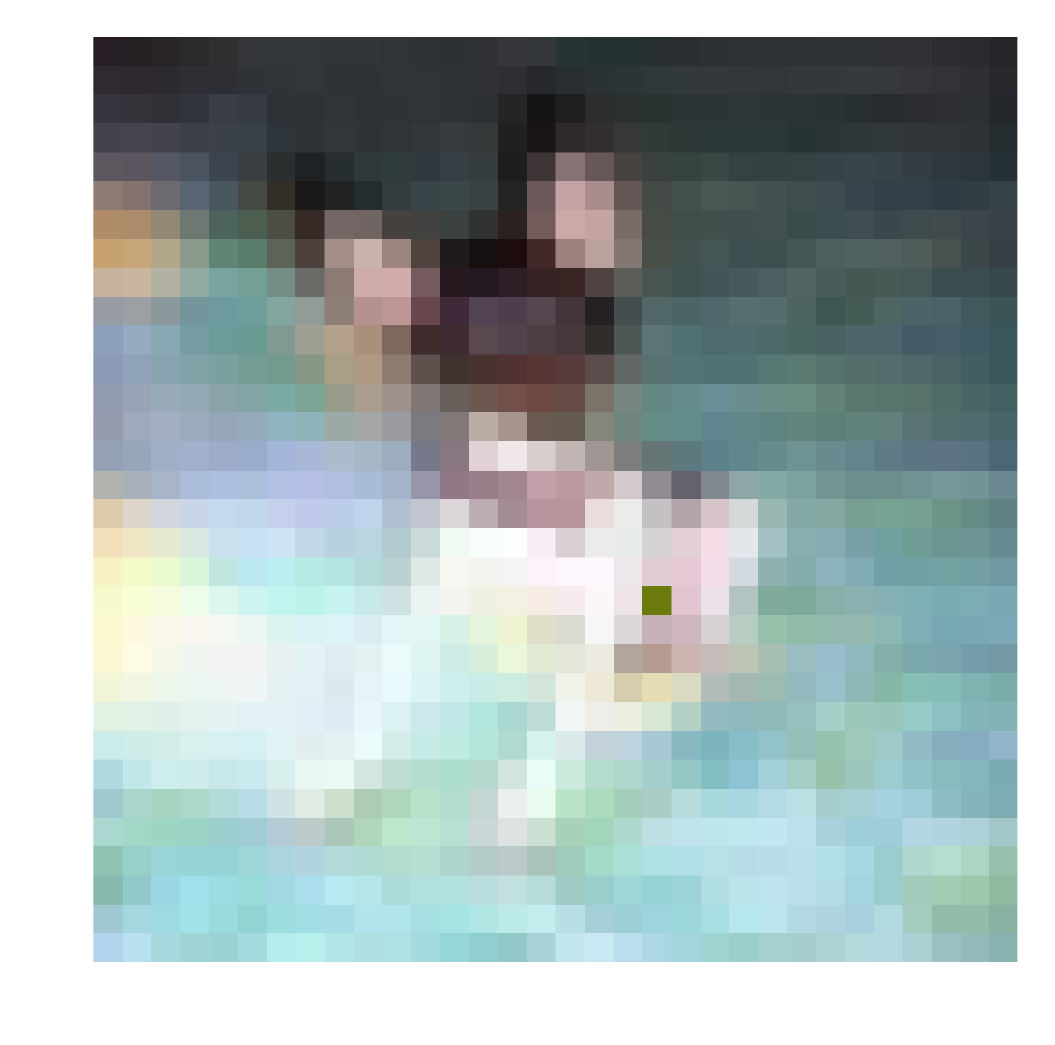}\!
    \caption*{cat (1)}
    \end{subfigure}
    \begin{subfigure}[b]{0.1\linewidth}
    \includegraphics[width=\linewidth]{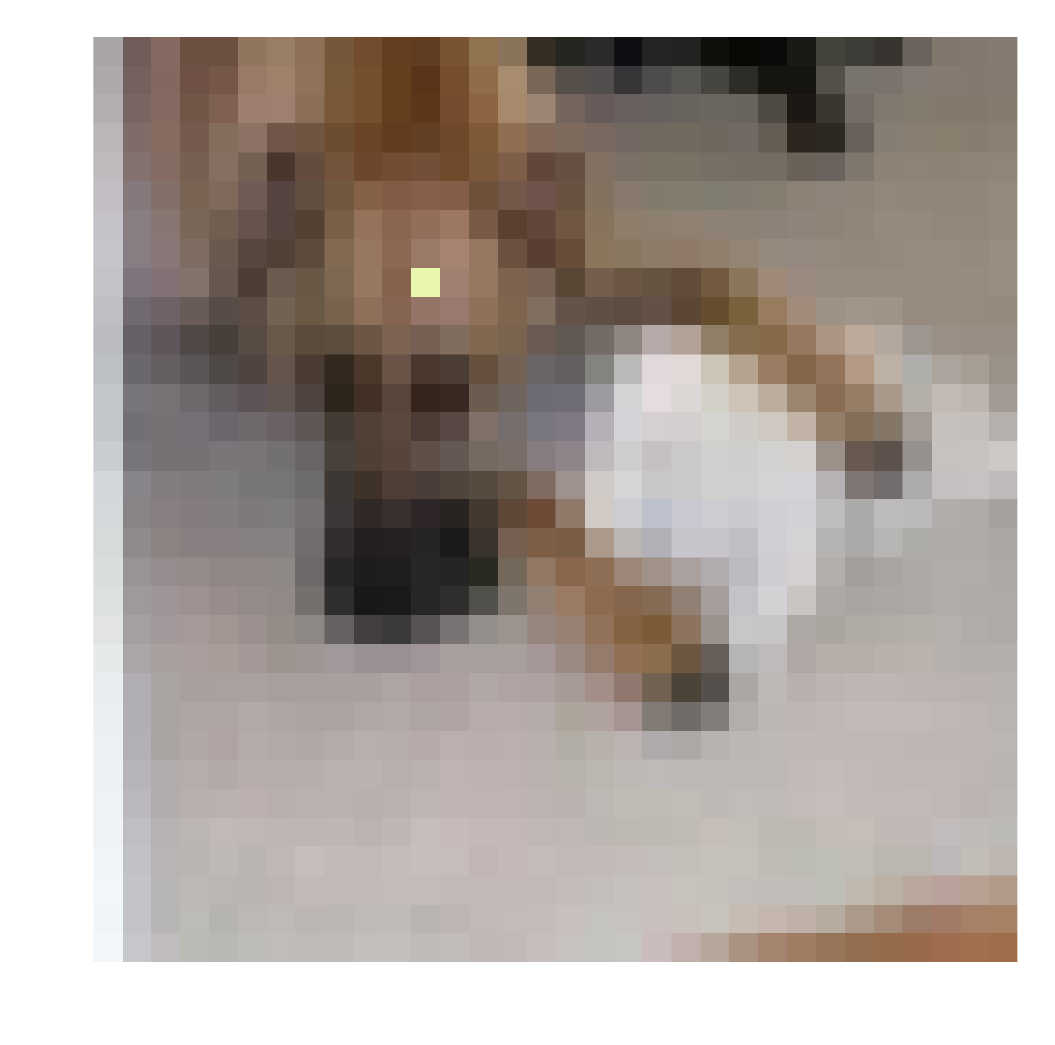}\!
    \caption*{cat (1)}
    \end{subfigure}
    \begin{subfigure}[b]{0.1\linewidth}
    \includegraphics[width=\linewidth]{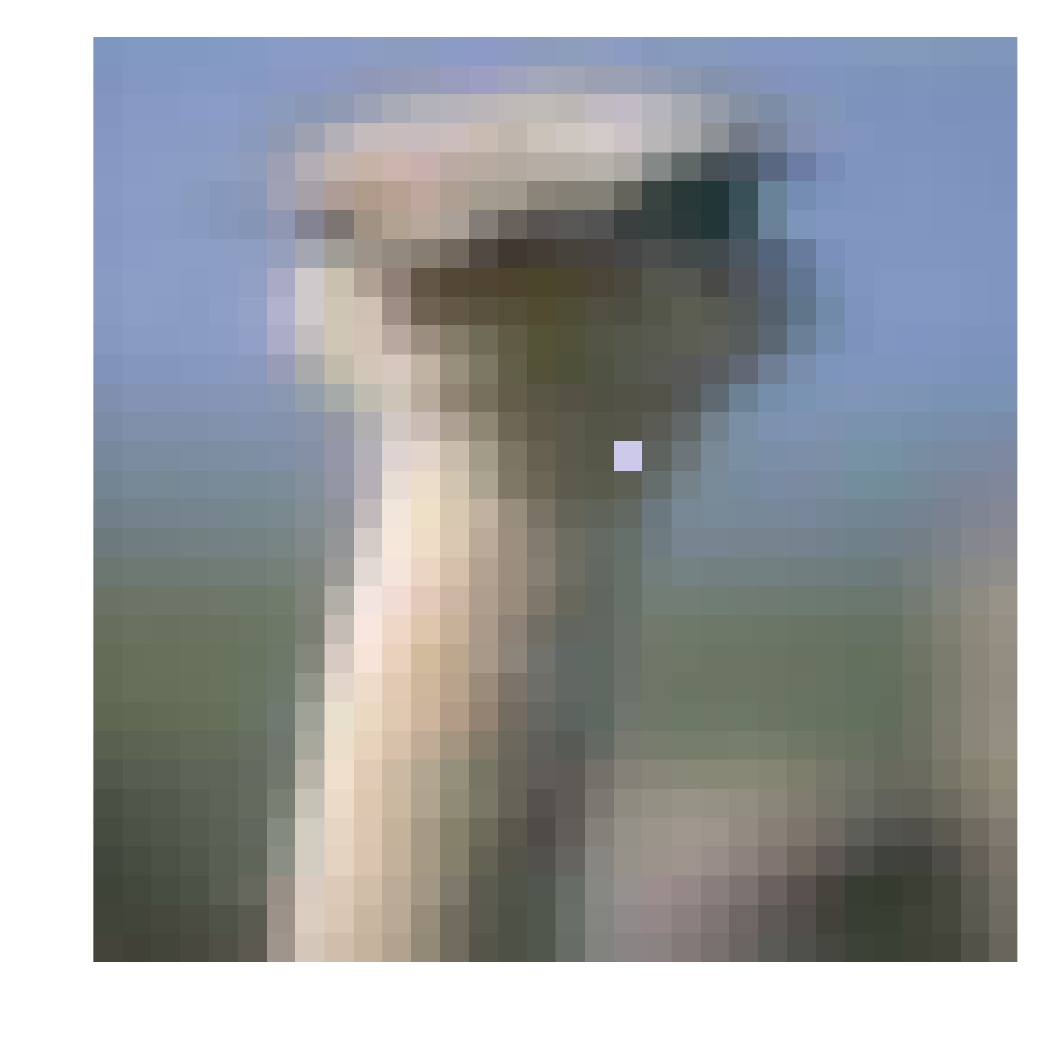}\!
    \caption*{plane (1)}
    \end{subfigure}
    \begin{subfigure}[b]{0.1\linewidth}
    \includegraphics[width=\linewidth]{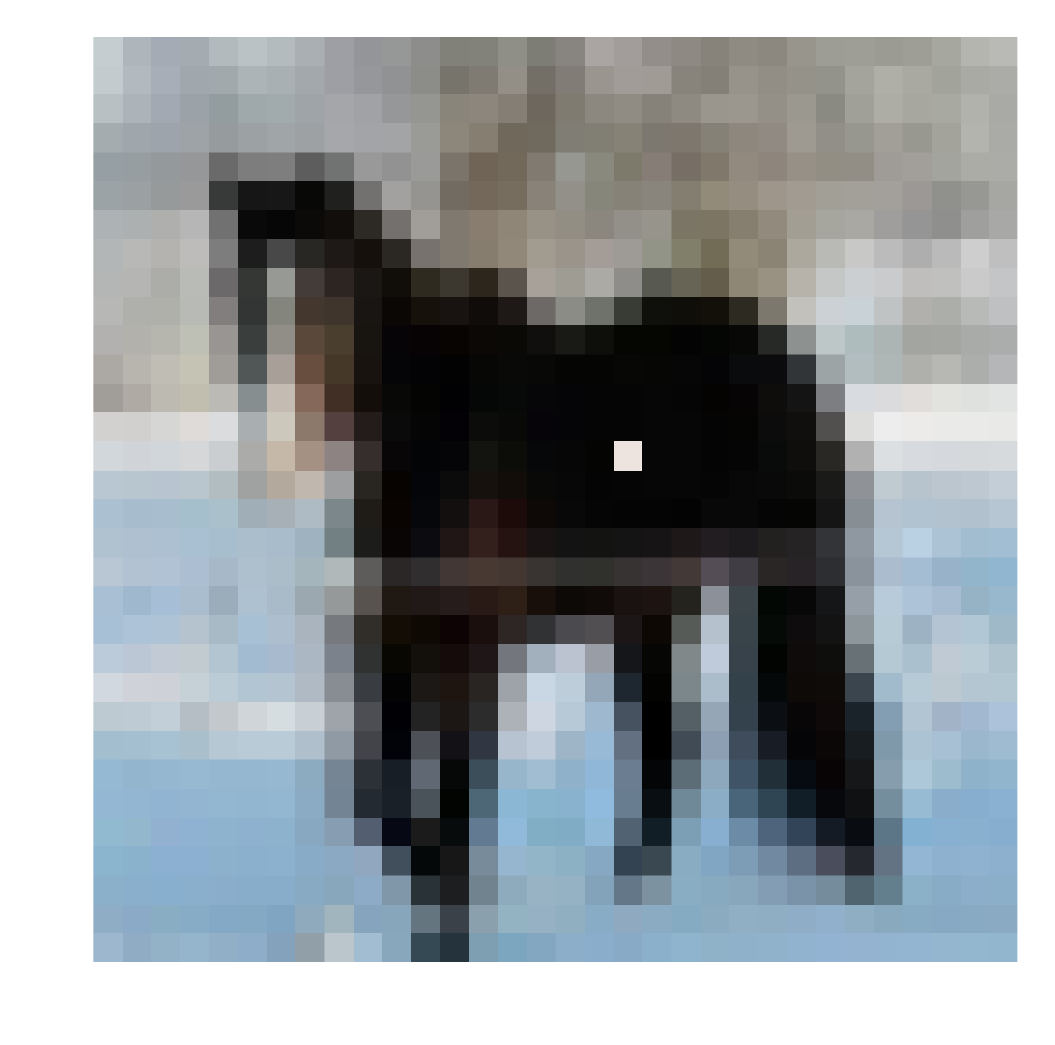}\!
    \caption*{dog (1)}
    \end{subfigure}
    
    \begin{subfigure}[b]{0.1\linewidth}
    \includegraphics[width=\linewidth]{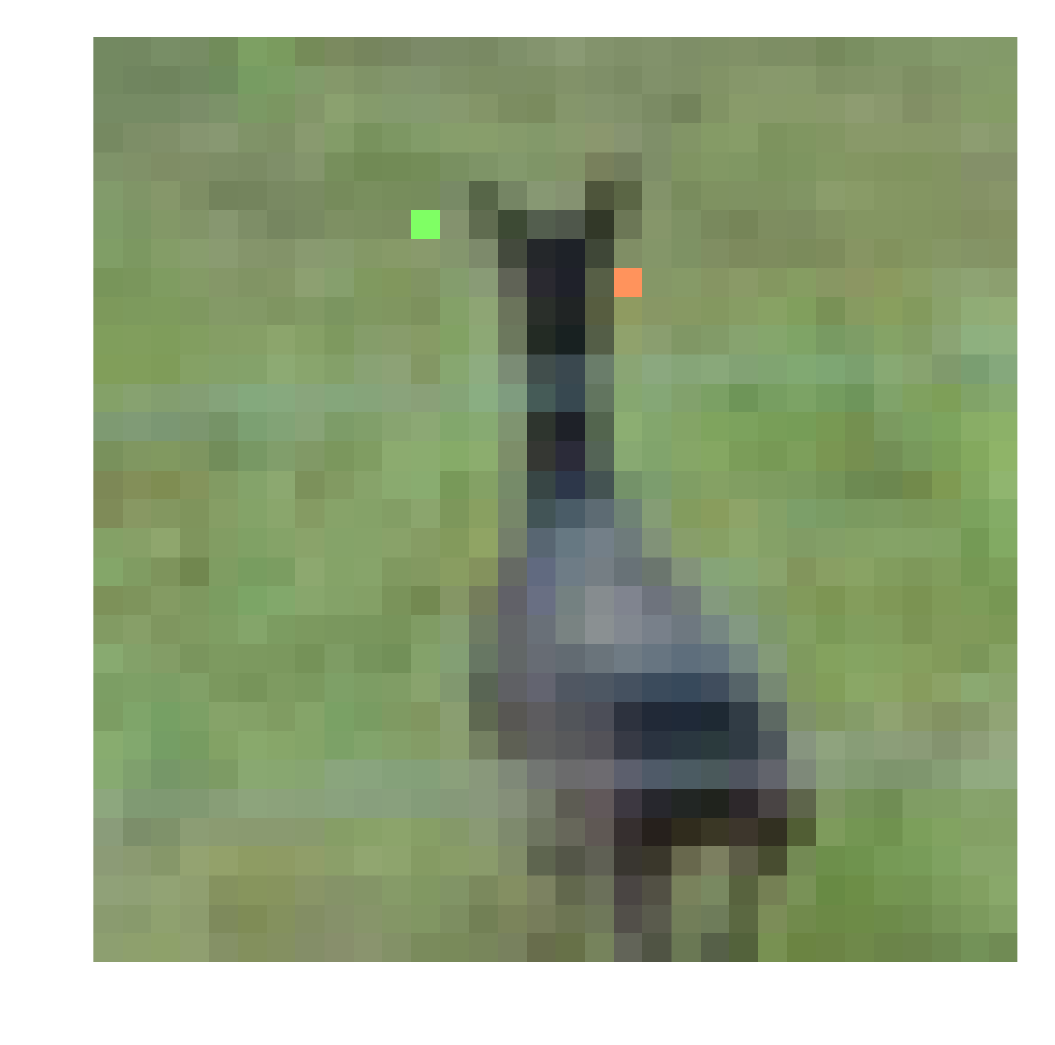}\!
    \caption*{deer (2)}
    \end{subfigure}
    \begin{subfigure}[b]{0.1\linewidth}
    \includegraphics[width=\linewidth]{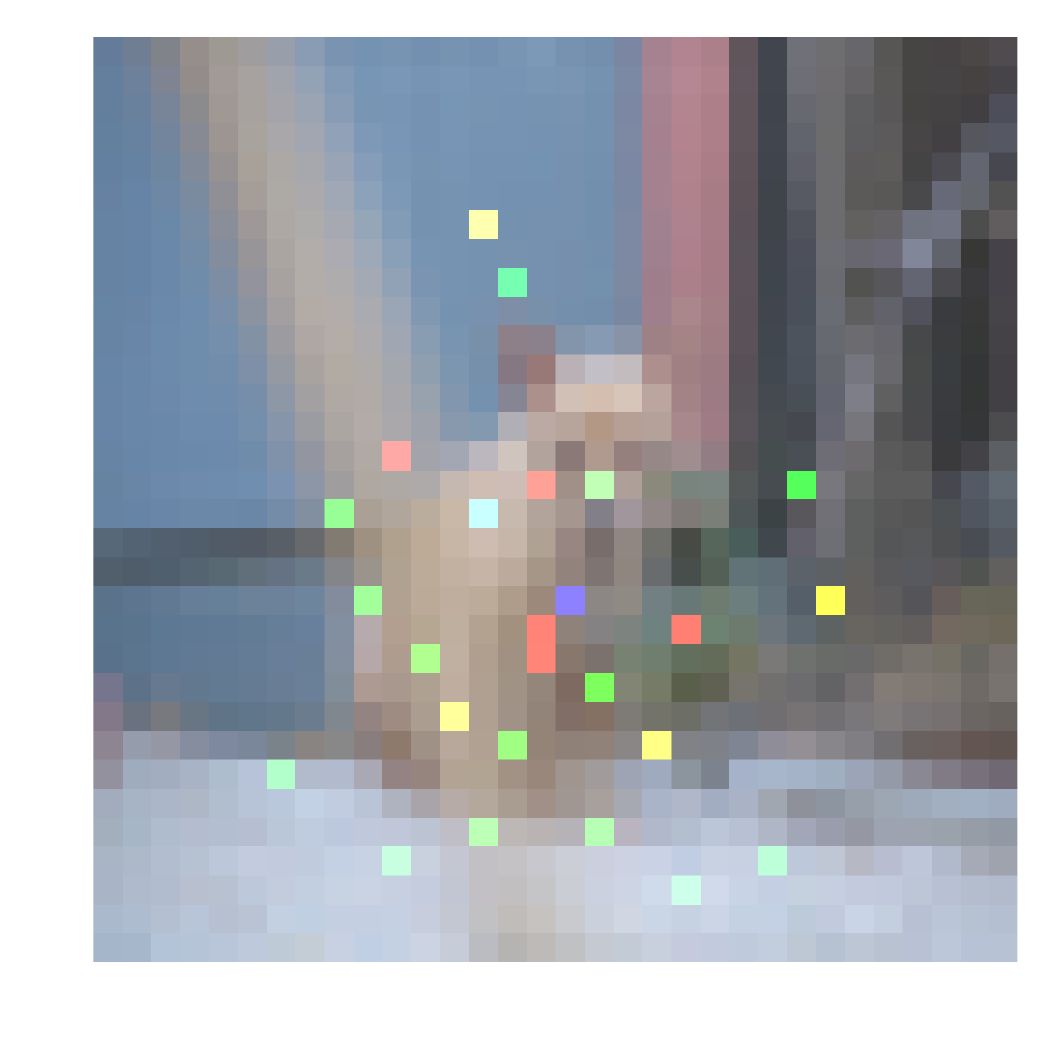}\!
    \caption*{truck (25)}
    \end{subfigure}
    \begin{subfigure}[b]{0.1\linewidth}
    \includegraphics[width=\linewidth]{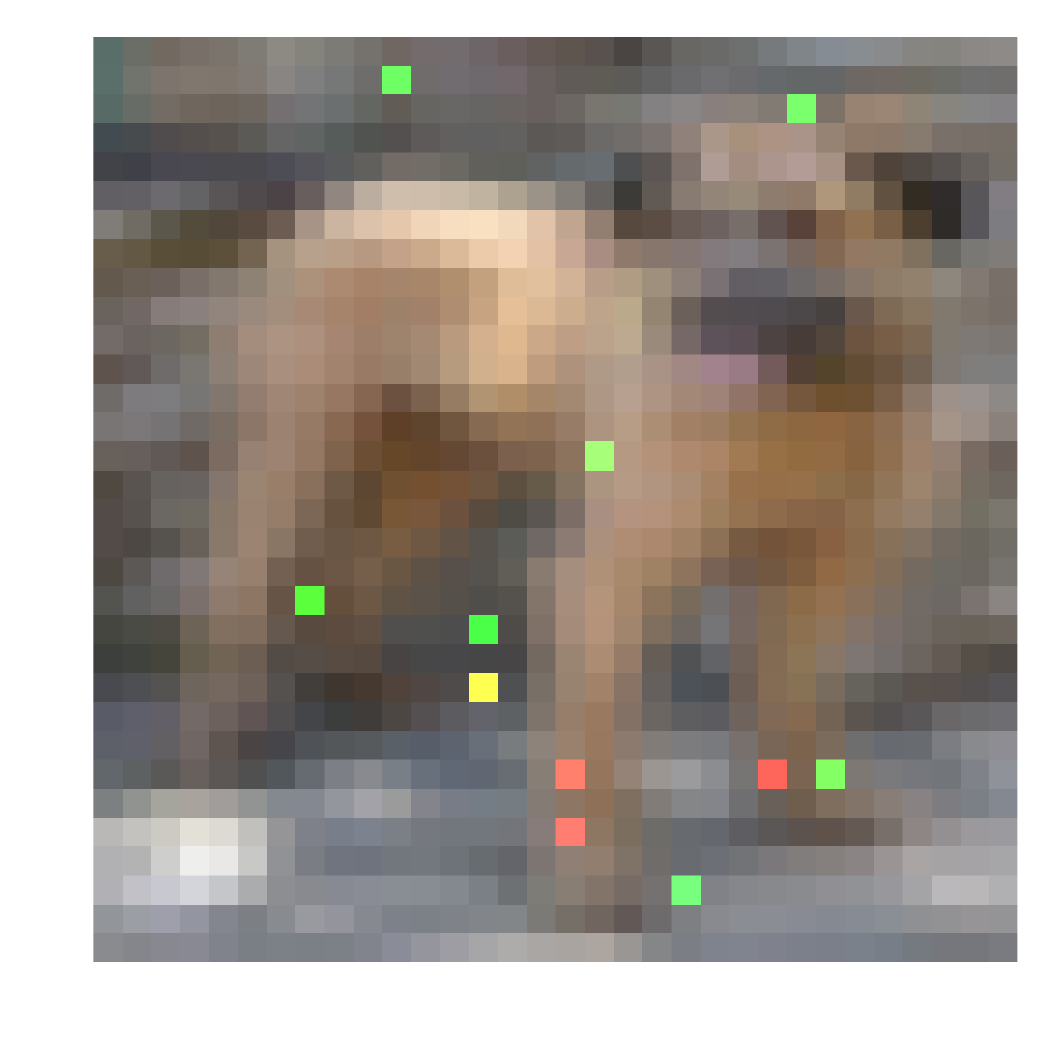}\!
    \caption*{deer (11)}
    \end{subfigure}
    \begin{subfigure}[b]{0.1\linewidth}
    \includegraphics[width=\linewidth]{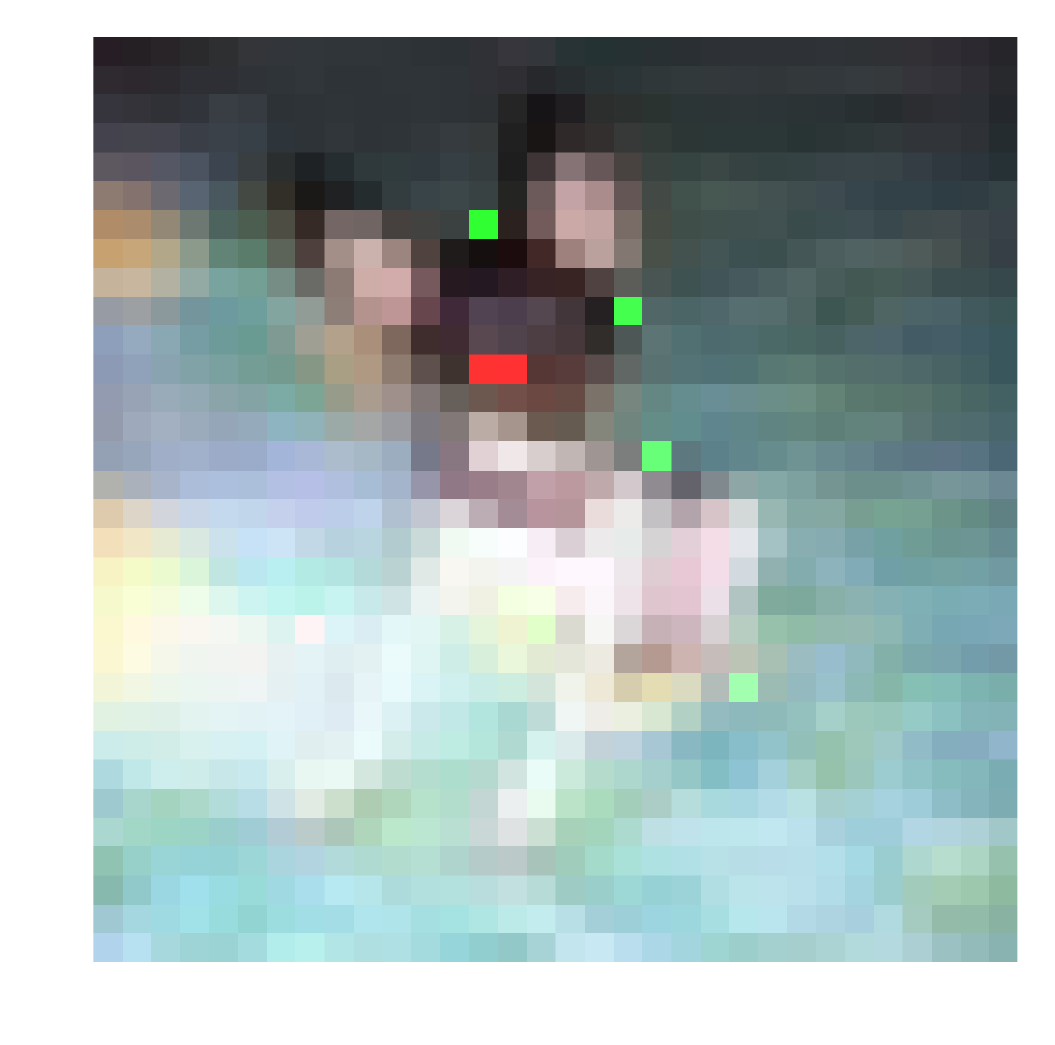}\!
    \caption*{deer (10)}
    \end{subfigure}
    \begin{subfigure}[b]{0.1\linewidth}
    \includegraphics[width=\linewidth]{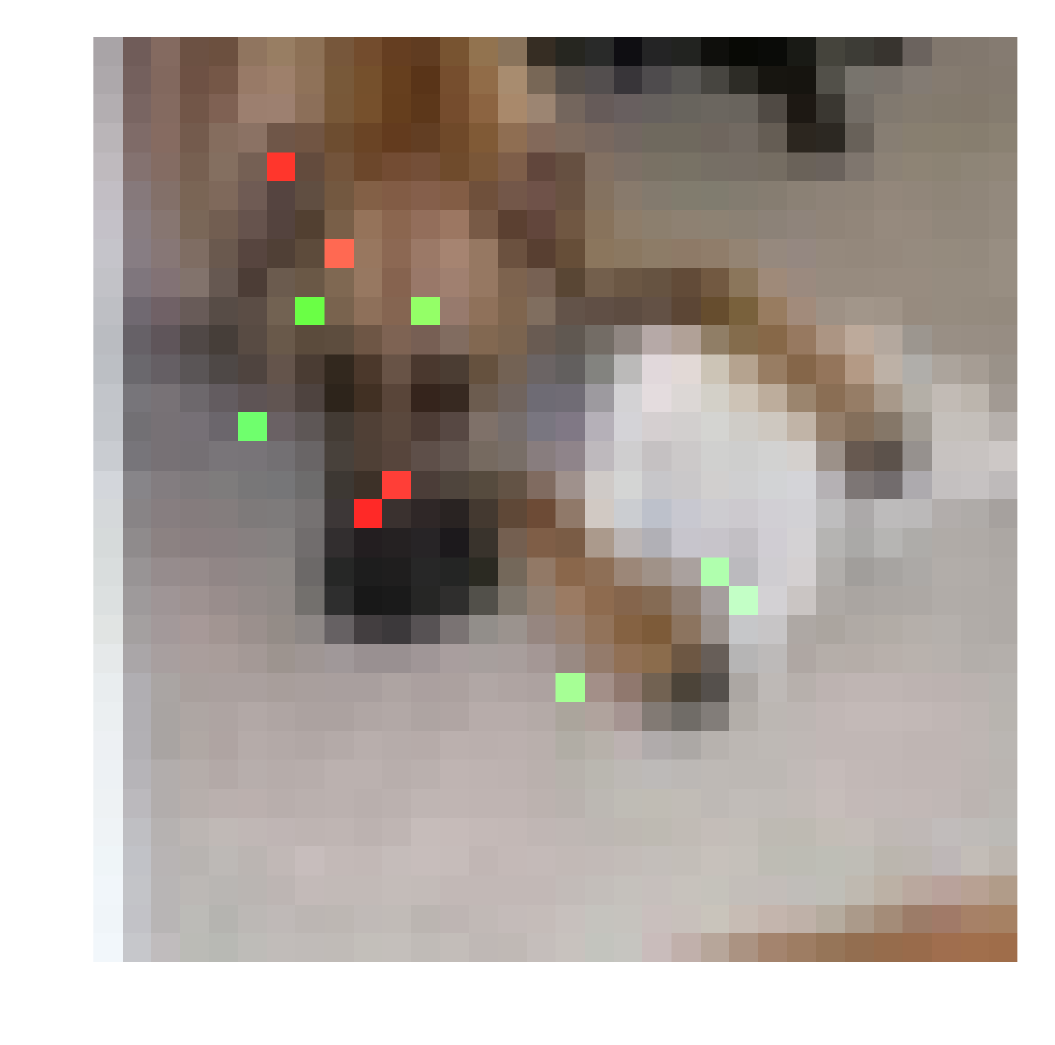}\!
    \caption*{cat (10)}
    \end{subfigure}
    \begin{subfigure}[b]{0.1\linewidth}
    \includegraphics[width=\linewidth]{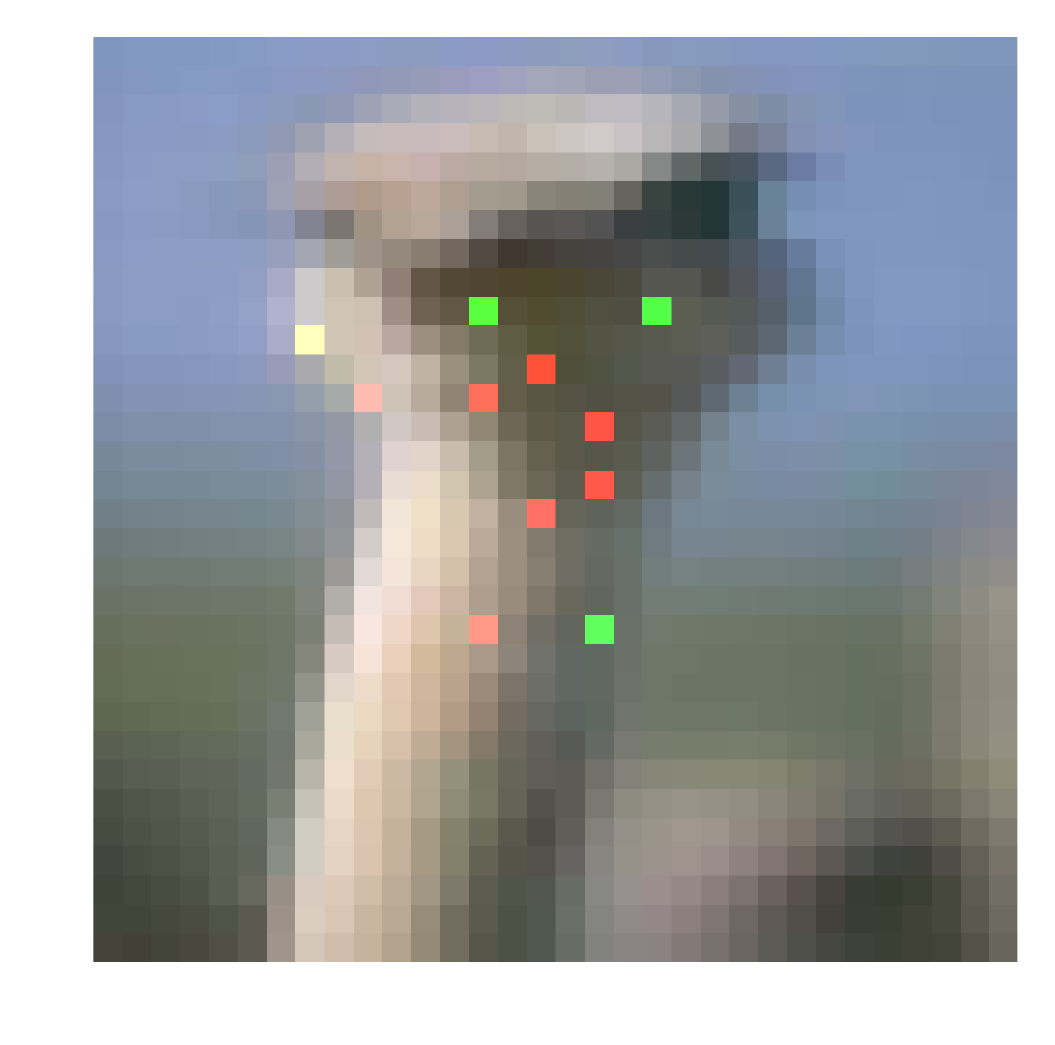}\!
    \caption*{deer (11)}
    \end{subfigure}
    \begin{subfigure}[b]{0.1\linewidth}
    \includegraphics[width=\linewidth]{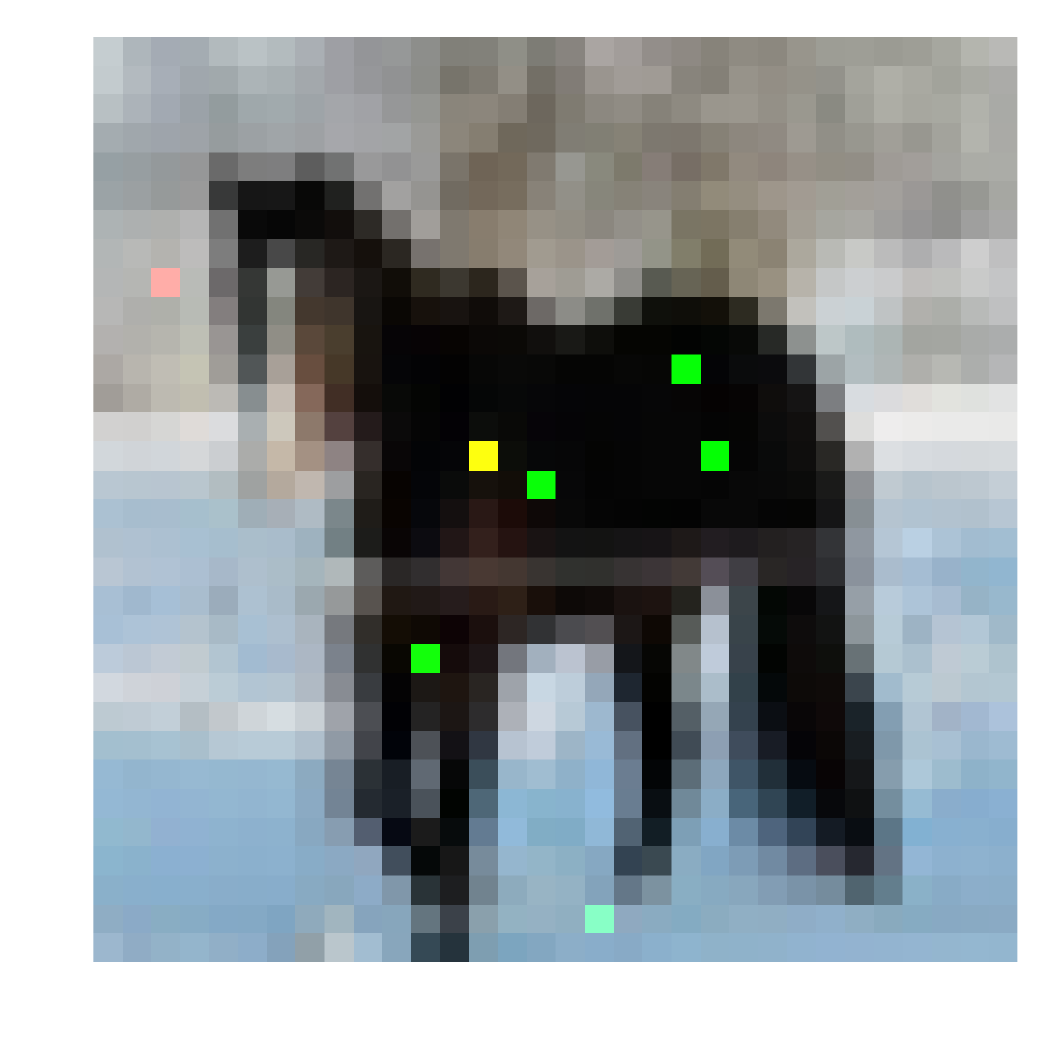}\!
    \caption*{cat (7)}
    \end{subfigure}
    
\caption{CIFAR-10 adversarial examples generated by (a) SparseFool (first row), (b) ``One pixel attack" (second row), and (c) JSMA (third row). The fooling label is shown below each image, and the number of perturbed pixels is written inside the parentheses.}
\label{fig:compare_cifar}
\end{figure}

In this section we illustrate adversarial examples generated by SparseFool, compared to the corresponding ones computed by JSMA and ``One-pixel attack". The results obtained for the MNIST and CIFAR-10 datasets are depicted in Figure~\ref{fig:compare_mnist} and Figure~\ref{fig:compare_cifar} respectively.

\end{appendices}

\end{document}